\documentclass[11pt, 
english, 
singlespacing, 
parskip, 
headsepline, 
]{DoctoralThesis} 

\usepackage[utf8]{inputenc} 
\usepackage[T1]{fontenc} 
\usepackage{mathpazo} 
\usepackage[autostyle=true]{csquotes} 
\usepackage[square,sort,comma,numbers]{natbib}

\usepackage{adjustbox}
\usepackage{amsmath}
\usepackage{amssymb}
\usepackage{bm}
\usepackage{comment}
\usepackage{enumitem}
\usepackage{empheq}
\usepackage{float}
\usepackage{graphicx}
\usepackage{mathtools}
\usepackage{microtype}
\usepackage[undotted]{minitoc}
\usepackage{multirow,makecell}
\usepackage{multicol}
\usepackage{pifont}
\usepackage[percent]{overpic}
\usepackage{subfigure}
\usepackage{wrapfig,lipsum}
\usepackage{xspace}
\usepackage{xfrac}
\usepackage{url}
\usepackage[pagebackref=true,bookmarks=false,hypertexnames=true]{hyperref}

\let\olddegree\degree  
\let\degree\relax      
\usepackage{gensymb}   
\let\degree\olddegree  

\makeatletter
\DeclareRobustCommand\onedot{\futurelet\@let@token\@onedot}
\def\@onedot{\ifx\@let@token.\else.\null\fi\xspace}
\def\eg{\emph{e.g}\onedot} 
\def\ie{\emph{i.e}\onedot} 
\def\etal{\emph{et al}\onedot}
\patchcmd\Hy@EveryPageBoxHook{\Hy@EveryPageAnchor}{\Hy@hypertexnamestrue\Hy@EveryPageAnchor}{}{\fail} 
\makeatother

\definecolor{darkgreen}{RGB}{0, 150, 0}
\definecolor{darkred}{RGB}{200, 0, 0}
\definecolor{darkblue}{RGB}{0, 0, 200}

\newcommand{\ch}{{\color{darkgreen} \ding{51}}}
\newcommand{\xm}{{\color{darkred} \ding{55}}}
\newcommand{\xmb}{{\color{darkblue} \ding{55}}}
\newcommand{\nap}{{\color{gray} NA}}
\newcommand{\lzero}{$L_0$\xspace}
\newcommand{\lone}{$L_1$\xspace}
\newcommand{\ltwo}{$L_2$\xspace}
\newcommand{\linfinity}{$L_\infty$\xspace}
\newcommand{\iPi}{\Pi^{-1}}
\newcommand{\lossfun}[1]{\rho\left( #1 \right)}
\newcommand{\partition}[1]{Z\left( #1 \right)}
\newcommand{\prob}[1]{p\left( #1 \right)}
\DeclarePairedDelimiter{\abs}{\lvert}{\rvert}
\newcommand{\matr}[1]{#1}                 
\renewcommand{\vec}[1]{\mathbf{#1}}         
\newcommand{\unit}[1]{\hat{#1}}
\newcommand{\unitv}[1]{\mathbf{\unit{#1}}}  
\newcommand{\spce}[1]{\mathbb{#1}}

\newcommand{\T}{\mathsf{T}}  
\newcommand{\atan}{\text{atan}}
\newcommand{\acos}{\text{acos}}

\thesistitle{Surround-View Cameras based \\ Holistic Visual Perception \\ for Automated Driving}
\supervisor{Prof. Dr.-Ing. Patrick Mäder} 
\examiner{} 
\degree{\textbf{Doctor of Engineering}}
\author{Dr.-Ing. Varun Ravi Kumar}
\addresses{} 
\subject{Computer Science}
\university{\href{https://www.tu-ilmenau.de/en/international/}{TECHNISCHE UNIVERSITÄT ILMENAU}} 
\department{\href{https://www.tu-ilmenau.de/secsy/}{Softwaretechnik für sicherheitskritische Systeme}}
\group{\href{https://informatik.tu-ilmenau.de/}{Fakultät für Informatik und Automatisierung}}
\faculty{\href{http://faculty.university.com}{Faculty Name}}

\AtBeginDocument{
\hypersetup{pdftitle=\ttitle} 
\hypersetup{pdfauthor=\authorname} 
\hypersetup{pdfkeywords=\keywordnames} 
}

\begin{document}
\frontmatter 
\pagestyle{plain} 
\begin{titlepage}
\begin{center}

\includegraphics[height=3cm]{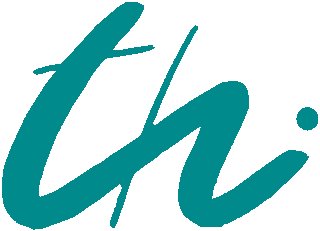}

\vspace*{.04\textheight}
{\scshape\LARGE \univname
\par}\vspace{0.5cm} 
\textsc{\Large Doctoral Thesis}\\[0.5cm] 

\HRule \\[0.3cm] 
{\huge 
\bfseries 
\ttitle
\par}
\vspace{0.3cm} 
\HRule \\[1.5cm] 

\vspace{-2em}
\begin{minipage}[t]{0.5\textwidth}
\begin{flushleft} \large
\emph{Reviewers:} \\
\href{https://scholar.google.com.ar/citations?user=LPzoSrsAAAAJ&hl=en}{\supname}\\
\href{https://scholar.google.com.ar/citations?user=MrI1EV4AAAAJ&hl=en}{Prof. Dr. John Mc Donald} \\
\href{https://www.researchgate.net/profile/Gunther-Notni}{Prof. Dr. rer. nat. Gunther Notni}
\end{flushleft}
\end{minipage}
\begin{minipage}[t]{0.4\textwidth}
\begin{flushright} \large
\emph{Author:}\\
\href{https://linkedin.com/in/rvarun7777}{\authorname}
\end{flushright}
\end{minipage}\\[3cm]

\vspace{-4em}
\includegraphics[height=2cm]{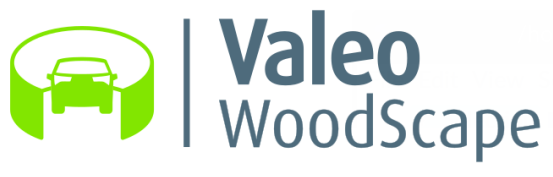}

\large 
\textit{A dissertation submitted in partial fulfillment \\ of the requirement for the degree of \\ \degreename}\\[0.3cm] 
\textit{in the}\\[0.2cm]
\groupname\\\deptname\\[0.5cm] 
\vspace{1cm}

\begin{minipage}[t]{0.6\textwidth}
\begin{flushleft} \large
DOI: 10.22032/dbt.50751 \\
URN: urn:nbn:de:gbv:ilm1-2021000421
\end{flushleft}
\end{minipage}
\begin{minipage}[t]{0.3\textwidth}
\begin{flushright} \large
\emph{Scientific Debate Day:} \\
{\large December 21, 2021}\\[4cm] 
\end{flushright}
\end{minipage}\\[3cm]

\end{center}
\end{titlepage}
\begin{declaration}
\addchaptertocentry{\authorshipname} 
\noindent I, \authorname, declare that this thesis titled, \enquote{\textbf{Multi-Task Near-Field Perception for Autonomous Driving Using Surround-View Fisheye Cameras}} and the work presented in it are my own. I confirm that:

I certify that I prepared the submitted thesis independently without undue assistance
of a third party and without the use of other than the indicated aids. Data and concepts
directly or indirectly taken over from other sources have been marked stating the
sources.

When selecting and evaluating the following materials, the persons listed below
helped me in the way described respectively for a charge/free of charge:

\begin{itemize}[nosep]
\item Senthil Yogamani
\item Prof. Dr. John McDonald
\item Marvin Klinger
\end{itemize}

Further persons were not involved in the content-material-related preparation of the thesis submitted. In particular, I have not used the assistance against payment offered by consultancies or placing services (doctoral consultants or other persons). I did not pay any money to persons directly or indirectly for work or services which are related to the content of the thesis submitted.

So far, the thesis has not been submitted identically or similarly to an examination office in Germany or abroad.

I have been notified that any incorrectness in the submitted declaration as mentioned above is assessed as an attempt to deceive and, according to § 7 para. 10 of the PhD regulations, this leads to a discontinuation of the doctoral procedure.\\ \\
 
\noindent Signed:\\
\rule[0.5em]{25em}{0.5pt} 
 
\noindent City, Date:\\
\rule[0.5em]{25em}{0.5pt} 
\end{declaration}

\cleardoublepage
\vspace*{0.2\textheight}
\noindent\enquote{\itshape You should not give up and we should not allow the problem to defeat us. To succeed in your mission, you must have single-minded devotion to your goal.}\bigbreak
\hfill Dr. A.P.J. Abdul Kalam
\dedicatory{Dedicated to my parents Ravi Kumar and Vinutha, who have worked hard and sacrificed their lives for a better future for me and offered unconditional love support during my tough times. Thank you so much for everything!}
\begin{abstract}
\addchaptertocentry{Abstract Deutsch} 

Die Bildung der Augen führte zum Urknall der Evolution. Die Dynamik änderte sich von einem primitiven Organismus, der auf den Kontakt mit der Nahrung wartete, zu einem Organismus, der durch visuelle Sensoren gesucht wurde. Das menschliche Auge ist eine der raffiniertesten Entwicklungen der Evolution, aber es hat immer noch Mängel. Der Mensch hat über Millionen von Jahren einen biologischen Wahrnehmungsalgorithmus entwickelt, der in der Lage ist, Autos zu fahren, Maschinen zu bedienen, Flugzeuge zu steuern und Schiffe zu navigieren. Die Automatisierung dieser Fähigkeiten für Computer ist entscheidend für verschiedene Anwendungen, darunter selbstfahrende Autos, Augmented Realität und architektonische Vermessung.

Die visuelle Nahfeldwahrnehmung im Kontext von selbstfahrenden Autos kann die Umgebung in einem Bereich von $0-10$ Metern und $360\degree$ Abdeckung um das Fahrzeug herum wahrnehmen. Sie ist eine entscheidende Entscheidungskomponente bei der Entwicklung eines sichereren automatisierten Fahrens. Jüngste Fortschritte im Bereich Computer Vision und Deep Learning in Verbindung mit hochwertigen Sensoren wie Kameras und LiDARs haben ausgereifte Lösungen für die visuelle Wahrnehmung hervorgebracht. Bisher stand die Fernfeldwahrnehmung im Vordergrund. Ein weiteres wichtiges Problem ist die begrenzte Rechenleistung, die für die Entwicklung von Echtzeit-Anwendungen zur Verfügung steht. Aufgrund dieses Engpasses kommt es häufig zu einem Kompromiss zwischen Leistung und Laufzeiteffizienz.

Wir konzentrieren uns auf die folgenden Themen, um diese anzugehen: 1) Entwicklung von Nahfeld-Wahrnehmungsalgorithmen mit hoher Leistung und geringer Rechenkomplexität für verschiedene visuelle Wahrnehmungsaufgaben wie \\ geometrische und semantische Aufgaben unter Verwendung von faltbaren neuronalen Netzen. 2) Verwendung von Multi-Task-Learning zur Überwindung von Rechenengpässen durch die gemeinsame Nutzung von initialen Faltungsschichten zwischen den Aufgaben und die Entwicklung von Optimierungsstrategien, die die Aufgaben ausbalancieren.

\end{abstract}
\begin{abstract}
\addchaptertocentry{\abstractname} 

The formation of eyes led to the big bang of evolution. The dynamics changed from a primitive organism waiting for the food to come into contact for eating food being sought after by visual sensors. The human eye is one of the most sophisticated developments of evolution, but it still has defects. Humans have evolved a biological perception algorithm capable of driving cars, operating machinery, piloting aircraft, and navigating ships over millions of years. Automating these capabilities for computers is critical for various applications, including self-driving cars, augmented reality, and architectural surveying.

Near-field visual perception in the context of self-driving cars can perceive the environment in a range of $0-10$ meters and $360\degree$ coverage around the vehicle. It is a critical decision-making component in the development of safer automated driving. Recent advances in computer vision and deep learning, in conjunction with high-quality sensors such as cameras and LiDARs, have fueled mature visual perception solutions. Until now, far-field perception has been the primary focus. Another significant issue is the limited processing power available for developing real-time applications. Because of this bottleneck, there is frequently a trade-off between performance and run-time efficiency.

We concentrate on the following issues in order to address them: 1) Developing near-field perception algorithms with high performance and low computational complexity for various visual perception tasks such as geometric and semantic tasks using convolutional neural networks. 2) Using Multi-Task Learning to overcome computational bottlenecks by sharing initial convolutional layers between tasks and developing optimization strategies that balance tasks.

\end{abstract}
\begin{publications}
\addchaptertocentry{\publicationname}

The research work presented in this thesis appears in the following publications, and the summarized contributions are solely from me. The following papers' chronology has been sorted on a top-down approach and descending order of the publication year.
\section*{Primary Authorship}
\begin{enumerate}
    \item Varun Ravi Kumar, Senthil Yogmani, Hazem Rashed, Ganesh Sitsu, Christian Witt, Isabelle Leang, Stefan Milz, Patrick Mäder.
    \href{https://arxiv.org/abs/2102.07448}{\enquote{\textit{\textbf{OmniDet: Surround View Cameras based Multi-task Visual Perception Network for Autonomous Driving.}}}} 
    In the IEEE Robotics and Automation Letters (RA-L) + IEEE International Conference on Robotics and Automation (ICRA), 2021 \cite{kumar2021omnidet}.\par
    
    \textit{Google Site:} \url{https://sites.google.com/view/omnidet/} \par
    
    \textbf{Contribution}:
    \begin{itemize}[nosep]
        \item Architect and lead developer of the first real-time six-task model for surround-view fisheye camera perception, out of which five were developed solely by me. For object detection, my contribution includes Open Neural Network Exchange (ONNX) model creation, setting up the training pipeline, and introducing a synergy between semantic segmentation and object detection.
        \item Designed the multi-task learning framework in PyTorch with various state-of-the-art multi-task loss functions. It potentially replaced the current MTL framework in Tensorflow developed by many developers worldwide in Valeo, and the new framework helped to win next-generation projects.
        \item Worked on creating the WoodScape dataset, including writing parsers and data loaders for each task. The dataset and the corresponding parsers will be released to the public on \href{https://github.com/valeoai/WoodScape}{Github} and will be part of the long-term maintenance.
        \item To encourage further research in developing multi-task perception algorithms, the code was made public on the \href{https://github.com/valeoai/WoodScape}{Github}, which proved to be quite popular in the vision community and significantly helped the discernibility of the approach.
        \item Integrated my novel approaches of distance estimation on fisheye camera images which are discussed in papers~\cite{kumar2020fisheyedistancenet, kumar2020unrectdepthnet, kumar2020syndistnet, kumar2021svdistnet, kumar2021fisheyedistancenet++}.
        \item Modelled real-time capable, efficient network-architectures for semantic segmentation, motion segmentation, and soiling segmentation.
        \item Performed embedded integration on NVIDIA's Jetson AGX with models exported to the ONNX format.
    \end{itemize}
    \item Varun Ravi Kumar, Ciar\'{a}n Eising, , Christian Witt, Senthil Yogamani, Patrick Mäder. \enquote{\textit{\textbf{Surround-view Fisheye Camera Perception for Automated Driving: Overview, Survey \& Challenges}}} 
    In-Review of the Journal of IEEE Transactions on Intelligent Transportation Systems, 2021.\par
     
    \textbf{Contribution}:
    \begin{itemize}[nosep]
        \item Summarized the perception tasks carried out in the thesis.
        \item Worked on describing the most commonly used fisheye camera projection models.
    \end{itemize}
    \item Varun Ravi Kumar, Marvin Klingner, Senthil Yogamani, Stefan Milz, Tim Fingscheidt, Patrick Mäder.  \href{https://arxiv.org/abs/2008.04017}{\enquote{\textit{\textbf{SynDistnet: Self-Supervised Monocular Fisheye Camera Distance Estimation Synergized with Semantic Segmentation for Autonomous Driving.}}}} 
    In the Proceedings of the IEEE/CVF Winter Conference on Applications of Computer Vision (WACV), 2021 \cite{kumar2020syndistnet}.\par
    
    \textit{Oral Talk}: \url{https://youtu.be/zL6zvtUy4cc} \\
    \textit{Google Site}: \url{https://sites.google.com/view/syndistnet/} \par
    
    \textbf{Contribution}:
    \begin{itemize}[nosep]
        \item Solely developed a novel architecture to learn self-supervised distance estimation synergized with semantic segmentation and self-attention modules.
        \item Research collaboration with Technische Universität Braunschweig to come up with a novel dynamic object filtering technique. Had fruitful discussions with Marvin Klinger (Ph.D. student) on depth estimation. The code and integration were carried out solely by me as Valeo's policy does not allow code sharing.
        \item As a second contribution compared to PackNet~\cite{guizilini2019packnet}, I came up with a one-stage training approach to filter dynamic objects by introducing synergy between distance estimation and semantic segmentation.
    \end{itemize}
    \item Varun Ravi Kumar, Marvin Klingner, Senthil Yogamani, Markus, Bach, Stefan Milz, Tim Fingscheidt, Patrick Mäder. \href{https://arxiv.org/abs/2104.04420}{\enquote{\textit{\textbf{SVDistNet: Self-Supervised Near-Field Distance Estimation on Surround View Fisheye Cameras.}}}} In the Journal of IEEE Transactions on Intelligent Transportation Systems. \par
    \textit{Google Site:} \url{https://sites.google.com/view/svdistnet/} \par
    
    \textbf{Contribution}:
    \begin{itemize}[nosep]
        \item Designed and developed a novel camera geometry adaptive multi-scale convolution for fisheye camera models to enable the training and deployment of distance estimation models on multiple surround-view cameras mounted on different viewpoints.
        \item With this design, I proposed a solution for surround-view fisheye cameras targeting large-scale industrial deployment. In other words, Valeo can target the design of a model for production that can be deployed in millions of vehicles having its own set of cameras.
        \item Demonstrated and showcased a single trained model for 12 fisheye cameras, which achieves the equivalent result as an individual specialized model that overfits to a particular camera model.
        \item Integrated and improved vector-based self-attention network module \cite{zhao2020exploring} compared to my previous work~\cite{kumar2020syndistnet} into the framework.
    \end{itemize}
    \item Varun Ravi Kumar, Senthil Yogamani, Stefan Milz, Patrick Mäder.  \href{https://arxiv.org/abs/2008.04017}{\enquote{\textit{\textbf{FisheyeDistanceNet++: Self-Supervised Scale-Aware Distance Estimation with Improved Training Speed and Accuracy through Loss Function Optimization.}}}} 
    In the Electronic Imaging Autonomous Vehicles and Machines (EI-AVM), 2021 \cite{kumar2021fisheyedistancenet++}.\par
    
    \textbf{Contribution}: 
    \begin{itemize}[nosep]
        \item As an incremental paper to FisheyeDistanceNet~\cite{kumar2020fisheyedistancenet} depicted the importance of using a general robust loss function instead of the \lone loss. I carried out the development of code and experiments.
        \item Performed extensive ablation studies on the impact of using normalization techniques on the results and designed optimal network architecture.
    \end{itemize}
    \item Varun Ravi Kumar, Sandesh Athni Hiremath, Markus Bach, Stefan Milz, Christian Witt, Clément Pinard, Senthil Yogamani, Patrick Mäder. 
    \href{https://arxiv.org/abs/1910.04076}{\enquote{\textit{\textbf{FisheyeDistanceNet: Self-Supervised Scale-Aware Distance Estimation using Monocular Fisheye Camera for Autonomous Driving.}}}} 
    In the IEEE International Conference on Robotics and Automation (ICRA), 2020 \cite{kumar2020fisheyedistancenet}. \par
    
    \textit{Oral Talk:} \url{https://youtu.be/qAsdpHP5e8c} \\
    \textit{Google Site:} \url{https://sites.google.com/view/fisheyedistancenet/} \par
    
    \textbf{Contribution}:
    \begin{itemize}[nosep]
        \item Formulated the novel self-supervised training strategy to infer a distance map from a sequence of distorted and unrectified raw fisheye camera images.
        \item With the research guidance from Clément Pinard, I came up with a novel solution to the scale factor uncertainty with the bolster from ego-motion velocity that allows outputting metric distance maps. This facilitates the map's practical use for self-driving cars and validates the training of the CNN. This contribution is considered the heart of the entire thesis, as this is vital to enable the training of a CNN to estimate metric distance on raw fisheye camera images. A significant idea that laid a foundation for the following research in this domain.
        \item Created a subset of the Woodscape dataset for distance estimation task, including recording the raw scenes and their post-processing. Wrote parsers and transformation scripts to obtain the ground-truth LiDAR samples to validate the network's estimates. 
        \item Incorporated my novel idea from~\cite{kumar2018near, kumar2018monocular} to solve the problem of occlusion due to different mounting positions of LiDAR and camera.
        \item Solely designed and developed novel ideas and network architecture to obtain state-of-the-art results on the KITTI~\cite{geiger2012we} and the WoodScape~\cite{yogamani2019woodscape} dataset.
    \end{itemize}
    \item Varun Ravi Kumar, Senthil Yogamani, Markus Bach, Christian Witt, Stefan Milz, Patrick Mäder.  \href{https://arxiv.org/abs/2007.06676}{\enquote{\textit{\textbf{UnRectDepthnet: Self-supervised Monocular Depth Estimation using a Generic Framework for Handling Common Camera Distortion Models.}}}} 
    In the International Conference on Intelligent Robots and Systems (IROS), 2020 \cite{kumar2020unrectdepthnet}.\par
    
    \textit{Oral Talk:} \url{https://youtu.be/3Br2KSWZRrY} \\
    \textit{Google Site:} \url{https://sites.google.com/view/unrectdepthnet/} \par
    
    \textbf{Contribution}:
    \begin{itemize}[nosep]
        \item Following upon the success of~\cite{kumar2020fisheyedistancenet}, created a novel generic end-to-end self-supervised training pipeline to estimate monocular depth maps on raw distorted images for various camera models.
        \item Demonstrated the first results of depth estimation directly on unrectified KITTI sequences and achieved state-of-the-art results on the KITTI depth estimation dataset among self-supervised methods.
    \end{itemize}
    \item Varun Ravi Kumar, Stefan Milz, Christian Witt, Martin Simon, Karl Amende, Johannes Petzold, Senthil Yogamani, Timo Pech. 
    \href{https://deepvision.data61.csiro.au/papers/5.pdf}{\enquote{\textit{\textbf{Near-field Depth Estimation using Monocular fisheye camera: A Semi-Supervised Learning Approach using Sparse LiDAR Data.}}}} 
    In the IEEE Conference on Computer Vision and Pattern Recognition Workshops (CVPRW), 2018 \cite{kumar2018near}.\par
    
    \textbf{Contribution}:
    \begin{itemize}[nosep]
        \item Built a joint fisheye camera and LiDAR dataset for supervised training of distance estimation.
        \item Adapted the training data to handle occlusions due to differences between camera and LiDAR viewpoint with a novel occlusion correction algorithm.
    \end{itemize}
    \item Varun Ravi Kumar, Stefan Milz, Christian Witt, Martin Simon, Karl Amende, Johannes Petzold, Senthil Yogamani, Timo Pech. 
    \href{https://arxiv.org/abs/1803.06192}{\enquote{\textit{\textbf{Monocular Fisheye Camera Depth Estimation using Sparse LiDAR Supervision.}}}} 
    In the 21st International Conference on Intelligent Transportation Systems (ITSC), 2018 \cite{kumar2018monocular}.\par
    
    \textit{Google Site:} \url{https://sites.google.com/view/FisheyeLiDARDepth/} \par
    
    \textbf{Contribution}:
    \begin{itemize}[nosep]
        \item Developed the first preliminary supervised approach work before the self-supervised FisheyeDistanceNet~\cite{kumar2020fisheyedistancenet}, to demonstrate fisheye camera distance estimation using CNN.
        \item Demonstrated a working prototype purely trained on sparse Velodyne LiDAR data.
        \item Tailored the loss function and training algorithm to handle sparse-depth data.
    \end{itemize}
\end{enumerate}
\section*{Secondary Authorship}
\begin{enumerate}
    \item Michal U\v{r}i\v{c}\'{a}\v{r}, Ganesh Sistu, Hazem Rashed, Anton\'{i}n Vobeck\'{y}, Varun Ravi Kumar, Pavel K\v{r}\'{i}\v{z}ek, Fabian B\"{u}rger, Senthil Yogamani. 
    \href{https://openaccess.thecvf.com/content/WACV2021/html/Uricar_Lets_Get_Dirty_GAN_Based_Data_Augmentation_for_Camera_Lens_WACV_2021_paper.html}{\enquote{\textit{\textbf{Let's Get Dirty: GAN Based Data Augmentation for Camera Lens Soiling Detection in Autonomous Driving.}}}} 
    In the Proceedings of the IEEE/CVF Winter Conference on Applications of Computer Vision (WACV), 2021 \cite{uricar2021let}. \par
    
    \textbf{Contribution}:
    \begin{itemize}[nosep]
        \item Part of the research discussions and reviewed the code and oral presentation.
        \item Supported the dataset integration into the model.
    \end{itemize}
    \item Hazem Rashed, Eslam Mohamed, Ganesh Sistu, Varun Ravi Kumar, Ciar\'{a}n Eising, Ahmad El-Sallab, Senthil Yogamani. 
    \href{https://openaccess.thecvf.com/content/WACV2021/html/Rashed_Generalized_Object_Detection_on_Fisheye_Cameras_for_Autonomous_Driving_Dataset_WACV_2021_paper.html}{\enquote{\textit{\textbf{Generalized Object Detection on Fisheye Cameras for Autonomous Driving: Dataset, Representations and Baseline.}}}} 
    In the Proceedings of the IEEE/CVF Winter Conference on Applications of Computer Vision (WACV), 2021 \cite{rashed2021generalized}.\par
    
    \textit{Google Site:} \url{https://sites.google.com/view/fisheyeyolo/} \par
    
    \textbf{Contribution}:
    \begin{itemize}[nosep]
        \item Major contribution in writing the research paper. 
        \item Designed the encoder for object detection and was part of the development of the YOLOv3 decoder.
        \item Integrated different representations of the bounding box to the framework and trained the standard bounding box representation.
        \item Active member of the research discussions and converted the model to ONNX to run it on an automotive embedded platform.
    \end{itemize}
    \item Ibrahim Sobh, Ahmed Hamed, Varun Ravi Kumar, and Senthil Yogamani. \href{https://www.overleaf.com/project/5ffa0d4afb4014f06c7d264a}{\enquote{\textit{\textbf{Adversarial Attacks on Multi-Task Visual Perception for Autonomous Driving}}}} 
    Journal of Imaging Science and Technology (JIST), 2021 \cite{sobh2021adversarial}.\par
    
    \textbf{Contribution}:
    \begin{itemize}[nosep]
        \item Major contribution in writing the research paper and brainstorming the adversarial attacks.
        \item Provided the inference and model boiler-plate code for white-box and black-box attacks from the \textit{OmniDet} framework.
    \end{itemize}
    \item Sebastian Houben, Stephanie Abrecht, Maram Akila, Andreas B{\"{a}}r, Felix Brockherde, Patrick Feifel, Tim Fingscheidt, Sujan Sai Gannamaneni, Seyed Eghbal Ghobadi, Ahmed Hammam, Anselm Haselhoff, Felix Hauser, Christian Heinzemann, Marco Hoffmann, Nikhil Kapoor, Falk Kappel, Marvin Klingner, Jan Kronenberger, Fabian K{\"{u}}ppers, Jonas L{\"{o}}hdefink, Michael Mlynarski, Michael Mock, Firas Mualla,
    Svetlana Pavlitskaya, Maximilian Poretschkin, Alexander Pohl, Varun Ravi Kumar, Julia Rosenzweig, Matthias Rottmann, Stefan R{\"{u}}ping, Timo S{\"{a}}mann,
    Jan David Schneider, Elena Schulz, Gesina Schwalbe, Joachim Sicking, Toshika Srivastava, Serin Varghese, Michael Weber, Sebastian Wirkert, Tim Wirtz and Matthias Woehrle. \href{https://arxiv.org/abs/2104.14235}{\enquote{\textit{\textbf{Inspect, Understand, Overcome: A Survey of Practical Methods for AI Safety}}}} 
    arXiv preprint arXiv:2104.14235 \cite{kia_2021}.\par
    
    \textbf{Contribution}:
    \begin{itemize}[nosep]
        \item Contributed the summary on Multi Task Learning.
    \end{itemize}
    \item Ashok Dahal, Varun Ravi Kumar, Senthil Yogamani and Ciar\'{a}n Eising. \href{https://arxiv.org/abs/1908.11789}{\enquote{\textit{\textbf{An Online Learning System for Wireless Charging Alignment using Surround-view Fisheye Cameras.}}}} 
    In the IEEE Transactions on Intelligent Transportation Systems, 2021.\par
    
    \textbf{Contribution}:
    \begin{itemize}[nosep]
        \item Major contribution in writing the research paper and brainstorming the idea.
        \item Provided the inference and model boiler-plate code from \textit{OmniDet} framework.
    \end{itemize}
    \item Mahesh M Dhananjaya, Varun Ravi Kumar and Senthil Yogamani. \href{https://arxiv.org/pdf/2104.14042}{\enquote{\textit{\textbf{Weather and Light Level Classification for Autonomous Driving: Dataset, Baseline and Active Learning}}}} 
    In-Review of the 24th International Conference on Intelligent Transportation Systems (ITSC), 2021 \cite{dhananjaya2021weather}.\par
    
    \textbf{Contribution}:
    \begin{itemize}[nosep]
        \item Major contribution in writing the research paper and mentored the intern during his master thesis.
        \item Provided the training and model boiler-plate code for the weather classification task from the \textit{OmniDet} framework.
    \end{itemize}
    \item Ashok Dahal, Eric Golab, Rajender Garlapati, Varun Ravi Kumar, Senthil Yogamani. \href{https://arxiv.org/abs/1908.11789}{\enquote{\textit{\textbf{RoadEdgeNet: Road Edge Detection System Using Surround View Camera Images.}}}} 
    In the Electronic Imaging Autonomous Vehicles and Machines (EI-AVM), 2021 \cite{dahal2021roadedgenet}.\par
    
    \textbf{Contribution}:
    \begin{itemize}[nosep]
        \item Part of the research discussions, reviewed the code and contribution in writing the research paper.
    \end{itemize}
    \item Hazem Rashed, Eslam Mohamed, Ganesh Sistu, Varun Ravi Kumar, Ciar\'{a}n Eising, Ahmad El-Sallab, Senthil Yogamani. \href{https://ml4ad.github.io/}{\enquote{\textit{\textbf{FisheyeYOLO: Object Detection on Fisheye Cameras for Autonomous Driving.}}}} 
    In the NeurIPS Workshop on Machine Learning for Autonomous Driving, 2021 \cite{rashedfisheyeyolo}.\par
    
    \textbf{Contribution}:
    \begin{itemize}[nosep]
        \item Major contribution in writing the short research paper derived from~\cite{rashed2021generalized}.
    \end{itemize}
    \item Arindam Das, Pavel K\v{r}\'{i}\v{z}ek, Ganesh Sistu, Fabian B\"{u}rger, Sankaralingam \\ Madasamy, Michal U\v{r}i\v{c}\'{a}\v{r}, Varun Ravi Kumar, Senthil Yogamani. \href{https://arxiv.org/abs/2007.00801}{\enquote{\textit{\textbf{TiledSoilingNet: Tile-level Soiling Detection on Automotive Surround-view Cameras Using Coverage Metric.}}}} 
    In the IEEE 23rd International Conference on Intelligent Transportation Systems (ITSC), 2020 \cite{das2020tiledsoilingnet}.\par
    
    \textbf{Contribution}:
    \begin{itemize}[nosep]
        \item Part of the research discussions and reviewed the code and oral presentation.
    \end{itemize}
    \item Marie Yahiaoui, Hazem Rashed, Letizia Mariotti, Ganesh Sistu, Ian Clancy, Lucie Yahiaoui, Varun Ravi Kumar, Senthil Yogamani. \href{https://arxiv.org/abs/1908.11789}{\enquote{\textit{\textbf{FisheyeModNet: Moving Object Detection on Surround-View Cameras for Autonomous Driving.}}}} 
    In the International Conference on Computer Vision (ICCV) Workshop on 360° Perception and Interaction, 2019 \cite{yahiaoui2019fisheyemodnet}. \par
    
    \textbf{Contribution}:
    \begin{itemize}[nosep]
        \item Re-implemented the entire approach with novel improvements in PyTorch~\cite{paszke2017automatic} and obtained better results than the former.
        \item Part of the original approach and discussions.
    \end{itemize}
\end{enumerate}
\end{publications}
\begin{publications}
\addchaptertocentry{Publications List}
\section*{Primary Authorship}

\begin{enumerate}
    \item V. Ravi Kumar, S. Yogamani, H. Rashed, G. Sitsu, C. Witt, I. Leang, S. Milz,and P. Mäder, “OmniDet:  Surround View Cameras based Multi-task VisualPerception Network for Autonomous Driving,” in \textit{IEEE Robotics and Automation Letters (RA-L) + 2021 IEEE International Conference on Robotics and Automation (ICRA),} vol. 6, no. 2, 2021, pp. 2830–2837.
    
    \item V. Ravi Kumar, C. Witt, S. Yogamani, and P. Mäder, “Surround-View Fisheye Camera Perception for Automated Driving: Overview, Survey \& Challenges,” in \textit{Review of the Journal of IEEE Transactions on Intelligent Transportation Systems, 2021}.
    
    \item V. Ravi Kumar, M. Klingner, S. Yogamani, M. Bach, S. Milz, T. Fingscheidt,
    and P. Mäder, “SVDistNet: Self-Supervised Near-Field Distance Estimation on
    Surround View Fisheye Cameras,” \textit{IEEE Transactions on Intelligent Transportation Systems, vol. abs/2104.04420, 2021}. 
    
    \item V. Ravi Kumar, M. Klingner, S. Yogamani, S. Milz, T. Fingscheidt, and P. Mader,
    “Syndistnet: Self-supervised monocular fisheye camera distance estimation synergized with semantic segmentation for autonomous driving,” in \textit{Proceedings
    of the IEEE/CVF Winter Conference on Applications of Computer Vision, 2021}, pp.
    61–71.
    
    \item V. Ravi Kumar, S. Yogamani, S. Milz, and P. Mäder, “FisheyeDistanceNet++:
    Self-Supervised Fisheye Distance Estimation with Self-Attention, Robust Loss
    Function and Camera View Generalization,” in \textit{Electronic Imaging. Society for
    Imaging Science and Technology, 2021}.
    
    \item V. Ravi Kumar, S. A. Hiremath, M. Bach, S. Milz, C. Witt, C. Pinard, S. Yogamani, and P. Mäder, “Fisheyedistancenet: Self-supervised scale-aware distance estimation using monocular fisheye camera for autonomous driving,” in \textit{2020 IEEE International Conference on Robotics and Automation (ICRA), 2020}, pp. 574–581
    
    \item  V. Ravi Kumar, S. Yogamani, M. Bach, C. Witt, S. Milz, and P. Mäder, “UnRectDepthNet: Self-Supervised Monocular Depth Estimation using a Generic Framework for Handling Common Camera Distortion Models,” in \textit{IEEE/RSJ International Conference on Intelligent Robots and Systems, (IROS) 2020}, pp. 8177–8183.
    
    \item V. Ravi Kumar, S. Milz, C. Witt, M. Simon, K. Amende, J. Petzold, S. Yogamani,
    and T. Pech, “Monocular fisheye camera depth estimation using sparse lidar
    supervision,” in \textit{21st International Conference on Intelligent Transportation Systems (ITSC), 2018, pp. 2853–2858}.
    
    \item V. Ravi Kumar, S. Milz, C. Witt, M. Simon, K. Amende, J. Petzold, S. Yogamani,
    and T. Pech, “Near-field depth estimation using monocular fisheye camera
    A semi-supervised learning approach using sparse LiDAR data,” in \textit{CVPR
    Workshop, vol. 7, 2018}.
\end{enumerate}
\section*{Secondary Authorship}
   \begin{enumerate}  
    
    \item M. U\v{r}i\v{c}\'{a}\v{r}, G. Sistu, H. Rashed, A. Vobeck\'{y}, V. Ravi Kumar, P. K\v{r}\'{i}\v{z}ek, F. B\"{u}rger,
    and S. Yogamani, “Let’s Get Dirty: GAN Based Data Augmentation for Camera
    Lens Soiling Detection in Autonomous Driving,” in \textit{Proceedings of the IEEE/CVF
    Winter Conference on Applications of Computer Vision, 2021, pp. 766–775}.
    
    \item Gallagher, Louis and Ravi Kumar, Varun and Yogamani, Senthil and McDonald, John ,B “A Hybrid Sparse-Dense Monocular SLAM System for Autonomous Driving,” in \textit{2021 European Conference on Mobile Robots (ECMR), 2021, pp. 1–8}.
    
    \item H. Rashed, E. Mohamed, G. Sistu, V. Ravi Kumar, C. Eising, A. El-Sallab,
    and S. Yogamani, “Generalized Object Detection on Fisheye Cameras for Autonomous Driving: Dataset, Representations and Baseline,” in \textit{Proceedings of
    the IEEE/CVF Winter Conference on Applications of Computer Vision, 2021, pp.
    2272–2280.}
    
    \item I. Sobh, A. Hamed, V. Ravi Kumar, and S. Yogamani, “Adversarial Attacks on
    Multi-task Visual Perception for Autonomous Driving,” in \textit{Review of the IEEE
    24th International Conference on Intelligent Transportation Systems (ITSC)}.
    
    \item A. Dahal, V. Ravi Kumar, S. Yogamani and C. Eising. “An Online Learning System for Wireless Charging Alignment using Surround-view Fisheye Cameras.” \textit{In the IEEE Transactions on Intelligent Transportation Systems, 2021.}

    \item Sebastian Houben, Stephanie Abrecht, Maram Akila, Andreas B{\"{a}}r, Felix Brockherde, Patrick Feifel, Tim Fingscheidt, Sujan Sai Gannamaneni, Seyed Eghbal Ghobadi, Ahmed Hammam, Anselm Haselhoff, Felix Hauser, Christian Heinzemann, Marco Hoffmann, Nikhil Kapoor, Falk Kappel, Marvin Klingner, Jan Kronenberger, Fabian K{\"{u}}ppers, Jonas L{\"{o}}hdefink, Michael Mlynarski, Michael Mock, Firas Mualla,
    Svetlana Pavlitskaya, Maximilian Poretschkin, Alexander Pohl, Varun Ravi Kumar, Julia Rosenzweig, Matthias Rottmann, Stefan R{\"{u}}ping, Timo S{\"{a}}mann,
    Jan David Schneider, Elena Schulz, Gesina Schwalbe, Joachim Sicking, Toshika Srivastava, Serin Varghese, Michael Weber, Sebastian Wirkert, Tim Wirtz and Matthias Woehrle. “Inspect, Understand, Overcome: A Survey of Practical Methods for AI Safety,” \textit{arXiv preprint arXiv:2104.14235, 2021}.\par
    
    \item M. M. Dhananjaya, V. R. Kumar, and S. Yogamani, “Weather and Light Level
    Classification for Autonomous Driving: Dataset, Baseline and Active Learning,”
    in \textit{Review of the IEEE 24\textsuperscript{th} International Conference on Intelligent Transportation Systems (ITSC). IEEE, 2021}.
    
    \item A. Dahal, E. Golab, R. Garlapati, V. Ravi Kumar, and S. Yogamani, “RoadEdgeNet: Road Edge Detection System Using Surround View Camera Images,” in
    \textit{Electronic Imaging. Society for Imaging Science and Technology, 2021}.
    
    \item A. Das, P. K\v{r}\'{i}\v{z}ek, G. Sistu, F. B\"{u}rger, S. Madasamy, M. U\v{r}i\v{c}\'{a}\v{r}, V. Ravi Kumar,
    and S. Yogamani, “TiledSoilingNet: Tile-level Soiling Detection on Automotive
    Surround-view Cameras Using Coverage Metric,” in \textit{IEEE 23\textsuperscript{rd} International Conference on Intelligent Transportation Systems (ITSC). IEEE, 2020, pp.1–6}.
    
    \item H. Rashed, E. Mohamed, G. Sistu, V. Ravi Kumar, C. Eising, A. El-Sallab, and
    S. Yogamani, “FisheyeYOLO: Object Detection on Fisheye Cameras for Autonomous Driving,” \textit{Machine Learning for Autonomous Driving NeurIPS 2020 Virtual Workshop, 2020}.
    
    \item M. Yahiaoui, H. Rashed, L. Mariotti, G. Sistu, I. Clancy, L. Yahiaoui, V. Ravi Kumar, and S. Yogamani, “FisheyeModNet: Moving object detection on Surround-View Cameras for Autonomous Driving,” \textit{arXiv preprint arXiv:1908.11789, 2019}.

\end{enumerate}
\end{publications}
\begin{acknowledgements}
\addchaptertocentry{\acknowledgementname}

First and foremost, I would like to express my sincere gratitude to my Ph.D. advisor \textit{Prof. Dr.-Ing. Patrick Mäder} for the continuous support of my Ph.D. study and research, for his patience, motivation, enthusiasm, and immense knowledge.\par

I am very grateful to \textit{Senthil Yogamani} from Valeo, Ireland, who ensured that all my work became publications. I am so lucky to get to work with someone who inspires me every day. My sincere thanks to him for his guidance and leadership. His advice helped me in research, writing this thesis, and research papers. I could not have imagined having a better mentor for my thesis study who believed in me during my entire work. I thank him for his kindness and valuable comments and discussions to complete work on time. Apart from the research talks, he lead me throughout the thesis during my tough times and motivated me to reach my goal in the end. Finally, thanks for being available 24/7 as a mentor and a friend.\par

There are many people from Valeo whom I would like to thank, whose contributions have helped me make this thesis possible. I would like to express my gratitude to my supervisor \textit{Dr. Stefan Milz} from Valeo, for the useful comments, remarks, and engagement throughout the Ph.D. thesis's learning process. \textit{Johannes Petzold}, my manager, and all my colleagues at Valeo, Kronach, for providing me with a wonderful environment to work in. Grateful to \textit{Clément Pinnard} for the initial research discussions on the \emph{SfM} framework and the discussions about depth estimation. Special thanks to \textit{Christian Witt} for teaching me advanced python and not so joyful rigorous code reviews and always being available for research discussions. \textit{Markus Bach} for all the math fun and research ideas, funny conversations about random things, and always being willing to help me. \textit{Kai Fischer}, for all the fun talks about the work and food, arguments about Python vs. C++. \textit{Sandesh Hiremath} for the initial research discussions. \textit{Martin Simon}, my team, lead for encouraging me at times during my thesis. \textit{Ganesh Sitsu} for all the awesome fun chats and the WoodScape dataset creation. \textit{Hazem Rashed} for all the hard work during MTL integration. \textit{Isabelle Leang} and \textit{Fabian Bürger} for the research discussions on MTL. \textit{Michal U\v{r}i\v{c}\'{a}\v{r}} for the discussions on GAN's. \textit{Marvin Klinger} from TU Braunschweig, for the research collaboration, interminable paper corrections, and writing. To all the other people who helped me during my Ph.D. time.\par

I would like to thank my friends and family. My parents, especially my mother, for being present for me throughout the years. Their prayers, sacrifices for educating me and preparing me for the future. They have been a constant support; as a result of which, I could pursue a Master's and a Ph.D. degree in Germany. I would like to thank my close friends \textit{Bharath Krishnaiah} and \textit{Jagadish Subramani}, who were part of my Master's cohort, currently working in Germany, for being supportive throughout my Ph.D. during my tough times and during the corona crisis. Thanks for all the food and memorable, joyful moments in life. To \textit{Pratik Kamble} for making sure I didn't sleep enough, for making me believe in myself, and for listening to all my research ideas late at night. To my childhood friend \textit{Sriram}, for being part of my entire educational journey from high school to engineering and reviewing my literature. Finally, I would like to express my thanks to those involved directly or indirectly in completing my project.\par

\end{acknowledgements}
\dominitoc
\tableofcontents 
\listoffigures 
\listoftables 
\mainmatter 
\pagestyle{thesis} 
\setcounter{mtc}{6}
\chapter{Introduction}
\label{Chapter1}
\minitoc


Since the early 1960s, we have been pursuing the fantasy of commuting between places while sitting in a driverless car with no manual intervention. Over the last decade, autonomous driving (AD) has piqued the interest of vehicle manufacturers more than ever before. The vast and ground-breaking advances in artificial intelligence (AI) and computer vision made possible by machine learning are the primary drivers of this developing trend.\par

Let us consider the case of an automobile. According to global statistics, approximately \textbf{3700} lives are lost due to road accidents every day (approximately \textbf{1.35 million} people per year) and \textbf{20} and \textbf{50 million} people are left with non-fatal injuries~\cite{who_accident_2021}. Out of those accidents, more than 70\% are caused due to human errors. Despite robust safety standards developed by the manufacturers and technology evolving massively, we have not reached an acceptable number of traffic accidents. What could be the possible reason? Do we have a long-term solution for this? Indeed, AI and autonomous systems could work as magic bullets in these situations. So, the basic principle involves machines taking control over everything. This would mean eradicating human interventions completely, which is the root cause of many of these problems. \par

\begin{figure}[!t]
  \centering
    \includegraphics[width=\textwidth]{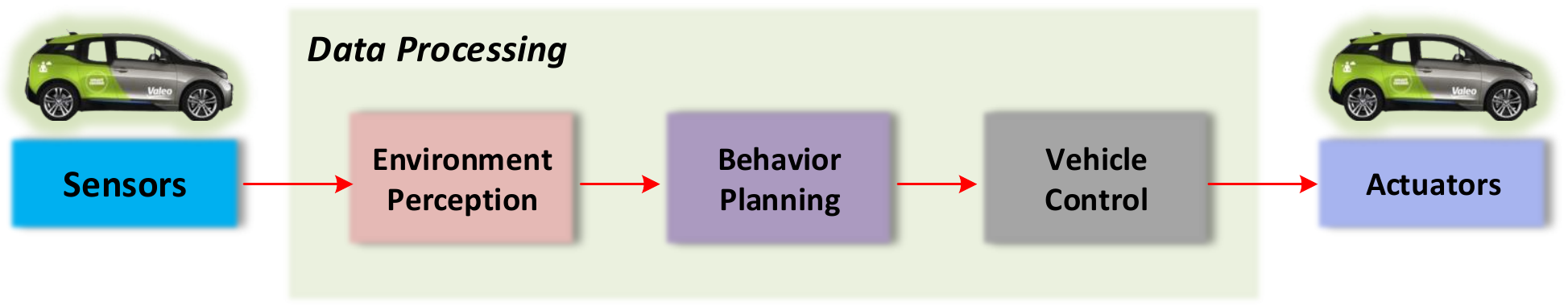}
    \caption[\bf Major components of an autonomous driving system.]{\textbf {Major components of an autonomous driving system~\cite{raaijmakers2017towards}.}}
    \label{fig:autonomous-system}
\end{figure}

An autonomous vehicle drives itself without the assistance of a human operator, using a collection of sensors, cameras, radar, and AI algorithms. Experts have identified five stages in the growth of self-driving vehicles. Each level defines how much a car may take over activities and obligations from its driver, as well as how the car and driver interact: 1) Driver assistance, 2) Semi-automated driving, 3) Highly automated driving, and 4) Fully automated driving 5) Complete automation~\cite{sae2014taxonomy}. The complete AD system can be roughly encompassed in five primary components across all five levels, as indicated in Figure~\ref{fig:autonomous-system}.\par

\textbf{Perception System in Action:} Sensors collect data from the environment and send it to the next ring of this chain, \emph{Environment Perception} where it is processed for the subsequent decision-making processes regarding the vehicle's next action. As a result, the extracted information at this point is critical for an autonomous vehicle. There are different types of sensors available to collect data from the environment. Our attention is drawn to image sensors, which convert light waves into signals that transfer information to build an image. Each camera is made up primarily of lenses and sensors. Most autonomous vehicle industry efforts are focused on advanced driver assistance systems (ADAS) since that is the first step for fully self-driving cars. The backbone of all ADAS applications is cameras, usually monocular or stereo vision systems. Regardless of the method, cameras are the basis for safe autonomous cars that can "\textit{see and drive}" themselves. Features, such as adaptive cruise control, can be implemented robustly as a fusion of radar or LiDAR data with cameras, usually for non-curvy roads and higher speeds. At present, this method has guided autonomous vehicles and ADAS to \textbf{Level 2} autonomy. In these situations, vehicles can control certain functions only — like emergency braking and advanced lane assist, and lane change. Still, in the end, humans are behind the wheel of these vehicles~\cite{Foresightauto_2021}.\par
The holistic scene understanding of the environment makes autonomous cars possible. Without this ability to sense the near-field environment, the autonomous systems would not have a way to know what speed to set, decide when to turn, when to make a lane switch or when to apply brakes. The complex, intelligent systems should have these abilities equipped to make split-second life-saving decisions. When an autonomous vehicle travels from source to destination, a navigator such as Google Maps or high-definition maps generates a high-level route. A series of connected nodes at finite distances make up this route. The vehicle moves from one node to another in a repetitive way until it reaches its destination. The sensing stage collects information about the surroundings using one of three sensors or a combination of these sensors such as cameras, radar, and LiDAR. Cameras are one of the most traditional means since they can visually perceive the scene in the same way humans do with vision. Radars are used in conjunction with cameras as an auxiliary method to detect large objects. LiDAR uses an array of light pulses to measure distance. Some use a combination of all three sensors. Perception involves the extraction of useful information from the raw data like lane positions, pedestrians, and other vehicles~\cite{siam2017deep}, moving objects detection~\cite{siam2018modnet} and recognition of drivable regions. Localization is the vehicle's ability to precisely know its position in the real world at decimeter accuracy~\cite{tripathi2020trained, milz2018visual}. In simple words, perception answers what is around the vehicle, and localization answers precisely where the car is. Path planning algorithms~\cite{lavalle2006planning} make use of this related information to define a path to navigate from one node to another.\par

Historically, most autonomous car companies (\eg, BMW, Audi, Toyota) have relied heavily on LiDAR since, until recently, neural networks were not powerful enough to handle multiple camera inputs. The laser sensors currently used to detect 3D objects in autonomous cars' paths are bulky, ugly, expensive, energy-inefficient – and highly accurate~\cite{camera_over_lidar_2021}. These LiDAR sensors are affixed to cars' roofs, increasing wind drag, a particular disadvantage for electric cars. They can add around $10,000\$$ to a car's cost. However, despite their drawbacks, most experts have considered LiDAR sensors the only plausible way for self-driving vehicles to safely perceive pedestrians, cars, and other hazards on the road~\cite{camera_over_lidar_2021}. As with human eyes, cameras capture the resolution, small details, and vividness of a scene with such detail that no other sensors, including radar, ultrasonic, and lasers, can match~\cite{Foresightauto_2021}. \textit{Tesla} is one of the most notable companies that has placed a big bet on cameras, integrating eight of them into each vehicle~\cite{tesla_2021}, along with a powerful deep neural network called HydraNets~\cite{mullapudi2018hydranets}. With the primary factor of cost and the inspiration from humans to mimic the way nature forged us to drive. \emph{This thesis will mainly focus on building a unified perception system from one of the three primary sensors} \ie\emph{using cameras only}.\par
\begin{figure}[!t]
  \centering
    \includegraphics[width=\textwidth]{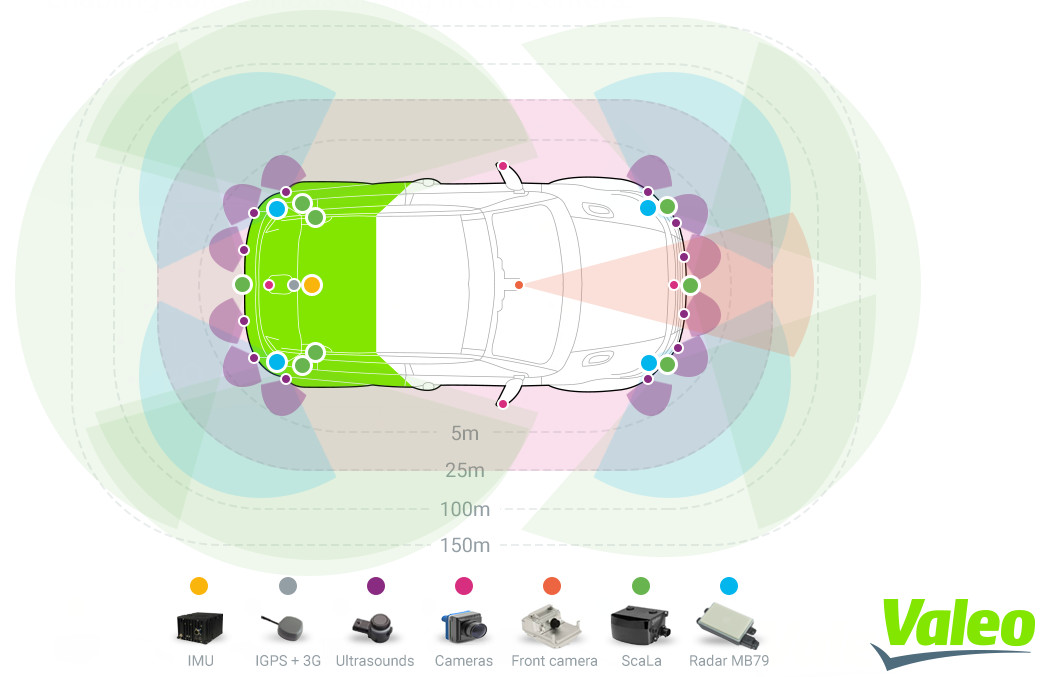}
    \caption{\bf Illustration of the near-field range and sensor suite of an exemplary self-driving car.} 
    \label{fig:near-field}
\end{figure}
\section{Goal}

\begin{figure}[!t]
  \centering
  \begin{minipage}[t]{0.495\textwidth}
    \centering
    \includegraphics[width=\textwidth]{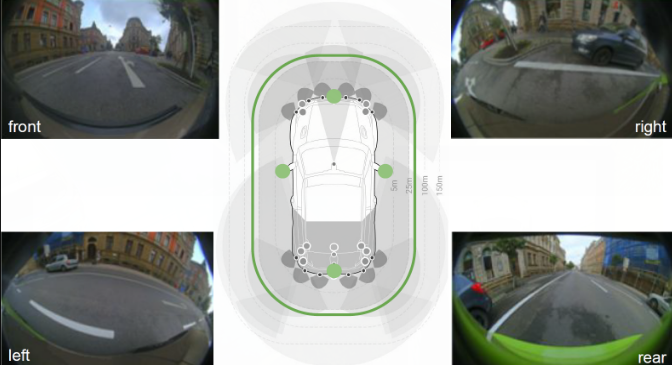}
    \caption{\bf Surround view system with four fisheye cameras.}
    \label{fig:surround-view}
  \end{minipage}%
  \hfill
  \begin{minipage}[t]{0.495\textwidth}
    \centering
    \includegraphics[width=\textwidth]{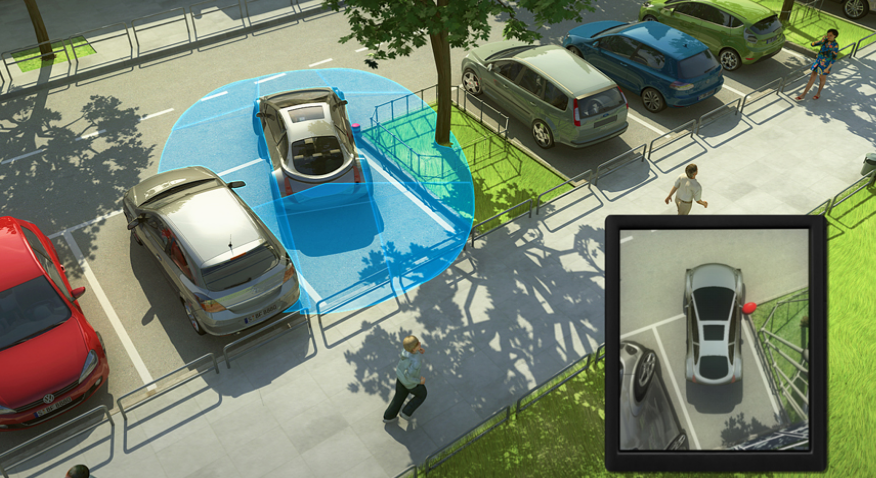}
    \caption[\bf Illustration of a tight parking scenario.]{\textbf{Illustration of a tight parking scenario.} Figure reproduced from~\cite{parking_2021}.}
    \label{fig:parking-scenario}
  \end{minipage}
\end{figure}

Near Field perception for AD is a region from 0-10 meters and $360\degree$ coverage around the vehicle as shown in Figure~\ref{fig:near-field}. Some of its use cases are automated parking, traffic jam assist, and urban driving. The sensor suite includes ultrasonics, fisheye-cameras, and radar (see~\ref{fig:near-field}). There are limited datasets and very little work on near-field perception tasks as the main focus is on far-field perception. In contrast to far-field, it is more challenging due to high precision object detection requirements of $10\,cm$. For example, let us look at the parking scenario in Figure~\ref{fig:parking-scenario}. The car needs to be parked in a tight space with partial object visibility and no room for error, requiring high precision. Four fisheye cameras are sufficient to cover the near-field perception as shown in Figure~\ref{fig:surround-view}.\par

Standard algorithms can not be extended easily on fisheye cameras due to their large radial distortion. There is very little work on the perception algorithms on fisheye cameras. Also, most of the current AD systems are \textbf{Level 2}. In this thesis, we focus on building a holistic $360\degree$ scene understanding for a near-field perception system that constitutes the necessary modules for a \textbf{Level 3} AD stack \textit{\textbf{using four fisheye cameras}}. The developed framework will be called \textit{OmniDet} for two reasons. Firstly, \textit{Omni} can be used for wide-angle omnidirectional cameras. Secondly, the word \textit{Omni} means all, and we are detecting all necessary objects for such a \textbf{Level 3} AD system. Thus, it is a framework that does a $360\degree$ Near-Field Detection.\par
The naive approach is rectifying the fisheye images and applying these algorithms. The standard question which arises when we talk about distortion in fisheye cameras is: \emph{Why do we not rectify the images?}
\begin{itemize}[nosep]
    \item Most algorithms are usually designed to work on rectified pinhole camera images.
    \item Removing distortion leads to a significant loss in the Field-of-View.
    \item For a horizontal Field-of-View (hFoV) greater than 180°, rays incident from behind the camera make it theoretically impossible to establish a complete mapping to a rectilinear viewport. Thus the rectification defeats the purpose of using a wide-angle fisheye lens.
    \item Resampling distortion artifacts are particularly strong in the periphery as a small region in the fisheye image is expanded to a larger region in the rectified image. The texture is lost, and noise is introduced.
\end{itemize}
A few of the most important questions and challenges that arise for near field perception are answered by these perception tasks:
\begin{itemize}[nosep]
    \item What is the geometry of the scene? Furthermore, How far is an object at a pixel level in an image? \textbf{Monocular depth estimation}
    \item What is around me? and What type of object at a pixel level in an image?  \textbf{Semantic Segmentation}
    \item How are different parts of the scene moving? or Is the object moving? \\ \textbf{Motion Segmentation}
    \item How to identify objects and locate them? \textbf{2D Object Detection}
    \item Can we perceive the world around us clearly? \textbf{Soiling Detection}
\end{itemize}
What if a simple sensor modality can provide all the cues listed above? \textit{A single \textbf{RGB Fisheye Camera}}. One of the other primary goals of this thesis is to target large-scale industrial deployment of these tasks in millions of cars; to achieve it, we need to consider some critical practical factors. The model needs to be robust to intrinsic camera variations arising due to manufacturing tolerances. Low-cost embedded hardware is needed for commercial reasons. Henceforth, the performance of these models needs to be real-time capable for driving scenarios. Model validation and safety certifications are required for compliance. Considering all this into account, we propose to use a single \textbf{multi-task learning (MTL)} (learn multiple tasks using a single input) model in contrast to using multiple models for different tasks. The MTL framework would be capable of learning geometry and semantics-related tasks from \textit{\textbf{monocular videos only}}. To further analyze the network's vulnerability against adversarial attacks, we plan to apply white and black box attacks for targeted and untargeted cases while attacking a task and inspecting the effect on all the others. The most challenging perception problem of all is the \textit{distance estimation} on raw fisheye cameras. We will explore this geometry task in detail and create novel methodologies to obtain distance maps on the fisheye camera. We also perform the first detailed study on object detection on fisheye cameras and explore various better representations of bounding boxes to adapt the fisheye camera's geometry. We also introduce synergy between depth, semantics, and object detection afterward and complete the MTL framework.\par

\section{Outline}

At first, we will look into the basics of the commonly used fisheye camera projection models in \textbf{Chapter}~\ref{Chapter2} followed by the basics and definition of the perception tasks. In \textbf{Chapter}~\ref{Chapter3} we will look into the background of all the perception tasks. In this thesis, we will focus mainly on the distance estimation on raw fisheye monocular videos in \textbf{Chapter}~\ref{Chapter4} and \textbf{Chapter}~\ref{Chapter5}. \textbf{Chapter}~\ref{Chapter4} shows novel methodologies to obtain distance maps on raw fisheye images using a CNN. We will tackle this task by setting up a \emph{structure-from-motion} (\emph{SfM}) framework and use the concept of view synthesis. The entire approach is self-supervised. The problem is more challenging than stereo-based approaches as the network also needs to solve the relative poses between the source images to reconstruct the target. This inherently contains the task of visual odometry, which will be part of our perception stack. At last, in this chapter, we generalize the approach to consider any camera geometry of choice described in Section~\ref{sec:projection-models} for the projection operation involved in the view-synthesis process. With this framework, we extend our work on public datasets such as KITTI~\cite{geiger2012we}, and achieve accurate depths, and outperform all previously published self-supervised methods.\par

In \textbf{Chapter}~\ref{Chapter5}, we focus on improving the geometry cues by leveraging semantics guidance. We reason about the \lone loss function and solve the infinite depth issue, which causes holes during inference, by incorporating the semantic information during view synthesis. We modify the reconstruction loss and replace it with a robust general loss function and obtain significant gains in accuracy. We examine how to leverage more directly the \textit{semantic} context of the scene to guide geometric representation learning while remaining in the self-supervised regime. One of the thesis's main goals is to target a real-time distance estimation convolutional neural network (CNN) design that can be deployed in millions of vehicles having its own set of cameras. Later in this chapter, to do so, we develop a novel camera geometry adaptive multi-scale convolution to incorporate the camera parameters into the self-supervised \emph{SfM} framework. We improve upon the previous work and obtain state-of-the-art results on the KITTI and WoodScape datasets.\par

In \textbf{Chapter}~\ref{Chapter6}, we look into the localization aspect of an autonomous car. We focus on the 2D object detection task in fisheye cameras and perform the first detailed study on object detection on fisheye cameras for AD scenarios. We present novel representations of bounding boxes on raw fisheye camera images \ie, oriented bounding box, ellipse, and generic polygon for object detection. To encourage further research in this direction, we will also make a public release of the dataset comprising $10,000$ images with annotations for all the object representations.\par

In \textbf{Chapter}~\ref{Chapter7}, we develop a whole scene understanding MTL framework for surround-view camera systems named \textit{Omnidet}. It would comprise all the six tasks listed in the thesis's goal. The developed CNN model is real-time capable on an automotive embedded platform. The entire perception system is \textit{using cameras only}. It is a distinct approach considering the immense challenge of AD: one which does not rely on infrastructures such as high-definition maps or extremely costly sensor payloads, yet can perform complex driving tasks using cameras only. It will be the first six-task perception network on fisheye cameras evaluated on the WoodScape dataset. We transfer the framework to a five-task network on the public datasets KITTI and CityScapes and establish state-of-the-art results on the KITTI Eigen depth benchmark and the KITTI odometry benchmark. We apply white and black box attacks on the MTL framework to further analyze the system's robustness on adversarial attacks, considering both targeted and untargeted scenarios.\par

In \textbf{Chapter}~\ref{Chapter8} we will identify the limitations of the developed system and discuss the challenges and shortcomings of the approach. Finally, in \textbf{Chapter}~\ref{Chapter9} we will draw up a conclusion based on the discussion brought in the preceding chapters.
\chapter{Background} 
\label{Chapter2}
\minitoc

This thesis aims to build a perception system that constitutes a \textbf{Level 3} AD stack using a multi-task CNN model, covering the necessary modules for near-field sensing use cases like parking or traffic jam assistance. We propose to design a multi-task model of six primary tasks necessary for an autonomous driving system: depth/distance estimation, visual odometry, semantic segmentation, motion segmentation, object detection, and lens soiling detection. This chapter will provide in-depth basic intuitions of the tasks and the terminologies used in the further chapters of the thesis. The latter part will focus on the datasets and the evaluation metrics. This specific background will determine our strategy on which quality measure is more meaningful regarding safety and path planning for automated parking.\par
Perception, localization and mapping, sensor fusion, path planning, and decision control are vital components of automated driving. Various perception tasks are required to provide a robust system covering a wide variety of scenarios. In general, a perception system comprises geometric and semantic scene understanding of the environment. Cameras are a dominant sensor for perception since roadway infrastructure is traditionally designed for human visual perception. Semantic tasks such as object detection~\cite{viola2001robust, girshick2015fast, redmon2016you, lin2017feature} (detecting pedestrians, vehicles, and cyclists, etc. with bounding boxes), semantic segmentation~\cite{long2015fully, noh2015learning, badrinarayanan2017segnet, segdasvisapp19} (pixel-wise labeling of road, lanes, and curbs, etc.) and soiling segmentation~\cite{uvrivcavr2019soilingnet, uricar2019challenges} (pixel-wise labeling of opaque and transparent) usually falls under the semantic branch, wherein fisheye cameras mounted low on a vehicle are susceptible to lens's soiling due to the splash of mud or water from the street. These are some of the significant tasks that help build a visual perception system.\par

The semantic tasks typically require a large annotated dataset covering various objects. However, it is practically infeasible to cover every possible object. Thus, generic object detection using geometric cues like motion or depth for rare objects is added to the perception system. Due to the multi-task training, these tasks will not only complement the detection of standard objects but also provide the final model with higher robustness.\par

Motion is a dominant cue in automotive scenes, and it requires at least two frames or the use of dense optical flow~\cite{farneback2003two, kundu2009moving, lin2014deep, dosovitskiy2015flownet, ilg2017flownet, vertens2017smsnet} (detect motion and estimate velocities of moving objects). In the case of geometric task \ie depth estimation~\cite{Eigen_14, eigen2015predicting, garg2016unsupervised, zhou2017unsupervised, monodepth17, godard2019digging} (distance in a real-world from ego vehicle). Finally, the visual odometry task is required to place the detected objects in a temporally consistent map.\par

In general, all the tasks mentioned earlier are mainly demonstrated on pinhole images in current research approaches. There is minimal work in the area of fisheye perception. In the following sections, we look into these tasks' background and ground principles and the difficulties in applying these tasks onto the fisheye cameras.\par
\section{Fisheye Camera and Geometry}
\label{sec:fisheye-cameras}

The development of fisheye cameras has a long history. Wood initially coined the term fisheye in 1908 and constructed a simple fisheye camera~\cite{wood1908fisheye}, a fact that is acknowledged in the naming of the recently released \textit{WoodScape} and SynWoodScape dataset of automotive fisheye camera videos~\cite{yogamani2019woodscape, sekkat2022synwoodscape} respectively. This water-based lens was replaced with a hemispherical lens by Bond~\cite{bond1922fisheye}, and thus began the optical development of fisheye cameras. Miyamoto~\cite{miyamoto1964fisheye} provided early insight into the modeling of geometric distortion in fisheye cameras, suggesting the use of equidistant, stereographic, and equisolid models. These models were already known in the field of cartography (\eg~\cite{thomas1952projections} and many others).\par
\textbf{Rise of Fisheye Cameras:} There has been a significant rise in the usage of fisheye cameras in various automotive applications~\cite{eising2021near, dahal2019deeptrailerassist, dahal2021online, horgan2015vision, cheke2022fisheyepixpro, kumar2021multi}, surveillance~\cite{drulea2014omnidirectional} and useful applications in robotics~\cite{caruso2015large}, including robotic localization~\cite{gu2014fisheyelocalisation, rashed2020fisheyeyolo}, simultaneous localization and mapping~\cite{gallagher2021hybrid, liu2014fisheyeslam, Ji2020PanoramicSLAM, tripathi2020trained}, visual odometry~\cite{Matsuki2018DirectSparseOdometry}, due to their large field-of-view (FoV). Recently, several computer vision tasks on fisheye cameras have been explored including object detection~\cite{sistu2019real, yahiaoui2019optimization, zhu2018robotobjectdetection}, semantic segmentation~\cite{ye2020universal, yahiaoui2019overview}, soiling detection~\cite{uvrivcavr2019soilingnet}, motion estimation~\cite{yahiaoui2019fisheyemodnet, mohamed2021monocular, rashed2019motion}, image restoration~\cite{uricar2019desoiling}, underwater robotics~\cite{meng2018underwater}, aerial robotics~\cite{qiu2017model} and many other related fields. Depth estimation is an essential task in autonomous driving as it is used to avoid obstacles and plan trajectories. While depth estimation has been substantially studied for narrow FoV cameras, it has barely been explored for fisheye cameras~\cite{kumar2018monocular, kumarcnn, kumarself, zioulis2018omnidepth}. Fisheye cameras offer a significantly wider FoV than standard cameras, often with $180\degree$ FoV or even greater. This can offer several advantages, in particular, that fewer cameras can be used to achieve complete surround-view coverage.\par
\textbf{Need for Application of Perception Tasks on Fisheye Cameras}

Surround-view fisheye cameras have been deployed in premium cars for over ten years, starting from visualization application on dashboard display units to provide near-field perception for automated parking. Fisheye cameras have a strong radial distortion that cannot be corrected without disadvantages, including reduced FoV and resampling distortion artifacts at the periphery~\cite{kumar2020unrectdepthnet}. Appearance variations of objects are larger due to the spatially variant distortion, particularly for close-by objects. Thus fisheye perception is a challenging task, and despite its prevalence, it is comparably less explored than pinhole cameras.\par
\begin{figure}[t]
  \centering
    \includegraphics[width=0.7\textwidth]{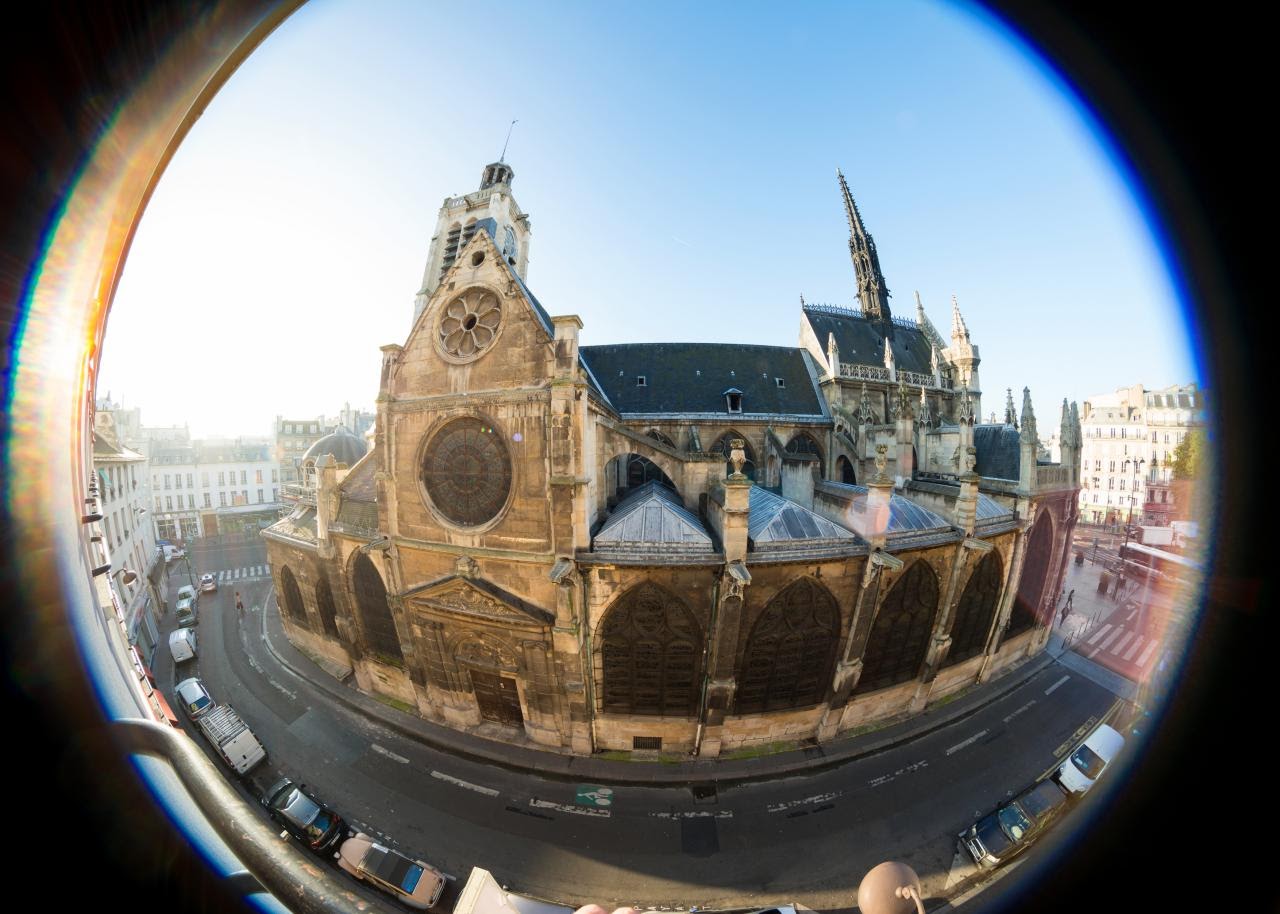}
    \caption[\bf An overview of an old church in Paris captured using a fisheye lens.]{\bf An overview of an old church in Paris captured using a fisheye lens. Figure reproduced from~\cite{fisheye_image_2021}.}
    \label{fig:fisheye-image}
\end{figure}
Different cameras with different FoV's are used to handle a wide variety of automotive use cases. The most common ones are around $100\degree$ hFoV cameras used for front camera sensing and $190\degree$ hFoV fisheye lens cameras for surround-view sensing. Due to their moderate to large FoV, these cameras suffer from lens distortion (see Figure~\ref{fig:fisheye-image}), whose main component is typically radial distortion and minor tangential distortion.\par
\subsection{Fisheye Lens Projection Models}
\label{sec:projection-models}

Several models have been developed to describe fisheye lenses. We can consider them in several classes. For example, we could consider a class of \textit{on-image} models, in which the fisheye projection is measured as a deviation from pinhole projection, \eg~\cite{basu1992pfet, devernay2001fovmodel}. Alternatively, we could consider a model in which the ray projection angle is manipulated at the projection center (\eg~\cite{yogamani2019woodscape, kannala2006fisheye}). We may refer to these models as \textit{ray deviation} models. Others still propose the use of a series of projections onto different surfaces to model fisheye distortion, for example~\cite{geyer2000ucm, khomutenko2016eucm, usenko2018doublesphere}, which we can refer to as \textit{geometric models}.

In the case of cameras with a more standard FoV, there is a very common geometry associated with them, the \textit{pinhole} model. One may first consider the intersection of a ray with a single planar surface at some fixed distance from the projection center. All models of the distortion due to the lens for such cameras are designed to shift the intersection point radially from the projection center on the plane (if one ignores tangential distortion). There are some varying proposals on what a model for this may look like, as discussed in the likes of~\cite{fitzgibbon2001divisionmodel}. However, the geometry for such standard FoV cameras is fixed as a ray-plane intersection. The role of geometric distortion correction is to find the deviation between the real camera and the hypothesized pinhole camera.\par

In a way, fisheye camera development in vision has been complicated by the lack of a single unifying geometry. There is a large set of models with different properties to describe fisheye projection and its radial distortion. This section aims at explaining some of these models, group them logically, and even highlight a couple of the more recently proposed models. They are almost re-derivations of existing models or describe projections that have already been known in different fields of science.\par

\textbf{Notation and Terminology}

Matrices are denoted by $\matr{A} \in \spce{R}^{m \times n}$. The usual notation for ordinary vectors $\vec{v} \in \spce{R}^n$ will be used, represented as $n$-tuples. Specifically, points in $\spce{R}^3$ will be denoted as $\vec{X}~=~(X,Y,Z)^\T$. We will use the same notation for points/vectors in $\spce{P}^n$, where necessary making it clear in the text that they should be interpreted as $(n\!+\!1)$-homogeneous vectors. Unit vectors are represented by the usual carat $\unitv{v} = (\unit{v}_1,\unit{v}_2,\unit{v}_3)^\T$.\par

We will rely heavily on points defined on the unit sphere embedded in Euclidean 3-space (itself embedded in $\spce{R}^3$). More formally, the unit sphere is ${S}^2 = \left\{\vec{u}\in\spce{R}^3 \, | \, \|\vec{u}\|=1\right\}$, and thus represented as a 3-vector $\vec{u} = (u_1,u_2,u_3)^\T$ of unit length. However, for clarity we forego the unit vector notation as it is implicit for points on the unit sphere.\par

\begin{figure}
  \centering
  \includegraphics[width=\linewidth]{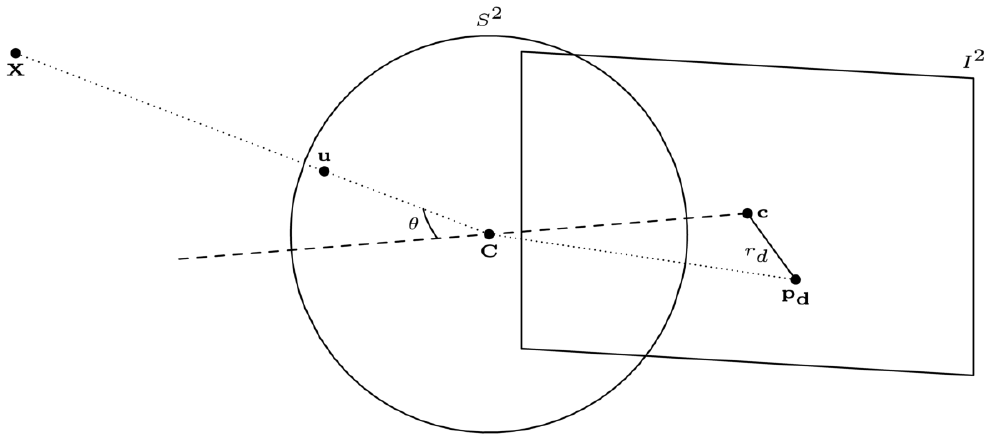}
  \caption[\bf Relationship between a fisheye image point and a point on the unit sphere.]{\textbf{Relationship between fisheye image point and point on the unit sphere.} $\mathbf{p}$ and $\mathbf{u}$ are equivalent points on the fisheye image and unit sphere, respectively, with $\mathbf{u}$ laying on the same ray as $\mathbf{X}$}.
  \label{fig:fisheye}
\end{figure}

In fisheye cameras, the lens views with FoVs $>\!\!180\degree$, and observed rays cannot all pass through a single flat image plane. Therefore, we cannot consider points on a projective plane, as they cannot encompass the entire FoV of a fisheye camera. We can define a mapping from defining a mapping from $\spce{R}^3$ to the fisheye image as $$\pi: \spce{R}^3 \rightarrow I^2$$
A true inverse is naturally not possible. However, we can define an unprojection mapping from the image domain to the unit central projective sphere
$$\pi^{-1}: {I}^2 \rightarrow {S}^2$$
Here, ${I}^2 \subset \spce{R}^2$ and ${S}^2 \subset \spce{R}^3$. Figure~\ref{fig:fisheye} demonstrates the relationship between the image plane and the unit sphere.\par
For the pinhole and the fisheye models that will follow, to simplify the math, note that we do not consider a separate $f$ parameter for each image direction (\ie, $f_x$ and $f_y$, we assume that the pixel aspect ratio is unit). However, it is easy to adapt the fisheye equations to include a non-unit aspect ratio, but that would not serve the purpose of this thesis. In addition, we do not consider the distortion center $\vec{c} = (c_x, c_y)^\T$, as this is just a translation on the image plane and does not affect the model performance.\par
\subsection{Classical Geometric Models}

We refer to the models discussed in this section as {\em classical}, as they have been researched for at least six decades~\cite{miyamoto1964fisheye}. One could also include the equisolid-angle model. However, we do not employ it for the scope of this work, and the readers are instead referred to~\cite{schneider2009fisheye, hughes2010fisheyeaccuracy}.\par

\subsubsection{Pinhole Camera Model}
\label{sec:pinhole-model}

For lenses with a moderate FoV (${<120\degree}$), the Brown–Conrady model~\cite{conrady1919decentred} is commonly used as it models both radial and tangential distortion. For larger FoV, this distortion model typically breaks down or requires very high polynomial orders. The KITTI dataset's calibration uses this model based on OpenCV's~\cite{opencv_library} implementation.
Assuming a point $\vec{X} = (X, Y, Z)^\T$ in the camera coordinate system, the pinhole model is
\begin{equation}  \label{eqn:pinhole}
    \vec{p} = \left(\frac{fX}{Z},\frac{fY}{Z}\right)^\T
\end{equation}
or, if we consider it as a radial function
\begin{equation} \label{eqn:pinholeradial}
    r_u = f_p\tan{\theta}
\end{equation}
where $\theta$ is the field angle of the projected ray, note that the parameter $f_p$ is sometimes referred to as the focal length. However, it has little to do with the optical focal length of the physical lens system (which can often be made up of many lens elements). $f_p$ is a scaling factor that converts from a projection surface at a unit distance from the project center to the pixel coordinates of the camera.\par

In this model, the projection function $X_c \mapsto \Pi(X_c) = p$ maps a 3D point $\vec{X} = (X,Y,Z)^\T$ in the camera coordinate system to a pixel $p = (i, j)^T$ in the image coordinates. It is calculated in the following way:
\begin{gather*}
   x = X / Z, \quad y = Y / Z \\
   x' = x (1 + k_1 r^2 + k_2 r^4 + k_3 r^6) + 2 p_1 x y + p_2(r^2 + 2 x^2) \\
   y' = y (1 + k_1 r^2 + k_2 r^4 + k_3 r^6) + p_1 (r^2 + 2 y^2) + 2 p_2 x y \\
   i = f_x \cdot x' + c_x,\quad
   j = f_y \cdot y' + c_y 
\end{gather*}
where $k_1$, $k_2$, and $k_3$ are radial distortion coefficients, $p_1$ and $p_2$ are tangential distortion coefficients of the lens, $r^2 = x^2 + y^2$, $f_x$, $f_y$ are the focal lengths and $c_x$, $c_y$ are the coordinates of the principal point.\par
\subsubsection{Equidistant Projection}

\begin{figure*}[t]
     \centering
    \includegraphics[width=0.48\textwidth]{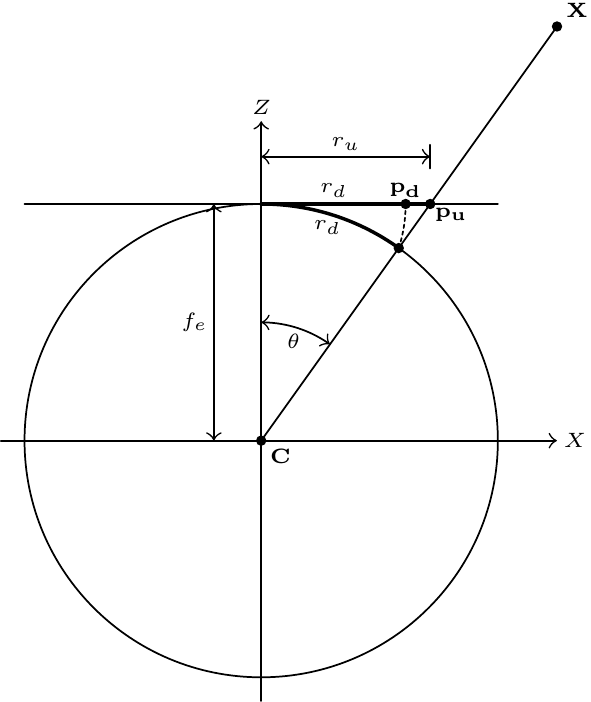}
    \includegraphics[width=0.48\textwidth]{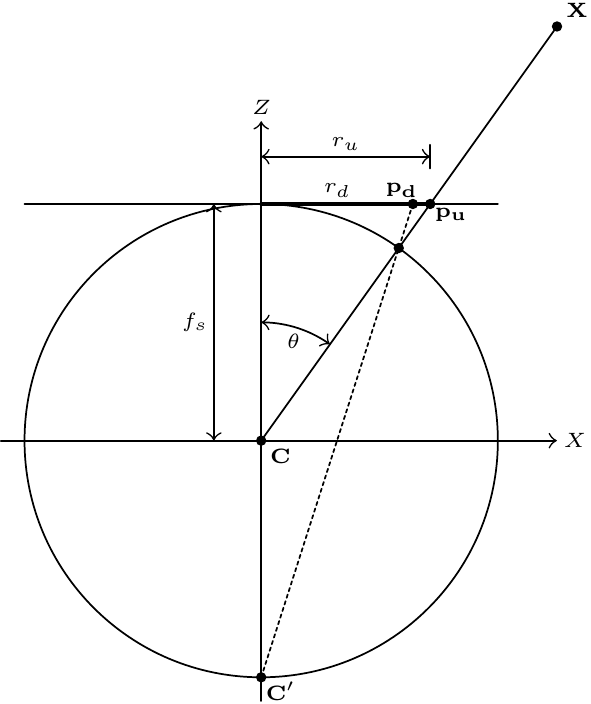} \\
    \caption{\bf Equidistant and Stereographic fisheye projection models.}
    \label{fig:equidistant-stereographic}
\end{figure*}

In the \textit{equidistant fisheye model}, the projected radius $r_d$ is related to the field angle $\theta$ through the simple scaling by the equidistant parameter $f_e$ (see Figure~\ref{fig:equidistant-stereographic}). \ie
\begin{equation}  \label{eqn:equidistantradial}
    r_d = f_e\theta 
\end{equation}
and thus
\begin{align} \label{eqn:equidistant}
    \pi(\vec{X}) &= \frac{f_e\theta}{d}\left[
    \begin{matrix} 
        X \\
        Y
    \end{matrix}
    \right] \nonumber \\
    d &= \sqrt{X^2 + Y^2} \nonumber \\
    \theta &= \acos \left(\frac{Z}{\sqrt{X^2 + Y^2 + Z^2}}\right)
\end{align}
\subsubsection{Stereographic Projection}

As with the equidistant model, in \textit{stereographic projection}, the center of the projection of $\vec{X}$ to the projection sphere is $\vec{C}$ (Figure~\ref{fig:equidistant-stereographic}). Consider that the image plane has a tangential point along the $Z$-axis (optical axis). There is a second central projection to the image plane in stereographic, with the antipodal point of the tangential point forming the center of projection. It is essentially a pinhole projection with a focal length of $2 f_s$. The stereographic projection is therefore described by
\begin{equation}  \label{eqn:stereographicradial}
    r_d = 2 f_s \tan\left(\frac{\theta}{2}\right)
\end{equation}
\begin{align}
    \pi(\vec{X}) & = \frac{r_d}{\sqrt{X^2 + Y^2}}\left[\begin{matrix} X \\ Y \end{matrix} \right] \nonumber \\
      & = \frac{2 f_s \tan{\frac{\theta}{2}}}{\sqrt{X^2 + Y^2}} \left[\begin{matrix} X \\ Y \end{matrix} \right] \nonumber \\
      & = \frac{2 f_s \tan{\left( \frac{1}{2} \atan{\left( \frac{\sqrt{X^2 + Y^2}}{Z} \right)} \right) }}{\sqrt{X^2 + Y^2}} \left[\begin{matrix} X \\ Y \end{matrix} \right] \\ \nonumber
      & = \frac{2 f_s \frac{\sqrt{X^2 + Y^2}}{\sqrt{X^2 + Y^2 + Z^2} + Z}}{\sqrt{X^2 + Y^2}} \left[\begin{matrix} X \\ Y \end{matrix} \right] \nonumber
\end{align}
\begin{equation}  \label{eqn:stereographic}
    \pi(\vec{X}) = \frac{2 f_s}{Z + ||\vec{X}||}\left[
    \begin{matrix} 
        X \\
        Y
    \end{matrix}
    \right]
\end{equation}
where $d$ and $\theta$ are equally defined as in (\ref{eqn:equidistant}). The inverse of (\ref{eqn:stereographicradial}), which we shall need later, is derived as
\begin{equation} \label{eqn:stereographicradialinv}
    \theta = 2 \atan\left(\frac{r_d}{2f_s}\right)
\end{equation}
\subsubsection{Orthographic Projection}

Similar to the previous projections models, the \textit{orthographic projection} begins with a projection to the sphere (Figure~\ref{fig:othographic-extended-othographic}). An orthogonal projection to the plane follows this. The orthographic projection is therefore described by
\begin{equation} \label{eqn:orthographicradial}
    r_d = f_o \sin\theta
\end{equation}
and thus
\begin{equation} \label{eqn:orthographic}
    \pi(\vec{X}) = \frac{f_o}{||\vec{X}||} \left[ 
    \begin{matrix} 
        X \\
        Y
    \end{matrix}
    \right]
\end{equation}
\subsubsection{Extended Orthographic Model}

\begin{figure*}[t]
     \centering
    \includegraphics[width=0.48\textwidth]{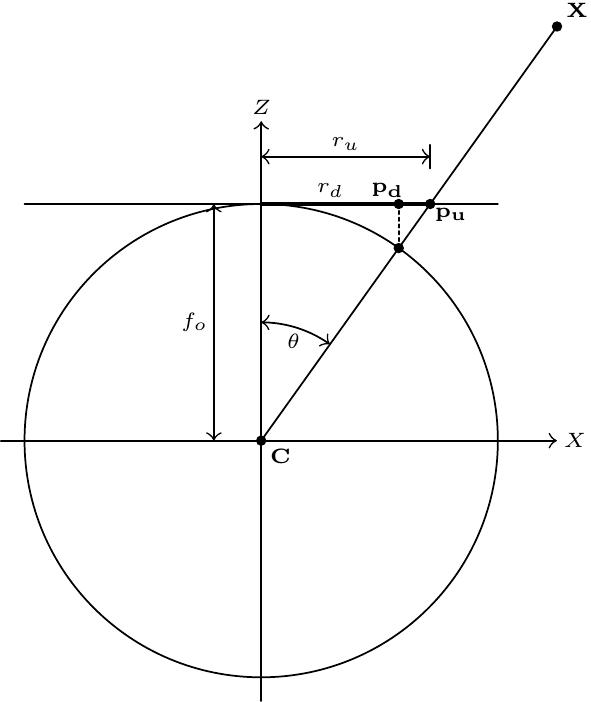}
    \includegraphics[width=0.48\textwidth]{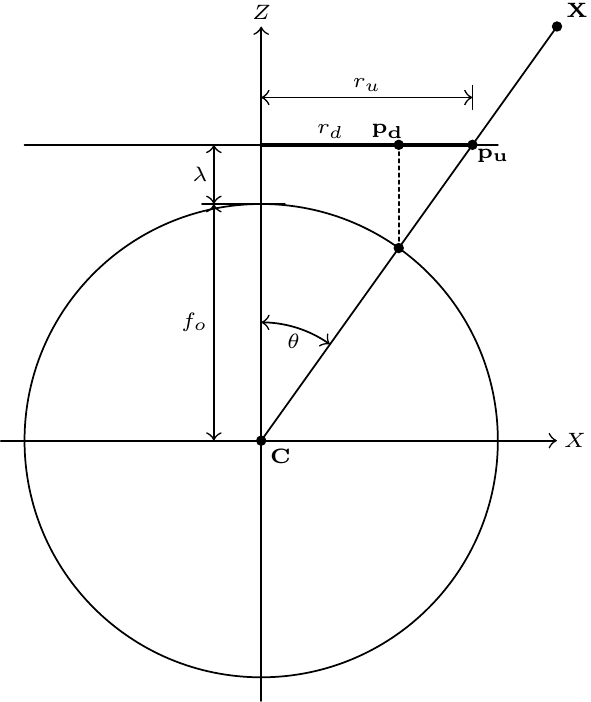} \\
    \caption{\bf Othographic and Extended Othographic fisheye projection models.}
    \label{fig:othographic-extended-othographic}
\end{figure*}

The \textit{Extended Orthographic Model}~\cite{kim2014model}, as demonstrated by Figure~\ref{fig:othographic-extended-othographic}, extends the classical orthographic model by freeing the projection plane from being tangential to the projection sphere, allowing an offset $\lambda$. The distorted projection remains the same as equations (\ref{eqn:orthographicradial}) and (\ref{eqn:orthographic}). However, the relationship between the distorted and undistorted radial distances is given by
\begin{equation}
    r_d = \frac{f_o}{\sqrt{(\lambda + f_o)^2 + r_u^2}}
\end{equation}
and thus is an on-image mapping given by
\begin{equation}
    \left[ \begin{matrix} 
        x_d \\
        y_d
    \end{matrix}
    \right] = \frac{f_o}{r_u \sqrt{(\lambda + f_o)^2 + r_u^2}}
    \left[ \begin{matrix} 
        x_u \\
        y_u
    \end{matrix}
    \right]
\end{equation}
\subsection{Algebraic Models}

We provide a short discussion on algebraic models of fisheye cameras, specifically polynomial models and the division model. We provide the polynomial model discussion for completeness, though we concentrate on the geometric models for the remainder of the chapter.\par
\subsubsection{Polynomial Models}

The classical \textit{Brown–Conrady model} of distortion for non-fisheye cameras~\cite{brown1966, conrady1919decentred} uses an odd-termed polynomial, $r_d = P_n(r_u)$, to describe the radial distortion on the image (\ie mapping $r_u$ to $r_d$), where $P_n$ represents some arbitrary $n$-th order polynomial. Despite its age, the Brown-Conrady model is the standard distortion model in software implementations for non-fisheye cameras~\cite{opencv_library, MATLAB:2021}. To account for fisheye distortion, an on-image polynomial model known as the \textit{Polynomial Fisheye Transform} (PFET) was proposed in~\cite{basu1992pfet}. The difference between the PFET and the Brown-Conrady model is that the PFET allows both odd and even exponents to account for the added distortion encountered in fisheye cameras.\par

A class of polynomial fisheye models exists, in which the mapping of the field angle to the image plane is via a polynomial, \ie $r_d = P_n(\theta)$. For example, Kannala-Brandt~\cite{kannala2006fisheye} (and as implemented in the popular \textit{OpenCV} software~\cite{opencv_library}) propose an polynomial model of order $n=5$, or more, with odd exponents only. In~\cite{yogamani2019woodscape}, an $n=4$ polynomial containing both even and odd exponents is proposed. Neither model used a constant coefficient term in the polynomial. In~\cite{ying2006poly} a fifth-order polynomial is proposed, but they reduce it to four independent parameters if the fisheye radius and the FoV are known. All of the above could be interpreted as a generalization of the equidistant model, which is a first-order polynomial. In this case, the projection sphere is replaced by some surface defined by the given polynomial. However, this is forcing a geometric interpretation with little utility.\par
\subsubsection{Division Model}

The \textit{division model}~\cite{fitzgibbon2001divisionmodel} of radial distortion gained some popularity due to the excellent property that, at least for the single parameter variant, straight lines project to circles in the image~\cite{Wildenauer2013DivisionModel, antunes2017unsupervised, bukhari2013automatic}. For many lenses, the single parameter variant performs very well~\cite{courbon2012evaluation}. It is given by
\begin{align} 
    r_u &= \frac{r_d}{1-ar_d^2} \label{eqn:divisionmodelradialund} \\
    r_d &= \frac{r_u}{1+ar_u^2} \label{eqn:divisionmodelradialdis}
\end{align}
where $a$ is the division model parameter. This was extended in~\cite{hughes2010equidistant} by adding an additional scaling parameter, which improved the modeling performance for certain types of the fisheye lenses. Note that the division model was presented as an \textit{on-image} mapping, though it can be expressed as the projection function
\begin{align}  
\label{eqn:divisionmodel}
    \pi(\vec{X}) &= \left[\begin{matrix}
        \frac{f r_d' X}{r_u'} \\
        \frac{f r_d' Y}{r_u'}
    \end{matrix}
    \right] \nonumber \\
    r_u' &= \sqrt{X^2 + Y^2} \nonumber \\
    r_d' &= \frac{r_u'}{1 + ar_u'}
\end{align}
\subsection{Geometric Models}

A set of more recent (at least, from the last couple of decades) fisheye camera models is also considered. Further reading on some of these models can be found, \eg, in~\cite{usenko2018doublesphere}, which also gives valuable information on the set of valid points for each projection.

\subsubsection{Field-of-View Model}

The FoV model~\cite{devernay2001fovmodel} is defined by
\begin{equation}  \label{eqn:fovmodelradial}
r_u = \frac{\tan(r_d \omega)}{2\tan\frac{\omega}{2}}
\end{equation}
where $r_u$ is the distance of the undistorted image point (pinhole image point) to the distortion center, $r_d$ is the corresponding distorted image point distance, and $\omega$ is the model parameter. The parameter $\omega$ approximates the camera field of view, though not exactly~\cite{devernay2001fovmodel}.\par

This is an \textit{on-image} model, like the Division Model, where $r_u$ and $r_d$ define undistorted and distorted radii on the image plane. Alternatively, it can be expressed as a projection function~\cite{usenko2018doublesphere}
\begin{align}  \label{eqn:fovmodel}
    \pi(\vec{X}) &= \left[\begin{matrix}
        \frac{f r_d' X}{r_u'} \\
        \frac{f r_d' Y}{r_u'}
    \end{matrix}
    \right] \nonumber \\
    r_u' &= \sqrt{X^2 + Y^2} \nonumber \\
    r_d' &= \frac{\atan2(2 r_u' \tan(\omega/2),z)}{\omega}
\end{align}
Note that $r_u'$ and $r_d'$ are related to the undistorted $r_u$ and distorted $r_d$ image plane radial distances, but are not quite the same, due to the scaling by $f$.\par
\subsubsection{Unified Camera Model}

The UCM was initially used to model catadioptric cameras~\cite{geyer2000ucm}. Later it was shown to be helpful when modeling fisheye cameras~\cite{ying2004division, courbon2007generic}. It has been shown to perform well across a range of lenses~\cite{Courbon2012FisheyeAnalysis}. The geometry of the projection is two-step. First, the point $\vec{X}$ is projected to a unit sphere, followed by a projection to a modeled pinhole camera. The UCM projection is given by
\begin{equation} 
\label{eqn:ucm}
    \pi(\vec{X}) = \frac{\gamma}{Z + \xi ||\vec{X}||} \left[
    \begin{matrix}
        X \\
        Y
    \end{matrix}
    \right]
\end{equation}
though~\cite{usenko2018doublesphere} propose a more numerically stable formulation. $\xi$ is the distance from the center of the unit sphere to the center of the pinhole projection, and $\gamma$ is the focal length of the secondary pinhole projection. If $\xi~=~0$, this model degrades to the pinhole model.\par
\subsubsection{Enhanced Unified Camera Model}

The UCM was extended by the Enhanced UCM~\cite{khomutenko2016eucm}, which replaced the spherical projection with a projection to an ellipsoid (or, in fact, a general quadratic surface), and was able to demonstrate some accuracy gain. The E-UCM is given by~\cite{usenko2018doublesphere}
\begin{equation} \label{eqn:eucm}
    \pi(\vec{X}) = \frac{f}{\alpha d + (1 - \alpha) Z} \left[
    \begin{matrix}
        X \\
        Y
    \end{matrix}
    \right]
\end{equation}
where $d = \sqrt{\beta(x^2 + y^2) + z^2}$.
\subsubsection{Double-Sphere Model}

Later still, the UCM was extended again by the double-sphere (DS) model~\cite{usenko2018doublesphere}, which added a second spherical projection to enable more complex modeling
\begin{align}  \label{eqn:ds}
    \pi(\vec{X}) &= \frac{f}{\alpha d_2 + (1 - \alpha)(\xi d_1 + Z)} \left[
    \begin{matrix}
        X \\
        Y
    \end{matrix}
    \right] \\
    d_1 &= \sqrt{x^2 + y^2 + z^2} \nonumber \\
    d_2 &= \sqrt{x^2 + y^2 + (\xi d_1 + Z)^2} \nonumber
\end{align}
Convincing results are presented in~\cite{usenko2018doublesphere} to demonstrate the effectiveness of the double-sphere model.
\subsubsection{Summary of Radial Distortion Models}

The radial distortion models are summarized below:
\begin{itemize}
\item Polynomial: $r(\theta) = a_1 \theta + a_2 \theta^2 + a_3 \theta^3 + a_4 \theta^4$
\item UCM: $r(\theta) = f\cdot\sin\theta / (\cos\theta + \xi)$
\item eUCM: $r(\theta) = f\cdot\frac{\sin\theta}{\cos\theta + \alpha\left(\sqrt{\beta\cdot \sin^2\theta + \cos^2\theta} - \cos\theta\right)}$ 
\item Rectilinear: $r(\theta) = f \cdot \tan\theta$
\item Stereographic: $r(\theta) = 2 f \cdot \tan(\theta/2)$ 
\item Double Sphere: $r(\theta) = f\cdot \frac{\sin\theta}{\alpha\sqrt{\sin^2\theta + (\xi + \cos\theta)^2} + (1-\alpha)(\xi + \cos\theta)}$
\end{itemize}
\subsection{Other Models}

While we have discussed many of the more popular fisheye projection models, the list is still incomplete. We have omitted some models are only rarely used. For example, Bakstein and Pajdla~\cite{bakstein2002panoramic} proposed two extensions to the classical models. Firstly, they allowed a second parameter in the stereographic model. Then by trying various combinations of classical models for a given fisheye camera, they propose what is essentially a weighted average of the stereographic and the equisolid angle projection. A logarithm-based \textit{Fisheye Transform} (FET) was also proposed in~\cite{basu1992pfet}, though the accuracy was low compared to other models. The hyperbolic sin-based model proposed in~\cite{pervs2002nonparametric}, and later used for wide-angle cameras~\cite{klanvcar2004wide}, is not discussed here. The cascaded one-parameter division model~\cite{mei2015radial} is also not mentioned.\par
\subsection{Similarity between Models}
\label{sec:commonality}

With the proliferation of fisheye models, it is natural to wonder if there is a commonality between some of the models or even if there has been repetition in the development of the models.\par

\textbf{General Perspective Projection and Fisheye Models}

The unified camera model is in a class of general vertical perspective projections of a sphere, which is known in the fields of geodesy and cartography~\cite{laubscher1965map, snyder1987maps}, with the addition of the trivial step of central projection to the spherical surface. The stereographic and the orthographic projections belong to this class as well. The stereographic projection has the pinhole projection center on the sphere's surface, while the orthographic projection has an infinite focal length (hence the term orthographic). The link between the stereographic projection and the UCM is described in~\cite{geyer2000ucm}.\par
\begin{figure}[!t]
  \centering
  \includegraphics[width=0.5\textwidth]{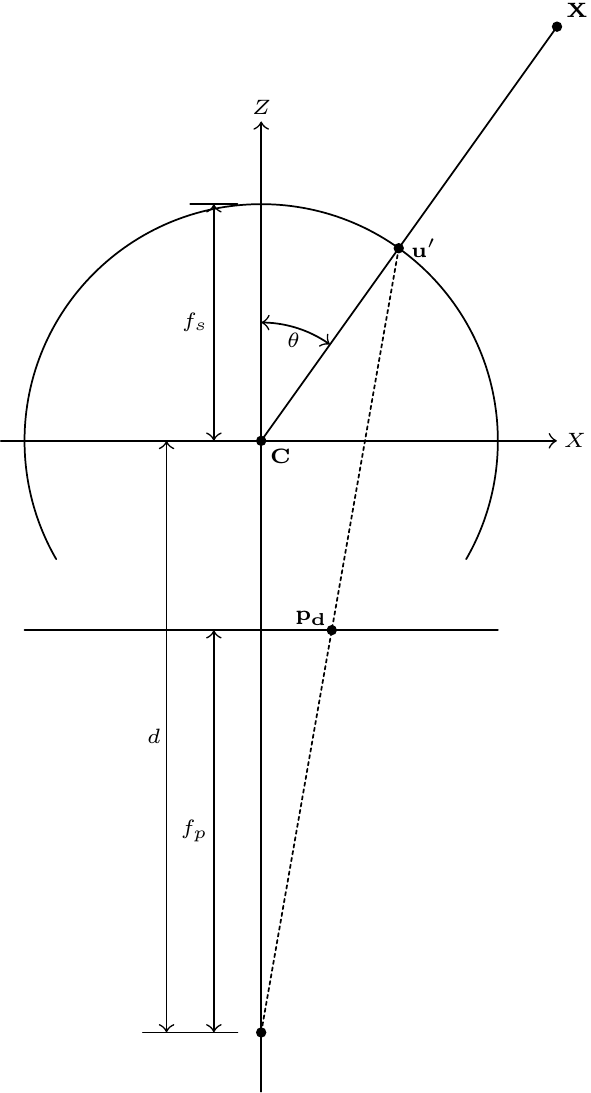}
  \caption{\bf The general perspective mapping.}
  \label{fig:generalperspective}
\end{figure}
Let us begin by examining the general vertical perspective projection, described by Figure~\ref{fig:generalperspective}. The pinhole camera is offset along the $Z$-axis by a distance of $d$. The projection to the sphere is given by
\begin{equation}
    \vec{u}' = f_s \frac{\vec{X}}{||\vec{X}||}
\end{equation}
Here we use $\vec{u}' = (u'_x, u'_y, u'_z)^\T$ for the point on the sphere of radius $f_s$, to distinguish it from $\vec{u}$ used previously to denote a point on the unit sphere. The point $\vec{p_d}$ is the pinhole projection of $\vec{u}'$
\begin{equation} \label{eqn:generalperspective}
    \pi(\vec{X}) = \frac{f_p}{u'_z + d} \left[\begin{matrix}
    u'_x \\
    u'_y
    \end{matrix} \right]
    =
     \frac{f_p}{Z + \frac{d}{f_s}||\vec{X}||} \left[\begin{matrix}
    X \\
    Y
    \end{matrix} \right]
\end{equation}
The $+d$ translates the point $\vec{u}'$ from the sphere to the pinhole coordinate system. Thus, with the two parameters $\gamma = f_p$ and the $\xi = d / f_s$, we have (\ref{eqn:ucm}), the UCM. Additionally, if we constrain the pinhole camera plane to be on the surface of the sphere (i.e. $d = f_s$), and make $f_p = 2f_s$, we get the stereographic equation (\ref{eqn:stereographic}).\par
The Enhanced UCM~\cite{khomutenko2016eucm} extended the UCM by projecting to an ellipsoid instead of a sphere. Again, this type of projection is known in geodesy and cartography for a long time~\cite{laubscher1965map, snyder1987maps} as \textit{ellipsoidal} general perspective projections. We will not re-derive the equations here but would refer the reader to the source material. As mentioned, the DS model~\cite{usenko2018doublesphere} extends the UCM by adding a second projection sphere to model more complex optics.\par

Thus the UCM, the E-UCM, and the DS models of fisheye lenses can be considered as generalizations of the stereographic camera model. It may be even more correct to say that they all (UCM, E-UCM, DS, division, and stereographic models) are part of a class of general perspective models. If we allow $f_s$ to approach infinity, then (\ref{eqn:generalperspective}) becomes the pinhole projection model. If we allow $f_p$ (and thus also $d$) to go to infinity, then we get the orthographic projection.\par

\begin{figure}[t]
  \centering
  \includegraphics[width=0.9\textwidth]{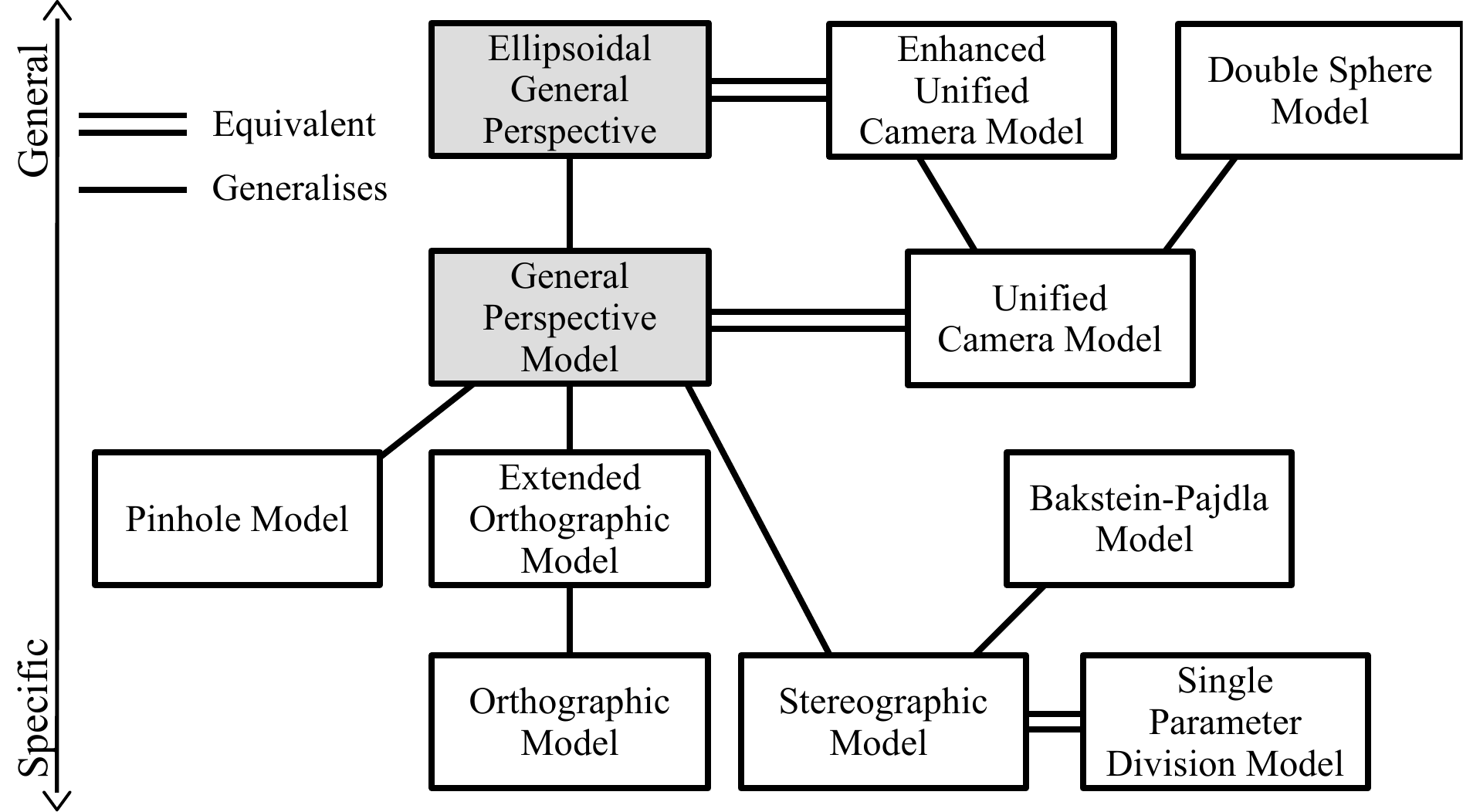}
  \caption[\bf The relationship between the various fisheye models and the general perspective projections]{\textbf{The relationship between the various fisheye models and the general perspective projections.} Double line indicates that two models are equivalent, and single line indicates a generalization/specialization.}
  \label{fig:relationships}
\end{figure}

Figure~\ref{fig:relationships} graphically shows the relationships between the various fisheye models and the general perspective projection. The Division Model stands out, as it was not initially designed as a geometric projection model, though its equivalence to the Stereographic Model is demonstrated in the next section. It should also be noted that the E-UCM is not restricted to ellipsoids but, depending on the parameters, may also be represented by hyperboloid or paraboloid projection surfaces. However, this is such a minor differentiation that we consider this still to be equivalent to the Ellipsoidal General Perspective Projection.\par

In Figure~\ref{fig:relationships}, we have attempted to provide a map of geometric fisheye models that are related to the General Perspective Projection. For a developer, this could be seen as a tool to guide the choice of model for a given application. One could attempt to use the simpler, more specialized models and, depending on the specific application, extend the development to one of the more general models in the case that errors remain high for a given camera model following calibration.\par
\textbf{Stereographic and Division Models}

We can combine the pinhole projection (\ref{eqn:pinholeradial}) with the inverse of the stereographic model (\ref{eqn:stereographicradialinv}) to give
\begin{equation}
r_u(r_d) = f_p\tan{\left(2\atan{\left(\frac{r_d}{2f_s}\right)}\right)}
\end{equation}
Elementary trigonometry yields
\begin{equation}
    r_u = \frac{f_s}{f_p} \frac{r_d}{1 - \frac{r_d^2}{4f_p^2}}
\end{equation}
Let us compare this with the single parameter division model. If we allow $a = 1 / 4f_p^2$, this is the same as the division model, (\ref{eqn:divisionmodelradialund}), up to a scaling factor $f_s / f_p$, which is discussed in~\cite{hughes2010fisheyeaccuracy}.\par
\textbf{Equidistant and Field-of-View Models}

Consider the radial pinhole projection given by (\ref{eqn:pinholeradial}), and the equidistant fisheye projection model (\ref{eqn:equidistantradial}). Combining the two to a similar form as the FoV model (\ref{eqn:fovmodelradial})
\begin{equation}  \label{eqn:equiundist}
r_u = f_p\tan\frac{r_d}{f_e}
\end{equation}
As $f_p$ and $f_e$ are free parameters, determined through calibration, we can set them to
\begin{equation} \label{eqn:fov2equi}
f_p = \frac{1}{2\tan\frac{\omega}{2}} \;\;\text{and}\;\; f_e = \frac{1}{\omega}
\end{equation}
Thus we see that (\ref{eqn:fovmodelradial}) and (\ref{eqn:equiundist}) are equivalent mapping functions. \ie, the equidistant and the field-of-view model are fundamentally the same model.\par
\textbf{Mapping of 3D to 2D for Fisheye Lenses}

For fisheye lenses, the mapping of 3D points to pixels universally requires a radial component $r\left(\theta\right)$ \cite{hughes2010fisheye}. The projection is a complex multi-stage process compared to regular lenses and thus we list the detailed steps:
\begin{enumerate}
    \item The point $X_c$ in camera coordinates is mapped to a unit vector
    as $S = (s_x, s_y, s_z)^T = X_c / \|X_c\|$.
    \item The incident angle against the optical axis (coincident with the $Z$-axis) $\theta = \frac{\pi}{2} - \arcsin\left(s_z\right)$ is computed.
    \item The radial function $r(\theta)$ to get the radius on the image plane (typically in pixels) is computed.
    \item Given the pixel distortion centre $(c_x,c_y)$, the pixel location is given by $i = r\cdot s_x / \rho + c_x$ and $j = r\cdot s_y / \rho + c_y$ with $\rho = \sqrt{s_x^2 + s_y^2}$. 
    \item (optional) Depending on the model used in Step 3, an additional distortion correction may needs to be applied.
\end{enumerate}
There is a great number of potential models for application with fisheye cameras. We have mentioned \textit{twenty} models, though this is not yet exhaustive. We have shown that there exists a strong relationship among many of the geometric models—at least seven of the models being related to or directly equivalent to the General Perspective Projection. In addition, we have shown that some of the more recently developed fisheye camera models are mathematically equivalent to the classical fisheye projection functions, being the stereographic and the equidistant models proposed decades ago. One could further theorize the existence of as yet undiscovered fisheye models. For example, from Figure~\ref{fig:relationships}, you could foresee the unification of the Ellipsoidal General Perspective model and the double sphere by developing a model that consists of sequential projections onto two quadratic surfaces.\par

The aim of this background is not to diminish the importance of any work in this space. At no point do we make any claim about the accuracy of any of the models. The accuracy of any model depends not just on the model itself, but on the application, on the lens and image sensor types, and the calibration procedure deployed. Instead, we would hope to provide a guide for further development (and perhaps unification) in fisheye projection models. Furthermore, perhaps it can be a valuable source of discussion for a developer considering which model is most appropriate for their camera and application.\par
\section{Single-Task Learning (STL)}

\subsection{Depth Estimation}

\begin{figure*}[!t]
  \centering
    \includegraphics[width=0.49\columnwidth]{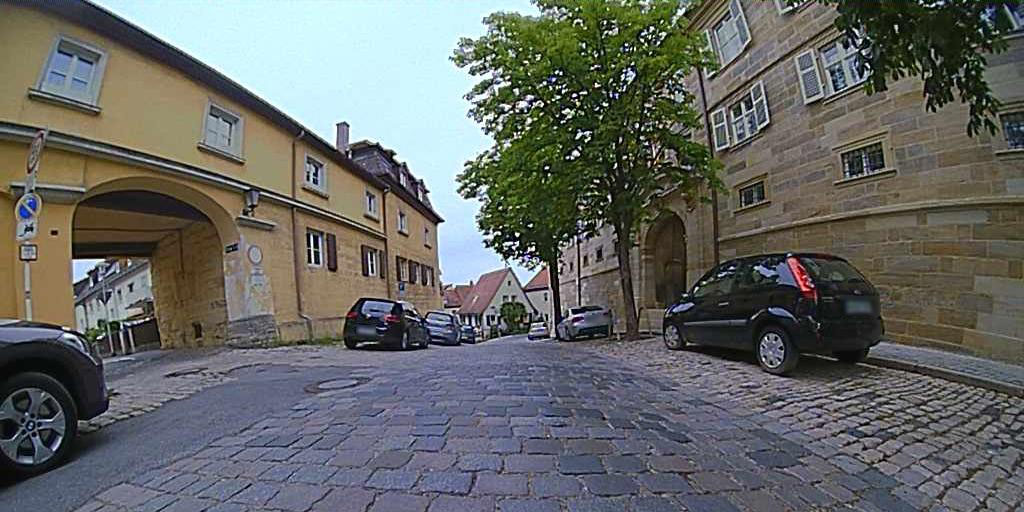}
    \includegraphics[width=0.49\columnwidth]{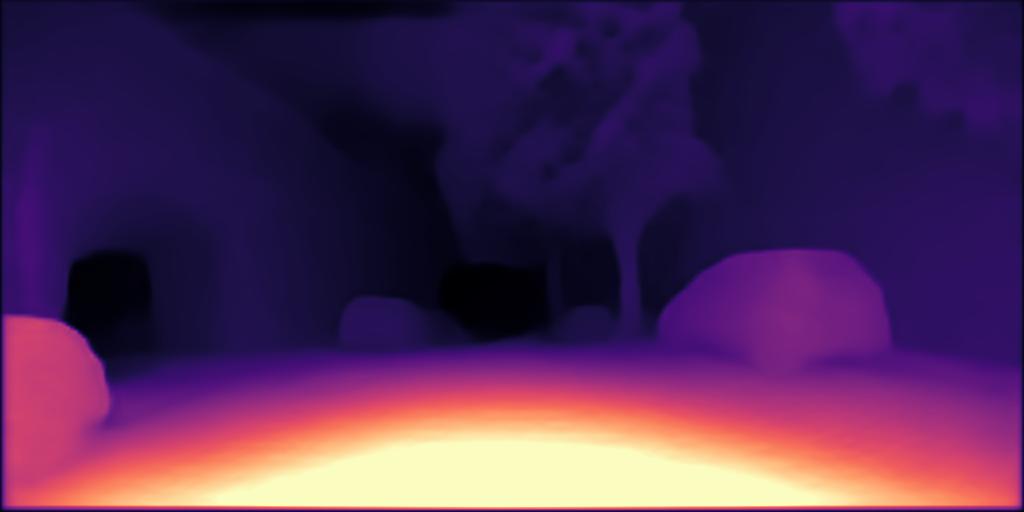} \\
    \caption[\bf A qualitative sample of distance/depth estimation on WoodScape.]{\bf A qualitative sample of distance/depth estimation on WoodScape~\cite{yogamani2019woodscape}.}
    \label{fig:depth-example}
\end{figure*}

\begin{figure}[t]
  \centering
    \includegraphics[width=\textwidth]{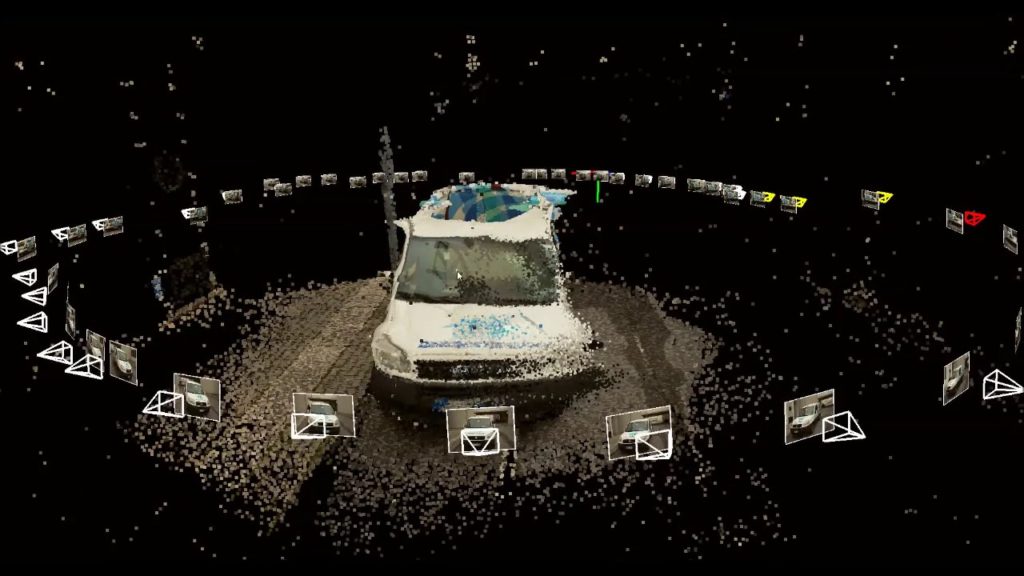}
    \caption[\bf An overview of reconstructed scene in 3D.]{\bf An overview of reconstructed scene in 3D. Figure reproduced from~\cite{3D_reconstruction_2021}.}
    \label{fig:3d-recon-image}
\end{figure}

Depth estimation involves estimating the distance to an object (or any
plane) at a pixel level as shown in Figure~\ref{fig:depth-example}.
For depth estimation, it is very useful to estimate the norm ($\sqrt{x^2+ y^2+ z^2}$) instead of $z$, because for fisheye images, the $z$ value can be (close to) zero for FoV > $180\degree$, which leads to mathematical problems, because all models have some direct or indirect division by $Z$. Instead, the norm is always zero (except for $x,y,z=0$) and allows a more numerical stable implementation. Calculating distance relative to a camera plane is still very challenging, but it is critical to unlocking exciting technologies such as autonomous driving, 3D scene reconstruction, and augmented reality. Distance estimation is a crucial requirement in robotics for performing various tasks such as perception, navigation, and planning. Another interesting application would be creating a 3D map that finds its application in Simultaneous Localisation and Mapping~\cite{mur2015orb, engel2014lsd}; computing depth helps us back-project images from different views into 3D as shown in Figure~\ref{fig:3d-recon-image}. The scene can then be restructured by registering and matching all of the points.\par
\subsubsection{How do we Perceive our World?}

\begin{figure}[!t]
  \centering
    \includegraphics[width=\textwidth]{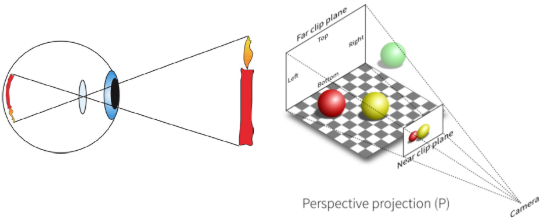}
    \caption[\bf Projecting onto the retina (left). Projecting onto the image plane (right).]{\bf Projecting onto the retina (left). Projecting onto the image plane (right)~\cite{depth_basics_2021}.}
    \label{fig:camera-plane}
\end{figure}

Let us begin by discussing how humans view distance in general. Since many of these approaches were derived from our human vision system, this will provide us valuable insights into depth estimation. The formation of an image is similar in both computer and human vision, as shown in Figure~\ref{fig:camera-plane}. In theory, as light rays from a source strike objects, they bounce off and move towards the back of our retina, where they are projected and processed in 2D~\cite{science_of_vision_2021}, similar to how an image is projected on an image plane. So, how do we calculate distance and comprehend our surroundings in 3D when the predicted scene is in 2D? For example, assume someone is about to punch; we would instinctively know when we are about to be hit and dodge it when his/her fist gets too close. Alternatively, when driving a car, we might somehow gauge when to step on the accelerator or hit the brakes to maintain a safe distance from so many other drivers and pedestrians~\cite{depth_basics_2021}. The mechanism at work here is that our brain begins to reason about the incoming visual signals by identifying patterns such as scale, texture, and motion in the scene referred to as \emph{Depth Cues}. There is no distance information in the image, but we can easily interpret and recover depth information. We can tell which parts of the scene are closer to us and which are farther away. Furthermore, these cues allow us to view objects and surfaces that are ostensibly flat on 2D images as 3D~\cite{science_of_vision_2021}.\par
\begin{figure}[!t]
  \centering
    \includegraphics[width=\textwidth]{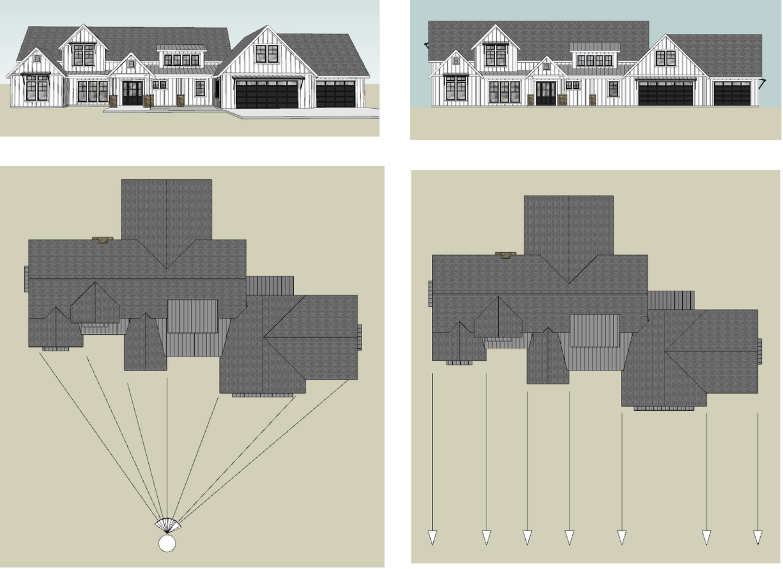}
    \caption[\bf Perspective Projection (Left). Orthographic Projection (Right)]{\bf Perspective Projection (Left). Orthographic Projection (Right). Figure reproduced from~\cite{perspective_othographic_2021}.}
    \label{fig:perspective-orthographic}
\end{figure}
\subsubsection{How to destroy depth from perspectives for human/computer vision?}

Depth ambiguity: Understanding depth cues starts with understanding how scenes are projected to perspective view in human and camera vision. An orthographic projection to front view or side view, on the other hand, loses all depth details. Let us consider Figure~\ref{fig:perspective-orthographic}, where an observer can tell which side of the house is closer to him/her, as seen in the left image. However, from the right image, it is impossible to discern relative distances. The background may also be on the same plane as the home. \par
\subsubsection{Inferring Depth Using Cues}

\begin{figure}[!t]
  \centering
    \includegraphics[width=\textwidth]{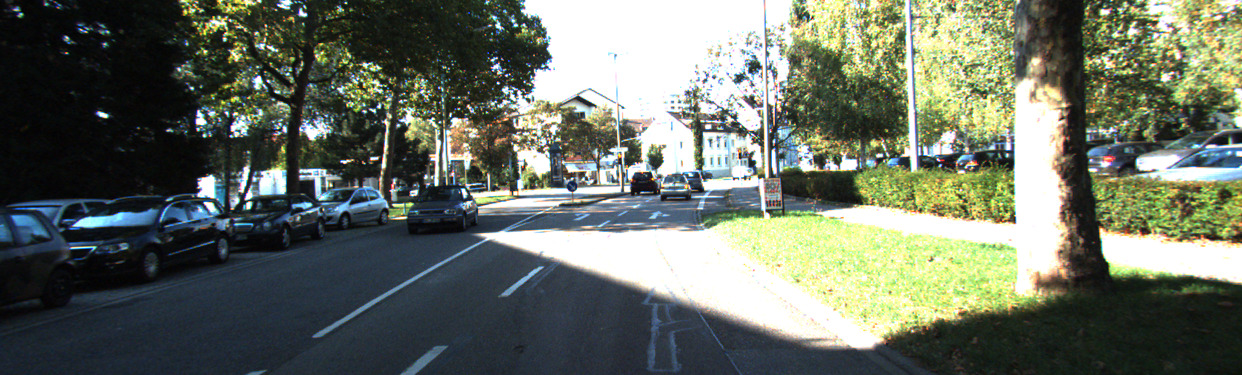}
    \caption[\bf A sample image from KITTI for the illustration of depth cues.]{\bf A sample image from KITTI~\cite{geiger2013vision} for the illustration of depth cues.}
    \label{fig:kitti-sample}
\end{figure}

Basically, there are four types of depth hints: static monocular, motion depth, binocular and physiological hints~\cite{kooi2004visual}. We use these signals subconsciously to perceive depth remarkably well. Our perception of depth from a single image mainly depends on the spatial arrangement of a scene. In Table~\ref{tab:depth-cues}, we summarize some cues that enable us to understand the distance between various objects. From our every day, it can already feel normal to us. Jonathan~\etal~\cite{gardner2010vertical} experimentally illustrates that when the horizon is visible, there is an overwhelming tendency for humans to exploit this cue to perceive depth quickly. By looking at the image in Figure~\ref{fig:kitti-sample}, we can summarize the following cues depicted in Table~\ref{tab:depth-cues}.
\begin{table}[t]
\centering
\begin{tabular}{@{}l|lll@{}}
\toprule
  \multicolumn{1}{c}{\textit{\begin{tabular}[c]{@{}c@{}}Monoscopic \\ Depth Cues\end{tabular}}} &
  \multicolumn{1}{c}{\cellcolor[HTML]{00b050} \textit{Examples}} &
  \multicolumn{1}{c}{\cellcolor[HTML]{7d9ebf}\textit{\begin{tabular}[c]{@{}c@{}}Appear\\ Nearer\end{tabular}}} &
  \multicolumn{1}{c}{\cellcolor[HTML]{e8715b}\textit{\begin{tabular}[c]{@{}c@{}}Appear\\ Farther\end{tabular}}} \\
\midrule
Size of objects    & Tree        & Larger       & Smaller  \\
Texture            & Grass patch & High Quality & Low Quality \& Blurry \\
Linear Perspective & Curb        & \xm          & Converge to Horizon \\
\bottomrule
\end{tabular}
\caption{\bf Analysis of Monoscopic Depth Cues on a sample from KITTI.}
\label{tab:depth-cues}
\end{table}
\begin{figure}[!t]
  \centering
    \includegraphics[width=0.6\textwidth]{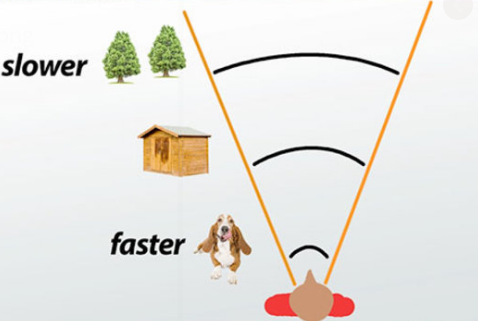}
    \caption[\bf Illustration of motion parallax.]{\bf Illustration of motion parallax. Figure reproduced from~\cite{motion_parallax_basics}.}
    \label{fig:motion-parallax}
\end{figure}
\subsubsection{Depth Cues from Motion (Motion Parallax)}

Motion parallax is a type of depth perception cue in which closer objects appear to move faster than further away objects. It is a type of monocular cue, a depth perception cue that can only be perceived with one eye. This is in contrast to binocular cues, which are depth perception cues that can only be perceived with both eyes open.

Motion parallax occurs when objects at different distances from us appear to move at different rates while we are moving (see Figure~\ref{fig:motion-parallax}). The speed with which an object moves is used to determine its distance. The closer an object is to us, the faster it appears to move. The further an object is from us, the slower it appears to move~\cite{motion_parallax_basics}
\subsubsection{Depth Cues from Stereo Vision (Binocular Parallax)}

\begin{figure}[!t]
  \centering
    \includegraphics[width=0.6\textwidth]{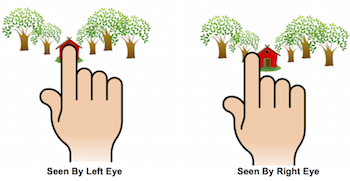}
    \caption[\bf Illustration of Retina disparity.]{\bf Illustration of Retina disparity~\cite{retina_disparity_2021}.}
    \label{fig:retina_disparity}
\end{figure}

\textbf{Retina Disparity:} Let us look into another fascinating phenomenon that enables us to understand the depth that can intuitively be grasped by demonstrating a simple experiment. If we place our index finger close to our face with one eye closed. Now, repeatedly if we close one and open the other. We can observe that our finger is moving! \emph{Retina disparity} refers to the disparity in vision between our left and right eyes. If we stick out our finger at arm's length and repeat the process, we should note that the shift in finger position becomes less noticeable as shown in~\ref{fig:retina_disparity}. This experiment provides us with some insight into how stereo vision works. \emph{Stereopsis} is the ability to perceive depth due to two different views of the world. The brain computes distance by comparing images from the retinas of the two eyes. The greater the disparity, the closer things are to us~\cite{depth_basics_2021}.
\subsubsection{Depth Estimation in Computer Vision}

Depth estimation aims to recover the three-dimensional structure and appearance of objects in imagery by obtaining a representation of the spatial structure of a scene. This is also known as the inverse problem~\cite{szeliski2010computer}, in which we attempt to recover certain unknowns despite the fact that there is insufficient knowledge to define the solution completely, \ie, the mapping between the 2D and 3D views is not unique. We discuss this issue in detail in the following sections. So, how do machines perceive depth? Can any of the ideas discussed above be transferred in some way? Back in the 1990s, the first algorithm with promising results was depth estimation using stereo vision. Since then, Dense stereo correspondence algorithms have made significant progress~\cite{okutomi1993multiple, boykov1998variable, birchfield1999depth}. The approaches involved using geometry to constrain and reproduce the concept of stereopsis mathematically and in real-time. Scharstein~\etal~\cite{scharstein2002taxonomy} summarizes most of these ideas in his survey.\par

Most research either exploits geometrical cues such as multi-view geometry or epipolar geometry~\cite{zhang1995robust} to learn depth. Monocular depth estimation has recently gained popularity due to the use of neural networks to learn a representation that distills depth directly~\cite{Eigen_14}. Accordingly, gradient-based approaches are used to learn depth cues implicitly. Aside from that, there has been significant progress in self-supervised depth estimation~\cite{garg2016unsupervised, zhou2017unsupervised, godard2019digging}, which is particularly exciting and revolutionary due to its wide trainability on arbitrary videos. In this approach, we train a model to predict depth by means of optimizing a proxy signal. In the training phase, we do not require any ground truth label and applicability on single images during inference. \par
\subsection{Object Detection}
\label{sec:object-detection-basics}

\begin{figure*}[t]
  \centering
    \includegraphics[width=0.49\columnwidth]{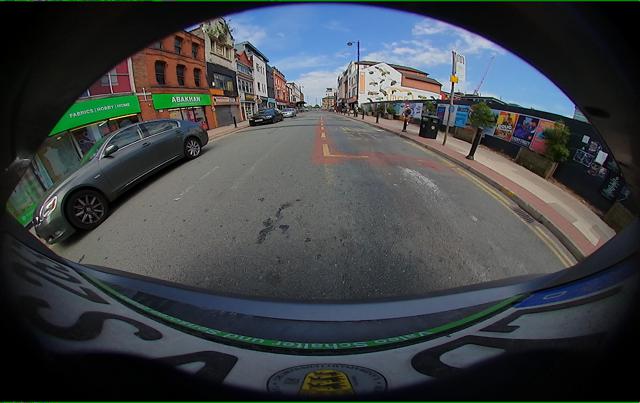}
    \includegraphics[width=0.49\columnwidth]{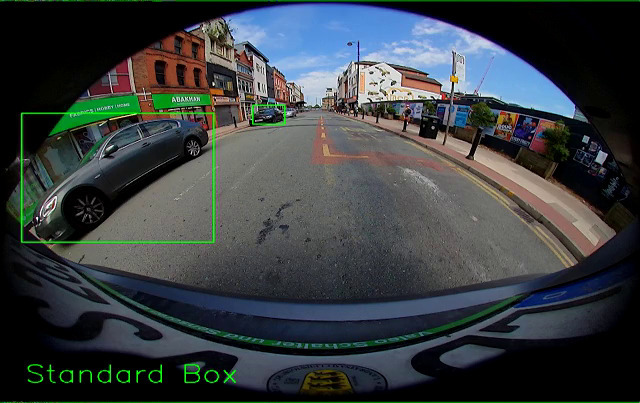} \\
    \caption[\bf A qualitative input sample (left) and object detection prediction (right) on WoodScape.]{\bf A qualitative input sample (left) and object detection prediction (right) on WoodScape \cite{yogamani2019woodscape}.}
    \label{fig:object-detection-sample}
\end{figure*}

Before we get started on creating a cutting-edge model, let us first define object detection. Let us pretend we are building a vehicle detection system for a self-driving car. Assume our car captures an image similar to the one in Figure~\ref{fig:object-detection-sample}. The image from the rear camera essentially conveys that it is a one-way street, and there is a car right next to our rear-end, and quick maneuvers to the right must be prohibited. We should slow down and allow the car to pass through if we need to steer right.\par

\textbf{So, what will the car's system do to ensure that this maneuver can occur safely?}

It can draw a bounding box around the cars so that the algorithm can determine where the cars are in the image and then decide which direction to proceed to prevent any mishaps.\par

We aim to perform object detection mainly:
\begin{itemize}[nosep]
    \item To detect where the objects are present in the image and locate them.
    \item To filter out the objects of interest.
\end{itemize}

\emph{Object detection} is therefore defined as a model that entails categorizing and localizing various objects in an input image based on their location (see Figure~\ref{fig:object-detection-sample}). It detects the presence of an object in an image and draws a box around it. This typically entails two processes: classifying an object's form and then drawing a box around that object.
Object detection tasks are efficiently solved using fully convolutional neural networks. CNN-based bounding box detection can be divided into two categories: single-stage and two-stage approaches. Single-stage methods regress box coordinates and class categories in a single pass. YOLO~\cite{redmon2016you} and SSD~\cite{liu2016ssd} are early adopters of single-stage approaches. On the other hand, two-stage networks use explicit loss functions for class-agnostic area proposals accompanied by accurate box coordinates regression. This group includes the R-CNN family of algorithms~\cite{ren2017faster}.\par
\subsection{Semantic Segmentation}
\label{sec:semantic-seg-basics}

\begin{figure}[!t]
  \centering
    \includegraphics[width=0.9\textwidth]{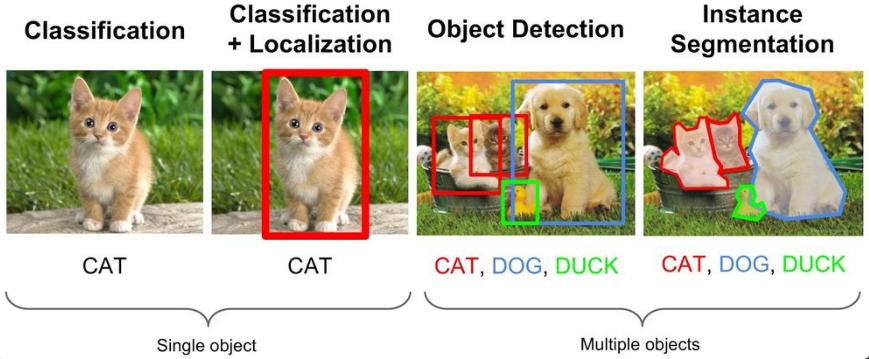}
    \caption[\bf Computer Vision tasks in ascending order of complexity.]{\bf Computer Vision tasks in ascending order of complexity. Figure reproduced from~\cite{order_of_complexity_2021}.}
    \label{fig:order-of-complexity-vision}
\end{figure}

Deep learning has been very effective when working with images as data, and it is now at the point that it outperforms humans in a variety of use-cases. In descending order of complexity, the fundamental problems humans have been involved in solving with computer vision are image classification, object detection, and segmentation, as shown in Figure~\ref{fig:order-of-complexity-vision}. In the task of image classification, we are simply interested in obtaining the labels of all the objects present in an image. Object detection, as discussed in Section~\ref{sec:object-detection-basics} takes it a step further by attempting to detect all objects that are present in an image and the position at which the objects are present using bounding boxes. Image segmentation is even more complex as a task, thus attempting to determine the exact boundary of the objects in the image~\cite{semantic_seg_basics_2021}.\par
\begin{figure*}[t]
  \centering
    \includegraphics[width=0.49\columnwidth]{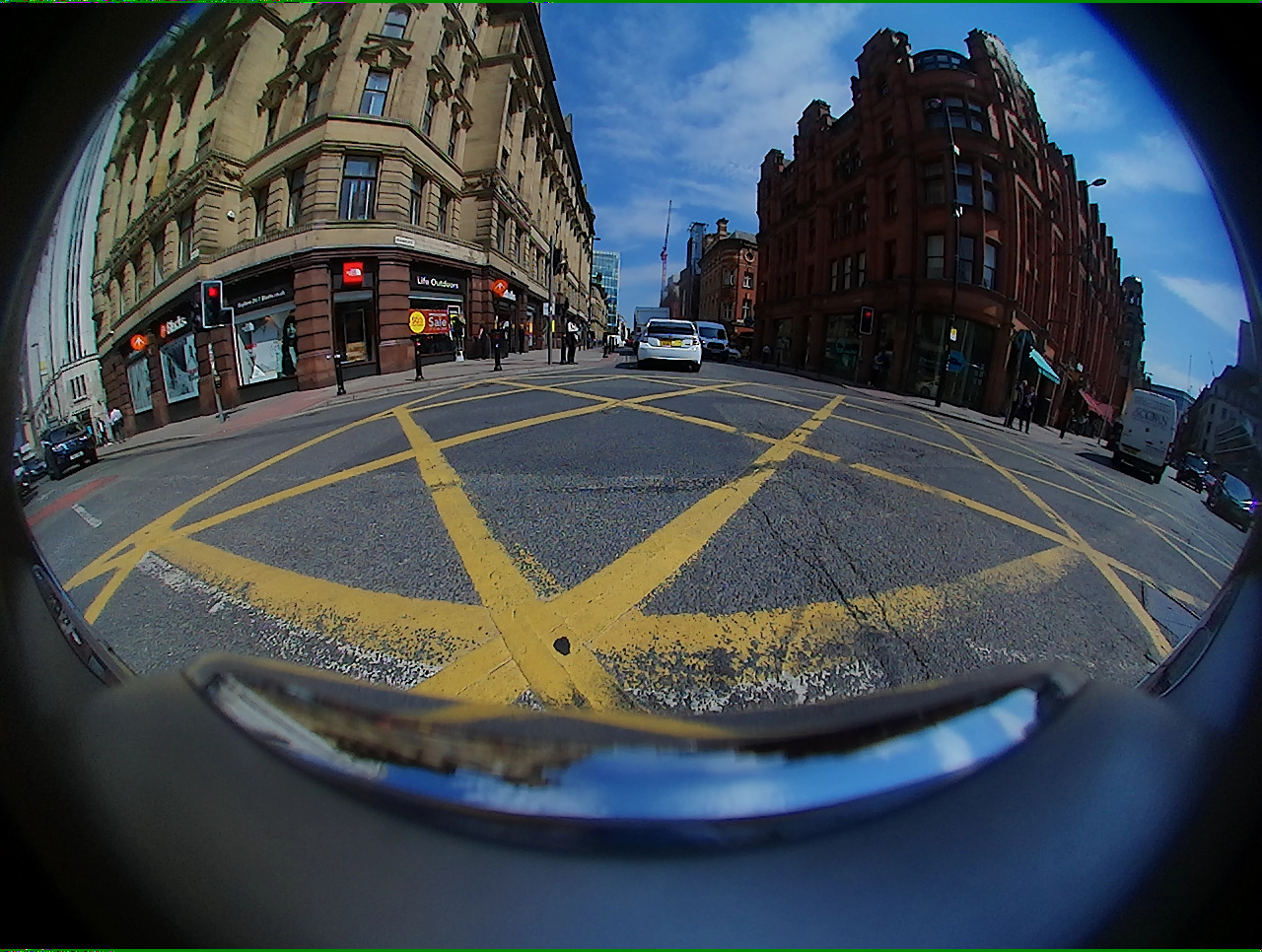}
    \includegraphics[width=0.49\columnwidth]{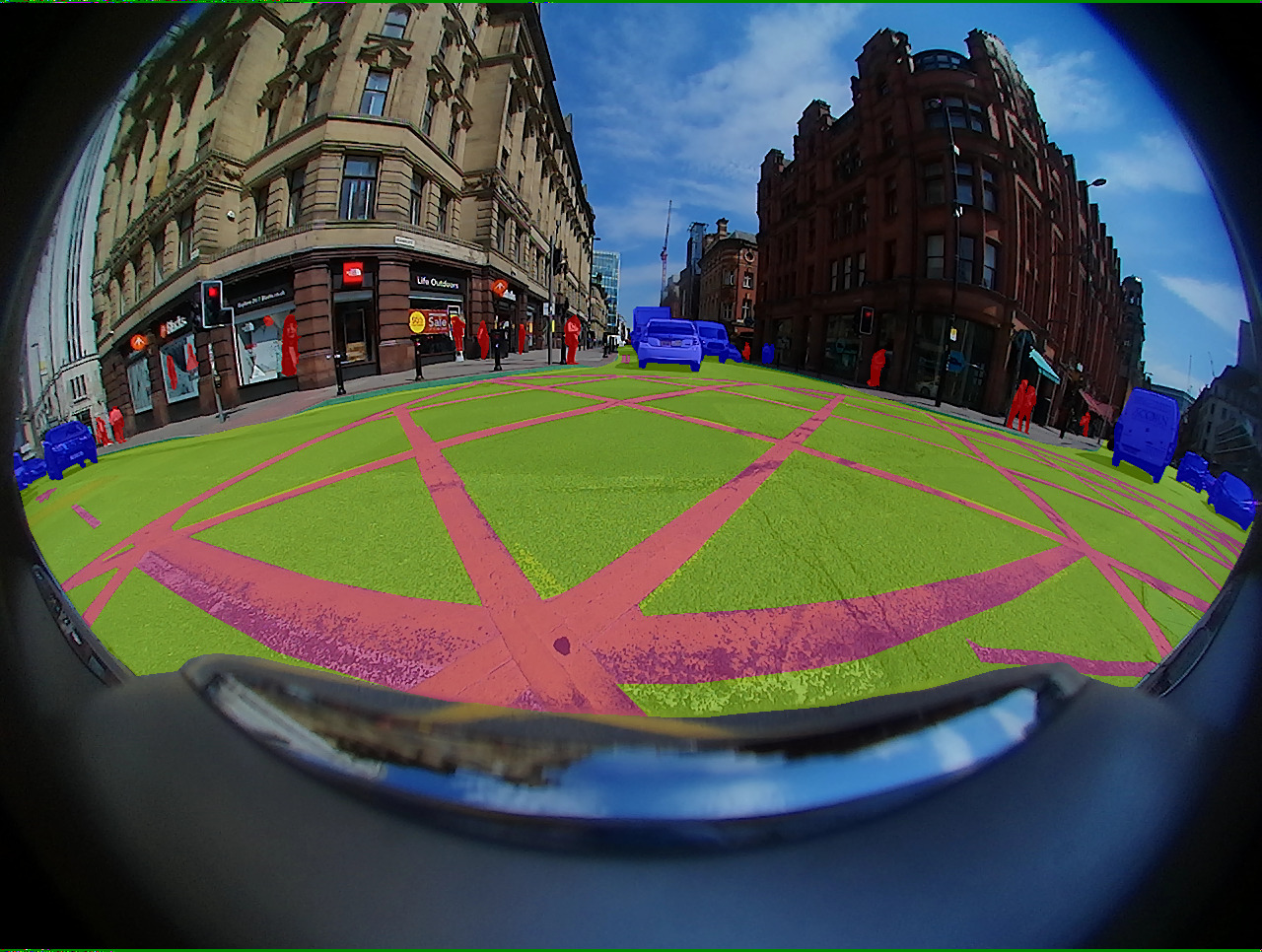} \\
    \caption[\bf A qualitative input sample (left) and semantic segmentation prediction (right) on WoodScape.]{\bf A qualitative input sample (left) and semantic segmentation prediction (right) on WoodScape \cite{yogamani2019woodscape}.}
    \label{fig:semantic-segmentation-sample}
\end{figure*}

\emph{Semantic segmentation} is the process of assigning a class label to each pixel in an image. These labels could refer to a person, road, curb, pole, and so on, as shown in Figure~\ref{fig:semantic-segmentation-sample}. It does not differ across different instances of the same object. For example, if an image contains two cars, semantic segmentation assigns the same label to all of the pixels in both cars. It provides dense pixel-by-pixel labeling of the image, resulting in scene comprehension. Semantic segmentation was once thought to be a difficult task. The development of accurate and efficient approaches was made possible with the help of fully convolutional neural networks (FCNs)~\cite{noh2015learning}. The level of sophistication of semantic segmentation has recently increased rapidly, and so has the computational power of embedded systems, allowing for commercial deployment. Segmentation models are highly useful for autonomous cars as we need to provide cars a detailed perception to enable them to understand their surroundings so that they can safely transition on our existing roads.\par
\subsection{Motion Segmentation}
\label{sec:motion-seg-basics}

\begin{figure}[!t]
	\centering
	\subfigure[Rear Cam (t-6)]
	{\includegraphics[width=0.45\linewidth]
	{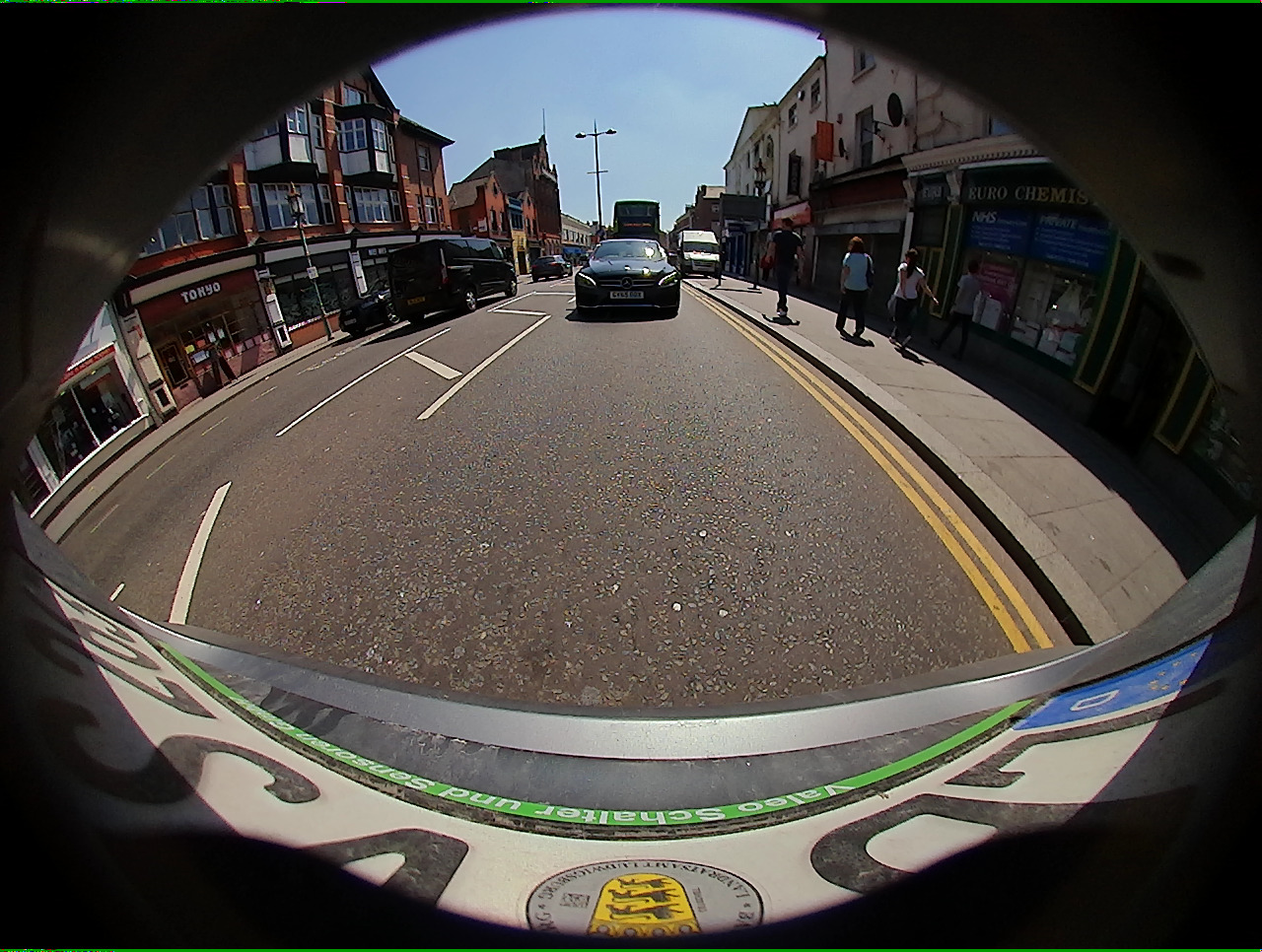}}\;\;
	\subfigure[Rear Cam (t)]
	{\includegraphics[width=0.45\linewidth]
	{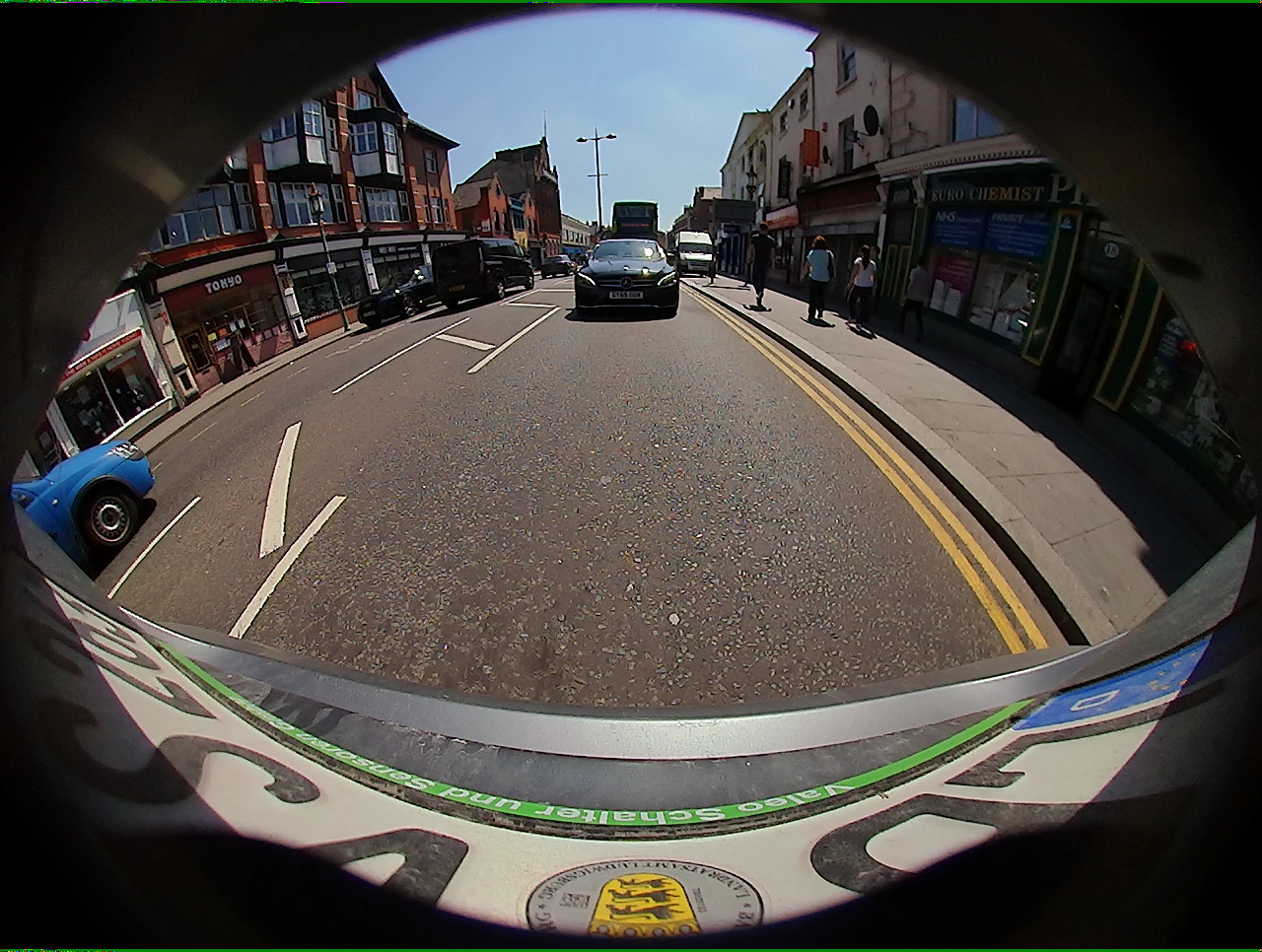}}\\
	\subfigure[Motion Estimate]
	{\includegraphics[width=0.45\linewidth]
	{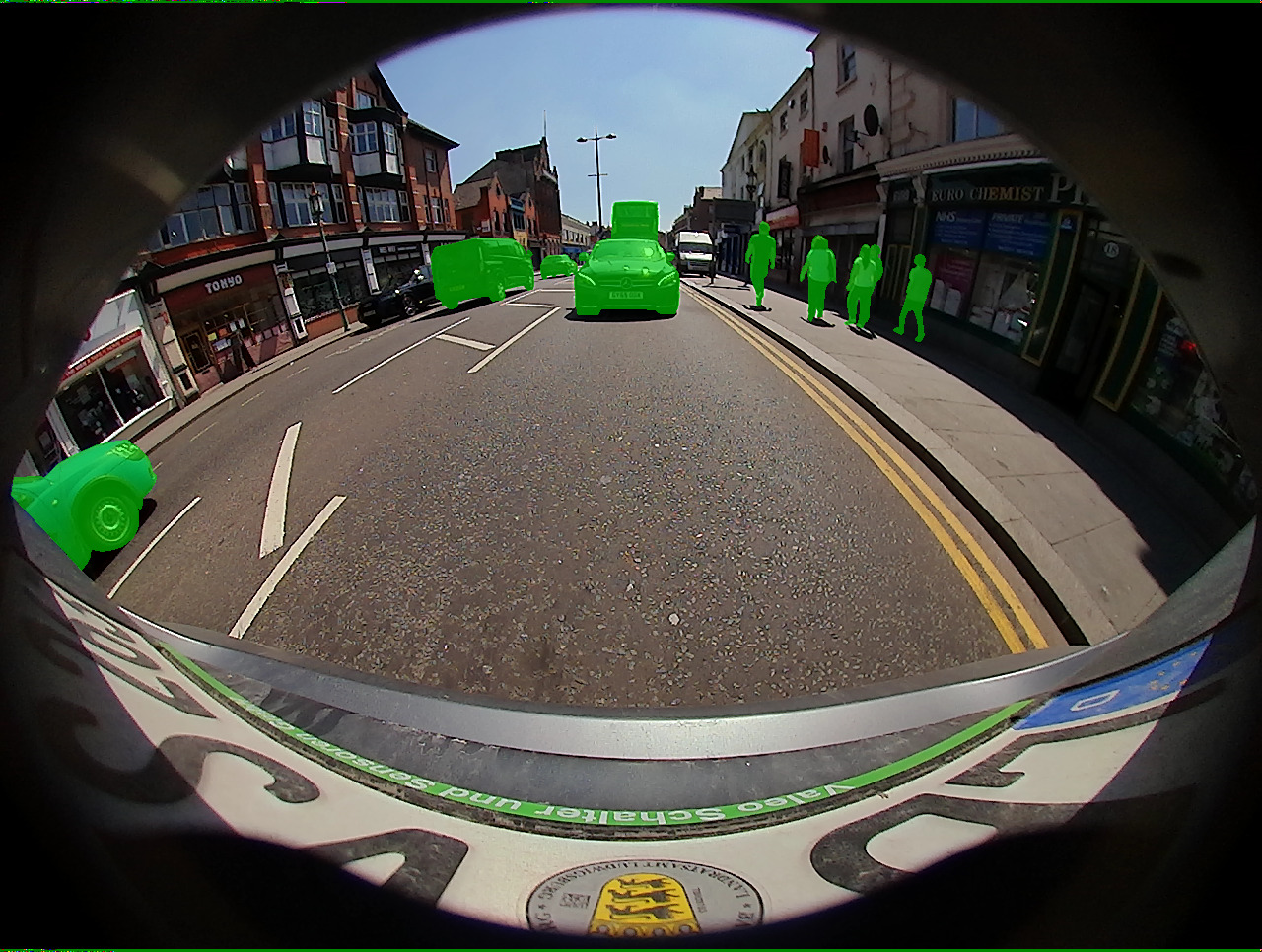}}
	\caption[\bf A qualitative sample of motion segmentation on WoodScape.]
	{\bf A qualitative sample of motion segmentation on WoodScape. (t-6) and (t) frames are showcased to visually spot dynamic objects segmented in the motion estimate.}
	\label{fig:motion-sample}
\end{figure}

\emph{Motion Segmentation} is defined as the task of identifying the independently moving objects (pixels) such as vehicles and persons etc., in a pair of sequences and separating them from the static background as shown in Figure~\ref{fig:motion-sample}. It involves assigning a class label to each pixel in an image to segment static and dynamic objects. This task is treated as a binary segmentation problem. There are two types of motion in an autonomous driving scene. The first one is the motion of the surrounding obstacles and the second is the motion of the ego-vehicle. The ego-motion might cause difficulties to successfully detect the moving objects because even static objects will be perceived as moving. Motion segmentation implies two tasks that are performed jointly. The first one is object detection. We highlight the interesting objects only of specific classes, \ie, pedestrians and vehicles, and discard any motion perceived from the background due to ego-motion. The second is motion classification, in which a binary classifier predicts whether the object is moving or static.\par

Because of the cameras' ego-motion on the moving vehicle, motion is a powerful cue in automotive driving, and detecting dynamic objects around the car is vital. Furthermore, it aids in detecting generic objects based on motion cues rather than appearance cues since there would still be unusual objects such as kangaroos or construction vehicles. The autonomous driving scenes are highly dynamic, where there are many moving objects interacting with each other forming a very complex environment to deal with. The detection and localization of moving obstacles are crucial for ADAS and autonomous vehicles. They are essential for emergency braking, support decision-making for its next step navigation, and for avoiding possible collisions~\cite{heimberger2017computer}. An autonomous vehicle has to estimate collision risk with other interacting objects in the environment and calculate an optional trajectory. Collision risk is typically higher for moving objects than static ones due to the need to estimate the future states and poses of the objects for decision making. This is particularly important for near-range objects around the vehicle, which are typically detected by a fisheye surround-view system that captures a 360$\degree$ view of the scene.\par
From a static observation point, the detection of moving obstacles is almost trivial as any non-zero optical flow will be due to motion in the scene or noise in the image. For a moving observer, the problem is challenging as the entire scene relative to the camera moves. It is also complicated when considering fisheye cameras, which exhibit complex motion patterns due to the non-linear projection and strong lens distortion. Fewer classes are movable, and this can be leveraged to improve classification accuracy. For example, object classes such as buildings or poles are static and will not have dominant motion vectors after ego-motion compensation.\par
\subsection{Soiling Segmentation}

\textbf{Level $3$} autonomous driving~\cite{sae2014taxonomy} stands out as a challenging goal of a large part of the computer vision and machine learning community. Due to this problem's difficulty, a combination of various sensors is necessary to build a safe and robust system. However, apart from the geometric and semantic tasks, there are other less "popular" problems slowly getting into attention, which have to be solved for the ultimate goal of the full \textit{Level $5$} autonomy. Cameras are an essential part of the sensor suite to achieve \textit{Level $3$} autonomous driving. Surround-view cameras are, however, directly exposed to the external environment and are vulnerable to get soiled as a consequence of bad weather conditions such as rain, fog, or snow~\cite{uvrivcavr2019soilingnet, uricar2019desoiling}. Furthermore, dust and mud have a significant impact on vision tasks' performance. Cameras have a much higher degradation in performance due to soiling compared to other sensors. Thus it is critical to accurately detect soiling on the cameras, particularly for higher autonomous driving levels. As the visual perception modules for autonomous driving are becoming more mature, there is much recent effort to make them more robust to adverse weather conditions. It can be seen by the popular CVPR workshop "Vision for all seasons''\footnote{\url{https://vision4allseasons.net/}}, which focuses on the performance of computer vision algorithms during adverse weather conditions for autonomous driving.\par
\begin{figure}[!t]
    \centering
    \includegraphics[width=\textwidth]{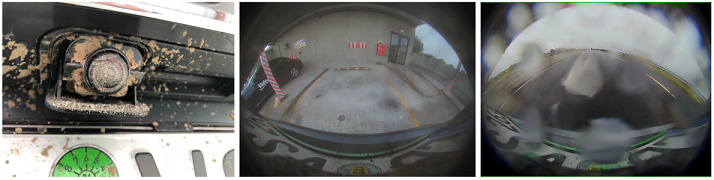}
    \caption[\bf Automotive surround-view cameras are exposed to harsh environmental setup.]{\textbf{Automotive surround-view cameras are exposed to harsh environmental setup.} Left: camera lens covered by mud. Middle: image produced by the soiled camera from the left picture. Right: camera lens soiled during heavy rain.}
    \label{fig:soiling_setup}
\end{figure}
\begin{figure}[!t]
    \centering
    \includegraphics[width=0.7\textwidth]{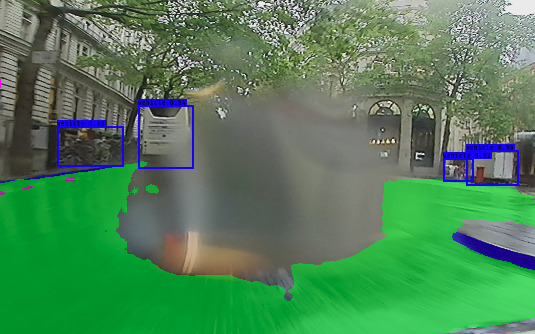}
    \caption[\bf Example of semi-transparent soiling in the form of a water drop on the camera lens.]
    {\textbf{Example of semi-transparent soiling in the form of a water drop on the camera lens.} The detection of the bus behind the water drop works still well, while the road segmentation (green) is highly degraded in the soiled region. In this scenario, a soiling detection algorithm is used to trigger a camera cleaning system that restores the lens hardware.}
    \label{fig:soiling_example_drop}
\end{figure}

One of these problems is the reliability of the sensory signal, which in the case of surround-view cameras means, \textit{inter alia}, the ability to detect soiling on the camera lens. Failure to recognize severe weather conditions leading to a deterioration of the image quality to such a level that any further image processing is unreliable~\cite{uricar2021let}. Figure~\ref{fig:soiling_setup} shows how the surround-view camera can get soiled and the corresponding image output, as well as an example of images taken during a~heavy rain. It usually happens when the tire splashes mud or water from the road or wind, depositing dust on the lens. Figure~\ref{fig:soiling_example_drop} shows an example of the strong impact of a significant water drop on the camera lens for object detection and semantic segmentation tasks.\par
\section{Multi-Task Learning (MTL)}

\begin{figure}[t]
  \centering
    \includegraphics[width=0.8\textwidth]{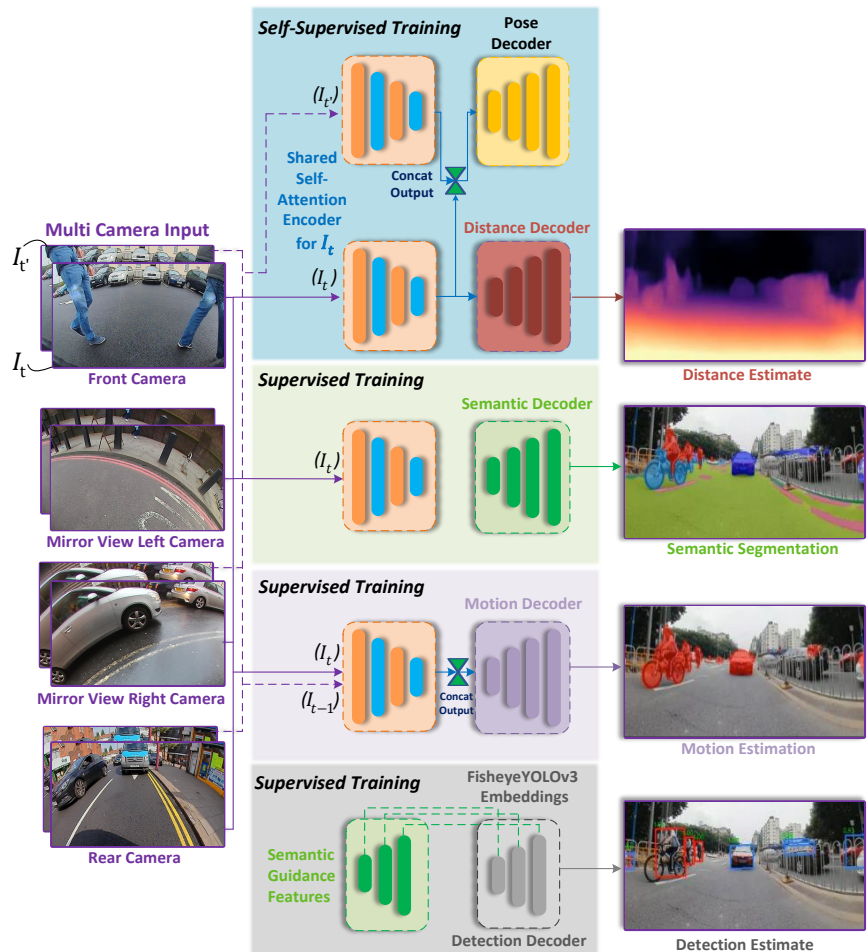}
    \caption[\bf Illustration of a multi-task architecture comprising of four tasks.]{\bf Illustration of a multi-task architecture comprising of four tasks~\cite{kumar2021omnidet}.}
    \label{fig:mtl_pipeline_high_level}
\end{figure}

\begin{figure}[!t]
  \resizebox{\columnwidth}{!}{
  \centering
\begin{tabular}{ccc}
 	\begin{overpic}[width=0.7\columnwidth]{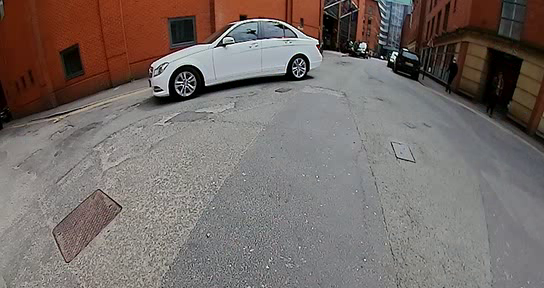}
    \put (0,2) {\colorbox{lightgray}{$\displaystyle\textcolor{black}{\text{(a)}}$}}
    \end{overpic}
    \begin{overpic}[width=0.7\columnwidth]{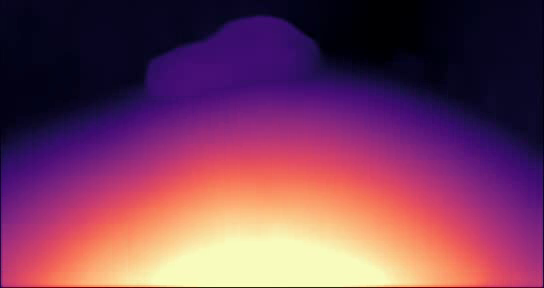}
    \put (0,2) {\colorbox{lightgray}{$\displaystyle\textcolor{black}{\text{(b)}}$}}
    \end{overpic} \\
    
    \begin{overpic}[width=0.7\columnwidth]{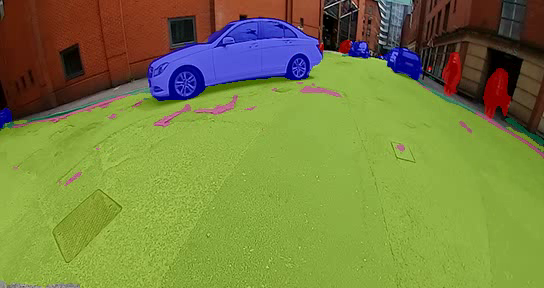}
    \put (0,2) {\colorbox{lightgray}{$\displaystyle\textcolor{black}{\text{(c)}}$}}
    \end{overpic} 
 	\begin{overpic}[width=0.7\columnwidth]{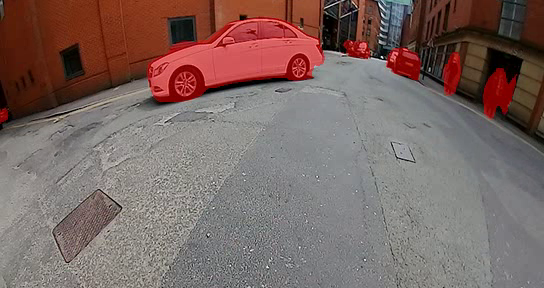}
    \put (0,2) {\colorbox{lightgray}{$\displaystyle\textcolor{black}{\text{(d)}}$}}
    \end{overpic} \\
    
    \begin{overpic}[width=0.7\columnwidth]{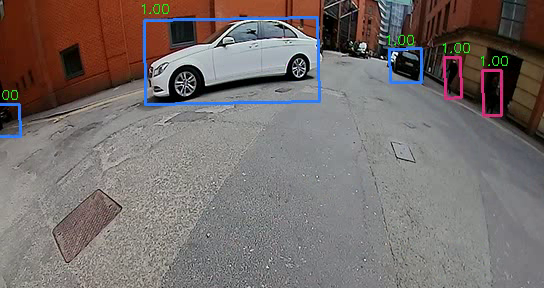}
    \put (0,2) {\colorbox{lightgray}{$\displaystyle\textcolor{black}{\text{(e)}}$}}
    \end{overpic}
    \begin{overpic}[width=0.7\columnwidth]{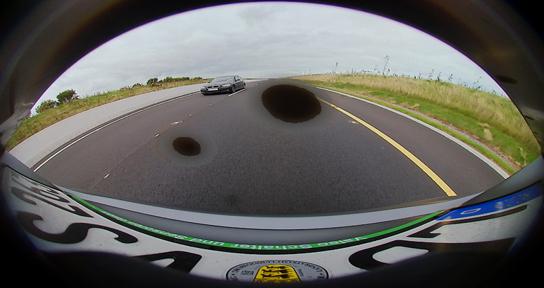}
    \put (0,2) {\colorbox{lightgray}{$\displaystyle\textcolor{black}{\text{(f)}}$}}
    \end{overpic}
\end{tabular}
}
\caption[\bf Sample Multi Task Learning perception output on raw fisheye images.]
{\textbf{Sample Multi Task Learning perception output on raw fisheye images.} (a) Rear-Camera Input Image, (b) Distance Estimate, (c) Semantic Segmentation, (d) Motion Estimation, (e) Standard Object Detection and (f) Soiling Segmentation.}
\label{fig:mtl_poster}
\end{figure}

We usually care about optimizing a specific metric in deep learning, such as a score on a specific benchmark or a business key performance indicator. To accomplish this, we usually train a single model or a group of models to perform our desired task. The models are then fine-tuned and tweaked until their output no longer improves. While we can typically produce satisfactory results, we neglect details that might help us do even better on the metric we care about by being highly focused on a single task. We can derive knowledge explicitly from the training signals of related tasks. We may improve our model's generalization on our original task by sharing representations between similar tasks. This method is known as \emph{Multi-Task Learning} (MTL)~\cite{mtl_basics_2021}. It is a sub-field of machine learning in which a shared model concurrently learns multiple tasks at once, as shown in Figure~\ref{fig:mtl_pipeline_high_level}. The Figure~\ref{fig:mtl_poster} illustrates the output from OmniDet~\cite{kumar2021omnidet} MTL framework.\par
In general, when we find ourselves optimizing more than one loss function, we are engaging in MTL (in contrast to single-task learning). In such cases, it is beneficial to think about what we are attempting to do clearly in MTL and gain conclusions from it. Even if we are just optimizing one loss, as is usually the case, there is likely to be an auxiliary task that will assist us in improving our main task. MTL's target is succinctly stated by Rich Caruana~\cite{caruana1998multitask}: "MTL enhances generalization by exploiting the domain-specific knowledge found in the training signals of similar tasks." MTL aims to solve computational bottlenecks in CNNs and increase computational performance by sharing the costly layers across all tasks.\par
\subsection{Motivation}

We may motivate MTL in various ways: It can be seen as biologically inspired by human learning. We often apply information gained from learning similar tasks while learning new tasks. An infant, for example, learns to recognize faces first and then applies this information to recognize other objects. From the standpoint of machine learning, we can inspire MTL by seeing it as a type of inductive transfer. Inductive transfer can help develop a model by adding an inductive bias that causes the model to favor some hypotheses over others. \lone regularization, for example, is a common type of inductive bias that results in a preference for sparse solutions. In MTL, the auxiliary tasks provide the inductive bias, causing the model to favor hypotheses that describe more than one task~\cite{mtl_basics_2021}.\par

From the automotive perspective in this thesis, our goal is to build a perception system
that constitutes a \textbf{Level 3} autonomous stack. Deploying an efficient multi-task model has several advantages over all the earlier discussed single-task models. We could attain significant gain in embedded performance, certification, validation, testing. Also, the deployment of a single MTL model is easier than several single-task learning (STL) models.\par
\subsection{Two MTL approaches for Deep Learning}
\label{sec:hard-soft-parameters-mtl}

\begin{figure}[!t]
	\centering
	\subfigure[][Hard parameter sharing]{\includegraphics[width=0.48\linewidth]{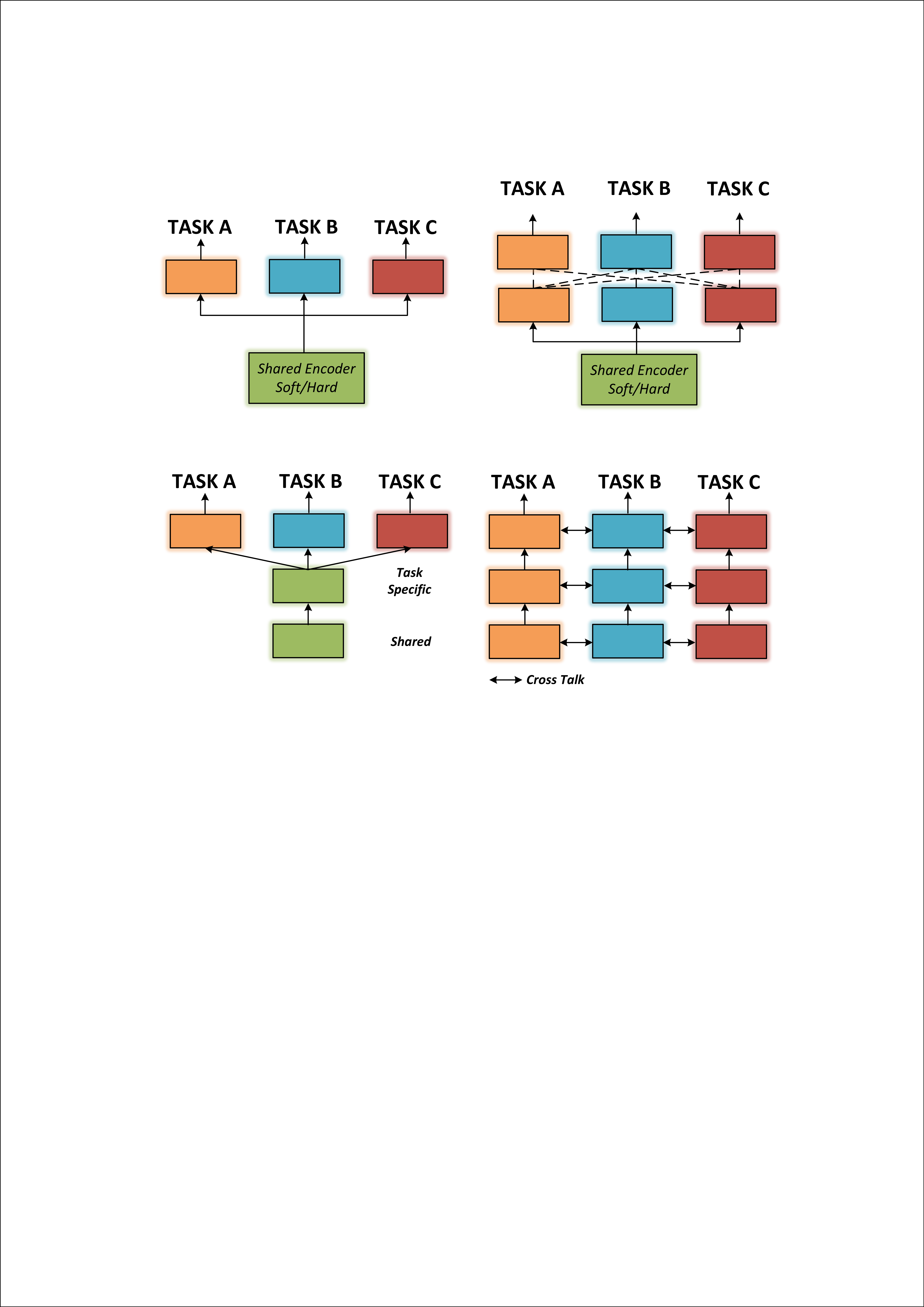}}\;\;
	\subfigure[][Soft parameter sharing]{\includegraphics[width=0.48\linewidth]{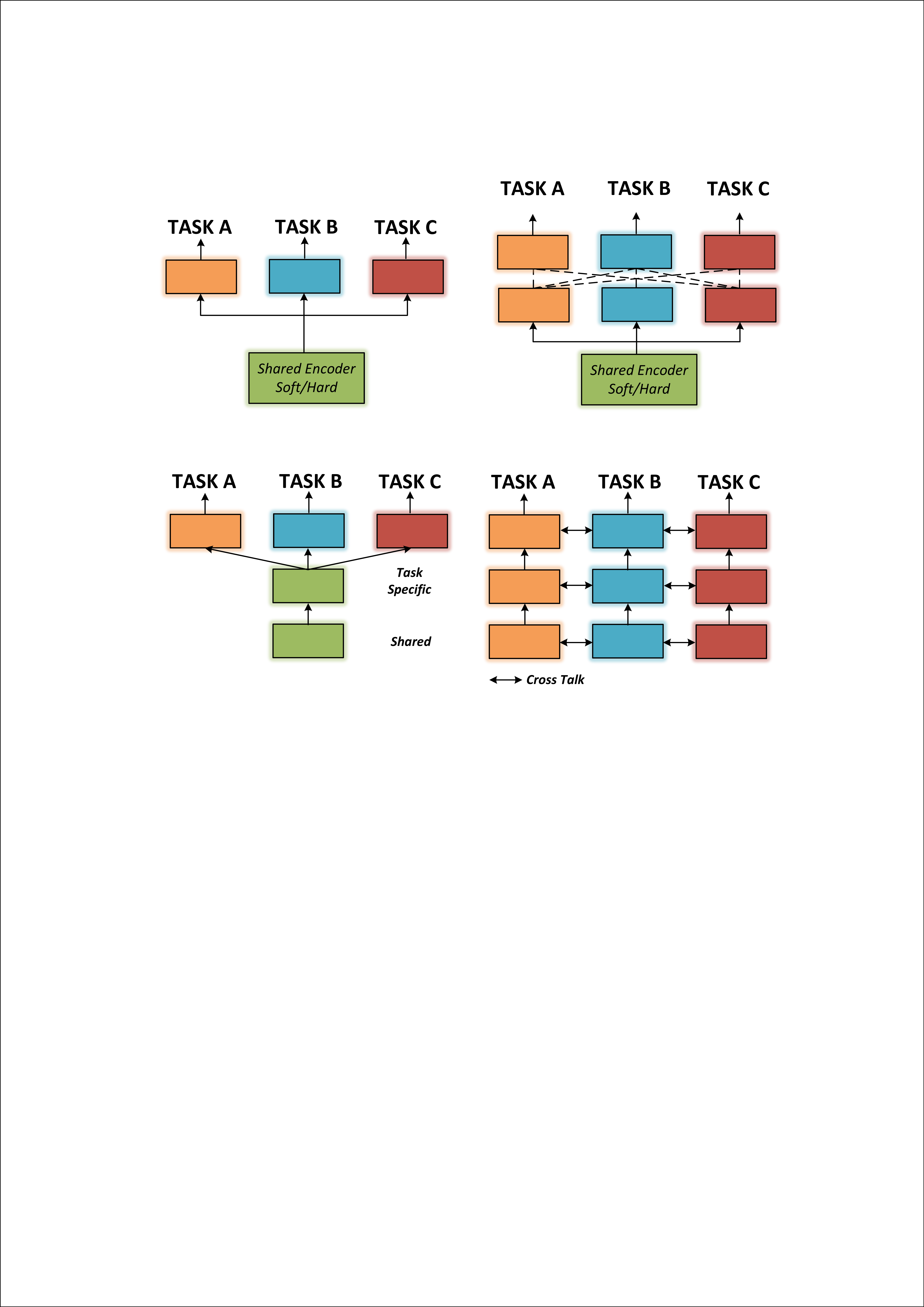}}
	\caption{\textbf{Multi-task learning employing neural networks has been divided into soft and hard parameter sharing schemes.}}
	\label{fig:mtl_soft_hard}
\end{figure}

So far, we have concentrated on theoretical motives for MTL. To put MTL concepts into context, we consider the two most popular approaches for performing MTL in Deep Neural Networks (DNNs). Initially, MTL architectures were classified into \emph{hard} or \emph{soft} parameter sharing methods. We illustrate the hard parameter sharing in Figure~\ref{fig:mtl_soft_hard} on the left, wherein the parameter set is divided into shared and task-specific parameters. The most popular approach to MTL in neural networks is hard parameter sharing, which goes back to~\cite{caruana1993multitask}. MTL networks employing hard parameter sharing usually comprise a shared encoder that branches out into task-specific heads~\cite{Kendall2018, sistu2019real, chen2018gradnorm, sener2018multi, neven2017fast, teichmann2018multinet, leang2020dynamic, chennupati2019auxnet}. Overfitting is significantly reduced by hard parameter sharing. Baxter~\etal~\cite{baxter1997bayesian} demonstrated that the risk of overfitting the shared parameters is an order N smaller than the risk of overfitting the task-specific parameters, \ie, the output layers. This makes intuitive sense: the more tasks we learn simultaneously, the more work the model, has to put into finding a representation that captures all of the tasks, and the less risk we have of overfitting on our original task~\cite{mtl_basics_2021}. The first hard parameter sharing model was introduced by UberNet~\cite{kokkinos2017ubernet} wherein they tried to jointly tackle a large number of low, mid, and high-level perception tasks.\par

On the other hand, in soft parameter sharing, each task has its own model with its own set of parameters, and a feature sharing mechanism handles the cross-task talk (see Figure~\ref{fig:mtl_soft_hard} (right)). The distance between the model's parameters is then regularized to allow the parameters to be similar. For example, Duong~\etal~\cite{duong2015low} employs the \ltwo norm for regularization, whereas Yang~\etal~\cite{yang2017trace} employs the trace norm. Cross-stitch networks~\cite{misra2016cross} proposed soft-parameter sharing in MTL architectures~\cite{mtl_basics_2021}.\par
\section{Adversarial Attacks}

\begin{figure}[t]
  \centering
    \includegraphics[width=\columnwidth]{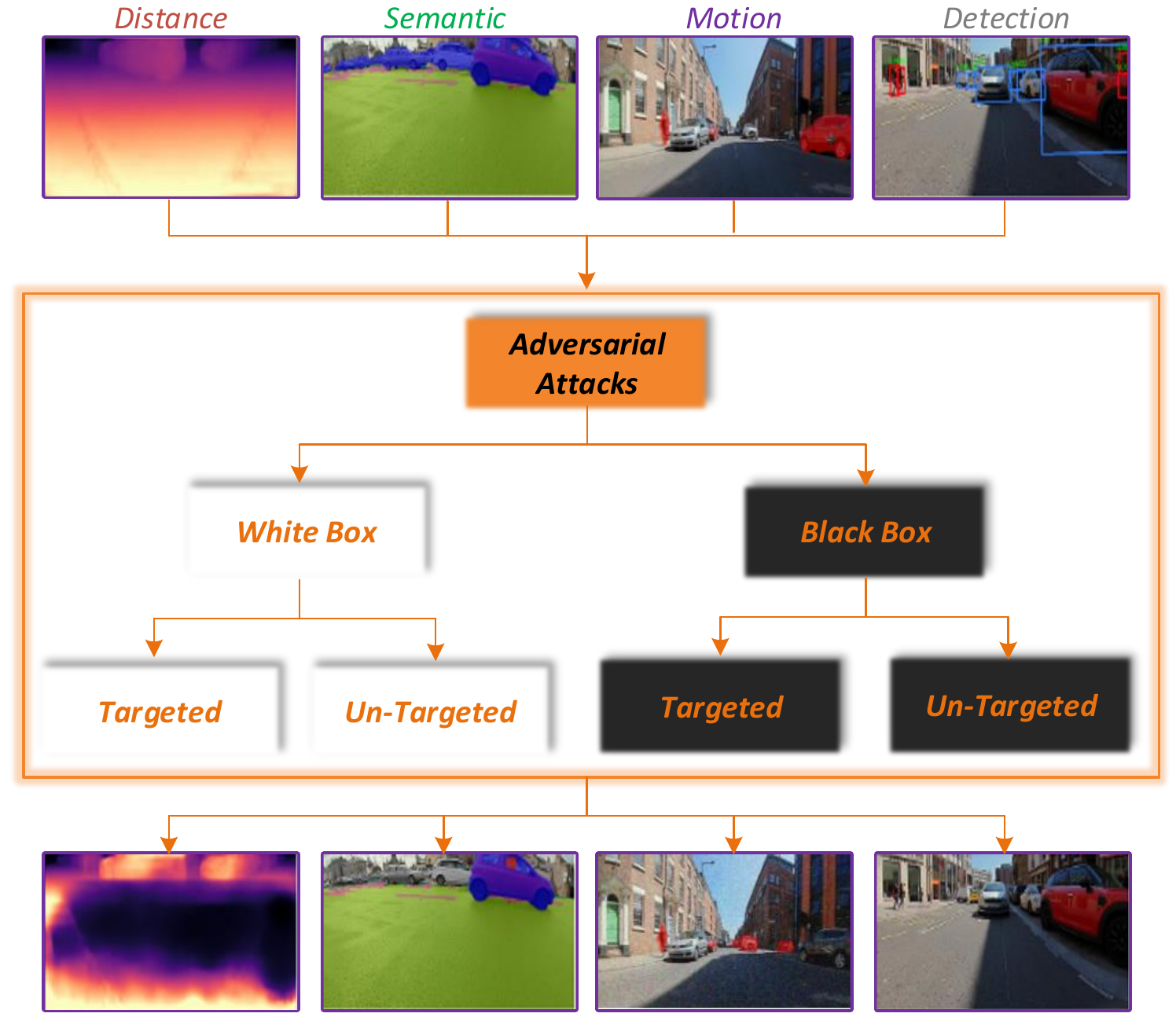}
    \caption[\bf Adversarial attacks on the OmniDet MTL model.]{\textbf{Adversarial attacks on the OmniDet \cite{kumar2021omnidet} MTL model.} Distance, segmentation, motion and detection perception tasks are attacked by white and black box methods with targeted and un-targeted objectives, resulting in incorrect model predictions.}
    \label{fig:omnidet_teaser}
\end{figure}

Autonomous Vehicles are expected to significantly reduce accidents~\cite{avsafetywho}, where visual perception systems are in the heart of these vehicles. Despite the notable achievements of DNNs in visual perception, we can easily fool the networks by adversarial examples that are imperceptible to the human eye but cause the network to fail. Hence, it is important to understand the robustness of MTL models over these attacks. Adversarial examples are usually created by deliberately employing imperceptibly small perturbations to the clean inputs, resulting in incorrect model outputs. This small perturbation is progressively amplified by a deep neural network (DNN) and usually yields wrong predictions. Generally speaking, attacks can be a white box or black box (see Figure~\ref{fig:omnidet_teaser}) depending on the adversary's knowledge (the agent who creates an adversarial example). White box attacks presume full knowledge of the targeted model's design, parameters, and, in some cases, training data. Gradients can thus be calculated efficiently in white box attacks using the back-propagation algorithm. In contrast, in Black box attacks, the adversary is unaware of the model parameters and has no access to the gradients. Furthermore, attacks can be targeted or untargeted based on the intention of the adversary. Targeted attacks try to fool the model into a specific predicted output. In contrast, untargeted attacks consider the predicted output irrelevant, and the main goal is to fool the model into any incorrect output.\par
\section{Datasets and Corresponding Benchmarks} 
\label{sec:benchmarks}

In this section, we briefly discuss the datasets employed in this thesis for the experiments and the metrics used for the tasks. 

\subsection{WoodScape}

\begin{figure}[t]
  \centering
    \includegraphics[width=\columnwidth]{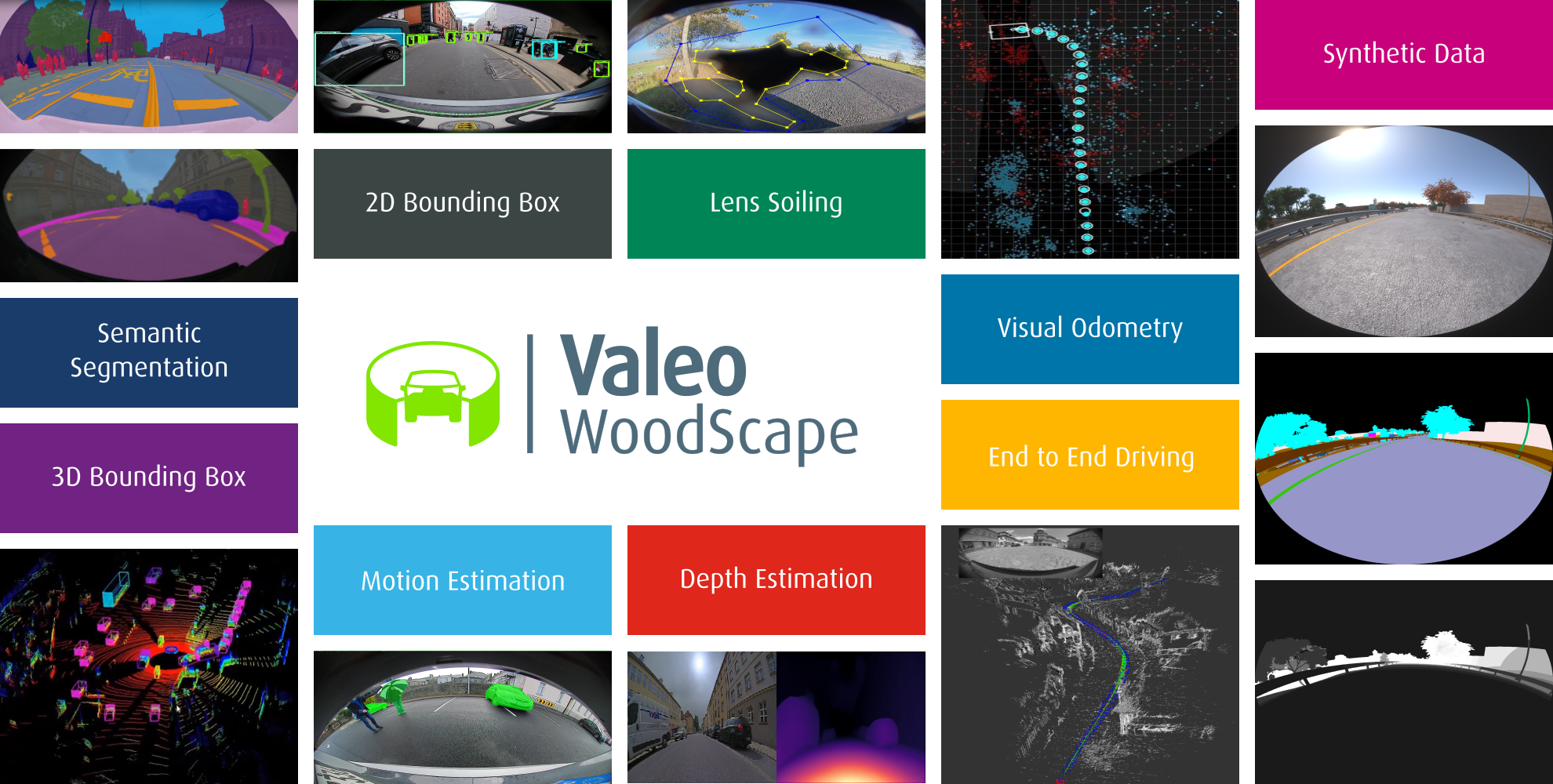}
    \caption{\bf Overview of the perception tasks in the WoodScape dataset.}
    \label{fig:woodscape_dataset}
\end{figure}

The dataset consists of 46,000 images sampled roughly equally from the four views and split into training, validation, and test in a 6:1:3 ratio. This dataset is used for OmniDet~\cite{kumar2021omnidet} as explained in Chapter~\ref{Chapter6}. A sub-set of 10,000 images from the dataset will be made public on \textbf{Github}\footnote{\url{https://github.com/valeoai/WoodScape}}. A baseline code is released along with the dataset on \textbf{GitHub} to encourage further research to the community in developing unified perception models for autonomous driving. It contains several perception tasks listed in Figure~\ref{fig:woodscape_dataset}. 2D box detection contains the five most essential categories of objects — \textit{pedestrians, vehicles, riders, traffic signs, and traffic lights}. Vehicles further have sub-classes, namely cars and large vehicles (trucks/buses). The polygon prediction task on raw fisheye is limited to only two classes — \textit{pedestrians and vehicles}. Unlike traffic lights and traffic signs, these categories are non-rigid in nature and quite diverse in appearance, making them suitable for polygon regression. We sample 24 points with high curvature values from each object instance contour for the polygon regression task. Learning these points helps to regress better polygon shapes, as these points at high curvature define the shape of the object contours. Semantic segmentation comprises of 6 classes on \textit{road, lanes, curbs, two-wheeled vehicles, vehicles, and persons}. The images are in RGB format with 1MPx resolution and $190\degree$ horizontal FoV. The dataset is captured in several European countries and the USA. For the experiments, we used only the vehicles' class. Further details about the dataset usage and demo code can be found on the WoodScape website \url{https://woodscape.valeo.com}.\par
\subsubsection{WoodScape -- Bamberg Dataset}
\label{sec:woodscape-bamberg}

The distance estimation dataset used in Chapter~\ref{Chapter3} and Chapter~\ref{Chapter4} for the work FisheyeDistanceNet~\cite{kumar2020fisheyedistancenet}, UnRectDepthNet~\cite{kumar2020unrectdepthnet}, SynDistNet~\cite{kumar2020syndistnet} and SVDistNet~\cite{kumar2021svdistnet} contains roughly $40,000$ raw images obtained with multiple fisheye cameras constituting a surround-view system and point clouds from a sparse Velodyne HDL-64E rotating 3D laser scanner as ground truth for the test set. The training set contains $39,038$ images collected by driving around various parts of Bavaria, Germany. The validation and the test split contain $1,214$ and $697$ images, respectively. The dataset distribution is similar to the KITTI Eigen split used in~\cite{godard2019digging,zhou2017unsupervised} for the pinhole model and explained in detail in Section~\ref{sec:kitti_eigen_split}. The training set comprises three scene categories: \textit{city}, \textit{residential} and \textit{sub-urban}. While training, these categories are randomly shuffled and fed to the network. We filter static scenes based on the vehicle's speed with a threshold of $2\,\text{km}/\text{h}$ to remove image frames that only observe minimal camera ego-motion since distance cannot be learned under these circumstances. Comparable to previous experiments on pinhole SfM~\cite{zhou2017unsupervised, godard2019digging}, we set the length of the training sequence to $3$.\par
\subsubsection{Occlusion Correction}
\label{Occlusion Correction}

\begin{figure}[!t]
 \centering
 \includegraphics[width=\textwidth, height=10cm, keepaspectratio]
                 {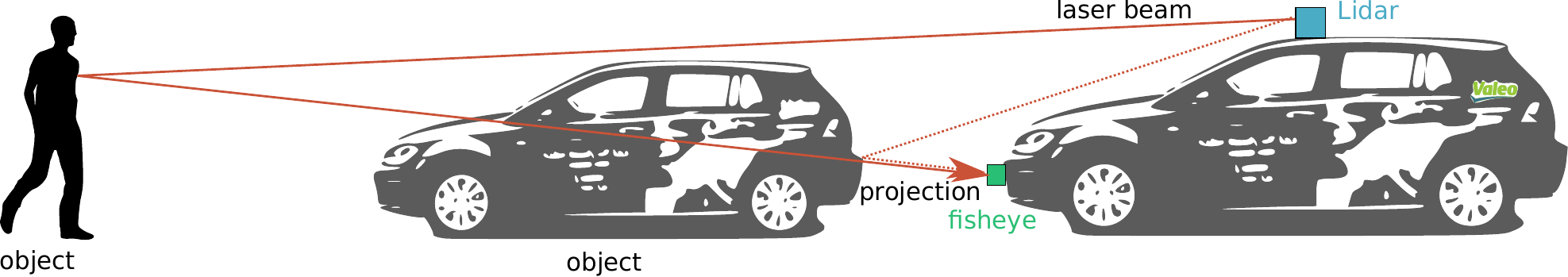}
    \caption[\bf Occlusion Scenario.]
            {\textbf{The LiDAR's viewpoint lies higher than the fisheye camera's viewpoint.} 3D points from the object (person) will be mapped, even though -- from the camera's point of view -- the point is occluded.}
    \label{OcclusionFig}
\end{figure}

The data's sensor fusion will be correct if both camera and the Velodyne LiDAR scanner observe the world from the same viewpoint. However, in our vehicle, the fisheye cameras are in the front, and LiDAR is placed at the top, as seen in Figure~\ref{OcclusionFig}. LiDAR perceives the environment behind objects that occlude the view of the camera. This problem of occlusion results in a wrong mapping of depth points that are not visible to the camera. It is harder to solve since occluded points are projected adjacently to disoccluded points.\par
\begin{figure}[!ht]
 \centering
 \includegraphics[width=\textwidth, height=10cm, keepaspectratio]
                 {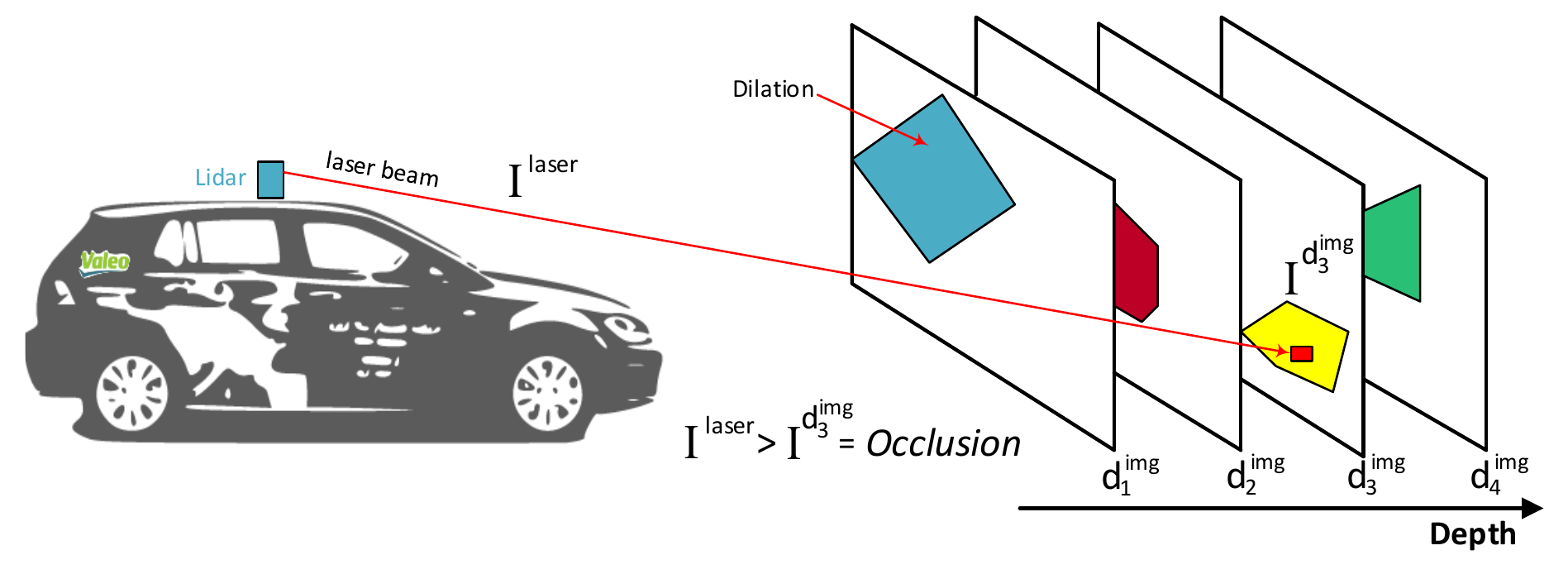}
    \caption[\bf Distance-based segmentation technique with morphological filters.]
            {\textbf{Distance-based segmentation technique with morphological filters.} The sliced sub-windows of the single depth image $d^{\text{img}}$, contains intensity values in ascending order.}
    \label{DilationFig}
\end{figure}
\begin{figure}[!t]
    \centering
    \resizebox{\textwidth}{!}{\begin{tabular}{ccc}
 	\begin{overpic}[width=0.8\columnwidth]{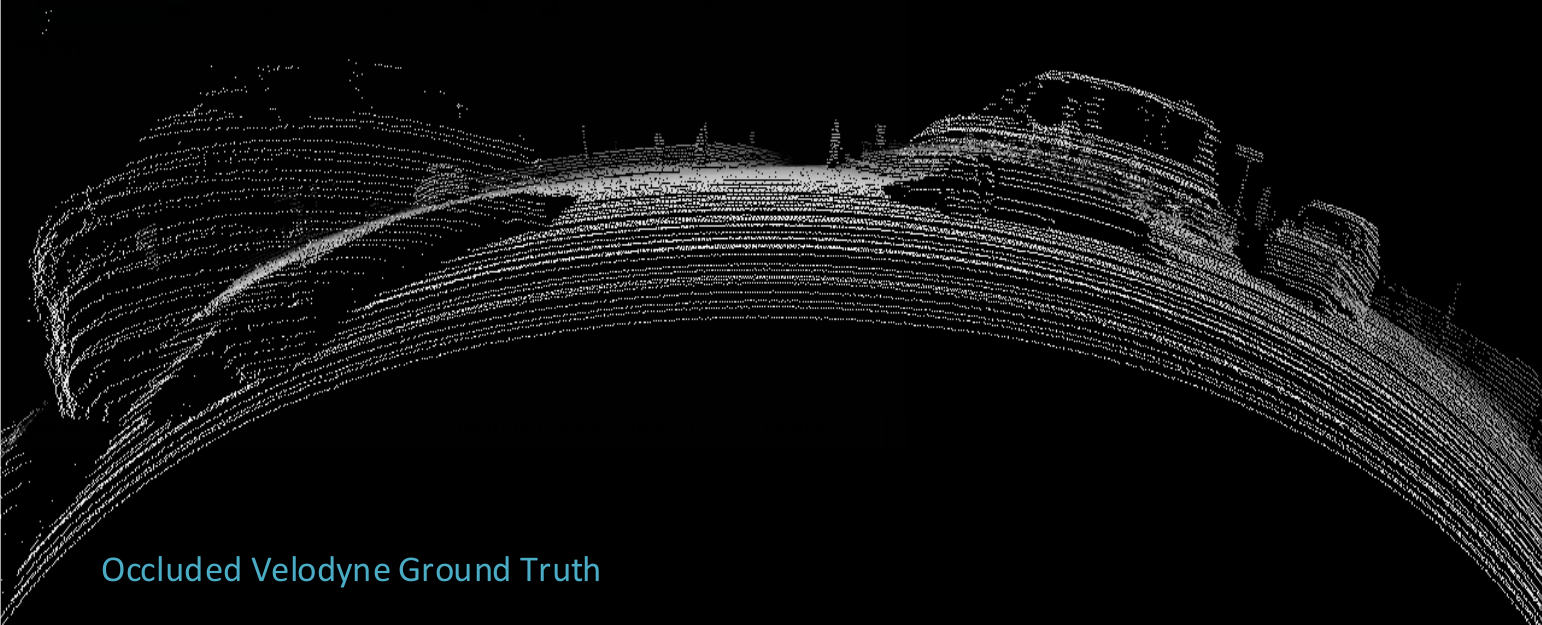}
    \put (0.25,2.25) {\colorbox{darkgreen}{$\displaystyle\textcolor{black}{\text{\footnotesize(a)}}$}}
    \end{overpic} \\
    \begin{overpic}[width=0.8\columnwidth]{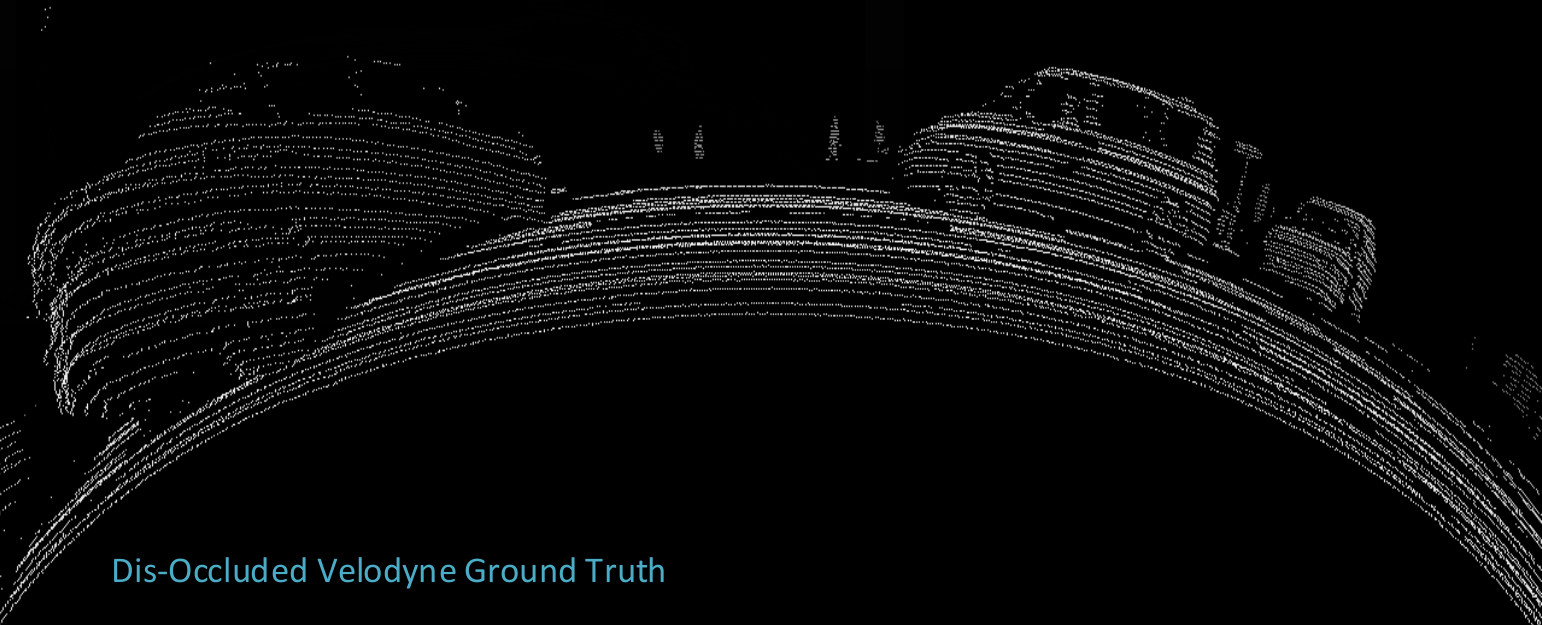}
    \put (0.25,2.25) {\colorbox{darkgreen}{$\displaystyle\textcolor{black}{\text{\footnotesize(b)}}$}}
    \end{overpic} \\
    
    \begin{overpic}[width=0.8\columnwidth]{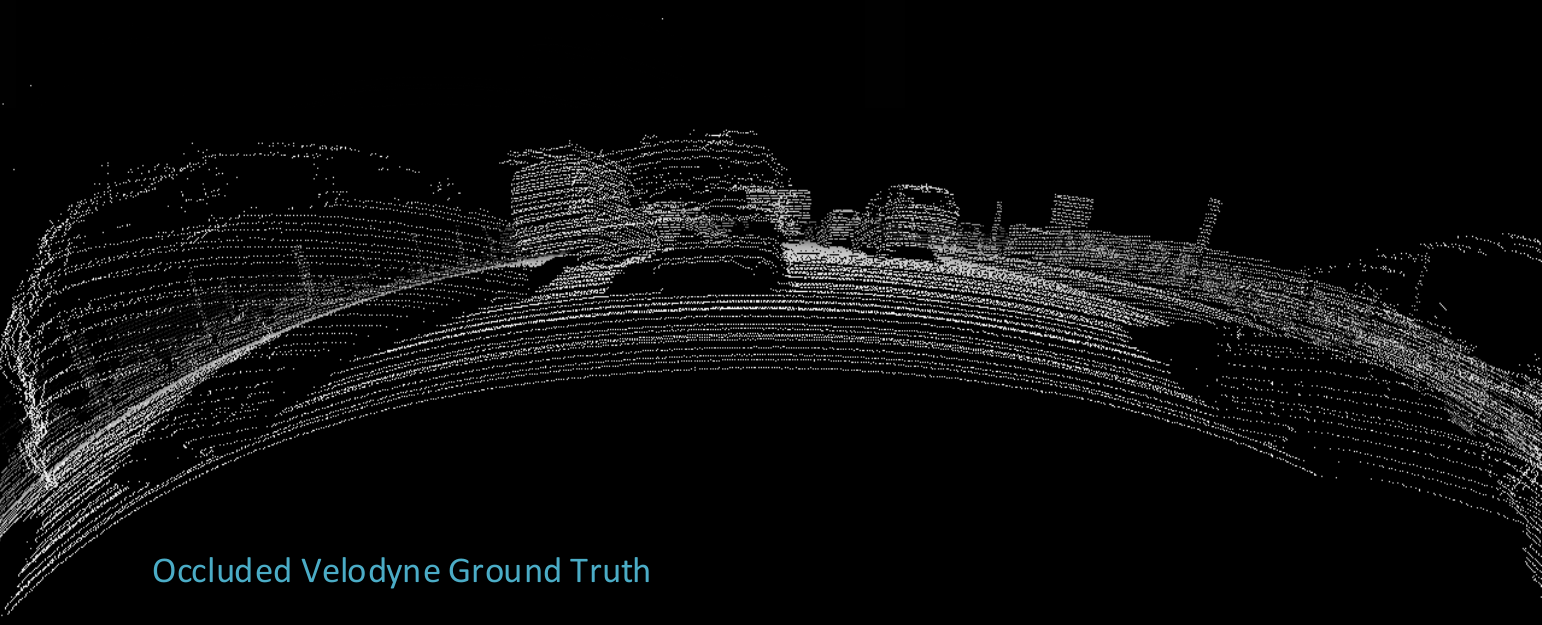}
    \put (0.25,2.25) {\colorbox{darkgreen}{$\displaystyle\textcolor{black}{\text{\footnotesize(c)}}$}}
    \end{overpic} \\
 	\begin{overpic}[width=0.8\columnwidth]{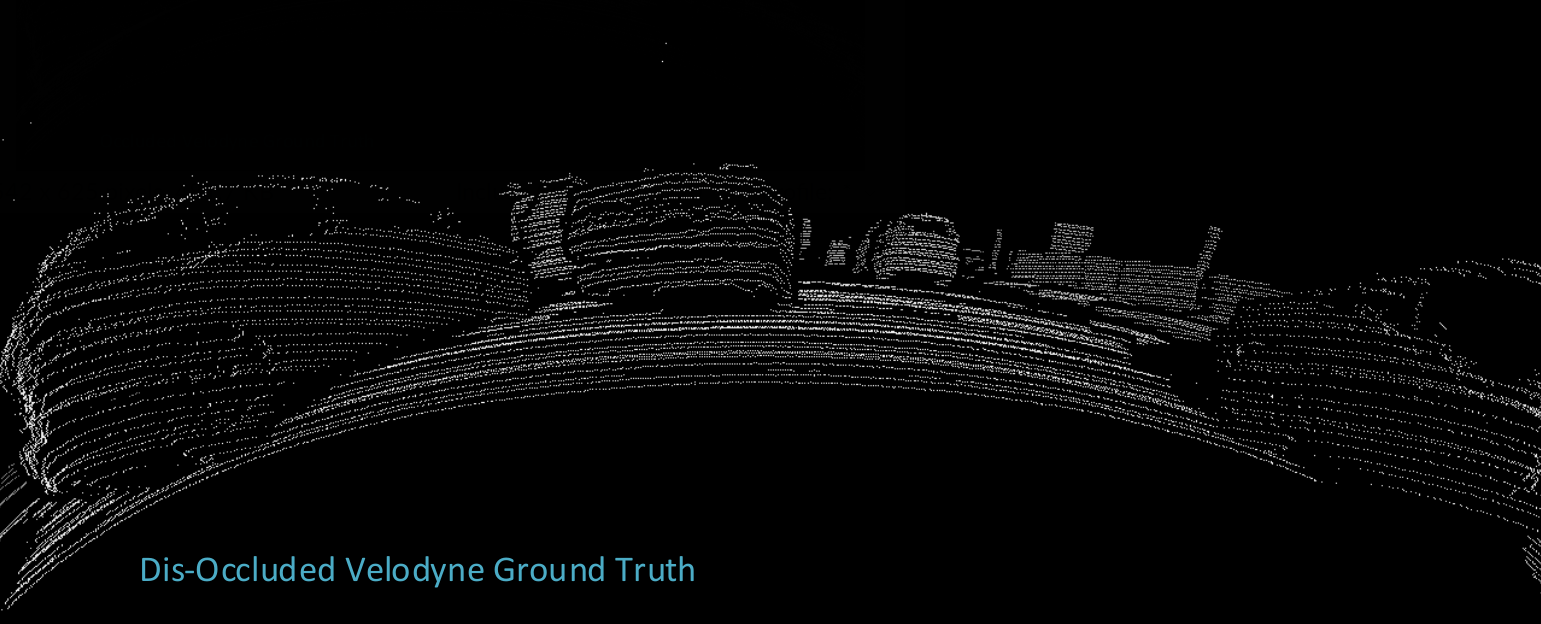}
    \put (0.25,2.25) {\colorbox{darkgreen}{$\displaystyle\textcolor{black}{\text{\footnotesize(d)}}$}}
    \end{overpic}
\end{tabular}}
    \caption[\bf Occlusion correction of ground-truth LiDAR distance maps.]
            {\textbf{Ground truth LiDAR values projected onto an image to obtain distance maps.} (a) and (c) Occluded LiDAR ground-truth maps, (b) and (d) Dis-Occluded LiDAR ground-truth maps}
\label{fig:occlusion-correction}
\end{figure}

To solve this problem, we adapted a distance-based segmentation technique from~\cite{kumar2018monocular} with morphological filters as shown in the Figure~\ref{DilationFig}. The depth image is denoted by $d^{\text{img}}$. Instead of directly mapping depth values on a single depth image $d^{\text{img}}$, we first split this image in $I$ sub-windows $d_i^{\text{img}}$, ${i = 1, \ldots, I}$ of size $1280 \times 800$ $(W \times H)$ in the image plane, based on the lasers' distance projected from the LiDAR. Each $d_i^{\text{img}}$ is mapped with depth-points in the form of intensity on the image plane in the increasing order of distance from the LiDAR's lasers. We apply a morphological filter that is dilation on each $d_i^{\text{img}}$ to fill the sparse regions. The occluded points are removed with a thresholding technique \ie $I^{\text{laser}} > I^{d_i^{\text{img}}}$ regarded as occlusion, where $I^{\text{laser}}$ is the LiDAR's laser intensity based on distance and $I^{d_i^{\text{img}}}$ is intensity of dilated sub-windows $d_i^{\text{img}}$ based on distance. The results are shown in Figure~\ref{fig:occlusion-correction}.
\subsection{KITTI}
\label{sec:kitti_eigen_split}

The KITTI dataset consists of $42,382$ stereo sequences with corresponding raw LiDAR scans, $7,481$ images with bounding box annotations, and $200$ training images with semantic annotations with a resolution of $1242 \times 375$. The unrectified dataset has a resolution of $1392 \times 512$. Geometric understanding tasks such as depth, flow, and pose estimation are widely benchmarked using KITTI. We use the data split according to Eigen~\etal~\cite{eigen2015predicting} for self-supervised depth estimation and filter the static frames as proposed by Zhou \etal~\cite{zhou2017unsupervised}. The resulting training and validation set contains $39,810$ and $4,424$ monocular triplets. We use the standard test set of $697$ images, which covers a total of $29$ scenes. We use a single camera intrinsic matrix for the entire training and validation set. The camera's principal point is centered, and the focal length is set as the average of all the focal lengths in KITTI. The length of the training sequence is set to $3$. We also use the $652$ test frames from the Eigen split with improved ground truth provided by~\cite{uhrig2017sparsity}. We utilize the KITTI split~\cite{monodepth17}, whose test set includes the $200$ training images from the KITTI 2015 Stereo dataset~\cite{menze2015object}. This test set's advantage is that it contains labels for depth and semantic segmentation, suitable to ablate the benefits of MTL.\par
\subsection{Cityscapes} 

Cityscapes dataset has $2,975$ training and $500$ validation images, with a resolution of $2048 \times 1024$ captured in $50$ different cities. It features higher resolution street images of higher quality compared to KITTI. It has a similar setting compared to KITTI but contains more dynamics scenes. The validation set in principle provides ground-truth labels for depth estimation and semantic segmentation tasks; the depth labels are obtained by a classical Semi-Global Matching~\cite{hirschmuller2005accurate} algorithm. The KITTI dataset's depth labels are physical measurements from a LiDAR sensor and thereby better suited for evaluating a depth estimation model. For pixel-wise semantic segmentation, the dataset contains $19$ classes. We extracted the 2D boxes from the instance polygons for the entire training and validation split.\par
\section{Evaluation}

This section explains the metrics used for all the visual perception tasks in this thesis.

\subsection{Metrics for Depth Estimation}

\begin{table*}[htpb]
\centering
\scalebox{1.0}{
\begin{tabular}{ll}
\toprule
$\textbf{Abs Rel}: \frac{1}{|N|}\sum_{i\in N}\frac{\mid d_{i}-d_{i}^{*}\mid}{d_{i}^{*}}$
&$\textbf{RMSE}:\sqrt{\frac{1}{|N|}\sum_{i\in N}\parallel d_{i}-d_{i}^{*} \parallel^{2}}$\\
$\textbf{Sq Rel}:\frac{1}{|N|}\sum_{i\in N}\frac{\parallel d_{i}-d_{i}^{*}\parallel^{2}}{d_{i}^{*}}$
&$\textbf{RMSE log}: \sqrt{\frac{1}{|N|}\sum_{i\in N}\parallel \log (d_{i})- \log (d_{i}^{*}) \parallel^{2}}$ \\
\multicolumn{2}{l}{\textbf{Accuracies} $ \delta_{t}:
\frac{1}{|N|}|\{d \in N| \: \max(\frac{d_{i}}{d_{i}^{*}}, \frac{d_{i}^{*}}{d_{i}}) \: < 1.25^t\}|\times 100\%$} \\
\bottomrule
\end{tabular}
}
\caption[\bf Performance indicators for depth evaluation.]
{{\textbf Performance indicators for depth evaluation.} where $d_{i}$ and $d_{i}^{*}$ denotes the ground truth and predicted depth value of pixel $i$, respectively. $N$ denotes the set of pixels with real-depth/distance values in an image, $|.|$ returns the number of the input set elements.}
\label{tab:depth-metrics}
\end{table*}

Table~\ref{tab:depth-metrics} indicates the performance indicators for depth evaluation. Saxena \etal~\cite{saxena2005learning, saxena2008make3d} proposed the first set of criteria defined to assess the quality of an estimated depth map. It consists of computing the percentage of pixels having a relative error $\delta$ lower than a certain threshold. A larger set of metrics were proposed by Eigen \etal~\cite{Eigen_14} for depth prediction given by:
\begin{itemize}
    \item \textbf{Absolute Relative error (Abs Rel)}: Normalizes per-pixel errors according to real depth, reducing the effect of large errors with the distance. It ignores the difference's sign, preventing positive and negative errors from canceling each other out. The error averaged over the entire test set is the score given in the literature. The metric is not specified for null measurements, which is irrelevant since depth is never zero.
    \item \textbf{Squared Relative error (Sq Rel)}: The squared term penalizes larger depth errors (\eg, near discontinuities). It is expressed likewise to the absolute relative error, except it has the effect of penalizing particularly bad predictions. In the case of Abs Rel, the symbol of the difference is ignored. Differences larger than one have a greater impact on Sq Rel than on Abs Rel, while differences smaller than one have a greater impact on Abs Rel than on Sq Rel. As a result, comparing the two tests should aid in determining whether the errors are caused by outliers or by several small offsets in the forecast. Sq Rel should not be used in isolation because the existence of outliers can lead to incorrect conclusions about a model's results. It is measured in the same units as the data.
    \item \textbf{Root Mean Squared Error (RMSE)}: A traditional metric for measuring regression errors. It measures the standard deviation of the error; a higher value implies widely varying prediction accuracy. The RMSE is defined as the squared root of the mean quadratic difference between the prediction and the ground truth. The RMSE, like Sq Rel, is susceptible to outliers. The squared root is used to ensure that the RMSE is expressed in the same unit as the data.
    \item \textbf{Root Mean Squared logarithmic error (RMSE log)}: The logarithm makes this error relative, reducing the effect of large errors with the distance. It is more invariant to the scale of the error and is meriting to note that it penalizes underestimates more than overestimates, which makes it less valuable of a metric for some applications of depth estimation, like autonomous navigation.
    \item \textbf{Three inlier thresholds} are used to determine the quality of the depth predictions. These are the percentage of predictions within some factor $\delta$ of the ground truth. The standard measures, used in these experiments, are $\delta < 1.25^{t}, t = \{1, 2, 3\}$. In other words, a value of $0.9$ for $\delta_{1}$ means that $90 \%$ of the pixels are such that the difference between the ground truth $d_{i}$ and the prediction $d_{i}^{*}$ is less than $25 \%$ of the smallest of the two (\ie, $25 \%$ of the prediction if $d_{i}<d_{i}^{*}$ and conversely). Contrary to previous metrics, the model performs better when the accuracy is higher.
\end{itemize}
\subsection{Metrics for Segmentation-Based Tasks}

Some basic concepts used by the metrics:

\begin{itemize}[nosep]
    \item True Positives (TP): A true positive is an outcome where the model correctly predicts the \textit{positive class} \ie, a correct Segmentation. Segmentation with IOU $\geq$ threshold.
    \item False Positives (FP): A false positive is an outcome where the model incorrectly predicts the \textit{positive class} \ie, a wrong Segmentation. Segmentation with IOU $<$ threshold.
    \item False Negatives (FN): A false negative is an outcome where the model incorrectly predicts the \textit{negative class} \ie, a ground truth not segmented.
    \item True Negative (TN): A true negative is an outcome where the model correctly predicts the \textit{negative class}.
\end{itemize}
\textbf{Pixel Accuracy (PA):} Also known as global accuracy~\cite{badrinarayanan2017segnet} is a straightforward metric that measures the ratio between the number of correctly classified pixels and the total number of pixels. The mean pixel accuracy (mPA) metric computes the proportion of right pixels on a per-class basis. mPA is also known as \emph{class average accuracy}~\cite{badrinarayanan2017segnet}. Accuracy is obtained by taking the ratio of correctly classified pixels with respect to the total pixels. The key drawback of employing this metric is that the outcome may appear favorable if one class outnumbers the other. If, for example, the background class covers 90\% of the input image, we can achieve 90\% accuracy by simply classifying every pixel as background.
\begin{equation}
\label{pa}
    PA=\frac{\sum_{j=1}^{k}{n_{jj}}}{\sum_{j=1}^{k}{t_{j}}}, \qquad mPA=\frac{1}{k}\sum_{j=1}^{k}\frac{n_{jj}}{t_{j}}
\end{equation}  
\begin{equation}
    Accuracy = \frac{TP+TN}{TP+TN+FP+FN}
\end{equation}
where $n_{jj}$ represents the total number of pixels classified and labeled as class \emph{j}. In other terms, $n_{jj}$ is the total number of \textit{True Positives} for class \emph{j}. The total number of pixels labeled as class \emph{j} is $t_{j}$~\cite{ulku2019survey}.\par
\begin{figure}[!t]
  \centering
    \includegraphics[width=0.4\textwidth]{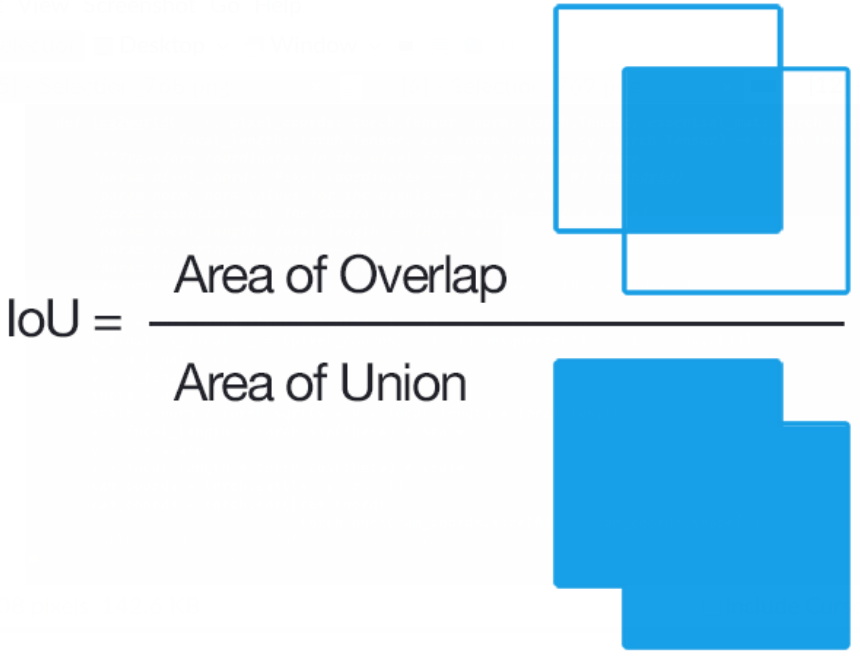}
    \caption[\bf Depiction of the Intersection over Union (IoU) metric.]{\textbf{Depiction of Intersection over Union (IoU) metric.} Figure reproduced from~\cite{iou_2021}.} 
    \label{fig:iou-metric}
\end{figure}

\textbf{Intersection over Union (IoU):} Also known as the Jaccard Index, the IoU statistic compares the similarity and diversity of sample sets. It is the ratio of the intersection of the pixel-wise classification results with the ground truth to their union as shown in Figure~\ref{fig:iou-metric}.
\begin{equation}
\label{iu}
    IoU=\frac{\sum_{j=1}^{k}{n_{jj}}}{\sum_{j=1}^{k}({n_{ij}+n_{ji}+n_{jj}})}, \qquad i \neq j
\end{equation}
where $n_{ij}$ denotes the number of pixels labeled as class \emph{i} but listed as class \emph{j}. They are, in other terms, \textit{False Positives} (false alarms) for class \emph{j}. Similarly, ${n_{ji}}$, the total number of pixels labeled as class \emph{j} but listed as class \emph{i}, represents the \textit{False Negatives} (misses) for class \emph{j}~\cite{ulku2019survey}. An extended version of IoU is widely in use. It is given by: \par
\textbf{Mean Intersection over Union (mIoU):} mIoU is the class-averaged IoU, \ie
\begin{equation}
    mIoU=\frac{1}{k}\sum_{j=1}^{k}\frac{n_{jj}}{n_{ij}+n_{ji}+n_{jj}}, \qquad i \neq j
\label{miou}    
\end{equation}
\subsection{Metrics for Object Detection}

Evaluation in object detection is difficult since there are two distinct tasks to measure:
\begin{itemize}[nosep]
    \item Determining the presence or absence of an object in the scene (classification).
    \item Determining the object's position (localization, a regression task).
\end{itemize}
Furthermore, there will be several classes in a standard data set, and their distribution will be non-uniform (\eg, there might be many more vehicles than traffic signals). As a result, biases would be introduced by a simplistic accuracy-based metric. It is also essential to determine the possibility of misclassifications. As a result, a "confidence score" or model score must be assigned to each bounding box observed, and the model must be evaluated at different levels of confidence. The Average Precision (AP) was created to meet these requirements. To comprehend the AP, one must first understand the Precision and Recall of a classifier~\cite{object_detection_metrics_2021}.\par

\emph{Precision} is defined as the "False Positive Rate," or the ratio of true object detections to the total number of objects predicted by the classifier. If the precision score is close to $1$, there is a good chance that whatever the classifier predicts as a positive detection is accurate. In other terms, Precision is a model's ability to identify only relevant objects. It is calculated as a percentage of correct positive predictions \ie,
\begin{equation}
\text{Precision}= \frac{\mathrm{TP}}{\mathrm{TP}+\mathrm{FP}}=\frac{\mathrm{TP}}{\text{all detections}}
\end{equation}
The "False Negative Rate," or the ratio of true object detections to the total number of items in the data collection, is measured by \emph{Recall}. If the recall score is close to $1$, the model can correctly detect almost all the objects in the dataset. In other terms, the ability of a model to identify all relevant cases is referred to as Recall (all ground truth bounding boxes). It is given as the percentage of true positives detected among all related ground truths and is calculated as follows:
\begin{equation}
\text{Recall}=\frac{\mathrm{TP}}{\mathrm{TP}+\mathrm{FN}}=\frac{\mathrm{TP}}{\text{all ground truths}}
\end{equation}
Finally, it is essential to remember that Precision and Recall have an inverse relationship. These metrics are affected by the model score threshold being set and the quality of the model. To compute the Average Precision (AP), the precision-recall curve for a specific class (\eg, car) is calculated from the model's detection output by varying the model score threshold that defines what is counted as a model-predicted positive detection of the class~\cite{object_detection_metrics_2021}.\par

\textbf{Average Precision (AP):} The method of computing AP by the PASCAL VOC challenge has improved since 2010. Currently, the PASCAL VOC challenge interpolation uses all data points, rather than interpolating just 11 equally spaced points (11-point interpolation method) as described in~\cite{everingham2010pascal}, and they propose to \emph{interpolate all data points}.\par

\textbf{11-point interpolation:} The 11-point interpolation tries to summarize the shape of the Precision $\times$ Recall curve by averaging the precision at a set of eleven equally spaced recall levels $[0,0.1,0.2, \ldots, 1]:$
\begin{equation}
\mathrm{AP}=\frac{1}{11} \sum_{r \in\{0,0.1, \ldots, 1\}} \rho_{\mathrm{interp}(r)}
\end{equation}
with
\begin{equation}
\rho_{\text {interp }}=\max _{\bar{r}: \bar{r} \geq r} \rho(\tilde{r})
\end{equation}
where $\rho(\tilde{r})$ is the measured precision at recall $\tilde{r}$. Instead of using the precision observed at each point, the AP is calculated by interpolating the precision only at the 11 levels $r$ taking the \textbf{maximum precision whose recall value is greater than $r$}~\cite{object_detection_metrics_2021a}.\par
\textbf{Interpolating all points:} Instead of just interpolating between the 11 equally spaced points, we might interpolate between all $n$ points in such a way that:
\begin{equation}
\sum_{n=0}\left(r_{n+1}-r_{n}\right) \rho_{\text {interp }}\left(r_{n+1}\right)
\end{equation}
with
\begin{equation}
\rho_{\text {interp}}\left(r_{n+1}\right)=\max _{\tilde{r}: \tilde{r} \geq r_{n+1}} \rho(\tilde{r})
\end{equation}
where $\rho(\tilde{r})$ represents the calculated precision at recall $\tilde{r}$.
Instead of using the precision observed at just a few points, the AP is now calculated by interpolating the precision at each step, with $r$ taking the maximum precision whose recall value is greater or equal to $r+1$. We measure the approximate region under the curve in this manner~\cite{object_detection_metrics_2021a}.\par

\textbf{Localization and Intersection over Union:} To assess the model's performance on the task of object localization, we must first determine how well the model predicted the object's position. Typically, this is accomplished by drawing a bounding box around the object of interest, but in some situations, an N-sided polygon or even pixel by pixel segmentation is used. The localization task is usually evaluated on the Intersection over Union threshold in both of these situations (IoU). In this thesis, for fisheye images, the object representation is mainly carried out using N-sided polygons. Therefore, we widely use (mIoU) as shown in Eq.~\ref{miou}.

Finally, now that we have defined AP and IoU thresholds, the mean Average Precision (mAP) score is computed by averaging the AP across all classes and/or IoU thresholds.
\chapter{Related Work}
\label{Chapter3}
\minitoc

\section{Depth Estimation}

Numerous deep learning models to infer scene depth from images have been proposed in recent years. They differ in many ways, including architecture and layer design, the required number of input images, training strategy, and datasets on which they are trained and tested. Among the various points used to categorize models, the number of images and the chosen training strategy have a strong influence on the quality of estimated depth maps, while the architecture design may affect the model's ability to generalize on unseen data~\cite{depth_estimation_survey_2020}.\par

When the training technique is used as a criterion for classification, three distinct classes emerge. To learn the task of depth prediction, supervised methods include ground truth depth maps associated with RGB inputs. Then there are unsupervised methods, which require only RGB images during training, and semi-supervised methods, which combine the two approaches and depend on ground truth depth maps when they are available in the training set. Models that are supervised are known to produce better results, but they come at a cost. As discussed in \cite{mahjourian2018unsupervised}, collecting datasets containing both RGB images and depth maps associated with them is difficult and error-prone. As a result, such models are often trained on a small number of datasets, restricting both research and implementation possibilities~\cite{depth_estimation_survey_2020}.\par

Methods are classified into the same two groups as non-learning approaches for the amount of RGB images retained, namely monocular and multi-view algorithms. The former can only infer depth from a single image, whereas the latter requires at least two. Monocular methods are less accurate. While some agree that the difference between the two types of algorithms can be bridged, others claim that monocular approaches can never catch up because there is no way to compensate for the lack of detail when only one image is used. Furthermore, monocular depth estimation is an ill-posed problem, meaning that any number of 3D structures will result in the same 2D observation. Nonetheless, applying smoothness constraints always yields acceptable results, often at the expense of fine details. Monocular algorithms produce up-to-scale depth maps, while multi-view methods can produce metric depth by depending on known real-world measurements such as the distance between two cameras in the recording setup~\cite{depth_estimation_survey_2020}.\par

These two characteristics are used in the following subsections to distinguish and present a broad range of state-of-the-art models.
\subsection{Supervised Monocular Methods}

Depth estimation is a crucial task for automated driving, and multi-view geometric approaches were traditionally used for computing depth. Some of the initial prototypes of automated driving relied primarily on depth estimation~\cite{franke1998autonomous}, and to enable accurate depth estimation, stereo cameras were used. As LiDAR-based depth perception is sparse and costly, image-based methods are of significant interest in perception systems regarding coverage density and redundancy. Eigen~\etal~\cite{Eigen_14} was one of the first to successfully prove that CNN's are capable of predicting depth from single images. Following its success, Eigen~\etal~\cite{eigen2015predicting} proposed a general multi-scale system that can handle tasks like depth map estimation, surface normal estimation, and semantic label prediction from a single image. Demon~\cite{ummenhofer_2017} is a network developed by Ummenhofer~\etal that can predict depth and egomotion, \ie, camera motion, from a pair of images taken by a single camera, thus solving the two-view SfM problem. Vijayanarasimhan~\etal~\cite{Vijayanarasimhan2017} proposed a different architecture called SfM-Net to solve the same \emph{SfM} problem. Their model can compute depth, egomotion, 3D rotations, and translations for the dynamic objects in the scene and motion masks for these dynamic objects, given a pair of images and camera intrinsics. SfM-Net is unique in that it can be used for a variety of training scenarios. It can be trained supervised, using egomotion or depth, or self-supervised, by learning its tasks without having access to ground truth data for the quantities it predicts. To infer depth from a single image, Liu~\etal~\cite{liu2015learning, liu2015deep} combined a convolutional network and a continuous conditional random field (CRF)~\cite{lafferty2001conditional}. Meanwhile, they suggested a super-pixel pooling method to speed up the CNN, and it aids in the design of the deeper network to increase depth estimation accuracy. Shelhamer~\etal~\cite{shelhamer2015scene} proposes a fully connected network (FCN) framework for monocular depth estimation, after which the proposed framework jointly optimizes the intrinsic factorization to recover the input image. Laina~\etal~\cite{laina2016deeper}, inspired by ResNet's~\cite {he2016deep} outstanding performance, used residual learning to learn the mapping relation between depth maps and single images, resulting in a network that is deeper and more accurate than previous works in depth estimation. Furthermore, up-sampling blocks replace fully connected layers in ResNet to increase the resolution of the projected depth map. Chen~\etal~\cite{chen2016} investigated a novel algorithm to tackle the problem of perceiving single-image depth estimation in the wild and also learn the camera intrinsics. Rather than using the supervised signal of depth as the ground truth, their networks are trained using relative depth annotations.\par

\textbf{Methods based on adversarial learning:} In recent years, the adversarial learning proposed in~\cite{goodfellow2014generative} has become a highly investigated research path due to its outstanding output on data generation~\cite{zhang2017stackgan}. A broad range of algorithms, theories, and applications have been developed, as summarized in~\cite{hong2019generative}. Stack GAN~\cite{huang2017stacked}, Conditional GAN~\cite{mirza2014conditional}, and Cycle GAN \cite{zhu2017unpaired} all based on~\cite{goodfellow2014generative}, are incorporated into depth estimation tasks and have a positive effect on the depth estimation~\cite{feng2019sganvo, jung2017depth, gwn2018generative}. Jung \etal; incorporate adversarial learning into monocular depth estimation tasks in~\cite{jung2017depth}. Here, a Global Net and a Refinement Net make up the generator, and these networks are designed to estimate global and local 3D structures from a single image. The predicted depth maps are then distinguished from the real ones using a discriminator, which is a typical type in supervised methods. For monocular depth estimation, conditional GAN is also used in~\cite{gwn2018generative}. The difference from~\cite{jung2017depth} is that a secondary GAN is used to produce a more refined depth map based on the image and a coarse approximate depth map.

The supervised methods can effectively learn a function to map 3D structures, and their scale details since the ground truth supervises them. However, the labeled training sets, which are difficult and costly to obtain, restrict  the application of these supervised methods~\cite{zhao2020monocular}.\par
\subsection{Self-Supervised Monocular Methods}

Although supervised methods learn their tasks efficiently using simple loss functions and typically produce good results, they also place strict limits on the datasets that can be used to train them. After all, they rely on ground truth depth maps that are compatible with the corresponding RGB images, which is difficult to achieve without errors. Furthermore, depending on the data acquisition system, the depth maps can be extremely sparse and contain numerous gaps, making learning challenging, if not impossible, in these areas. Self-Supervised approaches, on the other hand, are trained solely on RGB images and do not require any depth data, enabling training on a more significant number of datasets or making the recording of new ones easier~\cite{depth_estimation_survey_2020}. Here, current state-of-the-art approaches rely on neural networks~\cite{Fu2018, Zhang2019b}, which can even be trained in an entirely self-supervised fashion from sequential images~\cite{zhou2017unsupervised}, giving a clear advantage over supervised approaches in terms of applicability to arbitrary data domains.\par

The approaches of Garg~\etal~\cite{garg2016unsupervised}, and Zhou~\etal~\cite{zhou2017unsupervised} showed that it is possible to train networks in a self-supervised fashion by modeling depth as part of a geometric projection between stereo images and sequential images, respectively. A model based on view synthesis from a pair of rectified stereo images has also been proposed by Godard~\cite{monodepth17}. In two main ways, their solution, Monodepth, varies from that of Garg~\cite{garg2016unsupervised}. To avoid the Taylor approximation, which makes optimization difficult, they produce images using bilinear sampling, resulting in a fully sub-differentiable training loss. Second, they enforce a left-right consistency check in the form of an additional term in the training loss to resolve the ill-posed nature of monocular depth estimation. Many stereo algorithms use this form of consistency check as a post-processing stage. However, they were able to integrate it directly into the network, allowing for end-to-end learning. Zhou~\etal~\cite{zhou2017unsupervised} suggested a paradigm that overcomes the constraint on potentially accessible datasets. Instead of stereo images for training, it employs consecutive frames captured by a moving camera. Aside from disparity, the model must also estimate camera motion between frames. Camera motion is required by the models of Garg, as well as Monodepth~\cite{monodepth17}. However, it is constant and known ahead of time because the stereo cameras are calibrated before recording the dataset. The only details needed are the camera's intrinsic parameters. The core concept remains that of view synthesis, as well as the same bilinear sampling process used in Monodepth~\cite{monodepth17}.

The model of Mahjourian~\etal~\cite{mahjourian2018unsupervised} is designed to be trained on a broad range of datasets, similar to that of Zhou~\etal~\cite{zhou2017unsupervised}. Mahjourian~\etal~\cite{mahjourian2018unsupervised} predicts depth and egomotion, but it also considers explicitly the assumed 3D structure of the entire scene, rather than depending solely on a local photometric loss. Unlike the masks predicted by a dedicated network by Zhou~\cite{zhou2017unsupervised}, those of Mahjourian~\cite{mahjourian2018unsupervised} are computed analytically, making the overall learning problem more straightforward. Wang~\etal~\cite{Wang_2018_CVPR} discovered that unsupervised monocular models trained on monocular sequences, and constrained by the nature of the problem such as those used by Zhou~\etal~\cite{zhou2017unsupervised}, do not perform as well as unsupervised monocular models trained on rectified stereo datasets, such as those used by Godard~\etal~\cite{monodepth17}. To clarify the performance gap, they established two significant differences between stereo and monocular strategies: (i) unknown camera pose between frames and (ii) the uncertainty in scale inherent in all monocular models trained on monocular sequences. Both are only partly resolved by the pose network, and Wang~\etal~\cite{Wang_2018_CVPR} demonstrated that scale uncertainty induces divergence during training. This is due to the scale sensitivity of the depth regularization terms (\ie, the smoothing terms) used in the loss function, as shown empirically in their work. Furthermore, they argue that the pose network is superfluous and that it can be replaced by a differentiable and deterministic goal for pose prediction, as well as a simple normalization technique, both of which are widely used in Direct Visual Odometry (DVO)~\cite{depth_estimation_survey_2020}.

Indeed, pose estimation from depth is a well-studied problem with geometric properties and efficient algorithms. There are three benefits to replacing the pose network with one of them. It does not require any learning parameters, for starters, making the model simpler than those that rely on a pose network. Second, it establishes a direct relationship between the input dense depth map and the output pose prediction. In contrast, pose network models, except for Mahjourian~\etal~\cite{mahjourian2018unsupervised}, ignore scene geometry and produce camera pose estimates based solely on photometric appearance. Third, given a depth estimate, DVO solves for camera pose by minimizing the same view synthesis loss used to train the network, preventing an increase in computations. Wang~\etal~\cite{Wang_2018_CVPR} used a differentiable DVO (DDVO) algorithm similar to the inverse compositional spatial transformer network to enable backpropagation during training. Since it is a second-order gradient descent process, a good initialization point will lead to a better solution than one chosen at random. As a result, rather than starting from the identity pose, relying on a first estimate provided by a pre-trained pose network is likely to produce better results~\cite{depth_estimation_survey_2020}.

The initial concept has been extended by considering improved loss functions~\cite{Aleotti2018, barron2019general, monodepth17, godard2019digging, watson2019self, shu2020featdepth}, the application of generative adversarial networks (GANs)~\cite{Aleotti2018, cs2018monocular, Pilzer2018}, generated proxy labels from traditional stereo algorithms~\cite{Tosi2019}, or synthetic data~\cite{Bozorgtabar2019}. Other approaches proposed to use specialized architectures for self-supervised depth estimation~\cite{guizilini2019packnet, Wang_2018_CVPR, Zhou2019}, they apply teacher-student learning~\cite{Pilzer2019} to use test-time refinement strategies~\cite{Casser2019a, casser2019depth}, to employ recurrent neural networks~\cite{Wang2019, Zhang2019c}, or to predict the camera parameters~\cite{Gordon2019} to enable training across images from different cameras.\par
\subsection{Depth Estimation on Fisheye Cameras}

Recent approaches also investigated the application of self-supervised depth estimation to $360\degree$ images~\cite{Wang2020b, Jin2020}. Most of the works have solely focused on traditional 2D content captured with cameras following a typical pinhole projection model based on rectified image sequences. Omnidirectional ($360\degree$) content is now more easily and consistently produced thanks to the development of efficient spherical cameras and rigs and is seeing increased adoption in robotics and autonomous vehicles. Many real-world applications rely on more advanced camera geometries \eg, fisheye camera images.\par

With the surge of efficient and cheap wide-angle fisheye cameras and their larger FoV in contrast to pinhole cameras, there has been significant interest in the computer vision community to perform depth estimation from omnidirectional content similar to traditional 2D content via omnidirectional stereo~\cite{ma20153d, pathak2016dense, li2005spherical}. There is also a trend of integrating depth estimation tasks into multi-task models~\cite{sistu2019neurall, chennupati2019auxnet}. Most of the depth estimation methods were demonstrated in automated driving on rectified KITTI video sequences where barrel distortion was removed. The same multi-view geometry~\cite{hartley2003multiple} principles apply to $360\degree$ images equivalently as they apply to pinhole camera images. The underlying geometrical structure can be estimated by observing the scene from multiple perspectives and establishing correspondences between them. By taking into account the different projection models and defining the disparity as angular displacements, the traditional binocular or multi-view stereo~\cite{furukawa2015multi} problem is reformulated to binocular or multi-view spherical stereo~\cite{li2008binocular} for $360\degree$ cameras. It was recently~\cite{huang20176} demonstrated that using \emph{SfM}, $360\degree$ videos captured with a moving camera can be used to reconstruct a scene's geometry.\par

Current CNN processing pipelines can be applied to spherical input in two simple ways. Either directly on a projected (usually equirectangular) image or by projecting spherical content to the faces of a cube (cube map) and running CNN predictions on them, which are then merged by back-projecting them to the spherical domain. New techniques for applying CNNs to omnidirectional input have recently been presented. Given the difficulty of directly modeling the projection's distortion in typical CNNs while also achieving invariance to the viewpoint's rotation, \cite{khasanova2017graph} proposes a graph-based deep learning approach. Su~\etal~\cite{su2017learning} used a planar CNN to process $360\degree$ images in the equirectangular projection. They designed a novel approach by transferring appropriate convolution weights from an existing network trained on traditional 2D images to learn appropriate convolution weights for equirectangular projected spherical images. This conversion from the 2D to the $360\degree$ domain is achieved by enforcing consistency between the predictions of the 2D projected views and those in the $360\degree$ image. Recent work on convolutions~\cite{jeon2017active, dai2017deformable} that learn their shape, as well as their weights, has been applied to fisheye lenses~\cite{deng2019restricted}. However, apart from these works, applying self-supervised depth estimation to more advanced geometries, such as fisheye camera images, has not been investigated extensively yet.\par

Compared to the state-of-the-art approaches, in this thesis, we explore the \emph{SfM} approach to developing a self-supervised training strategy that aims to infer a distance map from a sequence of distorted and unrectified raw fisheye and pinhole images. We aim to develop a generic end-to-end self-supervised training pipeline to estimate monocular depth maps on raw distorted images for various camera models. We create a training framework for self-supervised distance estimation, which jointly trains and infers images from multiple fisheye cameras and viewpoints. Also, improve the photometric loss by a general and robust loss function. Further, we explore the avenues in introducing a novel architecture for the learning of self-supervised distance estimation synergized with semantic segmentation. We look upon the issues of the dynamic object impact on self-supervised distance estimation by using semantic guidance.\par
\section{Object Detection}

We can broadly classify the state-of-the-art object detection methods based on deep learning into two types: two-stage detectors and single-stage detectors. We provide an overview of the categorization of CNN-based object detection methods for object detection on pinhole camera images in Table~\ref{tab:object-detection-survey}. We categorize based on the number of stages involved in the framework, deformable part-based detection methods, and keypoint-based detectors.
\begin{table}[!t]
\centering
\begin{adjustbox}{width=\columnwidth}
\setlength{\tabcolsep}{0.1em}
\begin{tabular}{@{}lllll@{}}
\toprule
  \multicolumn{5}{c}{\cellcolor[HTML]{e5b9b5} \textbf{CNN based Object Detection methods}} \\ 
\midrule
  \multicolumn{1}{c}{\cellcolor[HTML]{00b0f0} \textit{
  \begin{tabular}[c]{@{}c@{}} Single stage \\ detectors\end{tabular}}} &
  \multicolumn{1}{c}{\cellcolor[HTML]{00b050} \textit{
  \begin{tabular}[c]{@{}c@{}} Two stage\\ detectors\end{tabular}}} &
  \multicolumn{1}{c}{\cellcolor[HTML]{ab9ac0} \textit{
  \begin{tabular}[c]{@{}c@{}} Multi-stage \\ detectors\end{tabular}}} &
  \multicolumn{1}{c}{\cellcolor[HTML]{7d9ebf} \textit{
  \begin{tabular}[c]{@{}c@{}} Deformable part \\ based detectors\end{tabular}}} &
  \multicolumn{1}{c}{\cellcolor[HTML]{a5a5a5} \textit{
  \begin{tabular}[c]{@{}c@{}} Keypoint \\ based detectors\end{tabular}}} \\
\midrule
SSD~\cite{liu2016ssd} & R-CNN~\cite{girshick2014rich} & Cascade R-CNN~\cite{cai2018cascade} & DPM-CNN~\cite{girshick2015deformable} & CornerNet~\cite{law2018cornernet} \\
Yolo9000~\cite{redmon2016you} & Fast R-CNN~\cite{girshick2015fast} & CRAFT~\cite{yang2016craft} & DeepIDNet~\cite{ouyang2015deepid} & ExtremeNet~\cite{zhou2019bottom} \\
Retinanet~\cite{lin2017focal} & Faster R-CNN~\cite{ren2017faster} & CC-Net~\cite{ouyang2017chained} & DP-FCN~\cite{mordan2017deformable} & CenterNet~\cite{duan2019centernet} \\
Squeezedet~\cite{wu2017squeezedet} & R-FCN~\cite{dai2016r} & Multipath Net~\cite{zagoruyko2016multipath}
& Deformable ConvNets~\cite{dai2017deformable} & \\
SPP-Net~\cite{he2015spatial} & & Multi-region CNN~\cite{gidaris2015object} & & \\
Overfeat~\cite{sermanet2014overfeat} & & HyperNet~\cite{kong2016hypernet} & & \\
DSSD~\cite{fu2017dssd} & & IoU-Net~\cite{jiang2018acquisition}  & & \\
MDSSD~\cite{Cui2020MDSSDMD} & & Hybrid task cascade~\cite{chen2019hybrid} & & \\
DETR~\cite{carion2020end} & & & & \\
EfficientDet~\cite{tan2020efficientdet} & & & & \\
\bottomrule
\end{tabular}
\end{adjustbox}
\caption{\bf An overview of the categorization of CNN-based object detection methods for object detection on pinhole camera images.}
\label{tab:object-detection-survey}
\end{table}
\subsection{Object Detection on Pinhole Cameras}

\subsubsection{Two-Stage Detectors}

The object detection task is divided into two stages in this approach: (i) extraction of Regions of Interest (ROIs) and (ii) classification and regression of the ROIs. Regions with CNN features (R-CNN) by Girshick~\etal~\cite{girshick2014rich} was the first to use a two-stage approach. It generates ROIs through selective search and classifies ROIs using a DCN-based classifier. It requires complex computations, which causes it to be slow and far from real-time capable. By extracting ROIs from feature maps, Fast R-CNN~\cite{girshick2015fast} and SPP-Net~\cite{he2015spatial} improved R-CNN~\cite{girshick2014rich}. SPP-Net used a Spatial Pyramid Pooling (SPP) layer to manage images of arbitrary sizes and aspect ratios. It applies an SPP layer over the feature maps produced by convolution layers and produces the fixed-length vectors required by fully connected layers. It removes the need for fixed-size inputs and can be used in any CNN-based classification model.
On the other hand, Fast R-CNN and SPP-Net are not end-to-end trainable since they depend on a region proposal approach. Faster R-CNN~\cite{ren2017faster} overcame this constraint by implementing the Region Proposal Network (RPN), which allowed end-to-end training. RPNs produce ROIs by regressing a series of reference boxes known as anchor boxes. R-FCN~\cite{dai2016r}, which replaces fully connected layers with Fully Convolutional Network (FCN), improves the efficiency of Faster R-CNN~\cite{ren2017faster}.\par
\subsubsection{Single-Stage Detectors}

In contrast to the two-stage method, single-stage detectors skip the RoI extraction stage and go straight to anchor box classification and regression. Specifically, the RoI pooling phase is omitted, and object detection is accomplished with a single network. Using a multi-scale, sliding window technique, Overfeat~\cite{sermanet2014overfeat} suggested a standardized framework to perform three tasks: classification, localization, and detection. "Overfeat," a feature extractor for vision applications, was introduced. You Only Look Once (YOLOv1)~\cite{redmon2016you} is a single-stage detector that divides the input image into grids and predicts the BB directly using regression and classification. YOLO9000 (YOLOv2)~\cite{redmon2017yolo9000} enhances efficiency by adding batch normalization and replacing YOLOv1's fully connected layers with anchor boxes for BB prediction. YOLOv3~\cite{YOLOV3}, which is quicker and more reliable than previous models, employs Darknet-53 as its feature extraction backbone. YOLOv3 can detect small objects with multi-scale predictions, which was a significant limitation in previous versions.\par

The Single Shot Multibox Detector (SSD)~\cite{liu2016ssd} overlays dense anchor boxes on the input image and extracts feature maps at different scales. The anchor boxes are then classified and regressed to predict BB. ResNet101~\cite{he2016deep} replaces the VGG network of SSD in DSSD~\cite{fu2017dssd}. It is then supplemented with a deconvolution module to merge feature maps from the beginning with the deconvolution layers. In terms of detecting small objects, it outperforms SSD. MDSSD~\cite{Cui2020MDSSDMD} expands DSSD with fusion blocks to handle feature maps at various scales. During training, RetinaNet~\cite{lin2017focal} implemented focal loss to resolve foreground and context class imbalance. It matches or exceeds the accuracy of cutting-edge two-stage detectors while operating at a faster rate. The design reuses RPN's 'anchors' and constructs an FCN with Feature Pyramid Network (FPN)~\cite{lin2017feature} on top of the ResNet backbone.\par

SqueezeDet~\cite{wu2017squeezedet} focuses on small model size, speed, and accuracy, making it ideal for object detection in autonomous driving. It used the Yolov1 detection pipeline and created a ConvDet layer to generate region proposals with fewer parameters than YOLOv1. EfficientDet~\cite{tan2020efficientdet} is a one-stage detector that is powered by EfficientNet~\cite{tan2019efficientnet}. It proposed a weighted bi-directional FPN to fuse multi-scale features and compound scaling, which jointly scales up all networks' depth, width, and resolution. Detection Transformer (DETR)~\cite{carion2020end} is a recent work on the direct set prediction paradigm. To provide unique predictions, it employs a transformer-based encoder-decoder architecture and a bi-partite matching loss function. On large objects, DETR outperforms Faster-RCNN~\cite{ren2017faster}.\par
\subsubsection{Multi-Stage Detectors}

As the name suggests, detectors in this group use a series of CNNs at different levels to detect objects. Cascade R-CNN~\cite{cai2018cascade} is a multi-stage R-CNN~\cite{girshick2014rich} extension. The training data is sampled at each point, and the IoU threshold is increased. It uses iterative BB regression to advance the hypotheses. CRAFT~\cite{yang2016craft} used a cascaded structure with RPN and two Fast R-CNNs~\cite{girshick2015fast} to introduce two functions, proposal generator and classifier, to improve the efficiency of proposal generation and detection. The CC-Net~\cite{ouyang2017chained} is made up of several cascade stages. It has two stages of the cascade (i) early cascade and (ii) contextual cascade. Since shallow layers reject easy ROIs, hard samples are easily managed by later stages. Fast R-CNN is also used as the foundation for a Multipath Net~\cite{zagoruyko2016multipath} with a few modifications such as skip connections, foveal regions, and improved loss functions. Information flows through several paths in the network, allowing the classifier to operate at various scales. Fast R-CNN suggested a multi-region CNN model with an iterative BB regression process that switches between box scoring and coordinate refinement. HyperNet~\cite{kong2016hypernet} is a multi-stage architecture that uses hyper feature maps to perform region proposal and object detection jointly. IoU-Net~\cite{jiang2018acquisition} achieves progressive BB regression using a standalone IoU predictor that can then be combined with any FPN-based CNN architecture for object detection.
\subsubsection{Deformable Part based Detectors}

Handling dynamic object deformation properties aids in improving detection efficiency. In~\cite{ouyang2013joint}, a deformation layer is proposed for pedestrian detection. Deformable Part Models (DPM)~\cite{girshick2015deformable} is a single CNN that constructs a distance transform pooling layer to map an input image pyramid to a detection score pyramid. DeepIDNet~\cite{ouyang2015deepid} proposed a CNN for object detection based on a deformable part by designing a def-pooling layer to learn the geometric deformations of all instances of a part. DP-FCN~\cite{mordan2017deformable} enhances it further by using a deformable part-based ROI pooling layer and deformation-aware localization. The model is fully convolutional end-to-end trainable. It focuses on discriminative elements. CNN's can learn dense spatial transformations with the aid of Deformable ConvNets~\cite{dai2017deformable}, which can then be used in object detection tasks. It introduced two modules: (i) deformable convolution and (ii) deformable ROI pooling, which aid CNNs in modeling geometric transformations efficiently.
\subsubsection{Keypoint based Detectors} 

In single-stage detectors, key points take the place of anchor boxes. To predict BBs, most one-stage detectors position dense anchor boxes over the image. RetinaNet~\cite{lin2017focal} and DSSD~\cite{fu2017dssd}, for example, require nearly 100k and 40k anchor boxes, respectively, thereby inducing hyperparameters. CornerNet~\cite{law2018cornernet} overcomes these limitations by detecting objects as paired key points representing the object's top left and bottom right corners. It employs an hourglass network as its backbone and employs corner pooling to locate corners. It uses associative embedding to group the key points. CenterNet~\cite{duan2019centernet} extends CornerNet by representing each object as a triplet – two corners and a central key point. It pioneered center pooling and cascade corner pooling to allow corners to recognize visual patterns of objects. ExtremeNet~\cite{zhou2019bottom}, on the other hand, detects BBs by looking at the topmost, leftmost, bottommost, rightmost, and the center of all objects.\par
\subsection{Object Detection on Fisheye Cameras}

Typically, automotive systems are equipped with a multi-camera network to cover the entire FoV around the vehicle~\cite{horgan2015vision}. The wide FoV of the fisheye image comes with the side effect of strong radial distortion. Objects at different angles from the optical axis look quite different, making the object detection task a larger challenge  than for pinhole cameras (see Figure~\ref{fig:object-detection-sample}). All CNN-based detectors are trained with standard pinhole camera images that are free of any optical aberrations and closely resemble objects' primary form and appearance in the real world. There have not been many attempts to specialize existing CNNs to detect deformed images produced and affected by fisheye lenses. A common practice is to rectify distortions in the image using a $4\textsuperscript{th}$ order polynomial~\cite{yogamani2019woodscape} model or unified camera model~\cite{khomutenko2016eucm}. However, undistortion comes with resampling distortion artifacts, especially at the periphery. In particular, the negative impact on computer vision due to the introduction of spurious frequency components is understood~\cite{LourencoSIFT}. Other more minor impacts include a reduced FoV and a non-rectangular image due to invalid pixels. Although semantic segmentation is an easier solution on fisheye images, object detection annotation costs are much lower~\cite{siam2017deep}.\par

Agarwal~\etal~\cite{agarwal2018recent} provides a detailed survey of current object detection methods and their challenges. Presumably, fisheye camera object detection is a much more complex problem. The rectangular bounding box (BB) fails to be a good representation due to the massive distortion in the scene. The size of the standard BBs in a fisheye image is almost double the size of the object of interest inside it. Instance segmentation can help to obtain accurate object contours. However, it is a different task that is computationally complex and more expensive to annotate. It also typically needs a BB estimation step. There are few works on object detection for fisheye camera images or closely related omnidirectional cameras. One of the main issues is the lack of a useful datasets, particularly for autonomous driving scenarios.
The recent fisheye object detection paper FisheyeDet~\cite{li2020fisheyedet} emphasizes the lack of a useful datasets, and they create a simulated fisheye camera dataset by applying distortions to the Pascal VOC dataset~\cite{everingham2010pascal}. FisheyeDet makes use of a 4-sided polygon representation aided by distortion shape matching. SphereNet~\cite{coors2018spherenet} and its variants~\cite{perraudin2019deepsphere, su2019kernel, jiang2019spherical} formulate CNNs on spherical surfaces. However, fisheye images do not follow spherical projection models, as seen by non-uniform distortion in horizontal and vertical directions. Deng~\etal~\cite{deng2017object} pioneered the use of deep learning to detect multi-class objects in fisheye images. Yang~\etal~\cite{yang2018object} compared the results of various detection algorithms that take equirectangular projection (ERP) images directly as inputs, demonstrating that the network produces only a certain accuracy without projecting ERP images into conventional 2D images.\par

Compared to the state-of-the-art approaches, in this thesis, we explore different object representations for fisheye object detection and design novel representations for fisheye images. We also release a dataset of 10,000 images with annotations for all the object representations. We perform an empirical study of our baseline, which can output different representations.
\section{Semantic Segmentation}

Semantic segmentation is the task of assigning dense semantic labels to images. It is of paramount significance in computer vision and has applications in autonomous driving, augmented reality, and human-computer interaction.

\subsection{Semantic Segmentation on Pinhole Cameras}

A multiscale convolutional network by Farabet~\etal~\cite{farabet2012learning} is fused with a segmentation framework in parallel (either superpixel or CRF-based). Because of a CRF block, computational efficiency is reduced. Pinheiro~\etal~\cite{pinheiro2014recurrent} built a recurrent architecture by using multiple instances of a CNN, each of which is fed with previous label predictions (obtained from the previous instance). There is a significant computational load when multiple instances (3 in their best-performing experiments) are fed. A standard milestone approach for semantic segmentation is the introduction of fully convolutional neural networks by Long~\etal~\cite{long2015fully}. The network architecture is a fully convolutional encoder structure (no fully connected layers) with skip connections at the final decision layer that fuse multiscale activations. Due to the lack of fully connected layers or a refinement block, it is comparably fast. DeepLabv1~\cite{chen2014semantic} a CNN with dilated convolutions is followed by a fully-connected (\ie, Dense) CRF. Near-real-time performance is achieved through fast and optimized computation~\cite{ulku2019survey}.

In parallel, layers of a pyramidal input are fed to separate FCNs for different scales by Eigen~\etal~\cite{eigen2015predicting}. These multiscale FCNs are also linked in series to provide pixel-by-pixel category, depth, and normal output simultaneously—lower computational efficiency as a result of progressive processing of a sequence of different scales. The UNet~\cite{ronneberger2015u} architecture uses an encoder-decoder (ED) structure with skip connections, connecting the same ED and final input-sized classification layer levels. Due to the lack of fully connected layers or a refinement block, the computation load is efficient. The SegNet~\cite{badrinarayanan2017segnet} structure is similar to UNet, but with skip connections that only transmit pooling indices (unlike U-Net, where skip connections concatenate the same level activations). Due to the lack of fully connected layers or a refinement block, the computation load is efficient. The DeconvNet~\cite{noh2015learning} structure is comprised of an ED (referred to as ‘the Conv./Deconv. Network') with no skip connections. The network's (convolutional) encoder component is transferred from the VGG-VD-16L~\cite{simonyan2015very}. Due to the lack of fully connected layers or a refinement block, the computation load is efficient. To perform pixel-wise labeling MSCG~\cite{yu2016multi}, designed multiscale context aggregation using only a rectangular prism of dilated convolutional layers without pooling or subsampling layers. Due to the lack of fully connected layers or a refinement block, the computation load is reduced. Zhen~\etal~\cite{zheng2015conditional} formulated a CRF-as-RNN, \ie, an FCN is followed by a CRF-as-RNN layer, which translates an iterative CRF algorithm into an RNN. Computational efficiency is limited due to the RNN block~\cite{ulku2019survey}.

FeatMap-Net~\cite{lin2016efficient} constitutes layers of a pyramidal input fed to parallel multiscale feature maps (\ie, CNNS), which were then fused in an upsample/concatenation (\ie, pyramid pooling) layer to provide the final feature map to a Dense CRF Layer. A well-thought-out but overburdened architecture results in moderate computational efficiency. Graph Long Short-Term Memory (LSTM)~\cite{liang2016semantic} is a LSTM generalization from sequential data to general graph-structured data for semantic segmentation, primarily of people. DAG-RNN~\cite{shuai2016dag} is a CNN+RNN network with a DAG structure that models long-term semantic dependencies among image units. The computational efficiency is significantly limited due to chain structured, sequential processing of pixels with a recurrent model. DeepLabv1 has been improved by Chen~\etal~\cite{chen2017deeplab}, with the addition of a 'dilated (atrous) spatial pyramid pooling (ASPP) layer. PSPNet~\cite{zhao2017pyramid} consists of a CNN followed by a pyramid pooling layer similar to~\cite{he2015spatial}, but it lacks a fully connected decision layer. As a result, computational performance is closer to FCN~\cite{long2015fully}. DeepLabv2 has been improved, with ASPP layer hyperparameters optimized and non-dense CRF layer, for faster operation by~\cite{chen2017rethinking}. Luo~\etal~\cite{luo2017deep} introduces dual learning for semantic image segmentation where one network predicts label maps/tags, while another uses these predictions to perform semantic segmentation. ResNet101~\cite{he2016deep} which is a larger backbone is used in both networks for preliminary feature extraction~\cite{ulku2019survey}.\par

GCN~\cite{peng2017large} uses large kernels to fuse high- and low-level features in a multiscale manner, powered by an initial ResNet-based~\cite{he2016deep} encoder. Stacked deconvolutional network \cite{fu2019stacked} is a UNET architecture made up of multiple shallow deconvolutional networks, known as SDN units, that are stacked one on top of each other to integrate contextual information and to ensure fine recovery of localized information. Discriminative Feature Network~\cite{yu2018learning} is made up of two sub-networks: Smooth Net (SN) and Border Net (BN). SN makes use of an attention module to handle global context, whereas BN makes use of a refinement block to handle borders. Due to an attention block, computational efficiency is limited. Multi-Scale Context Intertwining~\cite{lin2018multi} connects LSTM chains to aggregate features from different scales. Due to multiple RNN blocks, the computational efficiency is limited (\ie, LSTMs). DeepLab.v3+~\cite{chen2018encoder} is an improved version of DeepLab.v3 that employs a special encoder-decoder structure with dilated convolutions (rather than using Dense CRF for faster operation). Hierarchical Parsing Net~\cite{shi2018hierarchical} is a convolutional 'Appearance Feature Encoder,' while a 'Contextual Feature Encoder' made up of LSTMs generates superpixel features that are fed into a Softmax-based classification layer. Due to the use of multiple LSTMs, computational efficiency is limited. EncNet~\cite{zhang2018context} is a fully connected structure fed by dense feature maps (obtained from ResNet) and followed by a convolutional prediction layer to extract context. Fully connected layers limit computational performance within their "Context Encoding Module"~\cite{ulku2019survey}.

PSANet~\cite{zhao2018psanet} is a convolutional point-wise spatial attention (PSA) module connected to a pretrained convolutional encoder, allowing pixels to be interconnected via a self-adaptively learned attention map to provide global context. When compared to fully convolutional architectures (\eg, FCN), the addition of a PSA module reduces computational efficiency. EMANet152~\cite{li2019expectation} is made up of a novel attention module that converts input feature maps to output feature maps, providing global context. When compared to other attention governing architectures, it is computationally more efficient (\eg, PSANet). Kernel-Sharing Atrous Convolution~\cite{huang2019see} enables branches from different receptive fields to use the same kernel, allowing for more accessible branch communication and feature augmentation within the network. In Co-occurrent features (CFNet)~\cite{zhang2019co} a fine-grained spatial invariant representation is learned. The CFNet is constructed using a distribution of co-occurrent features for a given target in an image.\par

In this task, the right architectural design choice and multi-scale context modules are vital for performance. The former depend upon spatial pyramid pooling~\cite{he2015spatial, zhao2017pyramid} and atrous convolutions~\cite{chen2017rethinking, chen2017deeplab, chen2018encoder}. For the latter, the contemporary advancements in the design~\cite{huang2017densely, chollet2017xception} have improved the popular backbones~\cite{krizhevsky2012imagenet, simonyan2015very, he2016deep}. Despite their popularity, they require huge computational requirements and are not real-time capable for inference. The most popular design choice has been the encoder-decoder architecture~\cite{ronneberger2015u, badrinarayanan2017segnet}. To obtain state-of-the-art accuracy, recent approaches have incorporated Auto Machine Learning (AutoML)~\cite{liu2019auto, chen2018george}. ENet~\cite{paszke2016enet} is a compact real-time capable efficient network which follows an encoder-decoder architecture design. Bisenet~\cite{yu2018bisenet} is a two-path network designed to achieve fast inferences while obtaining high-resolution details. DABNet~\cite{li2019dabnet} combines the depth-wise separable filters and atrous-convolutions to obtain a decent trade-off between accuracy and efficiency. DfaNet~\cite{li2019dfanet} employs cascaded sub-stages to refine the segmentation predictions. To maximize larger networks' performance during inference, FCHardNet~\cite{chao2019hardnet} leverages on a new harmonic densely connected pattern.\par
\subsection{Semantic Segmentation on Fisheye Cameras}

Most of the popular semantic segmentation studies listed above are based on images taken by pinhole cameras. However, the urban traffic conditions are so complicated that more knowledge of the surroundings is needed, while the pinhole camera only has a narrow FoV. If a vehicle or pedestrian suddenly arrives from a blind spot, the safe operation of autonomous driving is hard to ensure. One of the methods is to enhance the significance of the information obtained for holistic scene comprehension. An Overlapping Pyramid Pooling module (OPP-Net) was presented by Deng~\etal~\cite{deng2017cnn} by employing several focal lengths to simulate different fisheye images with their corresponding annotations. In order to achieve real-time semantic segmentation, Saez~\etal~\cite{saez2019real} introduced an adaptation of Efficient Residual Factorized Network (ERFNet)~\cite{romera2017erfnet} to fisheye road images. The tests were performed on authentic fisheye images, but only qualitative results were revealed. Deng~\etal~\cite{deng2019restricted} used the identical approach to obtain road scene semantic segmentation of fisheye surround-view cameras using restricted deformable convolution. These models were trained on Cityscapes~\cite{cordts2016cityscapes} and SYNTHIA~\cite{ros2016synthia} datasets and tested on authentic fisheye images. However, to perform the segmentation directly on the fisheye images, we would require a large-scale finely-annotated fisheye image dataset to train the network. \textit{Henceforth, as a contribution to the research community, we release the WoodScape~\cite{yogamani2019woodscape} dataset.} Figure~\ref{fig:semantic-segmentation-sample} depicts the semantic segmentation task on the WoodScape dataset.\par
\subsubsection{Panoramic Images}

Xu~\etal~\cite{xu2019semantic} created a dataset of panoramic images by stitching images taken from different directions using synthetic images captured from SYNTHIA. The authors demonstrate that panoramic images improve segmentation results using these images. Yang~\etal~\cite{yang2019can} proposes a panoramic angular semantic segmentation framework by creating a data augmentation method by adding distortion to perspective images for the training set. After unfolding and partitioning the panoramic images, normal CNNs were used.\par
\subsubsection{Equirectangular Images}

Because of the simple transformation from spherical coordinates to planar coordinates, equirectangular representation is the most popular projection for $360\degree$ images. Classical CNNs designed for perspective images can be applied to equirectangular data. In polar regions, however, spherical input suffers from distortion. To address this issue, various approaches were proposed. SalNet360 was proposed by Monroy~\etal~\cite{monroy2018salnet360}, in which omnidirectional images were mapped to cube map by six faces projection and trained using standard CNNs to predict visual attention.
However, artifacts are produced when the cube map faces are recombined to create an omnidirectional image. Lai~\etal~\cite{lai2017semantic} converted panoramic videos to normal perspective images using semantic segmentation of equirectangular images. However, because highly accurate semantic labels was not required for this task, a frame-based fully convolutional network FCN~\cite{long2015fully} was used in this work. For the same task, they proposed the kernel transformer network, which efficiently transfers convolution kernels from perspective images to the equirectangular projection of $360\degree$ images. Tateno~\etal~\cite{tateno2018distortion} proposed a learning method for equirectangular images that uses a distortion-aware deformable convolution filter to estimate depth from a single image, and this method was also demonstrated on $360\degree$ semantic segmentation.\par
\subsubsection{Spherical Representations}

Due to distortions caused by the equirectangular representation, the most recent work on this topic has focused on the spherical presentation. Cohen~\etal~\cite{cohen2018spherical} created spherical convolutions by substituting sphere rotations for plane translations. Other works used the icosahedral spherical approximation, which is the most precise sphere discretization. To represent the discretization of the sphere, a spherical mesh is generated by dividing each face of a regular icosahedron into four equal triangles. In the case of triangle faces, Lee~\etal~\cite{lee2019spherephd} proposed an orientation-dependent kernel method, which was demonstrated through classification, detection, and semantic segmentation. Zhang~\etal~\cite{zhang2019orientation} proposed an orientation-aware CNN framework for semantic segmentation on omnidirectional images using icosahedron spheres. UGSCNN was proposed by Jiang~\etal~\cite{jiang2019spherical} to train spherical data mapped to an icosahedron mesh by replacing conventional convolution kernels with linear combinations of learnable weighted operators.\par
\section{Motion Segmentation} 
\label{sec:motion-related-work}

The classical approach to detecting moving objects is based on the scene's geometrical understanding, where the ego-vehicle motion and the displacement vectors of the pixels between two frames are known. Classical methods for moving object detection based on the geometrical understanding of the scene have been suggested, such as~\cite{menze2015object}, which was used to estimate object motion masks. To model the background motion in terms of homography, Wehrwein~\etal~\cite{wehrwein2017video} introduced assumptions about the camera motion model. Due to the errors caused by the restricted assumptions, such as camera translations, this method cannot be used in autonomous driving applications. Classical approaches have a lower performance than deep learning methods and a high level of difficulty due to the complicated pipelines used. The method of Menze~\etal~\cite{menze2015object}, for example, has a running time of $50$ minutes per frame, making it unsuitable for use in a real-time application such as autonomous driving~\cite{rashed2019fusemodnet}.\par

Jain~\etal~\cite{jain2017fusionseg} introduced a method for generic foreground segmentation that takes advantage of optical flow. This work is intended for generic object segmentation and does not concentrate on object classifications as \emph{Moving} or \emph{Static}. Drayer~\etal~\cite{drayer2016object} proposed an R-CNN detection-based video segmentation algorithm. Due to its sophistication, the technique is not feasible for autonomous driving applications, where it takes $8$ seconds to infer an image. Arguably the most famous constraint used in motion detection is the epipolar constraint~\cite{dey2012detection, clarke1996detection}, which can be combined with additional geometrical constraints to detect multiple types of motion~\cite{klappstein2006monocular}. However, even if the moving objects' geometry is well known, their detection still presents challenges caused by intrinsic geometrical limitations. Many methods for segmenting moving objects from stationary camera images have been suggested in~\cite{spagnolo2006moving, gao2010moving}. However, they cannot be directly applied to moving camera images because movement causes a dual motion appearance consisting of background motion and object motion. Methods that detect motion from freely moving cameras, in general, divide the image into coherent regions with homogeneous motion. The image is divided into the background and moving clusters during this process. These approaches can be divided into two types: optical flow-based and tracking-based approaches. Optical flow-based techniques~\cite{ranjan2019competitive, tosi2020distilled} determine whether a region's motion speed and direction are compatible with its radially surrounding pattern. Tracking-based methods, on the other hand~\cite{choi2012general, drayer2016object, lin2014deep, kundu2009moving}, tend to track and localize target points in successive frames. Object tracking produces movement trajectories, and by estimating the camera's ego-motion, objects can be separated from the background motion. Large processing pipelines are common in these methods, resulting in long computation times and coarse segmentations.\par

Chen~\etal~\cite{chen2016object} suggests a method for detecting object-level motion from a moving camera using two consecutive image frames and outputs 2D BBs. They create a robust context-aware motion descriptor that considers movement speed and object direction and combines it with an object classifier. The descriptor calculates the discrepancy between local optical flow histograms of objects and their surroundings, yielding a state of motion measurement. Dinesh~\etal~\cite{reddy2014semantic} propose a method for generating motion likelihoods based on depth and optical flow estimations while incorporating semantic and geometric constraints within a dense CRF. Fan~\etal~\cite{fan2016semantic} suggested a multistep framework in which first sparse image features in two consecutive stereo image pairs are extracted and matched. RANSAC is then used to classify the matched feature points as inliers caused by the camera and outliers caused by moving objects. The outliers are then clustered in a U-disparity chart, which provides object motion information. Finally, the motion information is combined with the semantic segmentation given by an FCN using a dense CRF. Long run times, ranging from a few seconds to minutes, are a major drawback of these methods, rendering them unsuitable for applications that involve near-real-time efficiency, such as autonomous driving. In the search to overcome the limitations of the classical approach, there has been good work in using CNN to solve the moving object detection problem, such as MODNet~\cite{siam2018modnet}, FisheyeMODNet~\cite{yahiaoui2019fisheyemodnet}, MPNet~\cite{wang2018mpnet}, OmegaNet~\cite{tosi2020distilled}, SMSnet~\cite{vertens2017smsnet} and Ranjan~\etal~\cite{ranjan2019competitive}. Siam~\etal~\cite{siam2018modnet, siam2018real} investigated motion segmentation using deep network architectures, but these networks depend solely on camera RGB images, which can fail in low-light conditions.\par

Given the use of fisheye cameras in surround-view systems, it is of utmost importance for research to explore this direction and provide a CNN architecture for moving object detection on fisheye images. One of the main challenges of detecting moving objects with a CNN is to make it scene agnostic so that the detection is based only on motion cues \& not on appearance cues. Figure~\ref{fig:motion-sample} depicts the motion segmentation task on the WoodScape dataset.\par
\section{Soiling Segmentation}
\label{sec:soiling-related-work}

There is very little work on the related but distinct lens soiling problem.  There are two types of soiled areas: opaque (mud, dust, snow) and translucent (water). Transparent soiling, in particular, can be challenging to detect due to the background's partial visibility. The two problems are similar in how they degrade image quality and can severely affect visual perception performance. However, there are substantial differences. The first significant difference is that soiling on the lens can be removed by a camera cleaning system that either sprays water or uses a more sophisticated ultrasonic hardware~\cite{uvrivcavr2019soilingnet}. Secondly, there is temporal consistency for soiling where mud or water droplets remain static typically or sometimes have low-frequency dynamics of moving water droplets compared to higher variability in adverse weather scenes. Thus this temporal structure can be exploited further for soiling scenarios. Finally, soiling can cause more severe degradation as opaque mud soiling can completely block the camera. Porav~\etal~\cite{porav2019can} discussed transparent soiling in a recent study in which a stereo camera was used in combination with a dripping water supply to simulate raindrops on the camera lens.
The authors also suggest a CNN-based de-raining algorithm. Sakaridis~\etal~\cite{sakaridis2018semantic} suggested a robust semantic segmentation algorithm that can handle foggy scenes. However, dealing with opaque soiling in this manner would be difficult. To boost the image quality, the third solution is to run a separate image restoration algorithm. De-raining \cite{li2019single, porav2019can, ren2019progressive, wang2019spatial, yang2019frame, yang2019scale} is a recent example of restoration algorithms in automotive scenarios. This will primarily help with partial soiling. Restoration algorithms may be single-image or video-based. The latter is more computationally costly, but it can benefit from the visibility of soiling occluded regions over time~\cite{uricar2019desoiling}.\par

In this thesis, we focus on the generic soiling detection task. Even disregarding camera cleaning, soiling detection is still needed to increase vision algorithms' uncertainty in the degraded areas. A more formal introduction to the soling detection and categorization is provided in~\cite{uvrivcavr2019soilingnet}. The problem is formalized as a multi-label classification task and discusses soiling detection applications, including camera cleaning. The authors present a proof of concept idea
on how GANs~\cite{goodfellow2014generative} could be applied for dealing with the insufficient data problem in terms of an advanced data augmentation. The authors also outline another potential usage of GANs in the AD area. U\v{r}i\v{c}\'{a}\v{r} \etal~\cite{uricar2019desoiling} provided a desoiling dataset benchmark. SoildNet was explored in~\cite{das2019soildnet} for embedded platform deployment.\par
\section{Multi-Task Learning}

\emph{Multi-task learning} (MTL) is carried out by learning commonly shared representations from multi-task supervisory signals. Many dense prediction tasks, \ie tasks that generate pixel-level predictions, have seen substantial performance improvements since the introduction of deep learning. Typically, these tasks are learned one at a time, with each task requiring its own neural network to be trained. However, by jointly tackling multiple tasks via a learned shared representation, recent MTL techniques have shown promising results regarding performance, computational complexity, and memory footprint.\par

MTL~\cite{caruana1998multitask} seeks to enhance generalization by incorporating domain-specific knowledge found in similar task training signals. MTL refers to the design of networks capable of learning mutual representations from multi-task supervisory signals. Compared to the single-task example, where each network solves only one task, multi-task networks provide several benefits. First, the resulting memory footprint is significantly reduced due to their intrinsic layer sharing. Second, since they explicitly avoid calculating the features in the shared layers multiple times for each task, they demonstrate faster inference speeds. Most significantly, whether the related tasks exchange complementary knowledge or function as a regularizer for one another, they have the potential to improve results~\cite{vandenhende2020multitask}.\par
\textbf{Non-Deep Learning-Based Methods}

Before the deep learning era, MTL works attempted to model the common information among tasks to improve generalization performance through joint task learning. To do so, they imposed constraints on the task parameter space, such as: task parameters should be close to each other in terms of some distance metric \cite{evgeniou2004regularized,  xue2007multi, jacob2008clustered, zhou2011clustered}, share a common probabilistic prior~\cite{bakker2003task,yu2005learning,lee2007learning,daume2009bayesian,kumar2012learning}, or reside in a low-dimensional subspace~\cite{argyriou2008convex,liu2012multi,jalali2010dirty} or manifold~\cite{agarwal2010learning}. These assumptions work well when all tasks are related~\cite{evgeniou2004regularized,ando2005framework,argyriou2008convex,rai2010infinite} and regularly scheduled, but they can degrade performance if the information is shared between unrelated tasks. The latter is a well-known MTL issue known as "\emph{negative transfer}." To address this issue, some of these studies chose to group tasks based on prior assumptions about their similarity or relatedness~\cite{vandenhende2020multitask}.\par
\textbf{Distilling Task Predictions in Deep Learning}
\label{subsubsec:distilling}

All of the works in Section~\ref{sec:hard-soft-parameters-mtl}: soft vs. hard have one thing in common: they \textit{directly} predict all task outputs from the same input in a single processing cycle. On the other hand, several recent studies used a multi-task network to make initial task predictions. They then used features from these initial predictions to improve each task output in a one-off or recursive manner. PAD-Net~\cite{xu2018pad} proposed using spatial attention to distill information from initial task predictions of other tasks before adding it as a residual to the task of interest. JTRL~\cite{zhang2018joint} chose to predict each task by using information from previous predictions to refine the features of another task at each iteration. PAP-Net~\cite{zhang2019pattern} built on this concept by employing a recursive procedure to propagate similar cross-talk and task-specific patterns discovered in the initial task predictions. They did so by using the affinity matrices of the initial predictions rather than the features themselves, as was the case previously~\cite{xu2018pad,zhang2018joint}. By separating inter-and intra-task patterns from each other, Zhou~\etal~\cite{zhou2020pattern} refined the use of pixel affinities to distill the information. To explicitly model the unique task interactions that occur at each scale, MTI-Net~\cite{vandenhende2020mti} used a multi-scale multi-modal distillation procedure~\cite{vandenhende2020multitask}.\par

Multi-task networks have traditionally been classified as using soft or hard parameter sharing techniques, as explained in Section~\ref{sec:hard-soft-parameters-mtl}. However, several recent works drew inspiration from both groups of works to collaboratively solve multiple pixel-level tasks. As a result, whether the soft vs. hard parameter sharing paradigm should still be used as the primary framework for classifying MTL architectures is debatable. An alternative taxonomy distinguishes between various architectures based on the interactions between tasks, \ie, network locations where information and features are shared or exchanged between tasks.\par

\begin{figure}[!t]
	\centering
	\subfigure[][Encoder-focused model]
	{\includegraphics[width=0.48\linewidth]{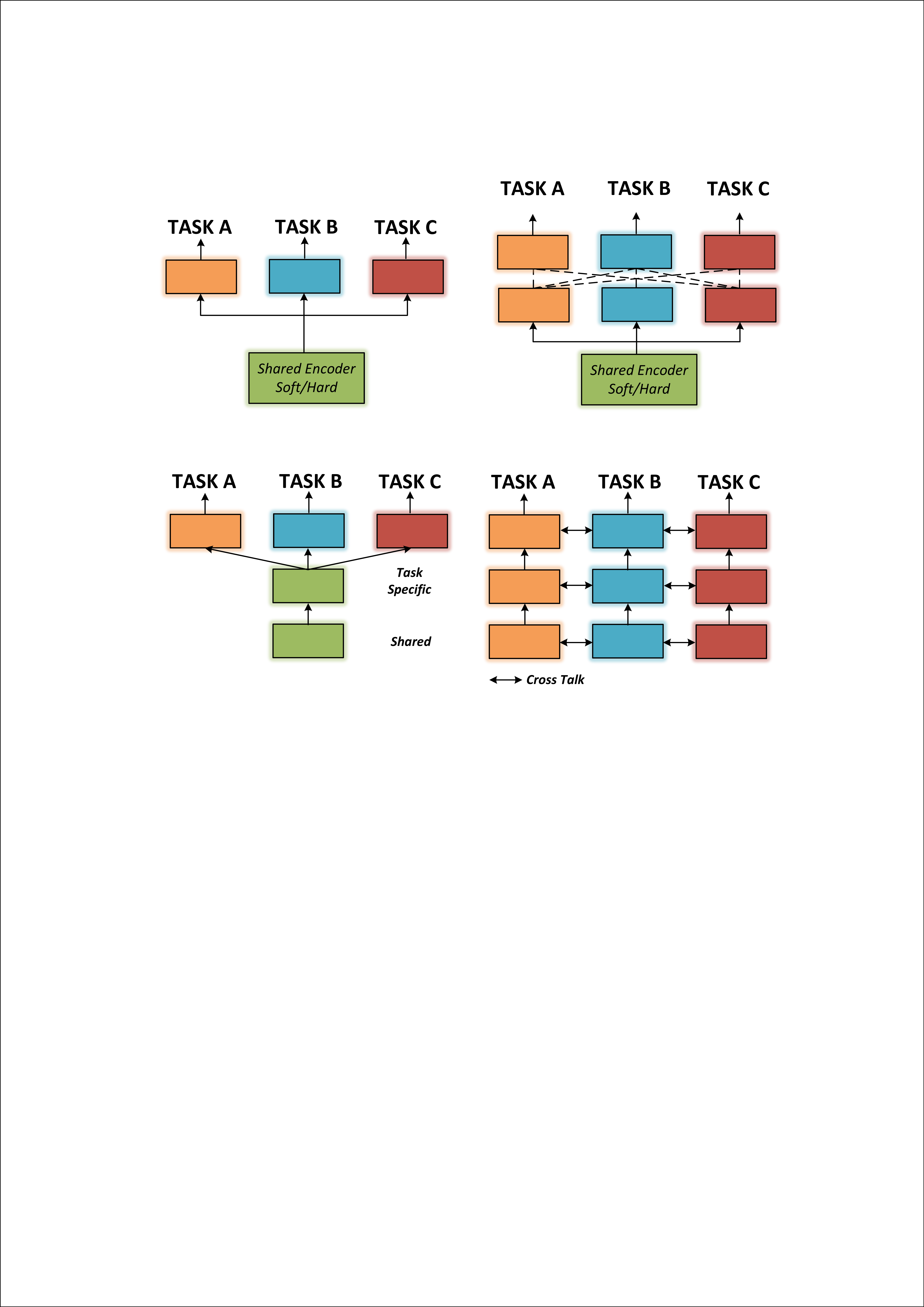}}\;\;
	\subfigure[][Decoder-focused model]
	{\includegraphics[width=0.48\linewidth]{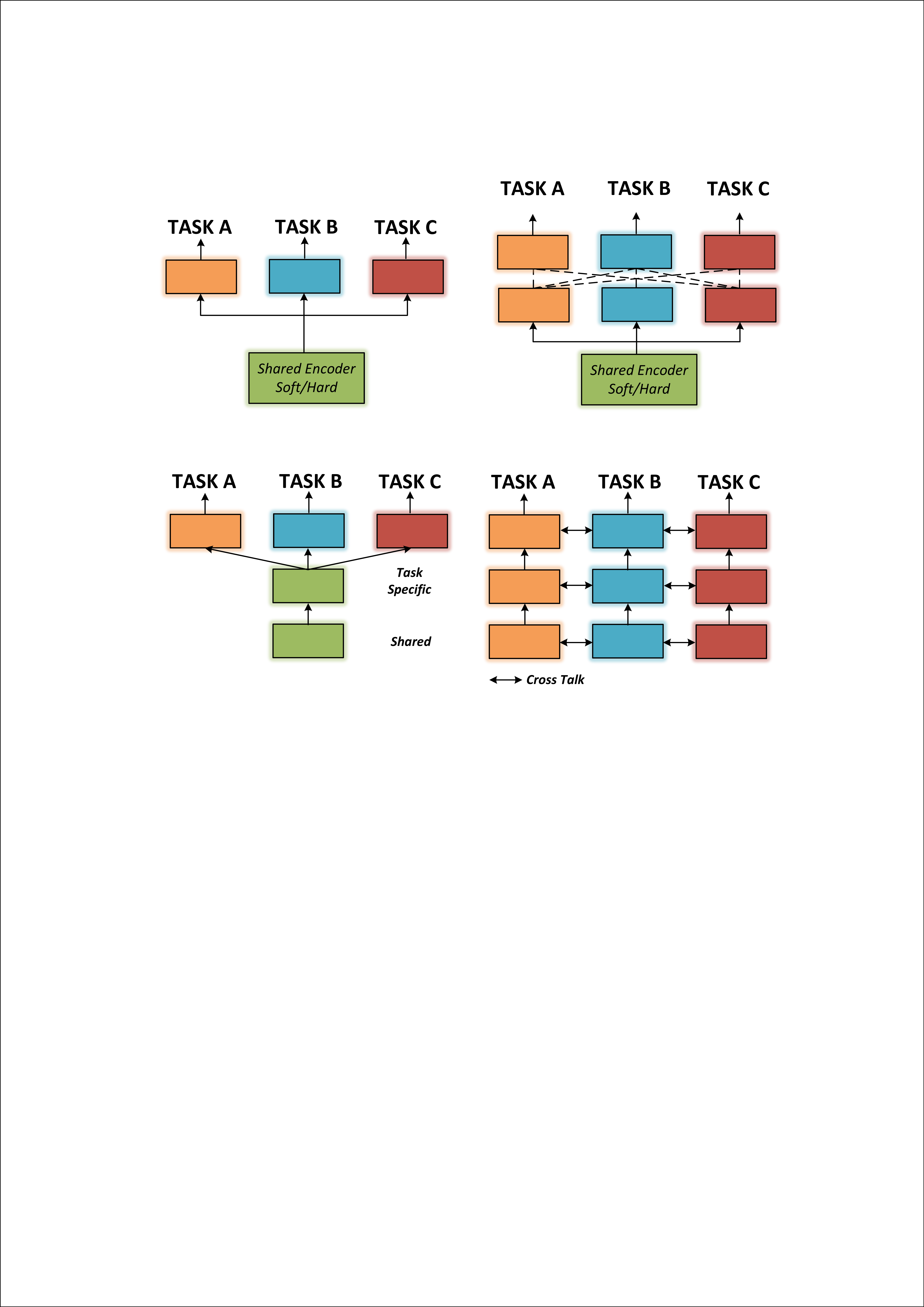}}
	\caption{\textbf{Illustration of the encoder and decoder-focused models depending on where the task synergies take place.}}
	\label{fig:mtl_arch}
\end{figure}

MTL is typically divided into two sections based on the proposed criterion: shared parameters and task-specific parameters. We allow the shared parameters to learn representing the commonalities between many tasks, while task-specific parameters are learned to perform independent task-specific processing. Shared parameters are commonly dubbed as \textit{encoders} (see Figure~\ref{fig:mtl_arch}) wherein they carry out the key feature extraction. The task-specific parameters are called \textit{decoders} (see Figure~\ref{fig:mtl_arch}) as they decode the vital information from the encoders. This method presents benefits such as enhanced data efficiency, reduced overfitting through shared representations, and fast learning by leveraging auxiliary information~\cite{crawshaw2020multi}. It is an efficient design pattern commonly used where most of the computation can be shared across all tasks~\cite{sistu2019neurall, vandenhende2020multitask}. Besides, learning features for multiple tasks can act as a regularizer, to improve generalization.\par
\subsection{Encoder-focused Architectures}
\label{subsec:encoder-focused}

The information is shared only in the encoder by employing either hard or soft-parameter sharing by the encoder-focused architectures before decoding each task with an independent task-specific head. Conversely, the decoder-focused architectures further exchange information during the decoding stage. Most recent works~\cite{neven2017fast, Kendall2018, chen2018gradnorm, sener2018multi, teichmann2018multinet} followed an ad-hoc approach by sharing an off-the-shelf backbone model in union with small task-specific heads. The activations were shared amongst all single-task models in the encoder by, \eg, cross-stitch networks  proposed in~\cite{misra2016cross}. Yuan~\etal~\cite{gao2019nddr} developed neural discriminative dimensionality reduction CNNs, a similar architecture as cross-stitch networks. It used a dimensionality reduction mechanism rather than employing a linear combination to fuse all single-task networks' activations. In combination with task-specific attention modules in the encoder, Liu~\etal~\cite{ liu2019end} introduced multi-task attention networks which employed a shared backbone model.\par
\subsection{Decoder-focused Architectures}
\label{subsec:decoder-focused}

Regarding decoder-focused architectures, PAD-Net~\cite{xu2018pad} was one of the first approaches. It performs multi-modal distillation through a spatial attention mechanism. Zhang~\etal~\cite{zhang2019pattern} introduced Pattern-Affinitive Propagation Networks (PAP-Net), which incorporated a similar architecture as PAD-Net. PAP-Net involves leveraging
pixel affinities in order to perform multi-modal distillation. Joint Task-Recursive Learning (JTRL)~\cite{zhang2018joint} recursively estimates two tasks at frequently higher scales to slowly improve the decisions based on past states. The architecture is similar to PAD-Net and PAP-Net; a multi-modal distillation mechanism is employed to couple information from earlier task predictions, through which later estimates are improved. Conversely, the JTRL model estimates two tasks sequentially rather than parallel and in an intertwined manner. In the above-listed decoder-focused works, multi-modal distillation was performed at a fixed scale, \ie, the backbone's last layer features. Nevertheless, Multi-Scale Task Interaction Networks (MTI-Net)\cite{vandenhende2020mti} indicated that this is somewhat a strict assumption. MTI-Net explicitly took into account task interactions at multiple scales.\par
\subsection{Other Approaches}

To enable synergies in the decoding stage, Multilinear relationship networks~\cite{long2017learning} employed tensor normal priors to the parameter set of the task-specific heads. Compared to~\cite{misra2016cross, gao2019nddr} where layers are aligned and shared according to the standard parallel ordering scheme, Elliot~\etal~\cite{meyerson2018beyond} proposed soft layer ordering a flexible sharing scheme across tasks and network depths. Yang~\etal~\cite{yang2016deep} generalized matrix factorization methods to MTL to learn cross-task sharing structures in every layer of the model. Clemens~\etal~\cite{rosenbaum2018routing} proposed routing networks as a principled way to determine the connectivity of a model's function blocks through routing. By learning binary masks, Piggyback~\cite{mallya2018piggyback} demonstrated how to adapt a single, fixed neural network to a multi-task network. Huang~\etal~\cite{huang2018gnas} presented an approach rooted in neural architecture search for the automated creation of a tree-based multi-attribute learning model. Bragman~\etal~\cite{bragman2019stochastic} re-formulated the convolution kernels in each layer of the network with stochastic filter groups to support either shared or task-specific behavior. In a similar style, Newell~\etal~\cite{newell2019feature} exhibited feature partitioning approaches to designate the convolution kernels in each layer of the model into different tasks. Overall, these works have a different scope within MTL, \eg, automate the network architecture design. Furthermore, they focus on solving multiple (binary) classification tasks rather than numerous dense prediction tasks. As a result, they fall outside the scope of this thesis.\par

Maninis~\etal~\cite{maninis2019attentive} presented \emph{Attentive Single-Tasking of Multiple Tasks} to take a single-tasking route for the MTL problem. \ie inside an MTL framework, they made separate forward passes, one for each task, that initiate shared responses among all tasks, complemented by residual responses that are task-specific. Moreover, adversarial training on the gradient level was applied to be statistically identical across tasks to overcome the negative transfer problem. A benefit of this method is that shared and task-specific knowledge inside the model can be easily disentangled. On the negative side, the tasks can not be estimated altogether, but only sequentially, which significantly increases the inference speed and somehow defies the purpose of MTL.\par

Recently, Mao~\etal~\cite{mao2020multitask} illustrated that multi-task learning improves adversarial robustness, which is critical for safety applications. In the automotive multi-task setting, MultiNet~\cite{teichmann2018multinet} was one of the first to demonstrate a three task network on KITTI, and most further works have primarily worked on a three task setting. In contrast to letting a network predict a single task, it is also possible to train a network to predict several tasks at once (see Figure~\ref{fig:mtl_poster}). It has shown to improve tasks such as, \eg, semantic segmentation, \cite{klingner2020improved, klingner2020self, sistu2019neurall, chennupati2019multinet++, chennupati2019auxnet}, domain adaptation~\cite{Bolte2019a, Ochs2019, Zhao2019}, instance segmentation: \cite{Kirillov2019}, and depth estimation \cite{eigen2015predicting, Kendall2018, Wang2020a, lu2020multi}.\par
\subsection{Previous MTL based Semantically-Guided Distance Estimation}

As the distance predictions are still imperfect due to the monocular cues such as occlusion, blur, haze, and different lighting conditions and the dynamic objects during the self-supervised optimizations between consecutive frames. Many approaches consider different scene understanding modalities, such as segmentation~\cite{Meng2019a, guizilini2019packnet, ranjan2019competitive} or optical flow~\cite{luo2019every, Chen2019b} within multi-task learning to guide and improve the distance estimation. As optical flow is usually also predicted in a self-supervised fashion~\cite{Liu2019}, it is, therefore, subject to similar limitations as the self-supervised distance estimation, which is why we focus on the joint learning of self-supervised distance estimation and semantic segmentation.\par

To improve the distance estimation task in MTL, many recent approaches aim to integrate optical flow into the self-supervised distance estimation training. This additional task can also be trained in a self-supervised fashion~\cite{Liu2019, Rhe2017}. In these approaches, both tasks are predicted simultaneously. Then losses are applied to enforce cross-task consistency~\cite{Liu2019a, luo2019every, Wang2019c, Yang2018c}, to enforce known geometric constraints~\cite{Chen2019b, ranjan2019competitive}, or to induce a modified reconstruction of the warped image~\cite{Chen2019b, yin2018geonet}. Although the typical approach is to compensate using optical flow, in this thesis we propose an alternative method to use semantic/motion segmentation instead for two reasons. (i) Semantic segmentation is a mature and common task in autonomous driving, which can be leveraged. (ii) Motion segmentation is an easier problem to solve as it falls under segmentation tasks (similar to semantic segmentation). It can be learned in a supervised fashion compared to optical flow, which is computationally more complex and harder to validate because of difficulties in obtaining ground truth.\par

Depth has no mathematical relationship with semantics. Several works \cite{Meng2019a, guizilini2019packnet, ranjan2019competitive}, however, follow this direction on the basis of the concept that semantic can guide depth estimation by offering specific cues. Sky, for example, should be placed far away and naturally with very high depth values. A shift in pixel mark will most likely signify an object's boundary, resulting in a noticeable change in depth. Several recent approaches also used semantic or instance segmentation techniques to identify moving objects and handle them accordingly inside the photometric loss \cite{Casser2019a, casser2019depth, Meng2019a, Vijayanarasimhan2017, guizilini2019packnet}. To this end, the segmentation masks are either given as an additional input to the network \cite{Meng2019a, guizilini2019packnet} or used to predict poses for each object separately between two consecutive frames \cite{casser2019depth, Casser2019a, Vijayanarasimhan2017} and apply a separate rigid transformation for each object. Avoiding an unfavorable two-step (pre)training procedure, other approaches in \cite{Chen2019a, Novosel2019, Yang2018b, Zhu2020} train both tasks in one multi-task network simultaneously, improving the performance by cross-task guidance between these two facets of scene understanding. Moreover, the segmentation masks can be projected between frames to enforce semantic consistency \cite{Chen2019a, Yang2018b}, or the edges can be enforced to appear in similar regions in both predictions \cite{Chen2019a, Zhu2020}. In this thesis, we propose to use this warping to discover frames with moving objects and learn their depth from these frames by applying a simple semantic masking technique.\par
\subsection{Optimization in MTL}
\label{sec:optimization-in-mtl}

\begin{table}[t]
\centering
\begin{adjustbox}{width=\textwidth}
\setlength{\tabcolsep}{0.2em}
\begin{tabular}{@{}lcclccl@{}}
\toprule
  \textbf{Method} &
  \cellcolor[HTML]{00b0f0} \textit{\begin{tabular}[c]{@{}c@{}} Balancing \\ Magnitudes\end{tabular}} &
  \cellcolor[HTML]{00b050} \textit{\begin{tabular}[c]{@{}c@{}} Balance \\ 
  Learning\end{tabular}} &
  \cellcolor[HTML]{ab9ac0} \textit{Prioritize} &
  \cellcolor[HTML]{7d9ebf} \textit{\begin{tabular}[c]{@{}c@{}} Gradients \\ Required\end{tabular}} &
  \cellcolor[HTML]{a5a5a5} \textit{\begin{tabular}[c]{@{}c@{}} No Extra \\ Tuning \end{tabular}} &
  \multicolumn{1}{c}{\cellcolor[HTML]{e5b9b5} \textit{Motivation}} \\ 
  \midrule
Uncertainty~\cite{Kendall2018}    & \ch  & \xm & Low Noise  & \xm & \ch & Homoscedastic uncertainty      \\
GradNorm~\cite{chen2018gradnorm}  & \ch  & \ch &            & \ch & \ch & Balance learning \& magnitudes \\
DWA~\cite{liu2019end}             & \xm  & \ch &            & \xm & \xm & Balance learning               \\
DTP~\cite{guo2018dynamic}         & \xm  & \xm & Difficult  & \xm & \xm & Prioritize difficult tasks     \\ 
\bottomrule
\end{tabular}
\end{adjustbox}
\caption[\textbf{Ablation of different task balancing techniques.}]
{\textbf{Ablation of different task balancing techniques~\cite{vandenhende2020multitask}.} Firstly, we contemplate if an approach balances the loss magnitudes (\emph{\textcolor[HTML]{00b0f0}{Balance Magnitudes}}) and/or the speed at which tasks are learned (\emph{\textcolor[HTML]{00b050}{Balance Learning}}). Also, we attest what tasks are prioritized during the training stage (\emph{\textcolor[HTML]{ab9ac0}{Prioritize}}) and followed by if the approach needs access to the task-specific gradients (\emph{\textcolor[HTML]{7d9ebf}{Gradients Required}}). Finally, we consider if the suggested approach needs additional tuning, \eg manually determining the weights, and KPIs (\emph{\textcolor[HTML]{a5a5a5}{No Extra Tuning}}).}
\label{tab:task-weighing-ablation}
\end{table}
In the previous section, we reviewed literature modeling MTL architectures that can learn multiple tasks together. However, a vital challenge in MTL arises from the optimization method itself. In particular, we need to thoughtfully weigh the joint learning of all tasks to circumvent a state where one or more tasks have an imperative influence on the network weights. This section considers various techniques that have studied this \textit{task weighing} problem. While initial works did weigh losses \cite{eigen2015predicting} or gradients \cite{Ganin2015} by an empirical factor, current approaches can estimate this scale factor automatically \cite{Kendall2018, chen2018gradnorm}.\par

\textbf{Homoscedastic uncertainty} to was incorporated to weigh the single-task losses by Kendall \etal~\cite{Kendall2018}. The homoscedastic uncertainty or \textit{task-dependent uncertainty} is not an output of the network rather a quantity that remains constant for different input samples of the corresponding task. Chen \etal~\cite{chen2018gradnorm} proposed \textbf{Gradient normalization} (GradNorm) to command the training of multi-task networks by stimulating the \textit{task-specific gradients} to be of similar magnitude. This encourages the network to learn all the assigned tasks at an equal pace. Similar to GradNorm, Liu \etal~\cite{liu2019end} proposed \textbf{Dynamic Weight Averaging} (DWA) to weigh the pace at which different tasks are learned. Compared to GradNorm, DWA requires only the task-specific loss values during training and eliminates the need to obtain the task-specific gradients separately by performing backward passes each time. The task weighting techniques~\cite{Kendall2018, chen2018gradnorm, liu2019end} chose to optimize the task-specific weights as part of a Gaussian likelihood objective. In contrast, \textbf{Dynamic Task Prioritization} (DTP)~\cite{guo2018dynamic} chose to prioritize the learning of 'challenging' tasks by designating them a higher task-specific weight.
The motive is that the network should pay more effort to learn the 'challenging' tasks. This approach is entirely opposite to Kendall's uncertainty weighting, where a higher weight is assigned to the 'easy' tasks. Kendall's approach suits better when tasks have noisy labeled data, while DTP performs better when we have the availability to clean ground-truth labels. Finally, the ablation of all the listed approaches is shown in Table~\ref{tab:task-weighing-ablation}.\par
\section{Adversarial Attacks}

For the first time, Szegedy~\etal~\cite{szegedy2013intriguing} demonstrated box-constrained L-BFGS, \ie, small perturbations in the images so that perturbed imagery can fool deep learning models. Fast gradient sign method (FGSM)~\cite{goodfellow2015explaining} is an example of a simple yet effective attack for generating adversarial instances. FGSM aims to fool the image classification by adding a small vector obtained by taking the sign of the gradient of the loss function. Moreover, it was shown that robust 3D adversarial objects could fool deep network classifiers in the physical world~\cite{athalye2018synthesizing}, despite the combination of viewpoint shifts, camera noise, and other natural transformations. The one-step methods cause images to be perturbed by taking a single significant step towards the classifier's loss (\ie, one-step gradient descent). They are iteratively taking multiple small steps while adjusting the direction after each step is an intuitive extension of this idea. This is precisely what the Basic Iterative Method \cite{kurakin2016adversarial} does. Kurakin~\etal~\cite{kurakin2016adversarial} also extended BIM to the Iterative Least-likely Class Method (ILCM).\par

Papernot~\etal~\cite{papernot2016limitations} also devised a jacobian-based saliency map attack (JSMA), an adversarial attack by limiting the perturbation's \lzero-norm. Physically, the goal is to change only a few pixels in the image rather than disrupt the entire image to fool the classifier. When only one pixel in an image is changed to fool the classifier, this is an extreme case of an adversarial attack. Surprisingly, Su~\etal~\cite{su2019one} claimed that by changing just one pixel per image, they were able to fool three different network models on 70.97\% of the images tested. Carlini and Wagner (C\&W)~\cite{carlini2016towards} proposed a set of three adversarial attacks in the aftermath of defensive distillation against adversarial perturbations~\cite{papernot2016distillation}. These attacks make perturbations almost imperceptible by limiting their \lzero, \ltwo, and \linfinity norms. It is demonstrated that defensive distillation for the targeted networks almost completely fails against these attacks. Moosavi-Dezfooli~\etal~\cite{moosavi2016deepfool} proposed to iteratively compute a minimal norm adversarial perturbation for a given image. DeepFool, their algorithm, starts with a clean image assumed to be in a region bounded by the classifier's decision boundaries. Whereas methods like FGSM~\cite{goodfellow2015explaining}, ILCM~\cite{kurakin2016adversarial}, DeepFool~\cite{moosavi2016deepfool}, and others compute perturbations to fool a network on a single image, Moosavi-Dezfooli~\etal~\cite{moosavi2017universal} compute 'universal' adversarial perturbations that can fool a network on 'any' image with high probability~\cite{akhtar2018threat}.\par

Sarkar~\etal~\cite{sarkar2017upset} proposed two black-box attack algorithms for targeted fooling of deep neural networks: UPSET: Universal Perturbations for Steering to Exact Targets and ANGRI: Antagonistic Network for Generating Rogue Images. Cisse~\etal~\cite{cisse2017houdini} proposed 'Houdini,' a method for deceiving gradient-based learning machines by generating adversarial examples that can be tailored to task losses. Houdini has also been demonstrated to be capable of successfully attacking a popular deep Automatic Speech Recognition system~\cite{amodei2016deep}. Feed-forward neural networks were trained by Baluja and Fischer~\cite{baluja2017adversarial} to generate adversarial examples against other targeted networks or sets of networks. Adversarial Transformation Networks were the name given to the trained models (ATNs). Hayex and Danezis~\cite{hayes2017machine} also used an ATN to learn adversarial examples for black-box attacks in the same direction~\cite{akhtar2018threat}. The summary of characteristics of various attacking methods are shown in Table~\ref{tab:adversarial-attacks-works}.\par
\begin{table}[t]
\centering
\begin{adjustbox}{width=0.8\columnwidth}
\begin{tabular}{@{}llcccl@{}}
\toprule
\textbf{Method}
& \cellcolor[HTML]{00b0f0} \textit{
\begin{tabular}[c]{@{}c@{}} Black/ \\ White box\end{tabular}} 
& \cellcolor[HTML]{00b050} \textit{
\begin{tabular}[c]{@{}c@{}} Image/ \\ Universal \end{tabular}}
& \cellcolor[HTML]{ab9ac0} \textit{
\begin{tabular}[c]{@{}c@{}} Perturbation \\ Norm \end{tabular}}
& \cellcolor[HTML]{7d9ebf} \textit{Learning} 
& \cellcolor[HTML]{a5a5a5} \textit{Strength} \\
\midrule
L-BFGS~\cite{szegedy2013intriguing} & \textit{wb\_target} & Image 
& $\ell_{\infty}$ & One shot  & $* * *$ \\
FGSM~\cite{goodfellow2015explaining} & \textit{wb\_target} & Image
& $\ell_{\infty}$ & One shot  & $* * *$ \\
BIM \& ILCM~\cite{kurakin2016adversarial} & \textit{wb\_untarget} & Image
& $\ell_{\infty}$ & Iterative & $* * * *$ \\
JSMA~\cite{papernot2016limitations} & \textit{bb\_untarget} & Image
& $\ell_{0}$ & Iterative & $* * *$ \\
C\&W attacks~\cite{carlini2016towards} & \textit{wb\_untarget} & Image
& $\ell_{2}, \ell_{\infty}$ & Iterative & $* * * *$ \\
DeepFool~\cite{moosavi2016deepfool} & \textit{bb\_target} & Universal 
& $\ell_{2}, \ell_{\infty}$ & Iterative & $* * * * *$ \\
\begin{tabular}[c]{@{}l@{}} Universal \\ perturbations~\cite{moosavi2017universal} \end{tabular}
& \textit{bb\_target} & Image& $\ell_{\infty}$ & Iterative & $* * * *$ \\
UPSET~\cite{sarkar2017upset} & \textit{bb\_target} & Image& $\ell_{2}, \ell_{\infty}$ & Iterative & $* * * *$ \\
ANGRI~\cite{sarkar2017upset} & \textit{wb\_target} & Image& $\ell_{\infty}$ & Iterative & $* * * *$ \\ 
\bottomrule
\end{tabular}
\end{adjustbox}
\caption[\bf Summary of the characteristics of various attacking methods.]{\textbf{A summary of the characteristics of various attacking methods:} The ‘perturbation norm' denotes the perturbations' restricted p-norm in order to make them imperceptible. The strength (higher for more asterisks) is based on the review of the literature~\cite{akhtar2018threat}.}
\label{tab:adversarial-attacks-works}
\end{table}
\subsubsection{Attacks on Geometric and Semantic Tasks}

Inspired by Moosavi-Dezfooli~\etal~\cite{moosavi2017universal}, Metzen~\etal~\cite{lu2017safetynet} demonstrated the existence of image-agnostic quasi-imperceptible perturbations that can fool a deep neural network into significantly corrupting the predicted image segmentation. Furthermore, they demonstrated that it is possible to compute noise vectors that can remove a specific class from the segmented classes while leaving most of the image segmentation intact (\eg, removing pedestrians from road scenes). Even though the "space of adversarial perturbations for semantic image segmentation is presumably smaller than image classification," the perturbations have been shown to generalize well for unseen validation images with high probability. Arnab~\etal~\cite{arnab2018robustness} investigated FGSM~\cite{goodfellow2015explaining} based adversarial attacks for semantic segmentation and discovered that several findings of these attacks for classification do not directly transfer to the segmentation task. Xie~\etal~\cite{xie2017adversarial} computed adversarial examples for semantic segmentation and object detection, observing that these tasks can be formulated as classifying multiple targets in an image - the target in segmentation is a pixel or a receptive field, and the target in detection is an object proposal. According to this viewpoint, their method, 'Dense Adversary Generation,' optimizes a loss function over a set of pixels/proposals to generate adversarial examples. The generated examples are tested for their ability to fool various deep learning-based segmentation and detection approaches. Their experimental results show that the generated perturbations not only fool the targeted networks but also generalize well across different network models~\cite{akhtar2018threat}. In addition, research on adversarial attacks for monocular depth estimation began in 2019. Van~\etal~\cite{dijk2019neural} developed some exceptional cases to investigate the internal mechanism of how networks perceive depth from images by constructing prominent fake images that can be identified at first glance. Yamanaka~\etal~\cite{yamanaka2020adversarial} demonstrated that the target area's depth is incorrectly estimated by overwriting the local area of the input image. Mopuri~\etal~\cite{mopuri2017fast} proposed a data-free method for creating universal adversarial examples for a specific CNN. Although their method effectively uses depth estimation and semantic segmentation, it is limited to a single network. It cannot produce universal examples that attack both tasks from different structures at the same time. Hu~\etal~\cite{hu2019analysis} investigated white-box adversarial attacks (I-FGSM) on depth estimation in an indoor setting and proposed a saliency map defense.\par
Fooling surveillance cameras was introduced in~\cite{thys2019fooling} where adversarial patches are designed to attack person detection. DAG algorithm~\cite{xie2017adversarial} is an example of generating adversarial attacks for semantic segmentation and object detection tasks. It was discovered that the perturbations are exchangeable across different networks, even though they were trained differently since they share some intrinsic structure that makes them susceptible to a common source of perturbations. In addition to camera sensors, potential vulnerabilities of LiDAR-based autonomous driving detection systems are explored in~\cite{cao2019adversarial}. Moreover, the 3D-printed adversarial objects showed effective physical attacks on LiDAR equipped vehicles, raising concerns about autonomous vehicles' safety. Robust Physical Perturbations (RP$_2$)~\cite{eykholt2018robust} is another example that generates robust visual adversarial perturbations under different physical conditions on road sign classifications.\par
\subsubsection{Defenses Against Adversarial Attacks}

On the other hand, adversarial robustness and defense methods of neural networks have been studied to improve these networks' resistance to different adversarial attacks. One method for defense is adversarial training, where adversarial examples besides the clean examples are used to train the model. Adversarial training can be seen as a sort of simple data augmentation. Despite being simple, it cannot cover all attack cases. In~\cite{dziugaite2016study}, it is demonstrated that JPEG compression can undo the small adversarial perturbations created by the FGSM. However, this method is not adequate for large perturbations. Xu~\etal~\cite{xu2018feature} proposed Feature-squeezing for detecting adversarial examples. The model is tested on both the original input and the input after being pre-processed by feature squeezers such as spatial smoothing. If the output difference exceeds a certain threshold, we identify the input as an adversarial example. Defense-GAN~\cite{samangouei2018defense} is another defense technique that employs generative adversarial networks (GAN)s~\cite{goodfellow2014generative}, in which it seeks a similar output to a given picture while ignoring adversarial perturbations. It is shown to be a feasible defense that relies on the GAN's expressiveness and generative power. However, training GANs is still a challenging task. Robust attacks and defenses are still challenging tasks and an active area of research. Most previous works on adversarial attacks focused on single task scenarios. However, in real-life situations, multi-task learning is adopted to solve several tasks at once. Accordingly, multi-task networks leverage the shared knowledge among tasks, leading to better performance, reduced storage, and faster inference. Moreover, it is shown that when models are trained on multiple tasks at once, they become more robust to adversarial attacks on individual tasks~\cite{mao2020multitask}. However, defense remains an open challenge.\par
\chapter{Geometric Tasks}
\label{Chapter4}
\minitoc
\section{Problem Definition}

In this chapter, the focus is on solving one of the most challenging perception problems for an autonomous car: \textit{to predict distance of vehicles around it}. With the knowledge from \textbf{Chapter}~\ref{Chapter2} wherein we discussed the geometry of the camera models in detail, we incorporate the camera model into the core of the CNN framework. We present a novel self-supervised scale-aware framework for learning Euclidean distance and ego-motion by exploiting geometrical constraints in a sequence of images extracted from raw monocular fisheye videos without applying rectification. This work was formally presented as \textit{FisheyeDistanceNet}~\cite{kumar2020fisheyedistancenet} as an oral at the \href{https://arxiv.org/abs/1910.04076}{ICRA} conference in 2020.\par

Building upon the success of this paper, we generalize the training framework to work with any camera model and propose a fully differentiable architecture that estimates the distance directly from raw unrectified images (shown in Figure~\ref{fig:unrectdepthnet-model_arch}) without the need for any pre-processing. This work was formally presented as \textit{UnRectDepthNet}~\cite{kumar2020unrectdepthnet} as an oral at the \href{https://arxiv.org/abs/2007.06676}{IROS} conference in 2020.\par

The method at the time of publication of these papers was state-of-the-art on KITTI and WoodScape datasets (refer Section~\ref{sec:benchmarks}) and showcased that it is possible to obtain distance maps on raw fisheye images without the need for rectification. This opens the door to rethink the need for rectification for fisheye images, and an in-detailed discussion about the motivation and problems encountered due to rectifying images is presented in Section~\ref{sec:motivation-for-working-on-raw-images}.\par
\section{Why is Predicting Depth so Difficult?}
\label{sec:depth-challenges}

Before we dive deep into the depth estimation framework, we try to grasp some of the primary depth estimation issues in this section. Some of the complex challenges that must be solved include correspondence matching, which can be difficult due to factors such as textureless regions, occlusion, non-Lambertian surfaces, and resolving ambiguous solutions, in which several 3D scenes can give the same scene on the image plane, implying that estimated depth is not unique. The key culprit is the loss of depth information when 3D views are projected to 2D images. Another issue arises when there are motion and dynamic objects. \emph{Lambertian surfaces} are those that tend to have the same brightness regardless of where they are viewed from. Owing to non-ideal diffuse reflection, the resulting brightness intensity in pictures depicting the same scene from two different perspectives may not be equal. \emph{Occlusion} occurs when an object is occluded in one view but not the other, and \emph{textureless region} occurs when several pixels have the same pixel intensity. The ability to retrieve distance information from a camera is very appealing due to its low production cost and dense representation. For the time being, the best alternative way to retrieve depth is to use an active range sensor such as Light Detection and Ranging (LiDAR). They are naturally high precision sensors that provide highly accurate depth information~\cite{depth_basics_2021}.\par

\subsubsection{Depth Estimation is an Ill-Posed Problem}

\begin{figure}[!t]
  \centering
  \begin{minipage}[t]{0.498\textwidth}
    \centering
    \includegraphics[width=\textwidth]{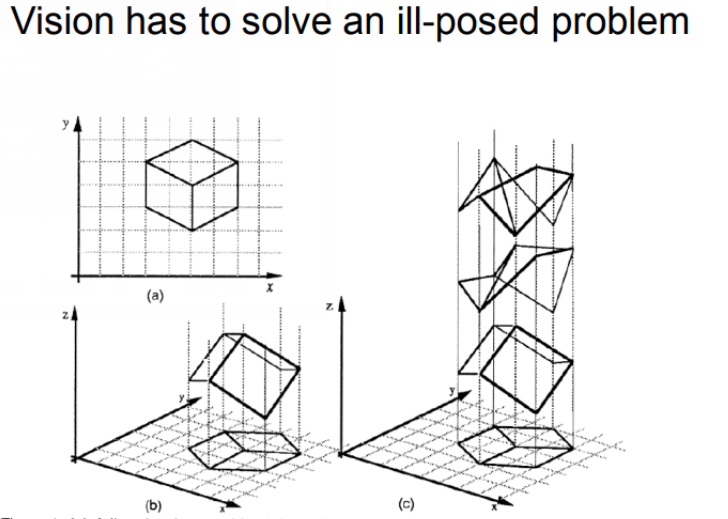}
    \caption[\bf An ill-posed problem in vision.]{\textbf{An ill-posed problem in vision}. Figure reproduced from~\cite{vision_ill_posed_2021}.}
  \label{fig:vision-ill-pose}
  \end{minipage}%
  \hfill
  \begin{minipage}[t]{0.498\textwidth}
    \centering
    \includegraphics[width=\textwidth]{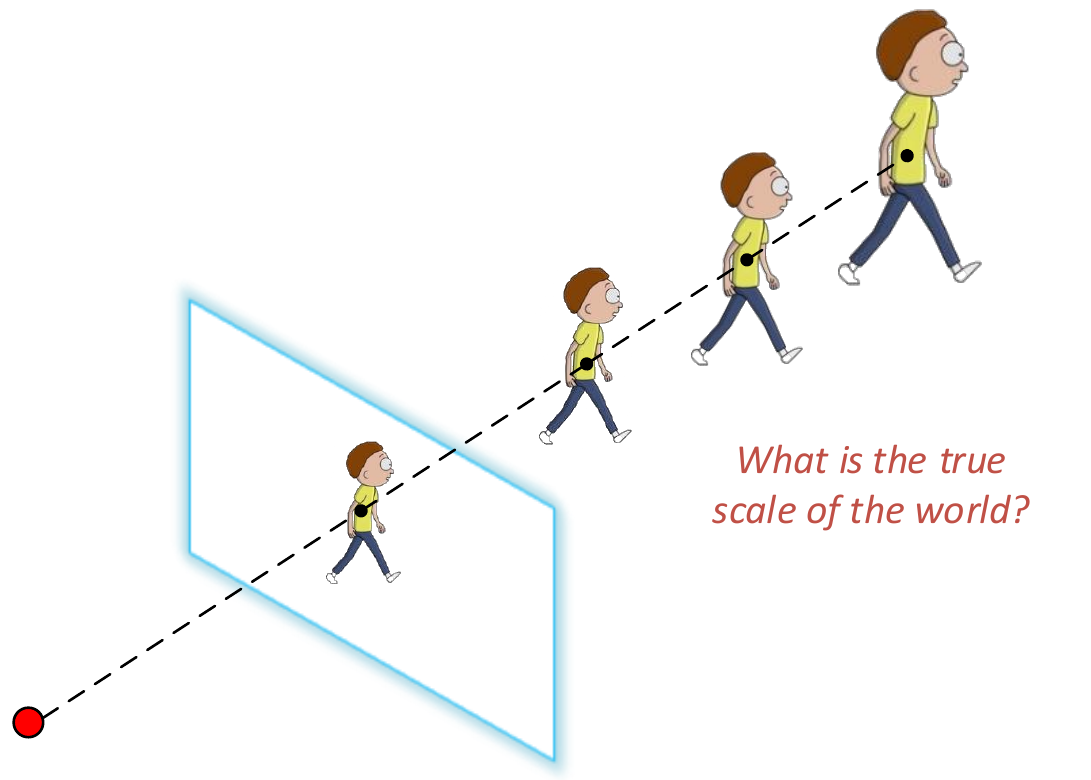}
    \caption{\bf Illustration of scale-ambiguity in depth estimation.}
    \label{fig:scale-ambiguity}
  \end{minipage}
\end{figure}

Many authors~\cite{garg2016unsupervised, zhou2017unsupervised, godard2019digging} would note that the problem of estimating depth from a single RGB image is an ill-posed inverse problem when researching monocular depth estimation. \ie, several 3D scenes observed in the world may also correspond to the same 2D plane as shown in Figures~\ref{fig:vision-ill-pose} and~\ref{fig:scale-ambiguity}. Figure~\ref{fig:vision-ill-pose} illustrates that (a) A line drawing provides information only about the x, y coordinates of points lying along the object contours. (b) The human visual system is usually able to reconstruct an object in three dimensions given only a single 2D projection (c) Any planar line drawing is geometrically consistent with infinitely many 3D structures.
\par
\textbf{Scale-Ambiguity for Monocular Depth Estimation:} For any given camera model, adjusting the focal length will proportionately scale the points on the image plane (see Figure~\ref{fig:scale-ambiguity}). Let us suppose we scale the entire scene points $X$, by a factor $k$ and, at the same time, scale the camera matrices $P$, by a factor of $1/k$, the projections of the scene points in the image remain exactly the same \ie,
\begin{align}
\label{eq:scale-ambiguity}
    x &= PX  \nonumber \\
      &= \frac{1}{k} P * kX = x
\end{align}
Eq.~\ref{eq:scale-ambiguity} depicts that we can never recover the exact scale of the actual scene from the image alone with monocular approaches. This does not preclude us from making specific predictions, but it does suggest that all depth values would be relative to one another. As a result, an absolute value is needed to serve as an anchor point, a measurement from another dedicated sensor, in order to obtain actual depth estimation. The depth can then be calculated by dividing the approximate anchor value by the measured one. Some of the possible solutions are:
\begin{itemize}
    \item Using an external sensor, such as LiDAR, Time-of-Flight, or stereo cameras, to measure the distance at least in one point. It is not a simple solution, and it necessitates embedded integration and accurate calibration to associate this calculation with the correct pixel in the image plane.
    \item If we assume depth consistency across training and testing dataset. It can be handy in datasets with high pose variability, such as KITTI \cite{geiger2013vision}, where the camera is \textit{still} at the same height and looking at the ground from the same perspective. This assumption will fail for drones and unmanned aerial vehicles.
    \item We can measure the movement magnitude employing dedicated sensors, such as an IMU or GPS in a vehicle, or speed from the wheels in a car, and compare it to apparent calculated movement, assuming that both depth and apparent movement are consistently estimated. The scale factor would be the resulting ratio.
\end{itemize}
The last option is preferred because movement estimation is a critical calculation in navigation for autonomous vehicles, and it is thus almost always possible to obtain the speed information from the odometry data from a car~\cite{clement19}.\par
\begin{figure}[!t]
  \centering
    \includegraphics[width=\textwidth]{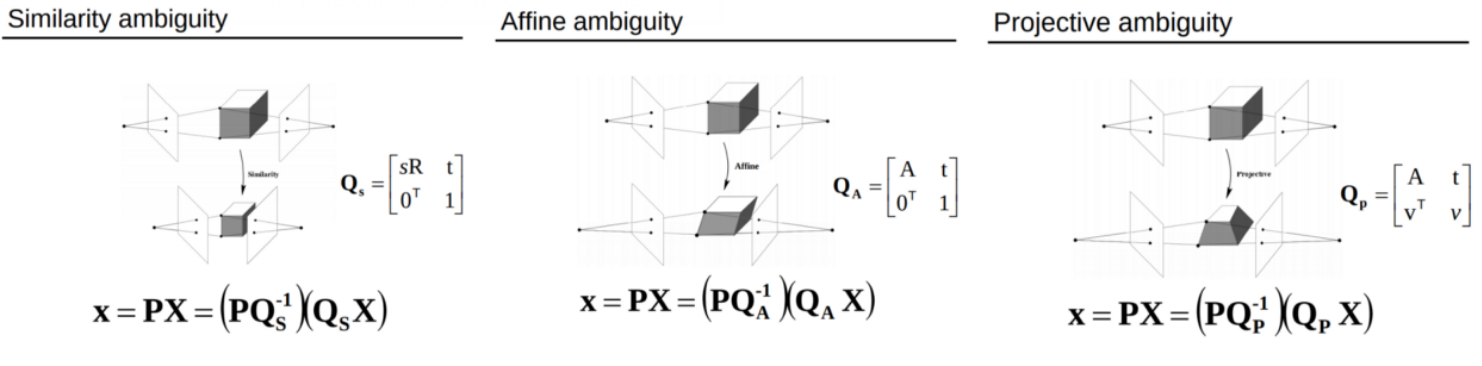}
    \caption[\bf Projection of an object after a transformation maps it to the same point in the plane.]{Projection of object after transformation maps to the same point in the plane. Figure reproduced from~\cite{projection_ambiguity_2021}.} 
    \label{fig:projection-ambiguity}
\end{figure}

\textbf{Ill-pose: Projection ambiguity} If we perform a geometric transformation on the scene as shown in Figure~\ref{fig:projection-ambiguity}, these points would likely map to the same location on the plane after the transformation. Once again, we are faced with the same problem~\cite{depth_basics_2021}.\par
\textbf{Dynamic Objects Violate Static World Assumption:} Dynamic objects in the scene complicate the estimation process even more for the \emph{SfM} framework. A moving camera and a series of static scenes are used to estimate depth via structure from motion. This assumption must hold true for pixel matching and alignment when there are moving objects in the scene, this assumption fails~\cite{depth_basics_2021}.\par
\textbf{Why is Predicting Distance on Fisheye Cameras Even More Difficult?}

Most state-of-the-art depth estimation works~\cite{garg2016unsupervised, zhou2017unsupervised, monodepth17, godard2019digging} have solely focused on the moderate FoV/pinhole camera models as described in Section~\ref{sec:pinhole-model} wherein the networks are limited to work only on rectified image sequences. As discussed in Section~\ref{sec:fisheye-cameras} fisheye cameras undergo large distortions. The \emph{SfM} framework and its core view synthesis approach are believed to work only on undistorted image pairs. For a pinhole projection model, $depth \propto 1 / disparity$. Henceforth, the network's sigmoid output $\sigma$ can be converted to depth with $D = 1 / ({a\sigma + b})$, where $a$ and $b$ are chosen to constrain $D$ between $0.1$ and $100$ units~\cite{godard2019digging}. For a spherical image, we can only obtain angular disparities~\cite{arican2007dense} by rectification. To perform distance estimation on raw fisheye images, we would require metric distance values to warp the source image $I_{t'}$ onto the target frame $I_t$. Due to the limitations of the monocular \emph{SfM} objective, both the monocular distance predictor $g_d$ and ego-motion predictor $g_\mathbf{x}$ predict \textit{scale-ambiguous} values as discussed in Section~\ref{sec:depth-challenges}. This would make it impossible to estimate distance maps on fisheye images.\par

Distance estimation, especially in the context of autonomous vehicles, has proven to be difficult due to a variety of factors discussed earlier, such as occlusion, dynamic objects in the scene, and imperfect stereo correspondence. The biggest enemy for stereo matching algorithms is a reflective, translucent, or mirror surface. For example, the windshield of a car often degrades matching and thus results in erroneous estimation. As a result, most companies continue to rely on LiDAR to extract distance reliably. However, the latest trend in the autonomous vehicle perception stack goes towards sensor fusion since each sensor has a distinct advantage in its feature extraction methods. Nonetheless, since the emergence of Deep Learning, this field has gained significant momentum and achieved impressive results. Many studies have been conducted to address these problems~\cite{depth_basics_2021}.\par

In this thesis, we will solve some of the critical problems of the vision community by proposing novel solutions to tackle the monocular Distance estimation's scale factor issues and overcome the community's notion of employing \emph{SfM} framework only on rectified sequences. We showcase that the \emph{SfM} approach can be extended to raw distorted fisheye camera images. \par
\subsubsection{Motivation for Working on Raw Fisheye Images}
\label{sec:motivation-for-working-on-raw-images}

Rectification is considered to be a fundamental step in dense depth estimation~\cite{hartley2003multiple}. In stereo cameras, epipolar rectification is performed to enable matching only in one direction along the horizontal scanline. This approach can also be extended to monocular cameras using two consecutive frames giving rise to motion stereo. These rectification steps also require the removal of non-linear distortion. Although it is convenient to work with rectilinear projections, there are practical issues that arise due to rectification. Rectification has also been transferred to CNN-based approaches as an inductive bias to simplify the learning. Yadati \etal~\cite{yadati2017multiscale} demonstrate that CNN-based two-view depth estimation is challenging without rectification and attempt to solve it in a more specific setting. To the best of our knowledge, all the methods reported on KITTI make use of barrel distortion corrected images. Automotive cameras such as fisheye surround-view cameras exhibit a strong distortion, and it is not easy to rectify their images. Recently, several tasks such as motion segmentation~\cite{yahiaoui2019fisheyemodnet} and soiling detection~\cite{uvrivcavr2019soilingnet} were demonstrated on fisheye images without rectification.\par

\begin{figure}[!t]
  \centering
  \newcommand{\turnwidth}{0.485\columnwidth}

\newcommand{\imlabel}[2]{\includegraphics[width=0.488\columnwidth]{#1}
\raisebox{2pt}{\makebox[-2pt][r]{\footnotesize #2}}}

\begin{tabular}{@{\hskip 0mm}c@{\hskip 1.5mm}c}
\centering
    \imlabel{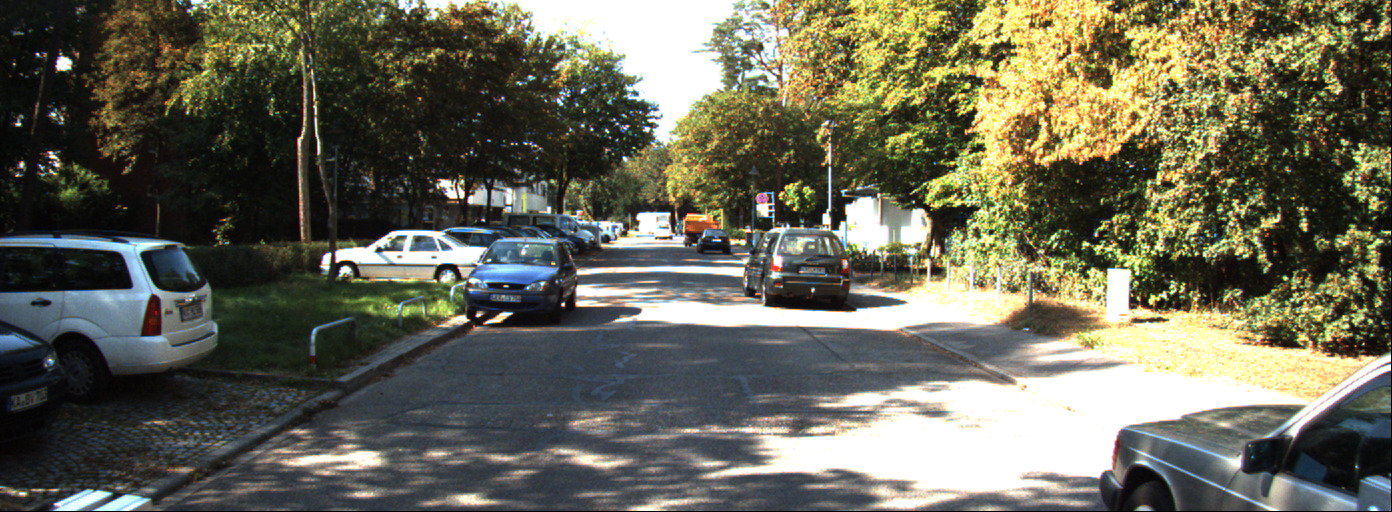} &
    \imlabel{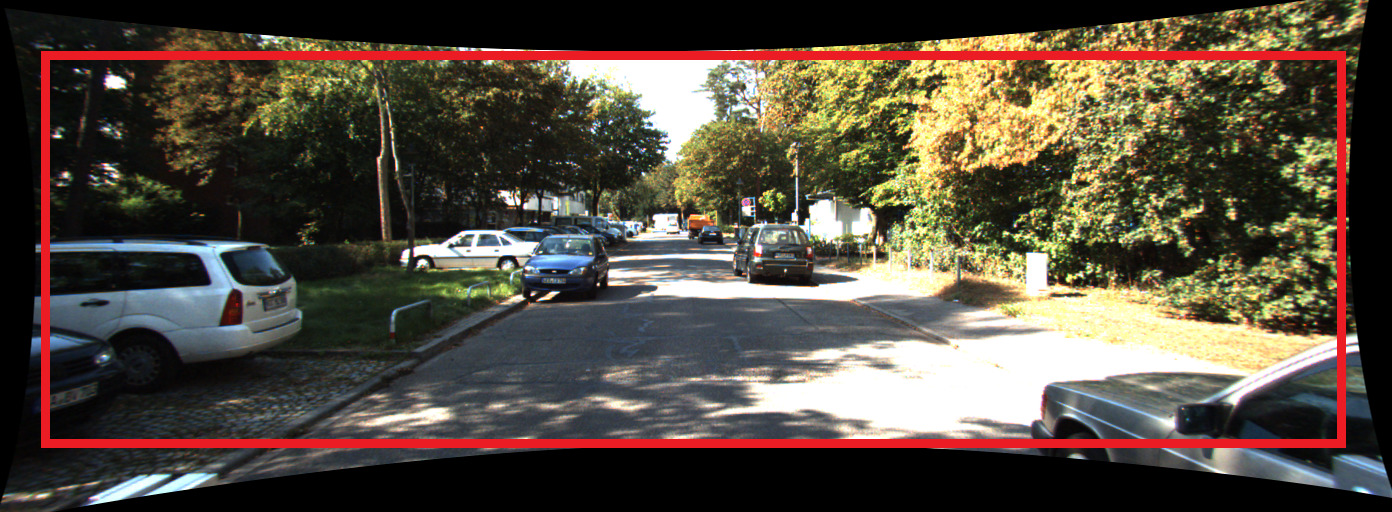}
    {} \\
    \imlabel{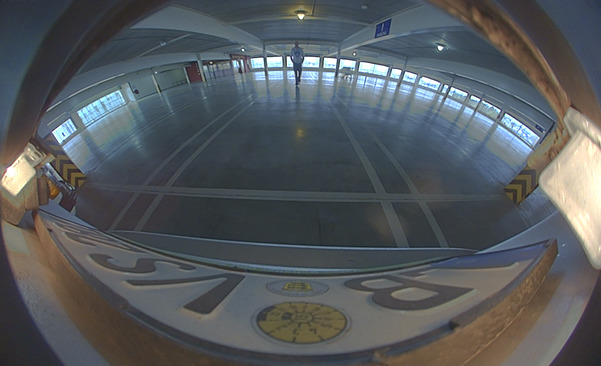}
    {} &
    \imlabel{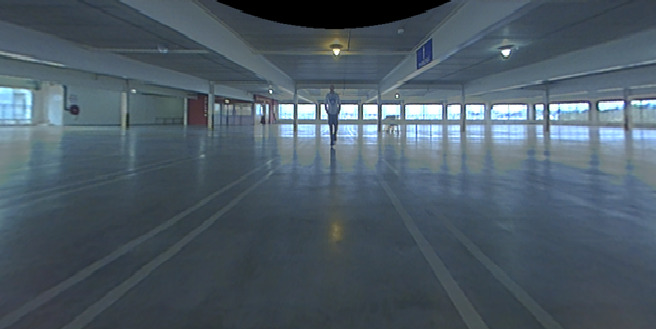}
    {} \\
\end{tabular}
  \caption{\bf Illustration of distortion correction in KITTI and WoodScape datasets.}
  \label{fig:undistort}
\end{figure}

\textbf{Practical Problems Encountered:} Real-world automotive cameras have lens distortion, and the typical approach is to remove the distortion and then apply standard camera projection models. However, in practice, this has several issues that are not dealt with in literature. Figure~\ref{fig:undistort} illustrates the rectification used in KITTI and WoodScape datasets. The first row shows a raw KITTI image with barrel distortion and the corresponding rectified image. The red box is used to crop out black pixels in the periphery, causing a loss in FoV. The second row shows a raw WoodScape image with strong fisheye lens distortion and the corresponding rectified image exhibiting a drastic loss of FoV. In the KITTI dataset, due to the barrel\footnote{KITTI~\cite{geiger2013vision} refers to it as pincushion distortion because of an error in the OpenCV documentation, which was fixed later.} distortion effect of the camera, images have been rectified and cropped to $1242\times375$ pixels. Cropping is performed after rectification to get a rectangular grid without any black pixels in the periphery. Thus, the rectified images' size is smaller than that of the raw images with $1392\times512$ pixels. Based on the number of the non-black, occupied pixels removed by the cropping, roughly $10\%$ of the image information is lost. This effect becomes more drastic for the WoodScape images with a much larger radial distortion where more than $30\%$ of the image information is lost\footnote{Other quasi-linear rectification methods like cylindrical rectification will preserve more information at cost of additional distortion.}. For a horizontal FoV greater than $180^\circ$, there are rays incident from behind the camera, making it theoretically impossible to establish a complete mapping to a rectilinear viewport. Thus the rectification defeats the purpose of using a wide-angle fisheye lens.\par
Reduced FoV is the most critical problem of undistortion, but there are further practical issues. The first one is resampling distortion, which is caused by interpolation errors during the warping step. This effect can be partially mitigated by a more advanced interpolation method~\cite{meijering2002chronology}. However, it is extreme in the periphery of fisheye lenses because a small region is expanded to a larger one in the warped image. Besides, the warping step is needed at inference time, which consumes significant computing power and memory bandwidth.\par

The other issue is related to calibration. In an industrial setup, millions of cameras are deployed, and they have manufacturing variations. The camera parameters (mainly focal length) can also vary due to high ambient temperatures when driving in a hot region. Thus a model that relies on rectification to correct the distortion could have errors. For instance, dataset capture and training are typically performed on one particular camera, and the model is deployed to work on millions of cameras in commercial vehicles. Thus rectification and cropping to a standard resolution as per the training camera are sub-optimal for a deployed camera. However, if CNN learns the distortion as part of the transfer function, it is only weakly encoded and expected to be more robust. To alleviate these issues, we are motivated to explore a distance estimation model that can work directly on raw images without needing rectification.\par
\section{Self-Supervised Scale-Aware Distance Estimation Framework}
\label{sec:fisheyedistancenet-framework}

Zhou~\etal's~\cite{zhou2017unsupervised} self-supervised monocular \emph{SfM} framework on which most self-supervised monocular depth estimation models build on, aims at learning:
\begin{enumerate}
\item a monocular distance model $g_d: I_t \to D$ predicting a scale-ambiguous distance $\hat D = g_d(I_t(p))$ per pixel $p$ in the target image $I_t$; and
\item an ego-motion predictor $g_x: (I_t, I_{t^\prime}) \to I_{t \to t^\prime}$ predicting a set of six degrees of freedom of the rigid transformations $T_{t \rightarrow t'} \in \text{SE(3)}$, between the target image $I_t$ and the set of reference images $I_{t^\prime}$.
Typically, $t' \in \{t+1, t-1\}$, \ie the frames $I_{t-1}$ and $I_{t+1}$ are used as reference images, although using a larger window is possible.
\end{enumerate}
This method is self-supervised because the ground truth is derived directly from the input signal—the RGB images in this case. There is no need for external data or signals to teach the network because the distance estimator $g_d$ is its teacher! A limitation of this approach is that both distance and pose are estimated up to an unknown scale factor in the monocular \emph{SfM} pipeline. One downside to this method is the scale ambiguity in both distance and pose estimation.\par
\begin{figure}[!t]
  \centering
  \newcommand{\turnwidth}{0.485\columnwidth}

\newcommand{\imlabel}[2]{\includegraphics[width=0.49\columnwidth]{#1}%
\raisebox{2pt}{\makebox[-2pt][r]{\footnotesize #2}}}

\begin{tabular}{@{\hskip 0mm}c@{\hskip 1.5mm}c}
\centering
\imlabel{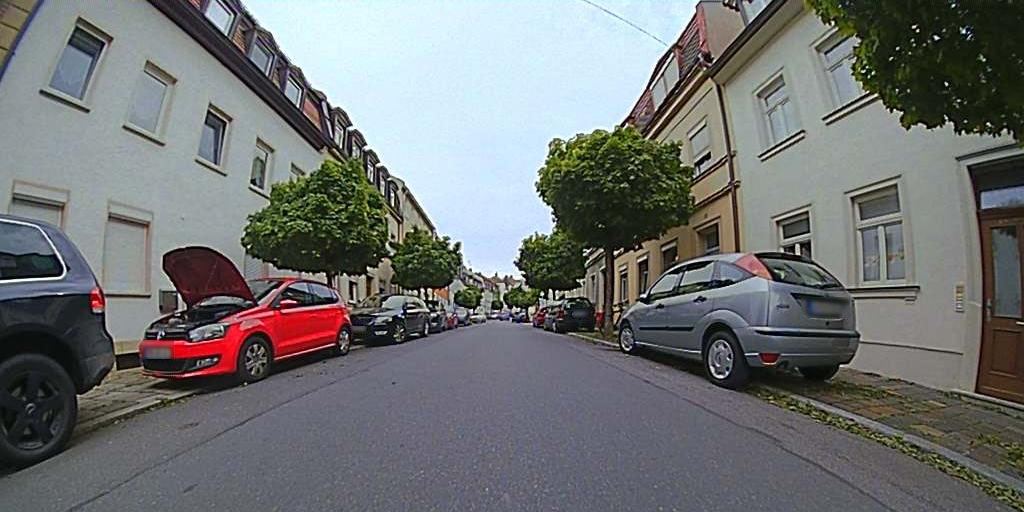}
{\textcolor{white}{WoodScape}} &
\imlabel{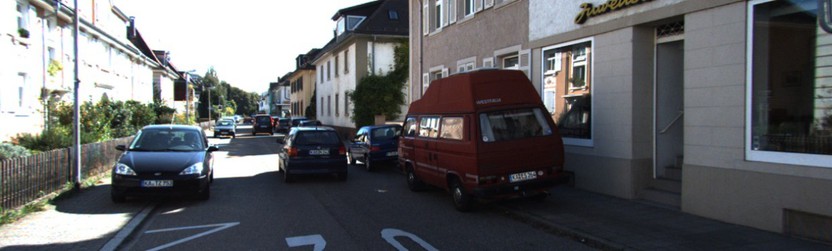}
{\textcolor{white}{KITTI}} \\

\imlabel{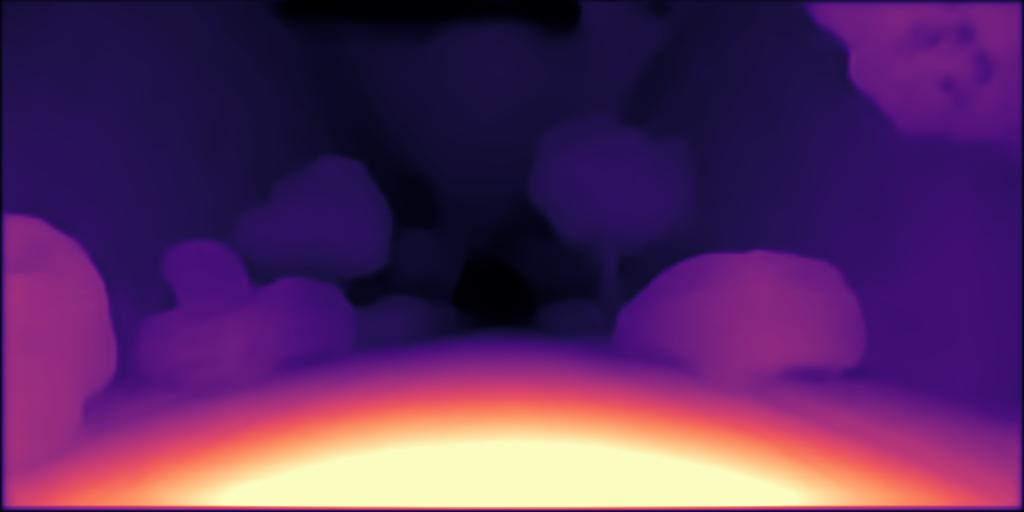}
{} &
\imlabel{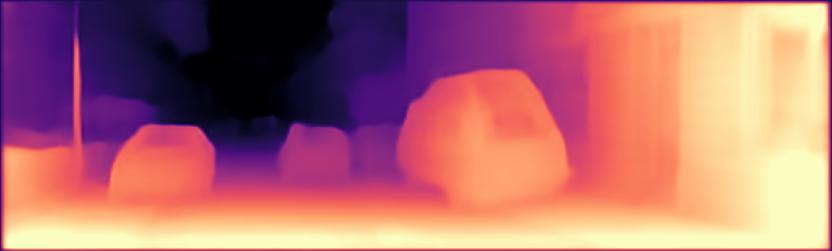}
{} \\
\end{tabular}

  \caption[\bf Distance and depth derived from a single fisheye image.]
          {\bf Distance and depth derived from a single fisheye image (left) and single pinhole image (right) respectively.}
  \label{fig:overview}
\end{figure}
In this thesis, we recover scale-aware distance directly for distorted images (see Figure~\ref{fig:overview}). The distance, which acts as an intermediary variable, is obtained from the network by constraining the model to perform image synthesis. As discussed earlier, distance estimation is an ill-posed problem as there could exist a large number of possible incorrect distances per pixel, which can also recreate the novel view, given the relative pose between $I_t$ and $I_{t^\prime}$.\par

\begin{figure}[!t]
  \centering
    \includegraphics[width=\columnwidth]{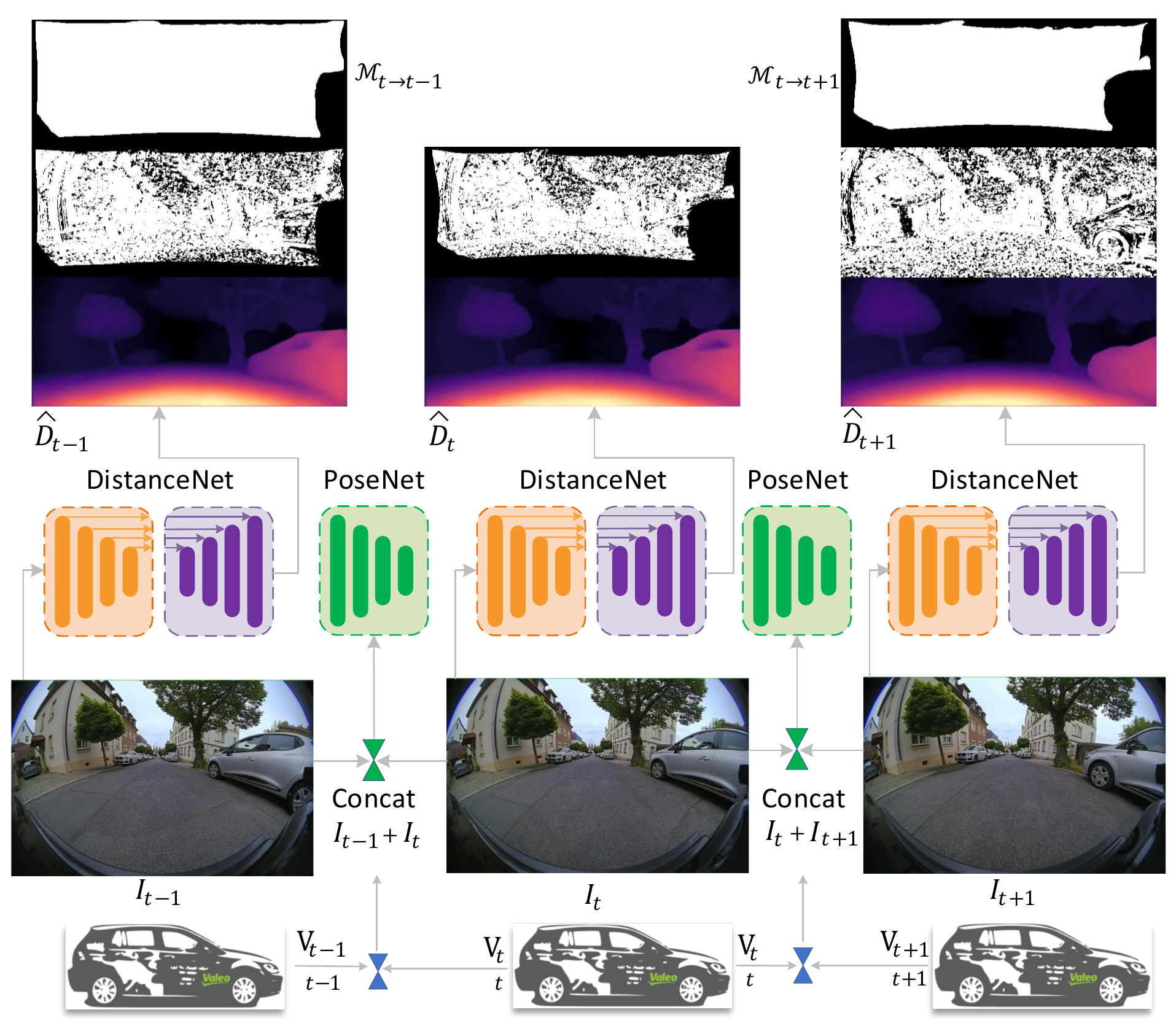}
    \caption[\bf  Overview of the FisheyeDistanceNet framework.]{\textbf{Overview of the FisheyeDistanceNet framework.} The first row represents the ego masks as described in Section~\ref{sec:mask}, $\mathcal{M}_{t \to t-1}$, $\mathcal{M}_{t \to t+1}$ indicate which pixel coordinates are valid when constructing $\hat I_{t-1 \to t}$ from $I_{t-1}$ and $\hat I_{t+1 \to t}$ from $I_{t+1}$ respectively. The second row indicates the masking of static pixels computed after two epochs, where black pixels are filtered from the photometric loss (\ie, $\omega=0$). It prevents dynamic objects at a similar speed as the ego car and low texture regions from contaminating the loss. The masks are computed for forward and backward sequences from the input sequence $S$ and reconstructed images using Eq.~\ref{eqn:staticmask} as described in Section \ref{sec:mask}. The third row represents the distance estimates corresponding to their input frames. The fourth row contains the encoder-decoder network architecture for distance and pose estimation. Finally, the vehicle's odometry data is used to resolve the scale factor issue.}
    \label{fig:model_arch}
\end{figure}

\textbf{View-synthesis} is used as a self-supervising technique, and the network is trained with the source images $I_{t-1}$ and $I_{t+1}$ to synthesize the appearance of a target image $I_t$ on raw fisheye images. For this, we need the projection function $\Pi$ of the chosen camera model, which maps a 3D point $X_c$ in camera coordinates to a pixel $p = \Pi(X_c)$ in image coordinates. An overview of projection models for different lens types can be found in Section~\ref{sec:projection-models}, and we specifically incorporate the Polynomial model. The corresponding unprojection function $\iPi$, which maps an image pixel $p$ and its distance estimate $\hat D$ to the 3D point $X_c = \iPi(p,\hat D)$, is also required. If $\iPi$ cannot be expressed in analytic form, a pre-calculated lookup table is used to ensure computational efficiency. A naive approach would be correcting raw fisheye images to piecewise or cylindrical projections and would essentially render the problem equivalent to Zhou \etal's work~\cite{zhou2017unsupervised}. In contrast, there is a simple yet efficient technique for obtaining scale-aware distance maps at the core of the approach. Figure~\ref{fig:model_arch} illustrates an overview of the \textit{FisheyeDistanceNet} method.\par

This section discusses the geometry of the problem and how it is used to obtain differentiable losses. We describe the scale-aware \textit{FisheyeDistanceNet} framework and its effects on the output distance estimates. Additionally, we provide an in-depth discussion of the various losses.\par
\subsection{Modeling of Fisheye Geometry}
\label{sec:modeling of fisheye geometry}

\subsubsection{Projection from Camera Coordinates to Image Coordinates}
\label{sec:cam2img}

The projection function $X_c \mapsto \Pi(X_c) = p$ of a 3D point $X_c = (x_c, y_c, z_c)^T$ in camera coordinates to a pixel $p = (u, v)^T$ in the image coordinates is obtained via a $4^\text{th}$ order polynomial in the following way:
\begin{align}
    \varphi &= \text{arctan2}(y_{c},x_{c}) \\
    \theta &= \frac{\pi}{2} - \text{arctan2} ({z_c},{r_c}) \\
    \varrho(\theta) &= k_1 \cdot \theta + k_2 \cdot \theta^2 + k_3 \cdot \theta^3 + k_4 \cdot \theta^4 
    \label{eq:poly4} \\
    p &= \begin{pmatrix} u \\ 
    v \end{pmatrix} = \begin{pmatrix} \varrho(\theta) \cdot \cos\varphi \cdot a_x + c_x \\
    \varrho(\theta) \cdot \sin\varphi \cdot a_y + c_y \end{pmatrix} 
\end{align}
where $r_c = \sqrt{x_c^2 + y_c^2}$, $\theta$ is the angle of incidence, $\varrho(\theta)$ is the mapping of incident angle to image radius, $(a_x, a_y)$ is the aspect ratio and $(c_x, c_y)$ is the principal point.\par
\subsubsection{Unprojection from Image Coordinates to Camera Coordinates}
\label{sec:img2cam}

The unprojection function $(p,\hat D) \mapsto \iPi(p,\hat D) = X_c$ of an image pixel $p = (u, v)^T$ and it's distance estimate $\hat D$ to the 3D point $X_c = (x_c, y_c, z_c)^T$ is obtained via the following steps. Letting $(x_i, y_i)^T = \big((u - c_x)/a_x, (v - c_y)/a_y \big)^T$, we obtain the angle of incidence $\theta$ by numerically calculating the $4^\text{th}$ order polynomial roots of $\varrho = \sqrt{x_i^2 + y_i^2}$ using the distortion coefficients $k_1, k_2, k_3, k_4$ (see Eq.~\ref{eq:poly4}). For training efficiency, we pre-calculate the roots and store them in a lookup table for all the pixel coordinates. Now, $\theta$ is used to get
\begin{equation}
    r_c = \hat D \cdot \sin\theta \quad \text{ and } \quad z_c = \hat D \cdot \cos\theta
\end{equation}
where the distance estimate $\hat D$ from the network represents the Euclidean distance $\| X_c \|$ = $\sqrt{x_c^2 + y_c^2 + z_c^2}$ of a 3D point $X_c$. The polar angle $\varphi$ and the $x_c$, $y_c$ components can be obtained as follows:
\begin{equation*}
    \begin{array}{lll}
           \varphi = \text{arctan2} (y_i, x_i),& \hspace{-.1cm} x_c = r_c \cdot \cos\varphi,& \hspace{-.1cm} y_c = r_c \cdot \sin\varphi.
    \end{array}
\end{equation*}
\subsection{Photometric Loss}
\label{sec:photometric loss}

Let us consider the image reconstruction error from a pair of images $I_{t^\prime}$ and $I_t$, distance estimate $\hat D_t$ at time $t$, and the relative pose for $I_t$, with respect to the source image $I_{t^\prime}$'s pose, as $T_{t \to t^\prime}$.
Using the distance estimate $\hat D_t$ of the network a point cloud $P_t$ is obtained via:
\begin{equation}
    \label{pt}
    P_t = \iPi(p_t, \hat D_t)
\end{equation}
where $\iPi$ represents the unprojection from image to camera coordinates as explained in Section~\ref{sec:img2cam}, $p_t$ the pixel set of image $I_t$. The pose estimate $T_{t \to t^\prime}$ from the pose network is used to get an estimate $\hat{P}_{t'} = T_{t \to t^\prime} P_t$ for the point cloud of the image $I_{t^\prime}$. $\hat{P}_{t^\prime}$ is then projected onto the fisheye camera at time $t'$ using the projection model $\Pi$ described in Section~\ref{sec:cam2img}. Combining transformation and projection with Eq.~\ref{pt} establishes a mapping from image coordinates $p_t=(u,v)^T$ at time $t$ to image coordinates $\hat p_{t^\prime}=(\hat u, \hat v)^T$ at time $t^\prime$. This mapping allows for the reconstruction $\hat I_{t' \to t}$ of the target frame $I_t$ by backward warping the source frame $I_{t^\prime}$.
\begin{align}
    \label{pixel_estimate}
    \hat{p}_{t'} = \Pi \big( T_{t \to t'} \iPi(p_t,\hat D_t) \big), \quad
    \hat{I}_{t' \to t}^{uv} = \Big\langle I_{t'}^{\hat{u}\hat{v}} \Big\rangle
\end{align}
Since the warped coordinates $\hat u, \hat v$ are continuous, we apply the differentiable spatial transformer network introduced by~\cite{jaderberg2015spatial} to compute $\hat{I}_{t' \to t}$ by performing bilinear interpolation of the four pixels from $I_{t^\prime}$ which lie close to $\hat p_{t'}$. The symbol $\big\langle\dots\big\rangle$ denotes the corresponding sampling operator.\par

Following~\cite{monodepth17, zhao2016loss} the image reconstruction error between the target image $I_t$ and the reconstructed target image $\hat I_{t' \to t}$ is calculated using the L1 pixel-wise loss term combined with Structural Similarity (SSIM)~\cite{wang2004image}, as the photometric loss $\mathcal{L}_{p}$ given by Eq.~\ref{eq:loss-photo} below.
\begin{align}
  \label{eq:loss-photo}
  \tilde{\mathcal{L}}_{p}(I_t,\hat I_{t' \to t}) &= \alpha~\frac{1 - \text{SSIM}(I_t,\hat I_{t' \to t}, \mathcal{M}_{t \to t^\prime})}{2} \nonumber \\
  &\quad+ (1-\alpha)~ \left\| (I_t - \hat I_{t' \to t}) \odot \mathcal{M}_{t \to t^\prime} \right\|_{l^1} \nonumber \\
    \mathcal{L}_{p} &= \min_{t^\prime \in \{t+1,t-1\}} \tilde{\mathcal{L}}_p(I_t, \hat I_{t' \to t})
\end{align}
where $\alpha = 0.85$, $\mathcal{M}_{t \to t^\prime}$ is the binary mask as discussed in Section~\ref{sec:mask} and the symbol $\odot$ denotes element-wise multiplication. Following~\cite{godard2019digging} instead of averaging the photometric error over all source images, we adopt per-pixel minimum. This significantly sharpens the occlusion boundaries and reduces the artifacts resulting in higher accuracy.\par

The self-supervised framework assumes a static scene, no occlusion, and change of appearance (\eg, brightness constancy). A large photometric cost is incurred, potentially worsening the performance if dynamic objects and occluded regions exist. These areas are treated as outliers similar to \cite{zhou2018unsupervised} and clip the photometric loss values to a $95^\text{th}$ percentile. Zero gradients are obtained for errors larger than $95\%$. This improves the optimization process and provides a way to strengthen the photometric error.\par
\subsection{Solving the Scale Factor Ambiguity}
\label{sec: scale-aware sfm}

To overcome the limitations of this \emph{SfM} framework as discussed in Section \ref{sec:depth-challenges} and to achieve scale-aware distance values, we normalize the pose network's estimate $T_{t \to t'}$ and scale it with $\Delta x$, the displacement magnitude relative to target frame $I_t$ which is calculated using vehicle's instantaneous velocity estimates $v_{t'}$ at time $t'$ and $v_t$ at time $t$. We also apply this technique on KITTI~\cite{geiger2013vision} to obtain metric depth maps.\par
\begin{equation}
    \overline{T}_{t \to t'} = \frac {T_{t \to t'}} {\|T_{t \to t'}\|} \cdot \Delta x
\end{equation}
\subsection{Masking Static Pixels and Ego Mask}
\label{sec:mask}

Following~\cite{godard2019digging}, we incorporate a masking approach to filter out static pixels that do not change their appearance from one frame to the other in the training sequence. The approach would filter out objects that move at the same speed as the ego-car and ignore the static frame when the ego-car stops moving. Similar to other approaches~\cite{godard2019digging, zhou2017unsupervised, Vijayanarasimhan2017, luo2019every} the per-pixel mask $\omega$ is applied to the loss by weighting the pixels selectively. Instead of being learned from the object motion~\cite{casser2019depth}, the mask is computed in the forward pass of the network, yielding a binary mask output where $\omega \in \{0, 1\}$. Wherever the photometric error of the warped image $\hat I_{t^\prime \to t}$ is not lower than that of the original unwarped source frame $I_{t^\prime}$ in each case compared to the target frame $I_t$, $\omega$ is set to ignore the loss of such pixels, \ie
\begin{align}
    \omega &= \big[ \, 
        \min_{t^\prime} pe(I_t, \hat I_{t^\prime \to t}) 
            < 
        \min_{t^\prime} pe(I_t, I_{t^\prime})     
            \, \big]
    \label{eqn:staticmask}
\end{align}
where $[\,]$ is the Iverson bracket. Additionally, we add a binary ego mask $\mathcal{M}_{t \to t^\prime}$ proposed in~\cite{mahjourian2018unsupervised} that ignores computing the photometric loss on the pixels that do not have a valid mapping \ie some pixel coordinates of the target image $I_t$ may not be projected onto the source image $I_{t'}$ given the estimated distance $\hat D_t$.\par
\subsection{Edge-Aware Smoothness Loss}
\label{sec:edge-smoothness}

In order to regularize distance and avoid divergent values in occluded or texture-less low-image gradient areas, we add a geometric smoothing loss. We adopt the edge-aware term similar to~\cite{monodepth17, mahjourian2018unsupervised, zou2018df}. The regularization term is imposed on the inverse distance map. Unlike previous works, the loss is not decayed for each pyramid level by a factor of $2$ due to down-sampling, as we use a super resolution network (see Section~\ref{sec:deformable super-resolution network})
\begin{equation}
    \mathcal{L}_{s}(\hat{D}_t) = | \partial_u \hat{D}^*_t | e^{-|\partial_u I_t|} + | \partial_v \hat{D}^*_t | e^{-|\partial_v I_t|}
\end{equation}
To discourage shrinking of estimated distance~\cite{Wang_2018_CVPR}, mean-normalized inverse distance of $I_t$ is considered, i.e. $\hat{D}^*_t = \hat{D}^{-1}_t / \overline{D}_t$, where $\overline{D}_t$ denotes the mean of $\hat{D}^{-1}_t := 1 /\hat{D}_t$.\par
\subsection{Cross-Sequence Distance Consistency Loss}

The \textit{SfM} setting uses an N-frame training snippet $S = \{ {I_1},{I_2}, \cdots ,{I_N}\} $ from a video as input. The \textit{FisheyeDistanceNet} can estimate the distance of each image in the training sequence. Another constraint can be enforced among the frames in $S$ since the distances of a 3D point estimated from different frames should be consistent.\par

Let us assume $\hat D_{t'}$ and $\hat D_{t}$ are the estimates of the images $I_{t'}$ and $I_t$ respectively. For each pixel ${{p}_{t}}\in {{I}_{t}}$, we can use Eq.~\ref{pixel_estimate} to obtain $\hat{p}_{t'}$. Since it's coordinates are real valued, we apply the differentiable spatial transformer network introduced by~\cite{jaderberg2015spatial} and estimate the distance value of $\hat{p}_{t'}$ by performing bilinear interpolation of the four pixel's values in $\hat D_{t'}$ which lie close to $\hat p_{t'}$. Let us denote the distance map obtained through this as ${{\hat{D}}_{t \to t'}}\left(p_{t}\right)$. 
Next, we can transform the point cloud in frame $I_t$ to frame $I_{t'}$ by first obtaining $P_t$ using Eq.~\ref{pt}. We transform the point cloud $P_t$ using the pose network's estimate via $\hat{P}_{t'} = T_{t \to t'} P_t$. Now, ${{D}_{t \to t'}}\left({{p}_{t}}\right) := \| \hat{P}_{t'} \|$ denotes the distance generated from point cloud $\hat{P}_{t'}$.
Ideally, ${{D}_{t \to t'}}\left( {{p}_{t}} \right)$ and ${{\hat{D}}_{t \to t'}}\left( p_{t} \right)$ should be equal. Therefore, we can define the following cross-sequence distance consistency loss (CSDCL) for the training sequence~$S$:
\begin{align} 
    \label{equ:dclf}
	\mathcal{L}_{dc} = \sum_{t=1}^{N-1} \sum_{t'=t+1}^{N} \bigg( &\sum_{p_t} \mathcal{M}_{t \to t^\prime}
	\left| D_{t \to t^\prime}\left(p_t \right) - \hat D_{t \to t^\prime}\left(p_t \right) \right| \nonumber \\
	+ &\sum_{p_{t'}} \mathcal{M}_{t' \to t} \left| D_{t' \to t}\left(p_{t'} \right) - \hat D_{t' \to t}\left(p_{t'} \right) \right| \bigg)
\end{align}
Eq.~\ref{equ:dclf} contains one term for which pixels and point clouds are warped forwards in time (from $t$ to $t'$) and one term for which they are warped backwards in time (from $t'$ to $t$).

In prior works~\cite{Vijayanarasimhan2017, zou2018df}, the consistency error is limited to only two frames, whereas we apply it to the entire training sequence $S$. This induces more constraints and enlarges the baseline, inherently improving the distance estimation~\cite{zhou2018unsupervised}.\par
\subsubsection{Backward Sequence}
\label{sec:backward sequence}

In the forward sequence, we synthesize the target frame $I_t$ with the source frames $I_{t-1}$ and $I_{t+1}$ (\ie as per above discussion $t' \in \{t+1, t-1\}$).
Analogously, a backward sequence is carried out using $I_{t-1}$ and $I_{t+1}$ as target frames and $I_t$ as source frame. We include warps $\hat I_{t \to t-1}$ and $\hat I_{t \to t+1}$, thereby inducing more constraints to avoid overfitting and resolve unknown distances in the border areas at the test time, as also observed in previous works~\cite{godard2019digging,zhou2017unsupervised,yin2018geonet}. We construct the loss for the additional backward sequence similar to the forward sequence. This comes at the cost of high computational effort and longer training time as we perform two forward and backward warps, which yields superior results on the WoodScape and KITTI dataset compared to the previous approaches~\cite{godard2019digging,zhou2017unsupervised} which train only with one forward sequence and one backward sequence.\par
\textbf{Depicting the importance of additional warps}

The reconstructed image $\hat{I}_{t' \to t}^{uv}$ will result in a zoom-in operation where the border distance values are useful and will get meaningful gradients to train with. However, center values will have much noise due to the low displacement. On the other hand, since $\hat{I}_{t \to t'}^{uv}$ will result in a zoom-out effect, the border distance values are insignificant and should be filtered out from the photometric loss because the pixels that have to be retrieved do not exist. Center distance, though, will have a warp that sample values from the border of the frame, \ie, with large displacement and the gradient will be less noisy than with $\hat{I}_{t' \to t}^{uv}$. Additional warps will induce more constraints to avoid overfitting and resolve unknown distances at the borders at test time.\par
\subsection{Final Training Loss}
\label{sec:fisheyedistancenet-final-loss}

The overall self-supervised \emph{SfM} objective consists of a photometric loss $\mathcal{L}_p$ imposed between the reconstructed target image $\hat{I}_{t' \to t}$ and the target image $I_t$, included once for the forward and once for the backward sequence, and a distance regularization term $\mathcal{L}_s$ ensuring edge-aware smoothing in the distance estimates. Finally, $\mathcal{L}_{dc}$ a cross-sequence distance consistency loss derived from the chain of frames in the training sequence $S$ is also included. To prevent the training objective from getting stuck in the local minima due to the gradient locality of the bilinear sampler~\cite{jaderberg2015spatial}, we adopt four scales to train the network as followed in~\cite{zhou2017unsupervised,monodepth17}. The final objective function is averaged per pixel, scale, and image batch.
\begin{align} 
    \label{equ:objective}
    \mathcal{L} &= \sum\limits_{n = 1}^4 {\frac{{\mathcal{L}_{n}}}{{{2^{n - 1}}}}} ,\\
    \mathcal{L}_{n} &= {}^n\mathcal{L}_{p}^f + {}^n\mathcal{L}_{p}^b + \gamma\ {}^n{\mathcal{L}_{dc}} + \beta\ {}^n{\mathcal{L}_{s}} \nonumber
\end{align}
\subsection{Handling Common Camera Distortion Models}

\begin{figure*}[!t]
    \centering
    \includegraphics[width=\textwidth]{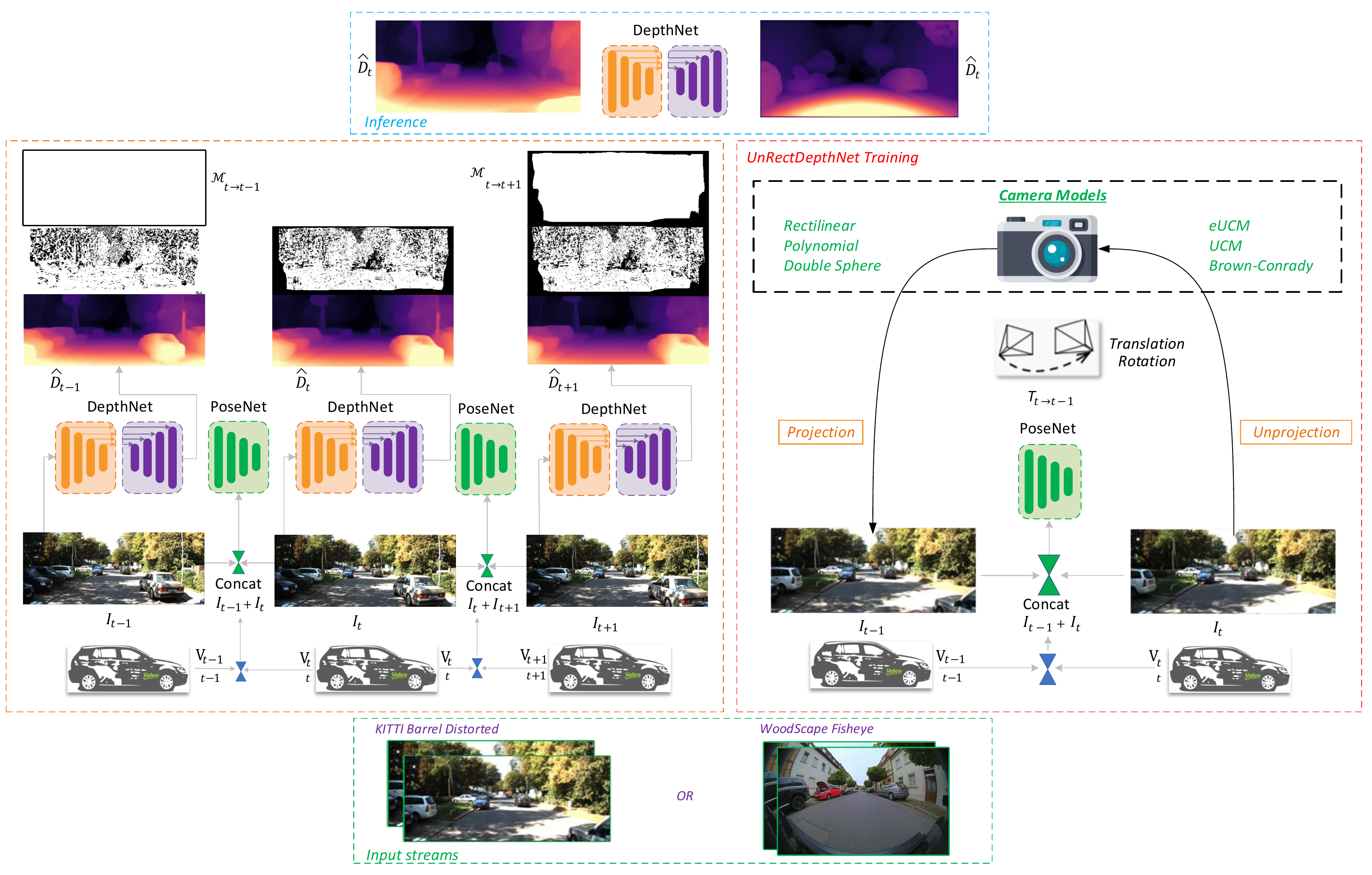}
    \caption[\bf Overview of the UnRectDepthNet framework.]
    {\textbf{Overview of the UnRectDepthNet: A generic depth estimation training framework that can handle various camera models.}}
    \label{fig:unrectdepthnet-model_arch}
\end{figure*}

\begin{figure}[!t]
  \centering
  \newcommand{\turnwidth}{0.485\columnwidth}

\newcommand{\imlabel}[2]{\includegraphics[width=0.488\columnwidth]{#1}
\raisebox{2pt}{\makebox[-2pt][l]{\footnotesize \sffamily #2}}}

\begin{tabular}{@{\hskip 0mm}c@{\hskip 1.5mm}c}
\centering
    \imlabel{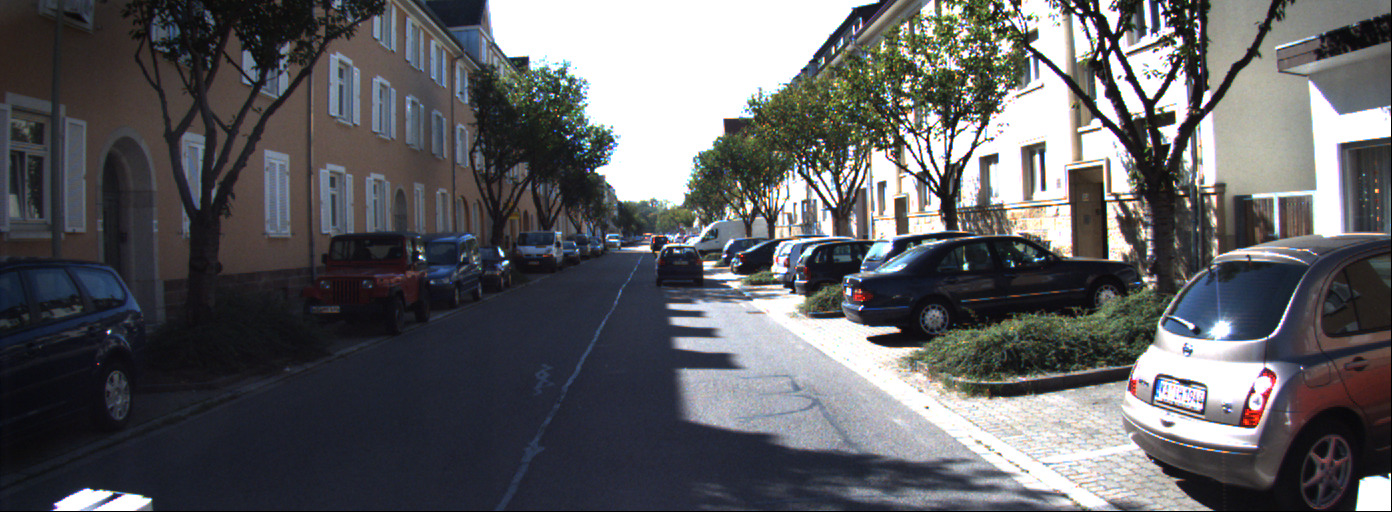}
    {\hspace{-0.47\columnwidth}\textcolor{white}{Unrectified}} &
    \imlabel{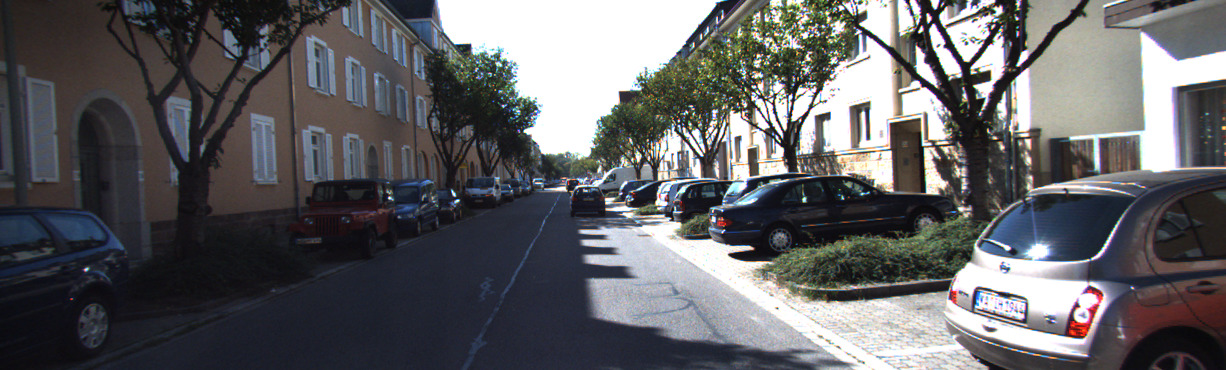}
    {\hspace{-0.47\columnwidth}\textcolor{white}{Rectified}} \\
    
    \imlabel{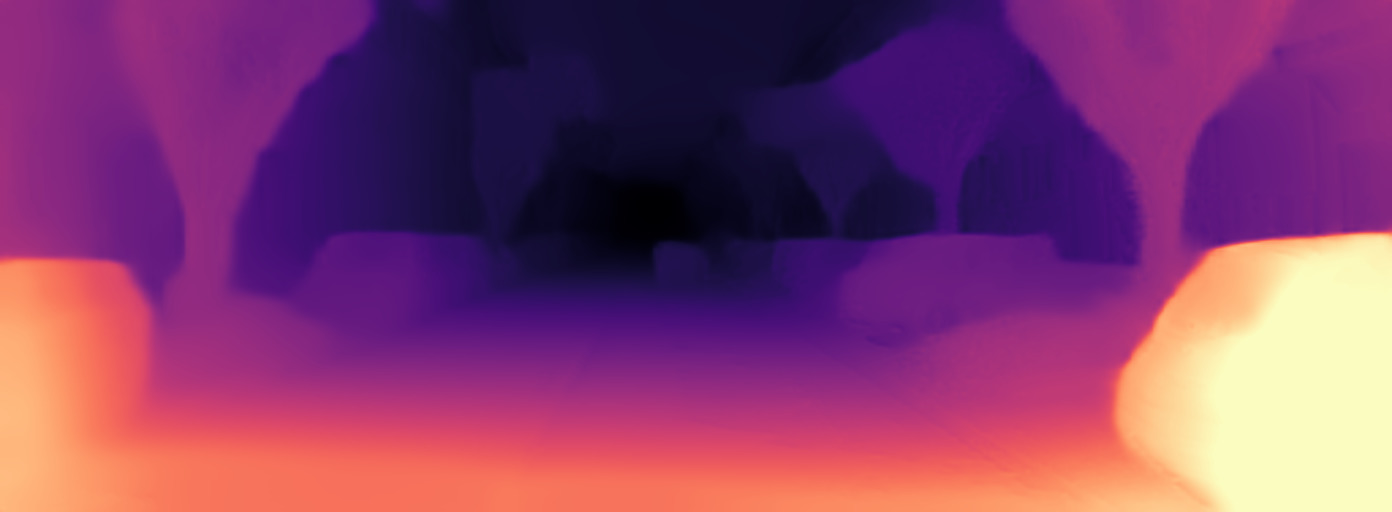}
    {} &
    \imlabel{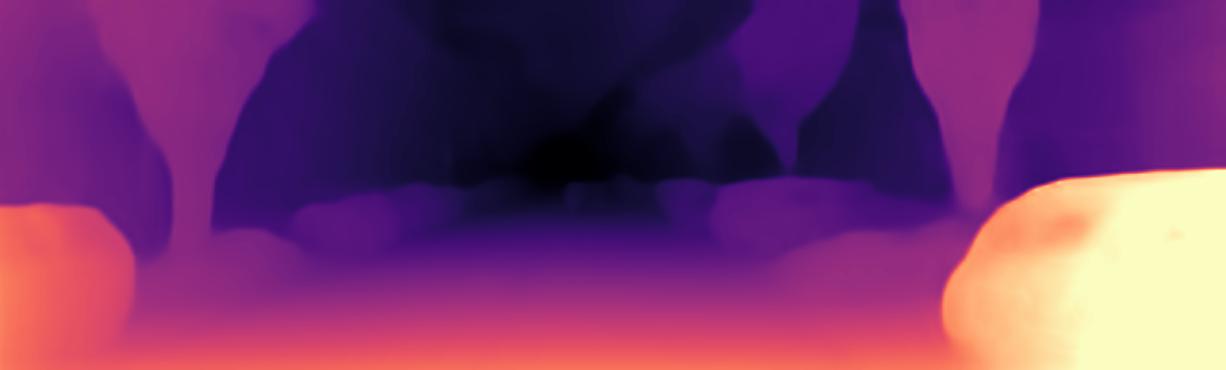}
    {} \\
\end{tabular}

  \caption[\bf Depth obtained from a single unrectified KITTI image.]
          {\bf Depth obtained from a single unrectified (left) and rectified KITTI image (right).}
  \label{fig:unrectdepthnet-overview}
\end{figure}

Up to now, for the \emph{SfM} framework to estimating depth, we employed a pinhole camera model on KITTI and a polynomial model WoodScape. Following the training regime's success, we illustrate an in-detail overview of the \textit{UnRectDepthNet}~\cite{kumar2020unrectdepthnet}: A self-supervised generic training framework for handling common camera distortion models discussed in Section~\ref{sec:projection-models}, to estimate depth directly from raw unrectified images is shown in Figure~\ref{fig:unrectdepthnet-model_arch}. The UnRectDepthNet training block on the right enables the usage of various camera models generically listed in the black box. The distortion is then handled internally in the unprojection and projection steps of the transformation from $I_t$ to $I_{t-1}$. We test the generic framework with barrel distorted KITTI images and distorted WoodScape fisheye video sequences in this regime. The block on the left indicates the entire workflow of the training pipeline where the top row depicts the ego masks as explained in Section~\ref{sec:mask}, $\mathcal{M}_{t \to t-1}$, $\mathcal{M}_{t \to t+1}$ represents the valid pixel coordinates while synthesizing $\hat I_{t-1 \to t}$ from $I_{t-1}$ and $\hat I_{t+1 \to t}$ from $I_{t+1}$ respectively. The following row showcases the masks used to filter static pixels obtained after training two epochs, and the black pixels are removed from the reconstruction loss. Dynamic objects moving at speeds similar to the ego car's and homogeneous areas are filtered out to prevent the erroneous signals in the reconstruction loss. The third row shows the depth predictions, where the scale ambiguity is resolved using the ego vehicle's odometry data as discussed in Section~\ref{sec: scale-aware sfm}. Finally, the top block illustrates the inference output. We compare a sample output from the framework on the unrectified (left) KITTI image vs. the standard rectified (right) KITTI image in Figure~\ref{fig:unrectdepthnet-overview}. We can see that our model handles the barrel distortion and outputs sharp depth maps without losing any FoV.\par
\section{Network Details}
\label{sec:deformable super-resolution network}

The distance estimation network is mainly based on the U-net architecture~\cite{ronneberger2015u}, an \textit{encoder-decoder} network with skip connections. After testing different ResNet family variants, such as ResNet50 with 25M parameters, we chose a ResNet18~\cite{he2016deep} as the encoder. The key aspect here is replacing normal convolutions with deformable convolutions since regular CNNs are inherently limited in modeling large, unknown geometric distortions due to their fixed structures, such as fixed filter kernels, fixed receptive field sizes, and fixed pooling kernels~\cite{dai2017deformable,zhu2019deformable}.\par
\begin{figure*}[!t]
    \centering
    \includegraphics[width=\textwidth]{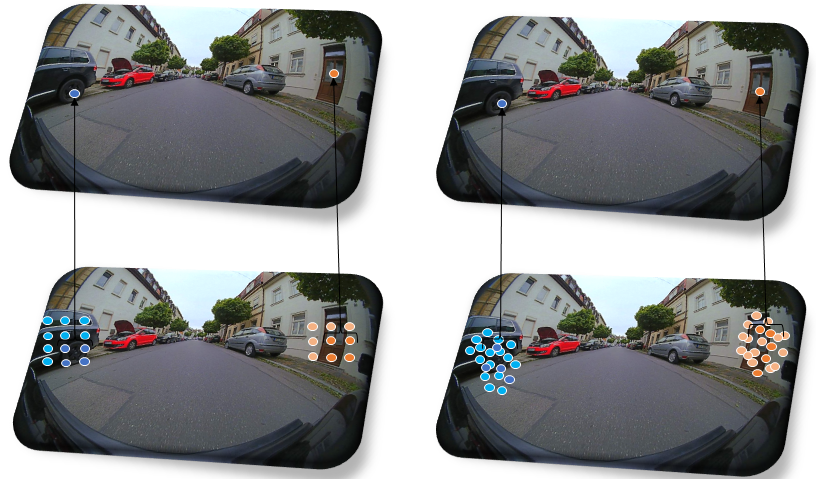}
    \caption[\bf Standard vs. Deformable convolution layer.]
    {\textbf{Standard vs. Deformable convolution layer.} Deformable convolution will pick the values at different locations for convolutions conditioned on the input image.}
    \label{fig:deformable-conv}
\end{figure*}
In a deep CNN, the upper layers encode high-level scene information with weak spatial information, including object- or category-level evidence. Features from the middle layers are expected to describe middle-level representations of object parts and retain spatial information. Features from the lower convolution layers encode low-level spatial visual information like edges, corners, circles. That means that the middle and lower layers are responsible for learning spatial structures. If the deformable convolution is applied to the lower or middle layers, the spatial structures are susceptible to fluctuation. The spatial correspondence between input images and output distance maps is difficult to be preserved. This is the spatial correspondence problem indicated in~\cite{li2017dense}, which is critical in pixel-wise distance estimation. Hence, deformable convolution is applied to the last few convolution layers as proposed by~\cite{dai2017deformable}.\par

To alleviate this problem, Zhu \etal~\cite{zhu2019deformable} proposed a better, more Deformable ConvNet (see Figure~\ref{fig:deformable-conv}) with enhanced modeling power that can effectively model geometric transformations. We incorporate the enhanced modulated deformable convolutions to the \textit{FisheyeDistanceNet} and \textit{PoseNet}.\par

In previous works~\cite{godard2019digging,zhou2017unsupervised,monodepth17, Wang_2018_CVPR,yin2018geonet}, the decoded features were upsampled via a nearest-neighbor interpolation or with learnable transposed convolutions. This process's main drawback is that it may lead to large errors at object boundaries in the upsampled distance map as the interpolation combines distance values of background and foreground. For effective and detailed preservation of the decoded features, we leverage the concept of sub-pixel convolutions~\cite{shi2016real} to the super-resolution network. We use pixel shuffle convolutions and replace the convolutional feature upsampling, performed via a nearest-neighbor interpolation or with learnable transposed convolutions. The resulting distance maps are super-resolved, have sharp boundaries, and expose more details of the scene.\par

The pose estimation network's backbone is based on~\cite{godard2019digging} and predicts rotation using Euler angle parameterization. Compared to previous works~\cite{zhou2017unsupervised,mahjourian2018unsupervised, Wang_2018_CVPR}, which stack the whole sequence as input and estimate the poses relative to the center image, we consider two consecutive images as input to the network, where $I_t$ is the target view, and $I_{t-1}$, $I_{t+1}$ are the source views. The output is a set of six DOF transformations between $I_{t-1}$ and $I_t$ as well as $I_{t}$ and $I_{t+1}$. We replace standard convolutions with deformable convolutions for the encoder-decoder setting.\par
\section{Experiments}

\subsection{Implementation Details}

We use Pytorch~\cite{paszke2017automatic} and employ the Ranger (RAdam \cite{liu2019variance} + LookAhead~\cite{zhang2019lookahead}) optimizer to minimize the training objective function (\ref{equ:objective}). The model is trained using Titan RTX with a batch size of 20 for 20 epochs, with an initial learning rate of $4 \times {10}^{-4}$ with OneCycleScheduler~\cite{smith2019super}. The network's sigmoid output $\sigma$ is converted to distance with $D = 1 / ({m \cdot \sigma + n})$ for pinhole model and $D = {m \cdot \sigma + n}$ for fisheye, where $m$ and $n$ are chosen such that $D$ is bounded between $0.1$ and $100$ units. The fisheye image's original input resolution is $1280\times800$ pixels; we crop it to $1024\times512$ to remove the vehicle's bumper, shadow, and other artifacts of the vehicle. Finally, the cropped image is downscaled to $512\times256$ before feeding it to the network. For the pinhole model on KITTI, we use $640\times192$ pixels as the network input. We use $608\times224$ pixels as the network input to maintain the original aspect ratio for KITTI distorted images. The loss weighting factors $\beta$ and $\gamma$ of smoothness and cross-sequence distance consistency loss are set to $0.001$. To remove checkerboard artifacts in the sub-pixel convolution~\cite{shi2016real}, the final convolutional layers are initialized before the pixel shuffle operation as described in~\cite{aitken2017checkerboard}.\par

We experimented with batch normalization~\cite{ioffe2015batch} and group normalization~\cite{wu2018group} layers in the encoder-decoder setting. We have found that group normalization with $G=32$ significantly improves the results~\cite{he2019rethinking}.
The smoothness weight term $\beta$ and cross-sequence distance consistency weight term $\gamma$ have been set to $0.001$. We applied deformable convolutions to the 3 x 3 Conv layers in stages conv3, conv4, and conv5 in ResNet18 and ResNet50, with 12 layers of deformable convolution in the encoder part compared to 3 layers in \cite{dai2017deformable}, all in the conv5 stage for ResNet50. We replaced the subsequent layers of the decoder with deformable convolutions for the distance and pose network. For the pinhole model, on the KITTI Eigen split in Section~\ref{sec:kitti_eigen_split}, we used regular convolutions instead of deformable convolutions.\par
\begin{figure*}[htbp]
  \centering
  \resizebox{\textwidth}{!}{
  \newcommand{\turnheightnew}{0.25\columnwidth}
\newcommand{\turnheight}{0.3\columnwidth}
\centering

\begin{tabular}{@{\hskip 0.5mm}c@{\hskip 0.5mm}c@{\hskip 0.5mm}c@{\hskip 0.5mm}c@{\hskip 0.5mm}c@{}}

{\rotatebox{90}{\hspace{12mm}Raw Input}} &
\includegraphics[height=\turnheight, width=71mm]{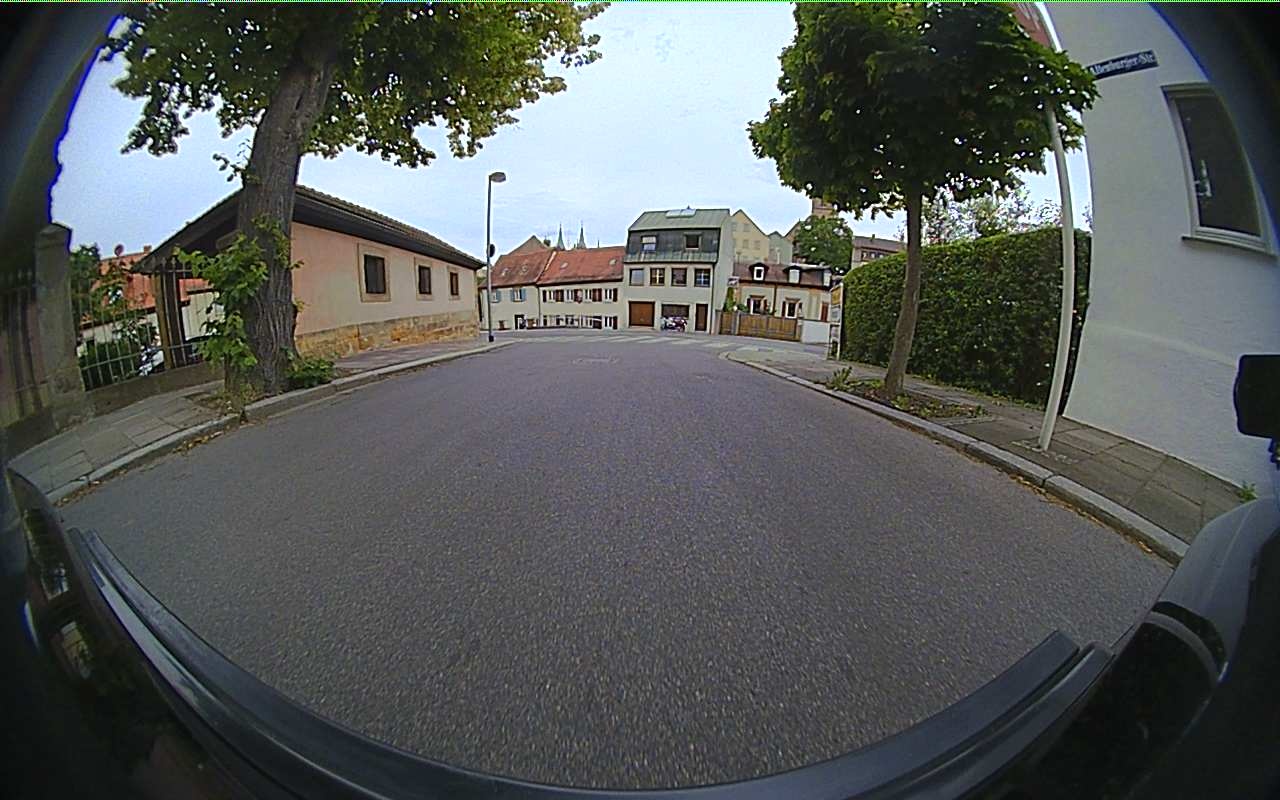} &
\includegraphics[height=\turnheight, width=71mm]{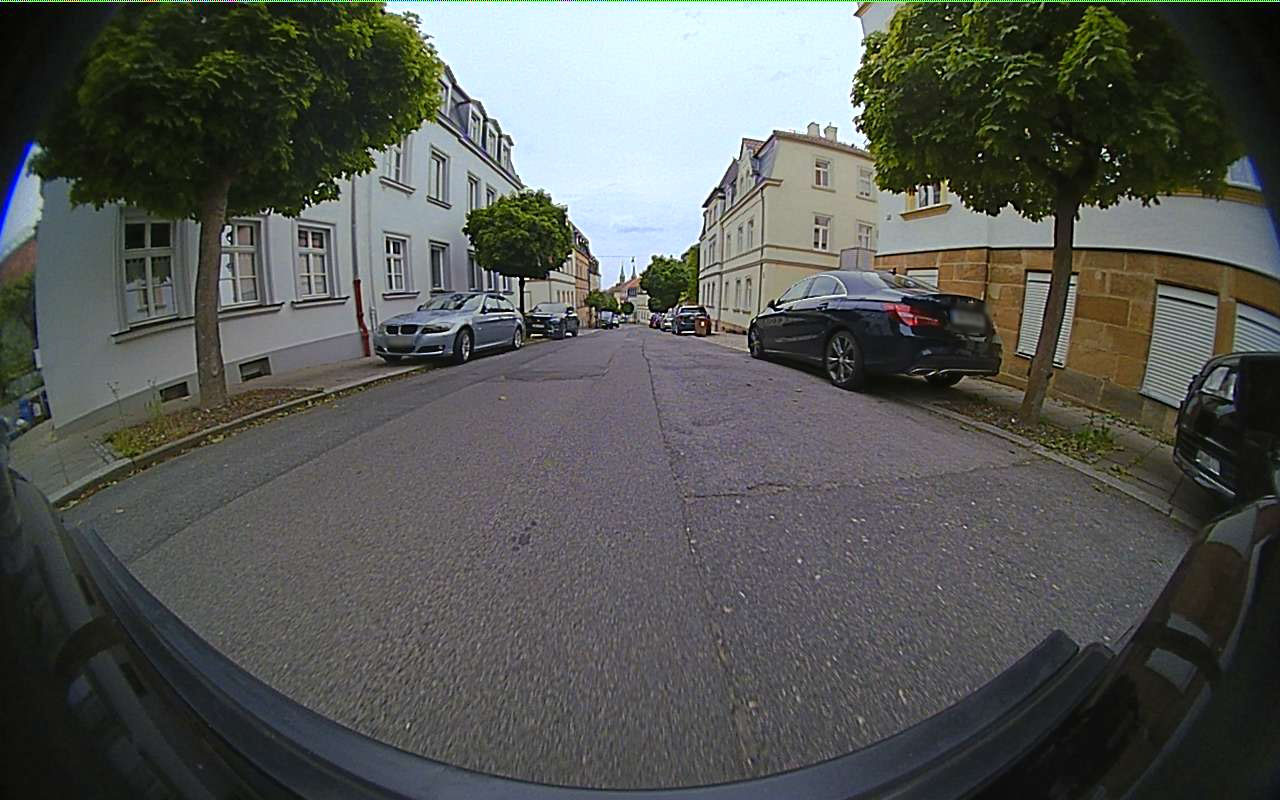} \\

{\rotatebox{90}{\hspace{4mm}Cropped Input}} &
\includegraphics[height=\turnheightnew]{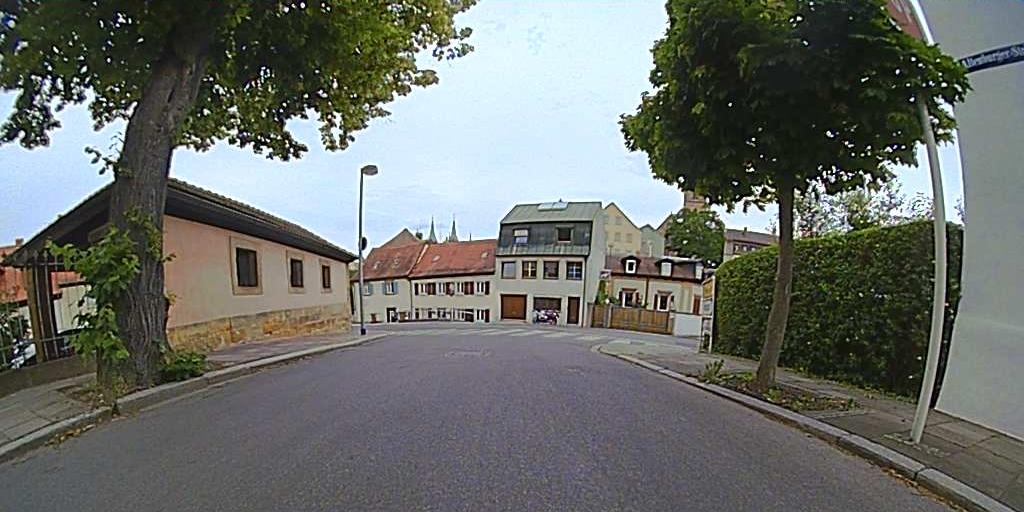} &
\includegraphics[height=\turnheightnew]{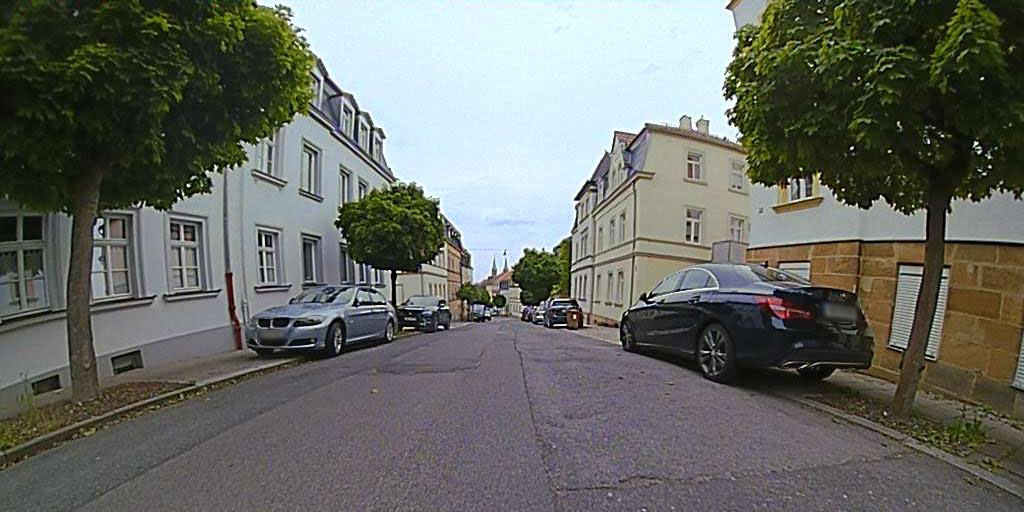} \\

{\rotatebox{90}{\hspace{0mm}\scriptsize}} &
\includegraphics[height=\turnheightnew]{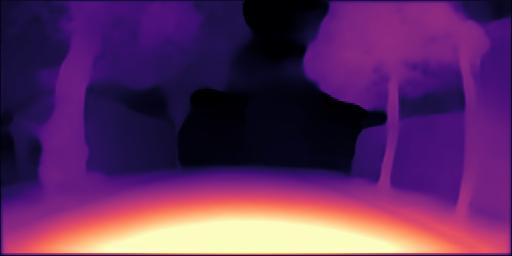} &
\includegraphics[height=\turnheightnew]{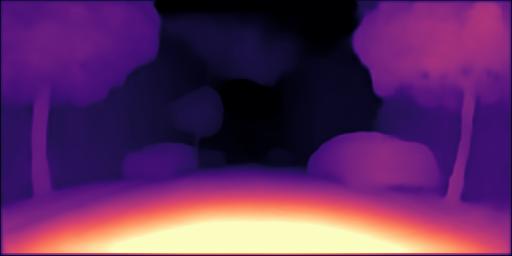} \\

{\rotatebox{90}{\hspace{12mm}Raw Input}} &
\includegraphics[height=\turnheight, width=71mm]{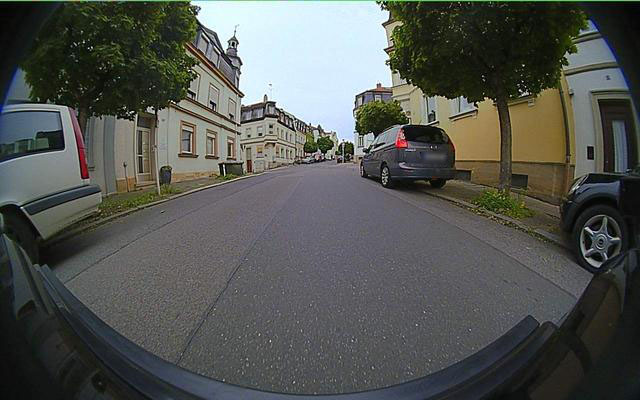} &
\includegraphics[height=\turnheight, width=71mm]{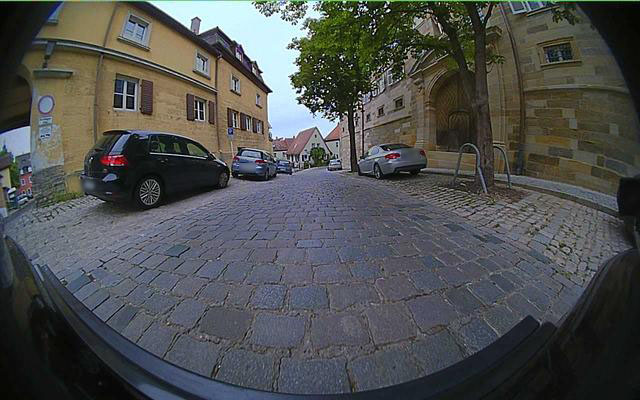}\\

{\rotatebox{90}{\hspace{4mm}Cropped Input}} &
\includegraphics[height=\turnheightnew]{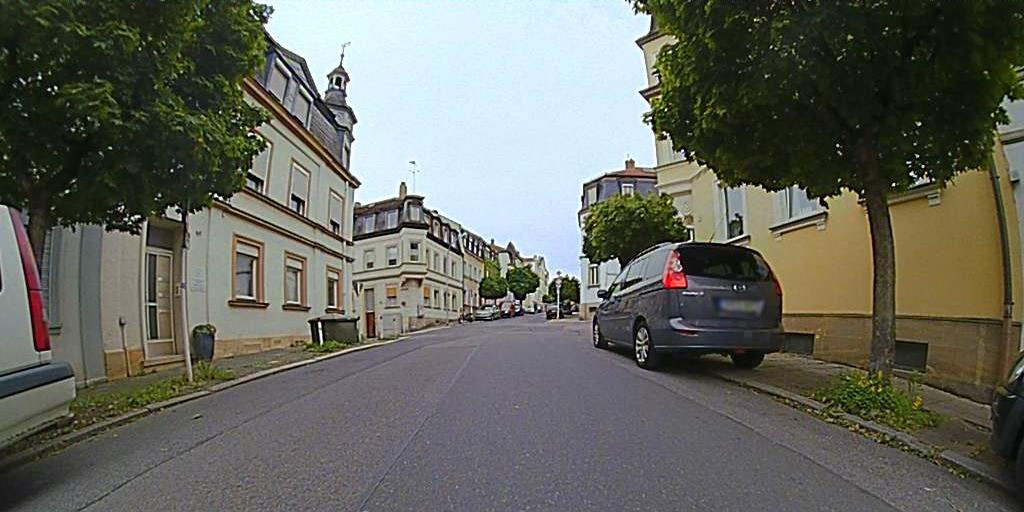} &
\includegraphics[height=\turnheightnew]{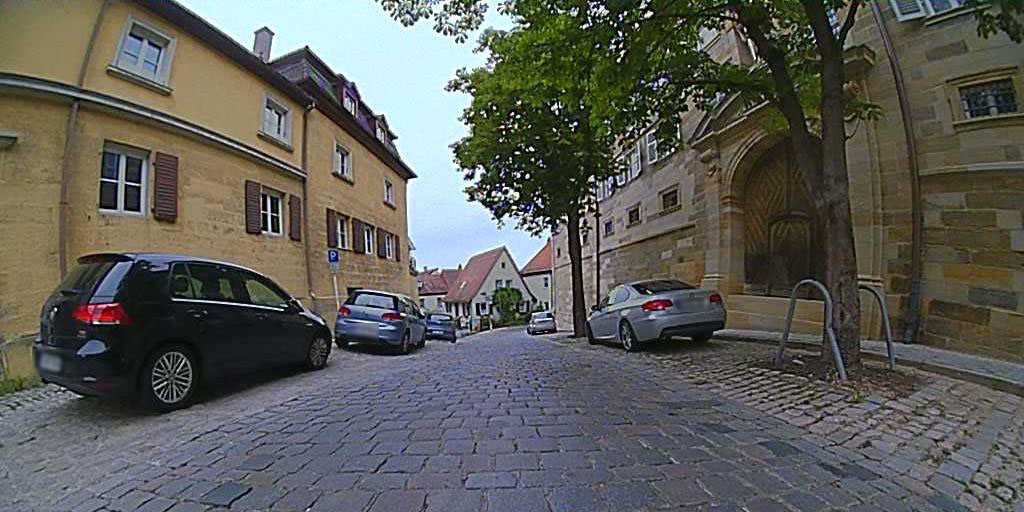}\\

{\rotatebox{90}{\hspace{0mm}\scriptsize}} &
\includegraphics[height=\turnheightnew]{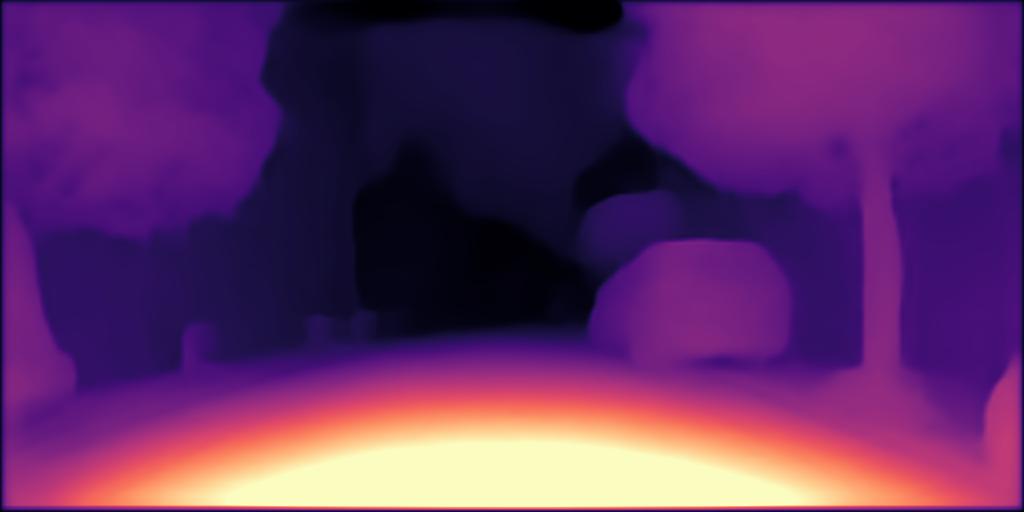} &
\includegraphics[height=\turnheightnew]{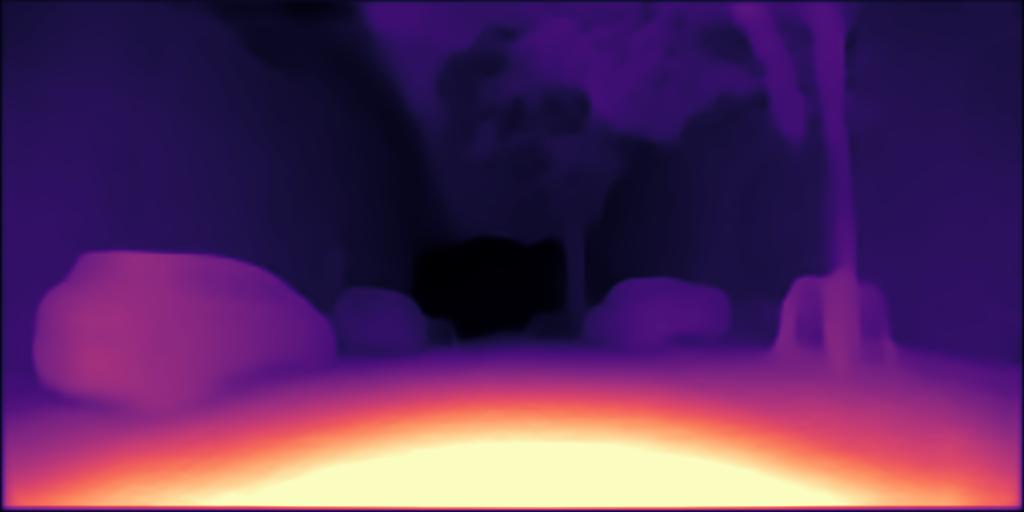} \\

\end{tabular}
}
  \caption[\bf Qualitative results of FisheyeDistanceNet on the WoodScape dataset.]
  {\textbf{Qualitative results of FisheyeDistanceNet on the WoodScape dataset.} For more qualitative results, we refer to this video: \url{https://youtu.be/Sgq1WzoOmXg}.}
  \label{fig:fisheye_qual}
\end{figure*}
\begin{figure*}[htbp]
  \centering
  \resizebox{\textwidth}{!}{
  \newcommand{\turnheightnew}{0.25\columnwidth}
\centering

\begin{tabular}{@{\hskip 0.5mm}c@{\hskip 0.5mm}c@{\hskip 0.5mm}c@{\hskip 0.5mm}}

{\rotatebox{90}{\hspace{5mm}\shortstack{\Large KITTI \\ \Large rectified}}} &
\includegraphics[height=\turnheightnew, width=90mm]{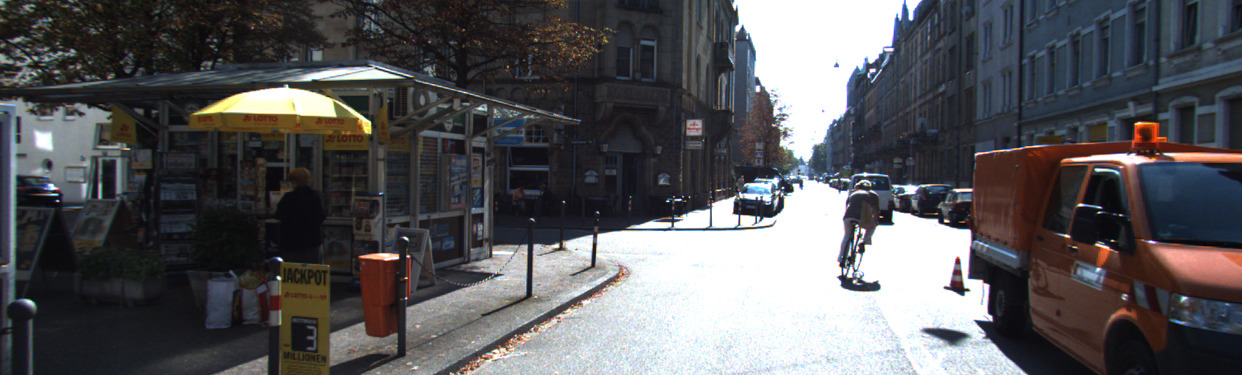} &
\includegraphics[height=\turnheightnew, width=90mm]{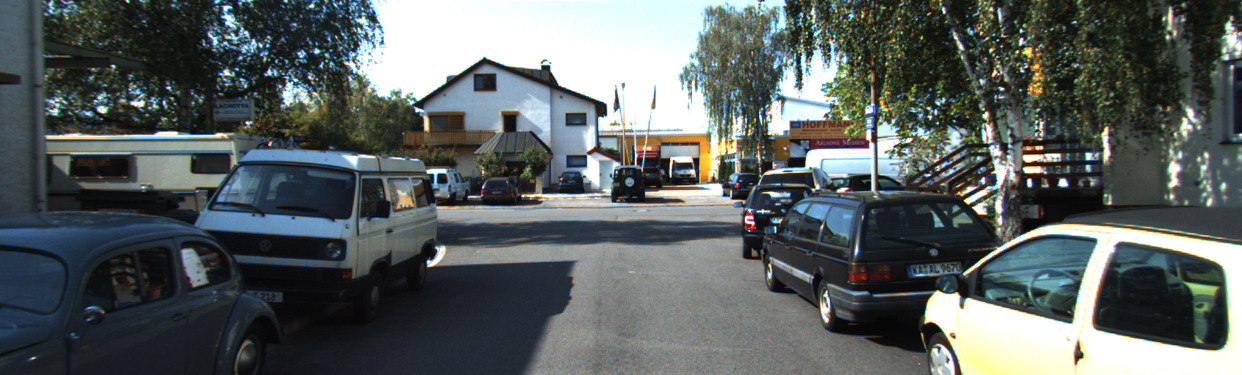} \\

{\rotatebox{90}{\hspace{0mm}\scriptsize}} &
\includegraphics[height=\turnheightnew, width=90mm]{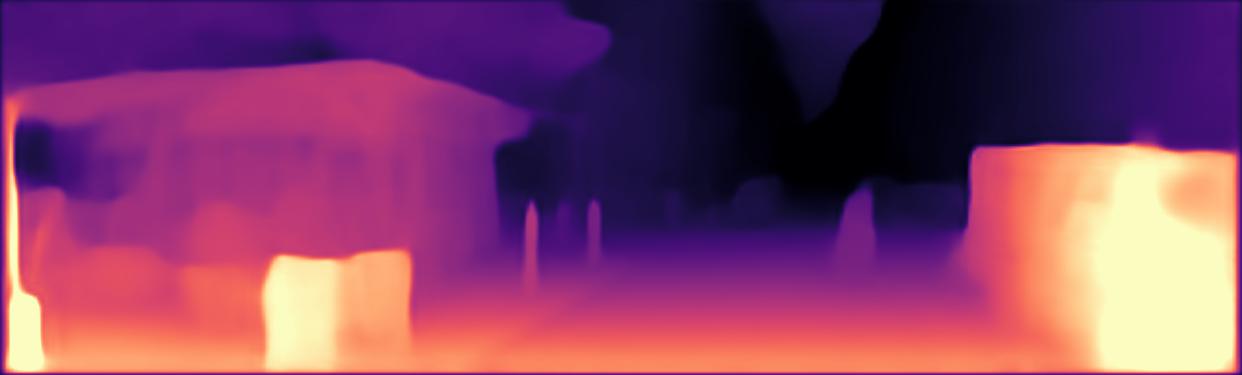} &
\includegraphics[height=\turnheightnew, width=90mm]{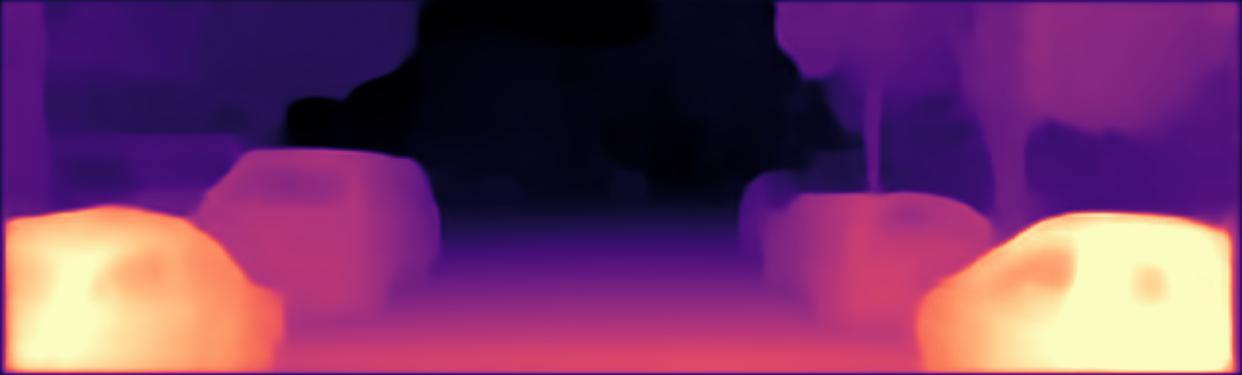} \\

{\rotatebox{90}{\hspace{3mm}\shortstack{\Large KITTI \\ \Large unrectified}}} &
\includegraphics[height=\turnheightnew]{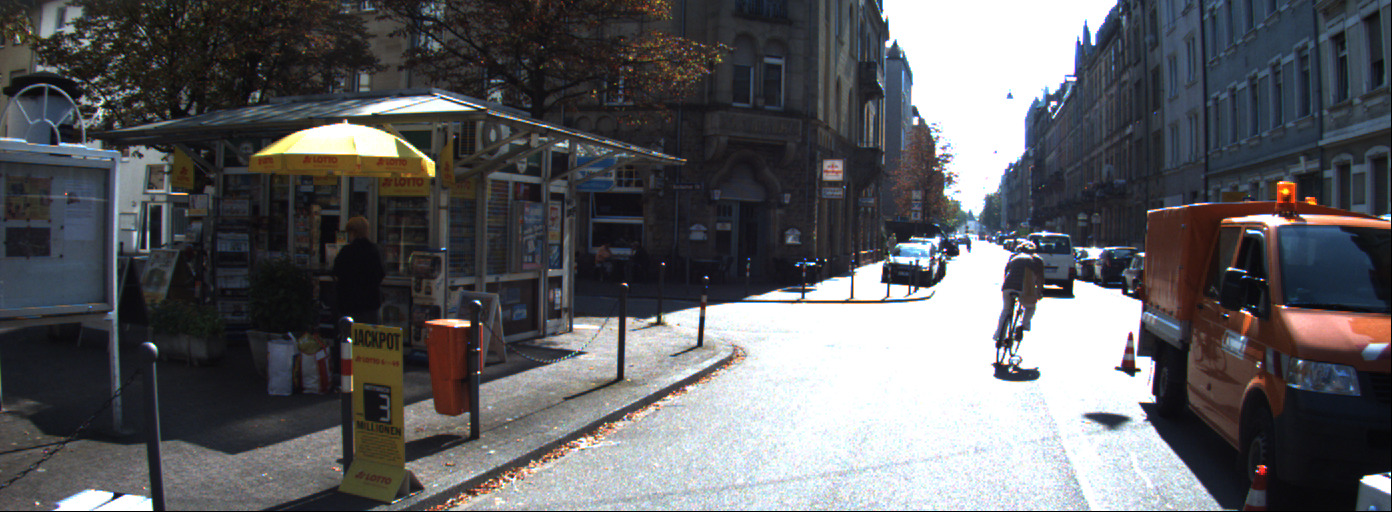} &
\includegraphics[height=\turnheightnew]{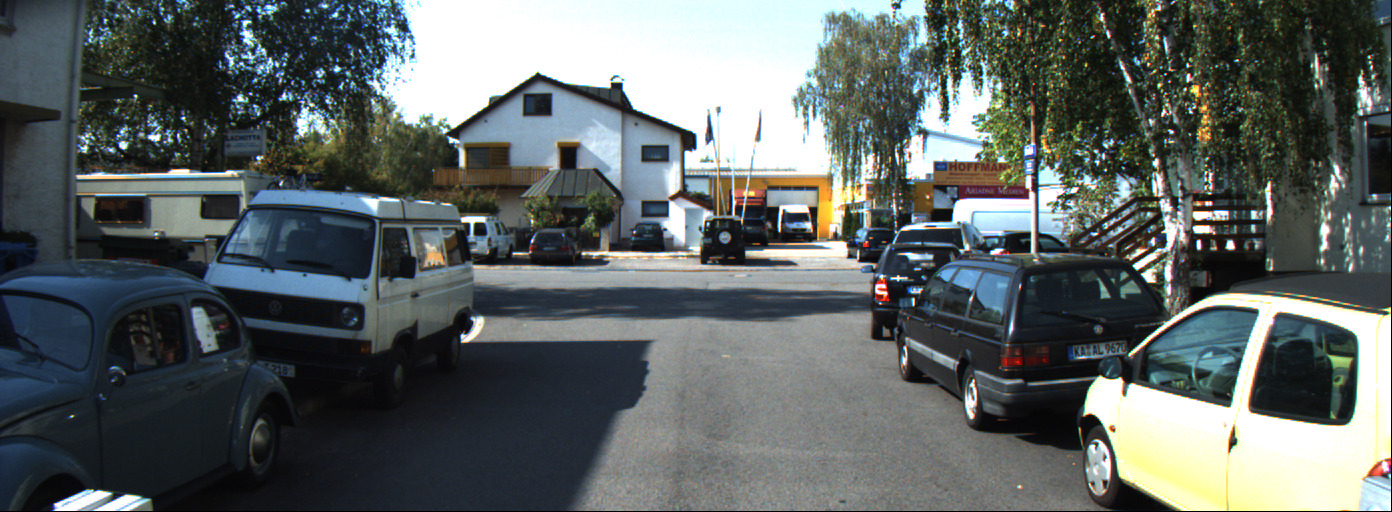} \\

{\rotatebox{90}{\hspace{0mm}\scriptsize}} &
\includegraphics[height=\turnheightnew]{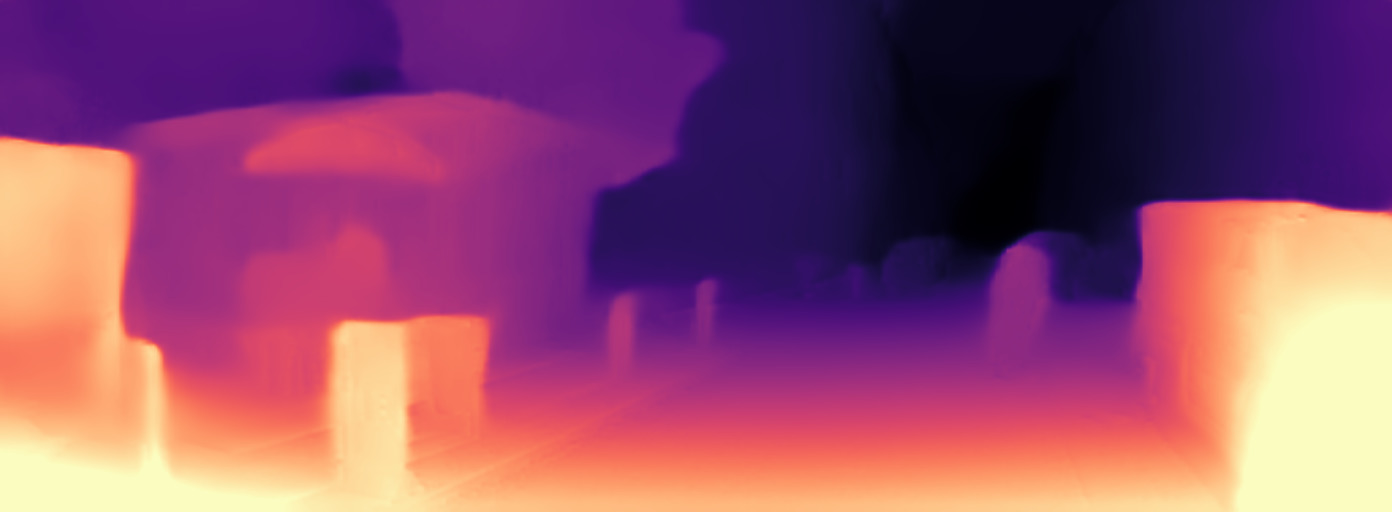} &
\includegraphics[height=\turnheightnew]{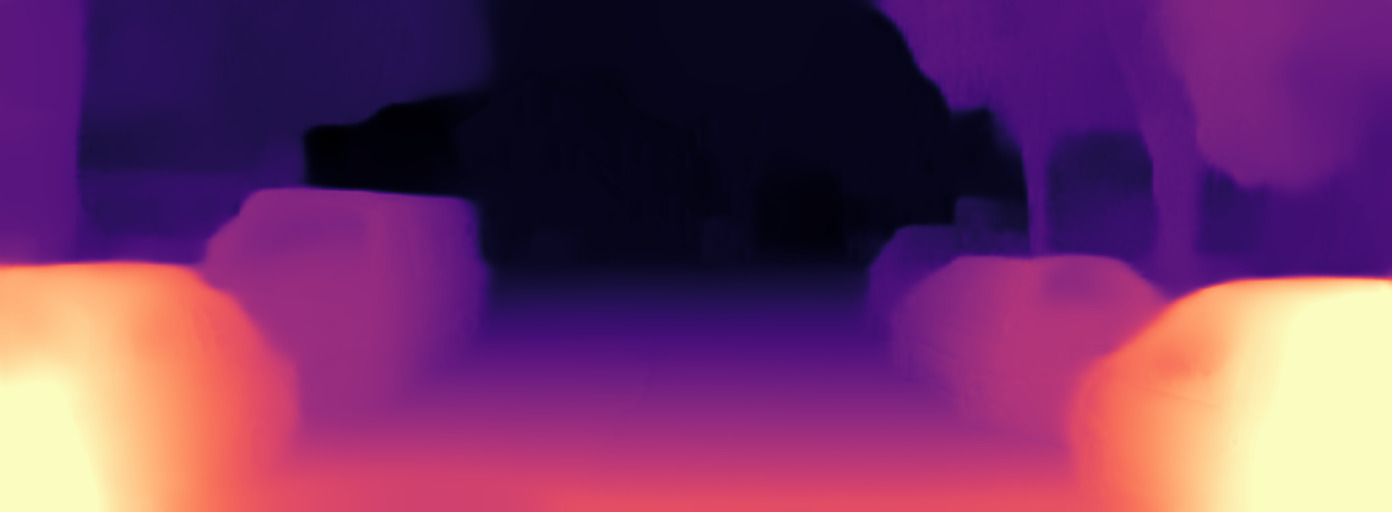} \\

{\rotatebox{90}{\hspace{7mm}\shortstack{\Large WoodScape \\ \Large cropped}}} &
\includegraphics[height=50mm, width=96mm]{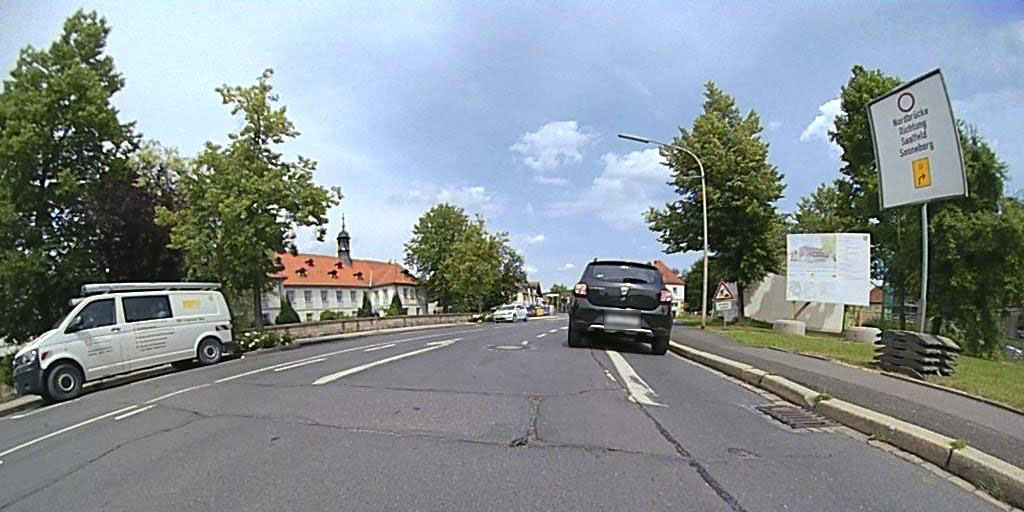} &
\includegraphics[height=50mm, width=96mm]{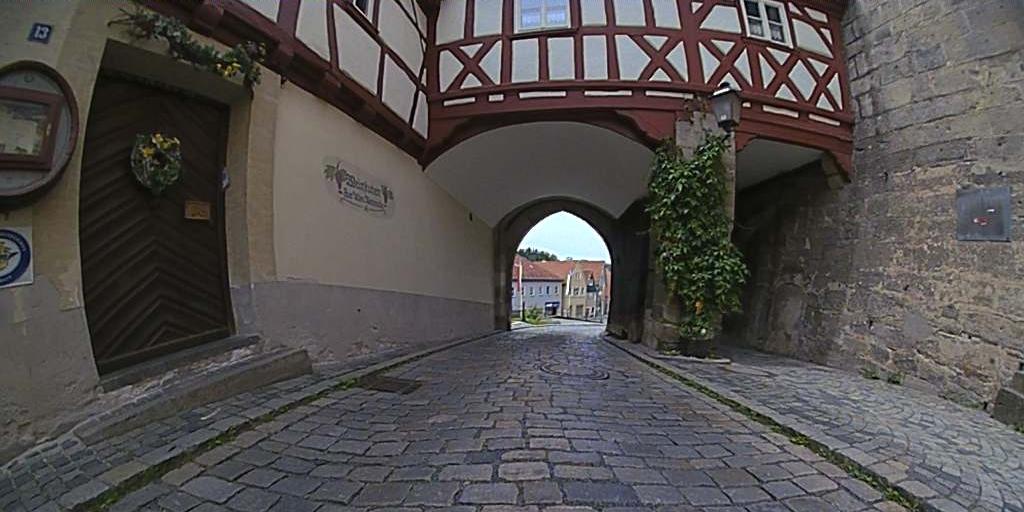} \\

{\rotatebox{90}{\hspace{0mm}}} &
\includegraphics[height=50mm, width=96mm]{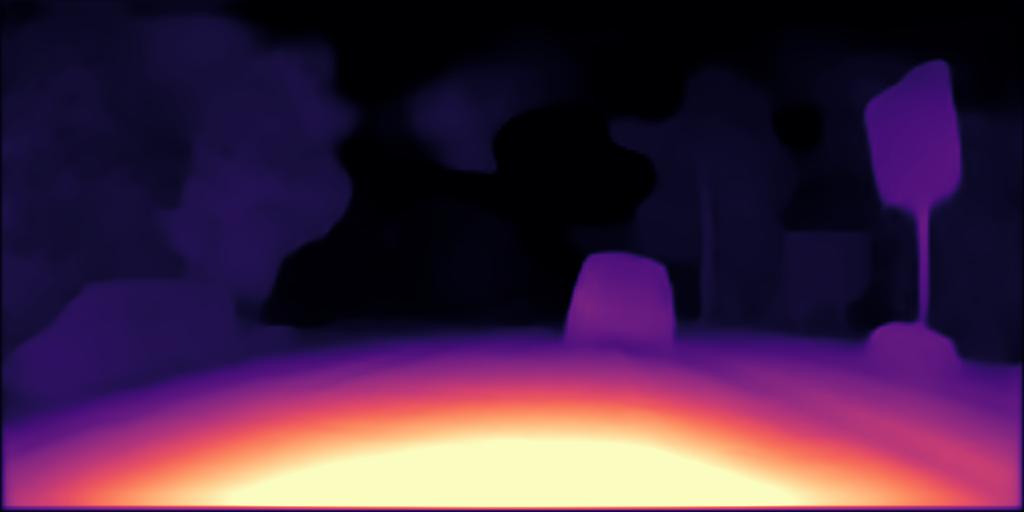} &
\includegraphics[height=50mm, width=96mm]{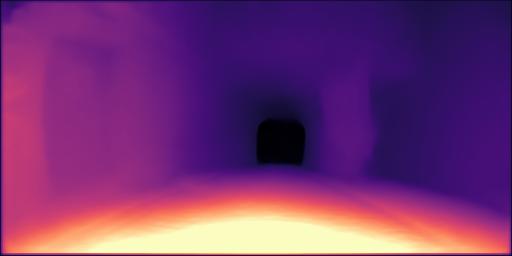} \\

\end{tabular}
}
  \caption[\bf Qualitative results comparison of UnRectDepthNet on KITTI and WoodScape dataset.]
  {\textbf{Qualitative results comparison of UnRectDepthNet on KITTI and WoodScape dataset.} The results on a distorted test video sequence indicate excellent performance, see \url{https://youtu.be/K6pbx3bU4Ss}.}
  \label{fig:qual}
\end{figure*}
\begin{figure*}[htbp]
  \centering
  \resizebox{\textwidth}{!}{
  \newcommand{\turnheightnew}{0.25\columnwidth}
\centering

\begin{tabular}{@{\hskip 0.5mm}c@{\hskip 0.5mm}c@{\hskip 0.5mm}c@{\hskip 0.5mm}c@{\hskip 0.5mm}c@{}}

{\rotatebox{90}{\hspace{5mm}\shortstack{\Large KITTI \\ \Large rectified}}} &
\includegraphics[height=\turnheightnew, width=90mm]{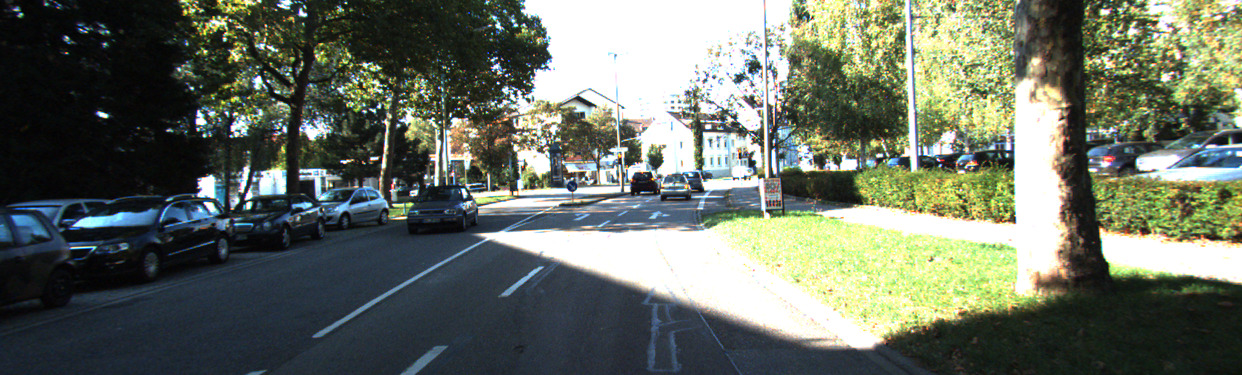} &
\includegraphics[height=\turnheightnew, width=90mm]{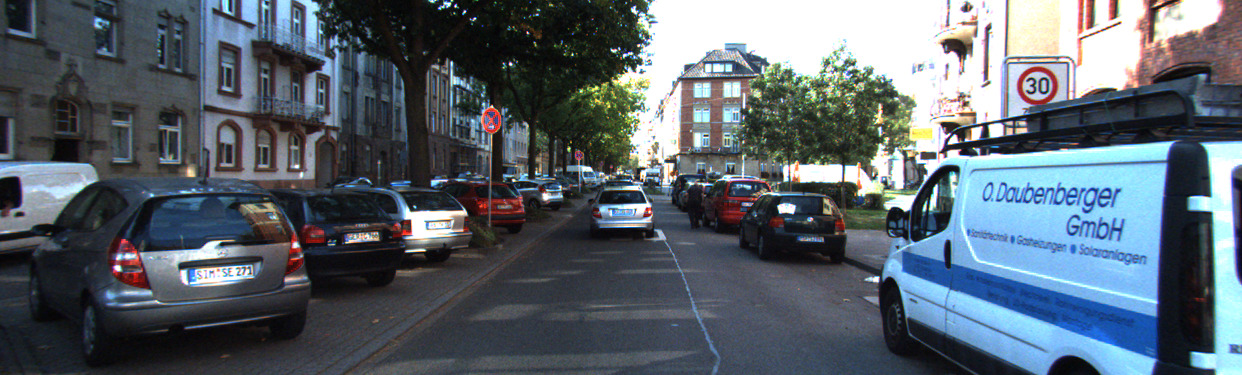}\\

{\rotatebox{90}{\hspace{0mm}\scriptsize}} &
\includegraphics[height=\turnheightnew, width=90mm]{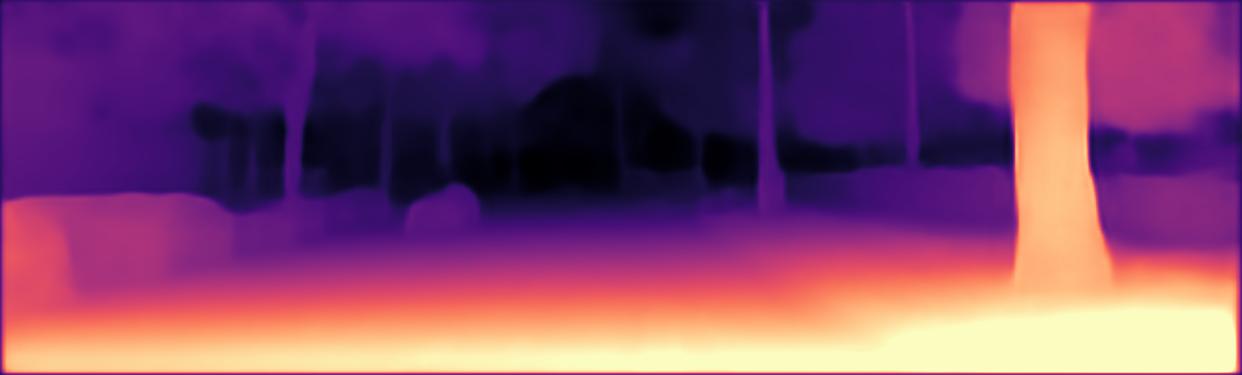} &
\includegraphics[height=\turnheightnew, width=90mm]{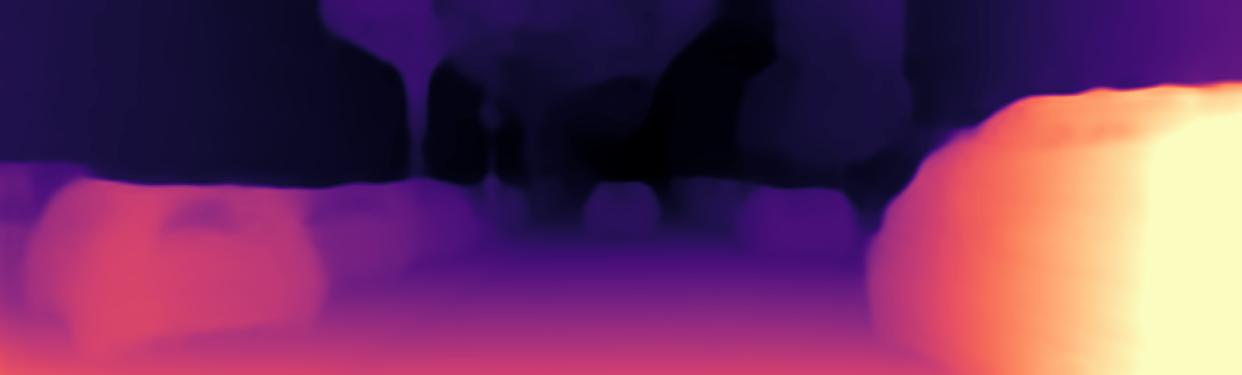} \\

{\rotatebox{90}{\hspace{3mm}\shortstack{\Large KITTI \\ \Large unrectified}}} &
\includegraphics[height=\turnheightnew]{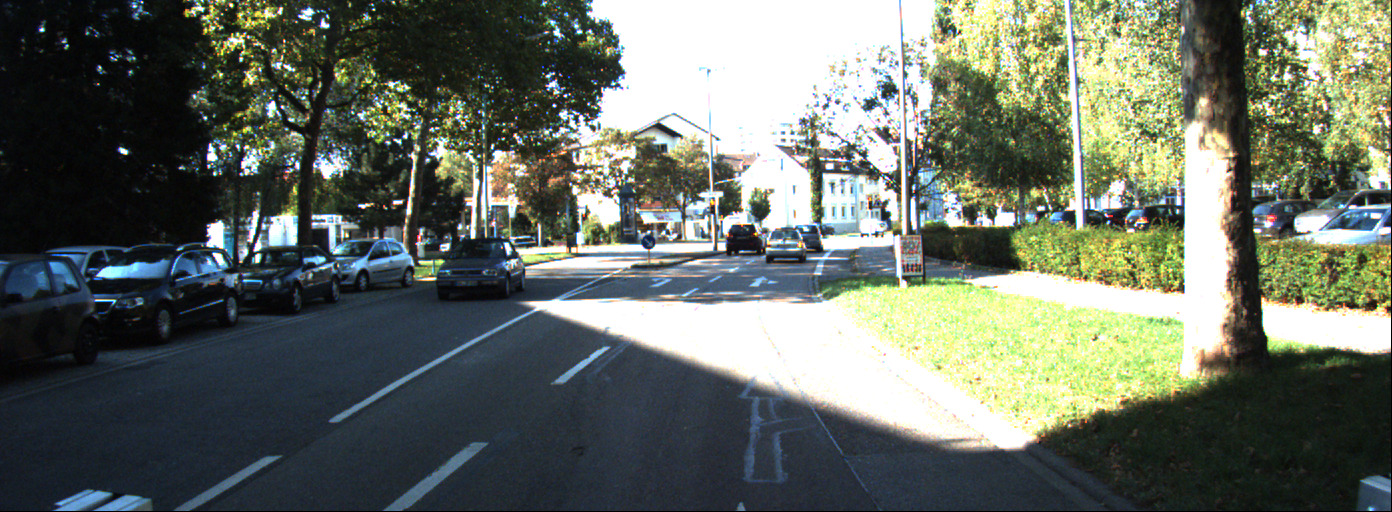} &
\includegraphics[height=\turnheightnew]{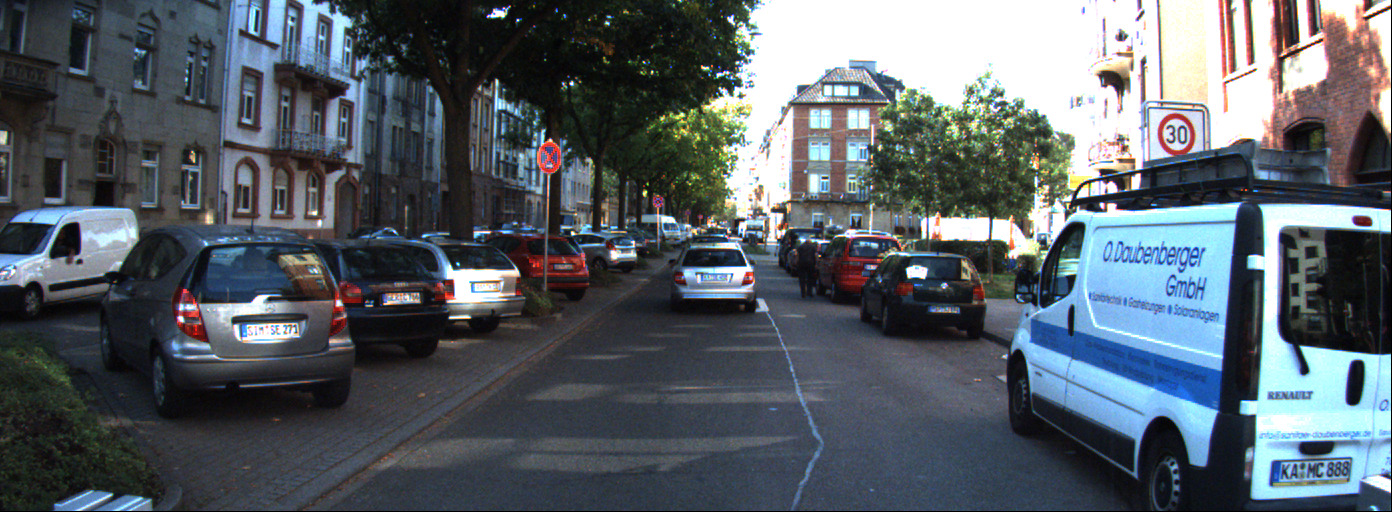}\\

{\rotatebox{90}{\hspace{0mm}\scriptsize}} &
\includegraphics[height=\turnheightnew]{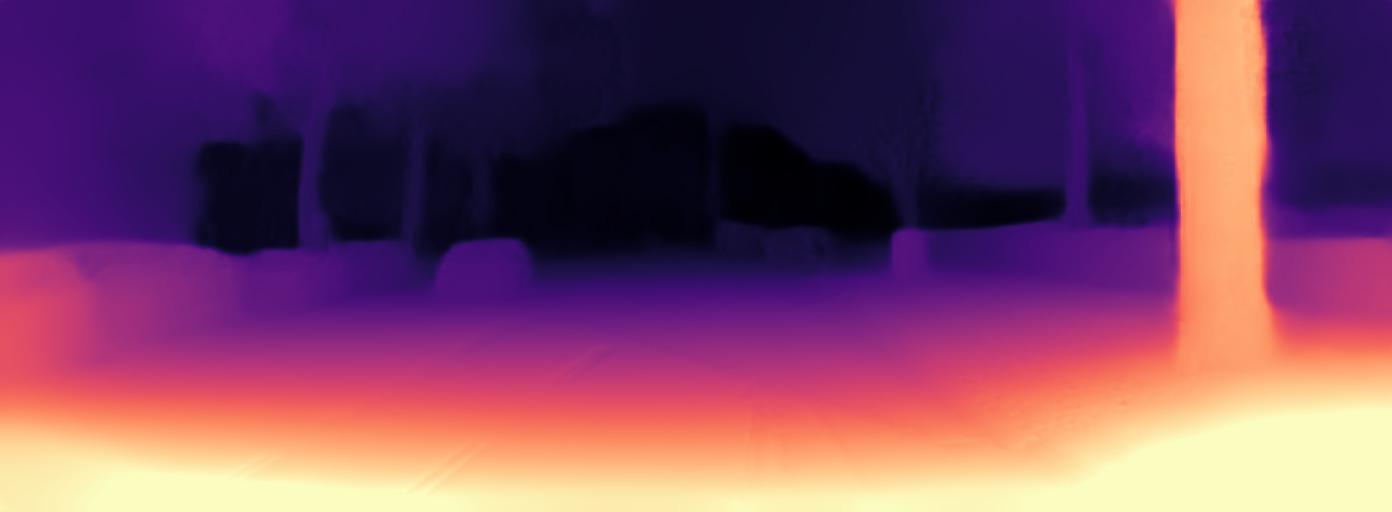} &
\includegraphics[height=\turnheightnew]{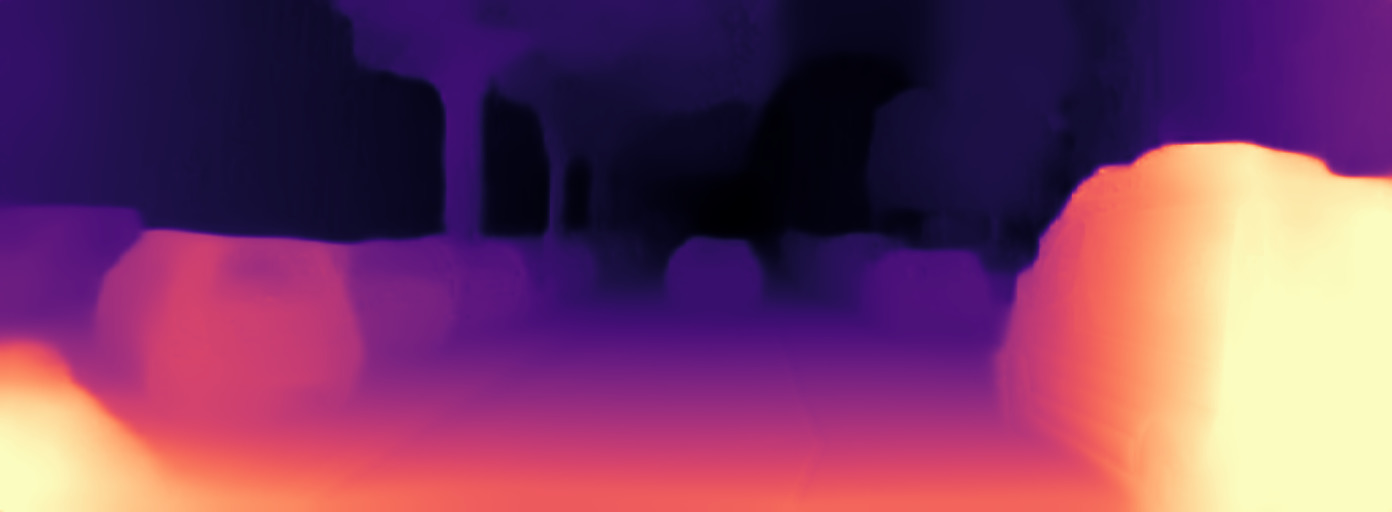} \\

{\rotatebox{90}{\hspace{7mm}\shortstack{\Large WoodScape \\ \Large cropped}}} &
\includegraphics[height=50mm, width=96mm]{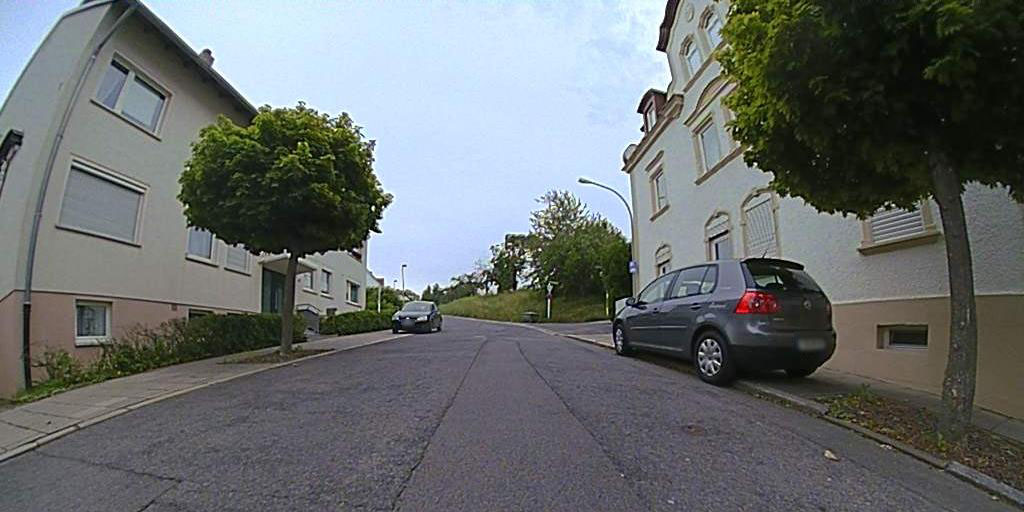} &
\includegraphics[height=50mm, width=96mm]{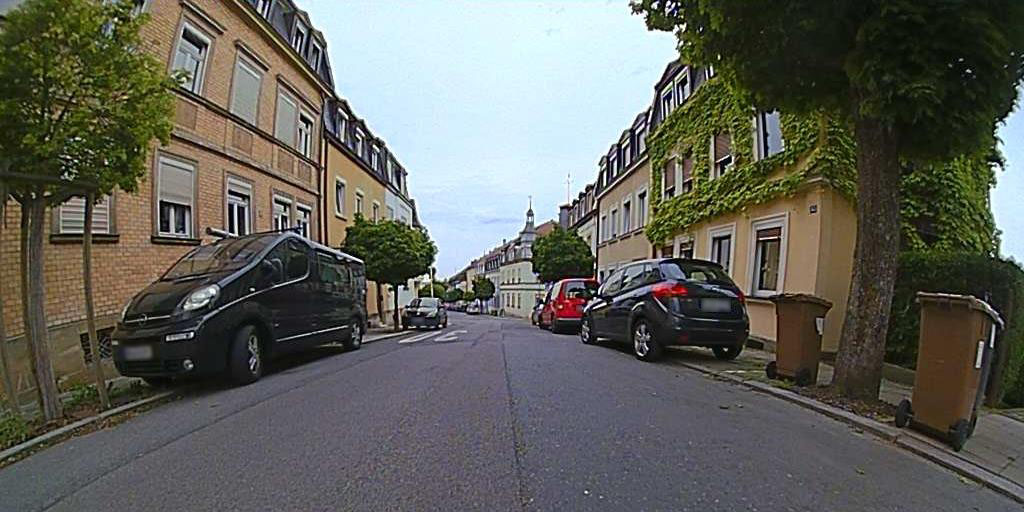}\\

{\rotatebox{90}{\hspace{0mm}}} &
\includegraphics[height=50mm, width=96mm]{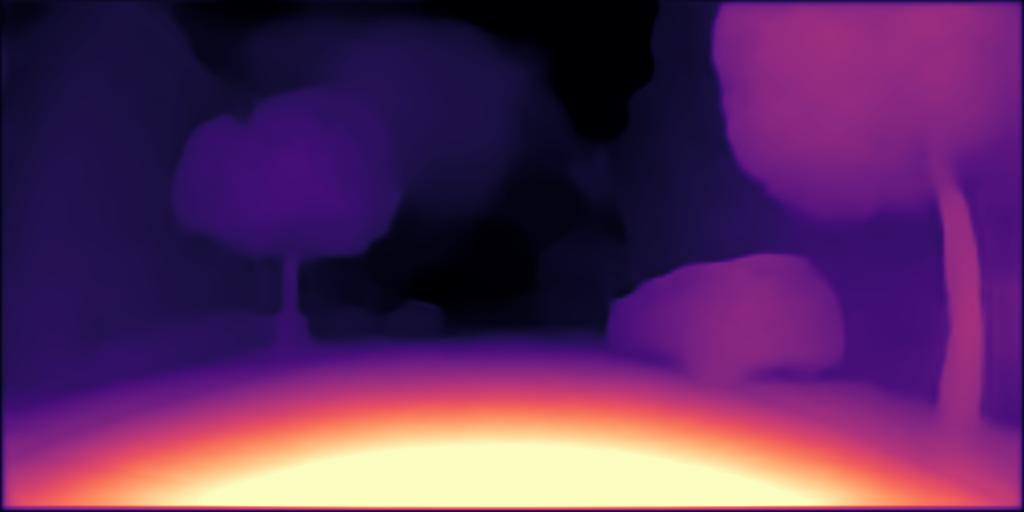} &
\includegraphics[height=50mm, width=96mm]{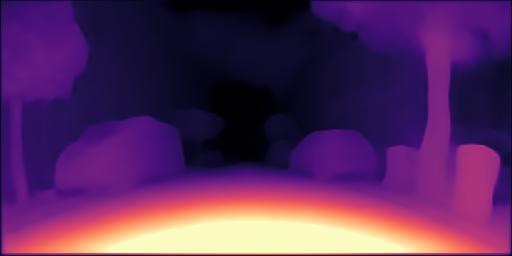}\\

\end{tabular}
}
  \caption{\bf Additional qualitative results comparison of UnRectDepthNet on KITTI and WoodScape dataset.}
  \label{fig:qual2}
\end{figure*}

\begin{table*}[!t]
\centering
\begin{adjustbox}{width=\columnwidth}
\begin{tabular}{lccccccc}
    \toprule
    & \cellcolor[HTML]{7d9ebf} Abs Rel 
    & \cellcolor[HTML]{7d9ebf} Sq Rel 
    & \cellcolor[HTML]{7d9ebf} RMSE 
    & \cellcolor[HTML]{7d9ebf} RMSE$_{log}$ 
    & \cellcolor[HTML]{e8715b} $\delta < 1.25$ 
    & \cellcolor[HTML]{e8715b} $\delta < 1.25^2$ 
    & \cellcolor[HTML]{e8715b} $\delta < 1.25^3$ \\
    \cmidrule(lr){2-5} \cmidrule(lr){6-8}
    \textbf{Approach} 
    & \multicolumn{4}{c}{\cellcolor[HTML]{7d9ebf} lower is better} & \multicolumn{3}{c}{\cellcolor[HTML]{e8715b} higher is better}\\
    \midrule
    \multicolumn{8}{c}{\cellcolor[HTML]{448BE9}\textbf{\textit{KITTI}}} \\
    \midrule
    Zhou~\cite{zhou2017unsupervised}\textdagger       & 0.183 & 1.595 & 6.709 & 0.270 & 0.734 & 0.902 & 0.959 \\
    Yang~\cite{yang2018unsupervised}                  & 0.182 & 1.481 & 6.501 & 0.267 & 0.725 & 0.906 & 0.963 \\
    Vid2depth~\cite{mahjourian2018unsupervised}       & 0.163 & 1.240 & 6.220 & 0.250 & 0.762 & 0.916 & 0.968 \\
    GeoNet~\cite{yin2018geonet}\textdagger            & 0.149 & 1.060 & 5.567 & 0.226 & 0.796 & 0.935 & 0.975 \\
    DDVO~\cite{Wang_2018_CVPR}                        & 0.151 & 1.257 & 5.583 & 0.228 & 0.810 & 0.936 & 0.974 \\
    DF-Net~\cite{zou2018df}                           & 0.150 & 1.124 & 5.507 & 0.223 & 0.806 & 0.933 & 0.973 \\
    Ranjan~\cite{ranjan2019competitive}               & 0.148 & 1.149 & 5.464 & 0.226 & 0.815 & 0.935 & 0.973 \\
    EPC++~\cite{luo2019every}                         & 0.141 & 1.029 & 5.350 & 0.216 & 0.816 & 0.941 & 0.976 \\
    Struct2depth `(M)'~\cite{casser2019depth}         & 0.141 & 1.026 & 5.291 & 0.215 & 0.816 & 0.945 & 0.979 \\
    Zhou~\cite{zhou2018unsupervised}                  & 0.139 & 1.057 & 5.213 & 0.214 & 0.831 & 0.940 & 0.975 \\
    PackNet-SfM~\cite{guizilini2019packnet}           & 0.120 & 0.892 & 4.898 & 0.196 & 0.864 & 0.954 & 0.980 \\
    Monodepth2~\cite{godard2019digging}               & \textbf{0.115} & 0.903 & 4.863 & 0.193 & \textbf{0.877} & 0.959 & 0.981 \\
    \textbf{FisheyeDistanceNet}                       & 0.117 & \textbf{0.867} & \textbf{4.739} & \textbf{0.190} & 0.869 & \textbf{0.960} & \textbf{0.982} \\
    \textbf{FisheyeDistanceNet} (1024 $\times$ 320)   & 0.109 & 0.788 & 4.669 & 0.185 & 0.889 & 0.964 & 0.982 \\
    \midrule
    \multicolumn{8}{c}{\cellcolor[HTML]{34FF34}\textbf{\textit{WoodScape}}} \\
    \midrule
    FisheyeDistanceNet cap $80\,\text{m}$             & 0.167 & 1.108 & 3.814 & 0.216 & 0.794 & 0.953 & 0.972 \\
    FisheyeDistanceNet cap $40\,\text{m}$             & 0.152 & 0.768 & 2.723 & 0.210 & 0.812 & 0.954 & 0.974 \\
    FisheyeDistanceNet cap $30\,\text{m}$             & 0.149 & 0.613 & 2.402 & 0.204 & 0.810 & 0.957 & 0.976 \\
    \bottomrule
\end{tabular}
\end{adjustbox}
\caption[\bf Quantitative results of leaderboard algorithms on KITTI and WoodScape dataset.]
{\textbf{Quantitative results of leaderboard algorithms on KITTI dataset~\cite{geiger2013vision} and FisheyeDistanceNet on WoodScape~\cite{yogamani2019woodscape}}. \textdagger~marks newer results reported on GitHub.}
\label{tab:results}
\end{table*}
\begin{table*}[!ht]
\renewcommand{\arraystretch}{0.87}
\centering
\begin{adjustbox}{width=\columnwidth}
\small
\setlength{\tabcolsep}{0.3em}
\begin{tabular}{c|lccccccccc}
\toprule
& \textbf{Method} 
& Resolution 
& Dataset 
& \cellcolor[HTML]{7d9ebf} Abs Rel 
& \cellcolor[HTML]{7d9ebf} Sq Rel 
& \cellcolor[HTML]{7d9ebf} RMSE 
& \cellcolor[HTML]{7d9ebf} RMSE$_{log}$ 
& \cellcolor[HTML]{e8715b}$\delta<1.25$
& \cellcolor[HTML]{e8715b}$\delta<1.25^2$ 
& \cellcolor[HTML]{e8715b}$\delta<1.25^3$ \\
\cmidrule(lr){5-8} \cmidrule(lr){9-11}
& & & & \multicolumn{4}{c}{\cellcolor[HTML]{7d9ebf} lower is better} & \multicolumn{3}{c}{\cellcolor[HTML]{e8715b} higher is better}\\
\toprule
\parbox[t]{2mm}{\multirow{14}{*}{\rotatebox[origin=c]{90}{Original~\cite{Eigen_14}}}}
& SfMLeaner~\cite{zhou2017unsupervised}            & 416 x 128  & K  & 0.183 & 1.595 & 6.709 & 0.270 & 0.734 & 0.902 & 0.959 \\
& Vid2depth~\cite{mahjourian2018unsupervised}      & 416 x 128  & K  & 0.163 & 1.240 & 6.220 & 0.250 & 0.762 & 0.916 & 0.968 \\
& DDVO~\cite{Wang_2018_CVPR}                       & 416 x 128  & K  & 0.151 & 1.257 & 5.583 & 0.228 & 0.810 & 0.936 & 0.974 \\
& EPC++~\cite{luo2019every}                        & 640 x 192  & K  & 0.141 & 1.029 & 5.350 & 0.216 & 0.816 & 0.941 & 0.976 \\
& Struct2Depth~\cite{casser2019depth}              & 416 x 128  & K  & 0.141 & 1.026 & 5.291 & 0.215 & 0.816 & 0.945 & 0.979 \\
& Monodepth2~\cite{godard2019digging}              & 640 x 192  & K  & 0.115 & 0.903 & 4.863 & 0.193 & 0.877 & 0.959 & 0.981 \\ 
& PackNet-SfM~\cite{guizilini2019packnet}          & 640 x 192  & K  & 0.111 & 0.785 & 4.601 & 0.189 & 0.878 & 0.960 & 0.982 \\
& Monodepth2~\cite{godard2019digging}              & 1024 x 320 & K  & 0.115 & 0.882 & 4.701 & 0.190 & 0.879 & 0.961 & 0.982 \\
\cmidrule{2-11}
& \textbf{UnRectDepthNet}                          & 640 x 192  & K  & \textbf{0.107} & \textbf{0.721} & \textbf{4.564} &\textbf{0.178} & \textbf{0.894} & \textbf{0.971} & \textbf{0.986} \\
& \textbf{UnRectDepthNet}                          & 1024 x 320 & K  & 0.103 & 0.705 & 4.386 & 0.164 & 0.897 & 0.980 & 0.989 \\
& \textbf{UnRectDepthNet}                          & 608 x 224  & KD & 0.102 & 0.720 & 4.559 & 0.183 & 0.892 & 0.973 & 0.988 \\
& \textbf{UnRectDepthNet}                          & 1216 x 448 & KD & 0.106 & 0.709 & 4.357 & 0.161 & 0.895 & 0.984 & 0.992 \\
& FisheyeDistanceNet~\cite{kumar2020fisheyedistancenet} & 512 x 256  & WS & 0.152 & 0.768 & 2.723 & \textbf{0.210} & 0.812 & 0.954 & 0.974 \\
& \textbf{UnRectDepthNet}                          & 512 x 256  & WS & \textbf{0.148} & \textbf{0.702} & \textbf{2.530} & 0.212 & \textbf{0.826} & \textbf{0.960} & \textbf{0.980} \\
\midrule
\parbox[t]{2mm}{\multirow{8}{*}{\rotatebox[origin=c]{90}{Improved~\cite{uhrig2017sparsity}}}}
& SfMLeaner~\cite{zhou2017unsupervised}            & 416 x 128 & K & 0.176 & 1.532 & 6.129 & 0.244 & 0.758 & 0.921 & 0.971 \\
& Vid2Depth~\cite{mahjourian2018unsupervised}      & 416 x 128 & K & 0.134 & 0.983 & 5.501 & 0.203 & 0.827 & 0.944 & 0.981 \\
& DDVO~\cite{Wang_2018_CVPR}                       & 416 x 128 & K & 0.126 & 0.866 & 4.932 & 0.185 & 0.851 & 0.958 & 0.986 \\
& EPC++~\cite{luo2019every}                        & 640 x 192 & K & 0.120 & 0.789 & 4.755 & 0.177 & 0.856 & 0.961 & 0.987 \\
& Monodepth2~\cite{godard2019digging}              & 640 x 192 & K & 0.090 & 0.545 & 3.942 & 0.137 & 0.914 & 0.983 & 0.995 \\
& PackNet-SfM~\cite{guizilini2019packnet}          & 640 x 192 & K & \textbf{0.078} & 0.420 & 3.485 & 0.121 & \textbf{0.931} & 0.986 & \textbf{0.996} \\
\cmidrule{2-11} 
& \textbf{UnRectDepthNet}                          & 640 x 192 & K  & 0.081 & \textbf{0.414} & \textbf{3.412} & \textbf{0.117} & 0.926 & \textbf{0.987} & \textbf{0.996} \\
& \textbf{UnRectDepthNet}                          & 640 x 224 & KD & 0.092 & 0.458 & 3.503 & 0.132 & 0.906 & 0.971 & 0.990 \\
\bottomrule
\end{tabular}
\end{adjustbox}
\caption[\bf Quantitative performance comparison of depth estimation in UnRectDepthNet.]
{\textbf{Quantitative performance comparison of UnRectDepthNet} for depths up to 80\,m for KITTI and 40\,m for WoodScape. In the Dataset column, K refers to KITTI~\cite{geiger2012we}, KD refers to the KITTI distorted~\cite{geiger2013vision}, and WS refers to WoodScape~\cite{yogamani2019woodscape} dataset. \textit{Original} refers to  depth maps defined in \cite{Eigen_14}, and \textit{Improved} refers to refined depth maps provided by \cite{uhrig2017sparsity}.}
\label{table:unrectdepthnet-results}
\end{table*}
We evaluate \textit{FisheyeDistanceNet} and \textit{UnRectDepthNet} distance and depth estimation results using the metrics illustrated in Table~\ref{tab:depth-metrics} on KITTI and WoodScape datasets as described in Section~\ref{sec:benchmarks} and report the results for less than $80\,m$ as indicated in~\cite{Eigen_14} for the pinhole model to facilitate a comparison. The quantitative results are shown in Tables~\ref{tab:results} and~\ref{table:unrectdepthnet-results} illustrates that the scale-aware self-supervised approach outperforms almost all the state-of-the-art monocular approaches. All the methods listed in the table are self-supervised approaches on monocular camera sequences. At inference time, all the approaches except \textit{FisheyeDistanceNet}, \textit{UnRectDepthNet}, and PackNet-SfM scale the estimated depths using median ground-truth LiDAR depth. We generalized our previous model \textit{FisheyeDistanceNet} in our new training framework and added additional features that improve results on WoodScape. For the fisheye dataset, we estimate distance rather than depth. The qualitative results are illustrated in Figures~\ref{fig:fisheye_qual}, \ref{fig:qual} and~\ref{fig:qual2} where the framework produces sharp distance maps on raw fisheye and pinhole images, respectively. The KITTI distorted results are better than most of the previous outcomes obtained with self-supervised approaches on the corresponding rectified dataset. We could not leverage the Cityscapes dataset into the training regime to benchmark the scale-aware framework due to the absence of odometry data.\par

Since the projection operators are different, previous \emph{SfM} approaches will not be feasible on the Woodscape dataset without adapting the network and projection model. It is important to note that due to the fisheye's geometry, it would not be a fair comparison to evaluate the distance estimates up to $80\,m$. The fisheye automotive cameras also undergo high data compression, and the dataset contains images of inferior quality compared with KITTI. The fisheye cameras can perform well up to a range of $40\,m$. Therefore, we also report results on a $30\,m$, and a $40\,m$ range (see Table~\ref{tab:results}).\par

We generalized the training methodology of this model to incorporate any arbitrary distortion model. We also tuned the network to the optimal hyperparameters using grid search and removed batch normalization in the decoder as we observed ghosting effects and holes in homogeneous areas. We calculated the minimum reconstruction error for the two warps of the backward sequence individually compared to a combined minimization for forward sequences since here the target frames are $I_{t'}$ ($t^\prime \in \{t+1, t-1\}$).\par
\begin{table*}[t!]
	\small
	\begin{center}
	\begin{adjustbox}{width=\columnwidth}
	\begin{tabular}{l|c|c|c|c|c|c|c|c|c|c|c|c}
	\toprule
	Method & FS & BS & SR & CSDCL & DCN 
	& \cellcolor[HTML]{7d9ebf} Abs Rel 
	& \cellcolor[HTML]{7d9ebf} Sq Rel 
	& \cellcolor[HTML]{7d9ebf} RMSE 
	& \cellcolor[HTML]{7d9ebf} RMSE$_{log}$ 
	& \cellcolor[HTML]{e8715b} $\delta < 1.25$ 
	& \cellcolor[HTML]{e8715b} $\delta < 1.25^2$ 
	& \cellcolor[HTML]{e8715b} $\delta < 1.25^3$\\
	\toprule
	Ours & \ch & \ch & \ch & \ch & \ch & 0.152 & 0.768 & 2.723 & 0.210 & 0.812 & 0.954 & 0.974 \\
	Ours & \ch & \xm & \ch & \ch & \ch & 0.172 & 0.829 & 2.925 & 0.243 & 0.802 & 0.952 & 0.970 \\
	Ours & \ch & \xm & \xm & \ch & \ch & 0.181 & 0.913 & 3.180 & 0.250 & 0.823 & 0.938 & 0.963 \\
	Ours & \ch & \xm & \xm & \xm & \ch & 0.190 & 0.997 & 3.266 & 0.258 & 0.796 & 0.930 & 0.963 \\
    Ours & \ch & \xm & \xm & \xm & \xm & 0.201 & 1.282 & 3.589 & 0.276 & 0.590 & 0.898 & 0.949 \\
	\bottomrule
\end{tabular}
\end{adjustbox}
\end{center}
\caption[\bf Ablation study on different variants of the FisheyeDistanceNet using the WoodScape dataset.]
{\textbf{Ablation study on different variants of the FisheyeDistanceNet using the WoodScape dataset.} BS, SR, CSDCL, and DCN represent a backward sequence, super-resolution network with PixelShuffle, or sub-pixel convolution initialized to convolution NN resize (ICNR)~\cite{aitken2017checkerboard}, cross-sequence distance consistency loss, and deformable convolutions respectively.}
\label{table:ablation}
\end{table*}
\subsection{Fisheye Ablation Study}

We conduct an ablation study to evaluate the importance of different components. We cap the distances at $40\,m$ and the input resolution is $512\times256$ pixels. We ablate the following components and report their impact on the distance evaluation metrics in Table~\ref{table:ablation}: 
\begin{itemize}
    \item \textit{Remove Backward Sequence}: The network is only trained for the forward sequence, which consists of two warps as explained in Section~\ref{sec:backward sequence}. This has a critical impact on the model's performance as the induced baseline during training decreases due to fewer warps. The only advantage of this is lesser training time.
    \item \textit{Additionally remove Super-Resolution using sub-pixel convolution}: Removal of sub-pixel convolution has a significant impact on Woodscape compared to KITTI. This is mainly attributed to the fisheye model, as far-away objects are tiny and cannot be resolved accurately with naive nearest-neighbor interpolation or transposed convolution~\cite{odena2016deconvolution}.
    \item \textit{Additionally remove cross-sequence distance consistency loss}: Removing the CSDCL mainly diminishes the baseline.
    \item \textit{Additionally remove deformable convolutions}: If we remove all the major components, especially deformable convolution layers~\cite{zhu2019deformable}, the model will fail miserably as the distortion introduced by the fisheye model will not be learned correctly by normal convolutional layers.
\end{itemize}
\begin{table*}[t!]
\small
\begin{center}
\begin{adjustbox}{width=\columnwidth}
	\begin{tabular}{l|c|c|c|c|c|c|c|c|c|c|c}
	\toprule
    Method & FS & BS & SR & CSDCL 
    & \cellcolor[HTML]{7d9ebf} Abs Rel 
    & \cellcolor[HTML]{7d9ebf} Sq Rel 
    & \cellcolor[HTML]{7d9ebf} RMSE 
    & \cellcolor[HTML]{7d9ebf} RMSE$_{log}$ 
    & \cellcolor[HTML]{e8715b} $\delta < 1.25$ 
    & \cellcolor[HTML]{e8715b} $\delta < 1.25^2$ 
    &  \cellcolor[HTML]{e8715b} $\delta < 1.25^3$\\
    \toprule
    Ours & \ch & \ch & \ch & \ch & 0.102 & 0.720 & 4.559 & 0.183 & 0.892 & 0.973 & 0.988 \\
    Ours & \ch & \xm & \ch & \ch & 0.131 & 0.856 & 4.933 & 0.198 & 0.853 & 0.954 & 0.968 \\
    Ours & \ch & \xm & \xm & \ch & 0.141 & 0.971 & 5.183 & 0.206 & 0.831 & 0.941 & 0.953 \\
    Ours & \ch & \xm & \xm & \xm & 0.144 & 1.011 & 5.204 & 0.225 & 0.822 & 0.945 & 0.949 \\
    \bottomrule
\end{tabular}
\end{adjustbox}
\end{center}
\caption[\bf Ablation study of UnrectdepthNet on the KITTI dataset.]
{\textbf{Ablation study of UnrectdepthNet on the KITTI dataset.} Depths are capped at 80\,m. FS, BS, SR, CSDCL indicate forward sequence, backward sequence, super-resolution network with PixelShuffle~\cite{aitken2017checkerboard} layers and cross-sequence depth consistency loss, respectively. The input resolution is $608\times224$ pixels.} 
\label{table:unrectdepthnet-kitti}
\end{table*}
\subsection{KITTI Distorted Ablation Study}

We perform an ablation study to understand the significance of different components used and tabulate in Table~\ref{table:unrectdepthnet-kitti}: 
\begin{itemize}
    \item \textit{Remove Backward Sequence}: The network is trained only for a forward sequence consisting of two warps, as explained in~\cite{kumar2020fisheyedistancenet}. The impact is significant in the border areas as fewer constraints are induced. The model inherently fails to resolve unknown depths in those areas at the test time, which was also observed in previous works~\cite{godard2019digging,zhou2017unsupervised,yin2018geonet}. The only advantage of this is lesser training time.
    \item \textit{Additionally remove Super-Resolution using sub-pixel convolution}: It has a significant effect as distant objects are small in fisheye cameras and cannot be resolved correctly with simple nearest-neighbor interpolation or transposed convolution~\cite{odena2016deconvolution}.
    \item \textit{Additionally remove cross-sequence depth consistency loss}: The removal of the CSDCL diminishes the baseline, induces fewer constraints, and the model is therefore not robust enough to yield accurate depth estimates.
\end{itemize}
\section{Conclusion}

This chapter presented two novel strategies in the field of self-supervised distance estimation. Firstly, a novel self-supervised training strategy to obtain metric distance maps on unrectified fisheye images. Secondly, a generic self-supervised training method for depth estimation handling distorted images. We showed that it is possible to support various commonly used automotive camera models in the framework and indicate empirical results on KITTI and WoodScape datasets. We show that the \textit{FisheyeDistanceNet} and \textit{UnRectDepthNet} establish a new state-of-the-art in the self-supervised monocular distance and depth estimation through extensive experiments on WoodScape and KITTI datasets, respectively. For KITTI, we show that depth estimation on unrectified images can produce the same accuracy as on rectified images. We obtain promising results, demonstrating the potential of using a CNN-based approach deployable in commercial automotive systems, particularly for replacing current classical depth estimation approaches.\par

In the following chapter, we will dig deeper into the choice of the self-supervised distance estimation loss functions and improve the model's robustness by incorporating semantic information. We will investigate how to leverage more directly the 2\textsuperscript{nd} class of perception algorithm: Semantic Segmentation to guide geometric representation learning and induce a synergy between both the tasks for autonomous driving. We will further extend this approach on surround-view cameras, and with novel techniques, we aim for a large scale deployment of the model.\par
\chapter{Geometry Meets Semantics}
\label{Chapter5}
\minitoc
\section{Problem Definition}

As in \textbf{Chapter}~\ref{Chapter4}, where we set up a \textit{SfM} framework for fisheye images to predict distance maps by performing view-synthesis, in this chapter, we examine how to leverage more directly the \textit{semantic} context of the scene to guide geometric representation learning while remaining in the self-supervised regime. Further, we dig deep into the loss functions and the architecture aspects for distance estimation and propose a near-field distance estimation solution on surround-view fisheye cameras targeting large-scale industrial deployment.\par

The first contribution in that sense is the novel application of a general and robust loss function proposed by~\cite{barron2019general} to the task of self-supervised distance estimation, which replaces the de facto standard of an \lone loss function used in previous approaches~\cite{casser2019depth, godard2019digging, guizilini2019packnet, kumar2020unrectdepthnet, kumar2020fisheyedistancenet}. The most important of all is a novel solution to filter out the dynamic objects from contaminating the photometric loss during training and the infinite distance issue during inference. This work was formally presented as \textit{SyndistNet}~\cite{kumar2020syndistnet} as an oral at the \href{https://arxiv.org/abs/2008.04017}{WACV} conference in 2021.\par

Typically, automotive perception systems use multiple cameras, with current systems having at least four cameras. The number is likely to increase to more than ten cameras for future generation systems. Such surround-view cameras are focused on near field sensing, which is typically used for low-speed applications such as parking or traffic jam assistance functions~\cite{heimberger2017computer}. The surround-view distance estimation framework will be facilitated by employing a single network on images from multiple cameras. A surround-view coverage of geometric information can be obtained for an autonomous vehicle by utilizing and post-processing the distance maps from all cameras.\par

Near-field distance estimation is a challenging problem because of distortion and partial visibility of close-by objects. Also, centimeter-level accuracy is required to enable precise low-speed maneuvers such as parking. Up to now, for the generation of high-quality distance estimates, one network per camera has to be trained, inducing unfeasible computational complexity with an increasing number of cameras. One of the thesis's main goals is to target the design of a model that can be deployed in millions of vehicles having its own set of cameras. To do so, we present a novel camera geometry adaptive multi-scale convolution to incorporate the camera parameters into the self-supervised distance estimation framework. This work was very influential from a product perspective to win next-generation projects and be influential in the academic community. This work formally presented as \textit{SVDistNet}~\cite{kumar2021svdistnet} to a journal at the \href{https://arxiv.org/abs/2104.04420}{T-ITS} in 2021.\par
\begin{figure}[!t]
  \centering
    \includegraphics[width=1.0\linewidth]{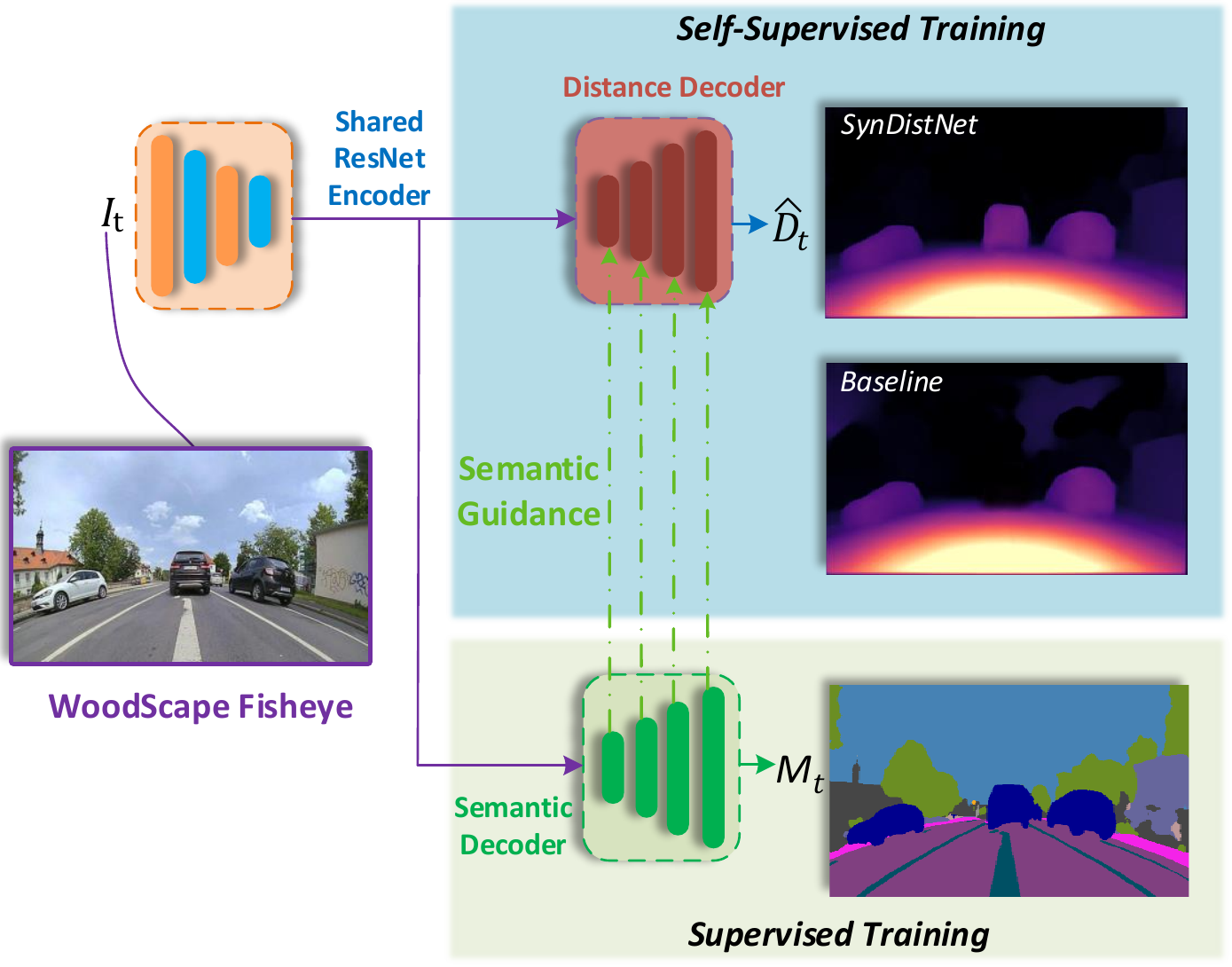}
    \caption[\bf Overview of the SynDistNet framework and a comparison with FisheyeDistanceNet baseline.]
    {\textbf{Overview of the SynDistNet framework and a comparison with FisheyeDistanceNet baseline.} We jointly predict the distance $\hat{D}_t$ and the semantic segmentation $M_t$ from a single input image $I_t$.}
    \label{fig:general_concept}
\end{figure}
As depicted in Figure~\ref{fig:general_concept}, dynamic objects induce a lot of unfavorable artifacts and hinder the photometric loss during the training, which results in infinite distance predictions, \eg, due to their violation of the static world assumption. Therefore, we use the segmentation masks to apply a simple semantic masking technique, based on the temporal consistency of consecutive frames, which delivers significantly improved results, \eg, concerning the infinite distance problem of objects moving at the same speed as the ego-camera. Compared to previous approaches, the semantically guided distance estimation in Figure~\ref{fig:general_concept} produces sharper depth edges and reasonable distance estimates for dynamic objects. Previous approaches~\cite{luo2019every, ranjan2019competitive} did predict these motion masks only implicitly as part of the projection model and therefore were limited to the projection model's fidelity.
\begin{figure*}[t]
  \centering
    \includegraphics[width=\textwidth]{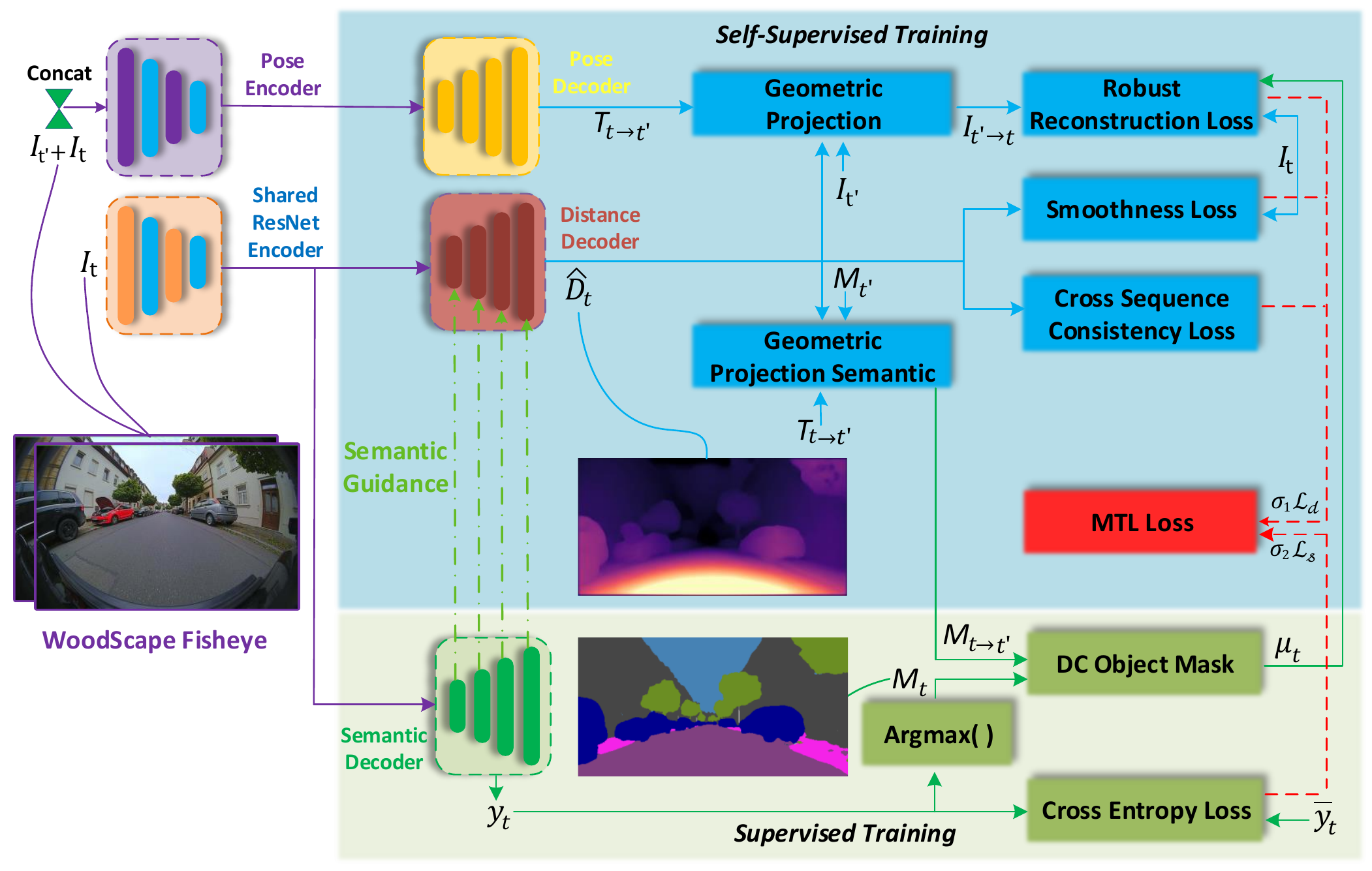}
    \caption[\bf Overview of the SynDistNet framework.]
            {\textbf{Overview of the SynDistNet framework} for the joint prediction of distance and semantic segmentation.}
    \label{fig:syndistnet-mtl_pipeline}
\end{figure*}
\section{Multi-Task Learning Framework}
\label{sec:syndistnet-method}

This section describes the framework for the multi-task learning of distance estimation and semantic segmentation as illustrated in Figure~\ref{fig:syndistnet-mtl_pipeline}. The upper part (blue blocks) describes the single steps required for distance estimation, while the green blocks describe the single steps needed to predict semantic segmentation. Both tasks are optimized inside a multi-task network by using the weighted total loss described in Eq.~\ref{eq:mtl_loss}. Following Section~\ref{sec:fisheyedistancenet-framework}, we set up the self-supervised monocular structure-from-motion (SfM) framework with extensions to enabling multiple cameras. View synthesis is performed by incorporating the polynomial projection model from Section~\ref{sec:modeling of fisheye geometry}. The same protocols are used to train the distance and pose estimation networks simultaneously. Section~\ref{sec:fisheyedistancenet-final-loss} describes the total self-supervised objective loss. In the following part, we will describe the different loss contributions in the context of fisheye camera images.\par
\subsection{Semantic Segmentation Baseline}

We define semantic segmentation as the task of assigning a pixel-wise label mask $M_{t}$ to an input image $I_t$, i.e. the same input as for distance estimation from a single image. Each pixel gets assigned a class label $s\in\mathcal{S} = \left\lbrace1,2,...,S\right\rbrace$ from the set of classes $\mathcal{S}$. In a supervised way, the network predicts a posterior probability $Y_t$ that a pixel belongs to a class $s\in \mathcal{S}$, which is then compared to the one-hot encoded ground truth labels $\overline{Y}_{t}$ inside the cross-entropy loss
\begin{equation}
\mathcal{L}_{ce} = -\sum_{s \in\mathcal{S}} \overline{Y}_{t,s} \cdot \log\left(Y_{t,s}\right)
\label{eq:crossentropy_loss}
\end{equation}
the final segmentation mask $M_{t}$ is then obtained by applying a pixel-wise $\operatorname{argmax}$ operation on the posterior probabilities $Y_{t,s}$. Note that we also use unrectified fisheye camera images, for which the segmentation task can however still be applied as shown in this work.\par
\subsection{Robust Reconstruction Loss}

Most state-of-the-art self-supervised depth estimation methods use heuristic loss functions. However, the optimal choice of a loss function is not well defined theoretically. In this section, we emphasize the need for the exploration of a better photometric loss function described in Section~\ref{sec:photometric loss} and explore a more generic robust loss function.\par

Towards developing a more robust loss function, we introduce the common notion of a per-pixel regression $\rho$ in the context of distance estimation, which is given by 
\begin{align}
    \rho \left( \xi \right) = \rho \left( \hat I_{t' \to t}-I_t \right)
\end{align}
while this general loss function can be implemented by a simple \lone loss as in the second term of Eq.~\ref{eq:loss-photo}, recently, a general and more robust loss function has been proposed by Barron~\cite{barron2019general}, which we use to replace the \lone term in Eq.~\ref{eq:loss-photo}. This function is a generalization of many common losses such as the \lone, $L_2$, Geman-McClure, Welsch/Leclerc, Cauchy/Lorentzian and Charbonnier loss functions. In this loss, robustness is introduced as a continuous parameter and it can be optimized within the loss function to improve the performance of regression tasks. The general form of the loss function is:
\begin{equation}
    \label{equ:robust_loss}
    f_{\mathrm{rob}}\left(\zeta, \rho, c\right) = \frac{\abs{\rho - 2}}{\rho} \left( \left( {\frac{\left( \sfrac{\zeta}{c} \right)^2}{\abs{\rho - 2}}} + 1 \right)^{\sfrac{\rho}{2}} - 1 \right)
\end{equation}
The free parameters in this loss function can be automatically adapted to any particular problem via data-driven optimization. 
To induce $\rho$ as a trainable parameter Barron~\cite{barron2019general} encapsulates the loss into a probability density function given by:
\begin{align}
     \prob{\zeta \;|\; \mu, \rho, c} &= {\frac {1}{c\partition{\rho}}} \exp \left( -\lossfun{\zeta - \mu, \rho, c} \right) \\
    \partition{\rho} &= \int_{-\infty}^{\infty} \exp \left(-\lossfun{\zeta, \rho, 1} \right)
\end{align}
where $\prob{\zeta \;|\; \mu, \rho, c}$ is only defined if $\rho \geq 0$, as $\partition{\rho}$ is divergent when
$\rho < 0$. Then the optimization function reduces to:
\begin{align}
     \arg \min_{\theta,\rho} -log(p(\zeta|\rho) = \rho \left(\zeta,\rho \right ) + log(Z(\rho))
\end{align}
where $log(Z(\rho))$ is an analytical function which is approximated with a cubic spline function. $\partition{\rho}$ is an important factor in the loss function as it reduces the cost of outliers. The loss of outliers decreases with the reduction of $\rho$. Correspondingly, the loss of inliers will increase. The main properties of the robust loss function are (i) It is monotonic with respect to its inputs $\abs{\zeta}$ and $\rho$ which is useful for graduated non-convexity. (ii) It is smooth respect to its inputs $\zeta$ and $\rho$ (\ie, in $C^\infty$). (iii) It has bounded first and second derivatives (no exploding gradients and easier pre-conditioning).
\begin{figure}[!t]
	\centering
	\subfigure[][Image \label{fig:mask_analysis_a}]{\includegraphics[width=0.48\linewidth]{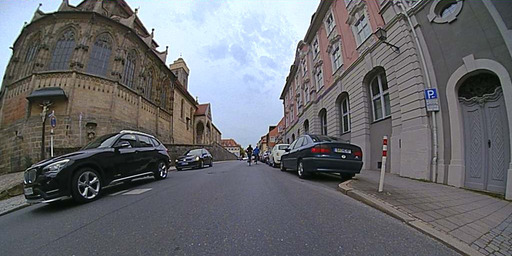}}\;\;
	\subfigure[][Segmentation \label{fig:mask_analysis_b}]{\includegraphics[width=0.48\linewidth]{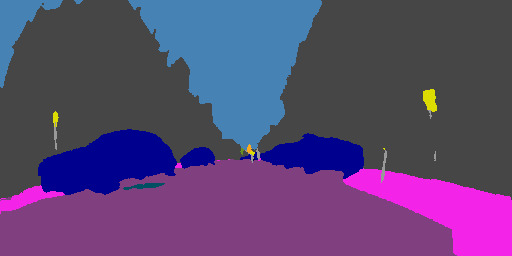}}\\
	\vspace{-3mm}
	\subfigure[][Projected image \label{fig:mask_analysis_c}]{\includegraphics[width=0.48\linewidth]{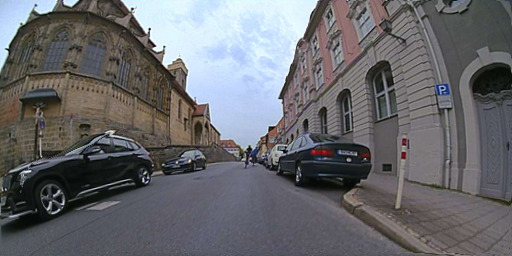}}\;\;
	\subfigure[][Projected segmentation \label{fig:mask_analysis_d}]{\includegraphics[width=0.48\linewidth]{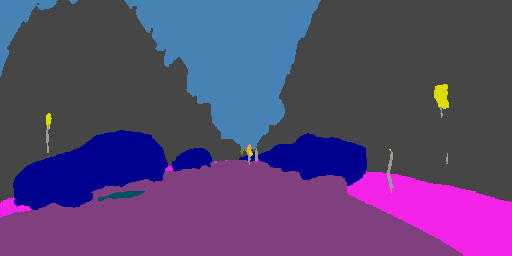}}\\
	\vspace{-3mm}
	\subfigure[][Photometric error \label{fig:mask_analysis_e}]{\includegraphics[width=0.48\linewidth]{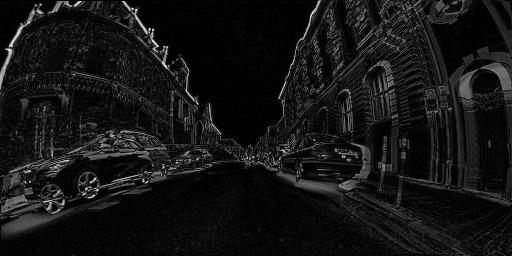}} \;\;
	\subfigure[][Dynamic object mask \label{fig:mask_analysis_f}]{\includegraphics[width=0.48\linewidth]{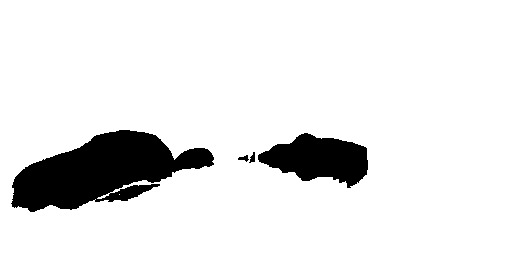}}\\
	\vspace{-3mm}
	\subfigure[][Distance Estimate \label{fig:mask_analysis_g}]{\includegraphics[width=0.48\linewidth]{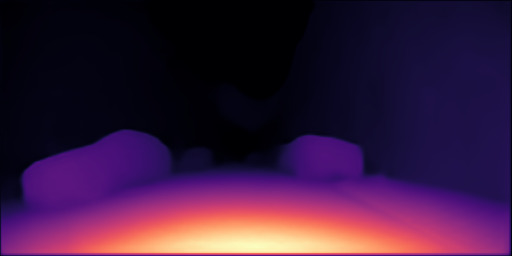}} \;\;
	\subfigure[][ Mask (f) applied on (e) \label{fig:mask_analysis_h}]{\includegraphics[width=0.48\linewidth]{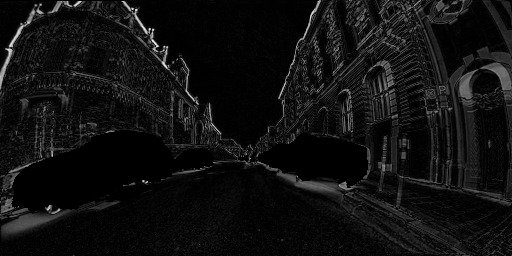}}\\
	\caption[\bf Application of the semantic masking methods.]
	        {\textbf{Application of the semantic masking methods to handle potentially dynamic objects.} The dynamic objects inside the segmentation masks from consecutive frames in (b) and (d) are accumulated to a dynamic object mask, which is used to mask the photometric error (e), as shown in (h).}
	\label{fig:mask_analysis}
\end{figure}
\subsection{Dealing With Dynamic Objects}
\label{sec:dynamic-object-mask}

Typically, the assumed static world model for projections between image frames is violated by dynamic objects' appearance. Thereby, we use the segmentation masks to exclude \textit{moving} potentially dynamic objects while \textit{non-moving} dynamic object should still contribute.\par 
In order to implement this, we aim at defining a pixel-wise mask $\mu_t$, which contains a $0$, if a pixel belongs to a dynamic object from the current frame $I_t$, or a wrongfully projected dynamic object from the reconstructed frames $\hat{I}_{t'\to t}$, and a $1$ otherwise. For calculating the mask, we start by predicting a semantic segmentation mask $M_t$ which corresponds to image $I_t$ and segmentation masks $M_{t'}$ for all images $I_{t'}$. We then use the same projections for the images and warp the segmentation masks (using nearest neighbor instead of bilinear sampling), yielding projected segmentation masks $M_{t' \to t}$. Then, also defining the set of dynamic object classes $\mathcal{S}_{\mathrm{DC}} \subset \mathcal{S}$ we can define $\mu_t$ by its pixel-wise elements at pixel location $uv$:
\begin{equation}
\!\!\mu_{t, uv} = 
\!\left\{
\begin{array}{l}
1 ,\; M_{t, uv} \notin \mathcal{S}_{\mathrm{DC}}\; \land \; M_{t'\rightarrow t,uv} \notin \mathcal{S}_{\mathrm{DC}} \\
0,\; \mathrm{else} \\
\end{array}
\right.
\label{eq:semantic_mask}
\end{equation}
The mask is then applied pixel-wise on the reconstruction loss defined in Eq.~\ref{eq:loss-photo} to mask out dynamic objects. However, as we only want to mask out \textit{moving} DC-objects, we detect them using the consistency of the target segmentation mask and the projected segmentation mask to judge whether dynamic objects are moving between consecutive frames (\eg, we intend to learn the distance of dynamic objects from parking cars, but not from driving ones). With this measure, we apply the dynamic object mask $\mu_t$ only to an imposed fraction $\epsilon$ of images, in which the objects are detected as mostly moving.\par
\subsection{Joint Optimization}
\label{sec:joint-optimization}

We incorporate the task weighting approach by Kendall \etal~\cite{Kendall2018}; we weigh the distance estimation and semantic segmentation loss terms for multi-task learning, which enforces homoscedastic (task) uncertainty. It is proven to be effective in weighing the losses from Eq.~\ref{equ:objective} and Eq.~\ref{eq:crossentropy_loss}:
\begin{align}
    \label{eq:mtl_loss}
    \frac{1}{2 \sigma_1^2} \mathcal{L}_{tot} + \frac{1}{2 \sigma_2^2} \mathcal{L}_{ce} + \log (1 + \sigma_1) + \log (1+ \sigma_2)
\end{align}
Homoscedastic uncertainty does not change with varying input data and is task-specific. We, therefore, learn this uncertainty and use it to downweigh each task. Increasing the noise parameter $\sigma_i$ reduces the weight for the respective task. Furthermore, $\sigma$ is a learnable parameter; the objective optimizes a more substantial uncertainty that should lead to a smaller contribution of the task's loss to the total loss. In this case, the different scales from the distance and semantic segmentation are weighed accordingly. The noise parameter $\sigma_1$ tied to distance estimation is relatively low compared to $\sigma_2$ of semantic segmentation, and the convergence occurs accordingly. Higher homoscedastic uncertainty leads to a lower impact on the task's network weight update. It is important to note that this technique is not limited to the joint learning of distance estimation and semantic segmentation but can also be applied to more tasks and arbitrary camera geometries.\par
\subsection{Post-Processing Technique}
\label{subsec:postprocessing}

\begin{figure}[t!]
  \centering
  \resizebox{\textwidth}{!}{
  \centering
\begin{tabular}{@{\hskip 0.5mm}c@{\hskip 0.5mm}c@{\hskip 0.5mm}c@{}}

\includegraphics[height=\textwidth]{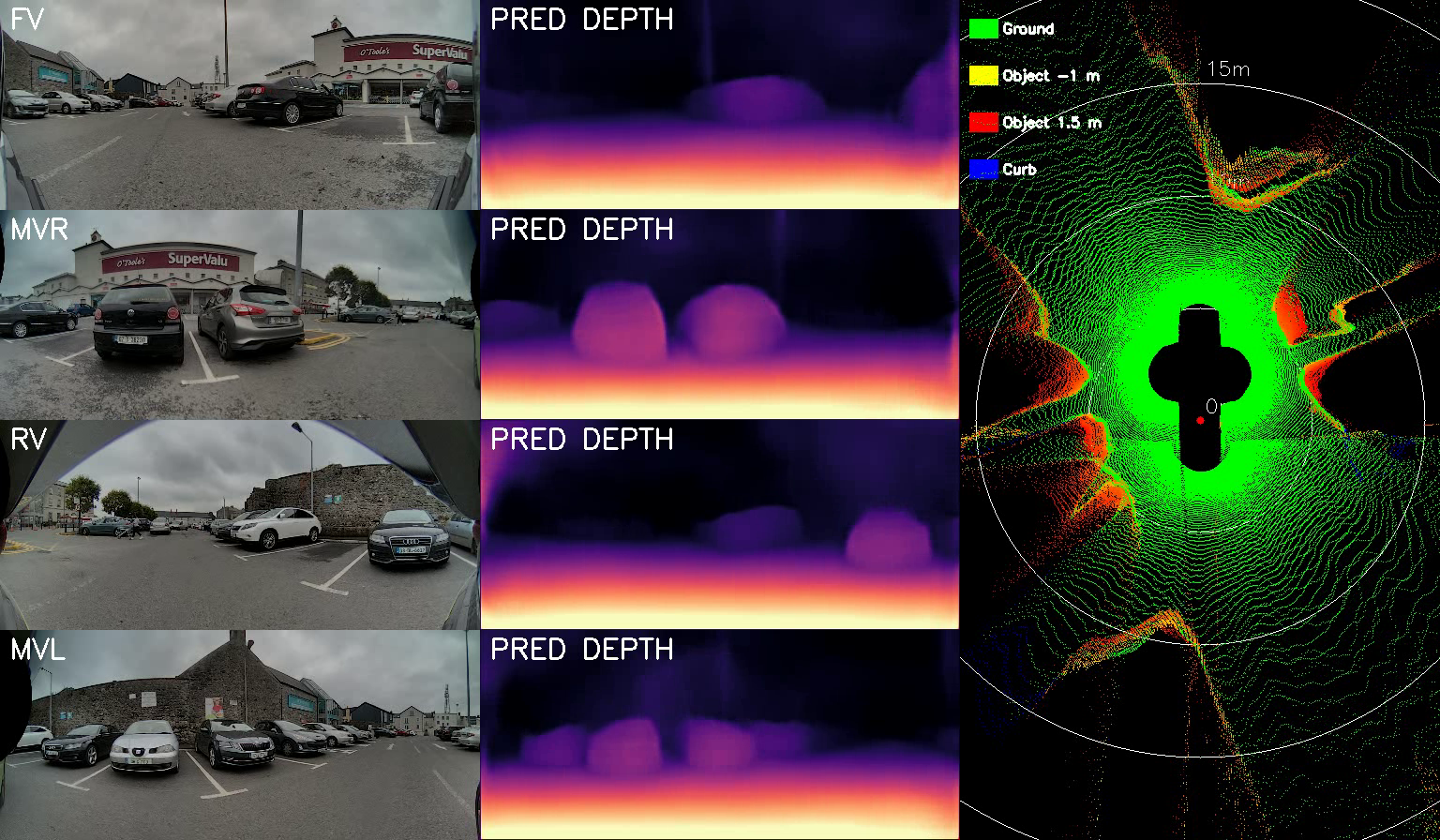} \\
\includegraphics[height=\textwidth]{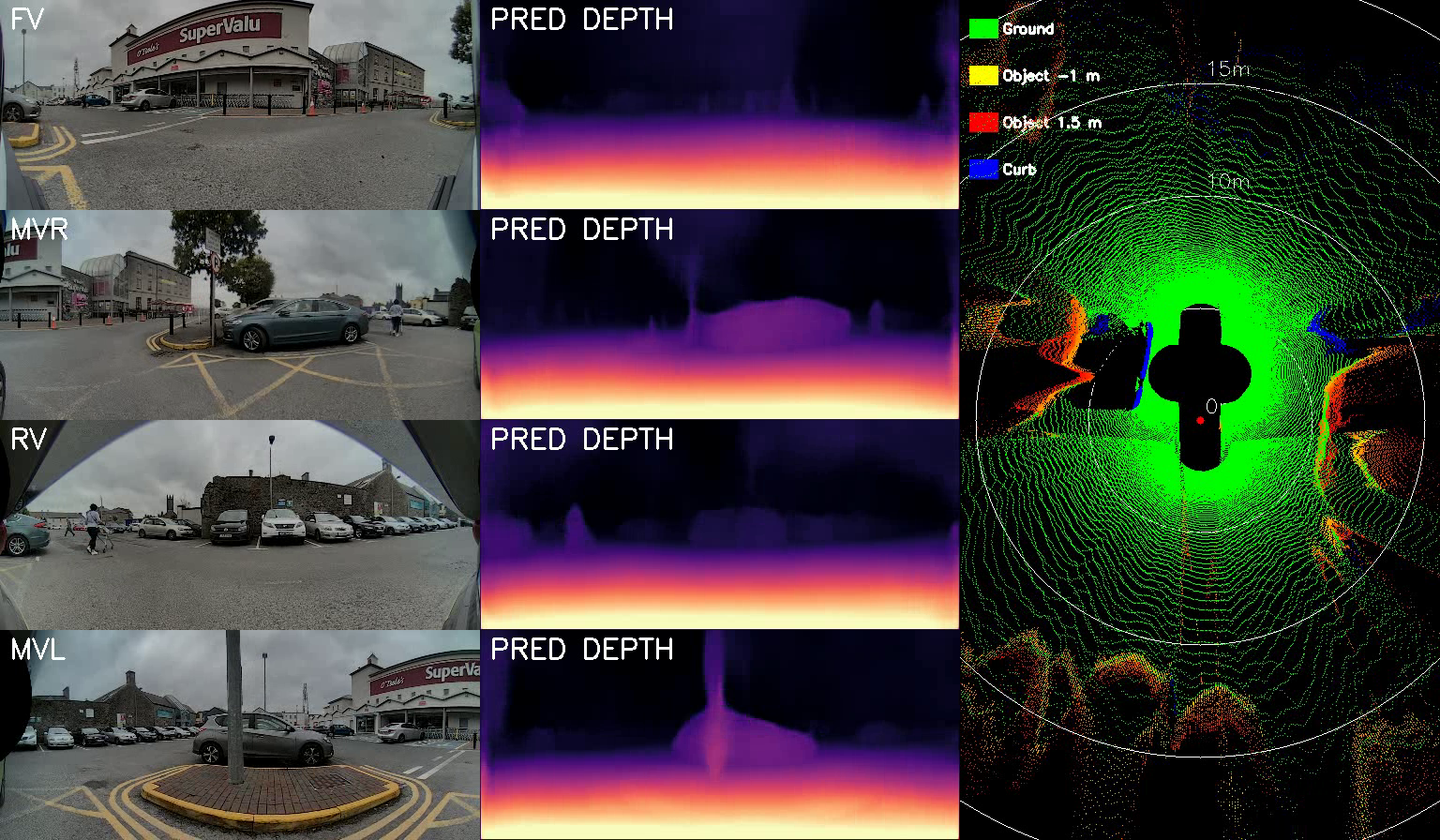}
\end{tabular}
}
   \caption[\bf Qualitative results of $360\degree$ bird's eye view distance output on an unseen sequence.]
           {\textbf{Qualitative results of $360\degree$ bird's eye view distance output on an unseen sequence}. Two snapshots from a sequence illustrating color-coded heights and its corresponding surround-view images are shown. The distance maps are converted to color-coded height maps (\ie, \textcolor[HTML]{00b050}{green} is ground surface, \textcolor[HTML]{FFC300}{yellow} is an object at $1\,m$, \textcolor[HTML]{FF0000}{red} is an object at a distance of $1.5\,m$ and \textcolor[HTML]{0000FF}{blue} indicates curb). Spatial and temporal smoothing operators were applied as discussed in subsection~\ref{subsec:postprocessing}.}
  \label{fig:post_processing_v1}
\end{figure}

\begin{figure}[t!]
  \centering
  \resizebox{\textwidth}{!}{
  \centering
\begin{tabular}{@{\hskip 0.5mm}c@{\hskip 0.5mm}c@{\hskip 0.5mm}c@{}}

\includegraphics[height=\textwidth]{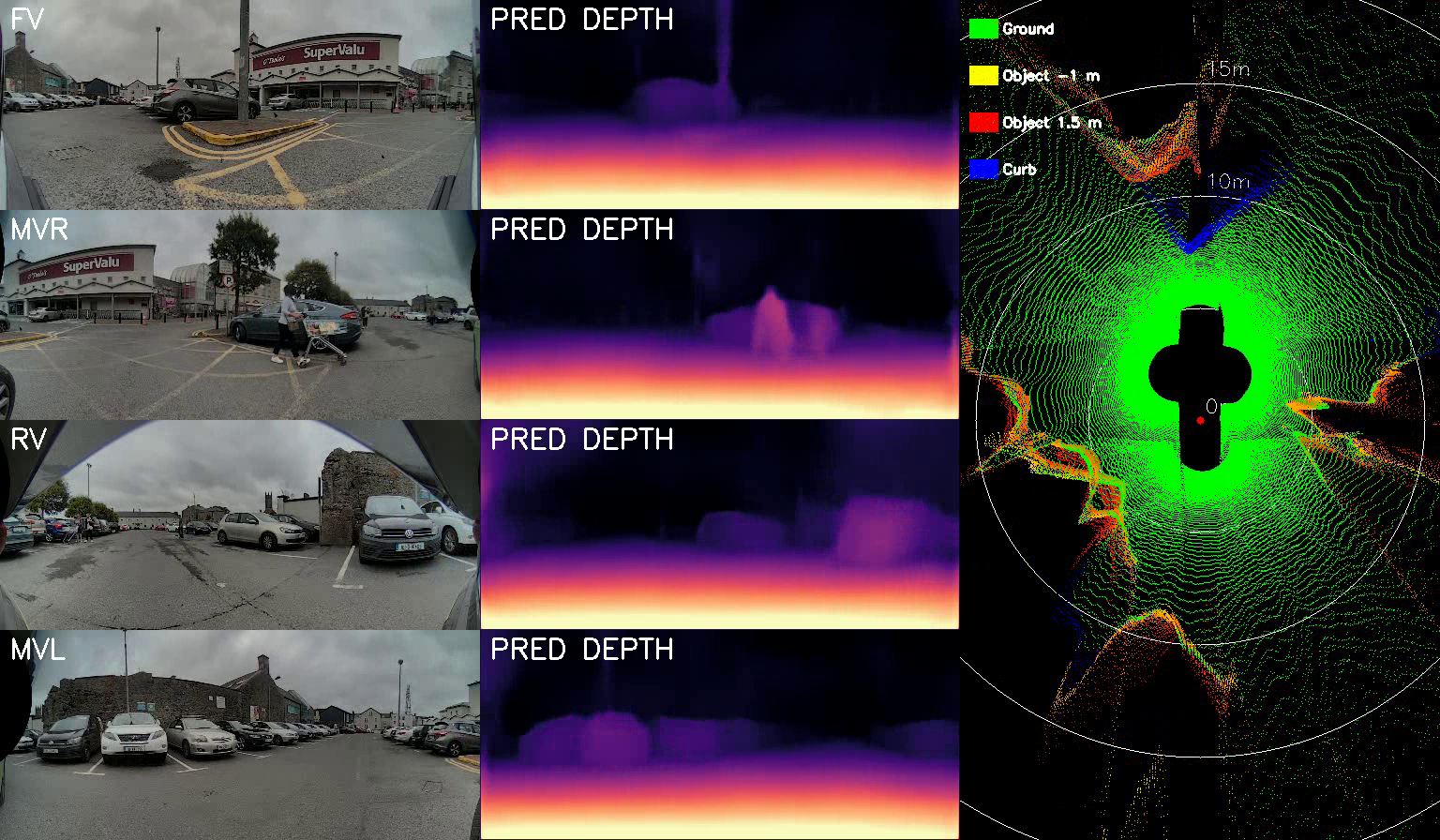} \\
\includegraphics[height=\textwidth]{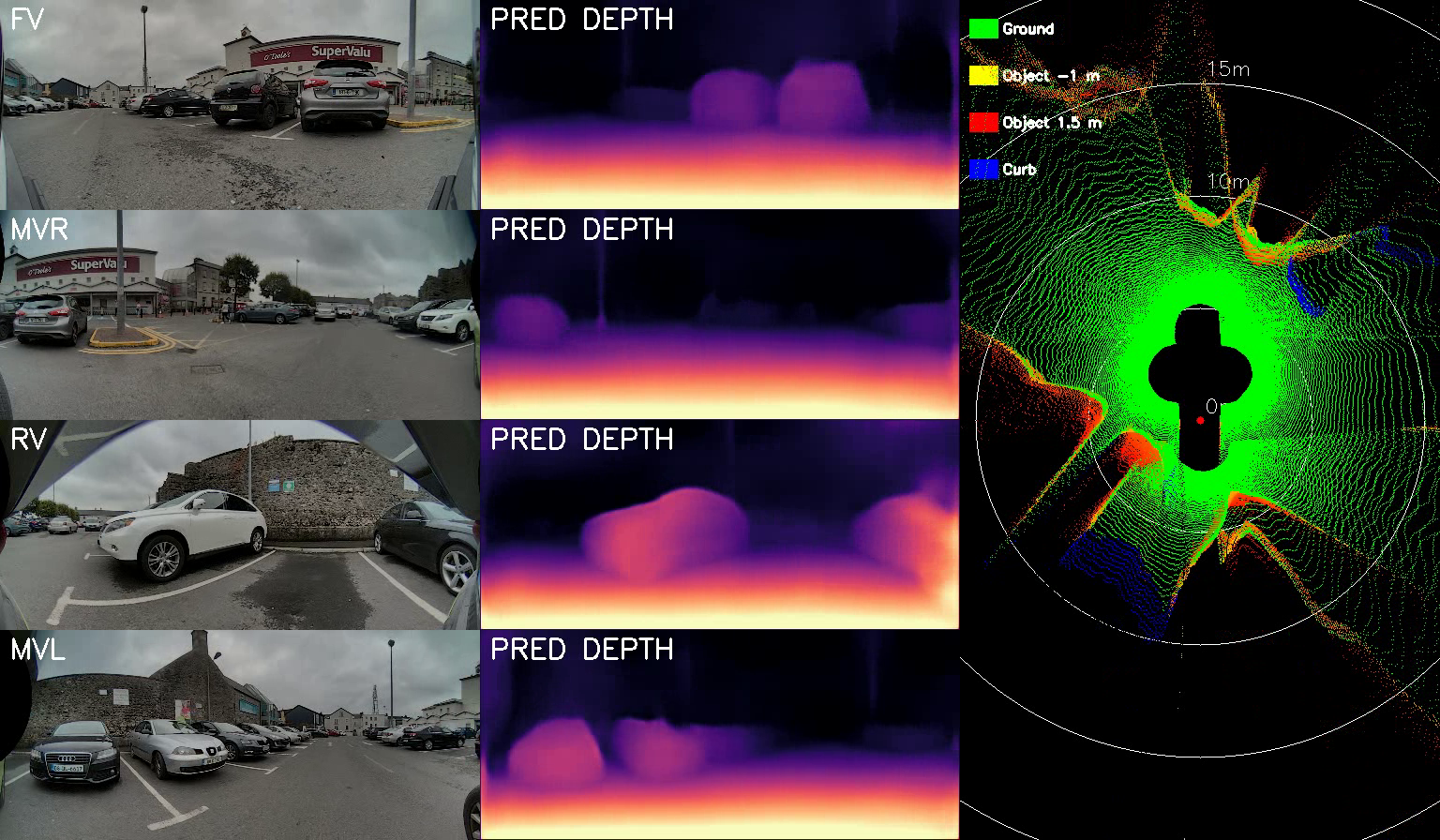}
\end{tabular}
}
   \caption[\bf Additional qualitative results of $360\degree$ bird's eye view distance output on an unseen sequence.]
           {\textbf{Additional qualitative results of $360\degree$ bird's eye view distance output on an unseen sequence}. Two snapshots from a sequence illustrating color-coded heights and its corresponding surround-view images are shown. The distance maps are converted to color-coded height maps (\ie, \textcolor[HTML]{00b050}{green} is ground surface, \textcolor[HTML]{FFC300}{yellow} is an object at $1\,m$, \textcolor[HTML]{FF0000}{red} is an object at a distance of $1.5\,m$ and \textcolor[HTML]{0000FF}{blue} indicates curb). Spatial and temporal smoothing operators were applied as discussed in subsection~\ref{subsec:postprocessing}.}
  \label{fig:post_processing_v2}
\end{figure}

\begin{figure*}[t!]
  \captionsetup{skip=2pt, belowskip=-8pt}
  \centering
  \resizebox{0.91\textwidth}{!}{\centering
\begin{tabular}{@{\hskip 0.5mm}c@{\hskip 0.5mm}c@{\hskip 0.5mm}c@{}}
\includegraphics[height=\textwidth]{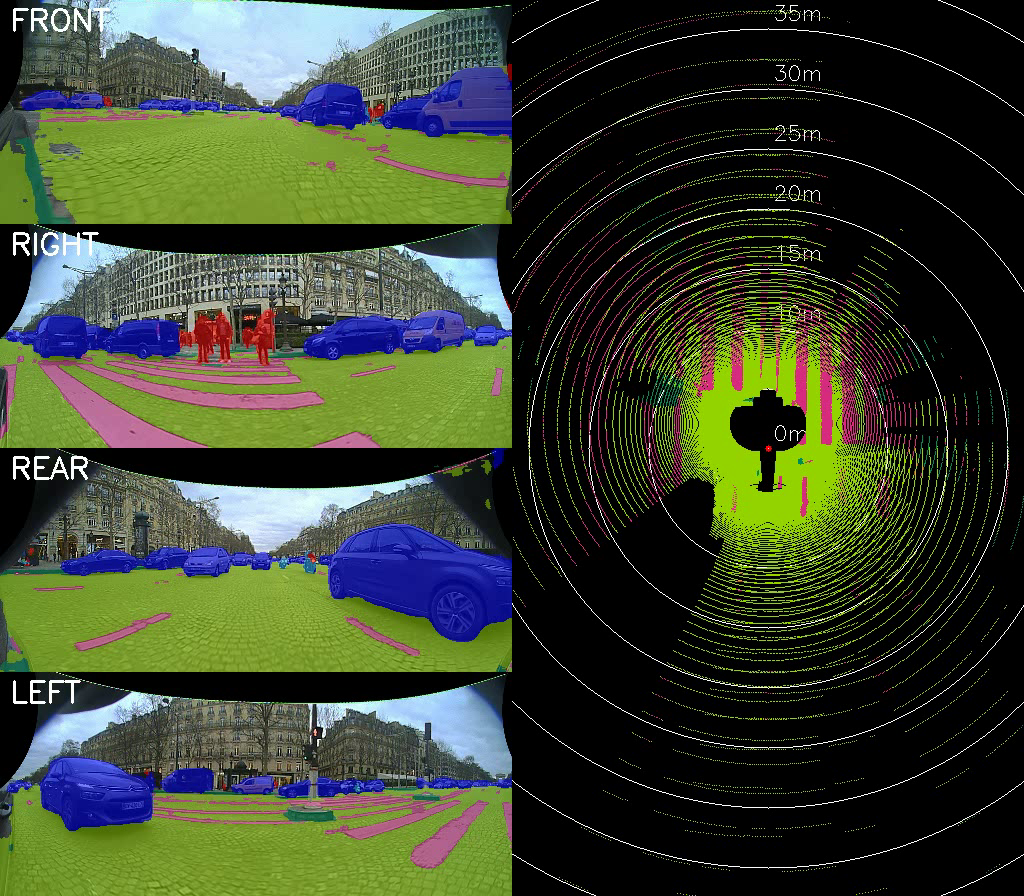} \\
\includegraphics[height=\textwidth]{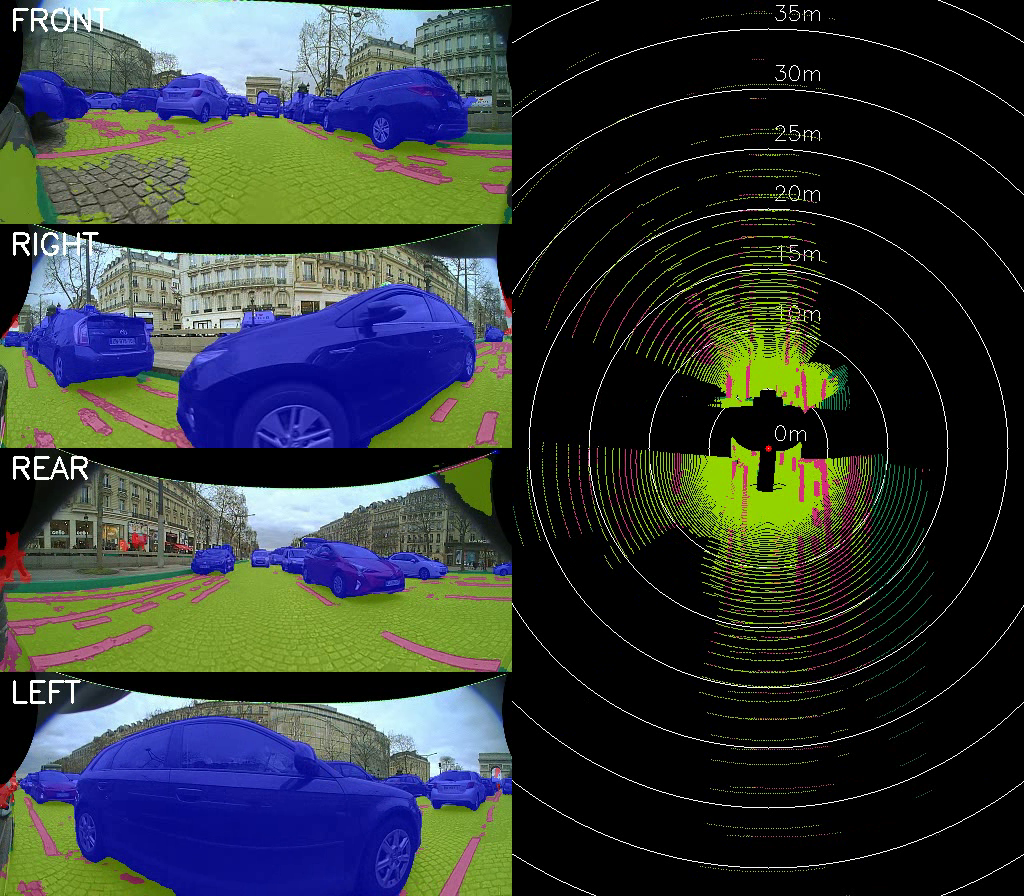}
\end{tabular}
}
  \caption[\bf Qualitative semantic results segmentation of $360\degree$ post-processed top-view output on an unseen sequence.]
          {\textbf{Qualitative semantic segmentation results of $360\degree$ post-processed top-view output on an unseen sequence}. The semantic maps are converted to color-coded height maps (\ie, \textcolor[HTML]{00b050}{green} is the road surface, \textcolor[HTML]{D72B81}{pink} are the lane markings, \textcolor[HTML]{297150}{dark green} are the curbs). The top-view maps aid to detect free space for the autonomous vehicle to navigate.}
  \label{fig:post_processing_semantic_v1}
\end{figure*}

\begin{figure*}[t!]
  \captionsetup{skip=2pt, belowskip=-8pt}
  \centering
  \resizebox{0.91\textwidth}{!}{\centering
\begin{tabular}{@{\hskip 0.5mm}c@{\hskip 0.5mm}c@{\hskip 0.5mm}c@{}}

\includegraphics[height=\textwidth]{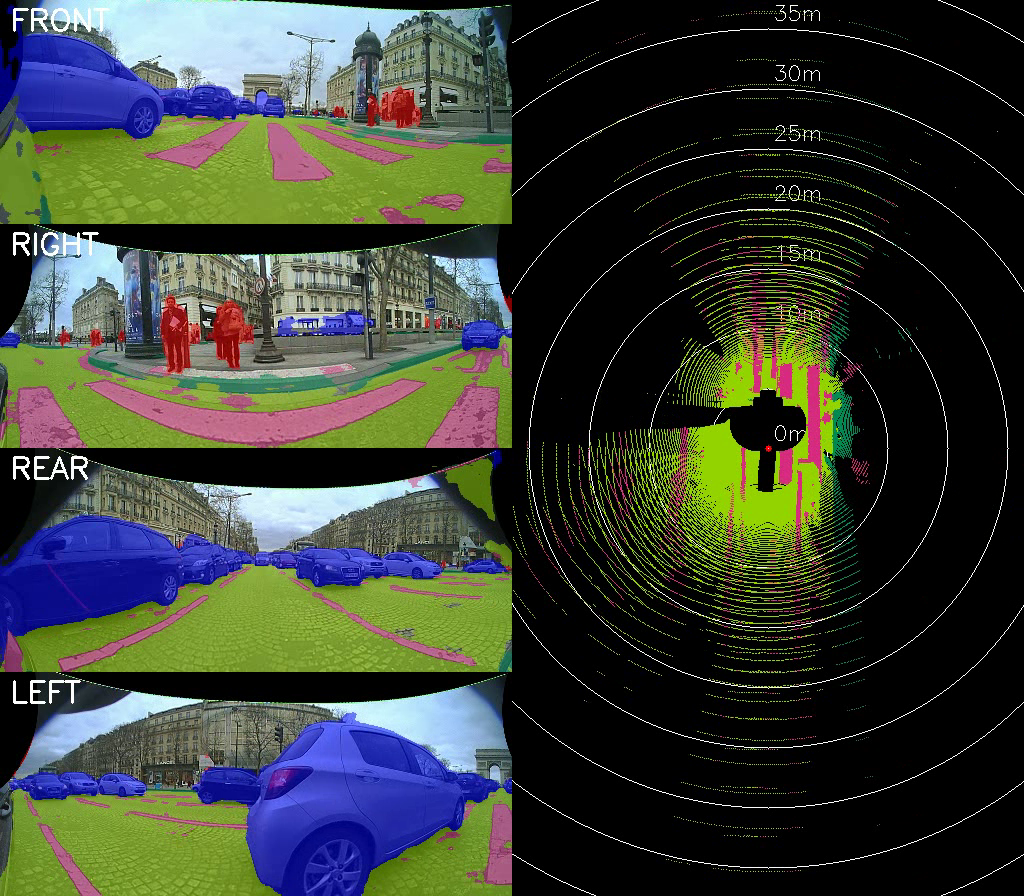} \\
\includegraphics[height=\textwidth]{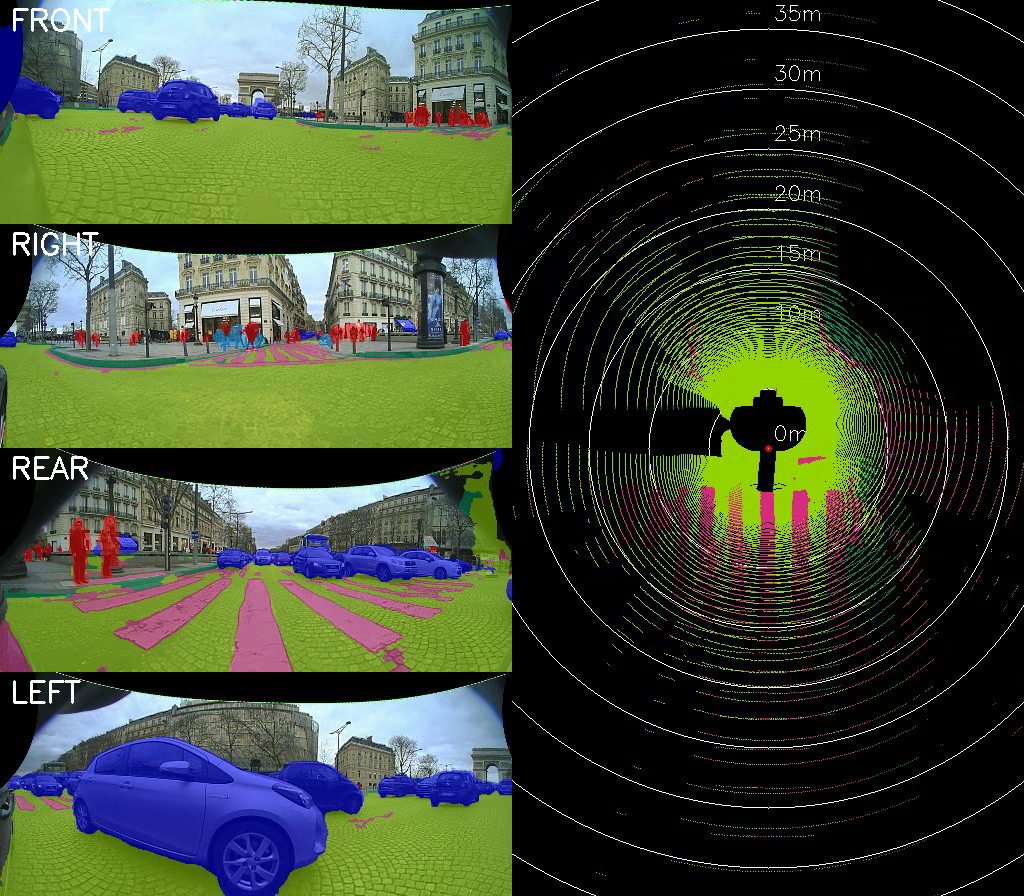}
\end{tabular}
}
  \caption[\bf Additional qualitative semantic segmentation results of $360\degree$ post-processed top-view outputs on an unseen sequence.]
          {\textbf{Additional qualitative semantic segmentation results of $360\degree$ post-processed top-view outputs on an unseen sequence}. The semantic maps are converted to color-coded height maps (\ie, \textcolor[HTML]{00b050}{green} is road surface, \textcolor[HTML]{D72B81}{pink} is the lane marking, \textcolor[HTML]{297150}{dark green} is the curb). The top-view maps aid to detect free space for the autonomous vehicle to navigate.}
  \label{fig:post_processing_semantic_v2}
\end{figure*}

This subsection provides a brief overview of how the distance maps get post-processed and converted to a representation directly used by motion planning. The prime purpose of surround-view cameras is to aid 360$\degree$ near-field sensing around the car. Low-speed maneuvering such as parking requires very high accuracy in the order of $5\,cm$ and the ability to detect small objects like curbs or potholes. We construct a 2D top-view heightmap grid at a $5\,cm$ and $10\,m$ range resolution by targeting these requirements. We project the distance and semantic segmentation maps from each image onto the top view and fill the height map cells with the height information. Due to a small overlap in the image's corners, we require a fusion scheme to combine the distance values. The spatial consistency of the scene across cameras can be exploited. We use a spatial smoothing filter to smoothen the current observation, further filtered using a temporal smoothing filter. The top-view post-processed distance and semantic maps are utilized for our \textbf{Level 3} planning module in the final step. The filtered height maps are illustrated in Figure~\ref{fig:svdistnet-mtl_pipeline}. Additional qualitative results are illustrated in Figures~\ref{fig:post_processing_v1}, \ref{fig:post_processing_v2}, \ref{fig:post_processing_semantic_v1} and \ref{fig:post_processing_semantic_v2}.\par
\section{Network Architecture}
\label{sec:network_details}

\begin{figure}[!t]
  \centering
    \includegraphics[width=\columnwidth]{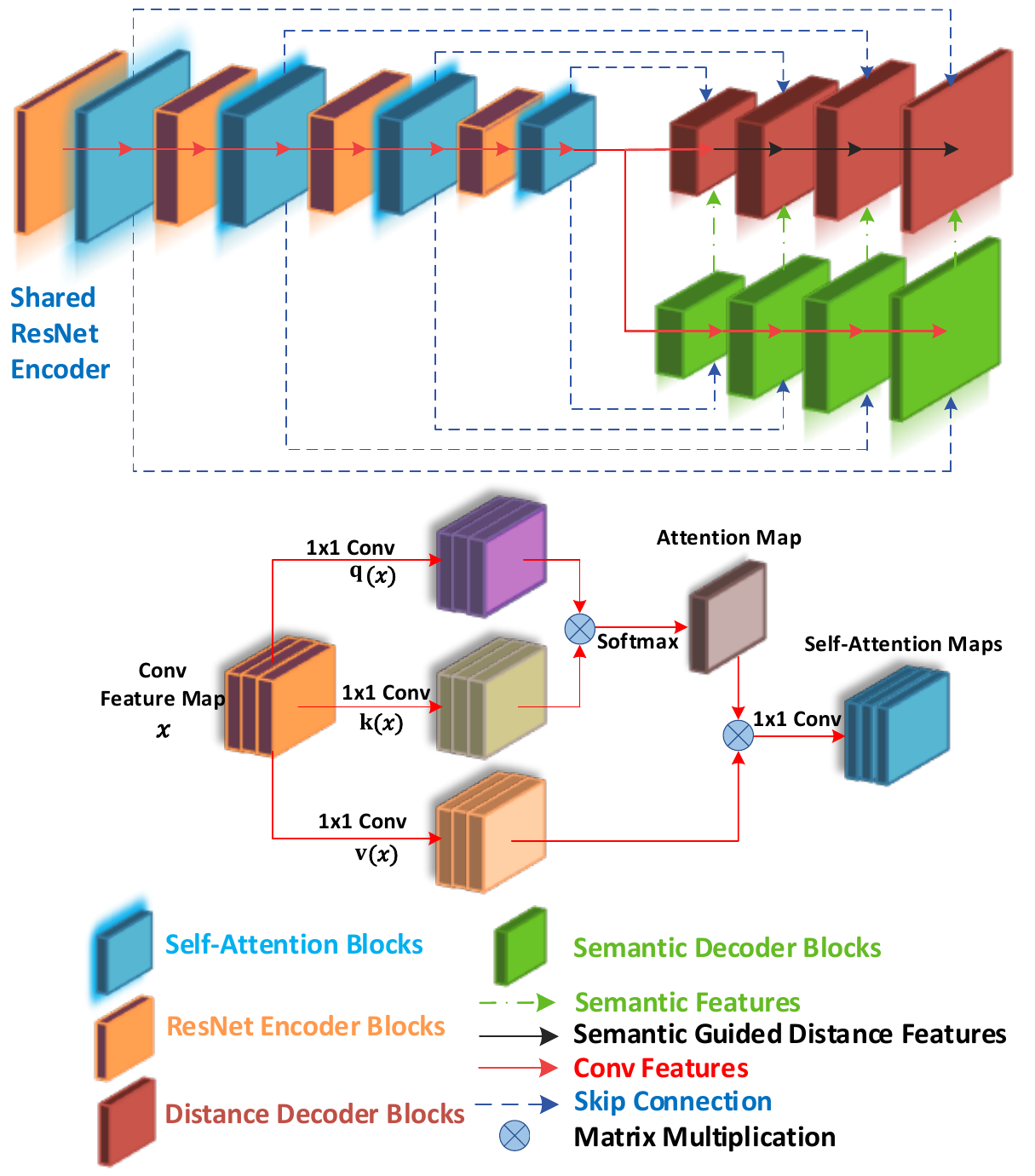}
    \caption{\textbf{Visualization of the SynDistNet network architecture.}}
    \label{fig:syndistnet_model_arch}
\end{figure}

This section explains the novel architecture for semantically guided self-supervised distance estimation utilizing Camera Geometry Tensor (CGT) to handle multiple viewpoints and changes in the camera's intrinsic. The baseline from~\cite{kumar2020fisheyedistancenet} used deformable convolutions to model the fisheye geometry to incorporate the distortion and improve the distance estimation accuracy. At first, we introduce a scalar-based self-attention encoder to obtain locally-attentive maps and a semantically guided decoder for the distance estimation using pixel-adaptive convolutions for the \textit{SynDistNet} network architecture as shown in Figure~\ref{fig:syndistnet_model_arch}, which can be trained in a one-stage fashion. Later, we incorporate improved vector-based self-attention modules from~\cite{zhao2020exploring} shown in Figure~\ref{fig:svdistnet-model_arch}. The figure provides an overview of the proposed network architecture used in the \textit{SVDistNet} framework. The encoder is a self-attention network with pairwise and patchwise variants as described in Section~\ref{sec:vector-self-attention}. At the same time, the decoder uses pixel-adaptive convolutions, which are complemented by the novel Camera Geometry Tensors. These networks efficiently adapt the weights across both spatial dimensions and channels. The complete training of both tasks is performed in a one-stage manner.\par
\subsection{Self-Attention Encoder}

\subsubsection{Scalar based Self-Attention Encoder}

Previous depth estimation networks~\cite{godard2019digging, zhou2017unsupervised} use normal convolutions for capturing local information in an image, but the convolutions' receptive field is comparably small. One of the major drawbacks is that convolution lacks rotation invariance. The kernel $K$'s footprint increases the number of parameters to be learned, and due to the filter's fixed nature, aggregation of neighborhood information can not adapt to its contents. Hu \etal~\cite{hu2019local} and Ramachandran \etal~\cite{ramachandran2019stand} perceived that self-attention could be a feasible choice for developing models for image perception rather than merely enhancing discrete convolutional operation layers. The authors present a self-attention layer which may replace convolution while reducing the number of parameters. Similar to a convolution, given a pixel $x_{uv} \in \mathbb{R}^{d_{in}}$ inside a feature map, the local region of pixels defined by positions ${a b} \in \mathcal{N}_k(uv)$ with spatial extent $k$ centered around $x_{uv}$ are extracted initially which is referred to as a memory block. For every memory block, the single-headed attention for computing the pixel output $z_{uv} \in \mathbb{R}^{d_{out}}$ is then calculated:
\begin{align}
\label{eq:attention}
z_{uv} = \sum_{\mathclap{ab \in \, \mathcal{N}_k(uv)}}
         \texttt{softmax}_{a b}\left(q_{i j}^\top k_{a b} \right) v_{a b}
\end{align}  
where $q_{uv} = W_Q x_{uv}$ are the \emph{queries}, \emph{keys} $k_{ab} = W_K x_{ab}$, and \emph{values} $v_{ab} = W_V x_{ab}$ are linear transformations of the pixel in position $uv$ and the neighborhood pixels. The learned transformations are denoted by the matrices W. $\texttt{softmax}_{a b}$ defines a softmax applied to all logits computed in the neighborhood of $uv$. $W_Q, W_K, W_V \in \mathbb{R}^{d_{out} \times d_{in}}$ are trainable transformation weights. There exists an issue in the above-discussed approach, as there is no positional information encoded in the attention block. Thus the Eq.~\ref{eq:attention} is invariant to permutations of the individual pixels. For perception tasks, it is typically helpful to consider spatial information in the pixel domain. For example, the detection of a pedestrian is composed of spotting faces and legs in a proper relative localization. The main advantage of using self-attention layers in the encoder illustrated in Figure~\ref{fig:syndistnet_model_arch} is that it induces a synergy between geometric and semantic features for distance estimation and semantic segmentation tasks. In~\cite{vaswani2017attention} sinusoidal embeddings are used to produce the absolute positional information. Following~\cite{ramachandran2019stand}, instead of attention with 2D relative position embeddings, we incorporate relative attention due to their better accuracy for computer vision tasks. The relative distances of the position $uv$ to every neighborhood pixel $(a,b)$ is calculated to obtain the relative embeddings. The calculated distances are split up into row and column distances $r_{a-i}$ and $r_{b-j}$ and the embeddings are concatenated to form $r_{a-i,b-j}$ and multiplied by the query $q_{uv}$:
\begin{equation} \label{equation:standard-self-attention}
z_{uv} = \sum_{\mathclap{a b \in \, \mathcal{N}_k(i j)}}
         \texttt{softmax}_{a b}\left( q_{i j}^\top  k_{a b} + q_{i j}^\top r_{a-i,b-j} \right) v_{a b}
\end{equation}
It ensures the weights calculated by the softmax function are modulated by both the key's relative distance and content from the query. Instead of focusing on the whole feature map, the attention layer only focuses on the memory block, which can be seen in the bottom part of Figure~\ref{fig:syndistnet_model_arch}.\par
\begin{figure}[!t]
  \centering
    \includegraphics[width=\columnwidth]{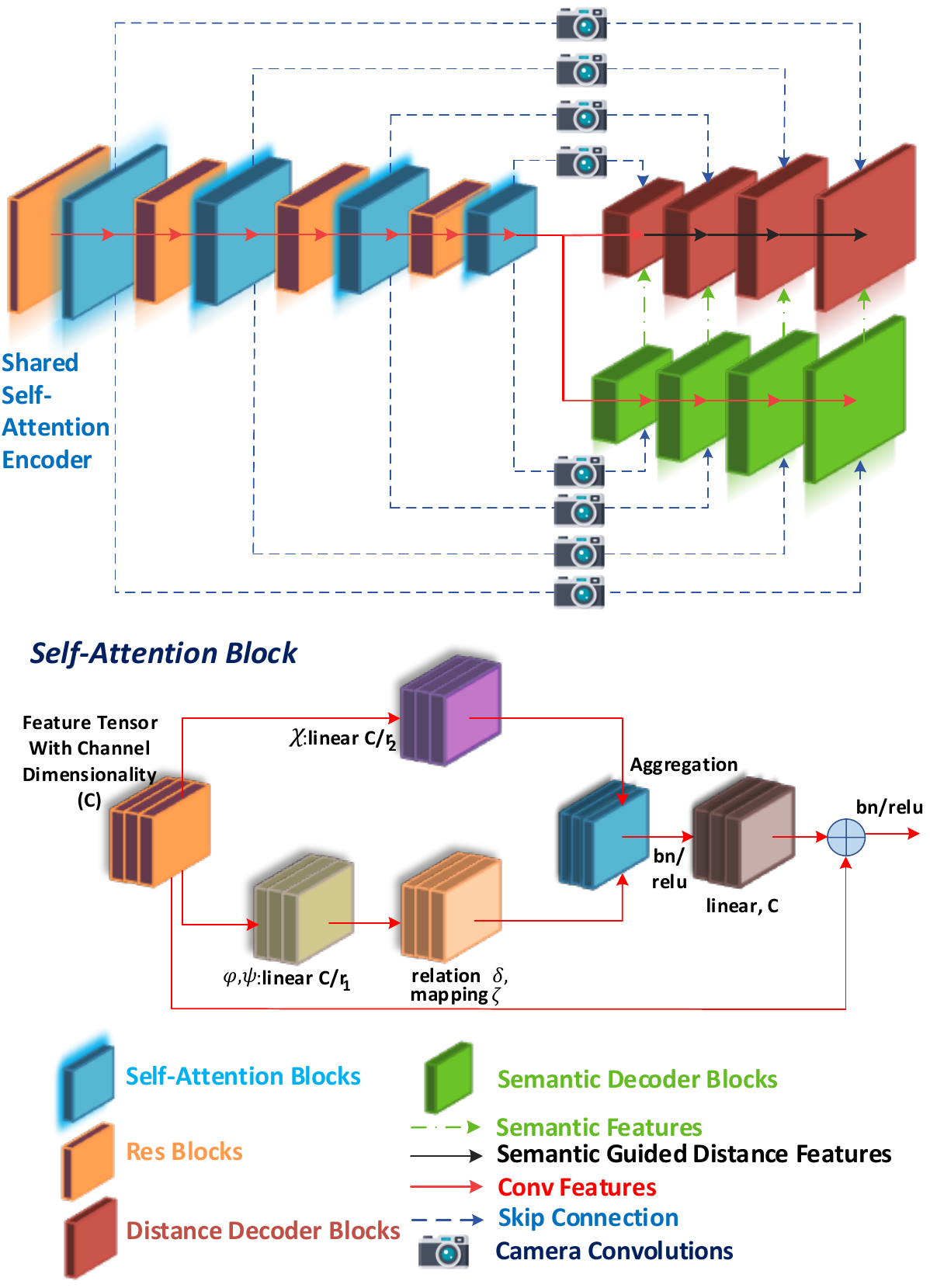}
    \caption{\textbf{Visualization of the SVDistNet network architecture.}}
    \label{fig:svdistnet-model_arch}
\end{figure}
\subsubsection{Vector based Self-Attention Encoders}
\label{sec:vector-self-attention}

Inspired by~\cite{zhao2020exploring}, we incorporate the self-attention network (SAN) backbone to the encoder and compare it with the standard ResNet18~\cite{he2016deep} and ResNet50~\cite{he2016deep} choice of encoder networks. Compared to the scalar attention self-attention layers explained earlier, which are only content-adaptive and not channel-adaptive. The vector attention work of Zhao~\cite{zhao2020exploring} has a smaller footprint than the ResNet family of standard encoder heads, and the vector attention is both content and channel adaptive. The authors showcase two convolution variants, namely \emph{pairwise} and \emph{patchwise}, which may replace convolution while reducing the number of parameters and giving advantages in terms of robustness and generalization.\par
\subsubsection{Pairwise Self-Attention} 

The \emph{pairwise} self-attention module is given by:
\begin{equation}
	\label{eq:pairwise}
	z_{uv} = \sum_{ab \in \mathcal{N}_r(uv)} \eta(x_{uv}, x_{ab}) \odot \chi(x_{ab})
\end{equation}
The location of the spatial index in the feature vector $x_{uv}$ is denoted by $uv$, $\odot$ is the Hadamard product, and the aggregation of the local footprint is $\mathcal{N}_r(uv)$.
The new feature $z_{uv} \in \mathbb{R}^{d_{out}}$ is constructed by aggregating the feature vectors specified by the set of indices by the footprint $\mathcal{N}_r(uv)$. The adaptive weight vectors $\eta(x_{uv}, x_{ab})$ aggregate the feature vectors $\chi(x_{ab})$ produced by the function $\chi$. The weights $\eta(x_{uv}, x_{ab})$ required to combine the transformed features $\chi(x_{ab})$ are computed by the function $\eta$. $\eta$ is decomposed to elucidate the different forms of self-attention and is given by:
\begin{equation}
	\label{eq:pairwise-decomposition}
	\eta(x_{uv}, x_{ab}) = \zeta(\delta(x_{uv},x_{ab}))
\end{equation}
The features $x_{uv}$ and $x_{ab}$ are expressed by a single vector outputted by the relation function $\delta$. This vector is used along with the function $\zeta$ to map a vector that can be combined with $\chi(x_{ab})$ as shown in Eq.~\ref{eq:pairwise}. The $\zeta$ function allows us to explore $\delta$ relationships that generate vectors of varying dimensionality that do not need to match the $\chi(x_{ab})$ dimensionality. In this work, we choose the relation function $\delta$ to be described by the \textit{Hadamard product} form from~\cite{zhao2020exploring}:
\begin{equation}
    \delta(x_{uv}, x_{ab}) = \varphi(x_{uv})\odot\psi(x_{ab})
\end{equation}
where $\varphi$ and $\psi$ are transformations that can be trained, which suit the dimensionality of the output. The dimensionality of $\delta(x_{uv},x_{ab})$ is the same as that of the transformation functions with the Hadamard product.\par
\subsubsection{Patchwise Self-Attention}

The \emph{patchwise} self-attention module is given by:
\begin{equation}
	\label{eq:patchwise}
	z_{uv} = \sum_{ab \in \mathcal{N}_r(uv)} \eta(x_{\mathcal{N}_r(uv)})_{ab} \odot \chi(x_{ab})
\end{equation}
In the footprint ${\mathcal{N}_r(uv)}$, the patch of feature vectors given by $x_{\mathcal{N}_r(uv)}$.
$\eta(x_{\mathcal{N}_r(uv)})$ and the patch $x_{\mathcal{N}_r(uv)}$ have the same spatial dimensionality. $\eta(x_{\mathcal{N}_r(uv)})_{ab}$ is a vector located at $ab$ in that tensor, corresponding to the $x_{ab}$ vector in $x_{\mathcal{N}_r(uv)}$ spatially. 
The patchwise self-attention concerning the features $x_{ab}$ is no longer a set operation compared to its pairwise counterpart. It is not permutation-invariant or cardinality-invariant: the $\eta(x_{\mathcal{N}_r(uv)})$ weight calculation will index the $x_{\mathcal{N}_r(uv)}$ feature vectors separately, by position, and can intermix information from feature vectors from various locations within the footprint. Therefore patchwise self-attention is especially more effective than standard convolution. $\eta(x_{\mathcal{N}_r(uv)})$ is decomposed by:
\begin{equation}
	\label{eq:patchwise-decomposition}
	\eta(x_{\mathcal{N}_r(uv)}) = \zeta(\delta(x_{\mathcal{N}_r(uv)})).
\end{equation}
The feature vector created by $\delta(x_{\mathcal{N}_r(uv)})$ is mapped by the function $\zeta$ to a tensor with the suitable dimensionality and it is comprised of weight vectors for all locations $ab$. Feature vectors $x_{ab}$ from the patch $x_{\mathcal{N}_r(uv)}$ are combined by the function $\delta$. We specifically incorporate the \textit{Concatenation} form from~\cite{zhao2020exploring} for the relation function $\delta$.
\begin{equation}
    \delta(x_{\mathcal{N}_r(uv)}) = [\varphi(x_{uv}), [\psi(x_{ab})]_{\forall ab \in \mathcal{N}_r(uv)}]
\end{equation}
The pairwise and patchwise self-attention operations can be used to build residual blocks~\cite{he2016deep} for the encoder that performs both feature aggregation and transformation.\par
\begin{figure}[!t]
  \centering
    \includegraphics[width=0.8\textwidth]{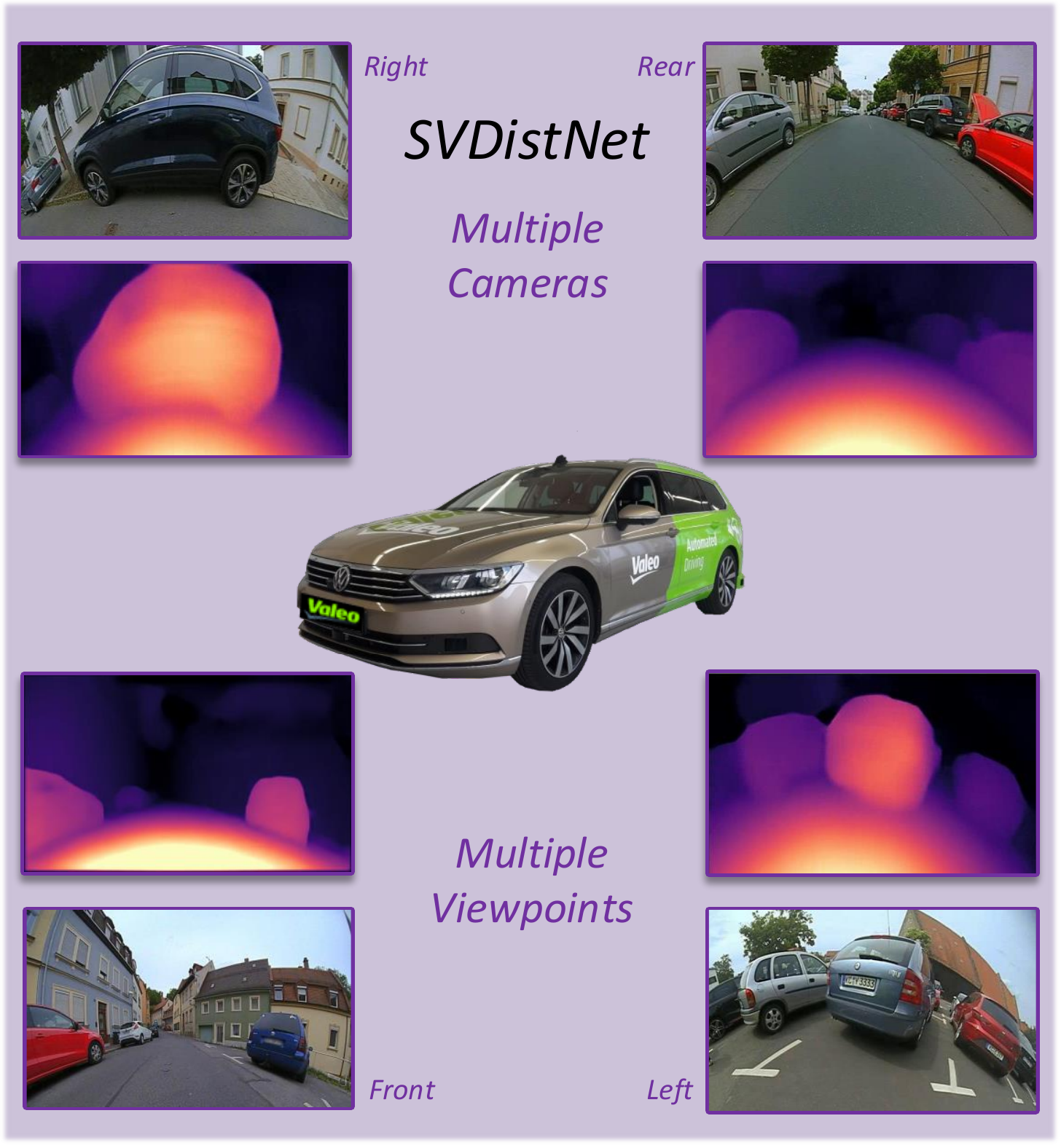} 
    \caption{\bf Illustration of distance estimation on multiple cameras and multiple viewpoints constituting SVDistNet.}
    \label{fig:svdistnet-general_concept}
\end{figure}
\subsection{Camera Geometry Tensor}
\label{sec:camera-geometry-tensor}

\subsubsection{Motivation}

We target the design of a model that can be deployed in millions of vehicles having its own set of cameras. Although the underlying camera intrinsics model is the same for a particular family of vehicles, there are variations due to manufacturing processes, which require the calibration of each camera instance. Even after deployment, calibration can vary due to high environmental temperature or due to aging. Thus a calibration adaptation mechanism in the model is essential.\par

This contrasts with public datasets, which have a single camera instance for both the training and test dataset. In the Woodscape dataset, there are 12 different cameras with slight intrinsic variations to evaluate this effect. There are four camera instances around the vehicle with different intrinsics, even for a single instance of a surround-view system. A single model for these four cameras instead of 4 individual models would also have several practical advantages such as:
\begin{itemize}[nosep]
    \item An improved efficiency on the embedded system requiring less memory and data rate to transmit.
    \item An improved training by access to a larger dataset and regularization through different views.
    \item Maintenance and certification of a single model instead of four.
\end{itemize}
\begin{figure*}[t]
  \centering
    \includegraphics[width=\textwidth]{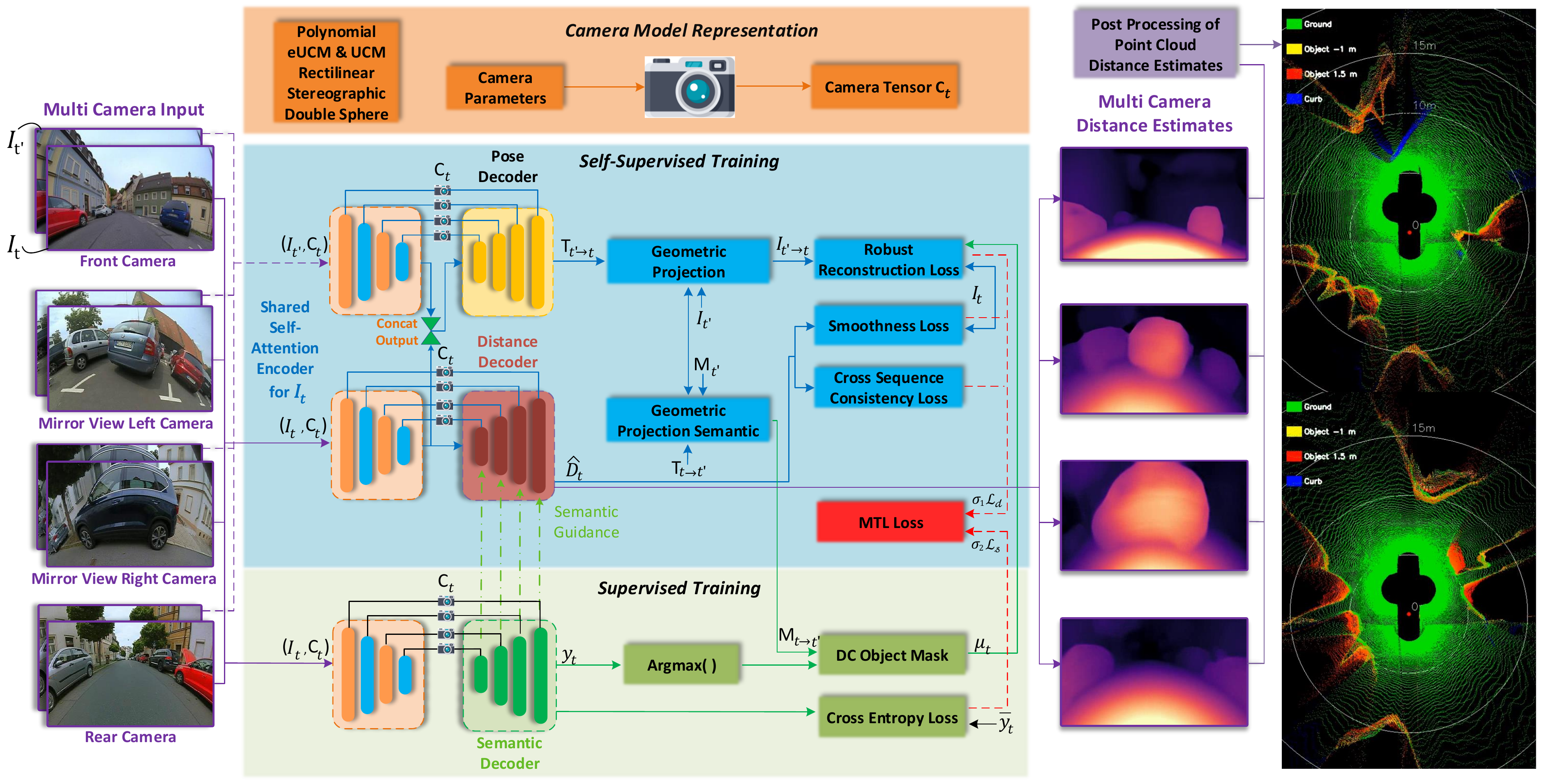}
    \caption[\bf Overview of SVDistNet: A surround-view based self-supervised distance estimation framework.]
    {\textbf{Overview of SVDistNet: A surround-view based self-supervised distance estimation framework} making use of semantic guidance and camera-geometry adaptive convolutions (\textcolor{orange}{orange} block).}
    \label{fig:svdistnet-mtl_pipeline}
\end{figure*}
Automated driving systems have a wide variety of cameras, typically around 10, placed in different car locations with different fields of view. Figure~\ref{fig:svdistnet-general_concept} shows sample distance estimation images of four cameras mounted on a car covering the entire $360\degree$ FoV surrounding the car. Instead of developing an individual model for each camera, developing a single model for all cameras is highly desirable, as discussed in the introduction. It is an unsolved problem, and we aim to solve this by incorporating camera geometry into distance estimation. We intend to convert all the camera geometry properties into a tensor called camera geometry tensor $C_t$ (CGT), which will then be passed to the CNN model at both training and inference. From the view of distance estimation, camera intrinsics is the primary model adaptation needed. However, the CGT notion is generic, and we plan to extend it to include camera extrinsic and camera motion (visual odometry) for improving other tasks. The closest work is CAM-Convs~\cite{Facil2019}, which uses camera-aware convolutions for pinhole cameras. We build upon this work and generalize to arbitrary camera geometries, including fisheye cameras.\par

The camera geometry adaptive mechanism is fundamental in the training process of the \textit{SVDistNet} as the four different cameras mounted on the car have different intrinsic parameters and viewpoints. The trained distance and pose estimation networks need to generalize when deployed on a different car with a change in multiple viewpoints and intrinsics. To achieve this, we introduce the CGT in the mapping from RGB features to 3D information for distance estimation and semantic segmentation, as shown in Figure~\ref{fig:svdistnet-model_arch}. We also add $C_t$ to the pose encoder, shown in Figure~\ref{fig:svdistnet-mtl_pipeline}. It is included in each stage and also applied to every skip connection. The framework consists of processing units to train the self-supervised distance estimation (\textcolor[HTML]{00b0f0}{blue} blocks) and semantic segmentation (\textcolor[HTML]{00b050}{green} blocks). The CGT (\textcolor{orange}{orange} block) helps \textit{SVDistNet} to yield distance maps on multiple camera-viewpoints and makes the network camera independent. $C_t$ can also be adapted to the standard camera models, as explained in Section~\ref{sec:camera-geometry-tensor}. Both modalities are weighted and optimized simultaneously by the multi-task loss from~\ref{eq:mtl_loss}. The proposed framework can obtain top-view geometric information by post-processing the predicted distance and semantic maps in \emph{3D space}.\par

The CGT is formulated in a three-step process:
For efficient training, the pixel-wise coordinates and angle of incidence maps are pre-computed. The normalized coordinates per pixel are used for these channels by incorporating information from the camera calibration. We concatenate these tensors and represent them by $C_t$ and pass it along with the input features to the SAN \emph{pairwise} and \emph{patchwise} operation modules. It comprises six channels in addition to the existing decoder channel inputs. The proposed approach can, in principle, be applied to any fisheye projection model of choice. We briefly discuss standard projection models which the CGT supports. For fisheye lenses, the mapping of 3D points to pixels universally requires a radial component $\varrho\left(\theta\right)$~\cite{hughes2010fisheye}. The polynomial model is the commonly used one, and relatively recent projection models are UCM (Unified Camera Model)~\cite{barreto2006unifying} and eUCM (Enhanced UCM) \cite{khomutenko2016eucm}. Rectilinear (representation of pinhole model) and Stereographic (mapping of a sphere to a plane) camera models are not suitable for fisheye lenses but provided for comparison. Double Sphere~\cite{usenko2018doublesphere} is a recently proposed model with a closed-form inverse with low computational complexity. For further information on radial distortion models we refer Section~\ref{sec:projection-models}.\par

The different maps included in the shared self-attention encoder are computed using the camera intrinsic parameters, where the distortion coefficients $a_1, a_2, a_3, a_4$ are used to create the angle of incidence maps $(a_x,a_y)$, $c_x, c_y$ are used to compute the principal point coordinate maps $(cc_x,cc_y)$ and the camera's sensor dimensions (width $w$ and height $h$) are utilized to formulate the normalized coordinate maps.\par
\subsubsection{Centered Coordinates $(cc_x,cc_y)$}
The information of the principal point position is fed to the SAN \emph{pairwise} and \emph{patchwise} operation modules by including the $cc_x$ and $cc_y$ coordinate channels centered at $(0,0)$. We formulate $cc_x$ and $cc_y$ channels as shown below:
\begin{equation}
\label{eq:ccx}
    cc_x=\begin{pmatrix}
           \smash{0 -c_x}\\
           \smash{1 -c_x}\\
           \smash{\vdots} \\
           \smash{w-c_x}
    \end{pmatrix}\cdot
    \begin{pmatrix}
           1 \\
           1 \\
           \smash{\vdots} \\
           1
    \end{pmatrix}_{(h+1)\times 1}^{\intercal} = 
    \begin{pmatrix} 
    -c_x & {\cdots} & -c_x \\ 
    {\vdots} & {\ddots} & {\vdots} \\ 
    w-c_x & {\cdots} & w-c_x 
    \end{pmatrix}
\end{equation}
\begin{equation}
\label{eq:ccy}
    cc_y=\begin{pmatrix}
           1 \\
           1 \\
           \smash{\vdots} \\
           1
    \end{pmatrix}_{(w+1)\times 1}\cdot\begin{pmatrix}
           \smash{0 -c_y} \\
           \smash{1 -c_y} \\
           \smash{\vdots} \\
           \smash{h - c_y}
    \end{pmatrix}^{\intercal}=
    \begin{pmatrix} 
    -c_y & \cdots & h - c_y \\ 
    \vdots & \ddots & \vdots \\ 
    -c_y & \cdots & h - c_y 
    \end{pmatrix}
\end{equation}
We concatenate $cc_x$ and $cc_y$ by resizing them using bilinear interpolation to match the input feature size.\par
\subsubsection{Angle of Incidence Maps $(a_x, a_y)$}

For the pinhole (Rectilinear) camera model, the horizontal and vertical angle of incidence maps are calculated from the $cc$ maps using the camera's focal length $f$
\begin{equation}
    \label{eq:ff}
    a_{ch}[i,j] = \arctan\Big(\frac{cc_{ch}[i,j]}{f}\Big)
\end{equation}
where $ch$ can be $x$ or $y$ (see Eq.~\ref{eq:ccx} and~\ref{eq:ccy}). For the different fisheye camera models, the angle of incidence maps can analogously be deduced by taking the inverse of the radial distortion functions $\varrho(\theta)$ listed above. Specifically, for the polynomial model used in this work, the angle of incidence $\theta$ is formulated by calculating the $4^\text{th}$ order polynomial roots of $\varrho = \sqrt{x_I^2 + y_I^2} = a_1 \theta + a_2 \theta^2 + a_3 \theta^3 + a_4 \theta^4$ through a numerical method. 
We store the pre-calculated roots in a lookup table for all the pixel coordinates to achieve training efficiency and create the $a_x$ and $a_y$ maps by setting $x_I = cc_x[i,j], y_I = 0$ and $x_I = 0, y_I = cc_y[i,j]$ respectively. As UCM, eUCM, and Double Sphere projection models can be inverted analytically, there is no need for a lookup table.\par
\subsubsection{Normalized Coordinates $(nc_{x}, nc_{y})$} 

Additionally, we add two channels of normalized coordinates~\cite{liu2018intriguing, Facil2019} whose values vary between $-1$ and $1$ linearly with respect to the image coordinates. The channels are independent of the camera sensor properties and characterize the spatial extent of the content in feature space in each direction (\eg, a value of the $\hat{x}$ channel close to $1$ indicates that the feature vector at this location is close to the right border).
\subsection{Semantically-Guided Distance Decoder}
\label{sec:Semantically-Guided-Distance-Decoder}

To address the limitations of regular convolutions, we follow the approaches of \cite{su2019pixel, guizilini2019packnet} in using pixel-adaptive convolutions (PAC) for semantic guidance inside the distance estimation branch of the multi-task network. This approach can break up the translation invariance of convolutions and incorporate spatially specific information of the semantic segmentation branch.\par

To this end, as shown in Figure~\ref{fig:syndistnet_model_arch} and Figure~\ref{fig:svdistnet-model_arch} we extract feature maps at different levels from the semantic segmentation branch of the multi-task network. These semantic feature maps are consequently used to guide the respective pixel-adaptive convolutional layer, following the formulation proposed in~\cite{su2019pixel} to process an input signal $x$ to be convolved:
\begin{equation}
x_{uv}' = \sum_{ab \in \mathcal{N}_k(i, j)} K(F_{uv},F_{ab}) W [r_{a-i,b-j}]x_{ab} + B
\end{equation}
where $\mathcal{N}_k(i, j)$ defines a $k \times k$ neighbourhood window around the pixel location $uv$ (distance $r_{a-i,b-j}$ between pixel locations), which is used as input to the convolution with weights $W$ (kernel size $k$), bias $B\in \mathbb{R}^1$ and kernel $K$, that is used in this case to calculate the correlation between the semantic guidance features $F \in \mathbb{R}^D$ from the segmentation network. We follow~\cite{guizilini2019packnet} in using a Gaussian kernel:
\begin{equation}
\label{eq:pac}
   K(F_{uv},F_{ab}) = \exp\left(-\frac{1}{2} (F_{uv} - F_{ab})^T \Sigma_{uvab}^{-1} (F_{uv} - F_{ab})\right)
\end{equation}
with covariance matrix $\Sigma_{uvab}$ between features $F_{uv}$ and $F_{ab}$, which is chosen as a diagonal matrix $\sigma^2 \cdot\mathbf{1}^D$, where $\sigma$ represents a learnable parameter for each convolutional filter.\par
In this work, we use pixel-adaptive convolutions to produce \emph{semantic-aware distance features}. The fixed information encoded in the semantic network is used to disambiguate geometric representations for the generation of multi-level distance features. Compared to previous approaches~\cite{casser2019depth, guizilini2019packnet}, we use features from the semantic segmentation branch that is trained simultaneously with the distance estimation branch introducing a more favorable one-stage training.\par
\subsection{Comparison of Convolution vs. Self-Attention}
\label{sec:comparison-to-cnn}

The pairwise models match or outperform the convolutional baselines while requiring similar or less parameters and FLOP budgets. The patchwise models perform even better in terms of computational complexity. For example, the patchwise SAN10 with $11.8$M params and $1.9$G FLOPS outperforms ResNet50 with $25.6$M params and $4.1$G FLOPS, a 54\% lower parameter and 44\% lower FLOP count. SAN10-patch models' parameter count is almost nearly equivalent to ResNet18 with $11.7$M params and $1.8$G FLOPS, whereas SAN15-patch with $20.5$M params and $3.3$G FLOPS is equivalent to ResNet50's parameter count. The SAN10-pairwise with $10.5$M params and $2.2$G FLOPS outperforms ResNet18 with a 9\% lower parameter count and 22\% higher FLOP count. We could leverage the usage of a more robust loss function compared to \lone to reduce training times on SAN10 by performing a single-scale image distance prediction in contrast to the multi-scale prediction in the previous works~\cite{kumar2020fisheyedistancenet, kumar2020unrectdepthnet}.\par
\section{Experiments}

\subsection{Ablative Experiments on Woodscape}

\subsubsection{Effect of Multi-Task Learning}
\begin{table}[t]
  \centering
    \begin{adjustbox}{width=0.75\columnwidth}
    \setlength{\tabcolsep}{0.2em}
      \begin{tabular}{lcccccc}
      \toprule
      \multicolumn{1}{l}{\textbf{Method}} 
      & \multicolumn{1}{c}{\cellcolor[HTML]{00b050}\textit{ Seg}} 
      & \multicolumn{1}{c}{\cellcolor[HTML]{00b0f0}\textit{Dist.}}
      & \multicolumn{1}{c}{\cellcolor[HTML]{ab9ac0}\textit{MTL}} 
      & \multicolumn{1}{c}{\cellcolor[HTML]{e5b9b5}
      \begin{tabular}[c]{@{}c@{}} \textit{MTL} \\ \textit{(Synergy)} \end{tabular}} 
      & \multicolumn{1}{c}{\cellcolor[HTML]{00b050}
      \begin{tabular}[c]{@{}c@{}} \textit{Seg.} \\ \textit{(mIoU)} \end{tabular}} 
      & \multicolumn{1}{c}{\cellcolor[HTML]{00b0f0}
      \begin{tabular}[c]{@{}c@{}} \textit{Distance} \\ \textit{(RMSE)} \end{tabular}} \\ 
      \toprule
      \multirow{4}{*}{\begin{tabular}[c]{@{}c@{}} SynDistNet~\cite{kumar2020syndistnet} \end{tabular}}
      & \ch & \xm & \xm & \xm & 76.8          & \xm   \\ 
      & \xm & \ch & \xm & \xm & \xm           & 2.316 \\
      & \ch & \ch & \ch & \xm & 78.3          & 2.128 \\ 
      & \ch & \ch & \xm & \ch & \textbf{81.5} & \textbf{1.714} \\
      \midrule
      \multirow{4}{*}{\begin{tabular}[c]{@{}c@{}} SVDistNet \\ (SAN10-patch) \end{tabular}}
      & \ch & \xm & \xm & \xm & 77.1          &  \xm   \\ 
      & \xm & \ch & \xm & \xm & \xm           & 2.153 \\
      & \ch & \ch & \ch & \xm & 78.6          & 1.861 \\ 
      & \ch & \ch & \xm & \ch & \textbf{82.3} & \textbf{1.532} \\
      \bottomrule
    \end{tabular}
\end{adjustbox}
\caption
{\bf Effect of the multi-task training approaches SynDistNet and SVDistNet compared with each other.}
\label{tab:svdistnet-mtl}
\end{table}

Table~\ref{tab:svdistnet-mtl} captures the primary goal of this work, which is to develop a synergistic multi-task network for semantic segmentation and distance estimation. ResNet-18 backbone was used for \textit{SynDistNet} and SAN-10 for \textit{SVDistNet}. We observe that both outputs improve through the multi-task training, mainly the distance estimation performance profits from the synergized semantic guidance. Firstly, we formulate single-task baselines for these tasks and build an essential shared encoder multi-task learning (MTL) baseline. The MTL results are slightly better than their respective single-task benchmarks demonstrating that shared encoder features can be learned for diverse tasks wherein segmentation captures semantic features and distance estimation captures geometric features. The proposed synergized MTL network \textit{SynDistNet} reduces distance RMSE by $25\%$ and improves segmentation accuracy by $4\%$ as shown in Table~\ref{tab:svdistnet-mtl}. We break down these results further using extensive ablation experiments.\par

Later we compare the \textit{SVDistnet} with \textit{SynDistNet} on the Woodscape dataset. The MTL results and the single-task benchmark for distance estimation are significantly better than \textit{SynDistNet} due to the usage of an improved self-attention encoder and the CGT. However, we observe only minimal gain for the semantic segmentation task. In the final experiment, we include the synergy between the distance and segmentation decoders. We observe that the content and channel adaptive self-attention encoder features can be learned jointly for these diverse tasks. The captured semantic features, used along with pixel-adaptive convolutions, guide the distance decoder to capture better geometric features. We break down these results further using ablation experiments.\par
\begin{table*}[!t]
  \centering{
  \small
  \begin{adjustbox}{width=\columnwidth}
  \setlength{\tabcolsep}{0.18em}
  \begin{tabular}{lcccccccccccc}
    \toprule
        \multirow{2}{*}{\textbf{Network}}
        & \multirow{2}{*}{\emph{RL}}
        & \multirow{2}{*}{\emph{Self-Attn}}
        & \multirow{2}{*}{\emph{SEM}}
        & \multirow{2}{*}{\emph{Mask}}
        & \multirow{2}{*}{\emph{CGT}}
        & \cellcolor[HTML]{7d9ebf} Abs$_{rel}$
        & \cellcolor[HTML]{7d9ebf} Sq$_{rel}$
        & \cellcolor[HTML]{7d9ebf} RMSE
        & \cellcolor[HTML]{7d9ebf} RMSE$_{log}$
        & \cellcolor[HTML]{e8715b} $\delta<1.25$
        & \cellcolor[HTML]{e8715b} $\delta<1.25^2$
        & \cellcolor[HTML]{e8715b} $\delta<1.25^3$ \\
        \cmidrule(lr){7-10} \cmidrule(lr){11-13}
         &  &   &  &  &  & \multicolumn{4}{c}{\cellcolor[HTML]{7d9ebf} lower is better} & \multicolumn{3}{c}{\cellcolor[HTML]{e8715b} higher is better} \\
    \toprule
        FisheyeDistanceNet~\cite{kumar2020fisheyedistancenet} 
        & \xm & \xm & \xm & \xm & \xm & 0.152 & 0.768 & 2.723 & 0.210 & 0.812 & 0.954 & 0.974 \\
    \midrule
        \multirow{6}{*}{SynDistNet (ResNet18)~\cite{kumar2020syndistnet}}
        & \ch & \xm & \xm & \xm & \xm & 0.142 & 0.537 & 2.316 & 0.179 & 0.878 & 0.971 & 0.985 \\
        & \ch & \xm & \xm & \ch & \xm & 0.133 & 0.491 & 2.264 & 0.168 & 0.868 & 0.976 & 0.988 \\
        & \ch & \ch & \xm & \ch & \xm & 0.121 & 0.429 & 2.128 & 0.155 & 0.875 & 0.980 & 0.990 \\
        & \ch & \ch & \ch & \xm & \xm & 0.105 & 0.396 & 1.976 & 0.143 & 0.878 & 0.982 & 0.992 \\
        & \ch & \ch & \ch & \ch & \xm & \textbf{0.076} & \textbf{0.368} & \textbf{1.714} & \textbf{0.127} & \textbf{0.891} & \textbf{0.988} & \textbf{0.994} \\
    \midrule
        \multirow{6}{*}{SynDistNet(ResNet50)~\cite{kumar2020syndistnet}}
        & \ch & \xm & \xm & \xm & \xm & 0.138 & 0.540 & 2.279 & 0.177 & 0.880 & 0.973 & 0.986 \\
        & \ch & \xm & \xm & \ch & \xm & 0.127 & 0.485 & 2.204 & 0.166 & 0.881 & 0.975 & 0.989 \\
        & \ch & \ch & \xm & \ch & \xm & 0.115 & 0.413 & 2.028 & 0.148 & 0.876 & 0.983 & 0.992 \\
        & \ch & \ch & \ch & \xm & \xm & 0.102 & 0.387 & 1.856 & 0.135 & 0.884 & 0.985 & 0.994 \\
        & \ch & \ch & \ch & \ch & \xm & \textbf{0.068} & \textbf{0.352} & \textbf{1.668} & \textbf{0.121} & \textbf{0.895} & \textbf{0.990} & \textbf{0.996} \\
    \midrule
        \multirow{6}{*}{SVDistNet (SAN10-patch)}
        & \ch & \ch & \xm & \xm & \xm & 0.128 & 0.469 & 2.153 & 0.164 & 0.875 & 0.974 & 0.986 \\
        & \ch & \ch & \xm & \ch & \xm & 0.114 & 0.413 & 2.022 & 0.149 & 0.878 & 0.982 & 0.990 \\
        & \ch & \ch & \xm & \ch & \ch & 0.101 & 0.378 & 1.861 & 0.133 & 0.884 & 0.984 & 0.991 \\
        & \ch & \ch & \ch & \xm & \xm & 0.094 & 0.345 & 1.789 & 0.128 & 0.887 & 0.985 & 0.992 \\
        & \ch & \ch & \ch & \xm & \ch & 0.082 & 0.316 & 1.682 & 0.119 & 0.890 & 0.987 & 0.993 \\
        & \ch & \ch & \ch & \ch & \xm & 0.074 & 0.343 & 1.641 & 0.112 & 0.892 & 0.985 & 0.994 \\
        & \ch & \ch & \ch & \ch & \ch &\textbf{0.057} & \textbf{0.315} & \textbf{1.532} & \textbf{0.108} & 
\textbf{0.896} & \textbf{0.986} & \textbf{0.996} \\
    \midrule
        \multirow{2}{*}{SVDistNet (SAN10-pair)}
        & \ch & \ch & \xm & \ch & \ch & 0.121 & 0.457 & 2.115 & 0.152 & 0.879 & 0.979 & 0.985 \\
        & \ch & \ch & \ch & \xm & \ch & 0.103 & 0.385 & 1.882 & 0.141 & 0.882 & 0.983 & 0.990 \\
        & \ch & \ch & \ch & \ch & \ch & \textbf{0.081} & \textbf{0.365} & \textbf{1.710} & \textbf{0.128} & \textbf{0.890} & \textbf{0.985} & \textbf{0.994} \\
    \midrule
        \multirow{6}{*}{SVDistNet (SAN19-patch)}
        & \ch & \ch & \xm & \xm & \xm & 0.121 & 0.437 & 2.127 & 0.153 & 0.878 & 0.976 & 0.989 \\
        & \ch & \ch & \xm & \ch & \xm & 0.109 & 0.408 & 2.006 & 0.145 & 0.880 & 0.982 & 0.992 \\
        & \ch & \ch & \xm & \ch & \ch & 0.098 & 0.372 & 1.849 & 0.138 & 0.884 & 0.983 & 0.991 \\
        & \ch & \ch & \ch & \xm & \xm & 0.091 & 0.351 & 1.773 & 0.129 & 0.886 & 0.986 & 0.993 \\
        & \ch & \ch & \ch & \xm & \ch & 0.070 & 0.305 & 1.669 & 0.108 & 0.893 & 0.986 & 0.994 \\
        & \ch & \ch & \ch & \ch & \xm & 0.067 & 0.296 & 1.578 & 0.106 & 0.895 & 0.985 & 0.994 \\
        & \ch & \ch & \ch & \ch & \ch &\textbf{0.048} & \textbf{0.277} & \textbf{1.486} & \textbf{0.086} & \textbf{0.901} & \textbf{0.991} & \textbf{0.996} \\
    \midrule
        \multirow{2}{*}{SVDistNet (SAN19-pair)}
        & \ch & \ch & \xm & \ch & \ch & 0.116 & 0.461 & 2.097 & 0.154 & 0.881 & 0.982 & 0.988 \\
        & \ch & \ch & \ch & \xm & \ch & 0.096 & 0.371 & 1.846 & 0.147 & 0.884 & 0.985 & 0.991 \\
        & \ch & \ch & \ch & \ch & \ch & \textbf{0.074} & \textbf{0.331} & \textbf{1.624} & \textbf{0.101} & \textbf{0.891} & \textbf{0.986} & \textbf{0.994} \\
        \bottomrule
  \end{tabular}
  \end{adjustbox}
  }
\caption[\bf Ablation study on the effect of the contributions up to our final SVDistNet model on the Woodscape.]
{\textbf{Ablation study on the effect of the contributions up to our final SVDistNet model on the Woodscape}. From our distance estimation baseline \cite{kumar2020fisheyedistancenet}, we incrementally add up the robust loss (RL), self-attention layers encoder heads (Self-Attn), semantic guidance in the decoder (SEM), dynamic object masking (Mask), and camera geometry tensor (CGT).}
\label{table:svdistnet-table_ablation}
\end{table*}

\begin{figure*}[htbp]
  \centering
  \resizebox{\textwidth}{!}{\newcommand{\turnheightnew}{0.25\columnwidth}
\centering

\begin{tabular}{@{\hskip 0.5mm}c@{\hskip 0.5mm}c@{\hskip 0.5mm}c@{\hskip 0.5mm}c@{\hskip 0.5mm}c@{}}

{\rotatebox{90}{\hspace{7mm}Raw Input}} &
\includegraphics[height=\turnheightnew]{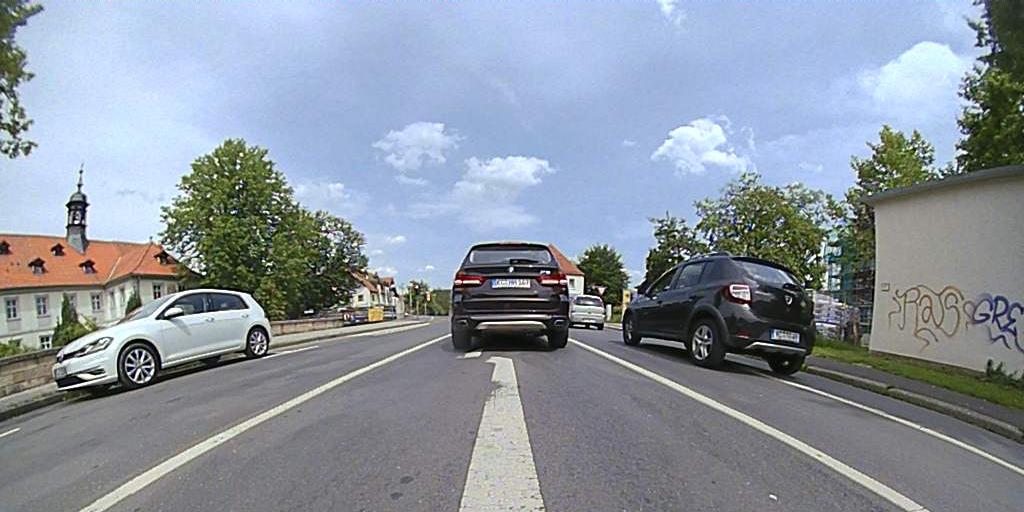} &
\includegraphics[height=\turnheightnew]{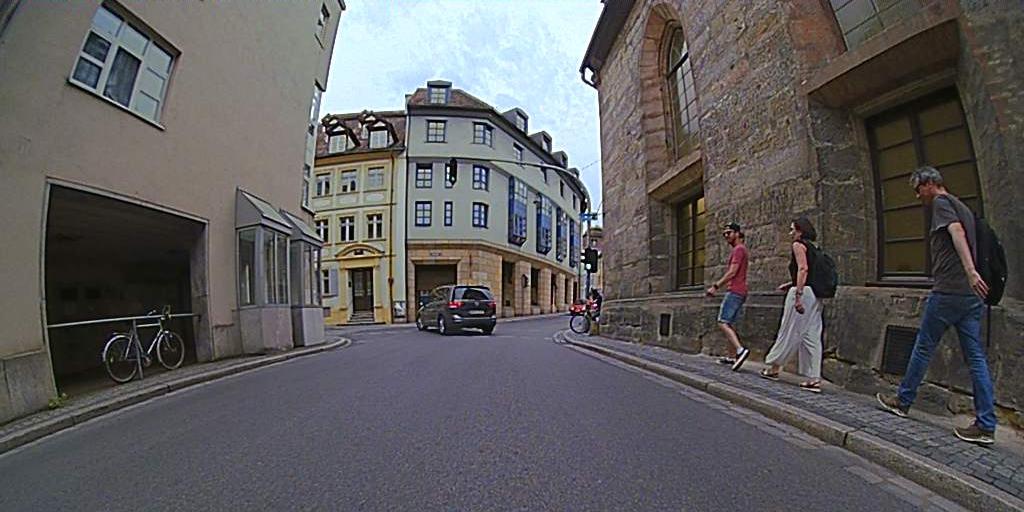} \\

{\rotatebox{90}{\hspace{0mm}\normalsize FisheyeDistanceNet}} &
\includegraphics[height=\turnheightnew]{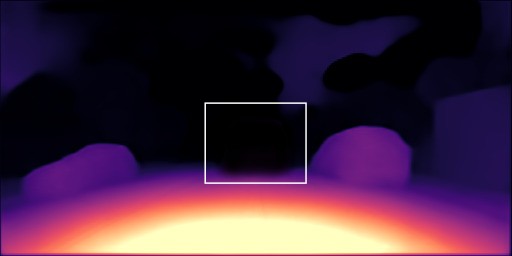} &
\includegraphics[height=\turnheightnew]{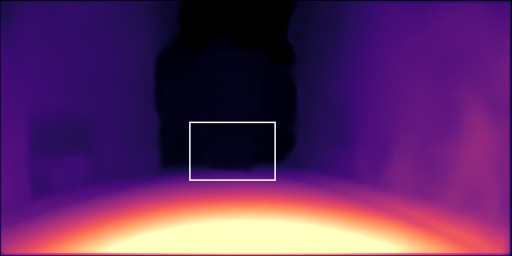} \\

{\rotatebox{90}{\hspace{7mm}SynDistNet}} &
\includegraphics[height=\turnheightnew]{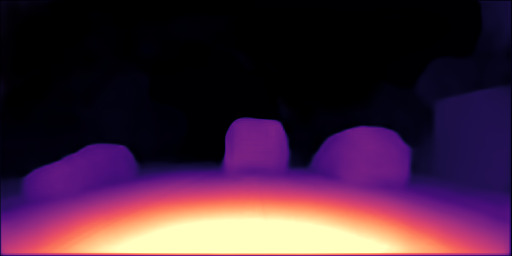} &
\includegraphics[height=\turnheightnew]{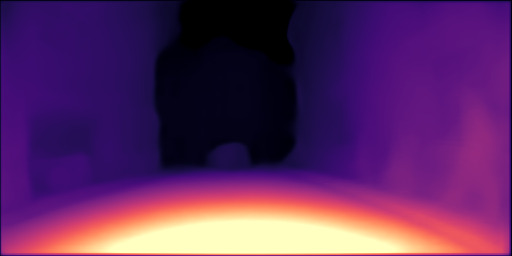} \\

{\rotatebox{90}{\hspace{7mm}Raw Input}} &
\includegraphics[height=\turnheightnew]{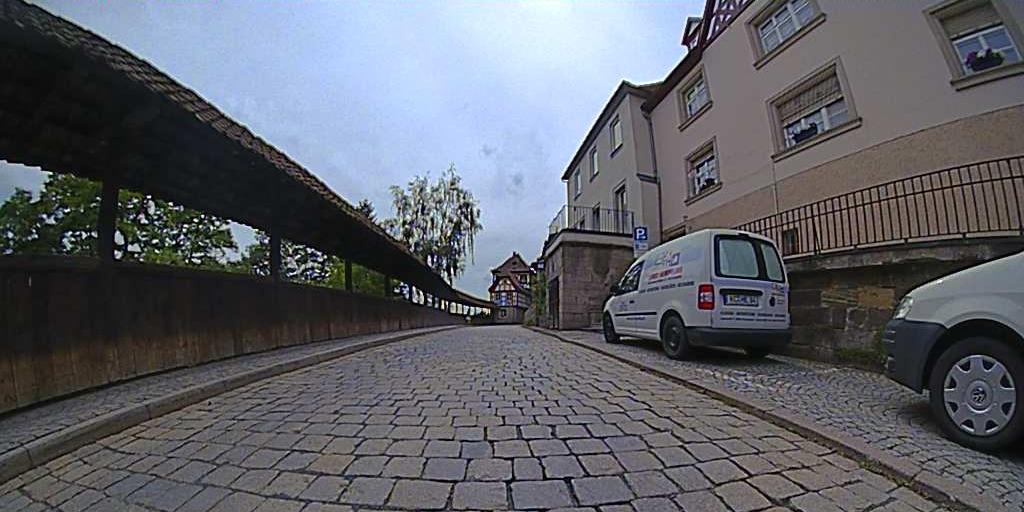} &
\includegraphics[height=\turnheightnew]{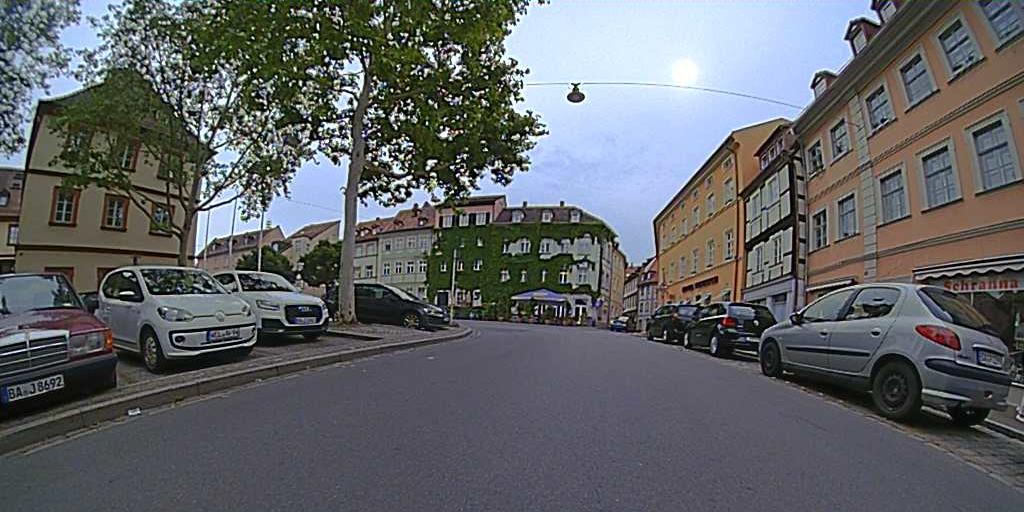}\\

{\rotatebox{90}{\hspace{0mm}\normalsize FisheyeDistanceNet}} &
\includegraphics[height=\turnheightnew]{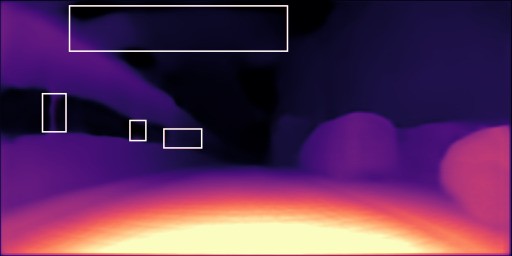} &
\includegraphics[height=\turnheightnew]{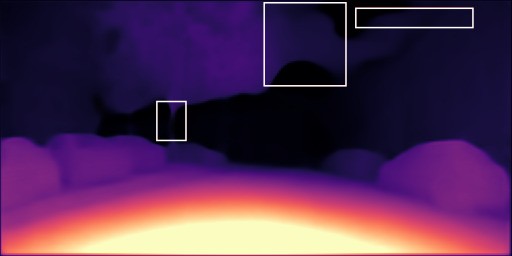}\\

{\rotatebox{90}{\hspace{7mm}SynDistNet}} &
\includegraphics[height=\turnheightnew]{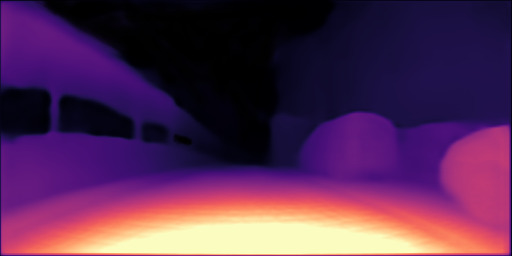} &
\includegraphics[height=\turnheightnew]{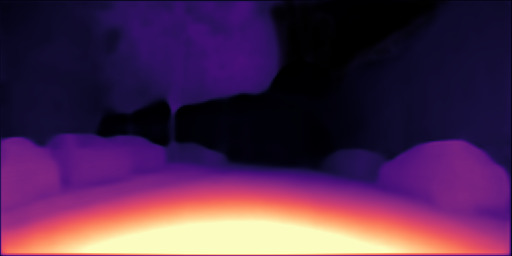} \\

\end{tabular}}
  \caption[\bf Qualitative result comparison of SynDistNet with FisheyeDistanceNet on the WoodScape.]
          {\textbf{Qualitative result comparison on the WoodScape} between the FisheyeDistanceNet and SynDistNet. SynDistNet can recover the distance of dynamic objects (left images), which eventually solves the infinite distance issue. In the 3\textsuperscript{rd} and 4\textsuperscript{th} rows, we can see that semantic guidance helps us to recover thin structures and resolves the distance of homogeneous areas, thereby outputting sharp distance maps on raw fisheye images.}
  \label{fig:syndistnet-fisheye_qual}
\end{figure*}

At first, for the ablation analysis, we consider two variants of ResNet encoder heads. Distance estimation results of these variants are shown in Table~\ref{table:svdistnet-table_ablation}. We showcase our improvements for various network architectures and, in particular, show the superiority of our \textit{SVDistNet} model over the \textit{SynDistNet} model as well as the positive effect of using the camera geometry tensor $C_t$. The distance estimates are capped to $40\,m$. We showcase the qualitative results of \textit{SynDistNet} compared to the \textit{FisheyeDistanceNet} model from \textbf{Chapter}~\ref{Chapter4} in Figure~\ref{fig:syndistnet-fisheye_qual}. Significant improvements in accuracy are obtained with the replacement of the \lone loss by a generic parameterized loss function. The impact of the mask is incremental in the WoodScape dataset. Still, it poses the potential to solve the infinite depth/distance issue and provides a way to improve the photometric loss. With the proposed self-attention-based encoder coupled with the semantically-guided decoder architecture, we can consistently improve the performance. Finally, with all the additions, we outperform \textit{FisheyeDistanceNet} for all considered metrics.\par

\begin{figure}[!t]
  \centering
  \resizebox{\columnwidth}{!}{
  \newcommand{\turnheightnew}{0.3\columnwidth}
\centering

\begin{tabular}{@{\hskip 0.4mm}c@{\hskip 0.4mm}c@{\hskip 0.4mm}c@{}}

{\rotatebox{90}{\hspace{10mm}Raw Input}} &
\includegraphics[height=\turnheightnew]{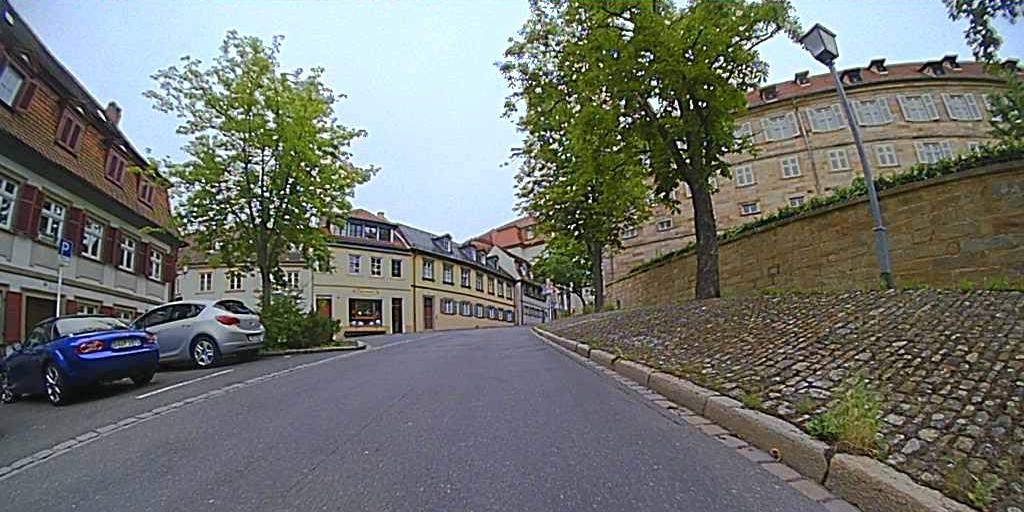} &
\includegraphics[height=\turnheightnew]{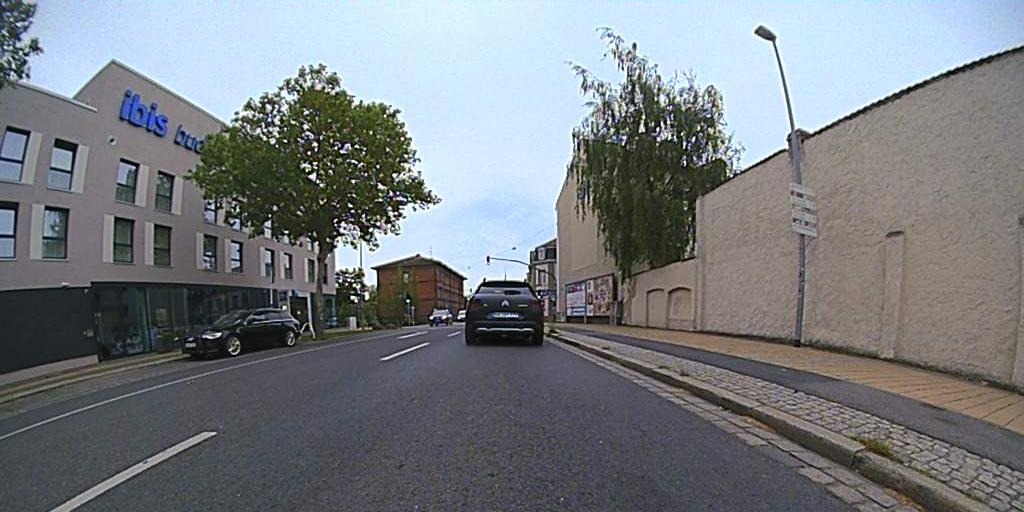} \\

{\rotatebox{90}{\hspace{3mm}FisheyeDistanceNet}} &
\includegraphics[height=\turnheightnew]{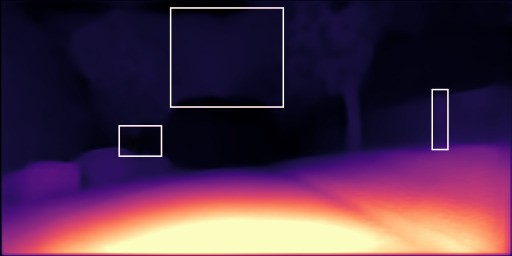} &
\includegraphics[height=\turnheightnew]{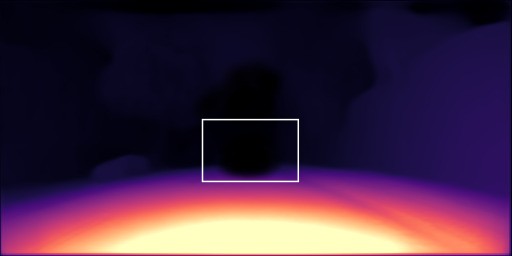} \\

{\rotatebox{90}{\hspace{10mm}SVDistNet}} &
\includegraphics[height=\turnheightnew]{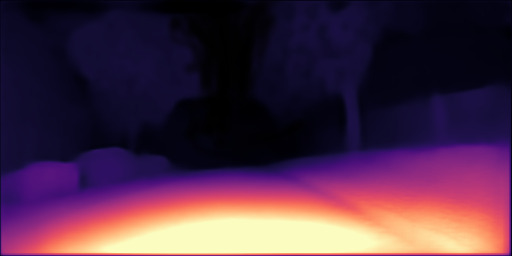} &
\includegraphics[height=\turnheightnew]{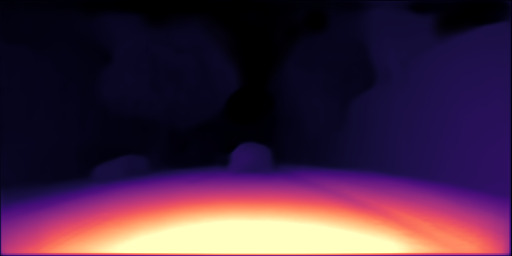} \\

{\rotatebox{90}{\hspace{10mm}SVDistNet}} &
\includegraphics[height=\turnheightnew]{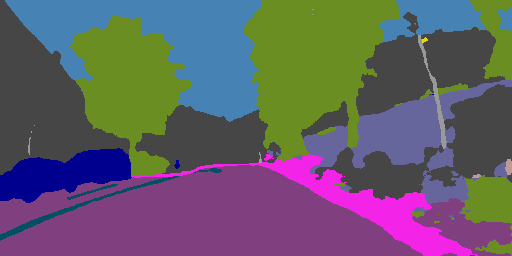} &
\includegraphics[height=\turnheightnew]{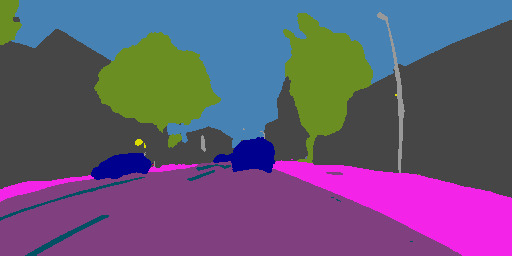} \\

\end{tabular}}
  \caption[\bf Evaluation of the SVDistNet model on WoodScape.]
          {\textbf{Evaluation of the SVDistNet model on WoodScape.} We observe that using semantic guidance inside the SVDistNet model helps to recover thin structures inside the distance map (left images). We also solve the infinite distance issue for dynamic objects by incorporating the mask described in Section~\ref{sec:dynamic-object-mask} (right images).}
  \label{fig:svdistnet-fisheye_qual}
\end{figure}

\begin{figure*}[!t]
  \centering
  \resizebox{\textwidth}{!}{
  \newcommand{\turnheightnew}{0.25\columnwidth}
\centering

\begin{tabular}{@{\hskip 0.5mm}c@{\hskip 0.5mm}c@{\hskip 0.5mm}c@{\hskip 0.5mm}c@{\hskip 0.5mm}c@{}}

{\rotatebox{90}{\hspace{0mm}}} &
\includegraphics[height=\turnheightnew]{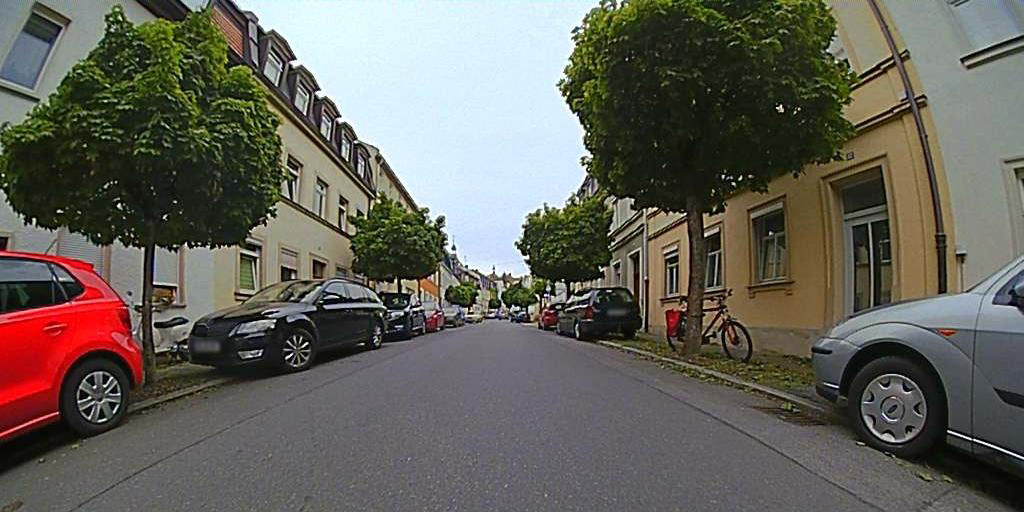} &
\includegraphics[height=\turnheightnew]{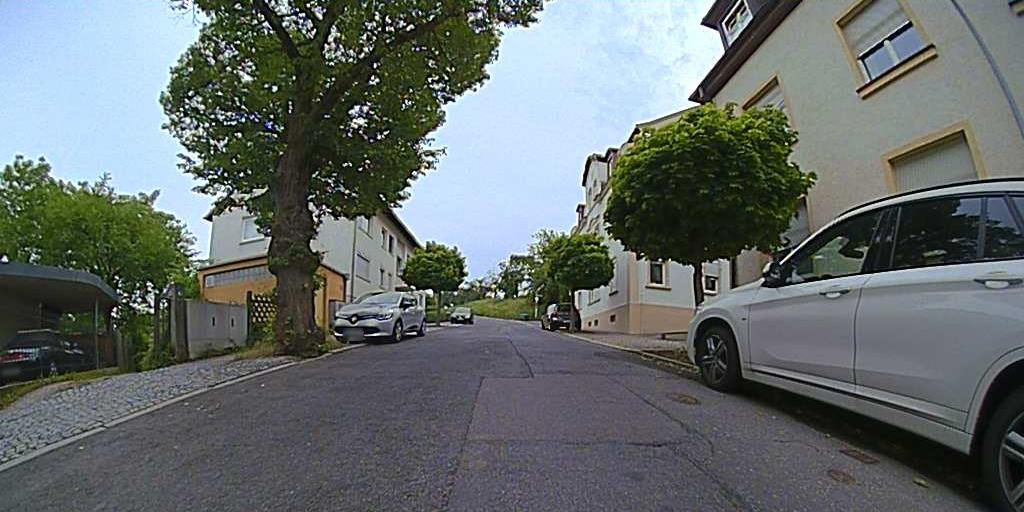} &
\includegraphics[height=\turnheightnew]{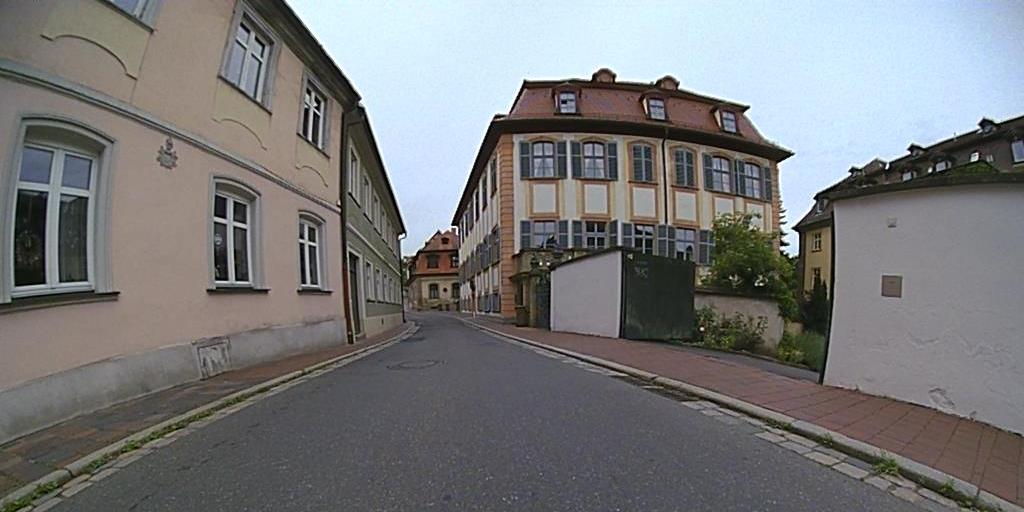} \\

{\rotatebox{90}{\hspace{0mm}}} &
\includegraphics[height=\turnheightnew]{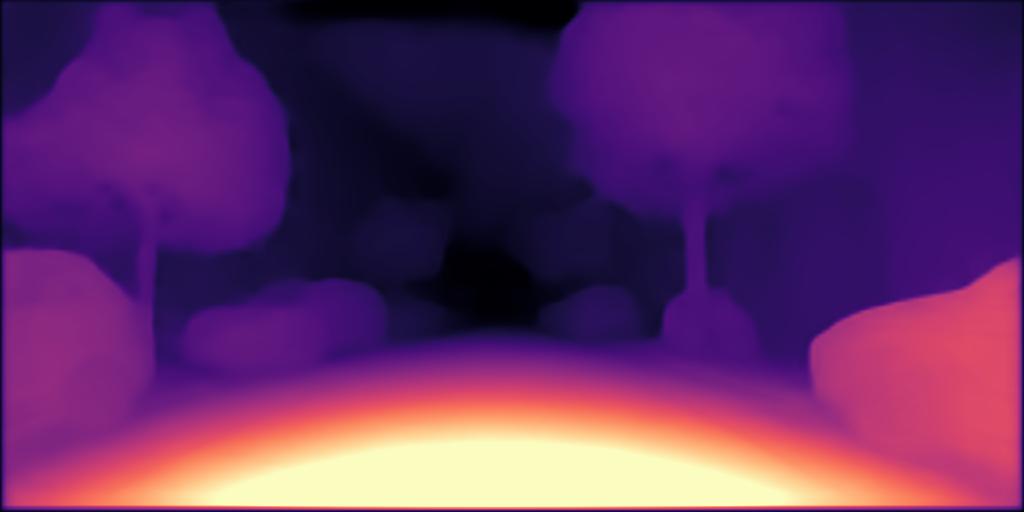} &
\includegraphics[height=\turnheightnew]{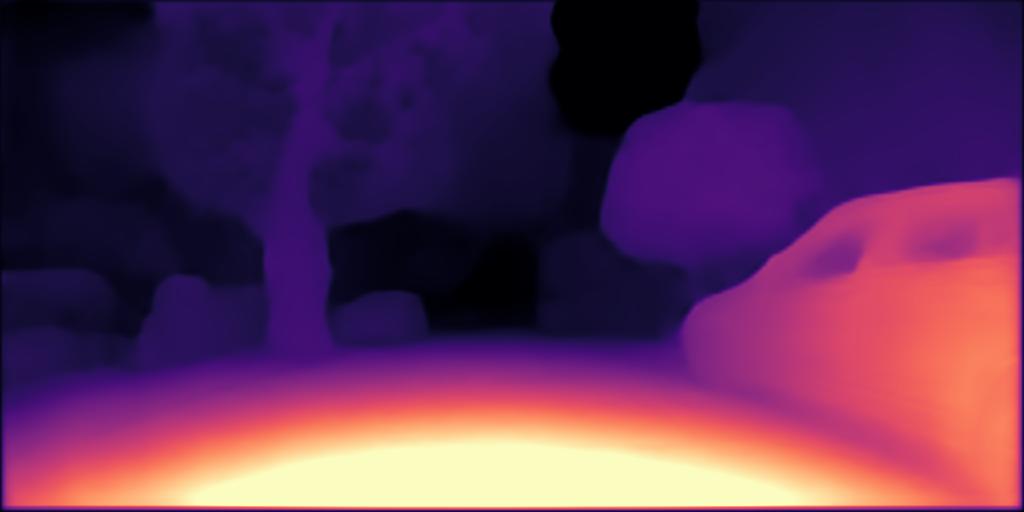} &
\includegraphics[height=\turnheightnew]{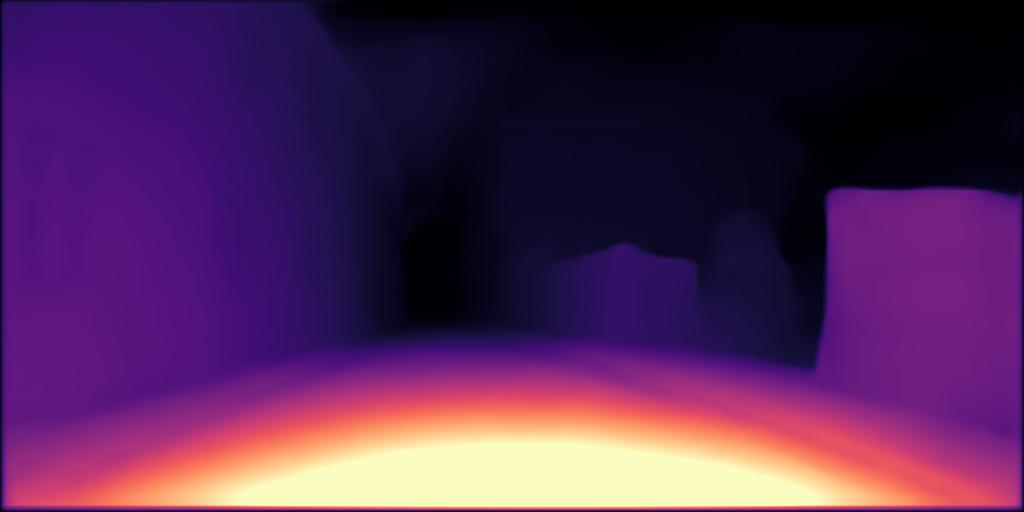} \\

{\rotatebox{90}{\hspace{0mm}\scriptsize}} &
\includegraphics[height=\turnheightnew]{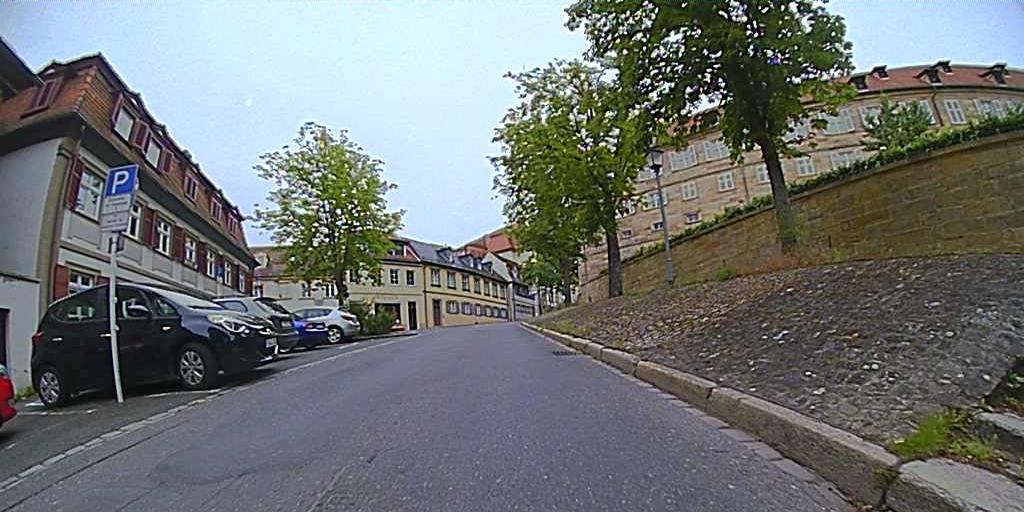} &
\includegraphics[height=\turnheightnew]{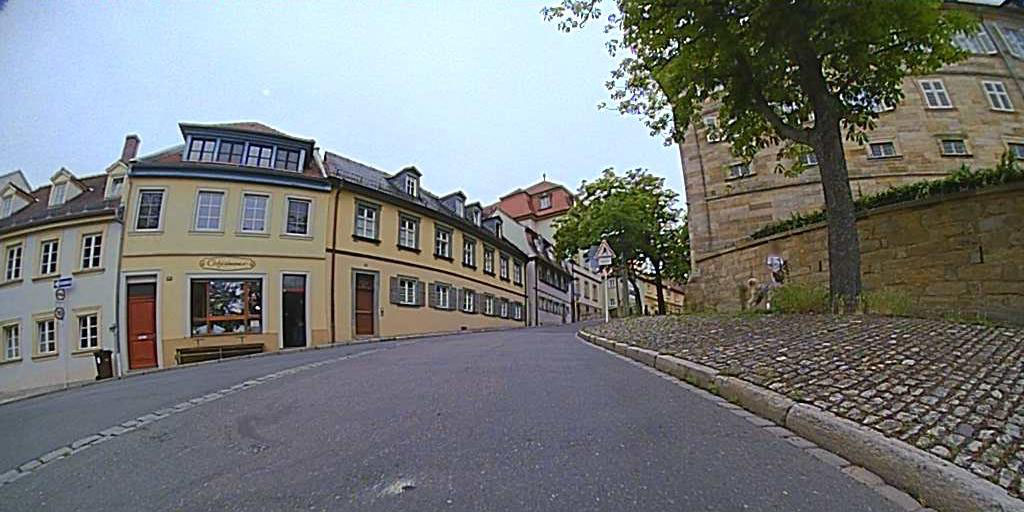} &
\includegraphics[height=\turnheightnew]{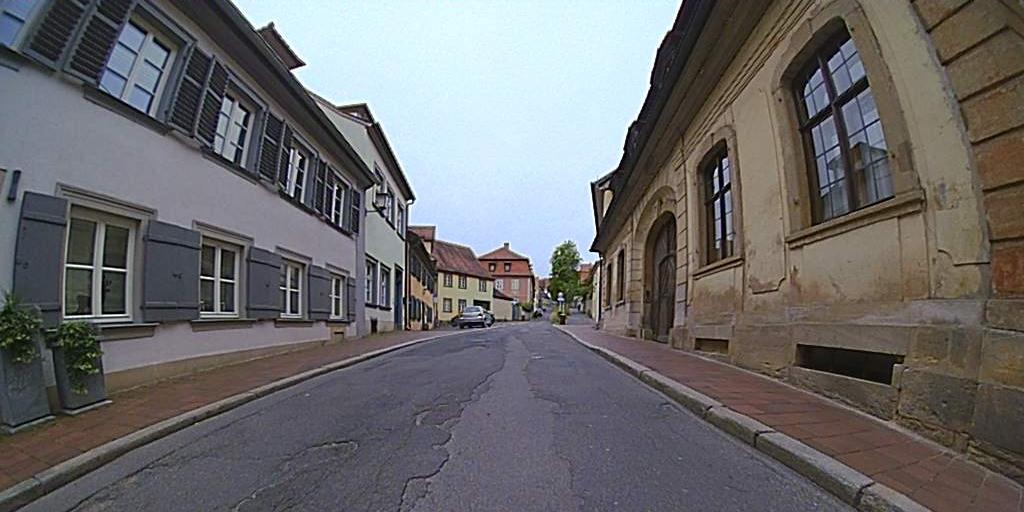} \\

{\rotatebox{90}{\hspace{0mm}\scriptsize}} &
\includegraphics[height=\turnheightnew]{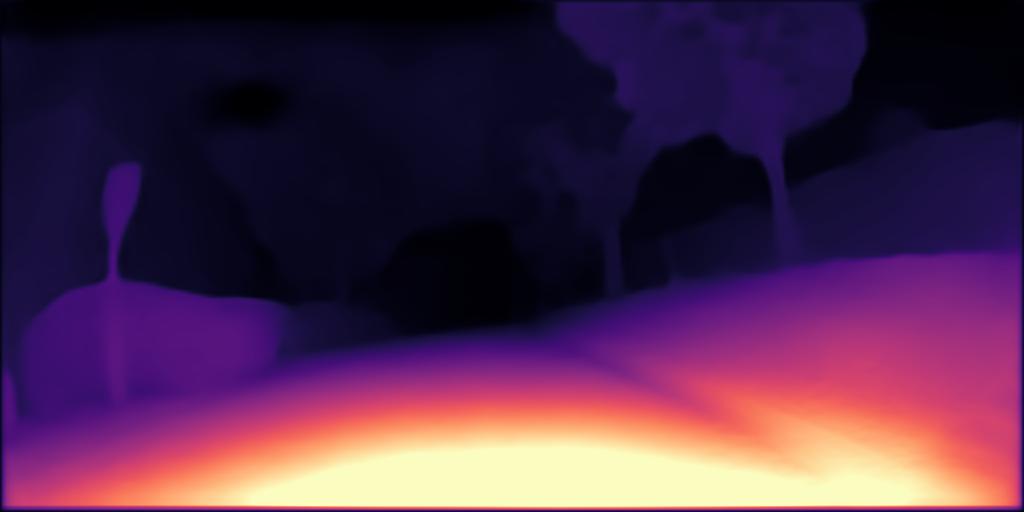} &
\includegraphics[height=\turnheightnew]{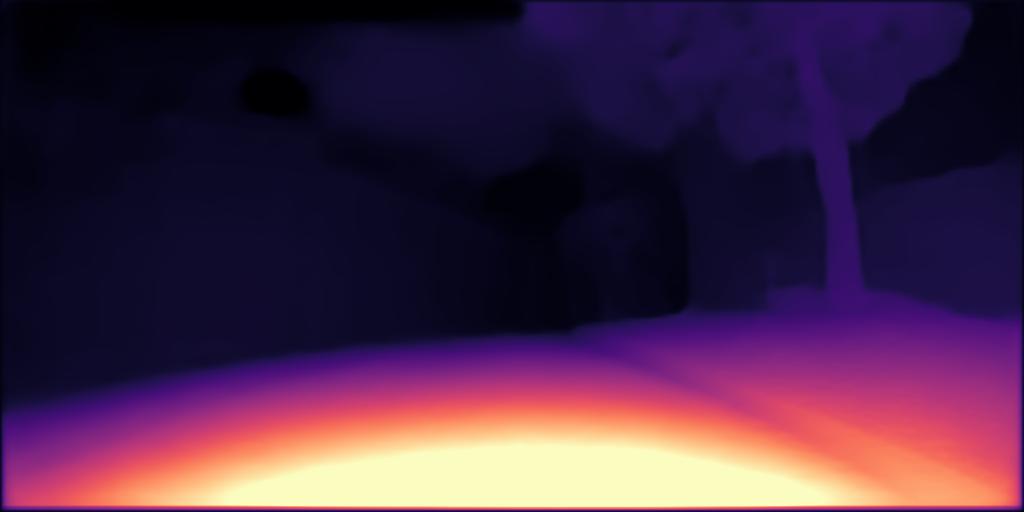} &
\includegraphics[height=\turnheightnew]{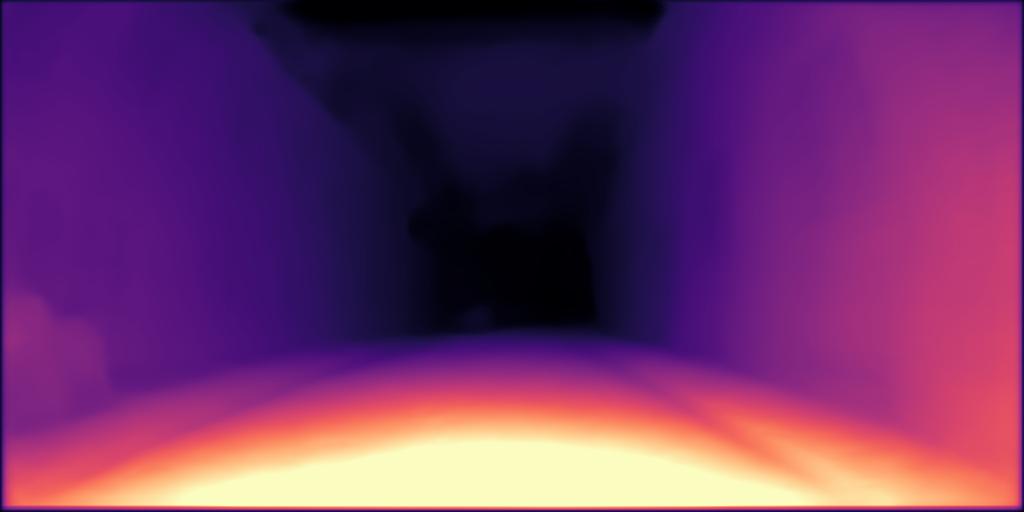} \\

{\rotatebox{90}{\hspace{0mm}\scriptsize}} &
\includegraphics[height=\turnheightnew]{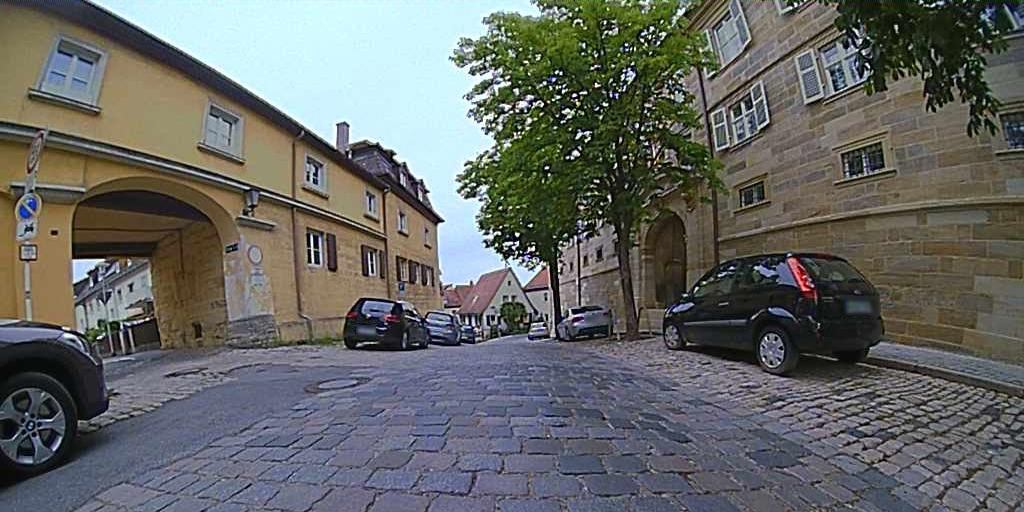} &
\includegraphics[height=\turnheightnew]{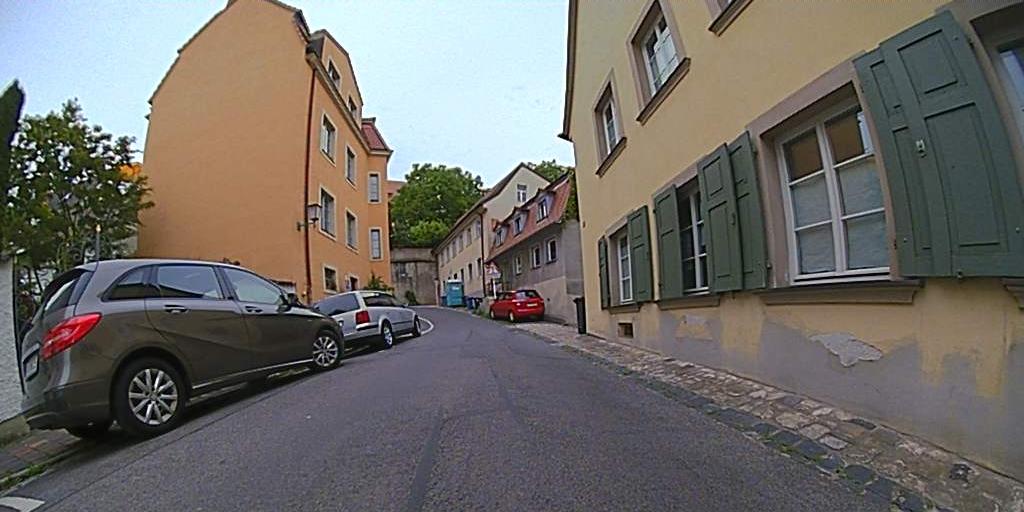} &
\includegraphics[height=\turnheightnew]{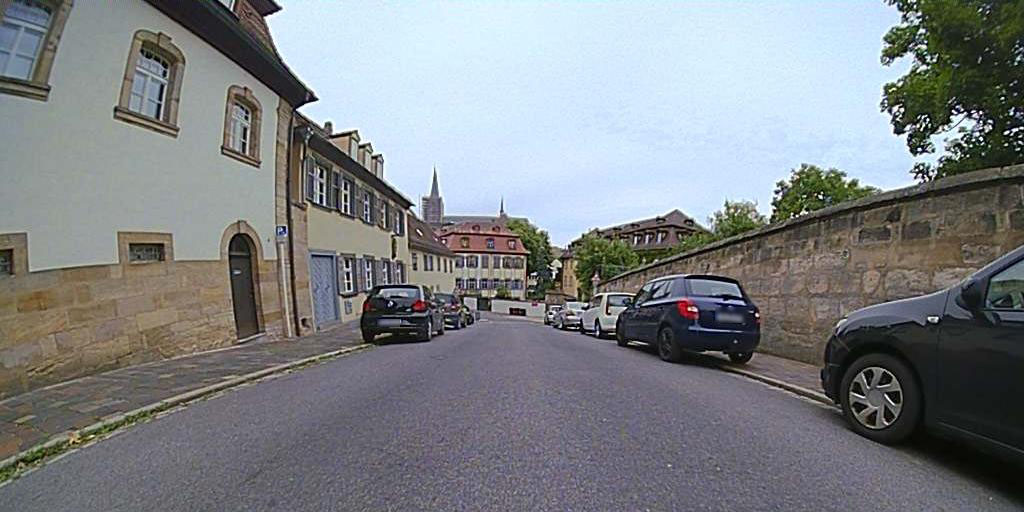} \\

{\rotatebox{90}{\hspace{0mm}\scriptsize}} &
\includegraphics[height=\turnheightnew]{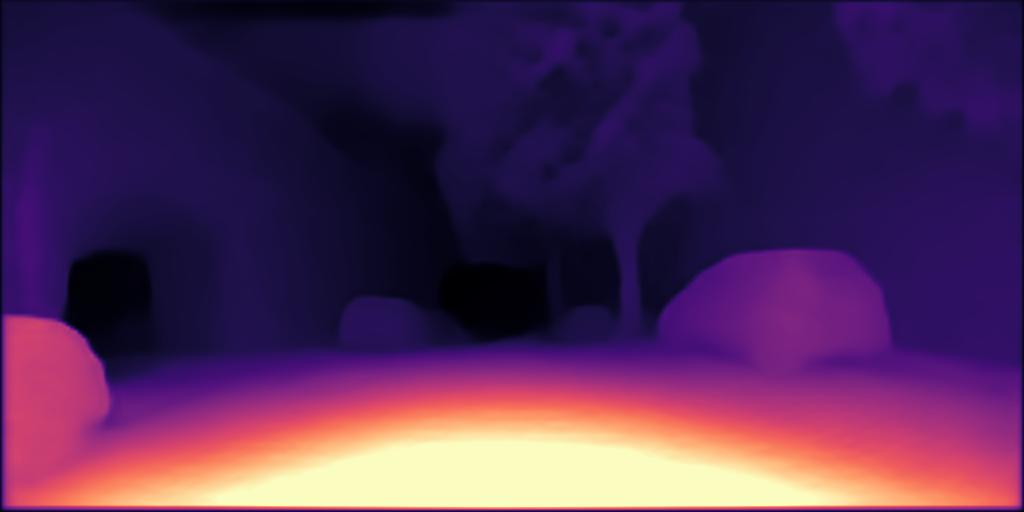} &
\includegraphics[height=\turnheightnew]{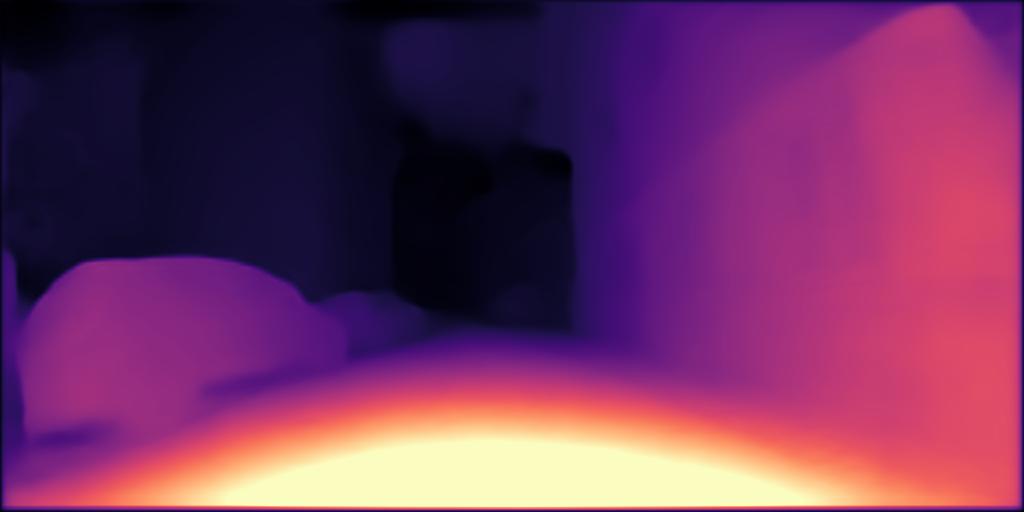} &
\includegraphics[height=\turnheightnew]{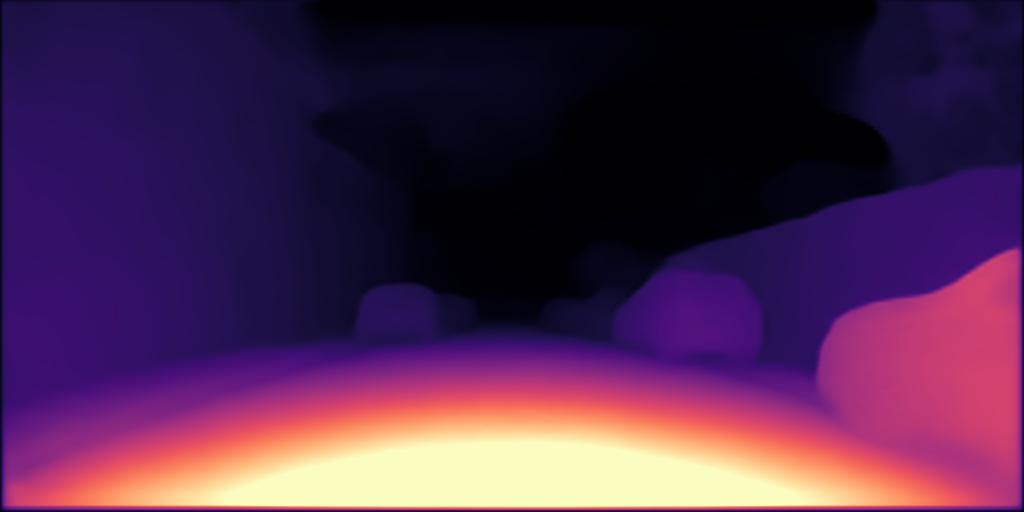} \\

{\rotatebox{90}{\hspace{0mm}\scriptsize}} &
\includegraphics[height=\turnheightnew]{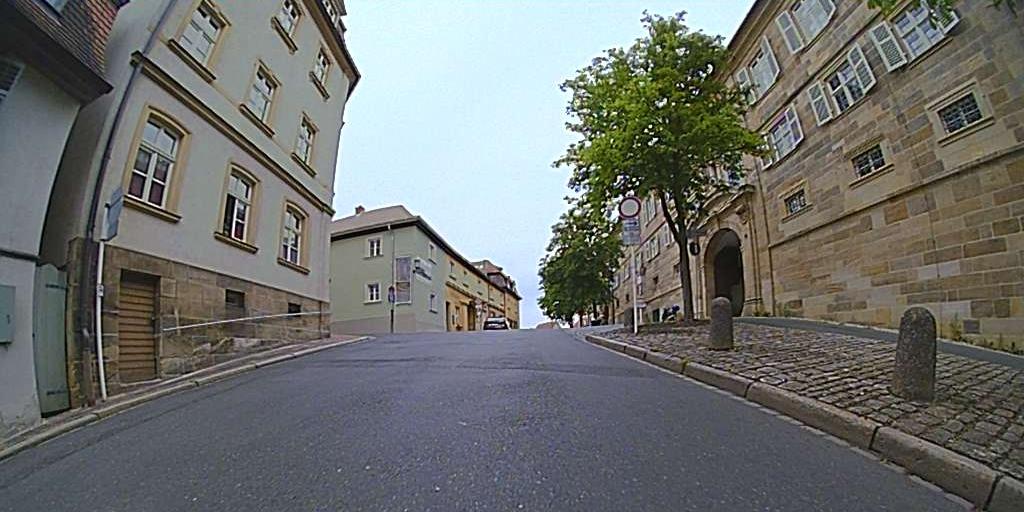} &
\includegraphics[height=\turnheightnew]{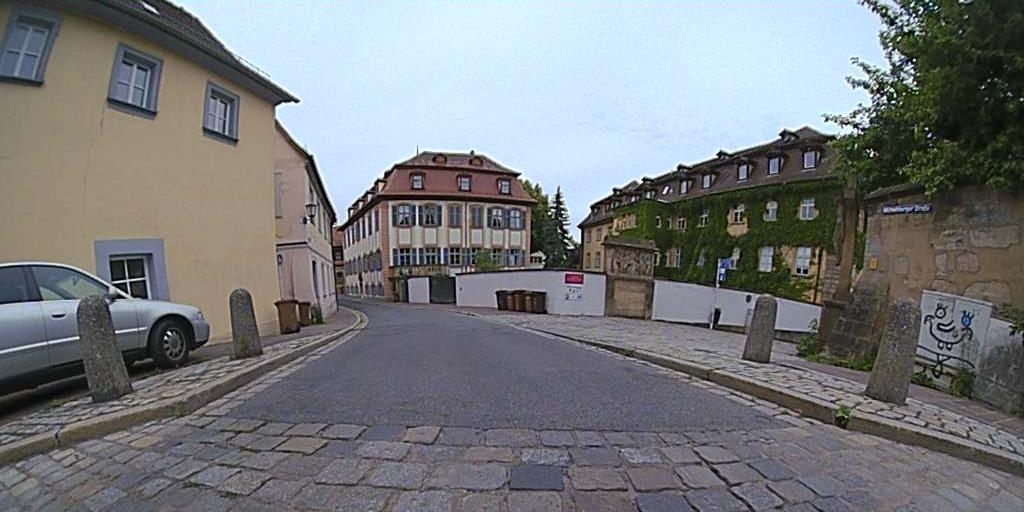} &
\includegraphics[height=\turnheightnew]{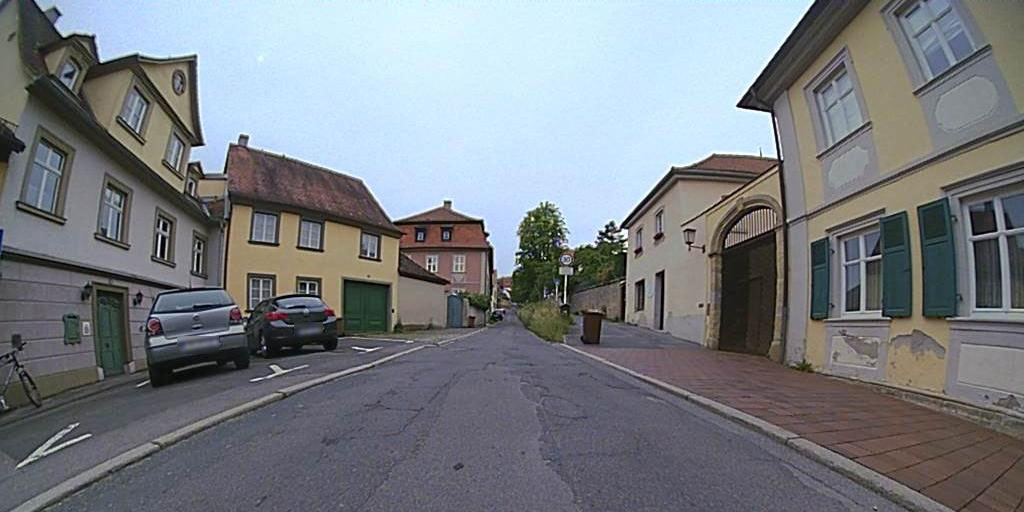} \\

{\rotatebox{90}{\hspace{0mm}\scriptsize}} &
\includegraphics[height=\turnheightnew]{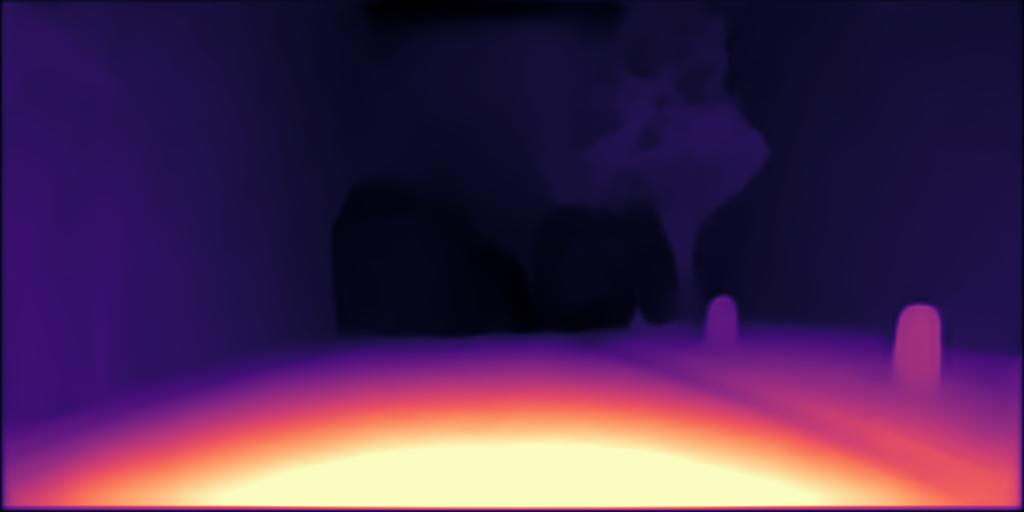} &
\includegraphics[height=\turnheightnew]{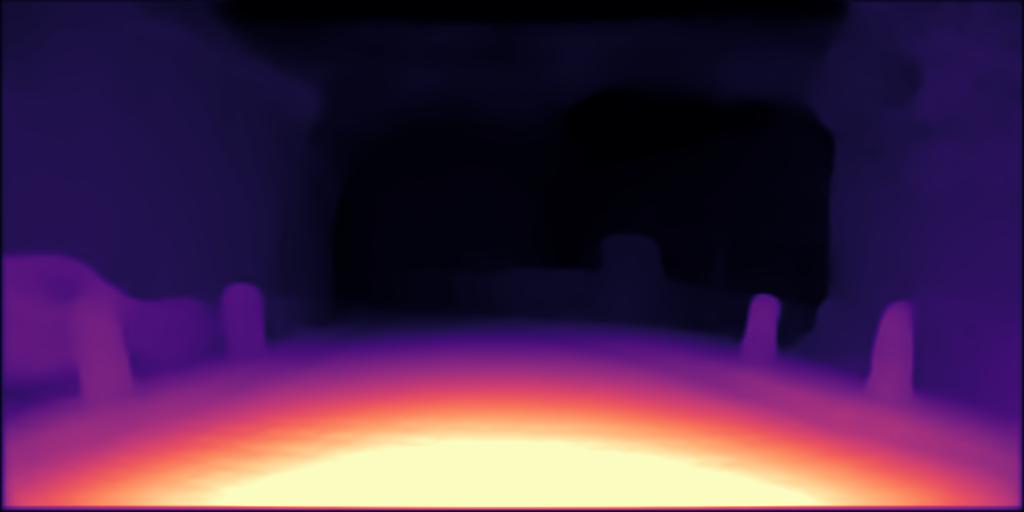} &
\includegraphics[height=\turnheightnew]{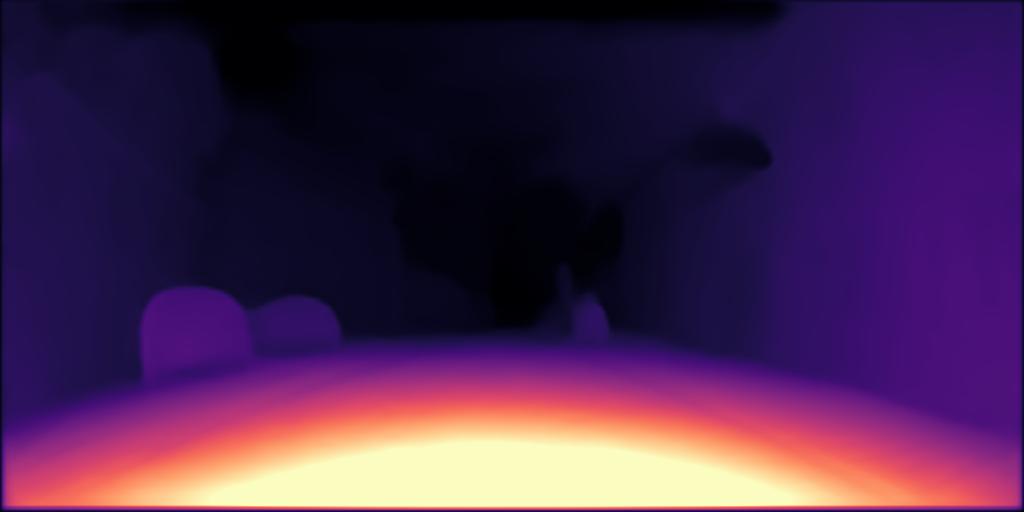} \\
\end{tabular}
}
  \caption[\bf Qualitative results of SVDistNet on WoodScape.]
          {\textbf{Qualitative results of SVDistNet on WoodScape.} In the 4\textsuperscript{th} row, we can see sharp curbs on the street, and in the 3\textsuperscript{rd} row, we can see that the model adapts to the extreme distortion induced by the fisheye camera and produces sharp distance maps. Finally, the model adapts to the most complex scenes in the last few rows and produces very sharp scale-aware distance maps. For more qualitative results, we refer to this video: \url{https://youtu.be/bmX0UcU9wtA}.}
  \label{fig:fisheye_additional_qual}
\end{figure*}

Additionally, we consider two variants of the self-attention encoder, namely pairwise and patchwise, as described in Section~\ref{sec:vector-self-attention}. We show the impact of specific contributions and their importance in the \textit{SVDistNet} framework compared to the previous work \textit{SynDistNet}. At first, we replace the \lone loss with a generic parameterized loss function and test it using the self-attention encoder's patchwise variant, wherein we gain notable improvements in accuracy. We use the dynamic object mask obtained through projecting semantic segmentation predictions as described in Section~\ref{sec:dynamic-object-mask} to filter all dynamic objects which contaminate the reconstruction loss. Additionally, this contribution possesses the potential to solve the infinite distance issue. We compare the baseline work \textit{FisheyeDistanceNet} trained only for distance estimation to \textit{SVDistNet}, which is trained in a multi-task fashion as shown in Figure~\ref{fig:svdistnet-fisheye_qual} illustrate the semantic mask's impact on the estimates with additional results of \textit{SVDistNet} in Figure~\ref{fig:fisheye_additional_qual}. When adding the CGT to this setting, we observe a significant increase in accuracy since we train multiple cameras with different camera intrinsics and viewing angles. The aforementioned training strategy makes the network camera-independent and better generalizes images taken from a different camera.\par

To further improve the multi-task setup, we perform a synergy of distance and segmentation decoders. We provide semantic features from the supervised task to the distance decoder's geometric features, where we still train the distance estimation in a self-supervised fashion. In this setting, we drop the dynamic object mask and still achieve improvements.\par

We improve the metric results further by adding the CGT to this setting, and we could surpass the accuracy obtained by the best setting in \textit{SynDistNet}. With the help of these vital features, we create an experiment comparable to \textit{SynDistNet}'s final setting, which consolidates all the listed features, excluding the CGT. We could attain better results than \textit{SynDistNet}'s ResNet50 results. It is important to note that ResNet50 is comparable to the SAN19-pair/patch encoder. However, we were able to outperform ResNet50 with the SAN10-patch encoder in terms of computational complexity (cf. Section \ref{sec:comparison-to-cnn}).\par

We complete the \textit{SVDistNet} model for the surround-view camera framework by introducing the CGT. In Figure~\ref{fig:fisheye_failure}, we show a few qualitative results of the failure cases having artifacts such as holes or merging of thin objects such as poles with the background. We also perform a few vital evaluations using the pairwise self-attention encoders. We were not able to obtain the same level of accuracy yielded by the patchwise self-attention encoder. The patchwise self-attention module is stringently more potent than the standard convolution, and the pairwise self-attention module matches or outperforms the convolutional equivalents. We perform the same vital ablation of the contributions with a higher-performing SAN19-patch encoder network. Finally, with all the additions, we outperform all previous works~\cite{kumar2020fisheyedistancenet, kumar2020unrectdepthnet, kumar2020syndistnet} for all considered metrics on the Woodscape dataset.\par
\subsubsection{Generalization Across Different Cameras}

\begin{figure*}[!t]
\centering
  \resizebox{\textwidth}{!}{
    \newcommand{\imlabel}[2]{\includegraphics[width=.49\linewidth]{#1}%
\raisebox{2pt}{\makebox[-2pt][r]{\small #2}}}

\scalebox{1}[1]{
\begin{tabular}{@{\hskip 1.5mm}c@{\hskip 1.5mm}c @{\hskip 1.5mm}c@{\hskip 1.5mm}c}
\centering
    \imlabel{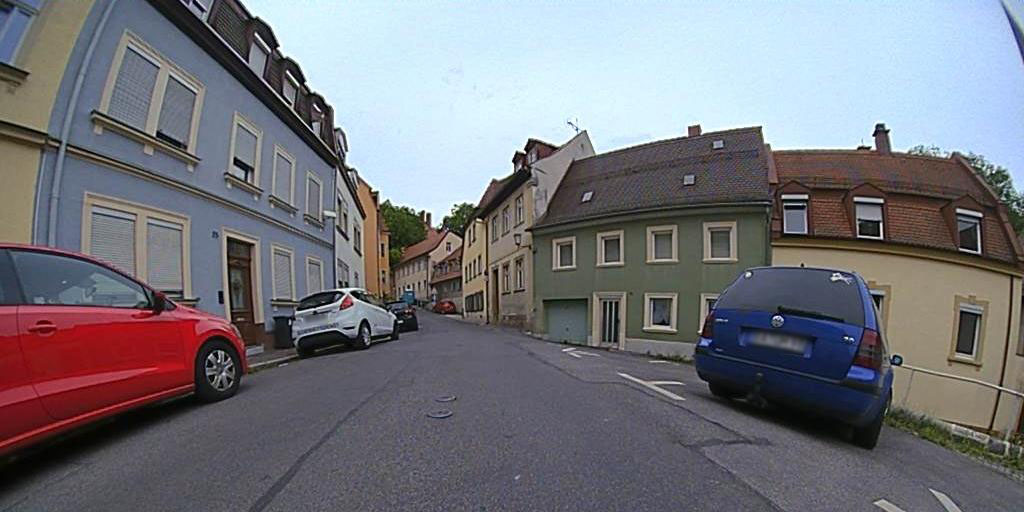}
    {\hspace{.24\textwidth}\textcolor{white}{Front Cam}} &
    \imlabel{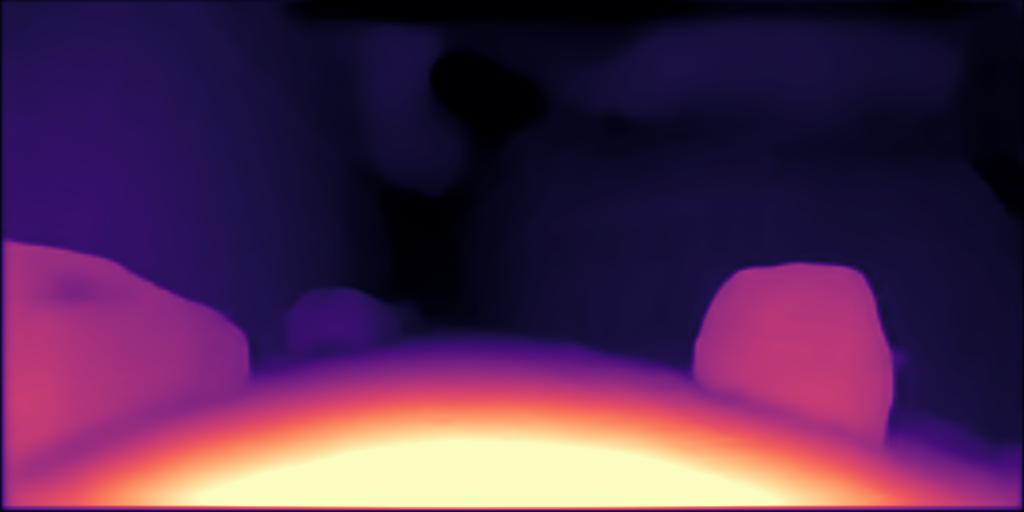}
    {} &
    \imlabel{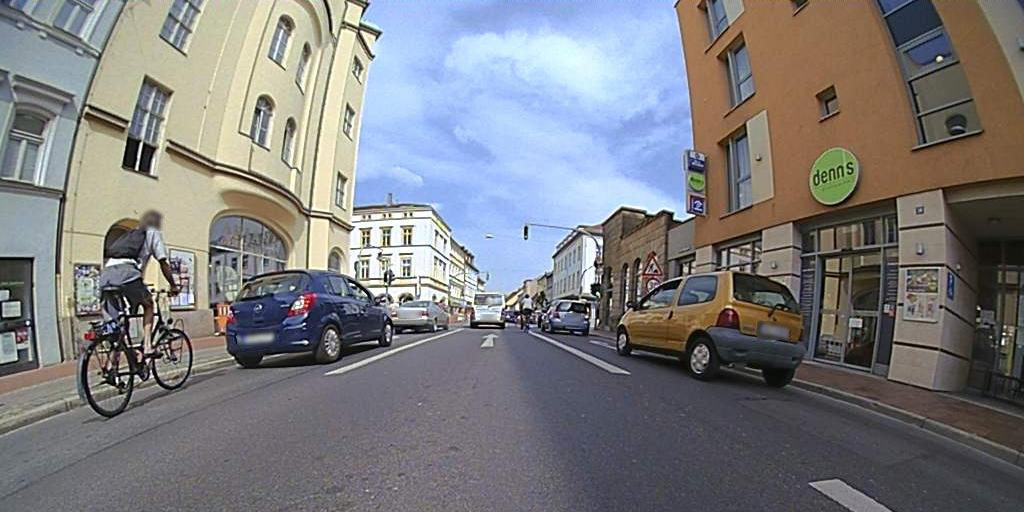}
    {} &
    \imlabel{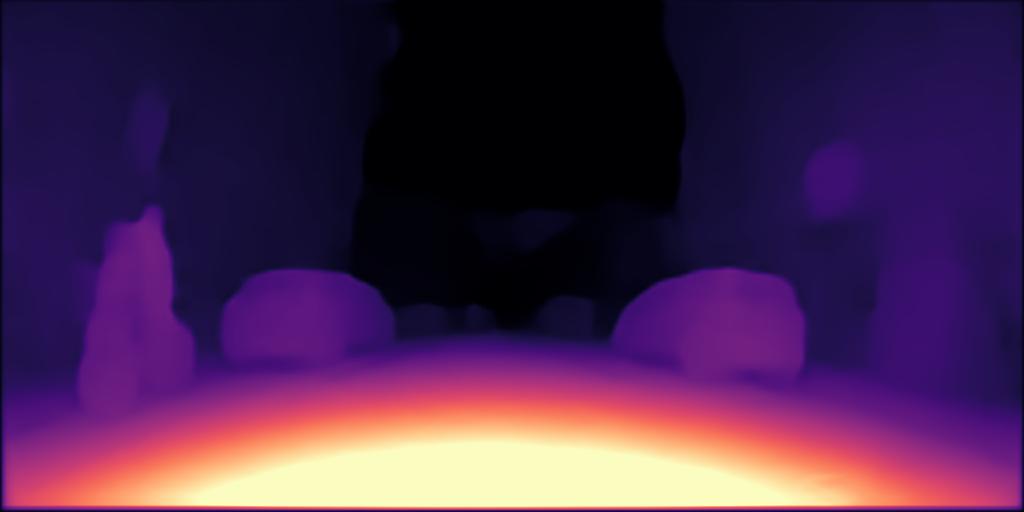}
    {} \\

    \imlabel{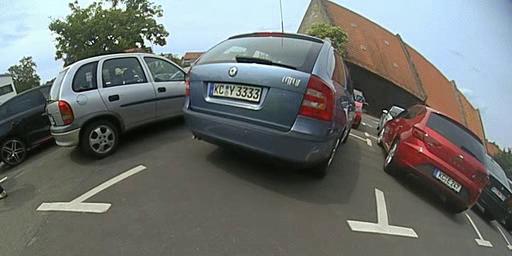}
    {\hspace{.24\textwidth}\textcolor{white}{Left Cam}} &
    \imlabel{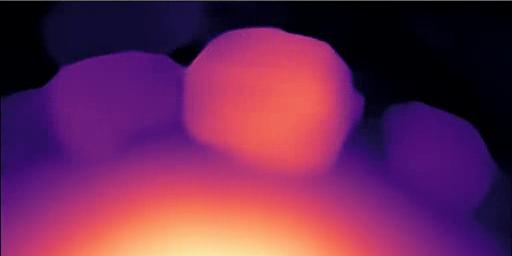}
    {} &
    \imlabel{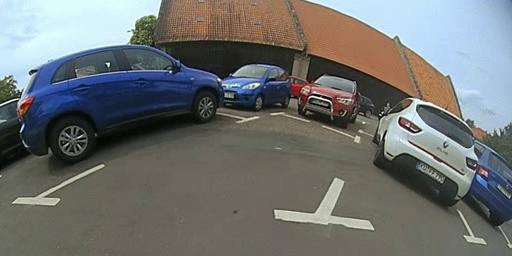}
    {} &
    \imlabel{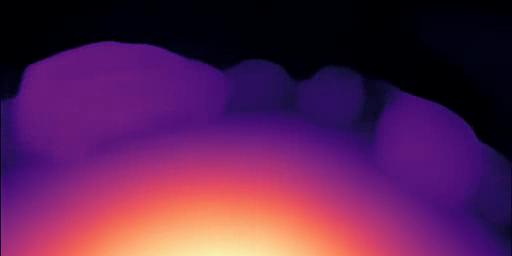}
    {} \\
    
    \imlabel{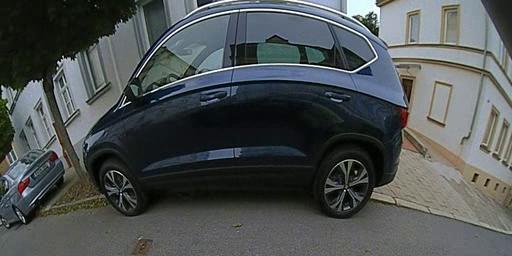}
    {\hspace{.24\textwidth}\textcolor{white}{Right Cam}} &
    \imlabel{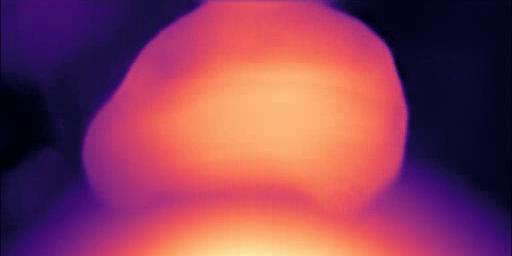}
    {} &
    \imlabel{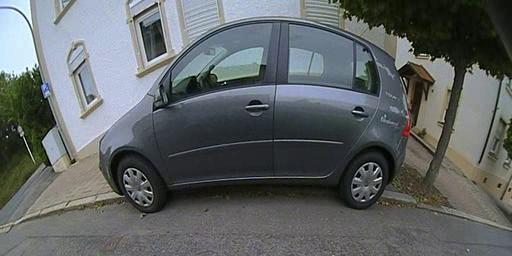}
    {} &
    \imlabel{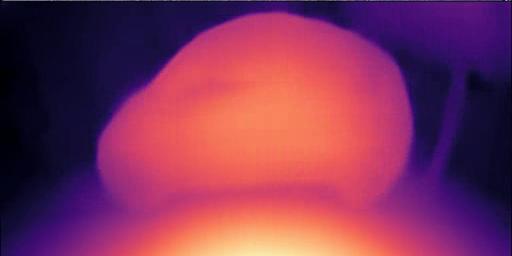}
    {} \\
    
    \imlabel{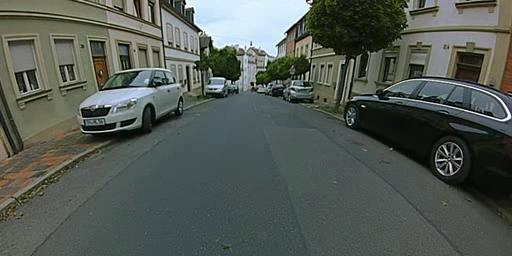}
    {\hspace{.24\textwidth}\textcolor{white}{Rear Cam}} &
    \imlabel{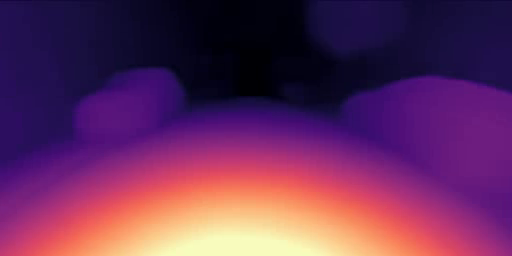}
    {} &
    \imlabel{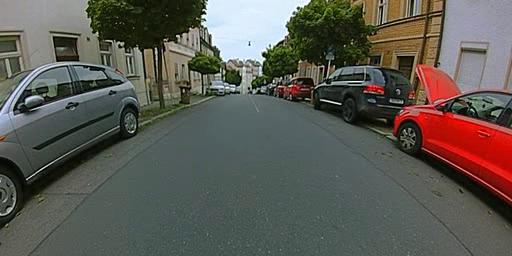}
    {} &
    \imlabel{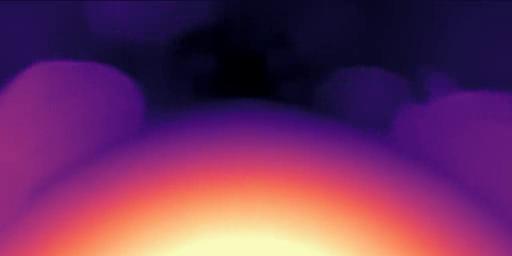}
    {} \\
\end{tabular}
}

}
\caption{\bf Distance estimation results on a surround-view camera system.}
  \label{fig:multi-cam}
\end{figure*}

\begin{figure}[htbp]
  \centering
  \newcommand{\turnheightnew}{0.5\columnwidth}
\centering
\begin{adjustbox}{width=\textwidth, totalheight=8.5in}
\begin{tabular}{@{\hskip 0.5mm}c@{\hskip 0.5mm}c@{\hskip 0.5mm}c@{}}

\includegraphics[height=\turnheightnew]{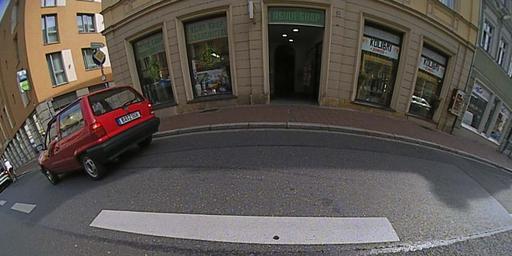} &
\includegraphics[height=\turnheightnew]{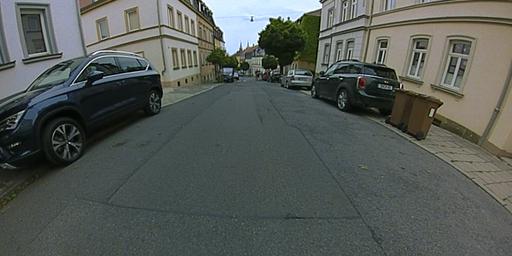} \\
\includegraphics[height=\turnheightnew]{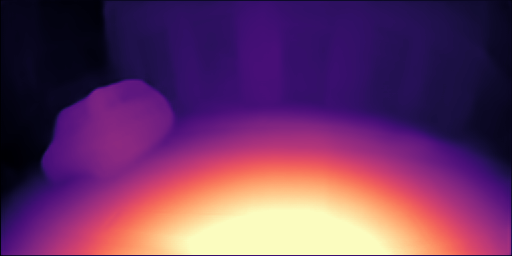} &
\includegraphics[height=\turnheightnew]{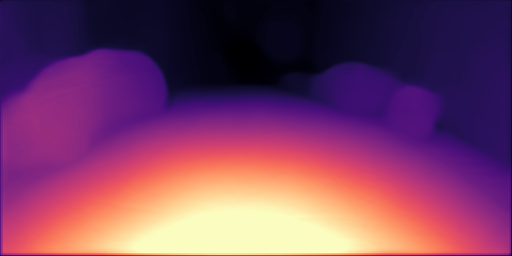} \\
\includegraphics[height=\turnheightnew]{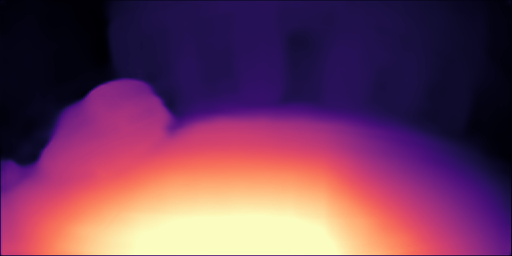} &
\includegraphics[height=\turnheightnew]{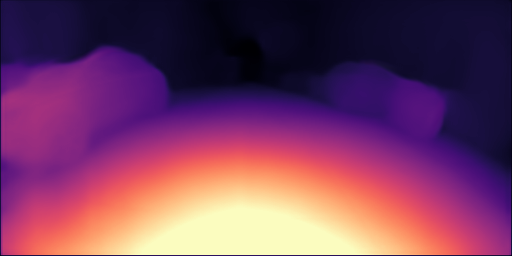} \\
\includegraphics[height=\turnheightnew]{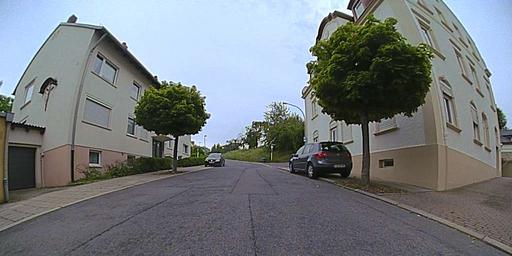} &
\includegraphics[height=\turnheightnew]{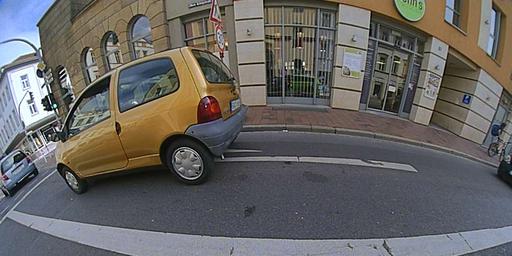} \\
\includegraphics[height=\turnheightnew]{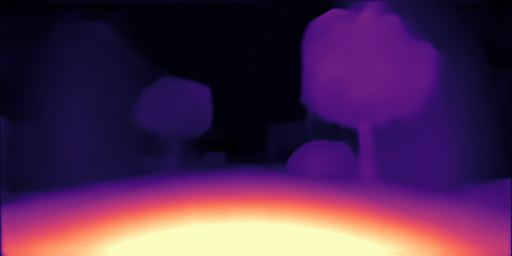} &
\includegraphics[height=\turnheightnew]{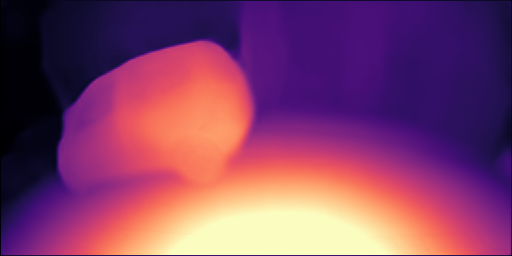} \\
\includegraphics[height=\turnheightnew]{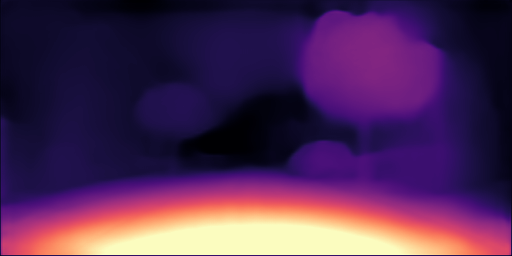} &
\includegraphics[height=\turnheightnew]{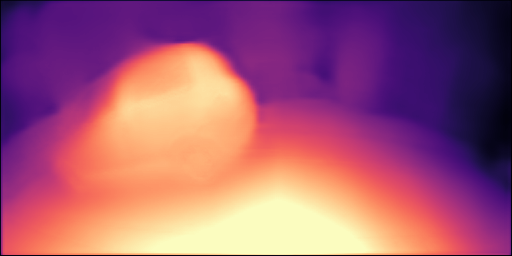} \\
\end{tabular}
\end{adjustbox}
  \captionsetup{skip=-1pt}
  \caption[\bf Qualitative results of SVDistNet on an unseen sequences.]
  {\textbf{Qualitative results of SVDistNet on an unseen sequences} from one of the test cars with a different camera intrinsic. The 1\textsuperscript{st} and 4\textsuperscript{th} row indicates the raw input images from the front and left camera. The 2\textsuperscript{nd}, 5\textsuperscript{th}, 3\textsuperscript{rd} and 6\textsuperscript{th} row indicate the distance estimates of the network trained with and without CGT respectively. Despite the notable variation in the camera parameters, the network with CGT outputs sharp distance maps on which edges are visible.}
  \label{fig:cam_tensor_qual}
\end{figure}

\begin{table}[t]
    \centering
    \small
    \begin{adjustbox}{width=0.8\columnwidth}
    \setlength{\tabcolsep}{0.2em}
      \begin{tabular}{lcccccccc}
	  \toprule
      \multirow{2}{*}{\textbf{Cams}}
      & \multirow{2}{*}{\emph{CGT}}
      & \cellcolor[HTML]{7d9ebf} Abs Rel
      & \cellcolor[HTML]{7d9ebf} Sq Rel
      & \cellcolor[HTML]{7d9ebf} RMSE 
      & \cellcolor[HTML]{7d9ebf} RMSE$_{log}$ 
      & \cellcolor[HTML]{e8715b} $\delta<1.25$ 
      & \cellcolor[HTML]{e8715b} $\delta<1.25^2$ 
      & \cellcolor[HTML]{e8715b} $\delta<1.25^3$ \\
      \cmidrule(lr){3-6} \cmidrule(lr){7-9}
      & & \multicolumn{4}{c}{\cellcolor[HTML]{7d9ebf} lower is better} & \multicolumn{3}{c}{\cellcolor[HTML]{e8715b} higher is better} \\
      \toprule
      \multirow{2}{*}{\begin{tabular}[c]{@{}c@{}} Front \end{tabular}}             
      & \xm & 0.074 & 0.343 & 1.641 & 0.112 & 0.892 & 0.985 & 0.994 \\
      & \ch & 0.057 & 0.315 & 1.532 & 0.108 & 0.896 & 0.987 & 0.996 \\
      \midrule
      \multirow{2}{*}{\begin{tabular}[c]{@{}c@{}} Rear \end{tabular}}              
      & \xm & 0.089 & 0.358 & 1.657 & 0.131 & 0.888 & 0.981 & 0.988 \\
      & \ch & 0.065 & 0.337 & 1.579 & 0.123 & 0.891 & 0.983 & 0.992 \\
      \midrule
      \multirow{2}{*}{\begin{tabular}[c]{@{}c@{}} Left \end{tabular}}  
      & \xm & 0.102 & 0.398 & 1.874 & 0.126 & 0.886 & 0.980 & 0.983 \\
      & \ch & 0.091 & 0.382 & 1.781 & 0.114 & 0.889 & 0.985 & 0.990 \\
      \midrule
      \multirow{2}{*}{\begin{tabular}[c]{@{}c@{}} Right \end{tabular}}  
      & \xm & 0.105 & 0.406 & 1.889 & 0.135 & 0.882 & 0.980 & 0.981 \\
      & \ch & 0.093 & 0.391 & 1.796 & 0.120 & 0.887 & 0.983 & 0.986 \\
      \bottomrule
  \end{tabular}
\end{adjustbox}
\caption[\bf Ablation study on multiple cameras concerning the usage of Camera Geometry Tensor using the WoodScape.]
{\textbf{Ablation study on multiple cameras} concerning the usage of Camera Geometry Tensor using the WoodScape.}
\label{table:multi-camera-ablation}
\end{table}
Table~\ref{table:multi-camera-ablation} depicts the generalization across different cameras from the surround-view setup using the CGT as described in Section~\ref{sec:camera-geometry-tensor}. The qualitative results of the model are shown in Figure~\ref{fig:multi-cam} where the same network is evaluated on four different fisheye cameras of a surround-view camera system. One can see that the \textit{SVDistNet} model generalizes well across different viewing angles and consistently produces high-quality distance outputs. The metrics of each camera significantly improve as during the model's training phase, sequences from different cameras help in generalization. For example, the front camera's distance estimates profit as the side cameras steer the network to generalize close and overlapping objects. We test the \textit{SVDistNet} model on an unseen sequence from one of the test cars whose cameras have different camera intrinsics than the ones used for training to examine the effect of the CGT as shown in Figure~\ref{fig:cam_tensor_qual}. Due to its usage, the network does not overfit to a particular camera intrinsic. It adapts to any changes from a family of unseen cameras deployed with a pre-calibrated camera setup. It leads to improved estimates and generalization of new cameras and allows training on images from different cameras.\par
\begin{table*}[!t]
\centering
\begin{adjustbox}{width=\columnwidth}
\small
\begin{tabular}{lcccccccc}
\toprule
\multirow{2}{*}{\textbf{Method}}  
& \multirow{2}{*}{$\rho$}
& \cellcolor[HTML]{7d9ebf} Abs Rel
& \cellcolor[HTML]{7d9ebf} Sq Rel
& \cellcolor[HTML]{7d9ebf} RMSE
& \cellcolor[HTML]{7d9ebf} RMSE$_{log}$
& \cellcolor[HTML]{e8715b} $\delta<1.25$
& \cellcolor[HTML]{e8715b} $\delta<1.25^2$
& \cellcolor[HTML]{e8715b} $\delta<1.25^3$ \\
\cmidrule(lr){3-6} \cmidrule(lr){7-9}
&
& \multicolumn{4}{c}{\cellcolor[HTML]{7d9ebf} lower is better} 
& \multicolumn{3}{c}{\cellcolor[HTML]{e8715b} higher is better}\\
\midrule
FisheyeDistanceNet~\cite{kumar2020fisheyedistancenet} 
& \xm & 0.152 & 0.768 & 2.723 & 0.210 & 0.812 & 0.954 & 0.974 \\
\hline
\multirow{4}{*}{\begin{tabular}[c]{@{}c@{}} SynDistNet \end{tabular}}       
& 1  & 0.148 & 0.642 & 2.615 & 0.203 & 0.824 & 0.960 & 0.978 \\
& 0  & 0.151 & 0.638 & 2.601 & 0.205 & 0.822 & 0.962 & 0.981 \\
& 2  & 0.154 & 0.631 & 2.532 & 0.198 & 0.832 & 0.965 & 0.981 \\
& $(0, 2)$  & \textbf{0.142} & \textbf{0.537} & \textbf{2.316} & \textbf{0.179} & \textbf{0.878} & \textbf{0.971} & \textbf{0.985} \\
\midrule
\multirow{2}{*}{\begin{tabular}[c]{@{}c@{}} SVDistNet \end{tabular}}       
& 2    & 0.145 & 0.564 & 2.279 & 0.186 & 0.863 & 0.966 & 0.982 \\
& 0, 2 & \textbf{0.128} & \textbf{0.469} & \textbf{2.153} & \textbf{0.164} & \textbf{0.875} & \textbf{0.974} & \textbf{0.986} \\
\bottomrule
\end{tabular}
\end{adjustbox}
\caption
{\bf Ablation study of the robust loss function of SynDistNet and SVDistNet on the WoodScape.}
\label{table:syndistnet-robust-ablation}
\end{table*}
\subsubsection{Robust loss function strategy}

Table~\ref{table:syndistnet-robust-ablation} ablates the usage of the robust loss function strategy on the frameworks \textit{SynDistNet} and \textit{SVDistNet}. We replace the \lone loss with several variants of the general loss function varying the parameter $\alpha$ and observe a significant performance improvement. The \lone loss is replaced with different variants of the robust general loss~\cite{barron2019general}, and we showcase that the usage of adaptive or annealed variants of the loss can significantly improve the performance.
The shape parameter $\rho$ is varied, keeping the scale fixed with a general distribution than a fixed Laplacian distribution. Instead of an RGB representation, following~\cite{barron2019general}, the YUV wavelet representations are used for the images, and the loss is applied to a YUV wavelet decomposition. The multi-scale training of the reconstruction loss described in Section~\ref{sec:fisheyedistancenet-final-loss} is dropped, which induces the sum of the loss means imposed at each scale in a $D$-level pyramid of side prediction since the robust loss function is a $D$ level normalized wavelet decomposition.
Compared to Barron \etal~\cite{barron2019general}, we retain the edge smoothness loss described in Section~\ref{sec:edge-smoothness} as it yielded better results. The fixed scale assumption is matched by setting the loss's scale $c$ fixed to $0.01$, which also roughly matches the shape of its \lone loss. 
For the fixed scale models in Table~\ref{table:syndistnet-robust-ablation}, we used a constant value for $\rho$. We observe an improvement in the performance, and there is no single value of $\rho$, which is optimal. 
In the adaptive $\rho \in (0, 2)$ variant, $\rho$ is made a free parameter and is allowed to be optimized along with the network weights during training. The adaptive plan of action outperforms the fixed strategies, which showcases the importance of allowing the model to regulate the robustness of its loss during training adaptively. A comparison of the adaptive model's performance with the fixed models indicates that no single set of $\alpha$ is optimal for all wavelet coefficients.\par
\begin{table}[t!]
\centering
  \small
    \setlength{\tabcolsep}{0.2em}
    \begin{tabular}{lcccc}
    \toprule
    \textbf{Method} 
    & \textit{Dataset} 
    & \multicolumn{1}{c}{\cellcolor[HTML]{7d9ebf}
    \begin{tabular}[c]{@{}c@{}} \textit{Network} \\ \textit{Resolution} \end{tabular}} &
    \multicolumn{1}{c}{
    \begin{tabular}[c]{@{}c@{}} \textit{Encoder} \\ \textit{head} \end{tabular}} &
    \multicolumn{1}{c}{\cellcolor[HTML]{e8715b}
    \begin{tabular}[c]{@{}c@{}} \textit{Inference} \\ \textit{(fps)} \end{tabular}} \\
    \toprule
    \multirow{3}{*}{\begin{tabular}[c]{@{}c@{}} FisheyeDistanceNet~\cite{kumar2020fisheyedistancenet}\end{tabular}}
      & K  & 640 x 192  & \multirow{3}{*}{\begin{tabular}[c]{@{}c@{}} ResNet-18\end{tabular}} & 84 \\
      & K  & 1024 x 320 & & 34 \\
      & WS & 512 x 256  & & 89 \\
    \midrule
    \multirow{3}{*}{\begin{tabular}[c]{@{}c@{}} UnrectDepthNet~\cite{kumar2020unrectdepthnet}\end{tabular}}
      & K  & 640 x 192  & \multirow{4}{*}{\begin{tabular}[c]{@{}c@{}} ResNet-50\end{tabular}} & 39 \\
      & K  & 1024 x 320 & & 16 \\
      & WS & 512 x 256  & & 42 \\
      & WS & 1024 x 512 & & 11 \\
    \midrule  
    \multirow{4}{*}{\begin{tabular}[c]{@{}c@{}} SynDistNet~\cite{kumar2020syndistnet}\end{tabular}}
      & K  & 640 x 192  & \multirow{3}{*}{\begin{tabular}[c]{@{}c@{}} ResNet-18\end{tabular}} & 82 \\
      & K  & 1024 x 320 & & 33 \\
      & WS & 512 x 256  & & 91 \\
      & WS & 1024 x 512 & & 23 \\
      & WS & 512 x 256  & ResNet50 & 42 \\
      \midrule
    \multirow{4}{*}{\begin{tabular}[c]{@{}c@{}} SVDistNet \end{tabular}}
      & K  & 640 x 192  & \multirow{2}{*}{\begin{tabular}[c]{@{}c@{}} SAN-10\end{tabular}} & 80 \\
      & K  & 1024 x 320 & & 30 \\
      & WS & 512 x 256  & & 87 \\
      & WS & 512 x 256  & SAN-19 & 45 \\
    \bottomrule
\end{tabular}
\caption[\bf Ablation study on inference time using ONNX 16-bit float precision models.]
{\textbf{Ablation study on inference time (frames per second)} using ONNX 16-bit float precision models on NVIDIA's Jetson AGX on the KITTI (K) and the WoodScape (WS) datasets.}
\label{table:inference-ablation}
\end{table} 
\begin{table}[t!]
\centering
\small
\begin{adjustbox}{width=0.75\columnwidth}
\setlength{\tabcolsep}{0.1em}
  \begin{tabular}{lccccccc}
	\toprule
    \multirow{2}{*}{\textbf{Method}} 
    & \cellcolor[HTML]{7d9ebf} Abs Rel
    & \cellcolor[HTML]{7d9ebf} Sq Rel 
    & \cellcolor[HTML]{7d9ebf} RMSE
    & \cellcolor[HTML]{7d9ebf} RMSE$_{log}$ 
    & \cellcolor[HTML]{e8715b} $\delta<1.25$ 
    & \cellcolor[HTML]{e8715b} $\delta<1.25^2$ 
    & \cellcolor[HTML]{e8715b} $\delta<1.25^3$ \\
    \cmidrule(lr){2-5} \cmidrule(lr){6-8}
    & \multicolumn{4}{c}{\cellcolor[HTML]{7d9ebf} lower is better}
    & \multicolumn{3}{c}{\cellcolor[HTML]{e8715b} higher is better} \\
    \toprule
     SAN10-patch & 0.044 & 0.302 & 1.274 & 0.097 & 0.906 & 0.987 & 0.995 \\
     SAN10-pair  & 0.069 & 0.353 & 1.543 & 0.111 & 0.901 & 0.985 & 0.993 \\
     SAN19-patch & 0.034 & 0.264 & 1.218 & 0.073 & 0.914 & 0.992 & 0.996 \\
     SAN19-pair  & 0.065 & 0.322 & 1.545 & 0.092 & 0.896 & 0.988 & 0.995 \\
    \bottomrule
  \end{tabular}
\end{adjustbox}
\caption[\bf Online refinement of SVDistNet distance estimates.]
{\textbf{Online refinement} of the network's distance estimates incorporating~\cite{casser2019depth}, where the model is trained during inference on the WoodScape dataset.}
\label{table:svdistnet-online-refine}
\end{table}
\subsubsection{Online Refinement and Run-Time Comparison}

In Table~\ref{table:inference-ablation}, we report the frame rate of all previous approaches and \textit{SVDistNet} using different input resolutions and a 16-bit float precision ONNX model on NVIDIA's Jetson AGX platform, operating in full power mode. All the models, including the one with the highest resolution, are real-time capable and can be deployed in a car. The primary advantage of a single-frame distance estimator is its broad applicability. Despite this, it comes at a cost when the continuous estimation of distances on image frames is misaligned or discontinuous, which is often the case. To overcome this issue, we follow the approach from~\cite{casser2019depth, Chen2019b, shu2020featdepth}, where we adapt the model in an online manner, mainly for practical autonomous systems. We train the model during inference by setting the batch size to 1. We feed in the inference image and its two adjacent frames, where we carry out the refinement as described in~\cite{casser2019depth}. We do not implement any data augmentation techniques during this phase. With this technique, using a minimal temporal chain of frames (\ie, three-frame snippets), the distance estimates enhance significantly, qualitatively, and quantitatively, as shown in Table~\ref{table:svdistnet-online-refine}. With a single frame's negligible delay, the \textit{SVDistNet} framework can operate in real-time, even when using online refinement.\par
\subsection{Ablative Experiments on KITTI}

\subsubsection{Pose Estimation Results}

\begin{table}[!t]
\centering
\begin{adjustbox}{width=0.75\columnwidth}
\begin{tabular}{@{}lcccc@{}}
\toprule
\multicolumn{1}{l}{\textbf{Method}} 
& \textit{\begin{tabular}[c]{@{}c@{}} No. of \\ Frames\end{tabular}} 
& \textit{GT} 
& \cellcolor[HTML]{7d9ebf} \textit{Sequence 09} 
& \cellcolor[HTML]{e8715b} \textit{Sequence 10} \\ 
\midrule
    ORB-SLAM~\cite{mur2015orb}                   & 5 & \ch & 0.014 $\pm$ 0.008 & 0.012 $\pm$ 0.011 \\
    DF-Net~\cite{zou2018df}                      & 5 & \ch & 0.017 $\pm$ 0.007 & 0.015 $\pm$ 0.009 \\
    SfMLearner~\cite{zhou2017unsupervised}       & 5 & \ch & 0.016 $\pm$ 0.009 & 0.013 $\pm$ 0.009 \\
    Klodt et al.~\cite{klodt2018supervising}     & 5 & \ch & 0.014 $\pm$ 0.007 & 0.013 $\pm$ 0.009 \\
    GeoNet~\cite{yin2018geonet}                  & 5 & \ch & 0.012 $\pm$ 0.007 & 0.012 $\pm$ 0.009 \\
    Struct2Depth~\cite{casser2019depth}          & 5 & \ch & 0.011 $\pm$ 0.006 & 0.011 $\pm$ 0.010 \\
    Ranjan~\cite{ranjan2019competitive}          & 5 & \ch & 0.011 $\pm$ 0.006 & 0.011 $\pm$ 0.010 \\ 
    PackNet-SfM~\cite{guizilini2019packnet}      & 5 & \ch & 0.010 $\pm$ 0.005 & 0.009 $\pm$ 0.008 \\
    PackNet-SfM~\cite{guizilini2019packnet}      & 5 & \xm & 0.014 $\pm$ 0.007 & 0.012 $\pm$ 0.008 \\
    SVDistNet                                    & 5 & \ch & \textbf{0.009} $\pm$ \textbf{0.004} & 0.008 $\pm$ \textbf{0.005} \\
    SVDistNet                                    & 5 & \xm & 0.010 $\pm$ 0.005 & 0.010 $\pm$ 0.008 \\
    \midrule
    DDVO~\cite{Wang_2018_CVPR}                   & 3 & \ch & 0.045 $\pm$ 0.108 & 0.033 $\pm$ 0.074 \\
    Vid2Depth~\cite{mahjourian2018unsupervised}  & 3 & \ch & 0.013 $\pm$ 0.010 & 0.012 $\pm$ 0.011 \\
    EPC++~\cite{luo2019every}                    & 3 & \ch & 0.013 $\pm$ 0.007 & 0.012 $\pm$ 0.008 \\
    SVDistNet                                    & 3 & \ch & \textbf{0.011} $\pm$ \textbf{0.006} & \textbf{0.010} $\pm$ \textbf{0.007} \\
    SVDistNet                                    & 3 & \xm & 0.012 $\pm$ 0.007 & 0.011 $\pm$ 0.008 \\
    \midrule
    Monodepth2~\cite{godard2019digging}          & 2 & \ch & 0.017 $\pm$ 0.008 & 0.015 $\pm$ 0.010 \\
    SVDistNet                                    & 2 & \ch & \textbf{0.015} $\pm$ \textbf{0.007} & \textbf{0.013} $\pm$ \textbf{0.007} \\
    SVDistNet                                    & 2 & \xm & 0.016 $\pm$ 0.008 & 0.014 $\pm$ 0.009 \\
    \bottomrule
    \end{tabular}
  \end{adjustbox}
\caption[\bf Evaluation of the pose estimation in SVDistNet on the KITTI Odometry benchmark.]
{\textbf{Evaluation of the pose estimation} on the KITTI Odometry Benchmark~\cite{geiger2013vision}.}
\label{table:svdistnet-pose-ate}
\end{table}
The PoseNet is an ego-motion predictor consisting of a SAN10-patch encoder. We apply the Siamese (twin network) notion where we feed $I_t$ and $I_t'$ individually to a shared self-attention encoder and concatenate the output features from the twin network before feeding it to the pose decoder as shown in Figure~\ref{fig:svdistnet-mtl_pipeline} predicting a relative pose between the images. Compared to the previous works~\cite{kumar2020fisheyedistancenet, kumar2020syndistnet, kumar2020unrectdepthnet}, where we used Euler angles, we chose quaternions to represent the 3D rotation. The design choice is mainly due to its continuous and smooth representation of rotation and smaller memory footprint than rotation matrices. Also, quaternions are much more efficient than both matrix, and angle/axis representations used in~\cite{godard2019digging, shu2020featdepth}.\par

In Table~\ref{table:svdistnet-pose-ate}, we report the average trajectory error (ATE) in meters, where we train the method on sequences 00-08 and evaluate on Sequences $09$ and $10$, same as for the baseline methods. For evaluation, we follow the evaluation protocol defined in~\cite{zhou2017unsupervised}. Note that all the methods except the \textit{SVDistNet} and PackNet-SfM utilize ground-truth at test-time to scale the prediction for a scale consistent result. We predict independent transformations for each of the four frame-to-frame transformations belonging to the five-frames set to evaluate the two-frame ego-motion model on the five-frame test sequences. We combine these different transformations to form local trajectories. We outperform the previous methods listed in Table~\ref{table:svdistnet-pose-ate}, mainly by applying the bundle adjustment framework inflicted by the cross-sequence distance consistency loss~\cite{kumar2020fisheyedistancenet}. It induces more constraints and simultaneously optimizes distances and camera pose for an implicitly extended training input sequence. This provides additional consistency constraints that are not induced by previous methods.\par
\subsubsection{State-of-the-Art Comparison on KITTI}

\begin{figure*}[!t]
  \captionsetup{skip=-1.2pt}
  \centering
  \newcommand{\turnheightnew}{0.23\columnwidth}
\centering
\begin{adjustbox}{width=\textwidth, totalheight=9.2in}
\begin{tabular}{@{\hskip 0.5mm}c@{\hskip 0.5mm}c@{\hskip 0.5mm}c@{\hskip 0.5mm}}

{\rotatebox{90}{\hspace{6mm}\large Raw Input}} &
\includegraphics[height=\turnheightnew]{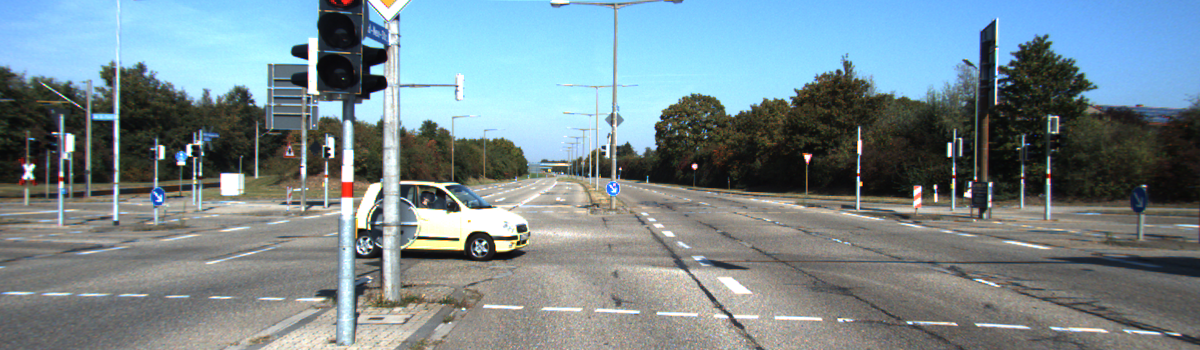} &
\includegraphics[height=\turnheightnew]{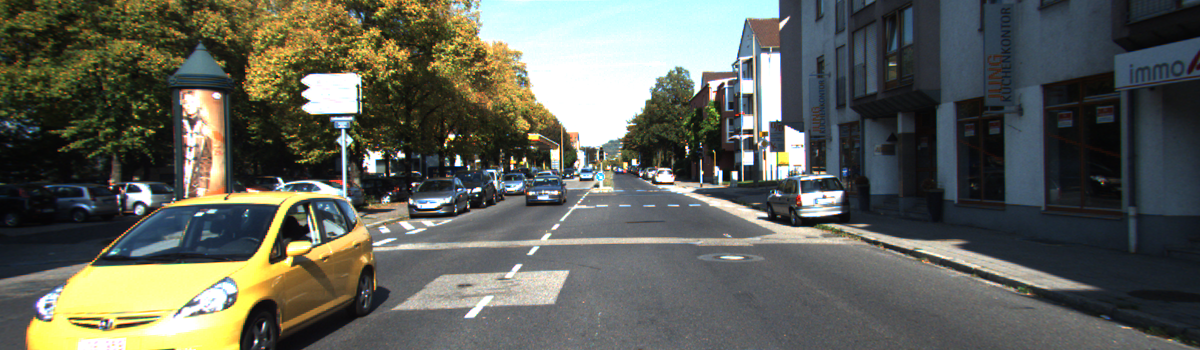} \\

{\rotatebox{90}{\hspace{6mm}\large SynDistNet}} &
\includegraphics[height=\turnheightnew]{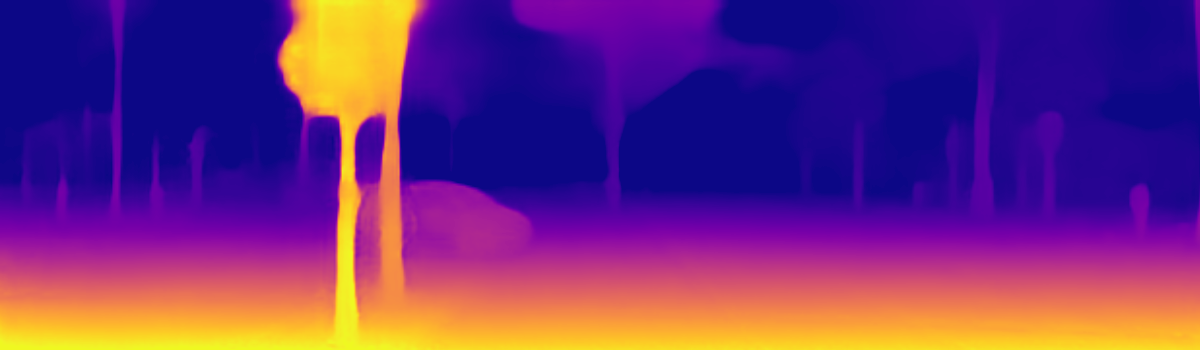} &
\includegraphics[height=\turnheightnew]{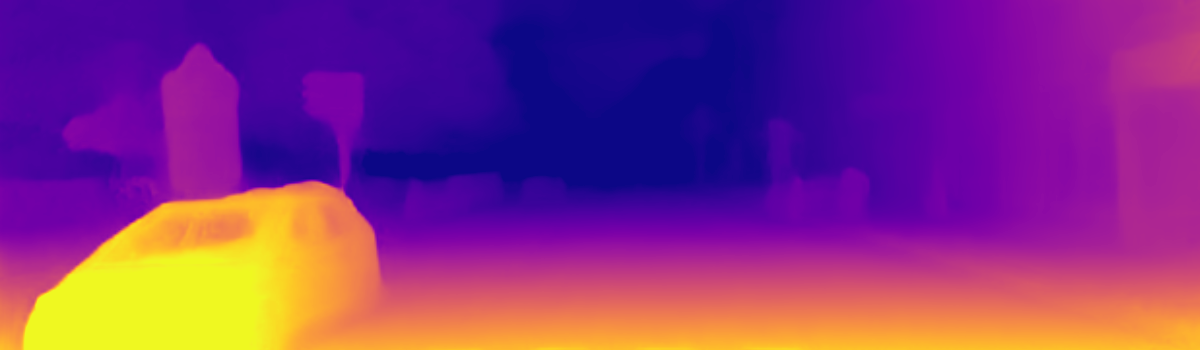} \\

{\rotatebox{90}{\hspace{3mm}\large SynDistNet}} &
\includegraphics[height=\turnheightnew]{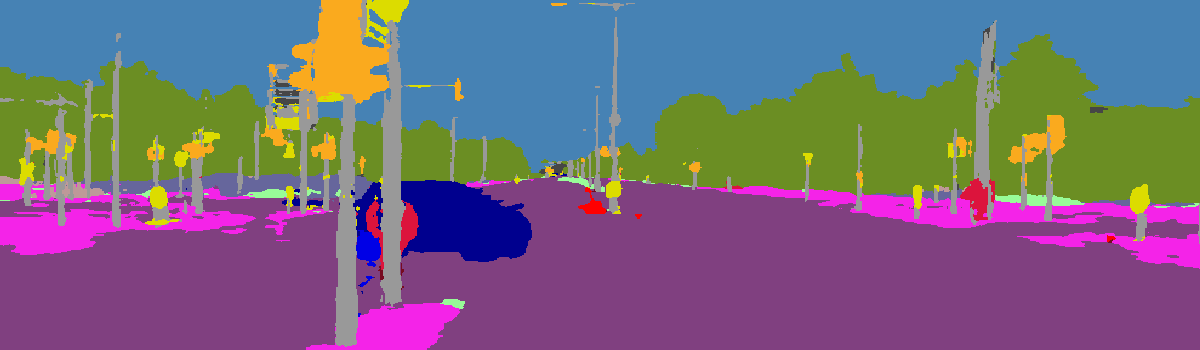} &
\includegraphics[height=\turnheightnew]{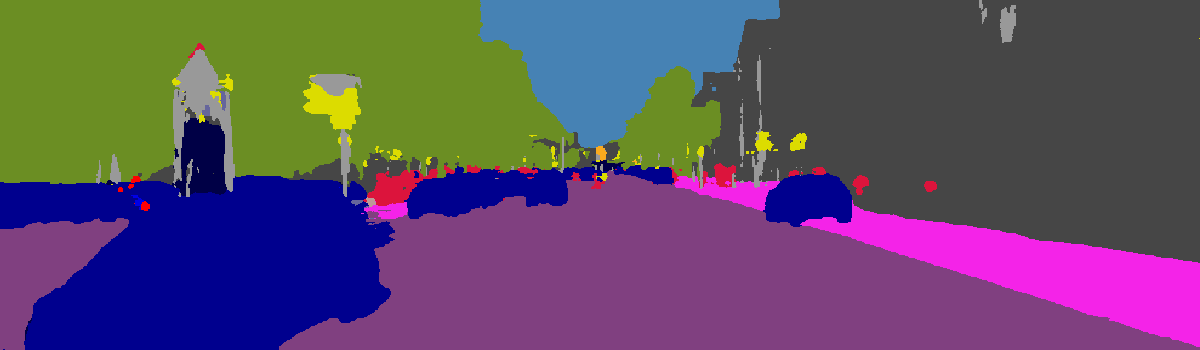} \\

{\rotatebox{90}{\hspace{6mm}\large Raw Input}} &
\includegraphics[height=\turnheightnew]{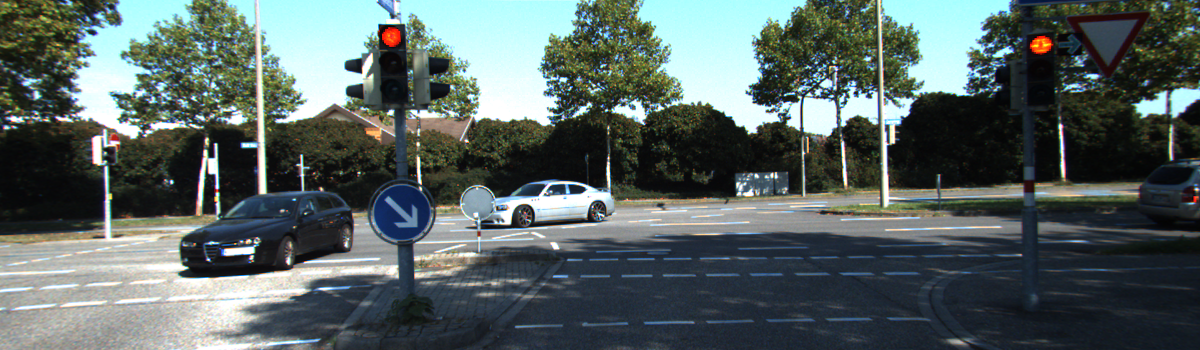} &
\includegraphics[height=\turnheightnew]{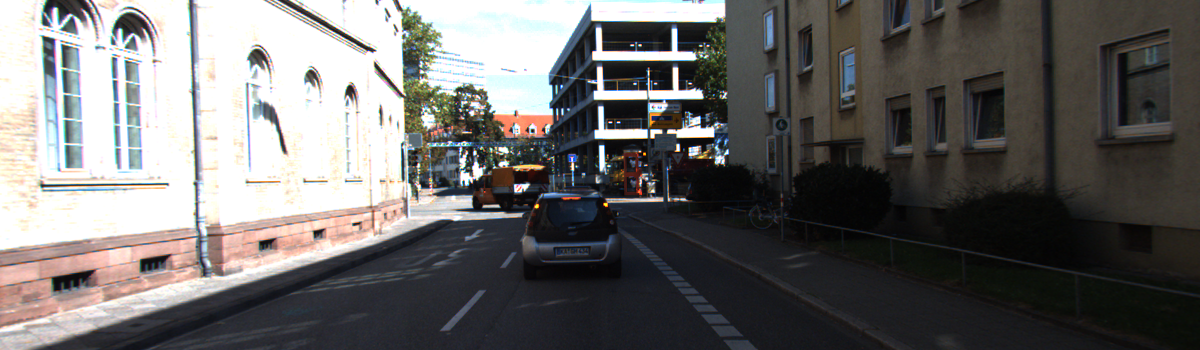}\\

{\rotatebox{90}{\hspace{6mm}\large SynDistNet}} &
\includegraphics[height=\turnheightnew]{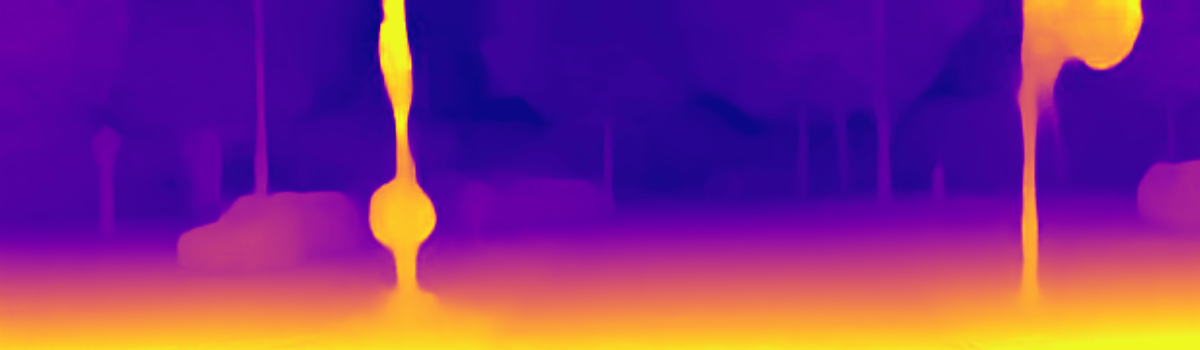} &
\includegraphics[height=\turnheightnew]{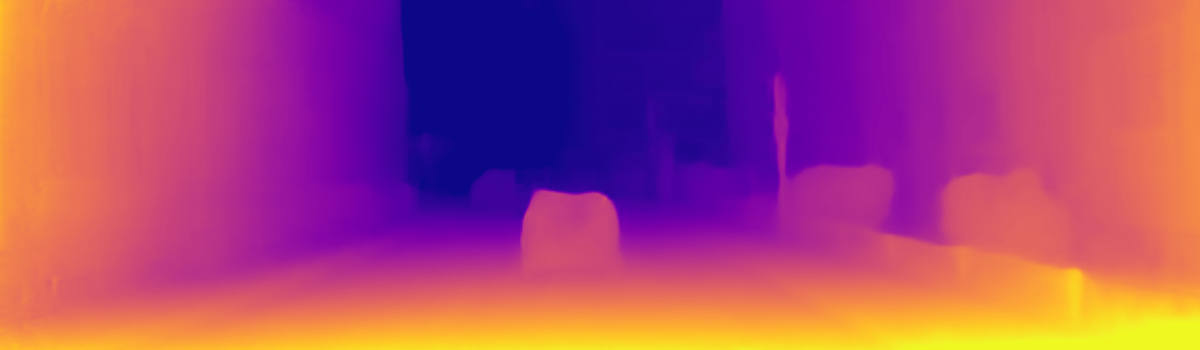}\\

{\rotatebox{90}{\hspace{6mm}\large SynDistNet}} &
\includegraphics[height=\turnheightnew]{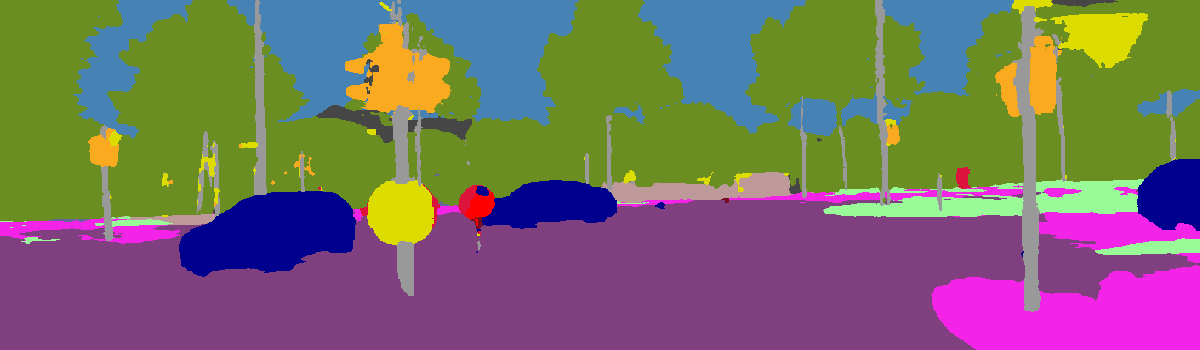} &
\includegraphics[height=\turnheightnew]{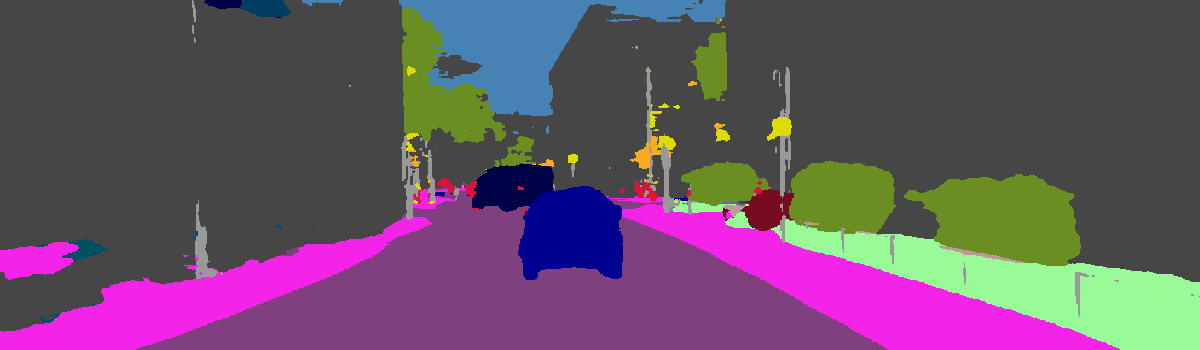} \\

{\rotatebox{90}{\hspace{6mm}\large Raw Input}} &
\includegraphics[height=\turnheightnew]{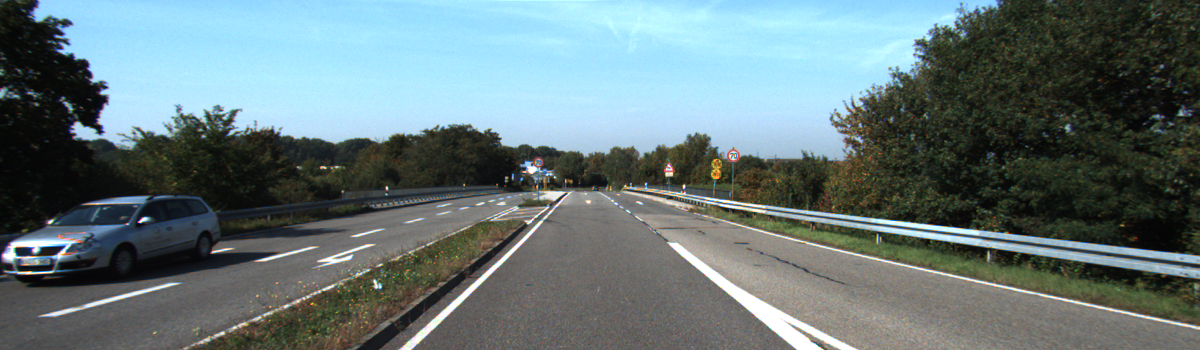} &
\includegraphics[height=\turnheightnew]{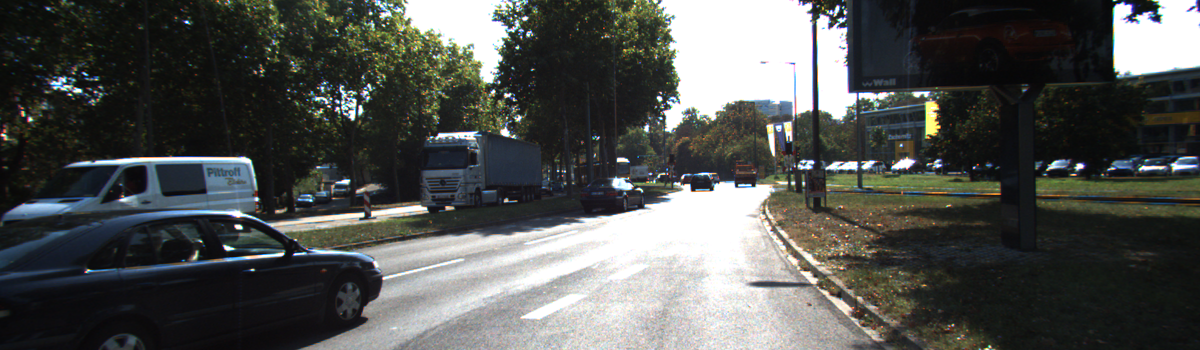} \\

{\rotatebox{90}{\hspace{6mm}\large SynDistNet}} &
\includegraphics[height=\turnheightnew]{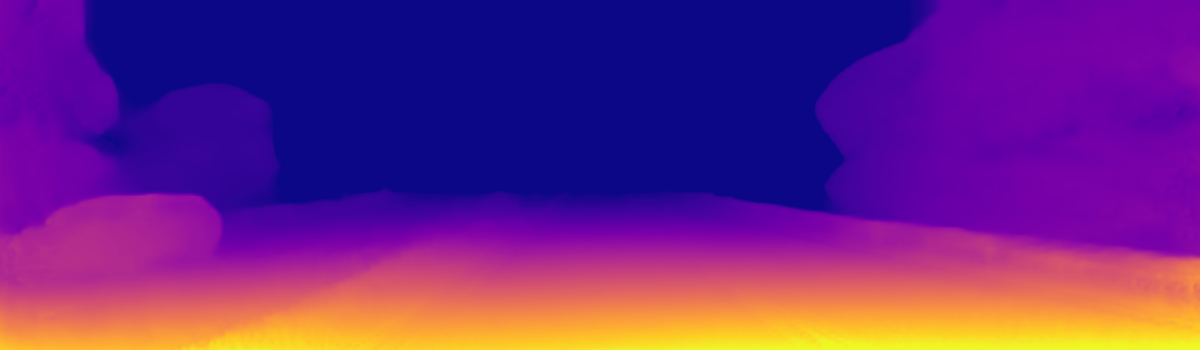} &
\includegraphics[height=\turnheightnew]{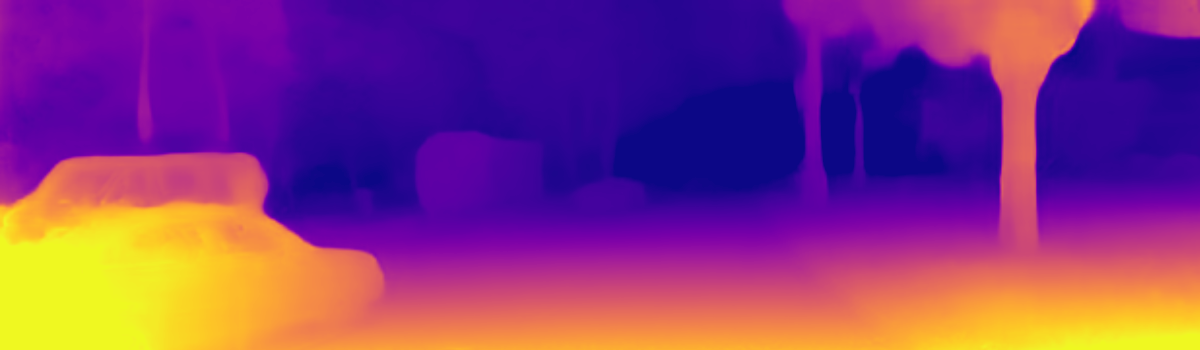} \\

{\rotatebox{90}{\hspace{6mm}\large SynDistNet}} &
\includegraphics[height=\turnheightnew]{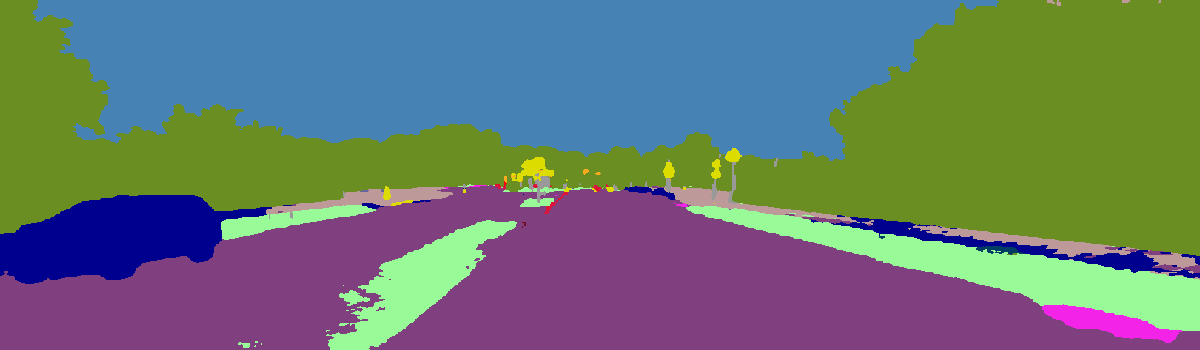} &
\includegraphics[height=\turnheightnew]{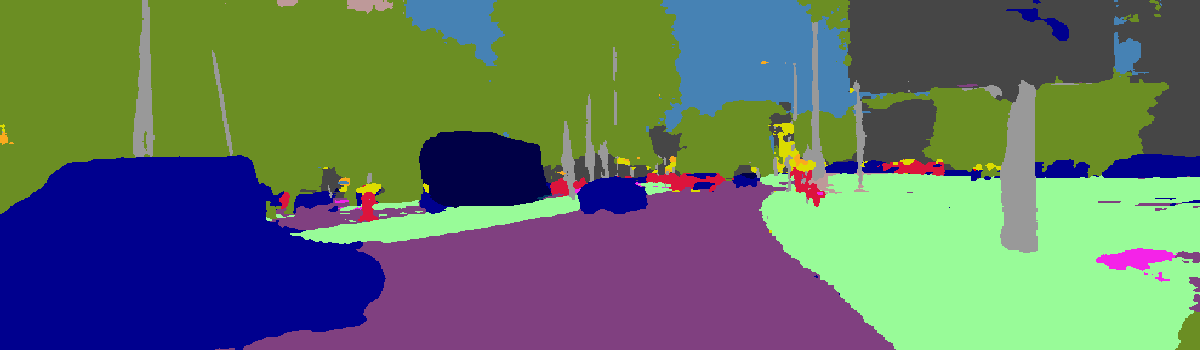} \\

{\rotatebox{90}{\hspace{6mm}\large Raw Input}} &
\includegraphics[height=\turnheightnew]{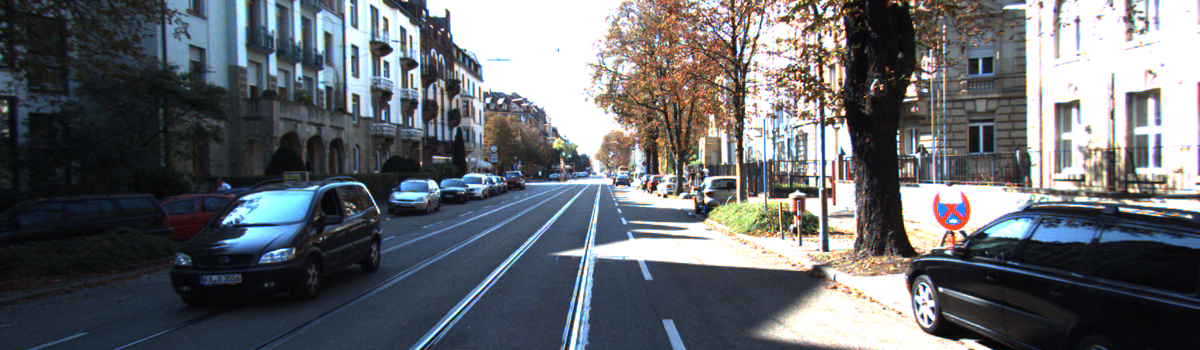} &
\includegraphics[height=\turnheightnew]{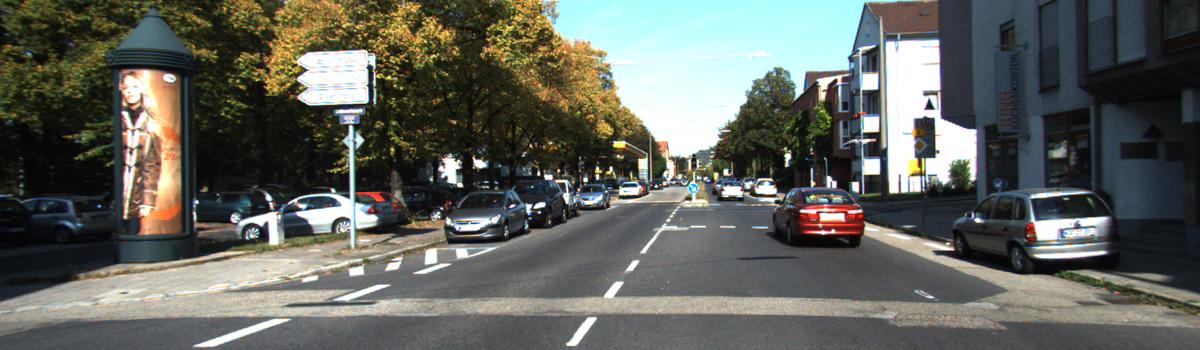} \\

{\rotatebox{90}{\hspace{6mm}\large SynDistNet}} &
\includegraphics[height=\turnheightnew]{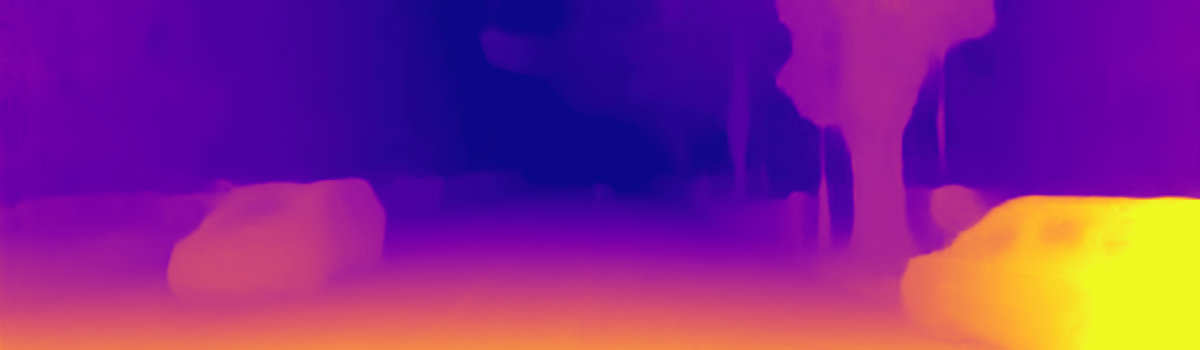} &
\includegraphics[height=\turnheightnew]{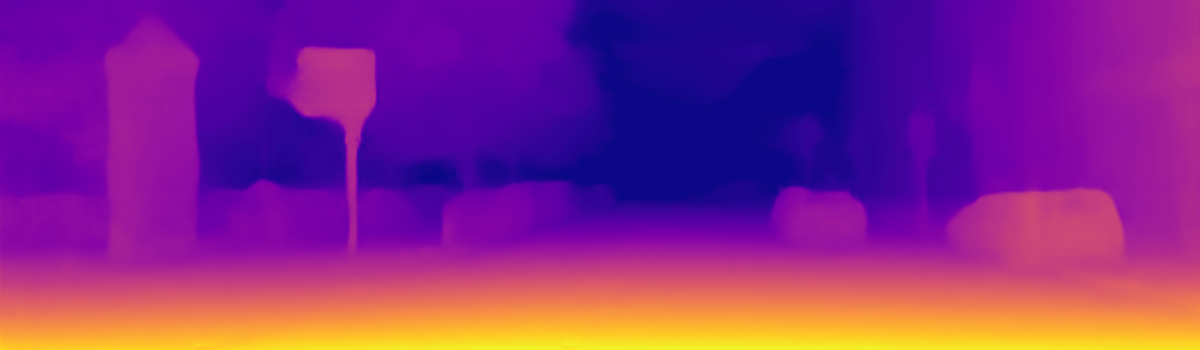} \\

{\rotatebox{90}{\hspace{6mm}\large SynDistNet}} &
\includegraphics[height=\turnheightnew]{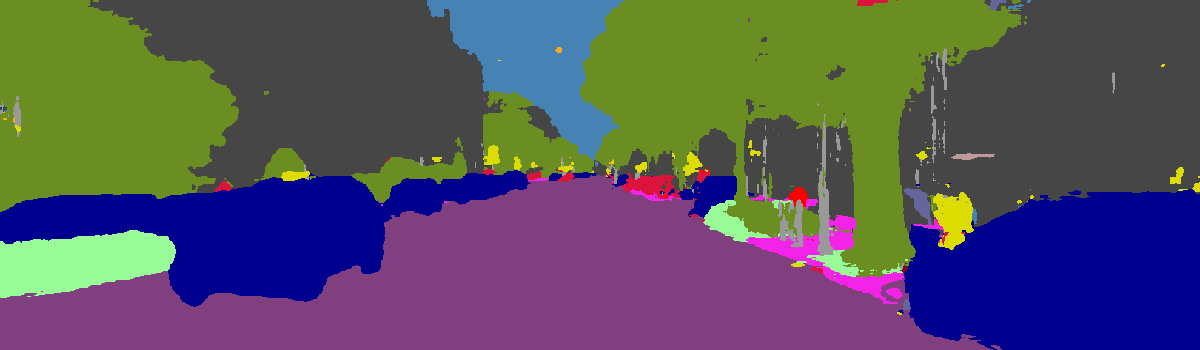} &
\includegraphics[height=\turnheightnew]{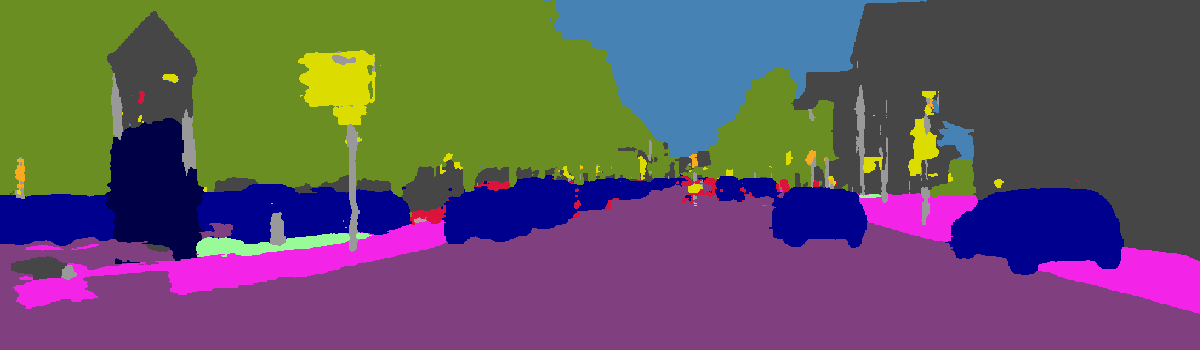} \\

\end{tabular}
\end{adjustbox}
  \caption[\bf Qualitative results of SynDistNet on the KITTI.]
          {\textbf{Qualitative results of SynDistNet on the KITTI.} We showcase depth estimation as well as semantic segmentation outputs.}
  \label{fig:KITTIMTLResults}
\end{figure*}

\begin{table*}[!t]
\centering
{
\small
\setlength{\tabcolsep}{0.3em}
\begin{adjustbox}{width=\columnwidth}
\begin{tabular}{c|lcccccccc}
\toprule
& \textbf{Method} 
& Resolution  
& \cellcolor[HTML]{7d9ebf} Abs Rel 
& \cellcolor[HTML]{7d9ebf} Sq Rel 
& \cellcolor[HTML]{7d9ebf} RMSE 
& \cellcolor[HTML]{7d9ebf} RMSE$_{log}$ 
& \cellcolor[HTML]{e8715b} $\delta<1.25$ 
& \cellcolor[HTML]{e8715b} $\delta<1.25^2$ 
& \cellcolor[HTML]{e8715b} $\delta<1.25^3$ \\
\cmidrule(lr){4-7} \cmidrule(lr){8-10} & & 
& \multicolumn{4}{c}{\cellcolor[HTML]{7d9ebf} lower is better}
& \multicolumn{3}{c}{\cellcolor[HTML]{e8715b} higher is better} \\
\midrule
  \multicolumn{10}{c}{\cellcolor[HTML]{448BE9}\textit{KITTI}}  \\
\midrule
\parbox[t]{2mm}{\multirow{10}{*}{\rotatebox[origin=c]{90}{Original~\cite{Eigen_14}}}}
& EPC++~\cite{luo2019every}                   & 640 x 192 & 0.141 & 1.029 & 5.350 & 0.216 & 0.816 & 0.941 & 0.976 \\
& Monodepth2~\cite{godard2019digging}         & 640 x 192 & 0.115 & 0.903 & 4.863 & 0.193 & 0.877 & 0.959 & 0.981 \\
& PackNet-SfM~\cite{guizilini2019packnet}     & 640 x 192 & 0.111 & 0.829 & 4.788 & 0.199 & 0.864 & 0.954 & 0.980 \\
& FisheyeDistanceNet~\cite{kumar2020fisheyedistancenet} & 640 x 192 & 0.117 & 0.867 & 4.739 & 0.190 & 0.869 & 0.960 & 0.982 \\
& UnRectDepthNet~\cite{kumar2020unrectdepthnet} & 640 x 192 & \textbf{0.107} & 0.721 & 4.564 & \textbf{0.178} & 0.894 & 0.971 & \textbf{0.986} \\
& \textbf{SynDistNet}                         & 640 x 192 & 0.109 & \textbf{0.718} & \textbf{4.516} & 0.180 & \textbf{0.896} & \textbf{0.973} & \textbf{0.986} \\
\cmidrule{2-10} 
& Monodepth2~\cite{godard2019digging}          & 1024 x 320 & 0.115 & 0.882 & 4.701 & 0.190 & 0.879 & 0.961 & 0.982 \\
& FisheyeDistanceNet~\cite{kumar2020fisheyedistancenet} & 1024 x 320 & 0.109 & 0.788 & 4.669 & 0.185 & 0.889 & 0.964 & 0.982 \\
& UnRectDepthNet~\cite{kumar2020unrectdepthnet} & 1024 x 320 & 0.103 & 0.705 & 4.386 & \textbf{0.164 }& 0.897 & \textbf{0.980} & 0.989 \\
& \textbf{SynDistNet}                         & 1024 x 320   & \textbf{0.102} & \textbf{0.701} & \textbf{4.347} & 0.166 & \textbf{0.901} & \textbf{0.980} & \textbf{0.990} \\
\midrule
\parbox[t]{2mm}{\multirow{8}{*}{\rotatebox[origin=c]{90}{Improved~\cite{uhrig2017sparsity}}}}
& SfMLeaner~\cite{zhou2017unsupervised}       & 416 x 128  & 0.176 & 1.532 & 6.129 & 0.244 & 0.758 & 0.921 & 0.971 \\
& Vid2Depth~\cite{mahjourian2018unsupervised} & 416 x 128  & 0.134 & 0.983 & 5.501 & 0.203 & 0.827 & 0.944 & 0.981 \\
& DDVO~\cite{Wang_2018_CVPR}                & 416 x 128  & 0.126 & 0.866 & 4.932 & 0.185 & 0.851 & 0.958 & 0.986 \\
\cmidrule{2-10}
& EPC++~\cite{luo2019every}                 & 640 x 192  & 0.120 & 0.789 & 4.755 & 0.177 & 0.856 & 0.961 & 0.987 \\
& Monodepth2~\cite{godard2019digging}       & 640 x 192  & 0.090 & 0.545 & 3.942 & 0.137 & 0.914 & 0.983 & 0.995 \\
& PackNet-SfM~\cite{guizilini2019packnet}   & 640 x 192  & 0.078 & 0.420 & 3.485 & 0.121 & \textbf{0.931} & 0.986 & \textbf{0.996} \\
& UnRectDepthNet~\cite{kumar2020unrectdepthnet} & 640 x 192 & 0.081 & 0.414 & 3.412 & 0.117 & 0.926 & 0.987 & \textbf{0.996} \\
& \textbf{SynDistNet}                       & 640 x 192 & \textbf{0.076} & \textbf{0.412} & \textbf{3.406} & \textbf{0.115} & \textbf{0.931} & \textbf{0.988} & \textbf{0.996} \\
\bottomrule
\end{tabular}
\end{adjustbox}
}
\caption[\bf Quantitative performance comparison of distance estimates in SynDistNet.]
{\textbf{Quantitative performance comparison of SynDistNet with other self-supervised monocular methods for depths up to 80\,m for the KITTI.} \textit{Original} uses raw depth maps as proposed by \cite{Eigen_14} for evaluation, and \textit{Improved} uses annotated depth maps from \cite{uhrig2017sparsity}.}
\label{tab:syndistnet-results}
\end{table*}
As there is little work on fisheye distance estimation, we evaluate the method on the extensively used KITTI dataset using the metrics illustrated in Section~\ref{tab:depth-metrics} by Eigen \etal~\cite{Eigen_14}. The quantitative results are shown in the Table~\ref{tab:syndistnet-results} illustrates that the improved scale-aware self-supervised approach outperforms all the state-of-the-art monocular approaches. Figure~\ref{fig:KITTIMTLResults} provides qualitative results of \textit{SynDistNet} on the KITTI test dataset for the segmentation and distance estimation tasks. More specifically, we improve the baseline \emph{FisheyeDistanceNet} with the usage of a general and adaptive loss function~\cite{barron2019general} which is showcased in Table~\ref{table:syndistnet-robust-ablation} and better architecture. Compared to PackNet-SfM~\cite{guizilini2019packnet}, which presumably uses a superior architecture than the ResNet18, where they estimate scale-aware depths with their velocity supervision loss using the ground truth poses for supervision, we only rely on speed and time data captured from the vehicle odometry, which is easier to obtain. The approach can be easily transferred to the domain of aerial robotics as well. We could achieve higher accuracy than PackNet, which can be seen in Table~\ref{tab:syndistnet-results}. At test-time, all methods excluding \textit{FisheyeDistanceNet}, PackNet-SfM, and \textit{SynDistNet} scale the estimated depths using median ground-truth LiDAR depth.\par

\begin{figure*}[!t]
  \resizebox{\textwidth}{!}{\newcommand{\turnheightnew}{0.12\columnwidth}
\centering

\begin{tabular}{@{\hskip 0.5mm}c@{\hskip 0.5mm}c@{\hskip 0.5mm}c@{\hskip 0.5mm}c@{\hskip 0.5mm}c@{}}

{\rotatebox{90}{\hspace{5mm}\normalsize
{Input}}} &
\includegraphics[height=\turnheightnew]{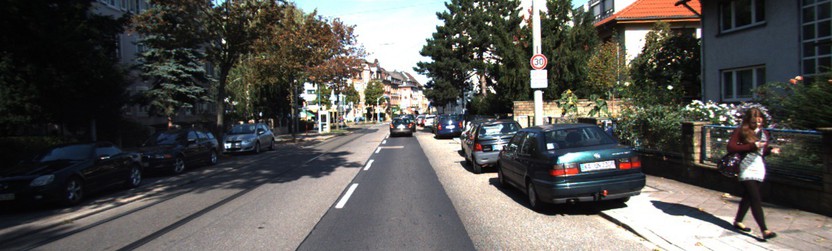} &
\includegraphics[height=\turnheightnew]{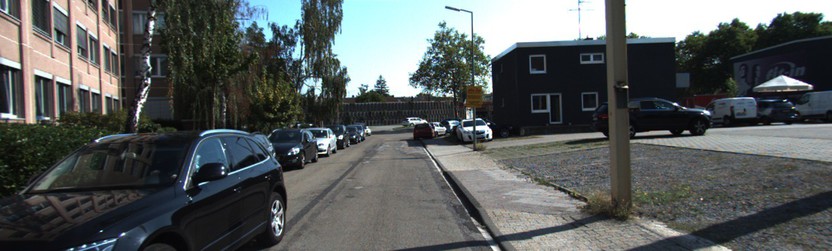} &
\includegraphics[height=\turnheightnew]{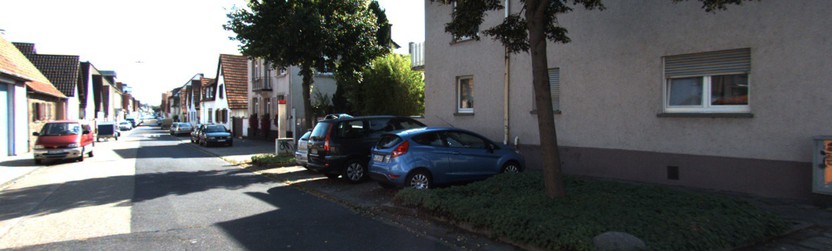} &
\includegraphics[height=\turnheightnew]{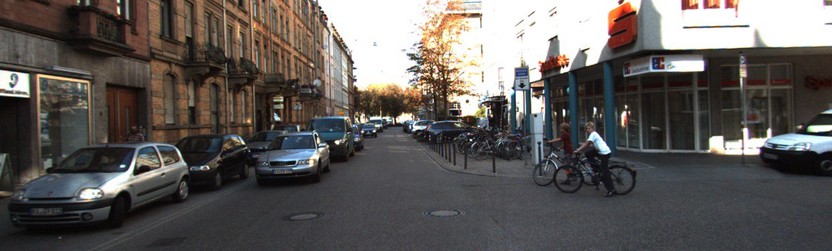}\\

{\rotatebox{90}{\hspace{0mm}\scriptsize
{Godard~\cite{monodepth17}}}} &
\includegraphics[height=\turnheightnew]{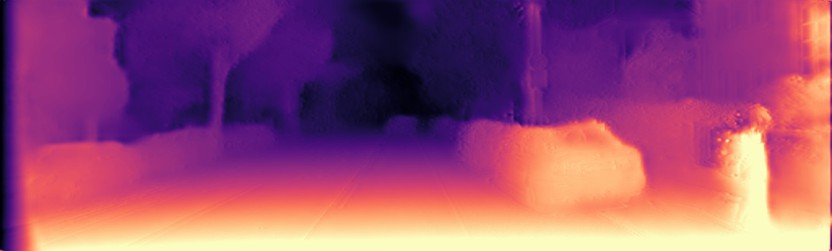} &
\includegraphics[height=\turnheightnew]{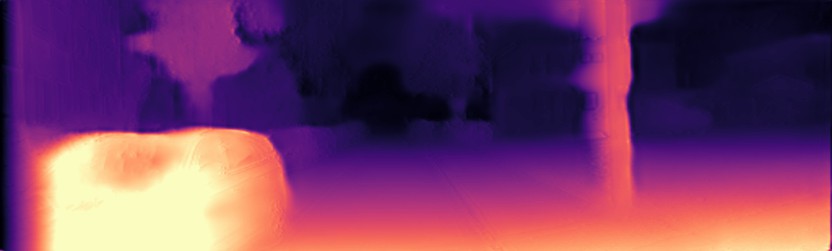} &
\includegraphics[height=\turnheightnew]{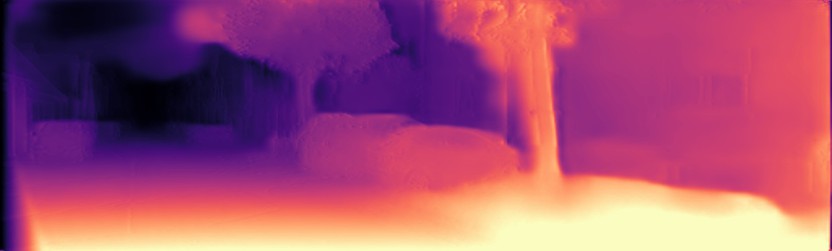} &
\includegraphics[height=\turnheightnew]{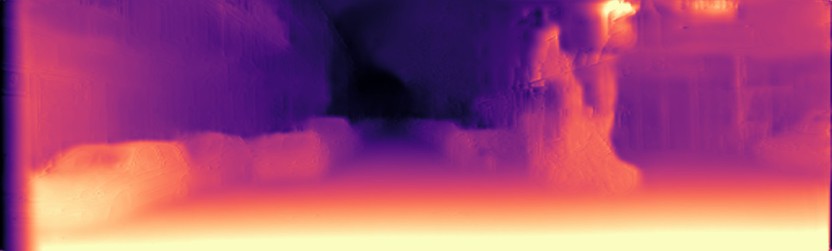} \\

{\rotatebox{90}{\hspace{0mm}\scriptsize
{Zhou~\cite{zhou2017unsupervised}}}} &
\includegraphics[height=\turnheightnew]{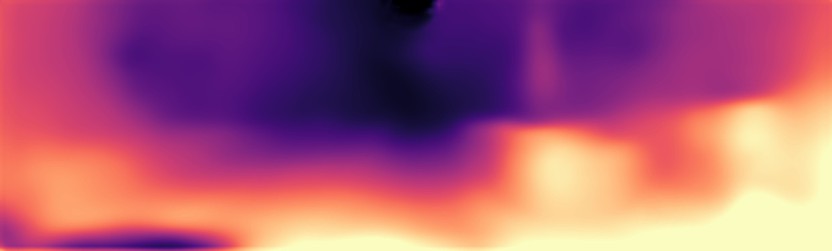} &
\includegraphics[height=\turnheightnew]{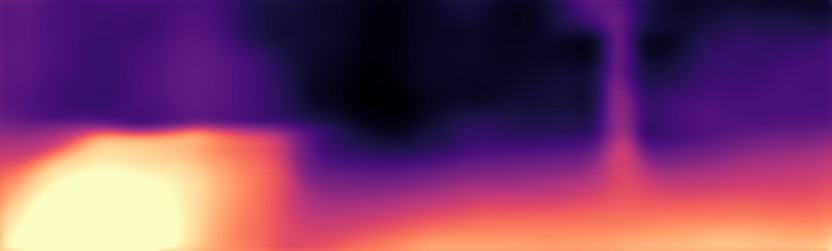} &
\includegraphics[height=\turnheightnew]{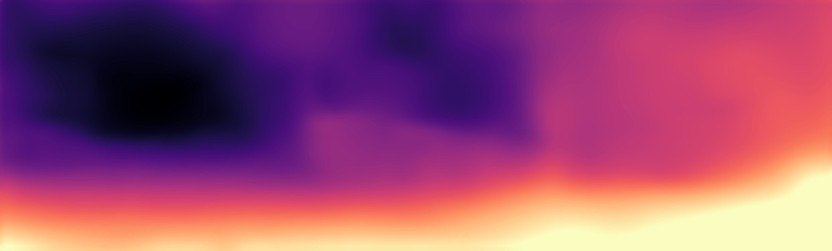} &
\includegraphics[height=\turnheightnew]{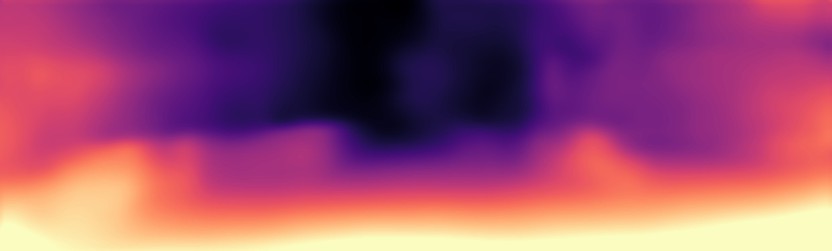} \\

{\rotatebox{90}{\hspace{0mm}\scriptsize
{DDVO~\cite{Wang_2018_CVPR}}}} &
\includegraphics[height=\turnheightnew]{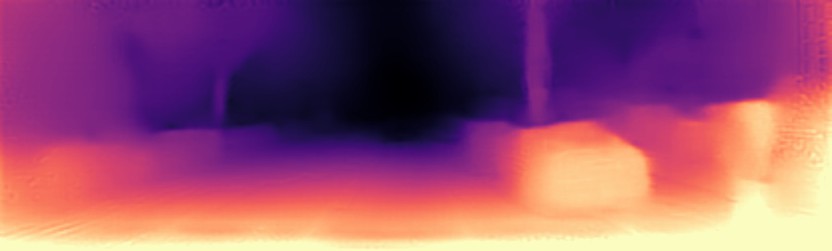} &
\includegraphics[height=\turnheightnew]{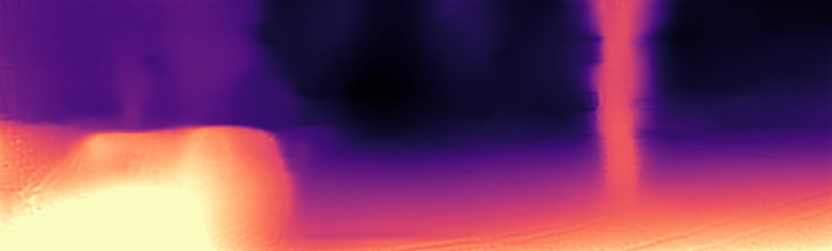} &
\includegraphics[height=\turnheightnew]{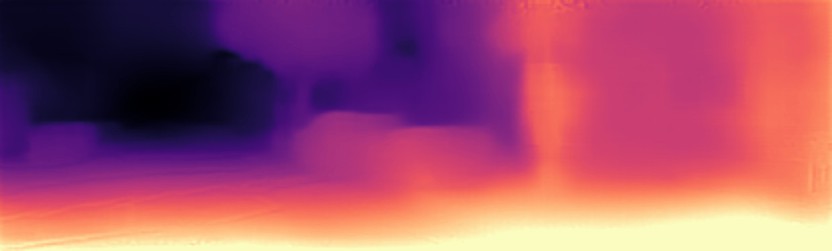} &
\includegraphics[height=\turnheightnew]{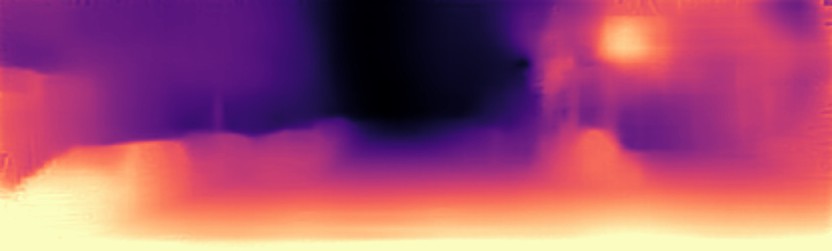} \\

{\rotatebox{90}{\hspace{0mm}\scriptsize
{GeoNet~\cite{yin2018geonet}}}} &
\includegraphics[height=\turnheightnew]{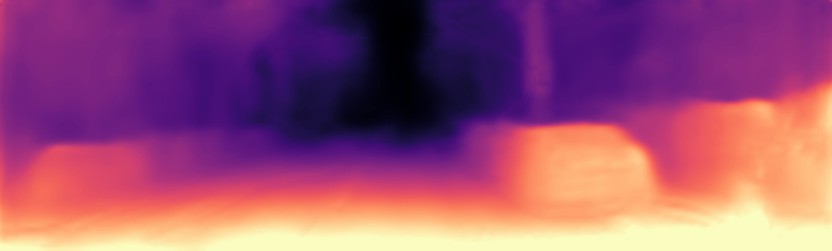} &
\includegraphics[height=\turnheightnew]{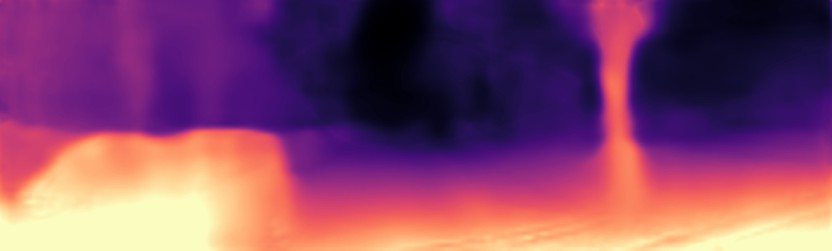} &
\includegraphics[height=\turnheightnew]{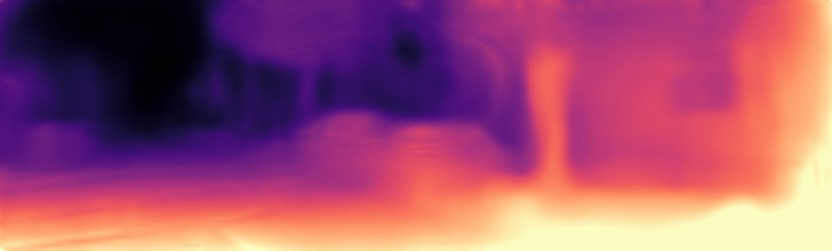} &
\includegraphics[height=\turnheightnew]{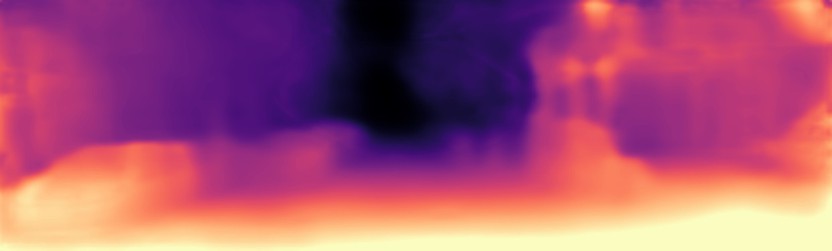} \\

{\rotatebox{90}{\hspace{0mm}\scriptsize
{Zhan~\cite{zhan2018unsupervised}}}} &
\includegraphics[height=\turnheightnew]{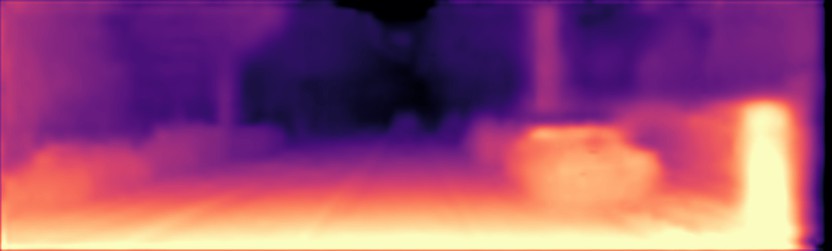} &
\includegraphics[height=\turnheightnew]{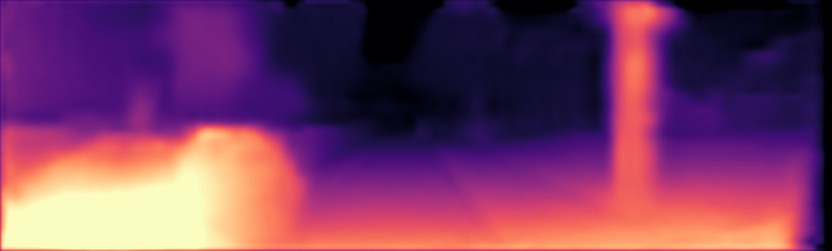} &
\includegraphics[height=\turnheightnew]{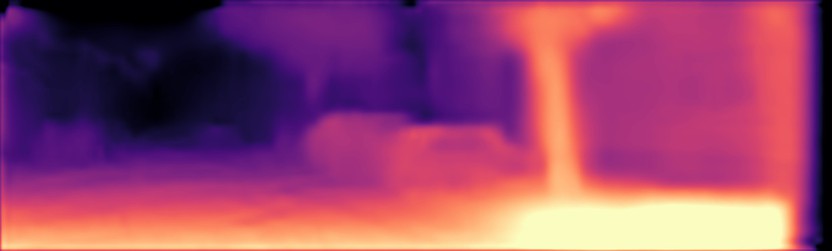} &
\includegraphics[height=\turnheightnew]{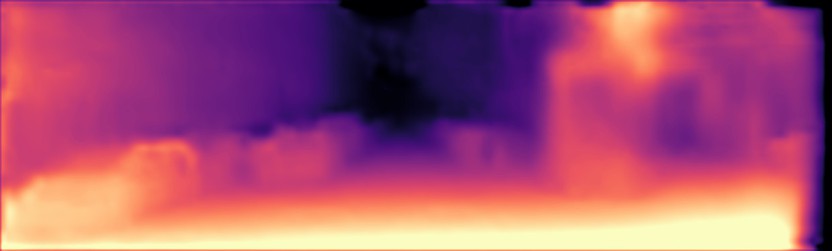} \\

{\rotatebox{90}{\hspace{0mm}\scriptsize
Ranjan~\cite{ranjan2019competitive}}} &
\includegraphics[height=\turnheightnew]{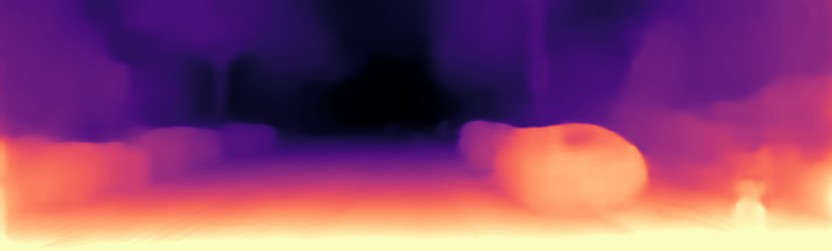} &
\includegraphics[height=\turnheightnew]{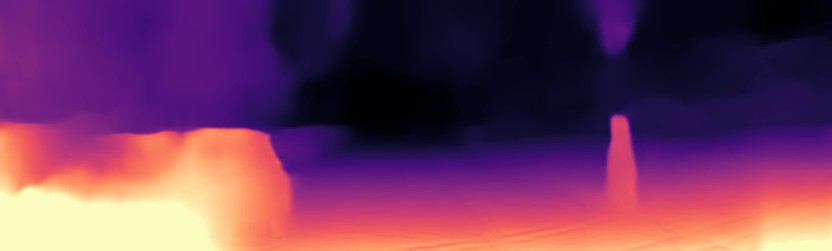} &
\includegraphics[height=\turnheightnew]{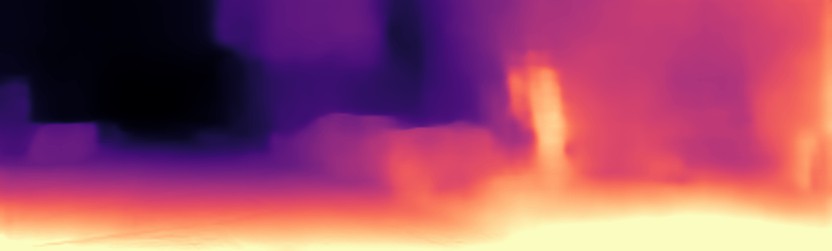} &
\includegraphics[height=\turnheightnew]{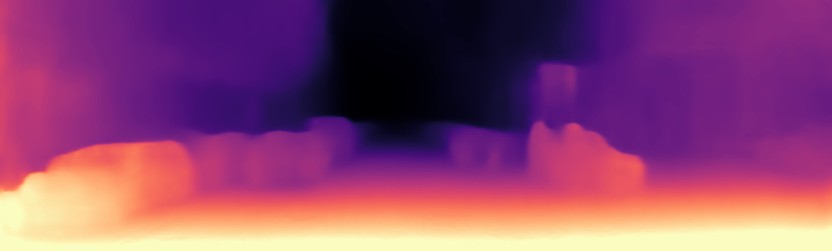} \\

{\rotatebox{90}{\hspace{0mm} \scriptsize
3Net-R50~\cite{luo2019every}}} &
\includegraphics[height=\turnheightnew]{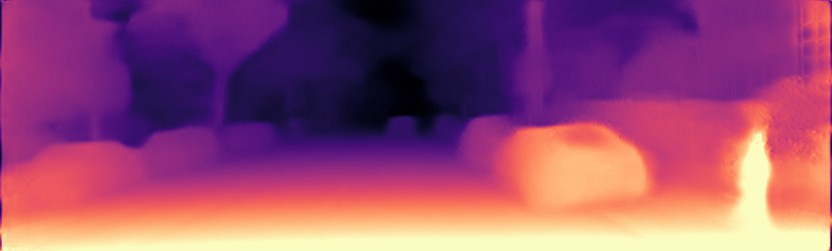} &
\includegraphics[height=\turnheightnew]{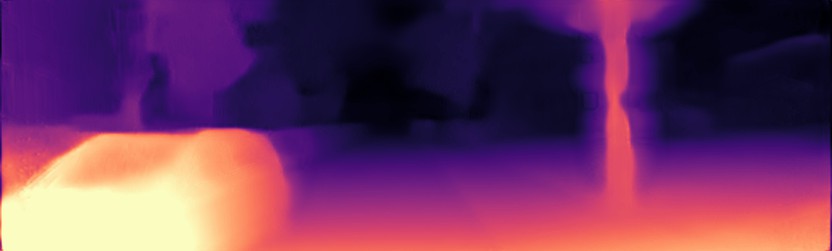} &
\includegraphics[height=\turnheightnew]{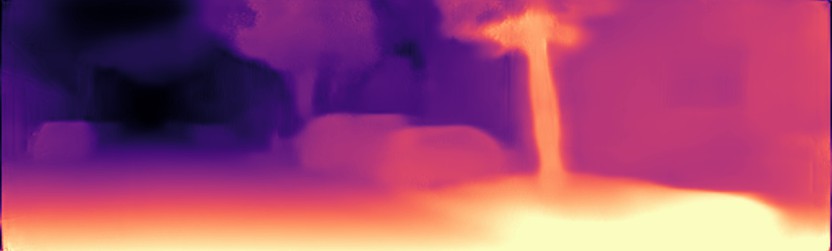} &
\includegraphics[height=\turnheightnew]{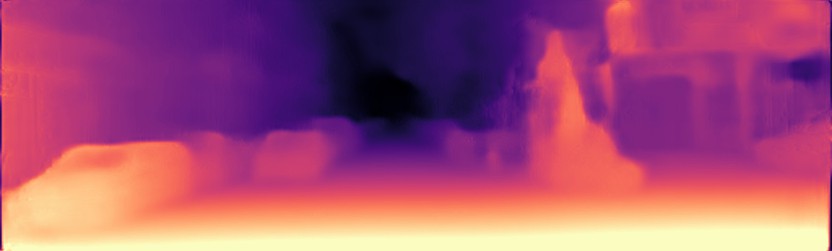} \\ 

{\rotatebox{90}{\hspace{0mm} \scriptsize
EPC++~\cite{luo2019every}}} &
\includegraphics[height=\turnheightnew]{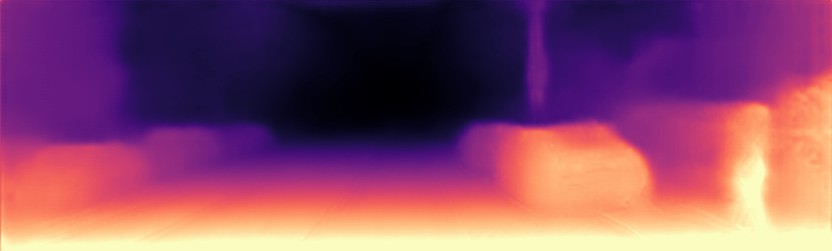} &
\includegraphics[height=\turnheightnew]{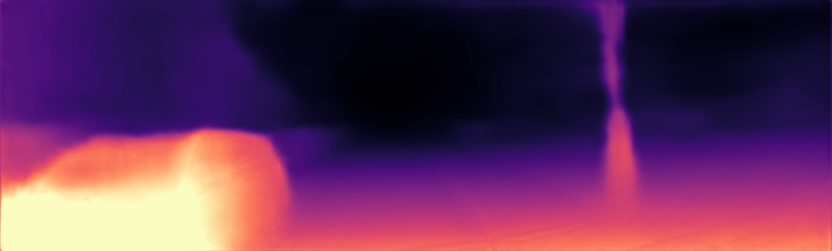} &
\includegraphics[height=\turnheightnew]{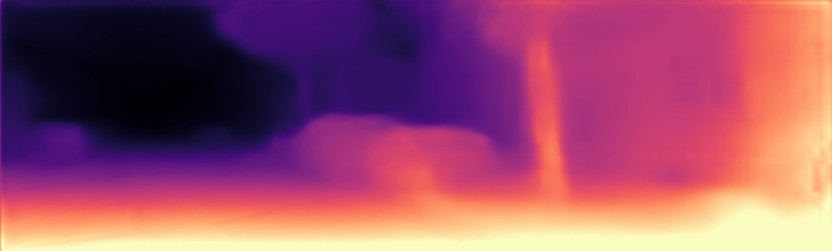} &
\includegraphics[height=\turnheightnew]{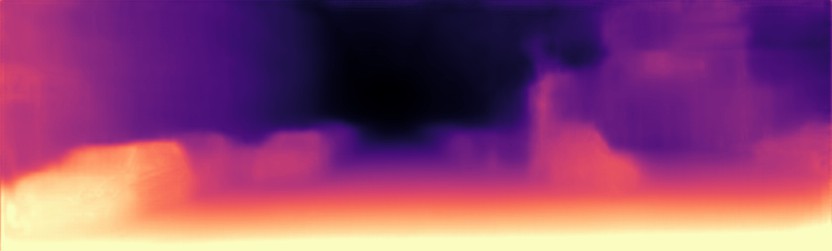} \\ 

{\rotatebox{90}{\hspace{0mm}\scriptsize
Monodepth2~\cite{godard2019digging}}} &
\includegraphics[height=\turnheightnew]{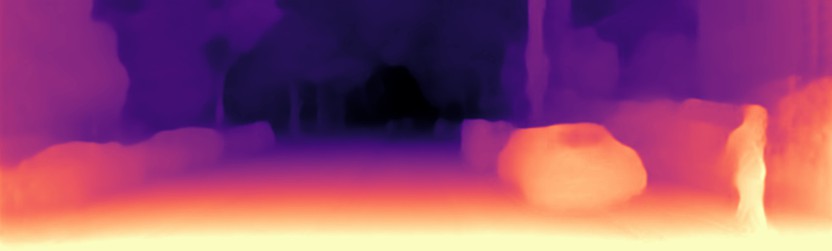} &
\includegraphics[height=\turnheightnew]{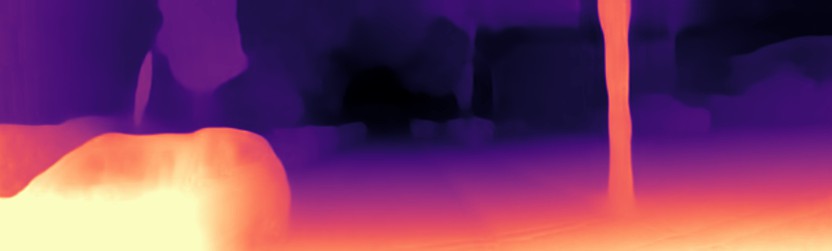} &
\includegraphics[height=\turnheightnew]{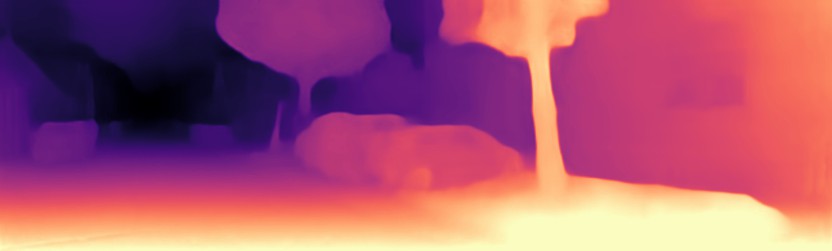} &
\includegraphics[height=\turnheightnew]{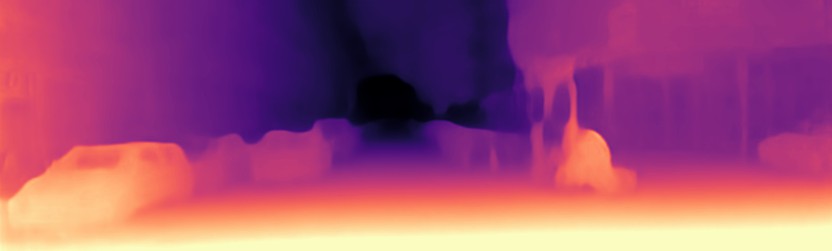} \\

{\rotatebox{90}{\hspace{0mm}\scriptsize
\textbf{SVDistNet}}} &
\includegraphics[height=\turnheightnew]{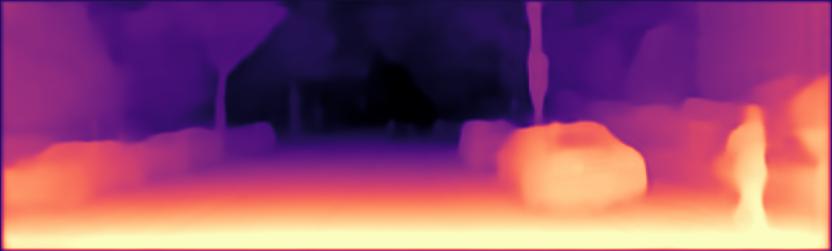} &
\includegraphics[height=\turnheightnew]{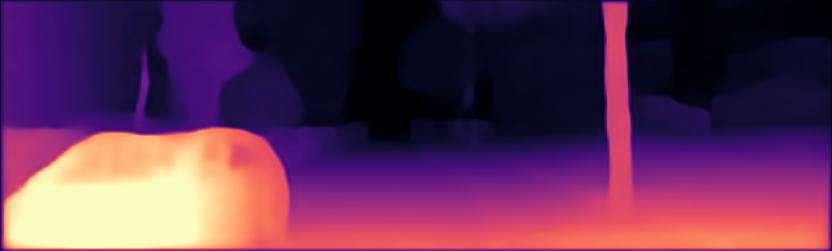} &
\includegraphics[height=\turnheightnew]{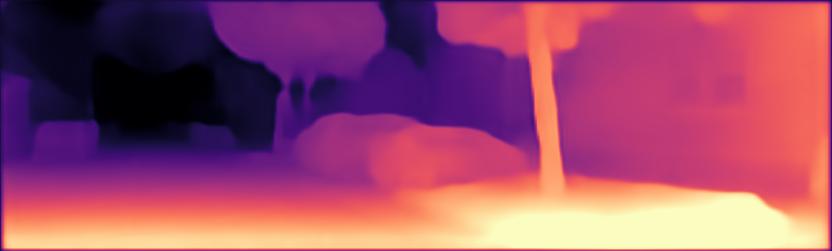} &
\includegraphics[height=\turnheightnew]{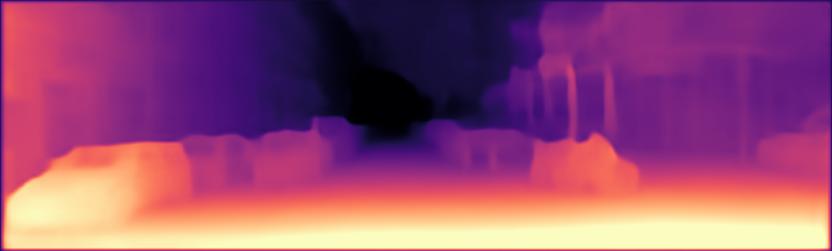} \\

\end{tabular}
}
  \caption{\bf Qualitative results of SVDistNet on KITTI compared with state-of-the-art algorithms.}
\label{fig:KITTIDepthComparison}
\end{figure*}

For an extensive overview of the previous monocular methods' results, including the surround-view approach CGT tensor, we create Table~\ref{tab:kitti-monocular-results}. First, we train and evaluate the depth maps generated from the LiDAR point clouds, where Table~\ref{tab:kitti-monocular-results} shows that with the use of the contributions, \textit{we outperform all previous methods}. Using the online refinement method from~\cite{casser2019depth}, we obtain a significant improvement, while the results are still superior to previous methods. When training and evaluating the improved KITTI labels for depth estimation, we can show a significant improvement compared to previous approaches. For \textit{SVDistNet}, we use the general camera tensor $C_t$ as described in Section~\ref{sec:camera-geometry-tensor} in the model, wherein instead of the angle of incidence maps, we employ the maps generated using Eq.~\ref{eq:ff} for pinhole cameras. We showcase a comparison of \textit{SVDistNet}'s estimates with leader-board algorithms in Figure~\ref{fig:KITTIDepthComparison}.\par
\begin{table*}[!ht]
\begin{adjustbox}{width=\columnwidth}
\renewcommand{\arraystretch}{0.87}
\centering
{
\small
\setlength{\tabcolsep}{0.3em}
\begin{tabular}{c|llccccccc}
\toprule
& \multirow{2}{*}{\textbf{Method}}
& Train
& \cellcolor[HTML]{7d9ebf}Abs Rel & \cellcolor[HTML]{7d9ebf}Sq Rel & \cellcolor[HTML]{7d9ebf}RMSE & \cellcolor[HTML]{7d9ebf}RMSE$_{log}$ & \cellcolor[HTML]{e8715b}$\delta<1.25$ & \cellcolor[HTML]{e8715b}$\delta<1.25^2$ & \cellcolor[HTML]{e8715b}$\delta<1.25^3$ \\
\cmidrule(lr){4-7} \cmidrule(lr){8-10}
& & & \multicolumn{4}{c}{\cellcolor[HTML]{7d9ebf}lower is better} & \multicolumn{3}{c}{\cellcolor[HTML]{e8715b}higher is better} \\
\toprule
\parbox[t]{2mm}{\multirow{33}{*}{\rotatebox[origin=c]{90}{Original~\cite{Eigen_14}}}}
& SfMLearner~\cite{zhou2017unsupervised}      & M & 0.208 & 1.768 & 6.958 & 0.283 & 0.678 & 0.885 & 0.957 \\
& DNC~\cite{yang2018unsupervised}             & M & 0.182 & 1.481 & 6.501 & 0.267 & 0.725 & 0.906 & 0.963 \\
& Vid2Depth~\cite{mahjourian2018unsupervised} & M & 0.163 & 1.240 & 6.220 & 0.250 & 0.762 & 0.916 & 0.968 \\
& LEGO~\cite{yang2018lego}                    & M & 0.162 & 1.352 & 6.276 & 0.252 & 0.783 & 0.921 & 0.969 \\
& Kumar~\cite{cs2018monocular}                & M & 0.211 & 1.979 & 6.154 & 0.263 & 0.731 & 0.897 & 0.959 \\
& Wang \etal~\cite{wang2019unsupervised}      & M & 0.158 & 1.277 & 5.858 & 0.233 & 0.785 & 0.929 & 0.973 \\
& GeoNet~\cite{yin2018geonet}                 & M & 0.155 & 1.296 & 5.857 & 0.233 & 0.793 & 0.931 & 0.973 \\
& Cycle-SfM~\cite{sun2019cycle}               & M & 0.162 & 1.349 & 5.847 & 0.239 & 0.784 & 0.925 & 0.969 \\
& Li \etal~\cite{li2019sequential}            & M & 0.150 & 1.127 & 5.564 & 0.229 & 0.823 & 0.936 & 0.974 \\
& DDVO~\cite{Wang_2018_CVPR}                  & M & 0.151 & 1.257 & 5.583 & 0.228 & 0.810 & 0.936 & 0.974 \\
& DF-Net~\cite{zou2018df}                     & M & 0.150 & 1.124 & 5.507 & 0.223 & 0.806 & 0.933 & 0.973 \\
& GANVO~\cite{almalioglu2019ganvo}            & M & 0.150 & 1.141 & 5.448 & 0.216 & 0.808 & 0.939 & 0.975 \\
& Bian~\cite{bian2019unsupervised}            & M & 0.137 & 1.089 & 5.439 & 0.217 & 0.830 & 0.942 & 0.975 \\
& EPC++~\cite{Yang2018c}                      & M & 0.141 & 1.029 & 5.350 & 0.216 & 0.816 & 0.941 & 0.976 \\
& CC~\cite{ranjan2019competitive}             & M & 0.140 & 1.070 & 5.326 & 0.217 & 0.826 & 0.941 & 0.975 \\
& Struct2Depth~\cite{casser2019depth}         & M & 0.141 & 1.036 & 5.291 & 0.215 & 0.816 & 0.945 & 0.979 \\
& LearnK~\cite{Gordon2019}                    & M & 0.128 & 0.959 & 5.230 & 0.212 & 0.845 & 0.947 & 0.976 \\
& SIGNet~\cite{Meng2019a}                     & M & 0.133 & 0.905 & 5.181 & 0.208 & 0.825 & 0.947 & 0.981 \\
& DualNet~\cite{Zhou2019}                     & M & 0.121 & 0.837 & 4.945 & 0.197 & 0.853 & 0.955 & 0.982 \\
& OmegaNet~\cite{tosi2020distilled}           & M & 0.126 & 0.835 & 4.937 & 0.199 & 0.844 & 0.953 & 0.982 \\
& SuperDepth~\cite{pillai2019superdepth}      & M & 0.116 & 1.055 & -     & 0.209 & 0.853 & 0.948 & 0.977 \\
& Monodepth2~\cite{godard2019digging}         & M & 0.115 & 0.903 & 4.863 & 0.193 & 0.877 & 0.959 & 0.981 \\
& PackNet-SfM~\cite{guizilini2019packnet}     & M & 0.111 & 0.829 & 4.788 & 0.199 & 0.864 & 0.954 & 0.980 \\
& FisheyeDistanceNet~\cite{kumar2020fisheyedistancenet} & M & 0.117 & 0.867 & 4.739 & 0.190 & 0.869 & 0.960 & 0.982 \\
& SGDepth~\cite{klingner2020self}             & M & 0.113 & 0.880 & 4.695 & 0.192 & 0.884 & 0.961 & 0.981 \\
& Patil \etal~\cite{patil2020don}             & M & 0.111 & 0.821 & 4.650 & 0.187 & 0.883 & 0.961 & 0.982 \\
& UnRectDepthNet~\cite{kumar2020unrectdepthnet} & M & 0.107 & 0.721 & 4.564 & 0.178 & 0.894 & 0.971 & 0.986 \\
& SynDistNet~\cite{kumar2020syndistnet}       & M & 0.109 & 0.718 & 4.516 & 0.180 & 0.896 & 0.973 & 0.986 \\
& Shu \etal~\cite{shu2020featdepth}           & M & 0.104 & 0.729 & 4.481 & 0.179 & 0.893 & 0.965 & 0.984 \\
& \textbf{SVDistNet}                          & M & \textbf{0.102} & \textbf{0.706} & \textbf{4.459} & \textbf{0.172} & \textbf{0.908} & \textbf{0.974} & \textbf{0.986} \\
\cmidrule{2-10}
& Struct2Depth~\cite{casser2019depth}         & M${^*}$ & 0.109 & 0.825 & 4.750 & 0.187 &0.874 & 0.958 & 0.983 \\
& GLNet~\cite{Chen2019b}                      & M$^{*}$ & 0.099 & 0.796 & 4.743 & 0.186 &0.884 & 0.955 & 0.979 \\
& Shu \etal~\cite{shu2020featdepth}           & M$^{*}$ & 0.088 & 0.712 & 4.137 & \textbf{0.169} & 0.915 & 0.965 & 0.982 \\
& \textbf{SVDistNet}                          & M${^*}$ & \textbf{0.086} & \textbf{0.701} & \textbf{4.118} & 0.170 & \textbf{0.919} & \textbf{0.976} & \textbf{0.985} \\
\midrule
\parbox[t]{2mm}{\multirow{11}{*}{\rotatebox[origin=c]{90}{Improved~\cite{uhrig2017sparsity}}}}
& SfMLearner~\cite{zhou2017unsupervised}    & M & 0.176 & 1.532 & 6.129 & 0.244 & 0.758 & 0.921 & 0.971 \\
& Vid2Depth~\cite{mahjourian2018unsupervised} & M & 0.134 & 0.983 & 5.501 & 0.203 & 0.827 & 0.944 & 0.981 \\
& GeoNet~\cite{yin2018geonet}               & M & 0.132 & 0.994 & 5.240 & 0.193 & 0.883 & 0.953 & 0.985 \\
& DDVO~\cite{Wang_2018_CVPR}                & M & 0.126 & 0.866 & 4.932 & 0.185 & 0.851 & 0.958 & 0.986 \\
& EPC++~\cite{Yang2018c}                    & M & 0.120 & 0.789 & 4.755 & 0.177 & 0.856 & 0.961 & 0.987 \\
& Monodepth2~\cite{godard2019digging}       & M & 0.090 & 0.545 & 3.942 & 0.137 & 0.914 & 0.983 & 0.995 \\
& PackNet-SfM~\cite{guizilini2019packnet}   & M & 0.078 & 0.420 & 3.485 & 0.121 & 0.931 & 0.986 & 0.996 \\
& UnRectDepthNet~\cite{kumar2020unrectdepthnet} & M & 0.081 & 0.414 & 3.412 & 0.117 & 0.926 & 0.987 & 0.996 \\
& SynDistNet~\cite{kumar2020syndistnet}     & M & 0.076 & 0.412 & 3.406 & 0.115 & 0.931 & 0.988 & 0.996 \\
& \textbf{SVDistNet}                        & M & \textbf{0.071} & \textbf{0.405} & \textbf{3.345} & \textbf{0.106} & \textbf{0.934} & \textbf{0.988} & \textbf{0.996} \\
& \textbf{SVDistNet}                        & M${^*}$ & \textbf{0.059} & \textbf{0.392} & \textbf{3.206} & \textbf{0.097} & \textbf{0.935} & \textbf{0.989} & \textbf{0.995} \\
\bottomrule
\end{tabular}
}
\end{adjustbox}
\caption[\bf Evaluation of SVDistNet on the KITTI dataset.]
{\textbf{Evaluation of the KITTI Eigen split} compared to most of the previous self-supervised monocular depth estimation methods. Following best practices, we cap depths at 80\,m. We also evaluate using the \textit{Original} depth maps generated from raw point clouds as proposed by \cite{Eigen_14} as well as \textit{Improved} annotated depth maps as introduced by~\cite{uhrig2017sparsity}. M indicates that the model is trained on monocular image sequences. M${^*}$ indicates the online refinement technique~\cite{casser2019depth}, where the model is trained during inference. Note that while most approaches use median scaling w.r.t. the ground truth at test-time for a scale-consistent prediction, we do not need to use this scaling method.}
\label{tab:kitti-monocular-results}
\end{table*}
\section{Conclusion}

Geometry and appearance are two crucial cues of scene understanding, \eg, in automotive scenes. This chapter developed a multi-task learning model to estimate metric distance and semantic segmentation in a synergized manner. Specifically, we leverage the semantic segmentation of potentially moving objects to remove wrongful projected objects inside the view synthesis step. We also proposed a novel architecture to semantically guide the distance estimation trainable in a one-stage fashion and introduce the application of a robust loss function application. The primary focus was to develop the proposed model for less explored fisheye cameras based on the WoodScape dataset.\par

This chapter discussed distance estimation in detail, which is a challenging and vital problem for autonomous driving. We have solved it successfully by developing novel methodologies and synergized multi-task learning approach, which is crucial for scene understanding. We advanced the problem of multi-camera distance estimation for surround-view fisheye cameras. We introduced a novel camera model adaptation mechanism wherein camera parameters are transformed into a tensor and used within the CNN model. The specific camera model parameters are used during training and inference. Using this technique, we demonstrate training of a single distance estimation model for twelve different cameras with different extrinsic and intrinsic parameters and achieve strongly improved results than training a specialized model for each camera variant. We demonstrated the effect of each proposed contribution individually and obtained state-of-the-art results on both WoodScape and KITTI datasets for self-supervised distance estimation. In the next chapter, we look into localization using perception algorithms. \ie, 2D object detection.\par
\chapter{Generalized Object Detection}
\label{Chapter6}
\minitoc
\section{Problem Definition}

Object detection is a comprehensively studied problem in autonomous driving. However, it has been relatively less explored in the case of fisheye cameras. The standard bounding box fails in fisheye cameras due to the strong radial distortion, particularly in the image's periphery. We explore better representations such as oriented bounding box, ellipse, and generic polygon for object detection in fisheye images. We use the IoU metric to compare these representations using accurate instance segmentation ground truth. We design a novel curved bounding box model that has optimal properties for fisheye distortion models. In this thesis, to the best of our knowledge, we perform the first detailed study on object detection based on fisheye camera images for autonomous driving scenarios. To encourage further research, we also made a public release of a dataset of 10,000 images with annotations for all considered object representations. This work was formally presented as \textit{Generalized Object Detection}~\cite{kumar2020syndistnet} as an oral at the \href{https://openaccess.thecvf.com/content/WACV2021/html/Rashed_Generalized_Object_Detection_on_Fisheye_Cameras_for_Autonomous_Driving_Dataset_WACV_2021_paper.html}{WACV} conference in 2021. \textit{My contribution to this work as a secondary author involved research idea discussion, writing code for (standard box model and training), code-reviews, generating a qualitative results video, and writing the research paper.}\par
\begin{figure}[!t]
  \centering
\begin{tabular}{cc}
 	\begin{overpic}[width=0.5\columnwidth]{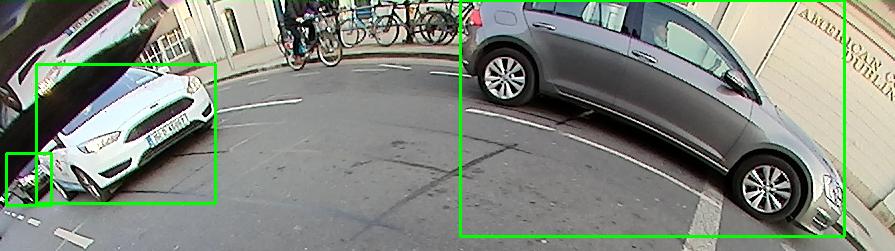}
    \put (0,23.3) {\colorbox{green}{$\displaystyle\textcolor{black}{\text{(a)}}$}}
    \end{overpic}
    \begin{overpic}[width=0.5\columnwidth]{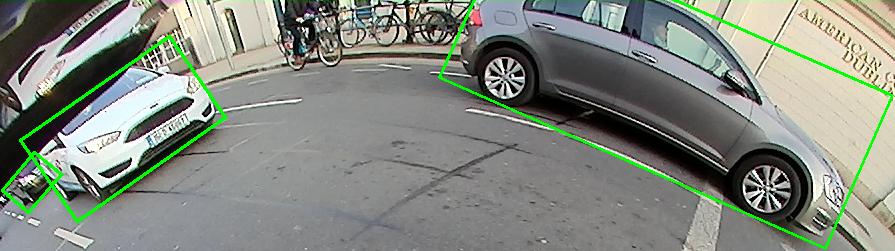}
    \put (0,23.3) {\colorbox{green}{$\displaystyle\textcolor{black}{\text{(b)}}$}}
    \end{overpic} \\
    \begin{overpic}[width=0.5\columnwidth]{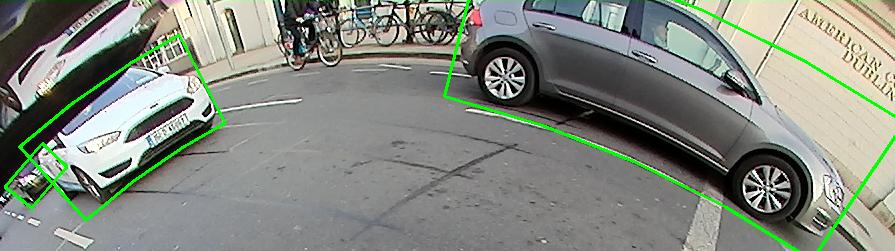}
    \put (0,23.3) {\colorbox{green}{$\displaystyle\textcolor{black}{\text{(c)}}$}}
    \end{overpic}
 	\begin{overpic}[width=0.5\columnwidth]{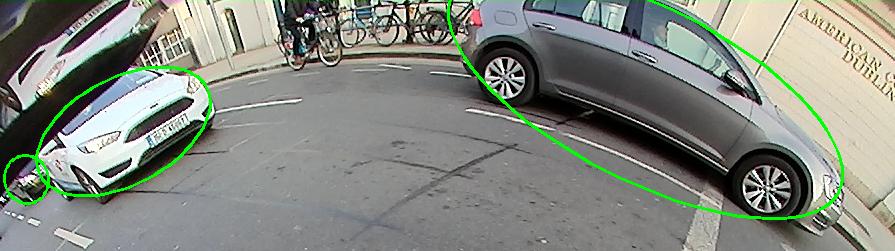}
    \put (0,23.3) {\colorbox{green}{$\displaystyle\textcolor{black}{\text{(d)}}$}}
    \end{overpic} \\
    \begin{overpic}[width=0.5\columnwidth]{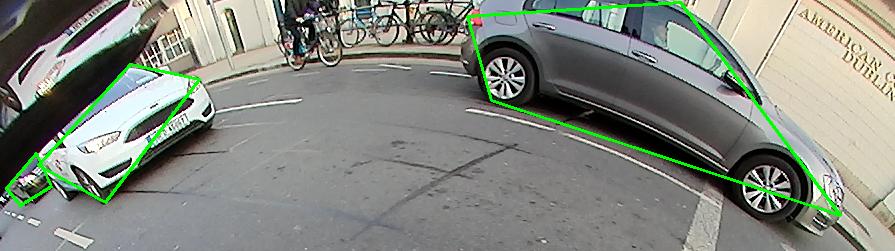}
    \put (0,23.3) {\colorbox{green}{$\displaystyle\textcolor{black}{\text{(e)}}$}}
    \end{overpic} 
    \begin{overpic}[width=0.5\columnwidth]{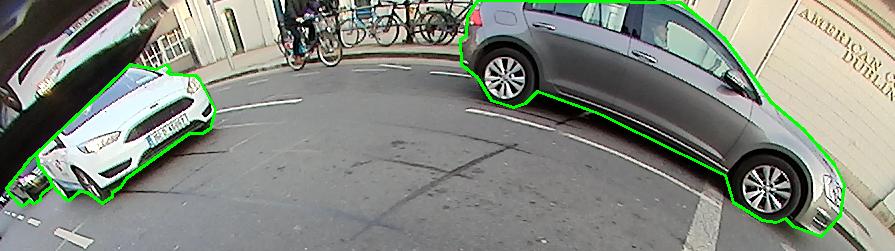}
    \put (0,23.3) {\colorbox{green}{$\displaystyle\textcolor{black}{\text{(f)}}$}}
    \end{overpic}
\end{tabular}
\caption[\bf Various 2D object detection representations on fisheye camera images.]
        {\textbf{Various 2D object detection representations on fisheye camera images.} (a) Standard Box, (b) Oriented Box, (c) Curved Box, (d) Ellipse, (e) 4-sided Polygon and (f) 24-sided Polygon.}
\label{fig:fisheye-yolo-main-fig}
\end{figure}
\vspace{5mm}
\section{Object Representations}
\label{sec:representations}

\subsection{Adaptation of Box Representations}

\subsubsection{Standard Box Representation}

The rectangular bounding box is the most common representation for object detection. They are aligned to the pixel grid axes, making them efficient to be regressed using a machine learning model. They are represented by four parameters ($\hat{x}$, $\hat{y}$, $\hat{w}$, $\hat{h}$), namely the box center, as well as width and height. It has the advantage of simplified, low-cost annotation. It also works in most cases, but it may capture a large non-object area within the box for complex shapes. This is particularly the case for distorted fisheye images, as shown in Figure~\ref{fig:fisheye-yolo-main-fig} (a).\par
\subsubsection{Oriented Box Representation}

The oriented box is a simple extension of the standard box with an additional parameter $\hat{\theta}$ to capture the box's rotation angle. It is also referred to as a tilted or rotated box. Lienhart~\etal~\cite{lienhart2002extended} adapted the Viola-Jones object detection framework to output rotated boxes. It is also commonly used in LiDAR top-view object detection methods~\cite{geiger2013vision}. The orientation of the ground-truth range spans the range of (-90\degree to +90\degree), where this rotation angle is defined with respect to the x-axis. For this study, we used instance segmentation contours to estimate the optimally oriented box as a minimum enclosing rectangle.\par
\subsubsection{Ellipse Representation}

The ellipse representation is closely related to an oriented box and can be represented using the same parameter set. Width and height parameters represent the major and minor axis of the ellipse. In contrast to an oriented box, the ellipse has a smaller area at the edge, and thus it is better for representing overlapping objects, as shown for the objects at the very left in Figure~\ref{fig:fisheye-yolo-main-fig}. It may also help to fit some objects such as vehicles better than a box. We created the ground truth by fitting a minimum enclosing ellipse to the ground truth instance segmentation contours. In parallel work, Ellipse R-CNN~\cite{dong2021ellipse} used ellipse representation for objects instead of boxes.\par
\subsection{Distortion Aware Representation} 
\label{sec:curvedbox}

This subsection aims to derive an optimal representation of objects undergoing radial distortion in fisheye images assuming a rectangular box is optimal for pinhole cameras. In the pinhole camera with no distortion, a straight line in the scene is imaged as a straight line in the image. However, a straight line in the scene is imaged as a curved segment in a fisheye image. The specific type of fisheye distortion determines the nature of the curved segment. The fisheye cameras from the Woodscape dataset we used are well represented and calibrated using a 4\textsuperscript{th} order polynomial model for the fisheye distortion \cite{yogamani2019woodscape}. We are aware that there have been many developments in fisheye camera models over the past few decades, \eg~\cite{kannala2006fisheye, brauerDivisionModel, khomutenko2016eucm}. In this section, we consider the $4\textsuperscript{th}$ order polynomial model and the division model only. The reason is that the $4\textsuperscript{th}$ order polynomial model is provided by the dataset that we use, and we examine the division model to understand if the use of circular arcs is valid under such fisheye projections.\par

In this case, the projection of a line on to the image can be described parametrically with higher order polynomial curves. Let us consider a much simpler model for the moment - a first-order polynomial (or equidistant) model of a fisheye camera. \ie $r' = a\theta$, where $r'$ is the radius on the image plane, and $\theta$ is the angle of the incident ray against the optical axis. If we consider the parametric equation $\mathbf{P}(t)$ of a line in 3D Euclidean space:
\begin{equation}
    \mathbf{P}(t) = \mathbf{D}t + \mathbf{Q}
\end{equation}
where $\mathbf{D} = [D_x, D_y, D_z]$ is the direction vector of the line and $\mathbf{Q} = [Q_x, Q_y, Q_z]$ is a point through which the line passes. Hughes~\etal~\cite{hughesFisheye} have shown that the projection on to a fisheye camera that adheres to equidistant distortion is described by:
\begin{equation}
    \mathbf{p}'(t) = \left[\begin{matrix}
        D_x t + Q_x \\
        D_y t + Q_y
    \end{matrix}\right]
    \frac{|\mathbf{p}'(t)|}{|\mathbf{p}(t)|}
\end{equation}
where
\begin{align}
    \frac{|\mathbf{p}'(t)|}{|\mathbf{p}(t)|} &=
    \frac{a \arctan{\left(\frac{d_{xy}(t)}{D_z t + Q_z}\right)}}{d_{xy}(t)} \\
    d_{xy}(t) &= \sqrt{(D_x t + Q_x)^2 + (D_y t + Q_y)^2}
\end{align}
$\mathbf{p}(t)$ is the projected line in a pinhole camera, and $\mathbf{p}'(t)$ is the distorted image of the line in a fisheye camera.
\begin{figure}[!t]
	\centering
	\subfigure[][Division model fit to the 4\textsuperscript{th} order polynomial model. Note that the two are almost indistinguishable and looks overlayed.]{\includegraphics[width=0.48\linewidth]{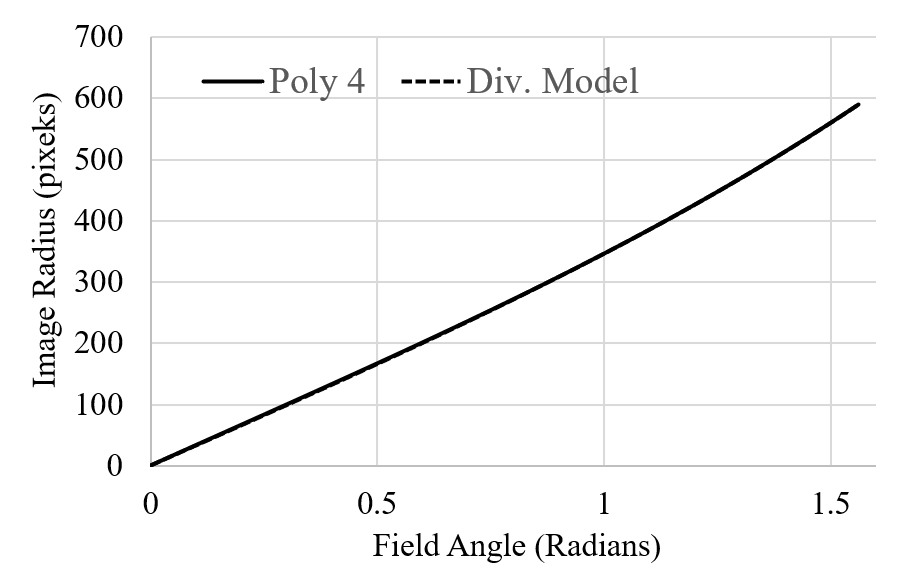}}\;\;
	\subfigure[][Residual error per field angle]{\includegraphics[width=0.48\linewidth]{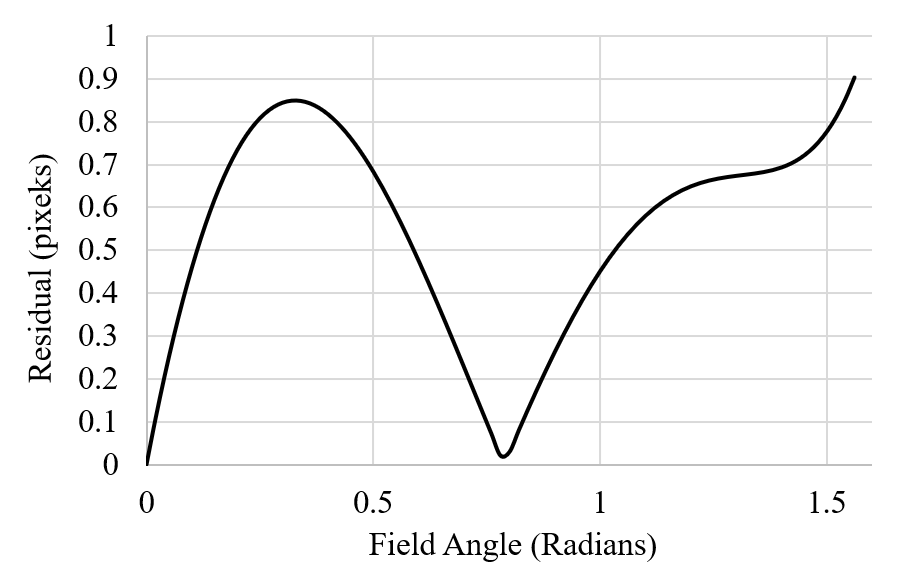}}
	\caption{\textbf{Approximation of the 4\textsuperscript{th} order radial distortion model by the division model.}}
	\label{fig:divmodel}
\end{figure}
This is a complex description of a straight line's projection, especially considering that we have ignored all but the first-order polynomial term. Therefore, it is highly desirable to describe straight lines' projection using a more straightforward geometric shape. Bräuer-Burchardt and Voss~\cite{brauerDivisionModel} show that if the first-order \textit{division model} can accurately describe the fisheye distortion, then we may use circles in the image to model the projected straight lines. As a note, the division model is generalized in \cite{scaramuzzaFisheye}, though it loses the property of straight line to circular arc projection. We should then consider how well the division model fits with the 4\textsuperscript{th} order polynomial model. In~\cite{hughesFisheye}, the authors adapt the division model slightly to include an additional scaling factor and prove that this does not impact the projection of a line to a circle. They show that the division model is a correct replacement for the equidistant fisheye model. Here we repeat this test but compare the division model to the 4\textsuperscript{th} order polynomial. The results are shown in Figure \ref{fig:divmodel}. As can be seen, the division model can map to the 4\textsuperscript{th} order polynomial with a maximum of $< 1$ pixel error. While this may not be accurate enough for applications in which sub-pixel error accuracy is desirable, it is sufficient for bounding box accuracy.\par
\begin{figure}[!t]
	\centering
	\subfigure[][A 4\textsuperscript{th}-degree polynomial model for radial distortion. We can visually notice that a box matures to a curved box, and it is justified theoretically in Section~\ref{sec:curvedbox}. ]{\includegraphics[width=0.48\linewidth]{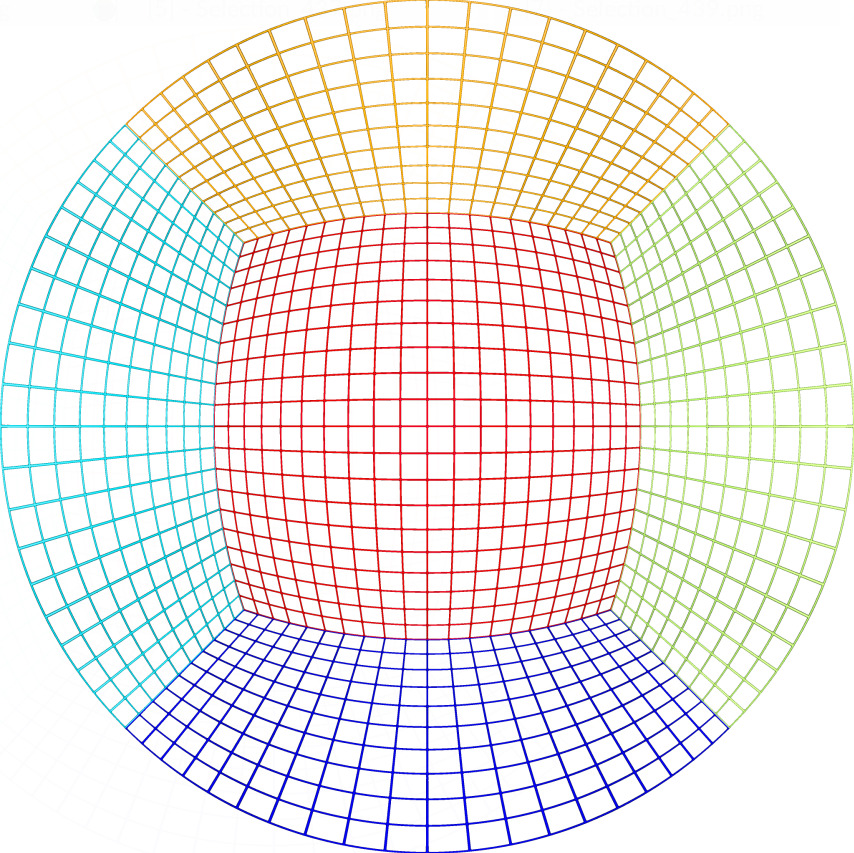}}\;\;
	\subfigure[][\textbf{A Curved Bounding Box} captures the radial distortion and obtains a better footpoint. The center of the circle can be equivalently re-parameterized using the object center ($\hat{x}$, $\hat{y}$).]{\includegraphics[width=0.48\linewidth]{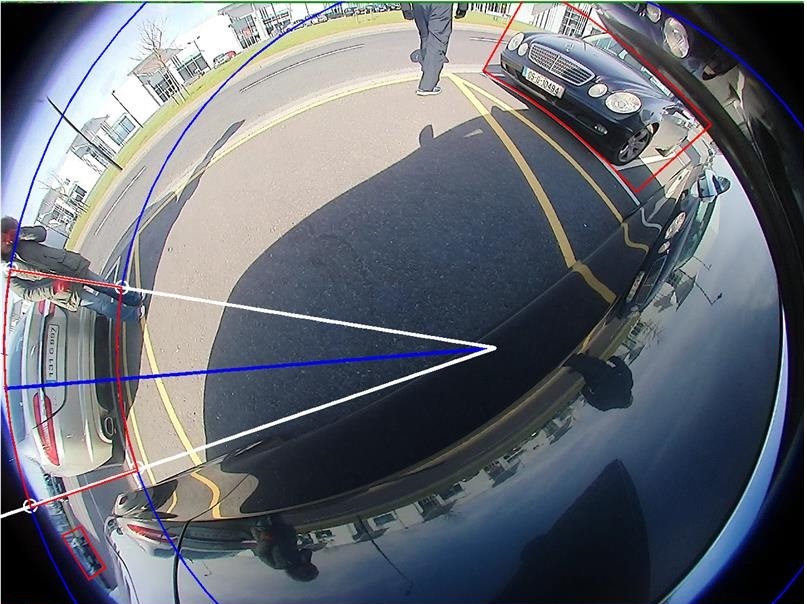}}
	\caption[\bf Illustration of fisheye distortion of projection of an Open Cube and Proposal of Curved Bounding Box]{\textbf{Illustration of fisheye distortion of projection of an Open Cube} and \textbf{Proposal of Curved Bounding Box} using a circle with an arbitrary center and radius.}
   \label{fig:curved_box_samples}
\end{figure}
Therefore, we propose a novel curved bounding box representation using circular arcs. Figure~\ref{fig:curved_box_samples} (left) provides a visual justification of circular arcs. We illustrate an open cube projection with grid lines where the straight lines become circular arcs after projection. Figure~\ref{fig:curved_box_samples} (right) illustrates the details of the curved bounding box. The blue line represents the axis, and the white lines intersect with the circles creating starting and ending points of the polygon. This representation allows two sides of the box to be curved, giving the flexibility to adapt to image distortion in fisheye cameras. It can also specialize in an oriented bounding box when there is no distortion for the objects near the principal point.\par

We create an automatic process to generate the representation that takes an object contour as an input. First, we generate an oriented box from the output contour. We choose a point that lies on the oriented box's axis line to represent a circle center. From the center, we create two circles intersecting with the corner points of the bounding box. We construct the polygon based on the two circles and the intersection points. To find the best circle center, we iterate over the axis line and choose the circle center, which forms a polygon with the minimum IoU with the instance mask. The output polygon can be represented by 6 parameters, namely, ($c_{1}$, $c_{2}$, $r_{1}$, $r_{2}$, $\theta_{1}$, $\theta_{2}$) representing the circle center, two radii and angles of the start and end points of the polygon relative to the horizontal x-axis. By simple algebraic manipulation, we can re-parameterize the curved box using the object center ($\hat{x}$, $\hat{y}$) following a typical box representation instead of the center of the circle.\par
\subsection{Generic Polygon Representations}

The polygon is a generic representation for an arbitrary shape and is typically used even, for instance segmentation annotation. Thus a polygon output can be seen as a coarse segmentation. We discuss two standard representations of a polygon and propose a novel extension that improves accuracy.\par
\begin{figure*}[!t]
\centering
   \includegraphics[width=0.99\linewidth]{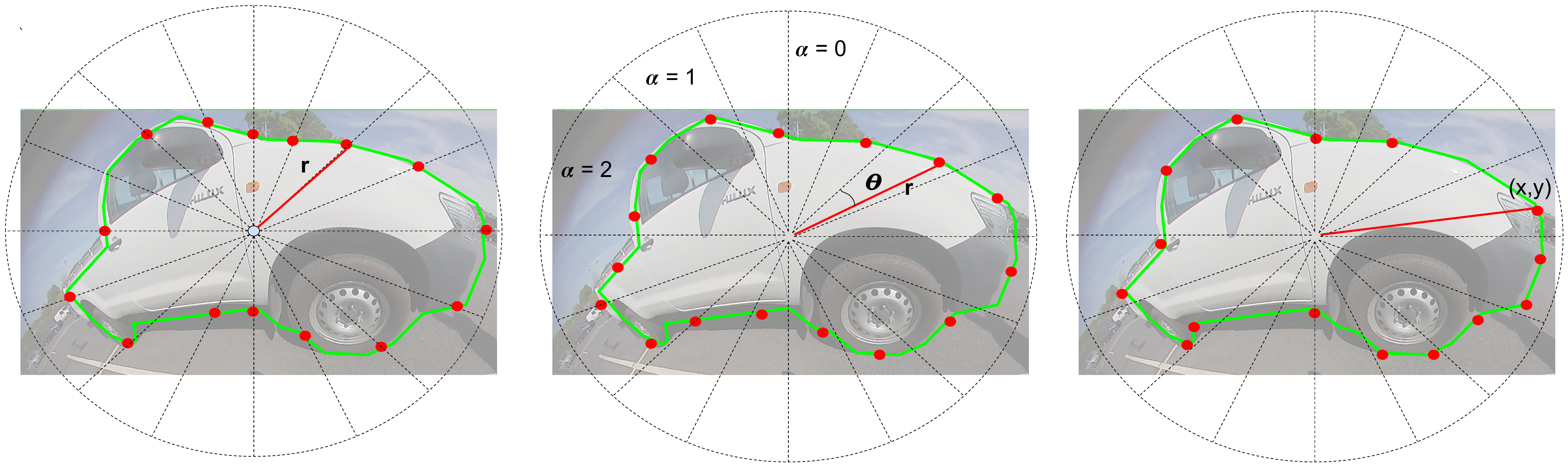}
   \caption[\bf Generic Polygon Representations.]
           {\textbf{Generic Polygon Representations.} 
            \textbf{Left:} Uniform angular sampling where the intersection of the polygon with the radial line is represented by one parameter per point (r). \textbf{Middle:} Uniform contour sampling using L2 distance. It can be parameterized in polar co-ordinates using 3 parameters ($r$, $\theta$, $\alpha$). $\alpha$ denotes the number of polygon vertices within the sector, and it may be used to simplify the training. Alternatively, two parameters (x,y) can be used, as shown in the figure on the right.
            \textbf{Right:} Variable step contour sampling. It is shown that the straight line in the bottom has fewer points than curved points, such as the wheel. This representation allows to maximize the utilization of vertices according to local curvature.}
\label{fig:polygon_representation}
\end{figure*}
\subsubsection{Uniform Angular Sampling}

The polar representation is quite similar to the PolarMask~\cite{polarmask}, and PolyYOLO~\cite{polyyolo} approaches. As illustrated in Figure~\ref{fig:polygon_representation} (left), the full angle range of $360\degree$ is split into $N$ equal parts where $N$ is the number of polygon vertices. Each polygon vertex is represented by the radial distance $r$ from the centroid of the object. Uniform angular sampling removes the need for encoding the $\theta$ parameter. The polygon is finally represented by its object center ($\hat{x}$, $\hat{y}$) and radius \{$r_{i}$\}.\par
\subsubsection{Uniform Perimeter Sampling}

In this representation, we divide the perimeter of the object contour equally to create $N$ vertices. Thus the polygon is represented by a set of vertices \{($x_{i}$, $y_{i}$)\} using the centroid of the object as the origin. PolyYOLO~\cite{polyyolo} showed that it is better to learn polar representation of the vertices \{($r_{i}$, $\theta_{i}$)\} instead. They define a parameter $\alpha$ to denote the presence or absence of a vertex in a sector, as shown in Figure~\ref{fig:polygon_representation} (middle). We extend this parameter to be the count of vertices in the sector.\par
\subsubsection{Curvature-adaptive Perimeter Sampling}

The original curve in the object contour between two vertices is approximated by a straight line in the polygon. For regions of high curvature, this is not a good approximation. Thus, we propose an adaptive sampling based on the curvature of the local contour. We distribute the vertices non-uniformly in order to represent the object contour best. Figure~\ref{fig:polygon_representation} (right) shows the effectiveness of this approach, where a larger number of vertices is used for higher curvature regions than straight lines, which can be represented by less vertices. We adopt the algorithm in~\cite{teh1989detection} to detect the dominant points in a given curved shape, which best represents the object. Then we reduce the set of points using the algorithm in~\cite{douglas1973algorithms} to get the most representative simplified curves. This way, the polygon has dense points on the curved parts and sparse points on the straight parts, which maximize the utilization of the predefined number of points per contour.\par
\section{FisheyeYOLO Network Architecture}

\begin{figure}[!t]
  \centering
  \includegraphics[width=\textwidth]{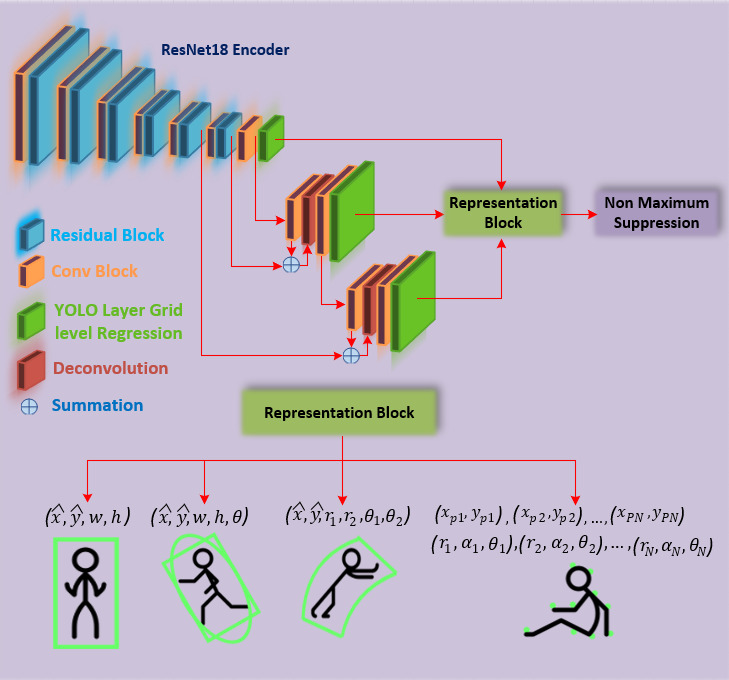}
  \caption[\bf FisheyeYOLO is an extension of YOLOv3.]
          {\textbf{FisheyeYOLO is an extension of YOLOv3} which can output different output representations discussed in Section~\ref{sec:representations}.}
\label{fig:generalized_yolo}
\end{figure}
We adapt the YOLOv3~\cite{YOLOV3} model to output different representations as discussed in Section \ref{sec:representations}. We call this model FisheyeYOLO, as illustrated in Figure \ref{fig:generalized_yolo}. The baseline bounding box model is the same as YOLOv3 \cite{YOLOV3}, except that we replace the Darknet53 encoder with a ResNet18 encoder. Similar to YOLOv3, object detection is performed at multiple scales. For each grid in each scale, object width ($\hat{w}$), height ($\hat{h}$), object center coordinates ($\hat{x}$, $\hat{y}$) and the object class is inferred. Finally, a non-maximum suppression is used to filter out the low confidence detections. Instead of using $L_{2}$ loss for categorical and objectness classification, we used standard categorical cross-entropy and binary entropy losses, respectively. The final loss is a combination of sub-losses:
\begin{align}
\mathcal{L}_{xy} &= \lambda_{coord} \sum_{i=0}^{S^2}\sum_{j=0}^{B} l_{ij}^{obj}[(x_{i}-\hat{x}_{_{i}})^2 + (y_{i}-\hat{y}_{_{i}})^2] \\
\mathcal{L}_{wh} &= \lambda_{coord} \sum_{i=0}^{S^2}\sum_{j=0}^{B} l_{ij}^{obj}[(\sqrt{w_{i}}- \sqrt{\hat{w}_{_{i}}})^2 + (\sqrt{h_{i}}-\sqrt{\hat{h}_{_{i}}})^2] + (\sqrt{h_{i}}-\sqrt{\hat{h}_{_{i}}})^2] \\
\mathcal{L}_{obj} &= -\sum_{i=0}^{S^2}\sum_{j=0}^{B} [C_{i}log(\hat{C}_{_{i}})] \\
\mathcal{L}_{class} &= -\sum_{i=0}^{S^2} l_{ij}^{obj}\sum_{c=\text{classes}}[c_{i,j}log(p({\hat{c}_{_{i,j}}}))] \\
\mathcal{L}_{total} &= \mathcal{L}_{xy} + \mathcal{L}_{wh} + \mathcal{L}_{obj} + \mathcal{L}_{class}
\end{align}
where height and width are predicted as offsets from pre-computed anchor boxes.
\begin{align}
     \hat{w} &= a_{w} * e^{f_w} \\
     \hat{h} &= a_{h} * e^{f_h} \\
     \hat{x} &= g_{x} + f_{x} \\
     \hat{h} &= g_{y} * f_{y}
\end{align}
where $a_{w}$, $a_{h}$ anchor box width and height. $f_{w}$, $f_{h}$, $f_{x}$, $f_{y}$ are the outputs from last layer of the network at grid location $g_{x}$, $g_{y}$.\par
In the case of oriented box or ellipse prediction, we define an additional loss function based on ellipse angle or orientation of the box. The loss function for oriented box and ellipse is:
\begin{align}
\mathcal{L}_{orn} &= \sum_{i=0}^{S^2}\sum_{j=0}^{B} l_{ij}^{obj}[\theta_{i}-\hat{\theta}_{_{i}}]^2 \\
\mathcal{L}_{total} &= \mathcal{L}_{xy} + \mathcal{L}_{wh} + \mathcal{L}_{obj} + \mathcal{L}_{class} + \mathbf{\mathcal{L}_{orn}}\label{eq:orn_loss}
\end{align}
where $\mathcal{L}_{total}$, is the total loss minimized for oriented box regression. In case of curved box, $\mathcal{L}_{wh}$ is replaced by $\mathcal{L}_{cods}$ in Eq. \eqref{eq:rtheta_loss}.\par
We also explored methods of learning orientation as a classification problem instead of a regression problem. One motivation is due to the discontinuity of angles at $90\degree$ due to wrapping around of angles. In this scenario, we discretized the orientation into 18 bins, where each bin represents a range of 10\degree with a tolerance of +-5\degree. To further improve the prediction, we design an IoU loss function that guides the model to minimize the difference in the area of the predicted box and the ground truth box. We compute the area of the predicted and ground truth rectangles and apply regression loss on those values. This loss maximizes the overlapping area between the prediction and the ground truth by improving the overall results. The IoU loss is,
\begin{align}
\mathcal{L}_{IoU} &= \lambda_{coord} \sum_{i=0}^{S^2}\sum_{j=0}^{B} l_{ij}^{obj}[(a_{i}-\hat{a}_{_{i}})^2]
\end{align}
where $a$ represents the area of the representation at hand. We report all the results related to these experiments in Table~\ref{tab:ablation}.\par
The polar polygon regression loss is,
\begin{align}
  \mathcal{L}_{cods} &= \sum_{i=0}^{S^2} \sum_{j=0}^{N} \hat{\alpha}_{ij} [(r_{i,j}-\hat{r}_{_{i,j}})^2 + (\theta_{i,j}-\hat{\theta}_{_{i,j}})^2] \label{eq:rtheta_loss}\\
  \mathcal{L}_{mask} &=-\sum_{i=0}^{S^2} \sum_{j=0}^{N} \alpha_{ij}
log(\hat{\alpha}_{ij}) \\
  \mathcal{L}_{total} &= \mathcal{L}_{xy} + \mathcal{L}_{obj} + \mathcal{L}_{class} + \mathbf{\mathcal{L}_{cods}} + \mathbf{\mathcal{L}_{mask}}
\end{align}
where N corresponds to the number of sampling points, each point is sampled with a step size of $360/N$ angle in polar coordinates, as shown in Figure~\ref{fig:polygon_representation}. The polar loss is similar to PolyYOLO \cite{polyyolo}, where each polygon point is (in red) is represented using three parameters $r$, $\theta$, and $\alpha$. Hence the total required parameters for $N$ sampling points are $3\times N$. The same is presented in Figure \ref{fig:polygon_representation} (middle).\par
In Cartesian representation, we regress over two parameters ($\hat{x}$, $\hat{y}$) for each polygon point. We further improve the predictions by adding the IoU loss function, which minimizes the area between the prediction and ground truth. We refer to both loss functions as localization loss $\mathcal{L}_{Localization}$. The combined loss for Cartesian polygon predictions is:
\begin{equation}
\label{insta_yolo_loss}
\mathcal{L}_{total} = \mathcal{L}_{Class} +\mathcal{L}_{Obj} +\mathcal{L}_{Localization}
\end{equation}
where $\mathcal{L}_{Obj}$ and $\mathcal{L}_{Class}$ are inherited from Yolov3 loss functions. According to the representation at hand, we perform the non-maximum suppression. We generate the predictions for all the representations; filter out the low confidence objects—computation of IoU of the output polygon with the list of outputs where high-IoU objects are filtered out.
\section{Experimental Results}

The objective of this work is to study various representations for the output of fisheye object detection. Conventional object detection algorithms evaluate their predictions against their ground truth, which is usually a bounding box. Unlike conventional evaluation, the first objective is to provide a better representation than a conventional bounding box. Therefore, we first evaluate the representations against the most accurate representation of the object, the ground-truth instance segmentation mask. We report the mIoU between a representation and the ground-truth instance mask.\par

Additionally, we qualitatively evaluate the representations in obtaining object intersection with the ground (footpoint). This is critical as it helps to localize the object in the map and provides more accurate vehicle trajectory planning. Finally, we report model speed in terms of frames-per-second (fps) as we focus on real-time performance. The distortion is higher in side cameras compared to front and rear cameras. Thus, we provide the evaluation on each camera separately. To simplify the baseline, we only evaluate the vehicle's class, although four classes are available in the dataset.\par
\subsubsection{Number of Polygon Points}

\begin{table}[!t]
\centering
\begin{adjustbox}{width=0.6\columnwidth}
\begin{tabular}{@{}l|c|c|c|c|c|c@{}}
\toprule
\textit{\begin{tabular}[c]{@{}l@{}} \# Vertices\end{tabular}} &
  \begin{tabular}[c]{@{}c@{}}4 \end{tabular} &
  \begin{tabular}[c]{@{}c@{}}12\end{tabular} &
  \begin{tabular}[c]{@{}c@{}}24\end{tabular} &
  \begin{tabular}[c]{@{}c@{}}36\end{tabular} &
  \begin{tabular}[c]{@{}c@{}}60\end{tabular} &
  \begin{tabular}[c]{@{}c@{}}120\end{tabular} \\\hline
mIoU &  85 & 85.3 & 86.6 & 91.8 & 94.2 & 98.4 \\ \bottomrule
\end{tabular}
\end{adjustbox}
\caption{\bf Analysis of the number of polygon vertices for representing the objects contour.}
\label{tab:number_points_analysis}
\end{table}
The polygon is a more generic representation of complex object shapes that arise in fisheye images. We perform a study to understand the effect of the number of vertices parameter in a polygon. We use a uniform perimeter sampling method to vary the number of vertices and compare the IoU using instance segmentation as ground truth. The results are tabulated in Table~\ref{tab:number_points_analysis} and the mIoU is calculated between the approximated polygon and the ground truth instance segmentation mask. A 24-sided polygon seems to provide a reasonable trade-off between the number of parameters and model accuracy. Although a 120-sided polygon provides a better ground truth with far too many points, it will be challenging to learn this representation, and it might even produce noisy overfitting. For the quantitative experiments, we fix the number of vertices to $24$ for each object. We observe no significant difference in fps due to increasing the number of vertices where the models run at $56$ fps on a standard NVIDIA TitanX GPU. It is due to the utilization of the YoloV3~\cite{YOLOV3} architecture, which performs the prediction at each grid cell in a parallel manner.\par
\subsubsection{Evaluation of Representation Capacity}

Table~\ref{tab:gt_table} compares the performance of different representations using its ground truth fit relative to the instance segmentation ground truth. We estimate the best fit for each representation using ground truth instance segmentation masks and then compute the mIoU to evaluate capacity. We also list the number of parameters used for each representation to provide a comparison w.r.t. complexity. This empirical metric demonstrates the maximum performance a representation can achieve regardless of the model complexity. As expected, a 24-sided polygon achieves the highest mIoU showing that it has the best representation capacity. The proposed curvature-adaptive polygon achieves a {2.2\%} improvement over the uniform sampling polygon with the same vertices. Polygon annotation is more expensive to collect, and it increases model complexity. Thus it is still interesting to also consider simple bounding box representations.\par
\begin{table}[t]
\centering
\begin{adjustbox}{width=0.8\columnwidth}
\begin{tabular}{lcccccc}
\toprule
\multicolumn{1}{l|}{\textit{\textbf{Representation}}} 
& \multicolumn{4}{c|}{\cellcolor[HTML]{7d9ebf} \textbf{mIoU}}
& \multicolumn{1}{l|}{\cellcolor[HTML]{e8715b} \textbf{mIoU}} 
& \cellcolor[HTML]{00b050} \textit{
\begin{tabular}[c]{@{}l@{}} \textbf{No. of} \\ \textbf{params} \end{tabular}} \\ 
\toprule
\multicolumn{1}{l|}{} & \multicolumn{1}{c|}{Front} & \multicolumn{1}{c|}{Rear} & \multicolumn{1}{c|}{Left} & \multicolumn{1}{c|}{Right} & & \\ 
\midrule
Standard Box  & 53.7 & 47.9 & 60.6 & 43.2 & 51.35& 4 \\
Curved Box    & 53.7 & 48.6 & 63.5 & 44.2 & 52.5 & 6 \\
Oriented Box  & 55.0 & 50.2 & 64.8 & 45.9 & 53.9 & 5 \\
Ellipse       & 56.5 & 51.7	& 66.5 & 47.5 &	55.5 & 5 \\
\hline
4-sided Polygon (uniform)  & 70.7 & 70.6 & 70.2 & 69.6 & 70.2 & 8 \\
24-sided Polygon (uniform) & 85.0 & 84.9 & 83.9 & 83.8 & 84.4 & 48 \\
24-sided Polygon (adaptive) & \textbf{87.2} & \textbf{87} & \textbf{86.2} & \textbf{86.1} & \textbf{86.6} & 48 \\
\bottomrule
\end{tabular}
\end{adjustbox}
\caption{\bf Evaluation of the representation capacity of various representations.}
\label{tab:gt_table}
\end{table}
Compared to the standard box representation, oriented box representation is approximately {2.5-4\%} efficient for the side cameras and {1.3-2.3\%} for front cameras. The ellipse representation improves the efficiency further by an additional 2\% for side cameras and 1-2\% for front cameras.
The curved box achieves a {1.15\%} improvement over the standard box.
However, it is slightly less than for the oriented box representation due to the constraint that two circular sides of the box share the same circle center, which adds some area inside the polygon, decreasing the IoU. In addition, the curvature is not modeled for the horizontal edges of the box. We plan to explore these extensions in future work to obtain a more optimal curved bounding box and leverage circular arcs' convergence at vanishing points.\par

The curved box's current version has the advantage of getting a tight bottom edge, capturing the footpoint for estimating the object's 3D location. The object's footpoint is captured almost entirely, as observed in qualitative results, especially for the side cameras where distortion is higher. This is important from an application perspective for vehicle navigation, as the footpoint is used for the projection of the object to the 3D world. 
Compared to polygon representation, curved-box representation has low annotation cost due to fewer representation points, which saves annotation effort.\par
\subsubsection{Quantitative Results}

Table~\ref{tab:ablation} shows our studies on methods predicting the orientation of the box or the ellipse efficiently. Angle classification and the added IoU loss significantly improved the mAP score relative to a standard baseline. The proposed variable step polygon representation provides a significant improvement of 2.7\%. At first, we train a model to regress over the box and its orientation, as specified in Eq. \eqref{eq:orn_loss}. In the second experiment, orientation prediction is addressed as a classification problem instead of regression as a possible solution to the discontinuity problem. We divide the orientation range of $180\degree$ into $18$ bins, where each bin represents $10\degree$, making this an 18 class classification problem. During inference, an acceptable error of +-5 degrees for each box is considered. Using this classification strategy, we improve performance by {1.6\%}. We are formulating orientation of box or ellipse prediction as a classification model where the IoU loss is found to be superior in performance compared to a direct regression. It yields a {2.9\%} improvement in accuracy. Hence we use this model as a standard representation for oriented box and ellipse prediction when comparing with other representations.\par
\begin{figure*}[!t]
  \captionsetup{skip=-0.3pt}
  \centering
  \newcommand{\turnheightnew}{0.33\columnwidth}

\begin{adjustbox}{max size={\textwidth}{\textheight}}
\begin{tabular}{@{\hskip 0.4mm}c@{\hskip 0.4mm}c@{\hskip 0.4mm}c@{\hskip 0.4mm}}

\includegraphics[height=\turnheightnew]{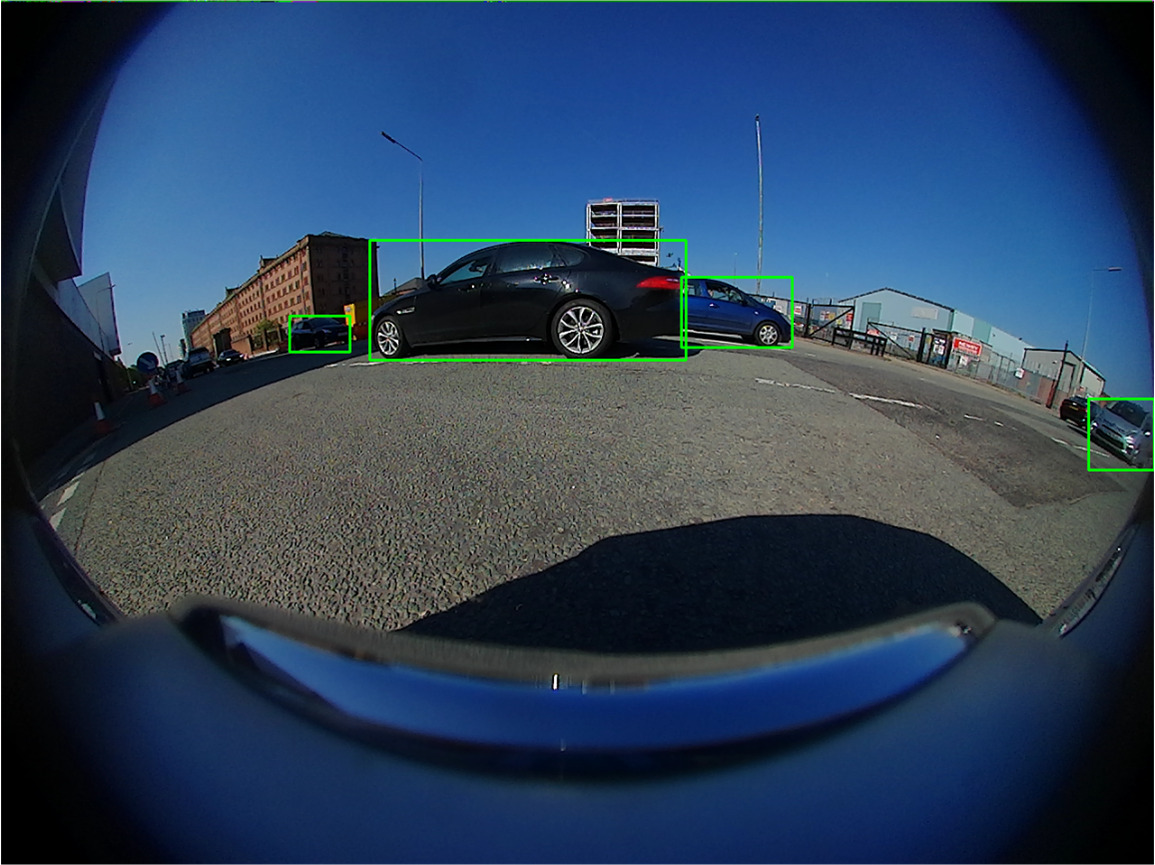} &
\includegraphics[height=\turnheightnew]{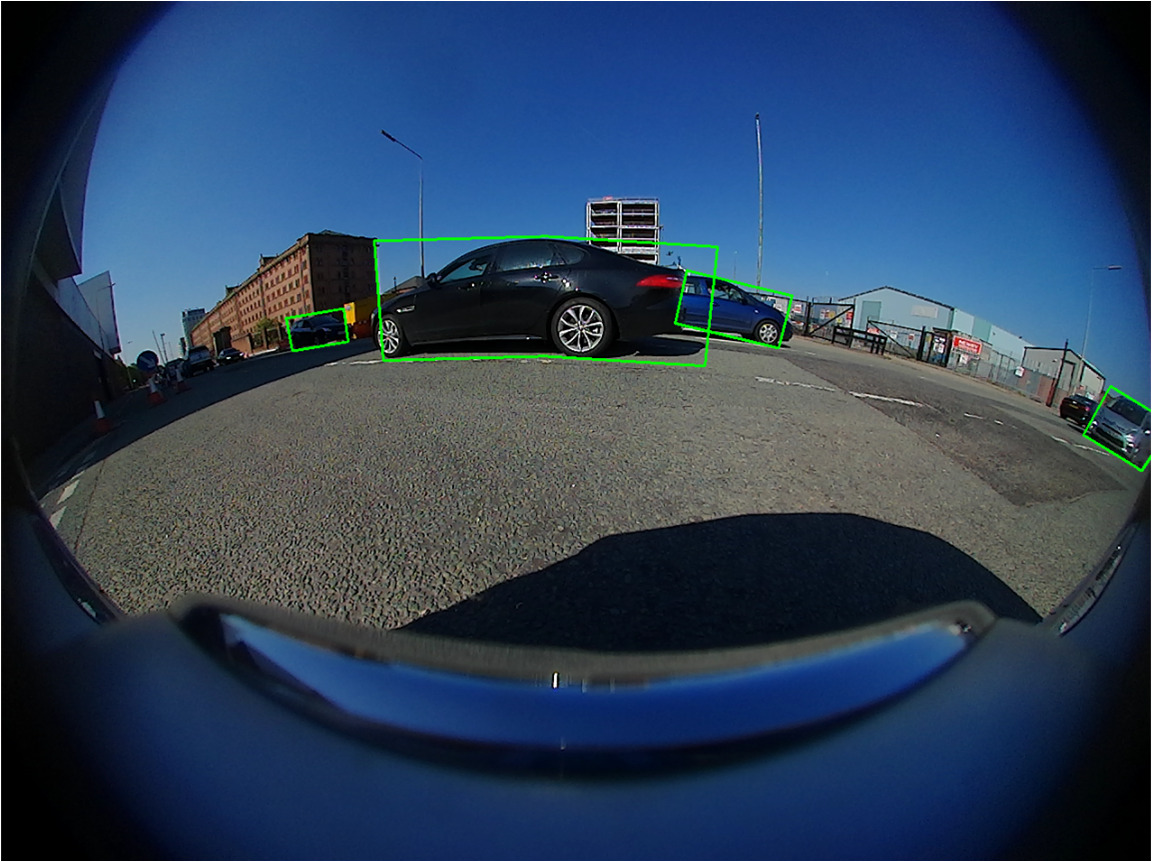} &
\includegraphics[height=\turnheightnew]{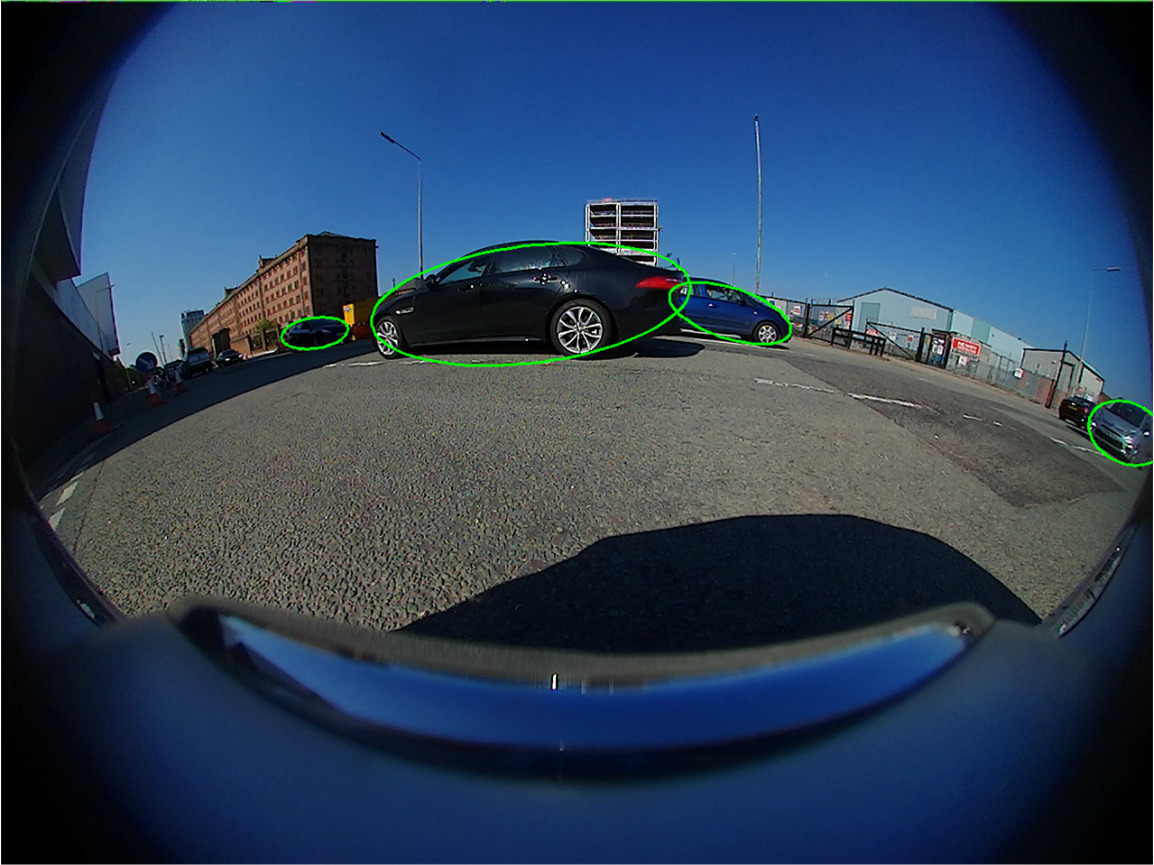} \\

\includegraphics[height=\turnheightnew]{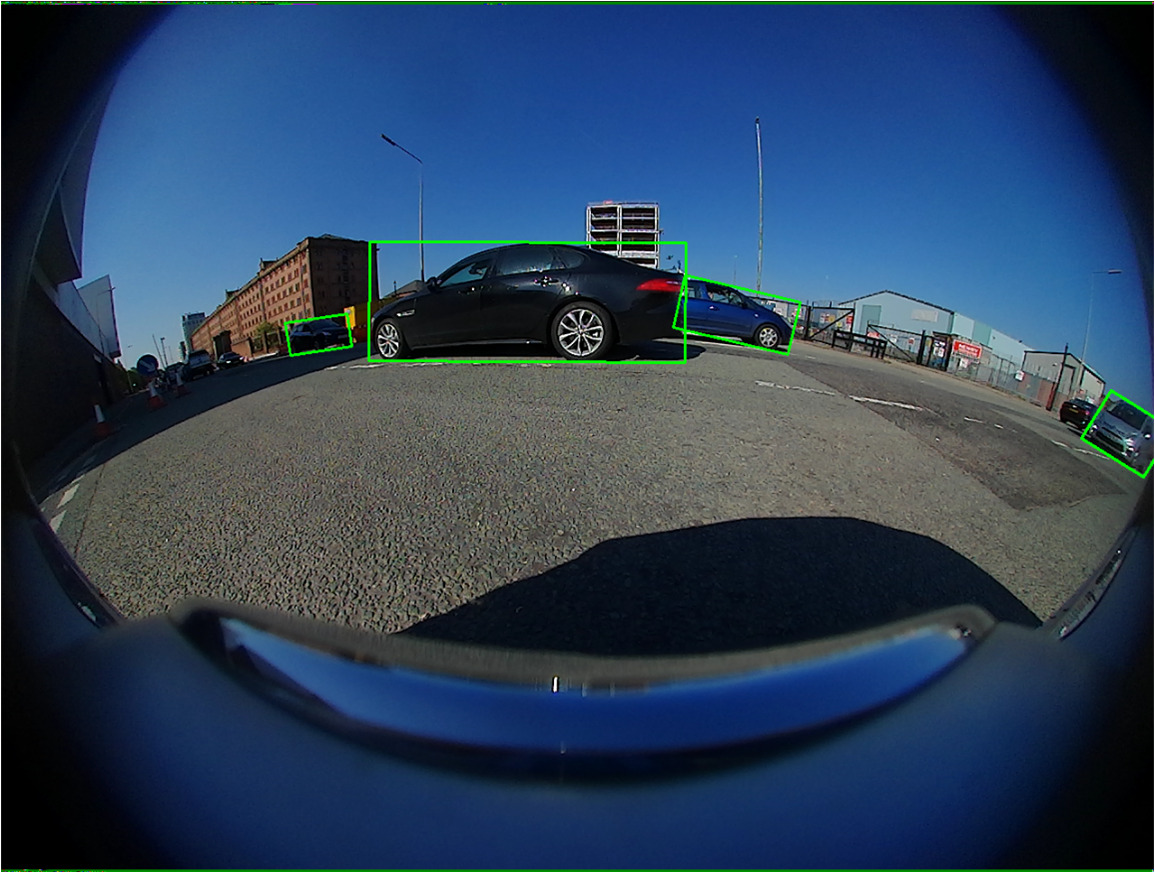} &
\includegraphics[height=\turnheightnew]{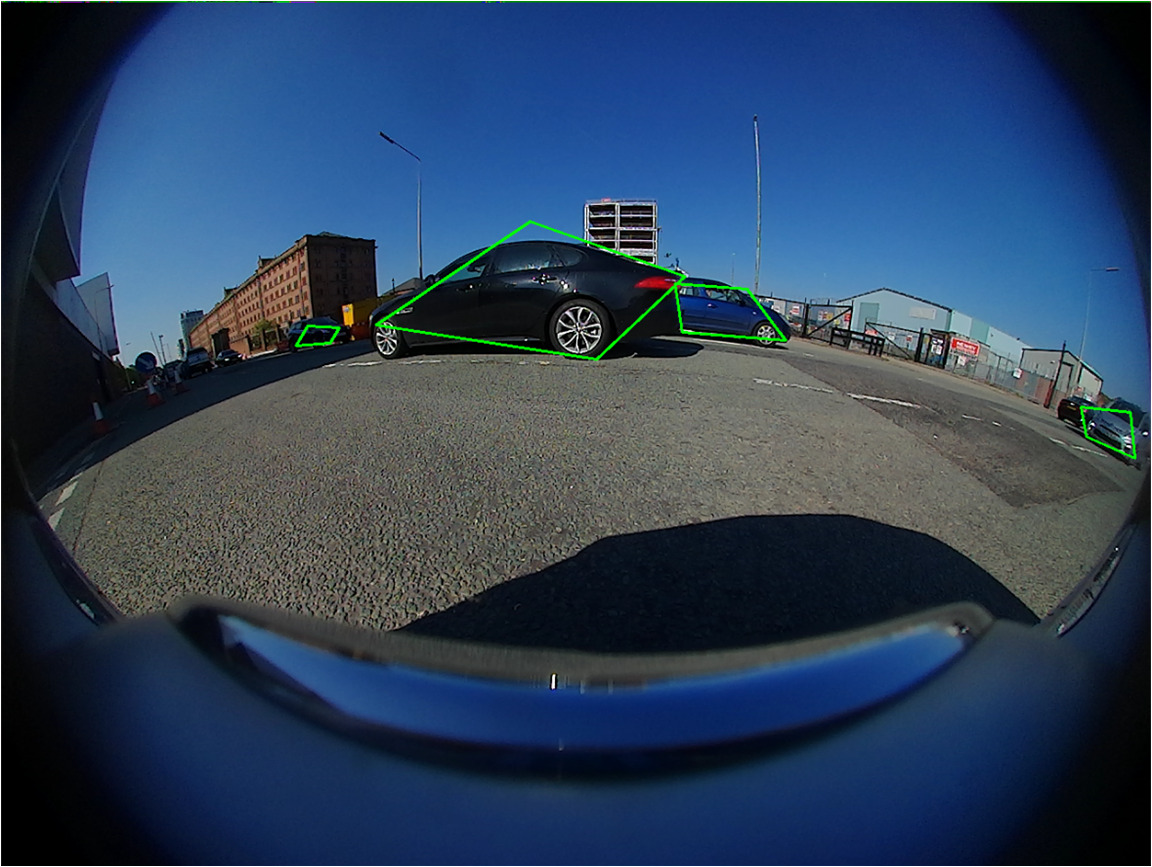} &
\includegraphics[height=\turnheightnew]{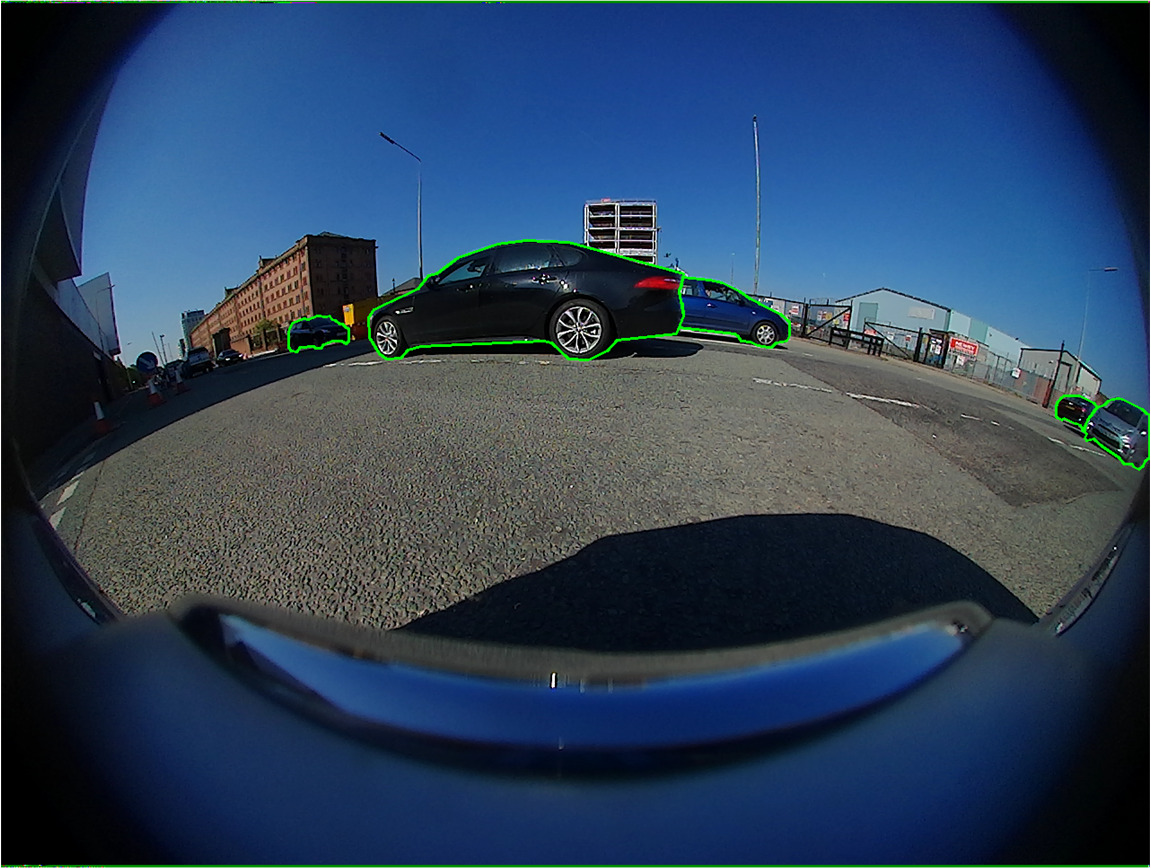} \\

\includegraphics[height=\turnheightnew]{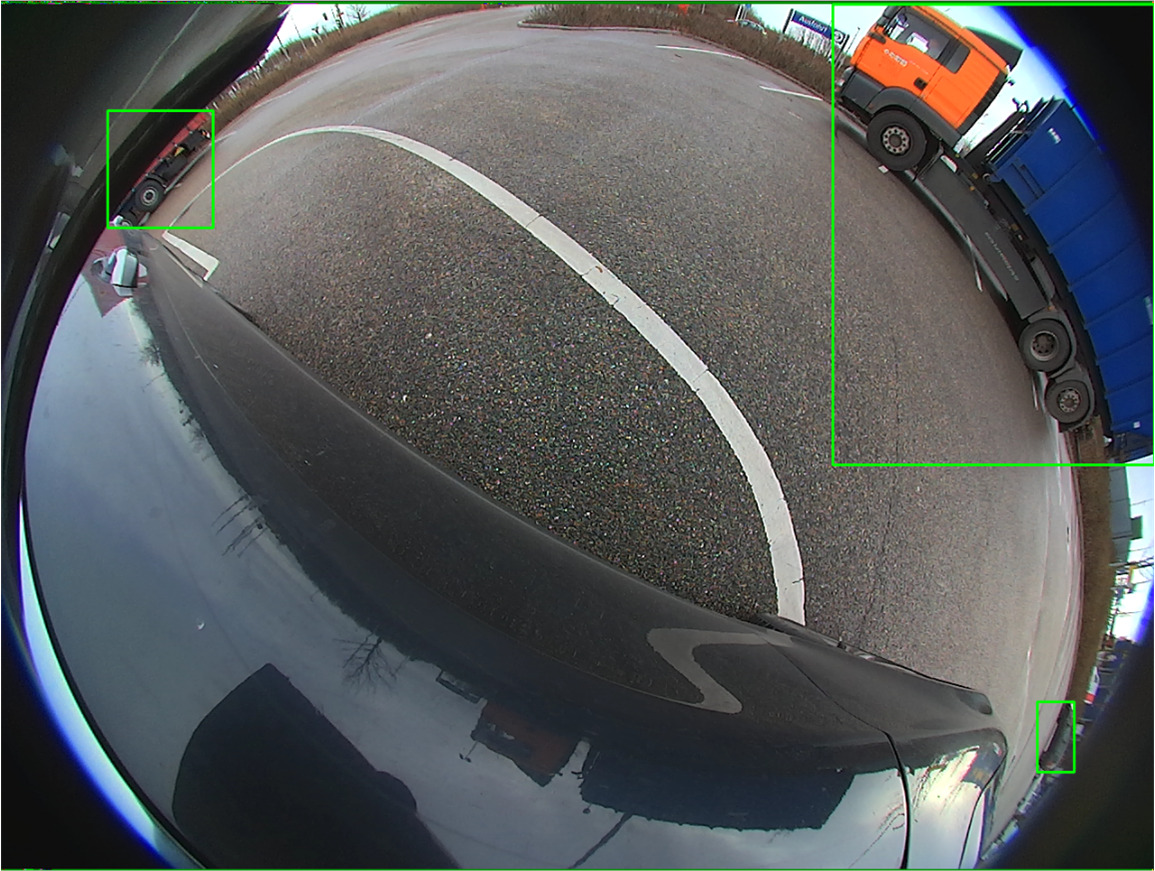} &
\includegraphics[height=\turnheightnew]{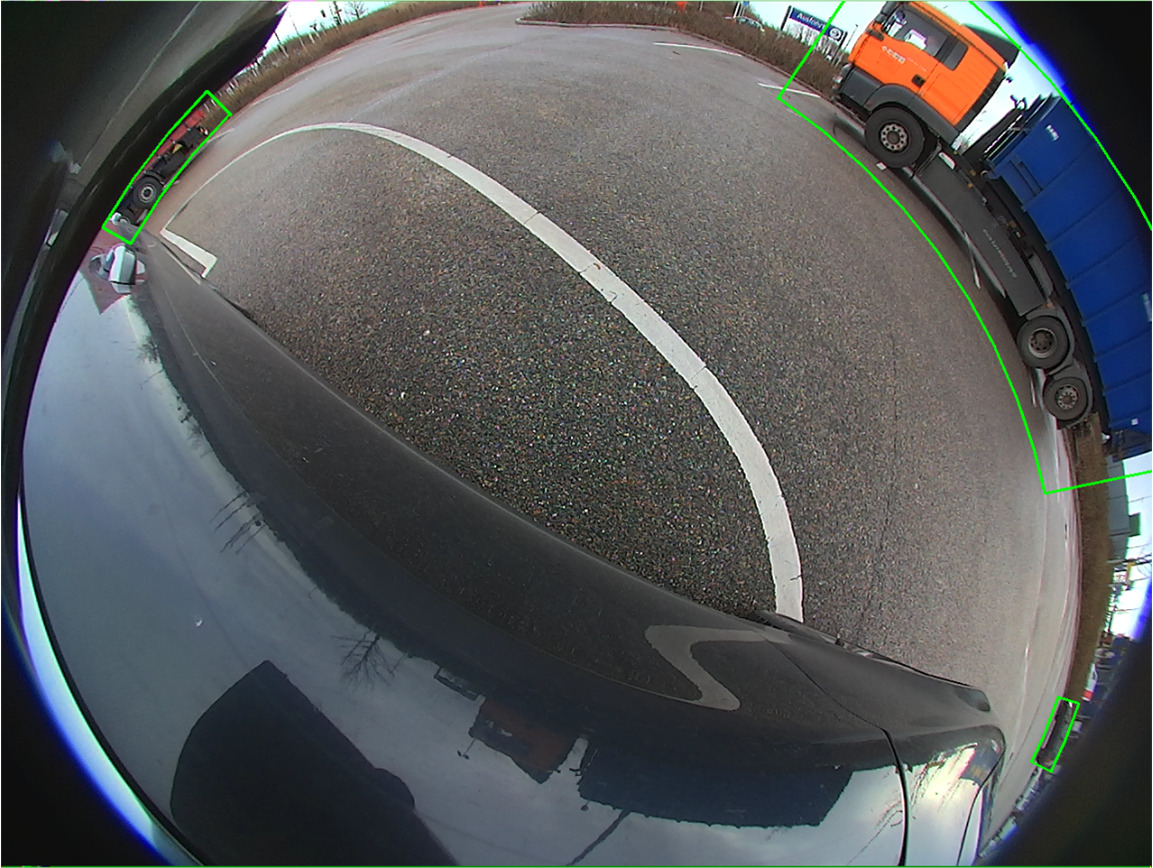} &
\includegraphics[height=\turnheightnew]{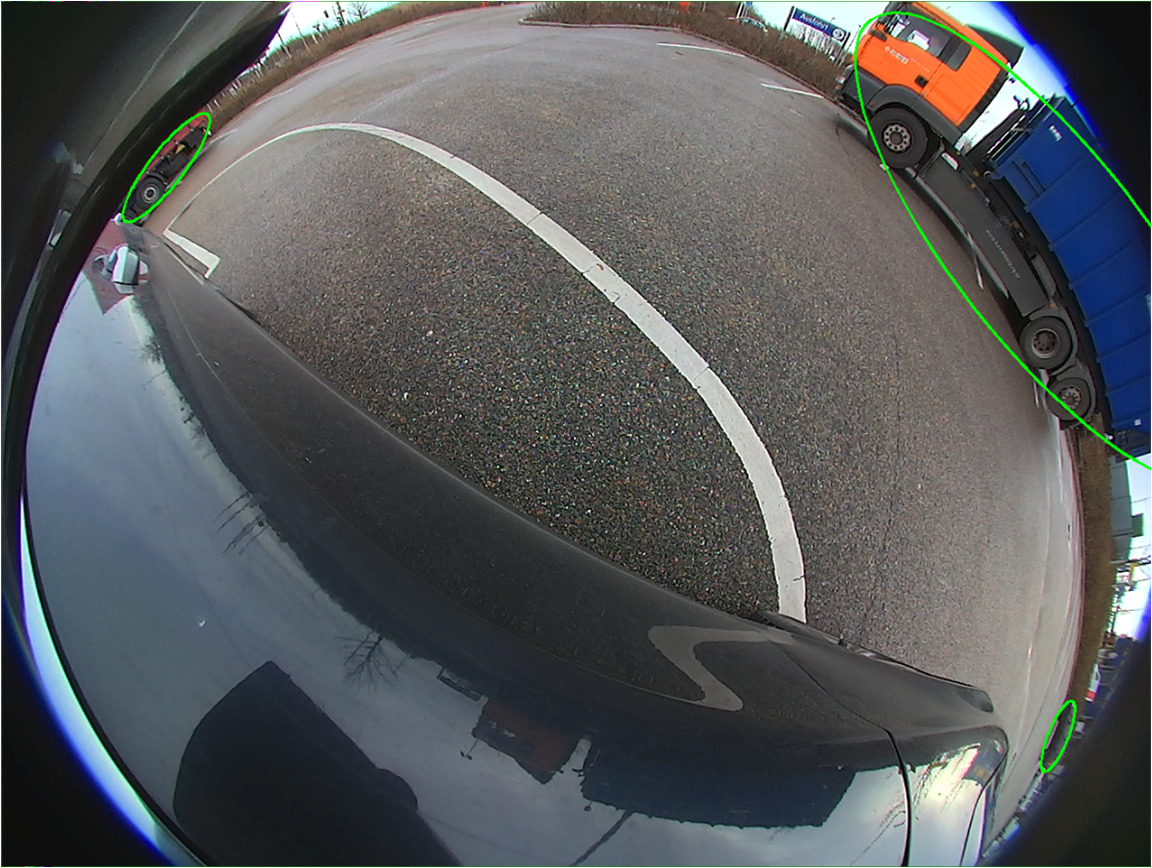} \\

\includegraphics[height=\turnheightnew]{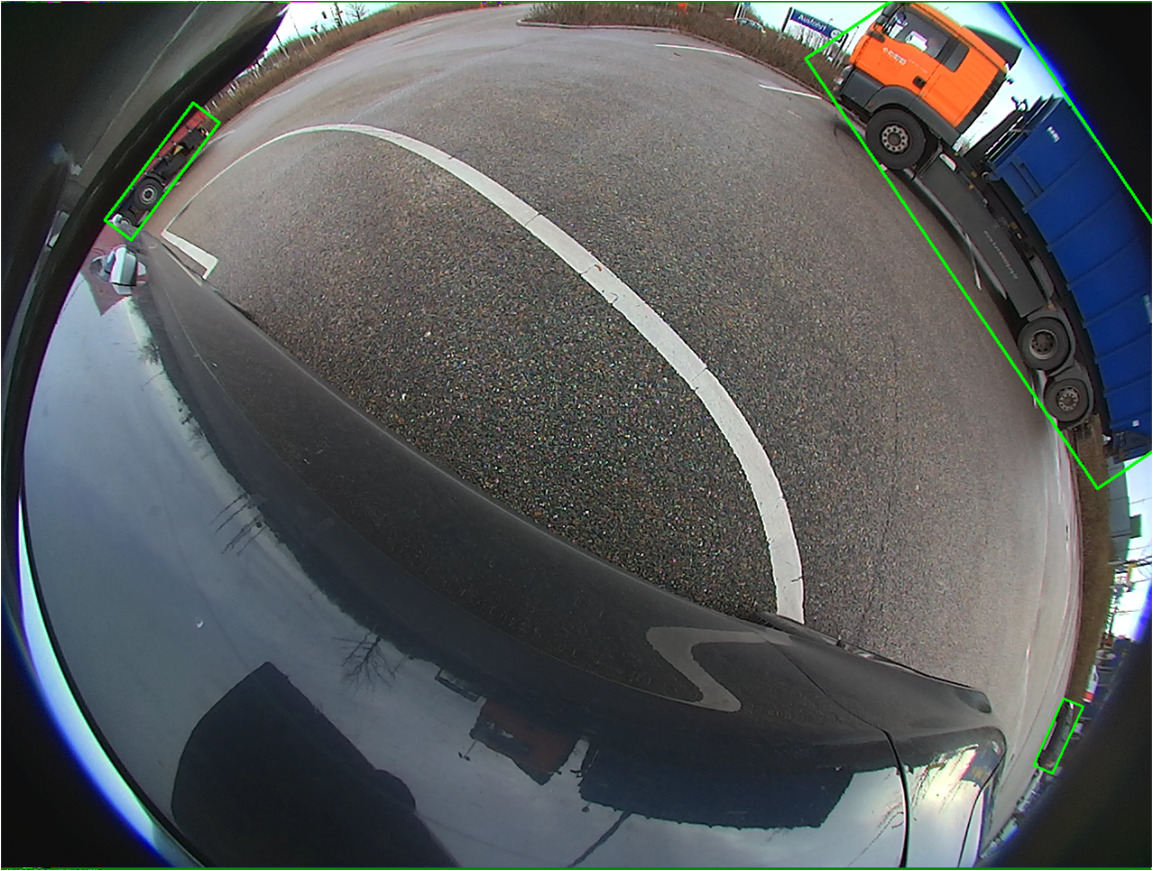} &
\includegraphics[height=\turnheightnew]{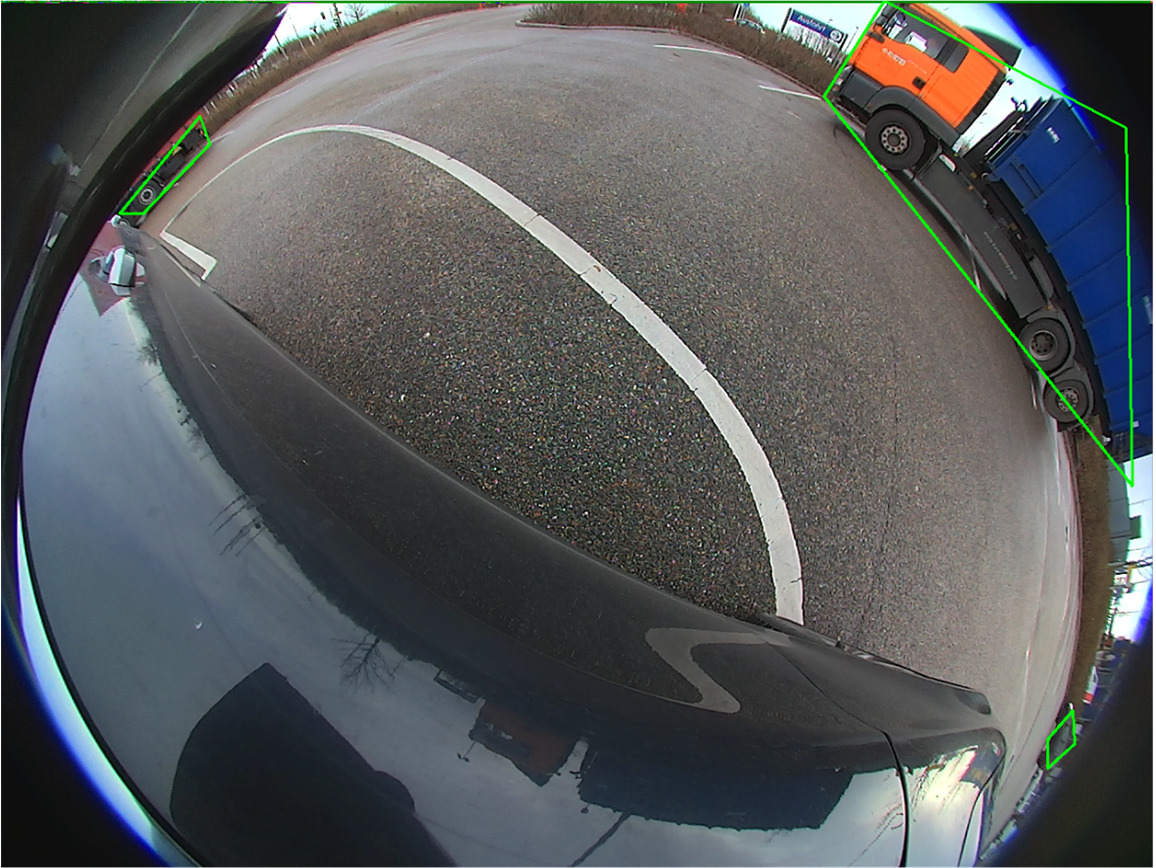} &
\includegraphics[height=\turnheightnew]{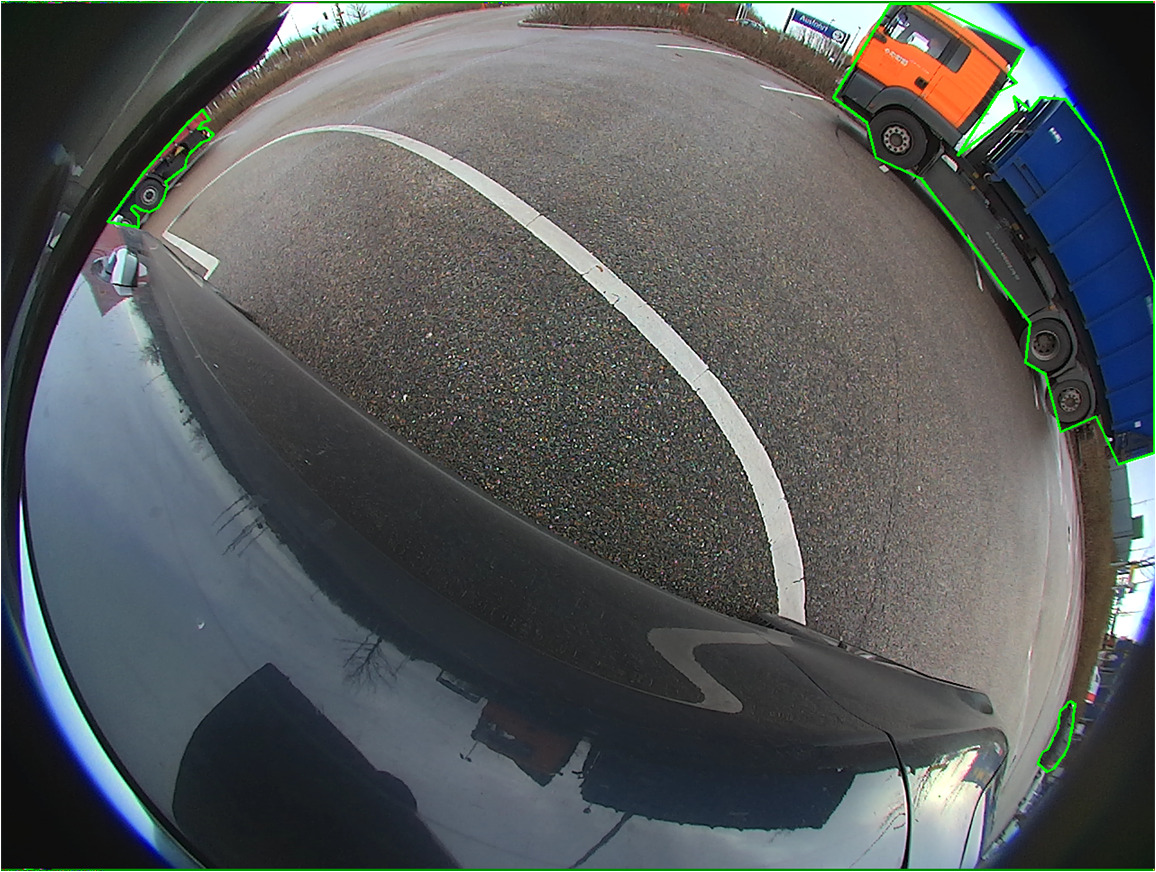} \\

\includegraphics[height=\turnheightnew]{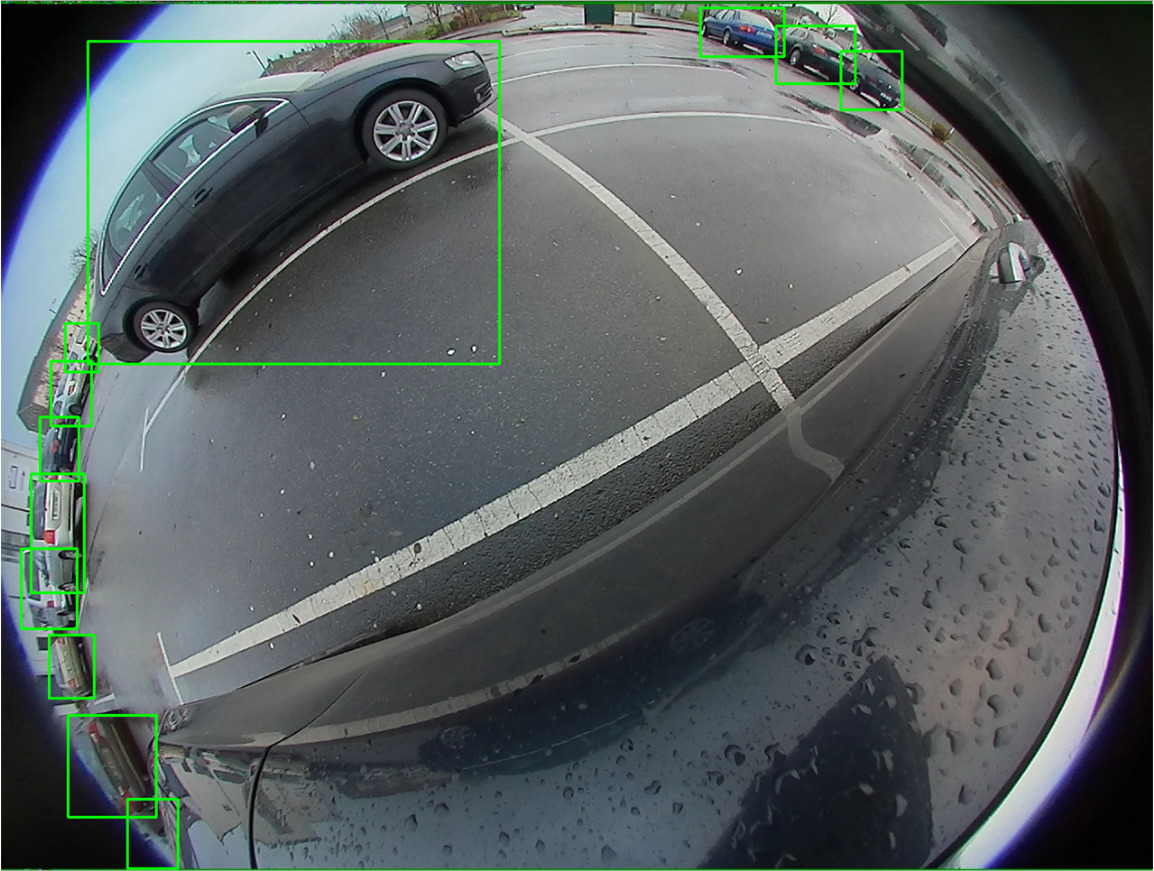} &
\includegraphics[height=\turnheightnew]{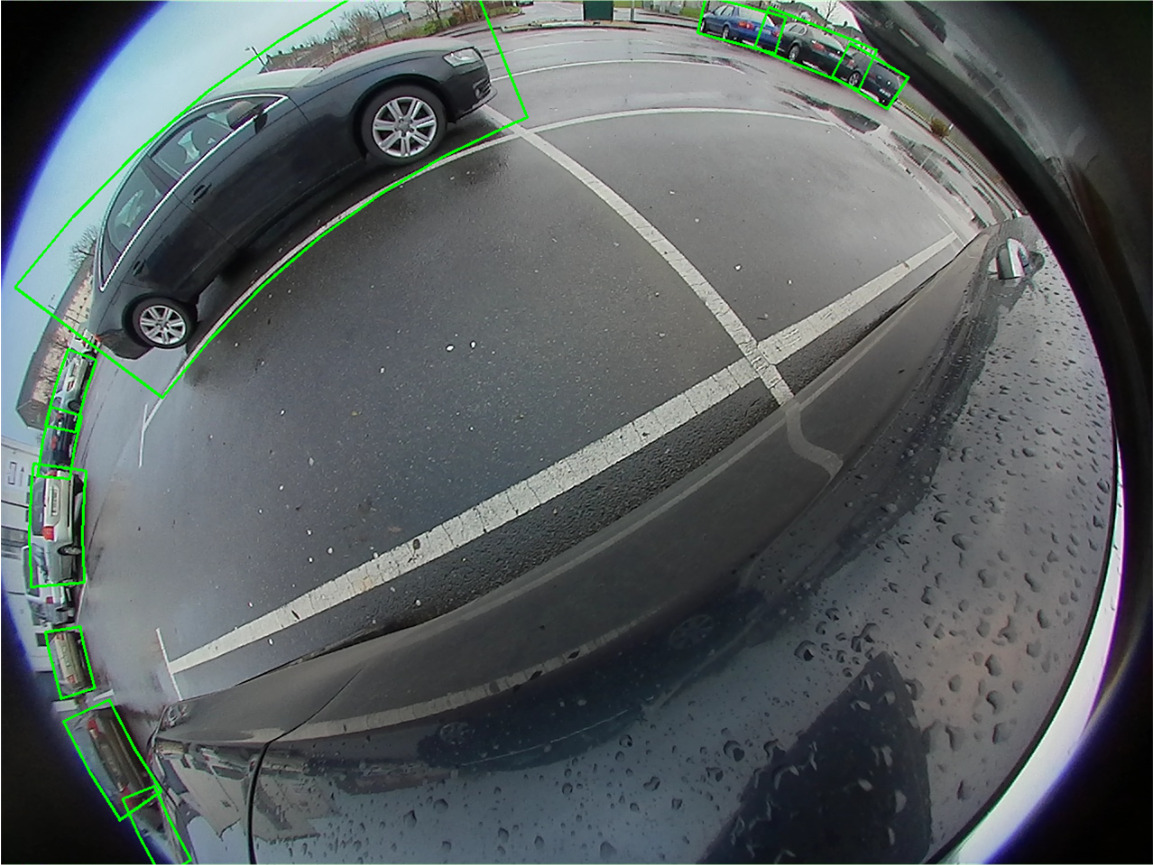} &
\includegraphics[height=\turnheightnew]{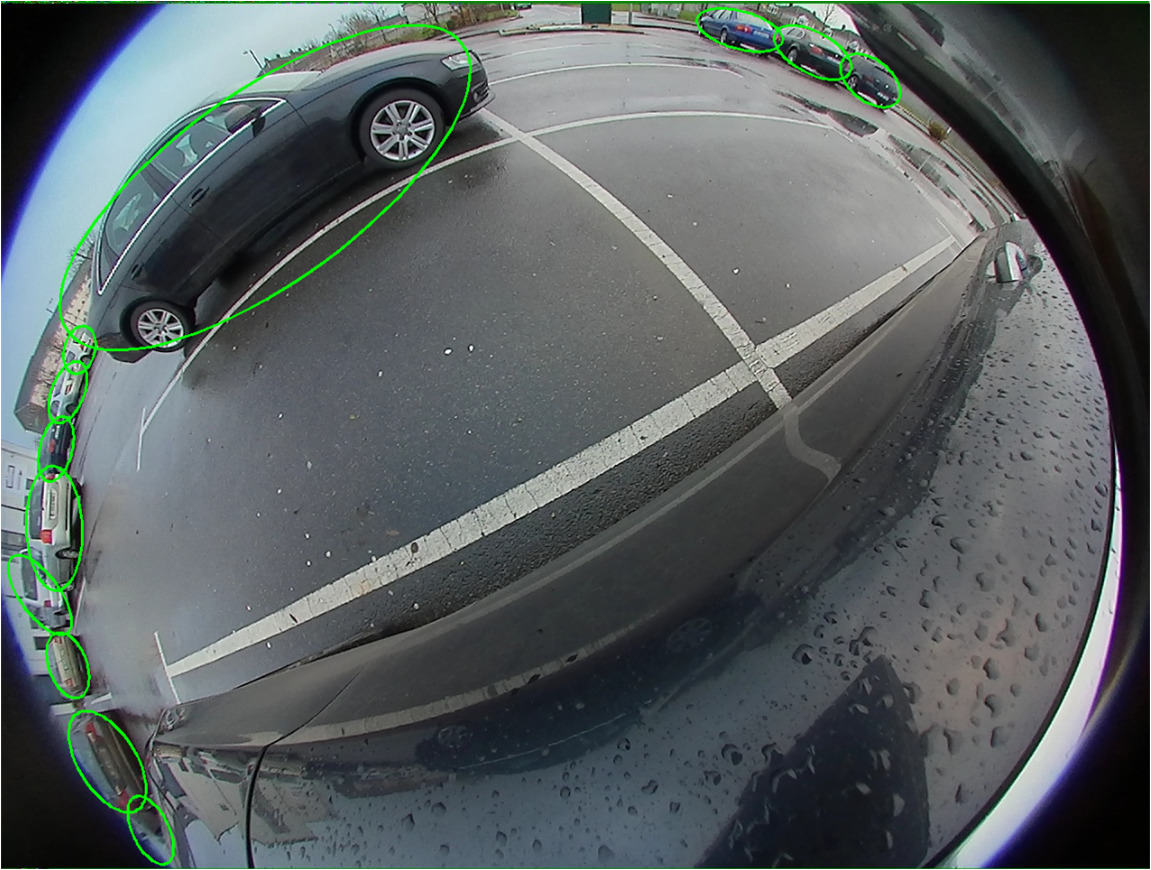} \\

\includegraphics[height=\turnheightnew]{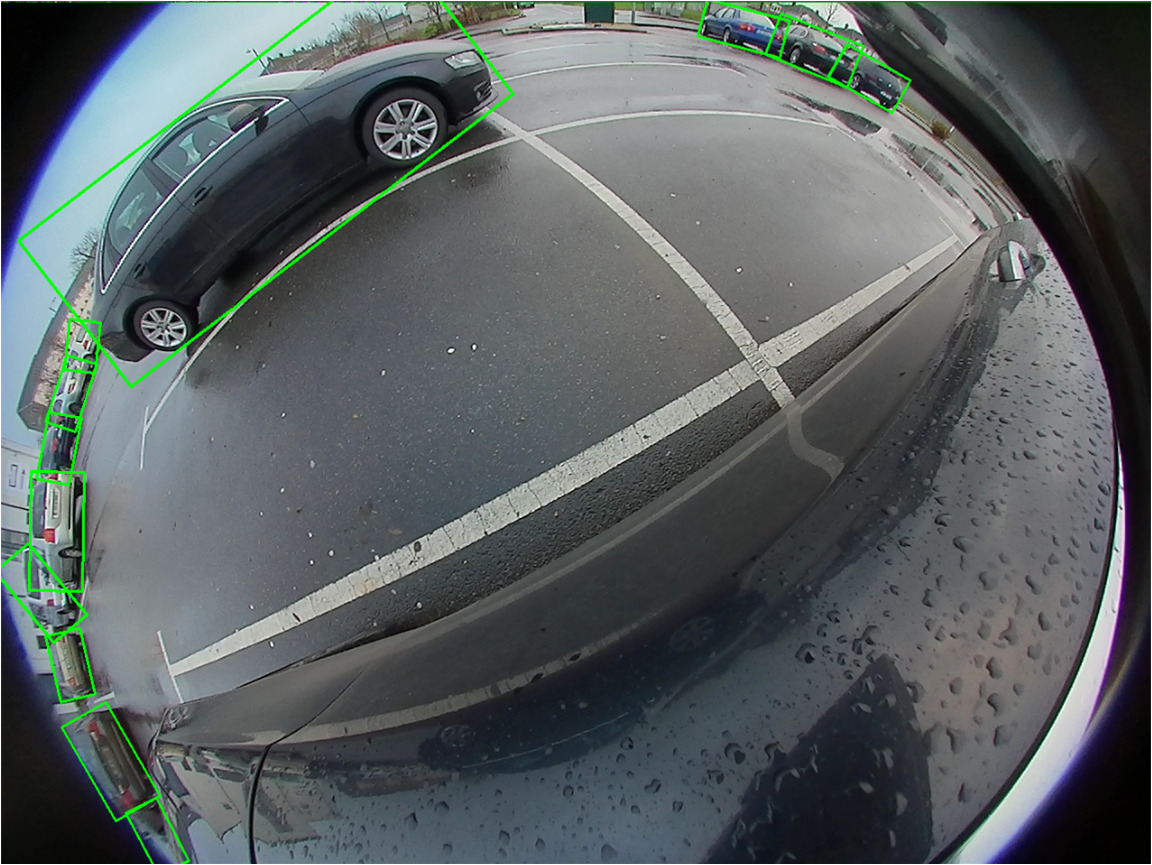} &
\includegraphics[height=\turnheightnew]{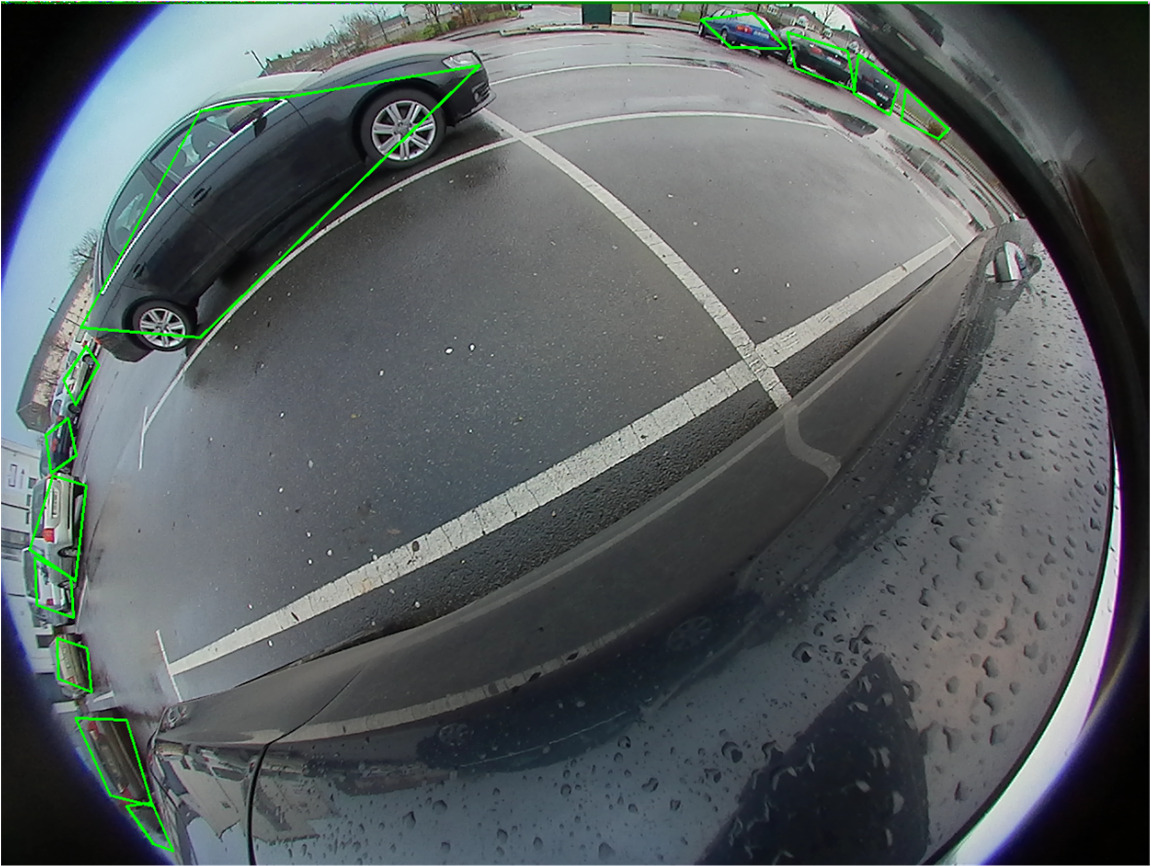} &
\includegraphics[height=\turnheightnew]{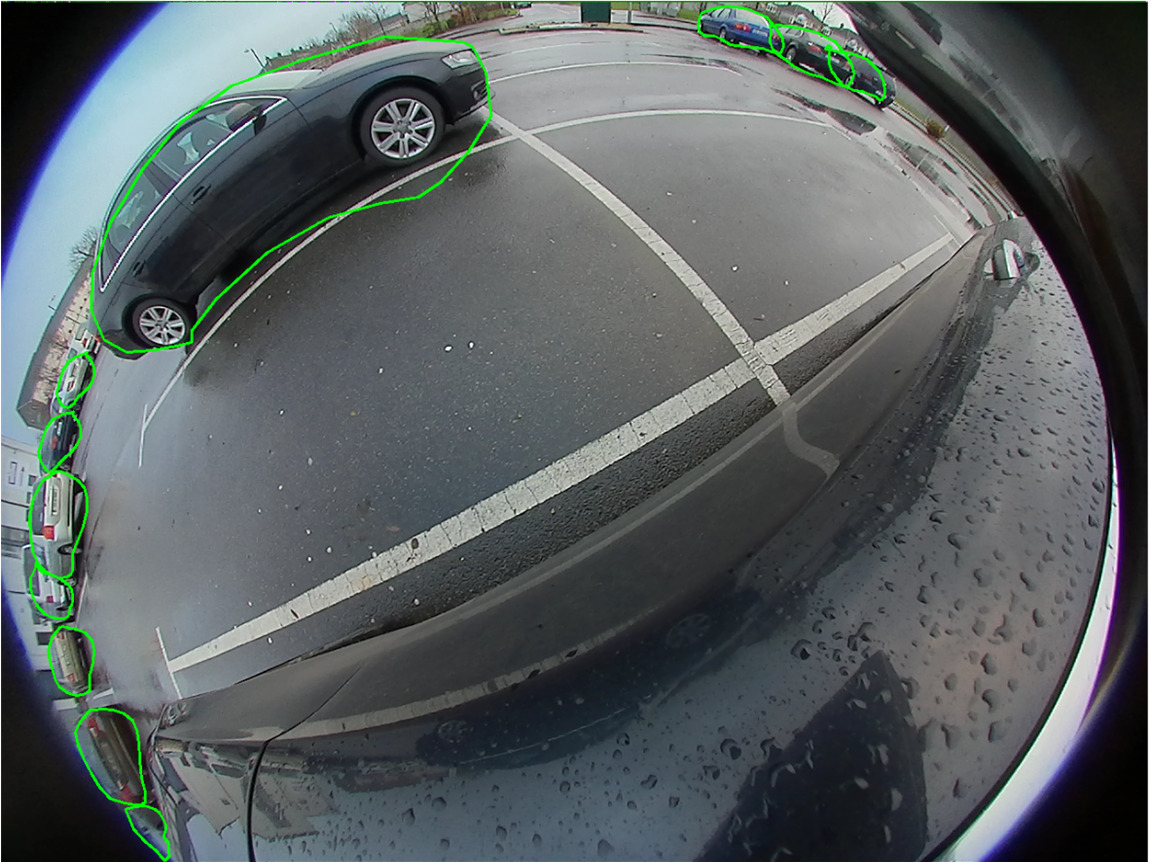}

\end{tabular}
\end{adjustbox}
  \caption[\bf Further qualitative results of the proposed model for different output representations.]
  {\textbf{Further qualitative results of the proposed model for different output representations.} The 1\textsuperscript{st}, 3\textsuperscript{rd} and 5\textsuperscript{th} rows show outputs for the Standard box, Oriented box and Ellipse representations. The 2\textsuperscript{nd}, 4\textsuperscript{th} and 6\textsuperscript{th} rows indicate Curved box and 4-point polygon. 24-point polygon representations. For more qualitative results, we refer to this video: \url{https://youtu.be/iLkOzvJpL-A}.}
  \label{fig:qualitative}
\end{figure*}

\begin{table}[!t]
\centering
\begin{adjustbox}{width=0.7\columnwidth}
\begin{tabular}{l|cccc|c}
\toprule
\multicolumn{1}{l|}{\textit{\textbf{Representation}}}
& \multicolumn{4}{c|}{\cellcolor[HTML]{7d9ebf} \textbf{IoU}}
& \multicolumn{1}{l}{\cellcolor[HTML]{e8715b} \textbf{mIoU}} \\
  \cline{2-5} 
  \multicolumn{1}{l|}{} &
  \multicolumn{1}{l|}{Front} &
  \multicolumn{1}{l|}{Rear} &
  \multicolumn{1}{l|}{Left} &
  \multicolumn{1}{l|}{Right} & \\ 
\midrule
YoloV3       & 32.5 & 32.1 & 34.2 & 27.8 & 31.6 \\
Curved Box   & 33.0 & 32.7 & 35.4 & 28.0 & 32.3 \\
Oriented Box & 33.9 & 33.5 & 37.2 & 30.1 & 33.6 \\
Ellipse      & 35.4 & 35.4 & 40.4 & 30.5 & 35.4 \\
24-sided Polygon & \textbf{44.4} & \textbf{46.8} & \textbf{44.7} & \textbf{42.7} & \textbf{44.65} \\
\bottomrule
\end{tabular}
\end{adjustbox}
\caption{\bf Quantitative results of the proposed model using different bounding box representations on the WoodScape dataset.}
\label{tab:prediction_table}
\end{table}
\begin{table}[!t]
\centering
\scalebox{0.7}{
\begin{adjustbox}{width=0.7\columnwidth}
\begin{tabular}{@{}lc@{}}
\toprule
\textit{\textbf{Representation}} & 
\textbf{mAP} \\ 
\midrule
\multicolumn{2}{c}{\cellcolor[HTML]{7d9ebf} Oriented Box} \\
\midrule
Orientation regression                & 39.0          \\
Orientation classification            & 40.6          \\
Orientation classification + IoU loss & \textbf{41.9} \\
\midrule
\multicolumn{2}{c}{\cellcolor[HTML]{e8715b} 24-sided Polygon} \\
\midrule
Uniform Angular    & 55.6   \\
Uniform Perimeter  & 55.4 \\
Adaptive Perimeter & \textbf{58.1} \\ 
\bottomrule
\end{tabular}
\end{adjustbox}
}
\caption{\bf Ablation study on the number of parameters in the oriented bounding box and the 24-point polygon representation.}
\label{tab:ablation}
\end{table}
Table~\ref{tab:prediction_table} demonstrates the prediction results on the proposed representations. The experiments are performed on the best performing model according to Table~\ref{tab:gt_table} and Table~\ref{tab:ablation}. Compared to the standard bounding box approach, the proposed oriented box and ellipse models improve the mIoU score on the test set by {2\%}, {1.8\%} respectively.
Ellipse prediction provides slightly better accuracy than the oriented box as it is not as sensitive to occlusions with other objects in the scene due to the absence of corners, which is demonstrated in Figure~\ref{fig:qualitative}.\par
\begin{figure*}[!t]
  \centering
  \newcommand{\turnheightnew}{0.25\columnwidth}

\begin{adjustbox}{max size={\textwidth}{\textheight}}
\begin{tabular}{@{\hskip 0.4mm}c@{\hskip 0.4mm}c@{\hskip 0.4mm}c@{\hskip 0.4mm}}

\includegraphics[height=\turnheightnew]{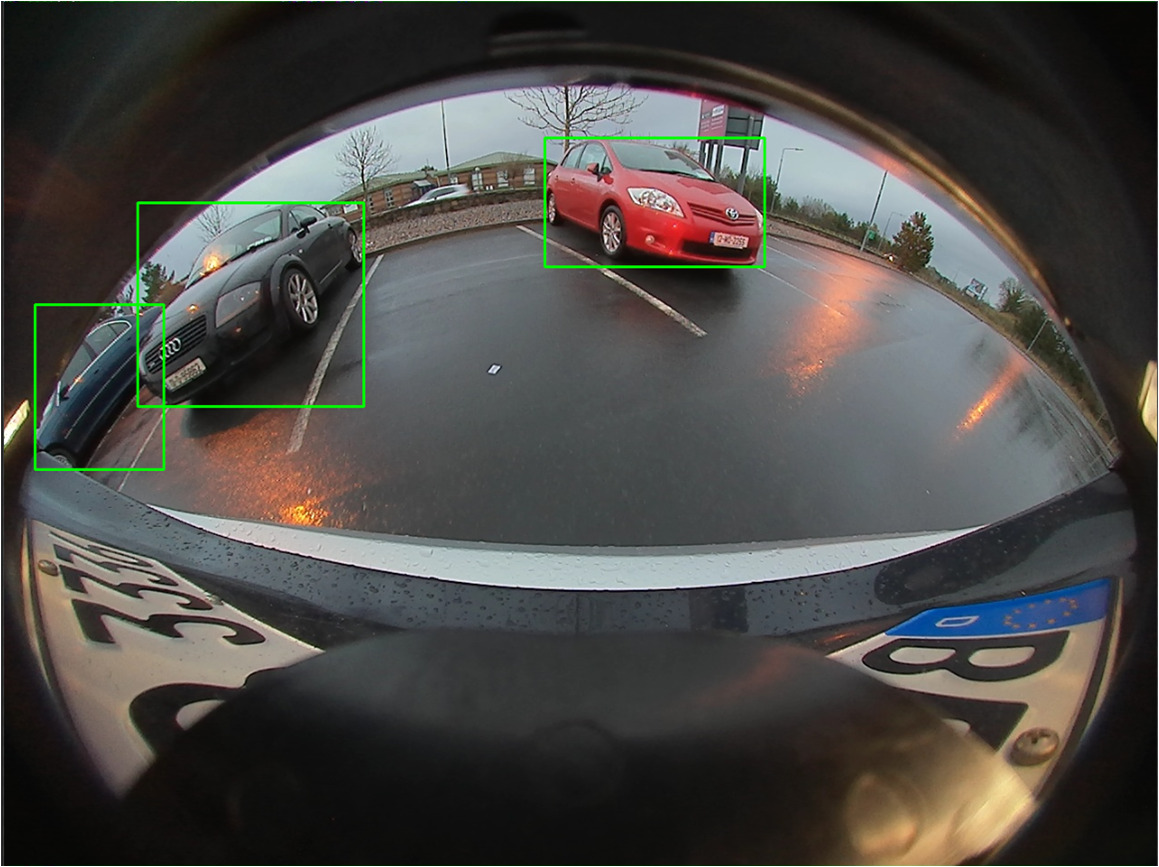} &
\includegraphics[height=\turnheightnew]{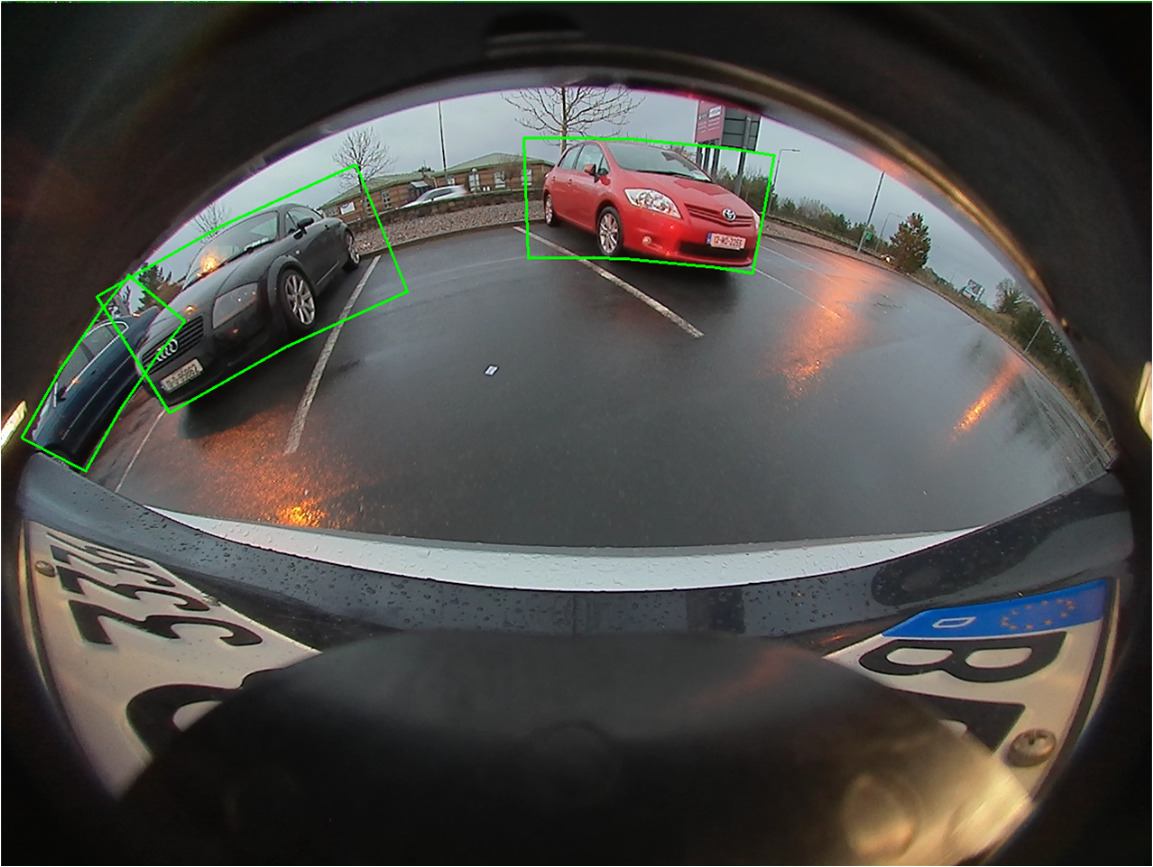} &
\includegraphics[height=\turnheightnew]{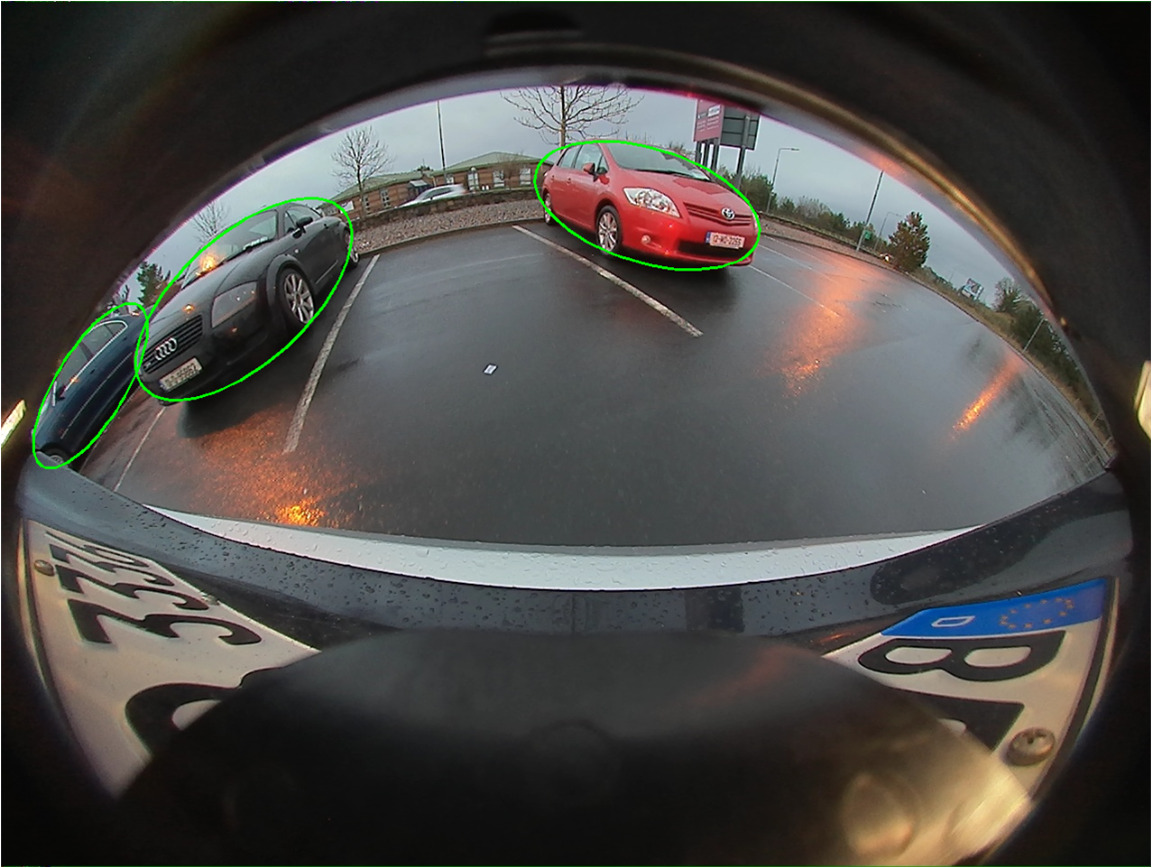} \\

\includegraphics[height=\turnheightnew]{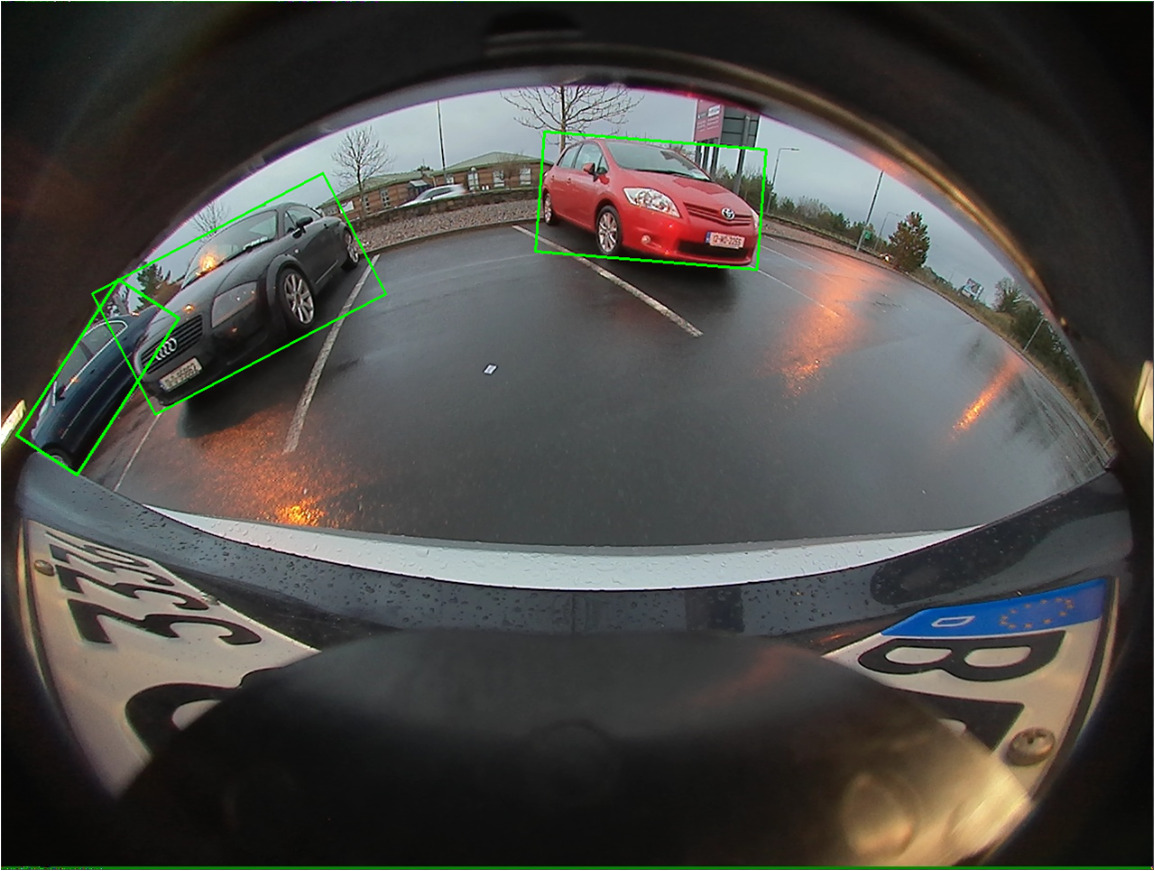} &
\includegraphics[height=\turnheightnew]{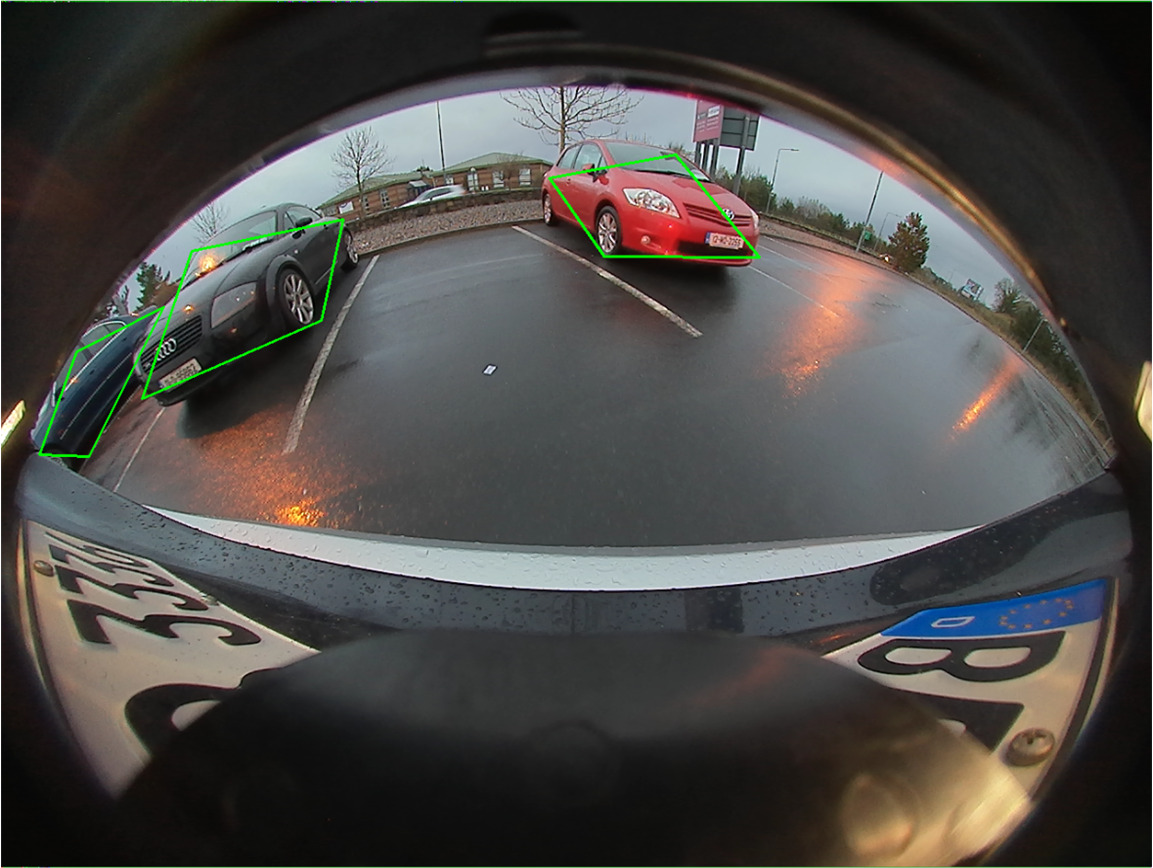} &
\includegraphics[height=\turnheightnew]{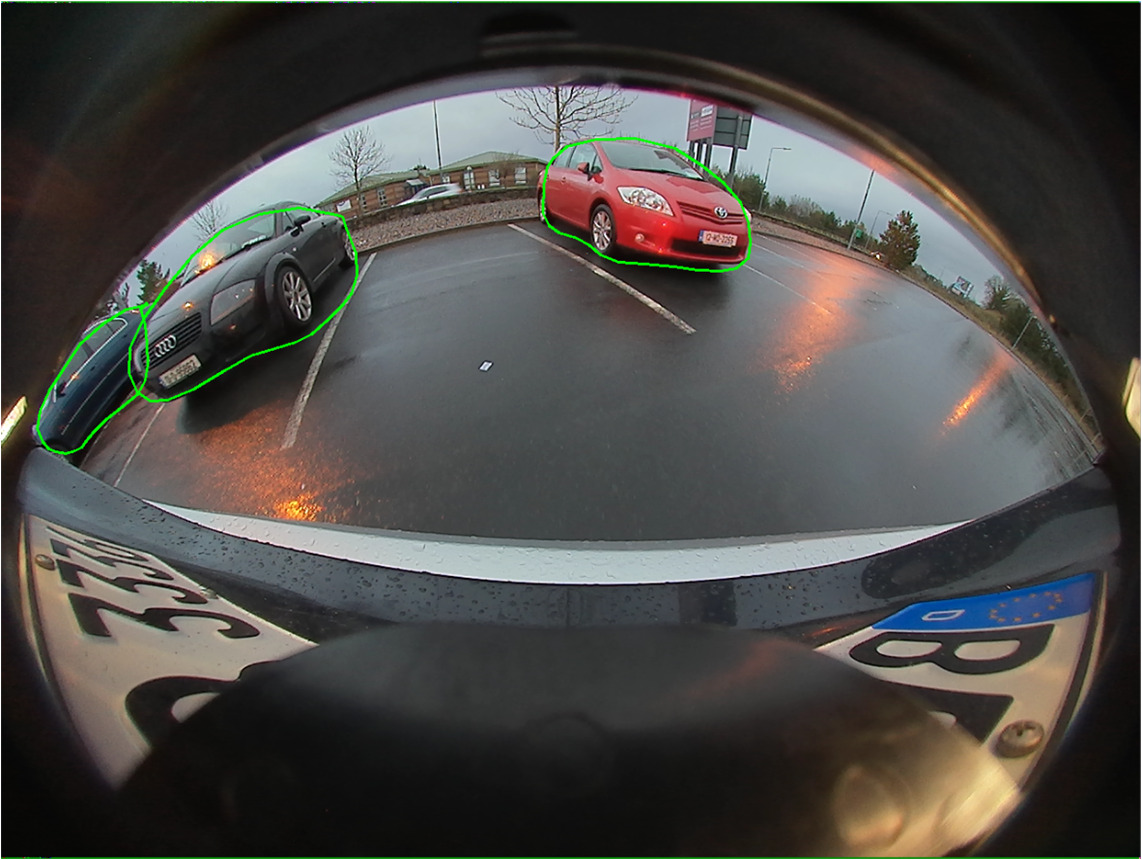}

\end{tabular}
\end{adjustbox}
  \caption[\bf Qualitative results of the proposed model for different output representations.]{\textbf{Qualitative results of the proposed model for different output representations.} The 1\textsuperscript{st} row shows the Standard box, Oriented box and Ellipse representations. The 2\textsuperscript{nd} row shows the Curved box, 4-point polygon and 24-point polygon representation.}
  \label{fig:qualitative_1}
\end{figure*}

Figure~\ref{fig:qualitative_1} and Figure~\ref{fig:qualitative} shows a visual evaluation of the proposed representations. Results show that the ellipse provides a decent easy-to-learn representation with a minimum number of parameters and minimum occlusion with the background objects compared to the oriented box representation. Unlike boxes, it allows a minimal representation for the object due to the absence of corners, which for instance, avoids incorrect occlusion with free parking slots, as shown in Figure~\ref{fig:qualitative_1} (Bottom). Polygon representation provides higher accuracy in terms of IoU with the instance mask. A four-point model provides high accuracy predictions with small objects as 4 points are sufficient to represent them. As the dataset contains a lot of small objects this representation demonstrates a good accuracy, which is shown in Tables~\ref{tab:gt_table} and \ref{tab:ablation}. Visually, large objects cannot be represented by a quadrilateral, as illustrated in Figure~\ref{fig:qualitative}. A higher number of sampling points on the polygon results in higher performance. However, the predicted masks are still prone to deformation due to minor errors in each point's localization.\par
\section{Conclusion}

In this chapter, we studied various representations for fisheye object detection. At a high level, we can split them into bounding box extensions and generic polygon representations. We explored the oriented bounding box and ellipse representations. Additionally, we designed a curved bounding box with optimal fisheye distortion properties. We proposed a curvature adaptive sampling method for polygon representations, which improves significantly over uniform sampling methods. Overall, the proposed model improves the relative mIoU accuracy significantly by {40\%} compared to a YOLOv3 baseline. We consider our method to be a baseline for further research into this area and will make the dataset with ground truth annotation for various representations publicly available. We hope this encourages further research in this area leading to a more mature object detection on raw fisheye imagery.\par

The final chapter will look into a holistic scene understanding of an autonomous car's environment using monocular fisheye camera videos. All the previous sensory sub-systems' perception algorithms will be integrated into a single fully functional real-time capable system. We will develop a framework that can reason jointly about geometry, motion, and semantics in order to estimate depth accurately, semantic segmentation and motion segmentation and localize in the real world with 2D object detection.\par

\chapter{Holistic 360° Scene Understanding}
\label{Chapter7}
\minitoc
\section{Introduction}

Surround-view fisheye cameras have been deployed in premium cars for over ten years, starting from visualization applications on dashboard display units to providing near-field perception for automated parking. Fisheye cameras have a strong radial distortion that cannot be corrected without disadvantages, including reduced FoV and re-sampling distortion artifacts at the periphery \cite{kumar2020unrectdepthnet}. Appearance variations of objects are larger due to the spatially variant distortion, particularly for close-by objects. It has been paramount to autonomous systems to comprehensively understand the surrounding environment using fisheye cameras. This chapter presents a multi-task visual perception network on unrectified fisheye images to enable the vehicle to sense its surrounding environment. It consists of six primary tasks necessary for an autonomous driving system: depth estimation, visual odometry, semantic segmentation, motion segmentation, object detection, and lens soiling detection.\par

In recent years DNNs have accomplished impressive success in various applications, including autonomous driving perception tasks. On the other hand, current deep neural networks are easily fooled by adversarial attacks. This vulnerability raises significant concerns, particularly in safety-critical applications. As a result, research into attacking and defending DNNs has gained much coverage. This chapter presents a detailed adversarial attack applied to the diverse multi-task visual perception. In the experiments, we consider both white and black box attacks for targeted and untargeted cases while attacking a task and inspecting the effect on all the others, in addition to inspecting the effect of applying a simple defense method.\par

This work was very influential from a product perspective to win next-generation projects and be influential in the academic community. This work was formally presented as \textit{OmniDet}~\cite{kumar2021omnidet} in a journal and a conference at the \href{https://arxiv.org/abs/2008.04017}{RA-L + ICRA} in 2021. The ablation on \textit{Adversarial Attacks}~\cite{sobh2021adversarial, klingner2022detecting} was presented at the \href{https://arxiv.org/abs/2008.04017}{ITSC} in 2021. To encourage further research in developing multi-task perception algorithms, the code was made public on the \href{https://github.com/valeoai/WoodScape}{Github}, which proved to be quite popular in the vision community and significantly helped the discernibility of the approach. Further details about the dataset usage and demo code can be found on the WoodScape website \url{https://woodscape.valeo.com}.\par
Autonomous Driving applications require various perception tasks to provide a robust system covering a wide variety of scenarios. Alternate ways to detect objects in parallel are necessary to achieve a high level of accuracy. For example, objects can be detected based on appearance, motion, and depth cues. Despite increasing computation power in automotive embedded systems, efficient design is always needed due to the increasing number of cameras and perception tasks. MTL is an efficient design pattern commonly used where most of the computation is shared across all the tasks~\cite{sistu2019neurall, vandenhende2020multitask}. Besides, learning features for multiple tasks can act as a regularizer, improving generalization. Recently, Mao~\etal~\cite{mao2020multitask} illustrated that multi-task learning improves adversarial robustness, which is critical for safety applications. In the automotive multi-task setting, MultiNet~\cite{teichmann2018multinet} was one of the first to demonstrate a three task network on the KITTI, and most further works have primarily worked on a three task setting.\par
\begin{figure}[!t]
  \resizebox{\columnwidth}{!}{
  \centering
\begin{tabular}{ccc}
 	\begin{overpic}[width=0.7\columnwidth]{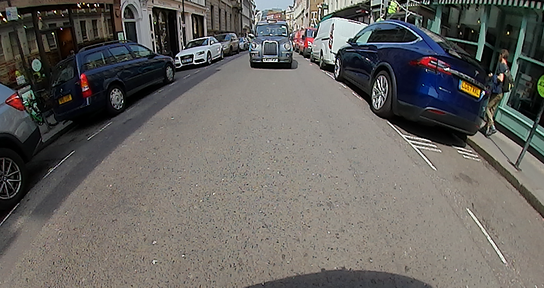}
    \put (0,2) {\colorbox{lightgray}{$\displaystyle\textcolor{black}{\text{(a)}}$}}
    \end{overpic}
    \begin{overpic}[width=0.7\columnwidth]{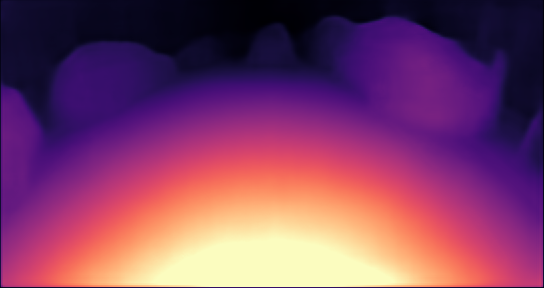}
    \put (0,2) {\colorbox{lightgray}{$\displaystyle\textcolor{black}{\text{(b)}}$}}
    \end{overpic} \\
    
    \begin{overpic}[width=0.7\columnwidth]{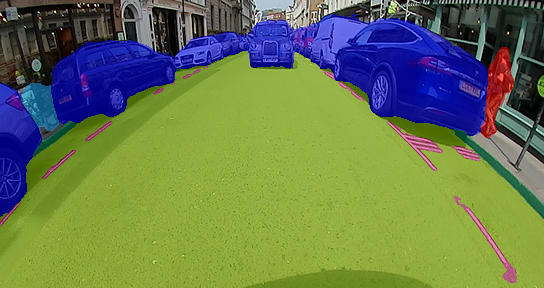}
    \put (0,2) {\colorbox{lightgray}{$\displaystyle\textcolor{black}{\text{(c)}}$}}
    \end{overpic} 
 	\begin{overpic}[width=0.7\columnwidth]{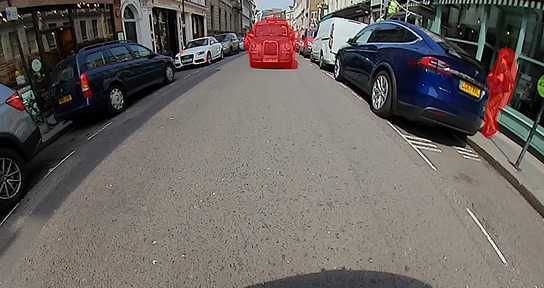}
    \put (0,2) {\colorbox{lightgray}{$\displaystyle\textcolor{black}{\text{(d)}}$}}
    \end{overpic} \\
    
    \begin{overpic}[width=0.7\columnwidth]{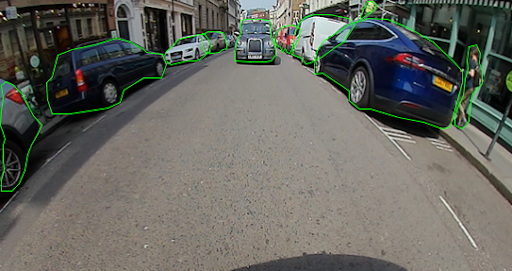}
    \put (0,2) {\colorbox{lightgray}{$\displaystyle\textcolor{black}{\text{(e)}}$}}
    \end{overpic}
    \begin{overpic}[width=0.7\columnwidth]{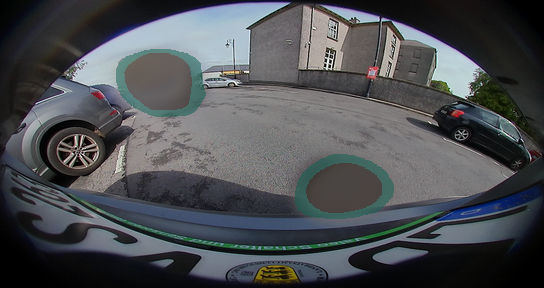}
    \put (0,2) {\colorbox{lightgray}{$\displaystyle\textcolor{black}{\text{(f)}}$}}
    \end{overpic}
\end{tabular}
}
\caption[\bf Real-time capable network estimates from the OmniDet framework on raw fisheye images.]
{\textbf{Real-time capable network estimates from the OmniDet framework on raw fisheye images.} (a) Rear-Camera Input Image, (b) Distance Estimate, (c) Semantic Segmentation, (d) Motion Estimation, (e) 24-sided Polygon Object Detection and (f) Soiling Segmentation.}
\label{fig:abstract}
\end{figure}
\section{Perception Tasks and Losses in MTL}

This thesis's final goal is to build a multi-task model covering the necessary modules in a \textit{Level 3} autonomous driving system for near-field sensing use cases such as Parking or Traffic Jam assist. This chapter builds upon the previous chapters focused on individual tasks. In general, there is minimal work in the area of fisheye perception. Specifically, there is only a research work on multi-task learning: FisheyeMultiNet~\cite{maddu2019fisheyemultinet} which discusses a simple three task network.\par

The perception system comprises semantic tasks, geometric tasks, and lens soiling detection (shown in Figure~\ref{fig:abstract}). The standard semantic tasks are object detection (pedestrians, vehicles, and cyclists) and semantic segmentation (road, lanes, and curbs). Fisheye cameras are mounted low on a vehicle and are susceptible to lens soiling due to splashing of mud or water from the road. Thus, it is vital to detect soiling on the camera lens and trigger a cleaning system~\cite{uvrivcavr2019soilingnet}.
The semantic tasks typically require a large annotated dataset covering various objects. It is practically infeasible practically to cover every possible object. Thus, generic object detection using geometric cues such as motion or depth for rare objects is typically used. They will also complement the detection of standard objects and provide more robustness. Thus we propose to include motion segmentation and depth estimation tasks. Motion is a dominant cue in automotive scenes, and it requires at least two frames or the use of dense optical flow~\cite{siam2018modnet}. Self-supervised methods have recently dominated depth estimation, which has also been demonstrated on fisheye images~\cite{kumar2020fisheyedistancenet}. Finally, the visual odometry task is required to place the detected objects in a temporally consistent map.\par
\subsection{Geometric Tasks}

\subsubsection{Self-Supervised Distance and Pose Estimation Networks}

We set up the self-supervised monocular SfM framework following Section~\ref{sec:fisheyedistancenet-framework} for distance estimation and pose estimation. View synthesis is performed by incorporating the polynomial projection model from Section~\ref{sec:modeling of fisheye geometry}. Section~\ref{sec:fisheyedistancenet-final-loss} describes the total self-supervised objective loss. Additionally, we include two more loss functions following~\cite{shu2020featdepth}. To prevent the training objective from getting stuck at multiple local minima for homogeneous areas, we incorporate feature-metric losses computed on $I_t$'s feature representations, where we learn the features using a self-attention autoencoder. The self-supervised loss landscapes are constrained to form proper convergence basins using the first-order $\mathcal{L}_{dis}$ and second-order derivatives $\mathcal{L}_{cvt}$ to regularize the target features. The total objective loss for distance estimation $\mathcal{L}_{dist}$ is calculated by averaging per pixel, scale, and image batch:
\begin{align}
    \mathcal{L}_{dist} &= \mathcal{L}_r(I_t,\hat{I}_{t'\to t}) + \beta~\mathcal{L}_s(\hat{D}_t) + \gamma~\mathcal{L}_{dc}(\hat{D}_t,\hat{D}_{t'}) \\
    &+ \mathcal{L}_r(\hat{F}_t,\hat{F}_{t'\to t}) + \omega~\mathcal{L}_{dis}(I_t, \hat{F}_t) 
    + \mu~\mathcal{L}_{cvt}(I_t, \hat{F}_t) \nonumber
    \label{eq:overall-loss}
\end{align}
where $\beta$, $\gamma$, $\omega$ and $\mu$ weigh the distance regularization $\mathcal{L}_s$, cross-sequence distance consistency $\mathcal{L}_{dc}$, discriminative $\mathcal{L}_{dis}$ and convergent $\mathcal{L}_{cvt}$ losses respectively. We calculate the image and feature reconstruction loss using the target $I_t$, estimated feature $\hat{F}_t$ frames, reconstructed target $\hat I_{t' \to t}$ and feature $\hat{F}_{t'\to t}$ frames. It is a linear combination of the general robust pixel-wise loss term~\cite{barron2019general} and the Structural Similarity (SSIM)~\cite{wang2004image} as described in~\cite{kumar2020syndistnet}.\par
\subsubsection{Discriminative loss} 
\label{sec:discriminative-loss}

Following~\cite{shu2020featdepth}, we define the feature representation by $\varphi_f(uv)$ with gradients $\frac{\partial \varphi(\widehat{uv})}{\partial \widehat{uv}}$ by ensuring that the learned features have relatively large slopes and gradients. For simplicity, the first-order derivative and second-order derivative with respect to image coordinates are denoted by $\nabla^1$ and $\nabla^2$, which equals $\partial_x+\partial_y$ and $\partial_{xx}+2\partial_{xy}+\partial_{yy}$ respectively. To do so, we constrain the first-order gradients of learned features and formulate it as $-\sum_p |\nabla^1 \varphi(uv)|_1$. As fisheye images have considerably larger homogeneous areas than rectilinear counterparts, this is an essential loss function that penalizes small slopes and emphasizes the low-texture regions using the image gradients. A higher penalty is imposed when similar prominent color areas are encountered.
\begin{equation}
    \mathcal{L}_{dis} = -\sum_{uv} e^{-|\nabla^1 I(uv)|_1} |\nabla^1 \varphi(uv)|_1
\end{equation}
$\mathcal{L}_{dis}$ forces $\varphi$ to be learned to satisfy the condition that $|\nabla^1 \varphi|_1$ should be relatively large at homogeneous areas, making gradient descent feasible at these regions as these regions receive larger weights. However, merely imposing the discriminative loss cannot guarantee that we move to the optimal solution during the gradient descent.\par

Merely replacing photometric values with features does not grasp the essence; we wish to have a landscape well-suited for optimization. For deep learning-based methods, whose optimization is mainly based on gradient descent approaches, two main factors of loss functions influence gradient descent's performance. Firstly, the right first-order gradients (slope) to ensure the right optimization direction and the small enough second-order gradients (curvature) to ensure a large convergence radius. However, the commonly used photometric loss cannot fully meet the above two requirements. Due to the raw image intensity limitations, photometric loss tends to have near-zero first-order gradients at low-texture regions. And due to the non-convex property of the image, the convergence radius of the photometric loss is very small (usually 1-2 pixels).\par
\subsubsection{Convergent loss}
\label{sec:converget-loss}

Since inconsistency exists among first-order gradients, \ie, spatially adjacent gradients point in opposite directions. Shu~\etal~\cite{shu2020featdepth} proposed a convergent loss $\mathcal{L}_{cvt}$ to have a relatively large convergence radius to enable gradient descent from a remote distance. This is achieved by formulating the loss to have consistent gradients during the optimization step by encouraging the smoothness of feature gradients and large convergence radii accordingly. Curvature is opposite to convergence radius, \ie; a large curvature corresponds with a small convergence radius, vice versa.
This property is of great importance for deep learning based methods; since their learning rates are pre-defined, large learning rates may step over a small convergence radius. $\mathcal{L}_{cvt}$ forces $\phi$ to have small curvatures, meanwhile encouraging the smoothness of learned feature gradients. $\mathcal{L}_{cvt}$ is formulated to penalize the second-order gradients, \ie, 
\begin{equation}
    \mathcal{L}_{cvt} =\sum_{uv} |\nabla^2 \varphi(uv)|_1
\end{equation}
\subsection{Generalized Object Detection}

\begin{figure*}[!ht]
  \centering
  \resizebox{\textwidth}{!}{\newcommand{\turnheightnew}{0.4\columnwidth}
\centering

\begin{tabular}{@{\hskip 0.4mm}c@{\hskip 0.4mm}c@{\hskip 0.4mm}c@{}}

{\rotatebox{90}{\hspace{13mm} \large Front Camera}} &
\includegraphics[height=\turnheightnew]{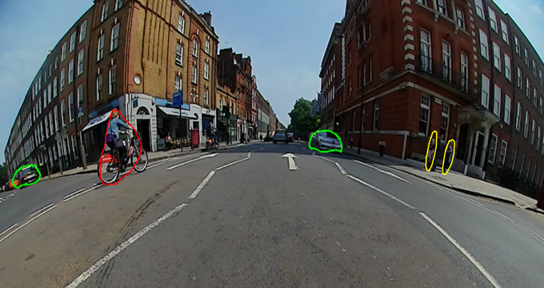} &
\includegraphics[height=\turnheightnew]{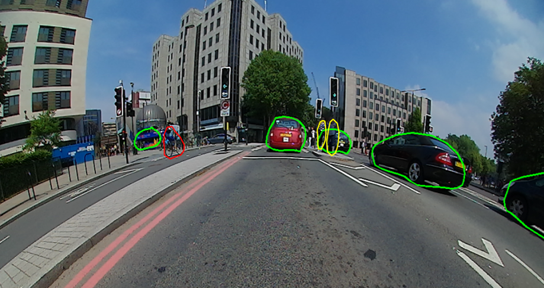} \\

{\rotatebox{90}{\hspace{13mm} \large Left Camera}} &
\includegraphics[height=\turnheightnew]{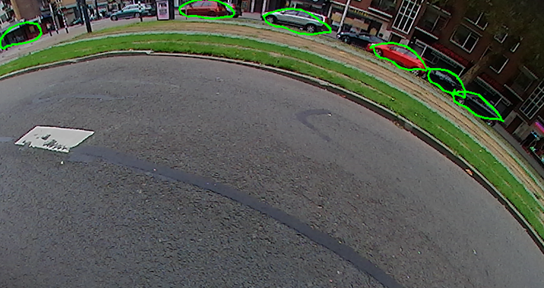} &
\includegraphics[height=\turnheightnew]{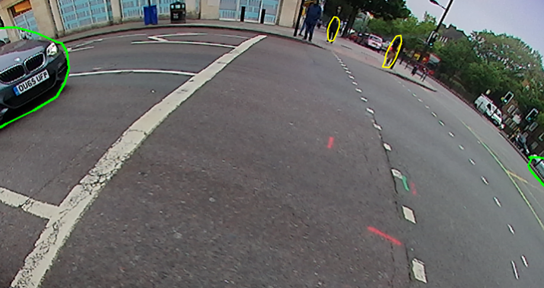} \\

{\rotatebox{90}{\hspace{13mm} \large Right Camera}} &
\includegraphics[height=\turnheightnew]{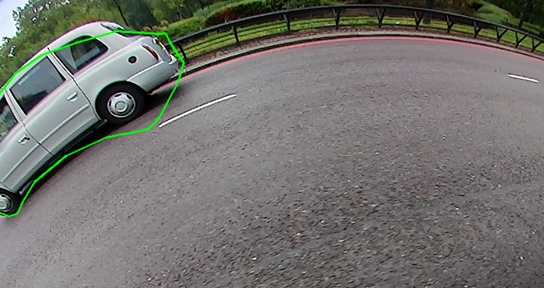} &
\includegraphics[height=\turnheightnew]{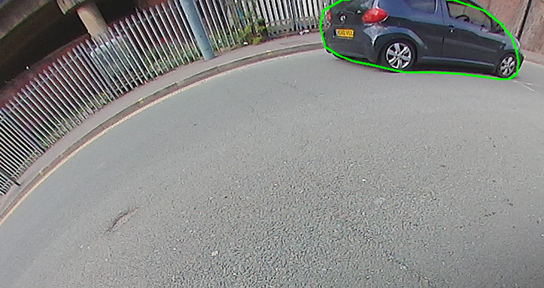} \\

{\rotatebox{90}{\hspace{13mm} \large Rear Camera}} &
\includegraphics[height=\turnheightnew]{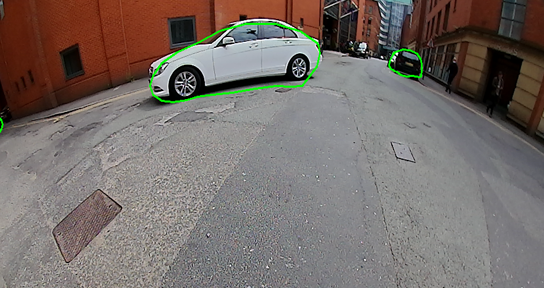} &
\includegraphics[height=\turnheightnew]{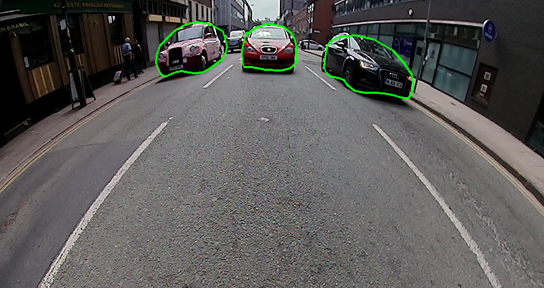}

\end{tabular}}
  \caption{\bf Qualitative results of 24-sided polygon-based objection detection on the WoodScape dataset.}
  \label{fig:omnidet-fisheye-yolo}
\end{figure*}
As discussed in the previous chapter~\ref{Chapter6}, the standard bounding box representation fails in fisheye cameras due to heavy radial distortion, particularly in the periphery. We explored different output representations for fisheye images, including oriented bounding boxes, curved boxes, ellipses, and polygons. We have integrated this model in our MTL framework, where we use a 24-sided polygon representation for object detection. We briefly summarize the details here and refer to the previous chapter~\ref{Chapter6} for more details on generalized object detection for fisheye camera images. We adapted the YOLOv3~\cite{YOLOV3} decoder to output polygons as shown in Figure~\ref{fig:omnidet-fisheye-yolo} and other representations listed above for a uniform comparison.\par
\subsection{Segmentation Tasks}
\label{sec:segmentation-tasks}

Three of the tasks are modeled as segmentation problems. Semantic and soiling segmentation having seven and four output classes, respectively, on the WoodScape dataset. In the following subsections, we briefly look into soiling and motion segmentation tasks.\par
\begin{figure*}[!t]
  \centering
  \newcommand{\turnheightnew}{0.25\columnwidth}

\begin{adjustbox}{max size={\textwidth}{\textheight}}
\begin{tabular}{@{\hskip 0.4mm}c@{\hskip 0.4mm}c@{\hskip 0.4mm}c@{\hskip 0.4mm}c@{\hskip 0.4mm}}

\includegraphics[height=\turnheightnew]{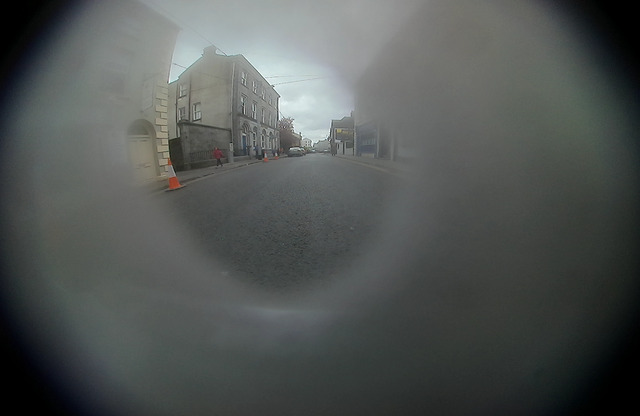} &
\includegraphics[height=\turnheightnew]{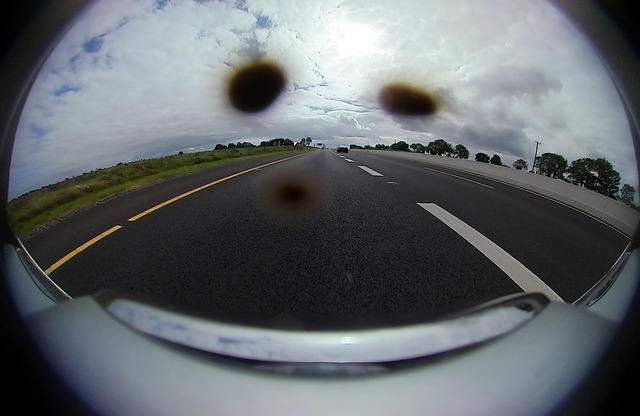} &
\includegraphics[height=\turnheightnew]{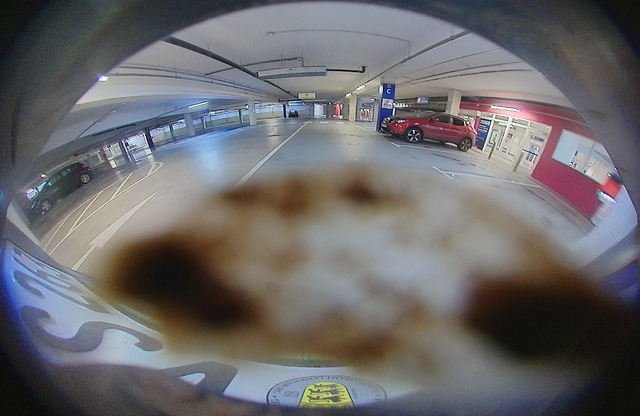} &
\includegraphics[height=\turnheightnew]{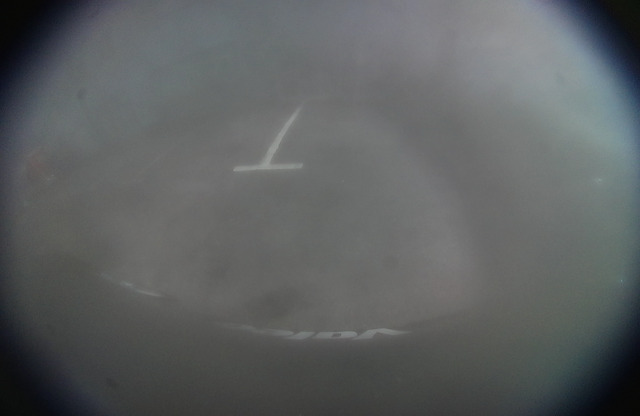} \\

\includegraphics[height=\turnheightnew]{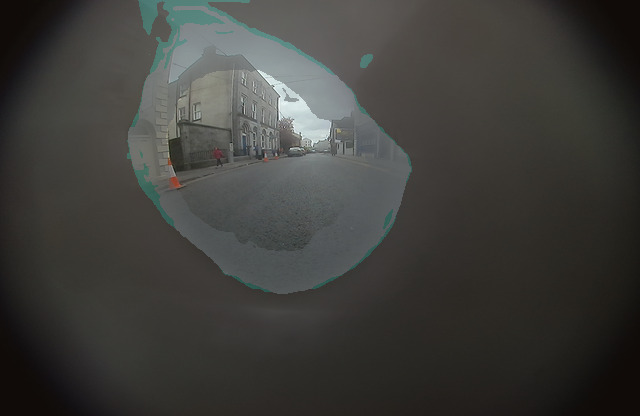} &
\includegraphics[height=\turnheightnew]{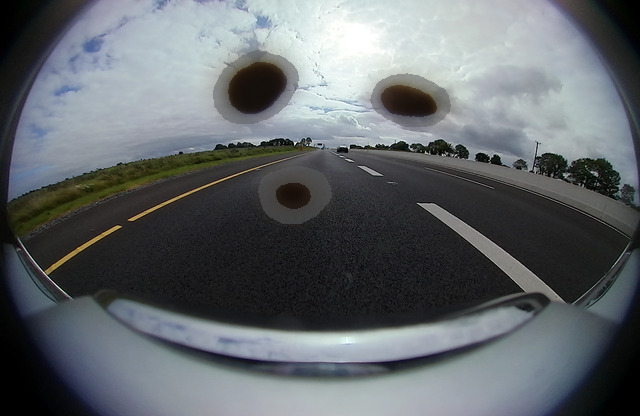} &
\includegraphics[height=\turnheightnew]{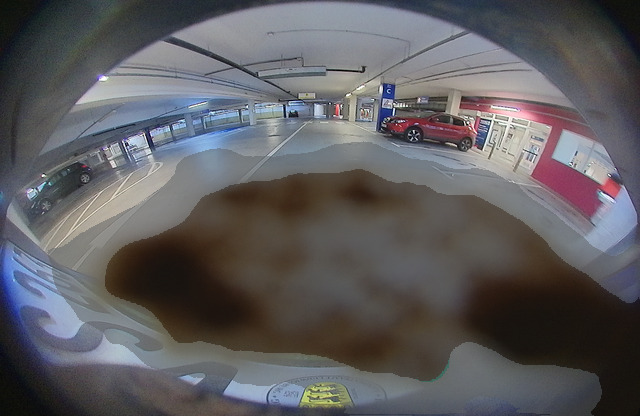} &
\includegraphics[height=\turnheightnew]{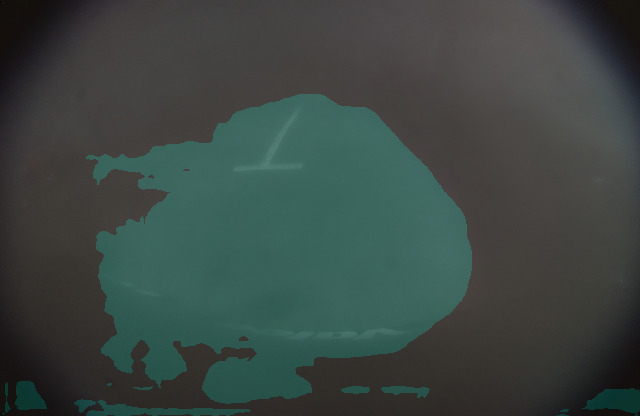} \\

\includegraphics[height=\turnheightnew]{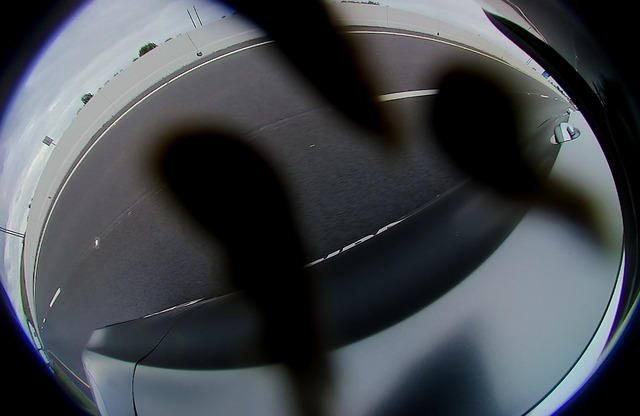} &
\includegraphics[height=\turnheightnew]{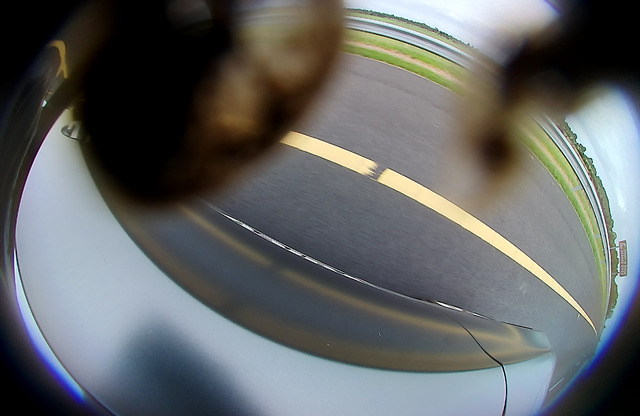} &
\includegraphics[height=\turnheightnew]{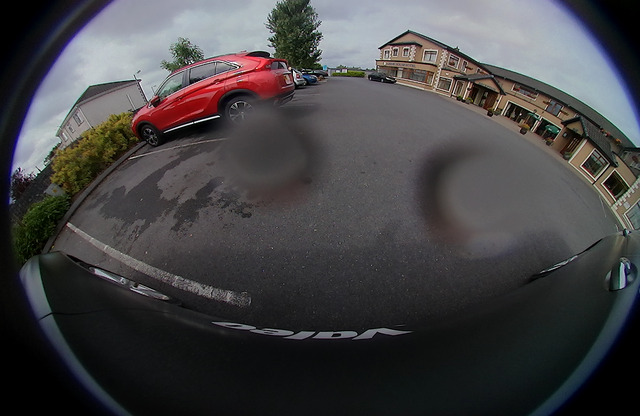} &
\includegraphics[height=\turnheightnew]{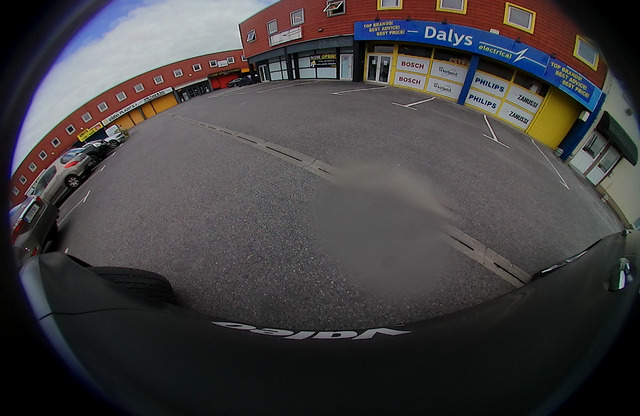} \\

\includegraphics[height=\turnheightnew]{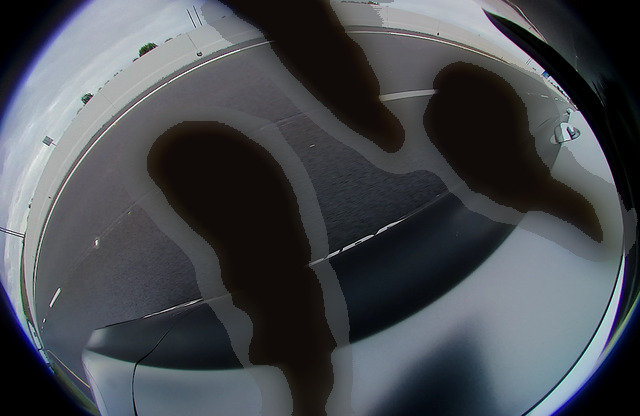} &
\includegraphics[height=\turnheightnew]{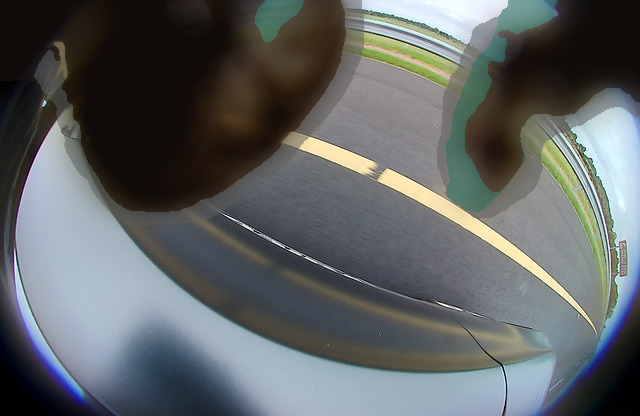} &
\includegraphics[height=\turnheightnew]{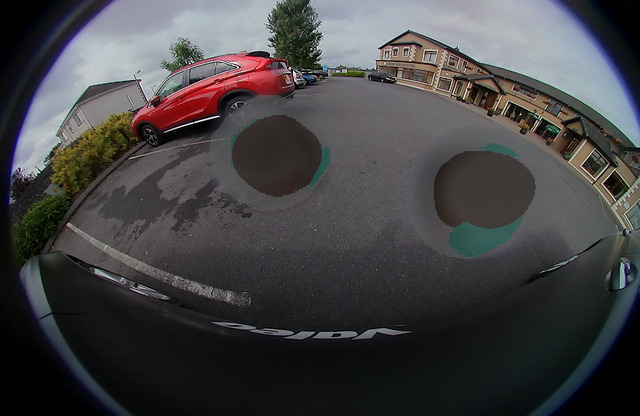} &
\includegraphics[height=\turnheightnew]{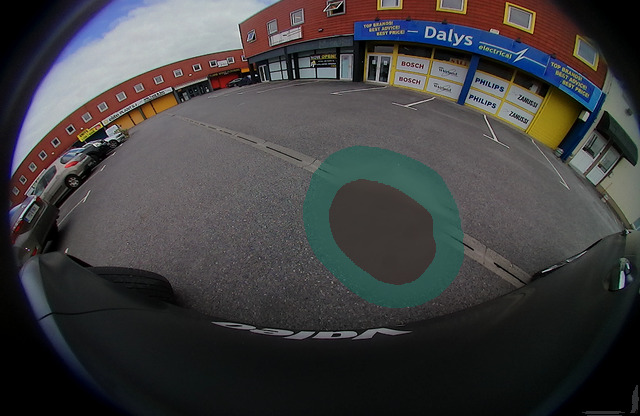} \\

\end{tabular}
\end{adjustbox}
  \caption[\bf Qualitative results of soiling segmentation on the WoodScape dataset.]
  {\bf Qualitative results of soiling segmentation on the WoodScape dataset. The 1\textsuperscript{st} and 3\textsuperscript{rd} rows indicate input images from front, rear, left and right cameras. The 2\textsuperscript{nd} and 4\textsuperscript{th} rows indicate the corresponding estimates.}
  \label{fig:omnidet-soiling-raw}
\end{figure*}
\subsubsection{Soiling Segmentation}

As discussed in Section~\ref{sec:soiling-related-work}, in this thesis, we focus on soiling caused by various unwanted particles reposing on the camera lens. These particles' source is mostly mud, dirt, water, or foam created by a detergent. Based on the state of aggregation, such soiling can be either static (\eg, highly viscous mud tends to dry up very quickly, so it does not change its position on the output image over time) or dynamic (mostly water and foam). We want to emphasize that the soiling detection task is necessary for an autonomous driving system as it is used to trigger a camera cleaning system that restores the visibility through the lens~\cite{uvrivcavr2019soilingnet}. It complements building segmentation or object detection models robust to soiling without an explicit soiling detection step. Coming to the loss function employed, soiling segmentation falls under the supervised learning category and is trained using the \textit{Lovasz-Softmax}~\cite{berman2018lovasz} loss. The qualitative results of the soiling segmentation are illustrated in Figure~\ref{fig:omnidet-soiling-raw}.\par
\subsubsection{Motion Segmentation}

\begin{figure}[htbp]
	\vspace{-1.8em}
	\centering
	\subfigure[Front Cam (t-6)]
	{\includegraphics[width=0.45\linewidth]
	{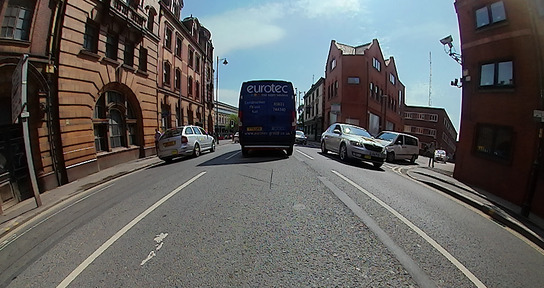}}\;\;
	\subfigure[Rear Cam (t-6)]
	{\includegraphics[width=0.45\linewidth]{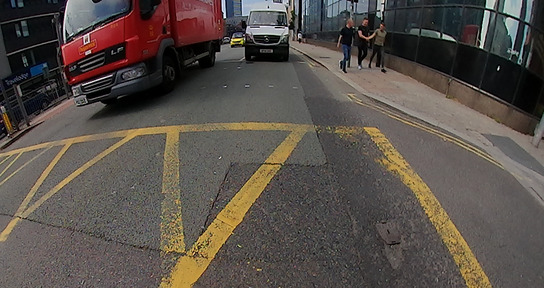}}\\
	\vspace{-3.5mm}
	\subfigure[Front Cam (t)]
	{\includegraphics[width=0.45\linewidth]{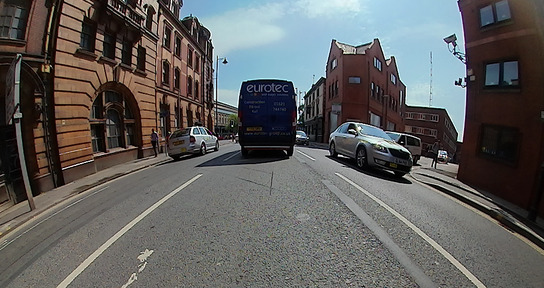}}\;\;
	\subfigure[Rear Cam (t)]
	{\includegraphics[width=0.45\linewidth]{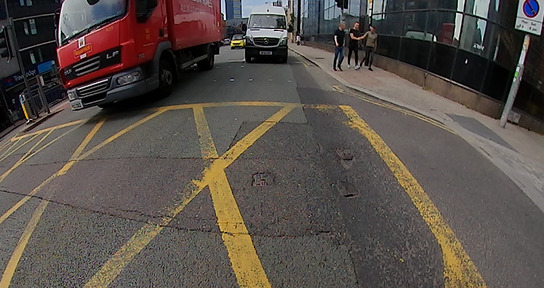}}\\
	\vspace{-3.6mm}
	\subfigure[Motion Estimate]
	{\includegraphics[width=0.45\linewidth]{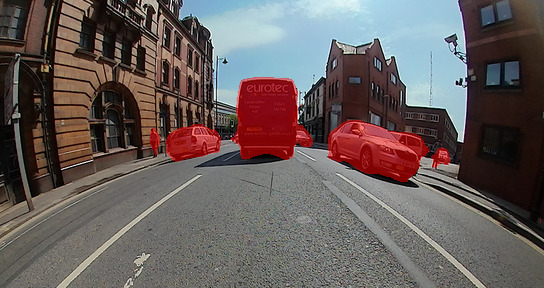}}\;\;
	\subfigure[Motion Estimate]
	{\includegraphics[width=0.45\linewidth]{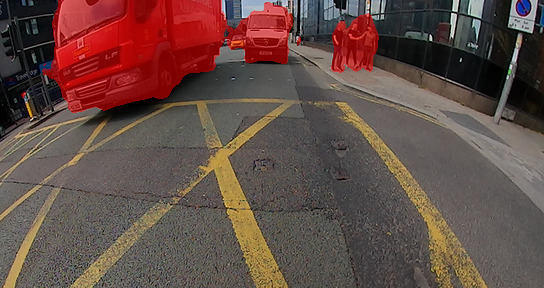}}\\
	\vspace{-3.7mm}
	\subfigure[Left Cam (t-6)]
	{\includegraphics[width=0.45\linewidth]{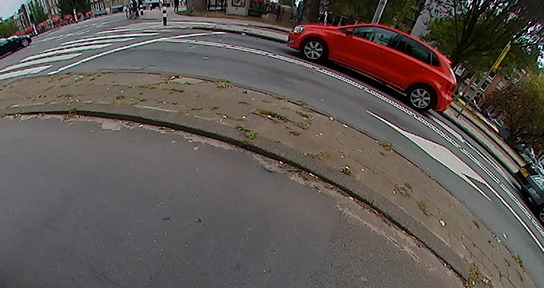}}\;\;
	\subfigure[Right Cam (t-6)]
	{\includegraphics[width=0.45\linewidth]{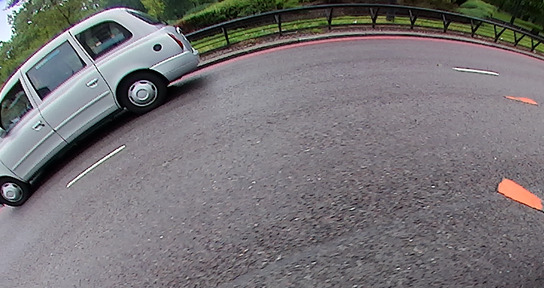}}\\
	\vspace{-3.7mm}
	\subfigure[Left Cam (t)]
	{\includegraphics[width=0.45\linewidth]{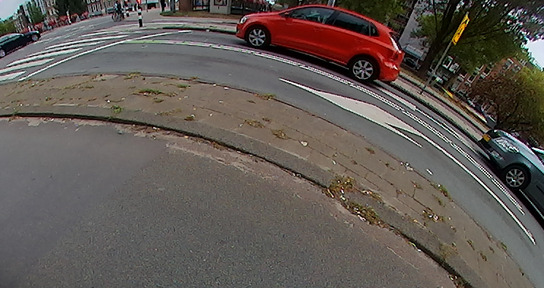}}\;\;
	\subfigure[Right Cam (t)]
	{\includegraphics[width=0.45\linewidth]{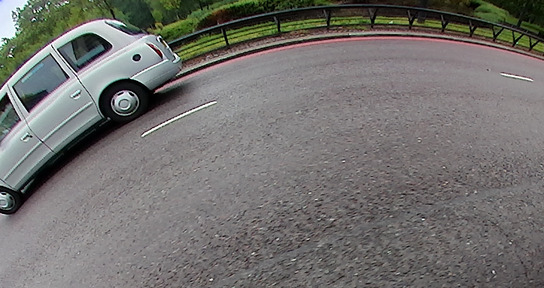}}\\
	\vspace{-3.7mm}
	\subfigure[Motion Estimate]
	{\includegraphics[width=0.45\linewidth]{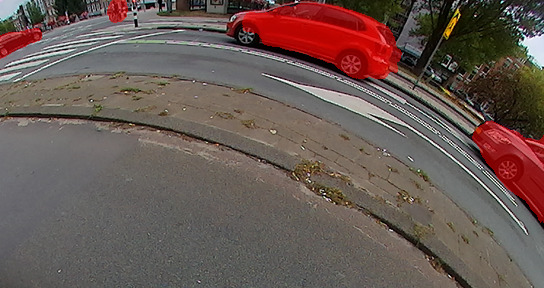}}\;\;
	\subfigure[Motion Estimate]
	{\includegraphics[width=0.45\linewidth]{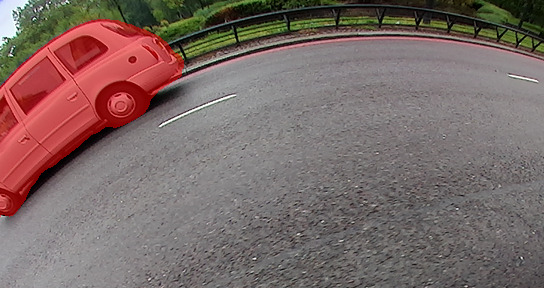}}\\
	\vspace{-3.7mm}
	\caption{\bf Qualitative results of motion segmentation on WoodScape. (t-6) and (t) frames are showcased to visually spot dynamic objects segmented in the motion estimate.}
	\label{fig:motion-raw}
\end{figure}
As discussed in Section~\ref{sec:motion-related-work}, in this thesis, we propose a CNN architecture for moving object detection using fisheye images that were captured in an autonomous driving environment. Motion segmentation uses two frames and outputs a binary moving or static mask. During training, the network predicts the posterior probability $Y_t$, which is optimized in a supervised fashion by \textit{Lovasz-Softmax}~\cite{berman2018lovasz} loss, and \textit{Focal}~\cite{lin2017focal} loss for handling class imbalance instead of the cross-entropy loss. We obtain the final segmentation mask $M_{t}$ by applying a pixel-wise $\operatorname{argmax}$ operation on the posterior probabilities. The qualitative results of the motion segmentation are illustrated in Figure~\ref{fig:motion-raw}.\par
\section{Network Details of the OmniDet MTL Framework} 
\label{sec:network-details-omnidet}

\begin{figure*}[!t]
  \centering
    \includegraphics[width=\textwidth]{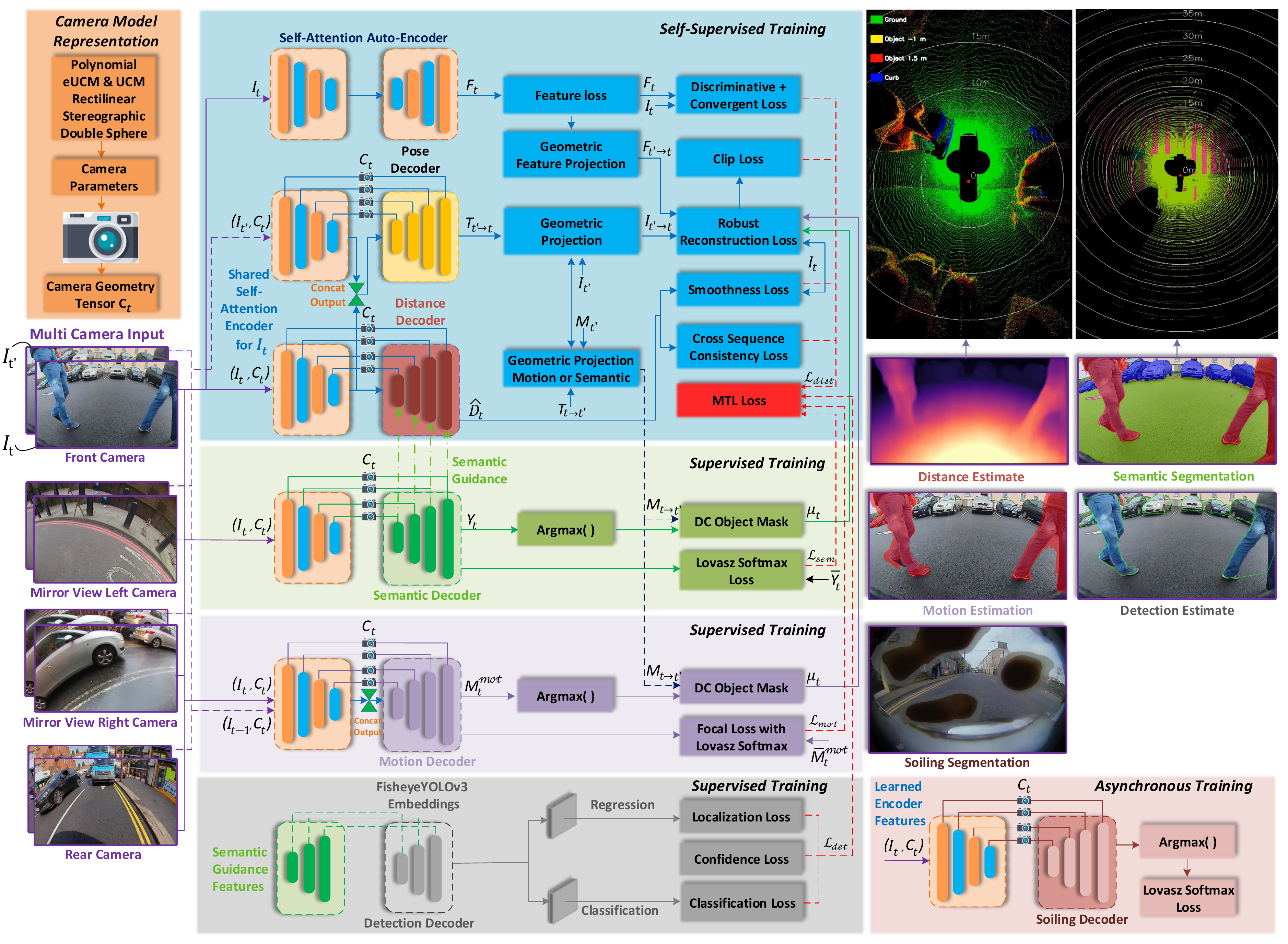}
    \caption[\bf Overview of the OmniDet framework.]
            {\textbf{Overview of OmniDet: A surround-view cameras based multi-task visual perception framework.} The distance estimation task (\textcolor[HTML]{00b0f0}{blue} block) makes use of semantic guidance and dynamic object masking from semantic/motion estimation (\textcolor[HTML]{00b050}{green} and \textcolor[HTML]{ab9ac0}{blue haze} block) and camera-geometry adaptive convolutions (\textcolor{orange}{orange} block). Additionally, we guide the detection decoder features (\textcolor{gray}{gray} block) with the semantic features. The encoder block (shown in the same color) is common for all the tasks. The framework consists of processing blocks to train the self-supervised distance estimation (\textcolor[HTML]{00b0f0}{blue} blocks) and semantic segmentation (\textcolor[HTML]{00b050}{green} blocks), motion segmentation (\textcolor[HTML]{ab9ac0}{blue haze} blocks), polygon-based fisheye object detection (\textcolor{gray}{gray} blocks), and the asynchronous task of soiling segmentation (\textcolor[HTML]{d9958f}{rose fog} block). We obtain top view geometric information by post-processing the predicted distance and semantic maps in 3D space. The camera geometry tensor $C_t$ (\textcolor{orange}{orange} block) helps OmniDet to yield distance maps on multiple camera-viewpoints and makes the network camera independent. $C_t$ can also be adapted to the standard camera models, as explained in Section~\ref{sec:camera-geometry-tensor}.}
    \label{fig:mtl_pipeline}
\end{figure*}
Encoder-decoder architectures are commonly used for dense prediction tasks. We use this type of architecture as it easily extends to a shared encoder for multiple tasks. Figure~\ref{fig:mtl_pipeline} provides an overview of the surround-view cameras based multi-task visual perception framework. We design the encoder by incorporating vector attention-based pairwise and patchwise self-attention encoders from~\cite{zhao2020exploring} as described in Section~\ref{sec:vector-self-attention}. These networks efficiently adapt the weights across both spatial dimensions and channels. We adapt the Siamese (twin network) approach for the motion prediction network, where we concatenate the source and target frame features and pass them to the super-resolution motion decoder. As the weights are shared in the Siamese encoder, the previous frame's encoder can be saved and re-used instead of re-computing the features. Inspired by~\cite{shu2020featdepth}, we develop an auxiliary self-attention auto-encoder for single-view reconstruction. We employ the novel CGT described in Section~\ref{sec:camera-geometry-tensor} to handle multiple viewpoints and changes in the camera's intrinsic distance estimation. Secondly, we employ synergized decoders via cross-task connections to improve each other's performance.\par

\begin{figure*}[!t]
  \centering
    \includegraphics[width=\textwidth]{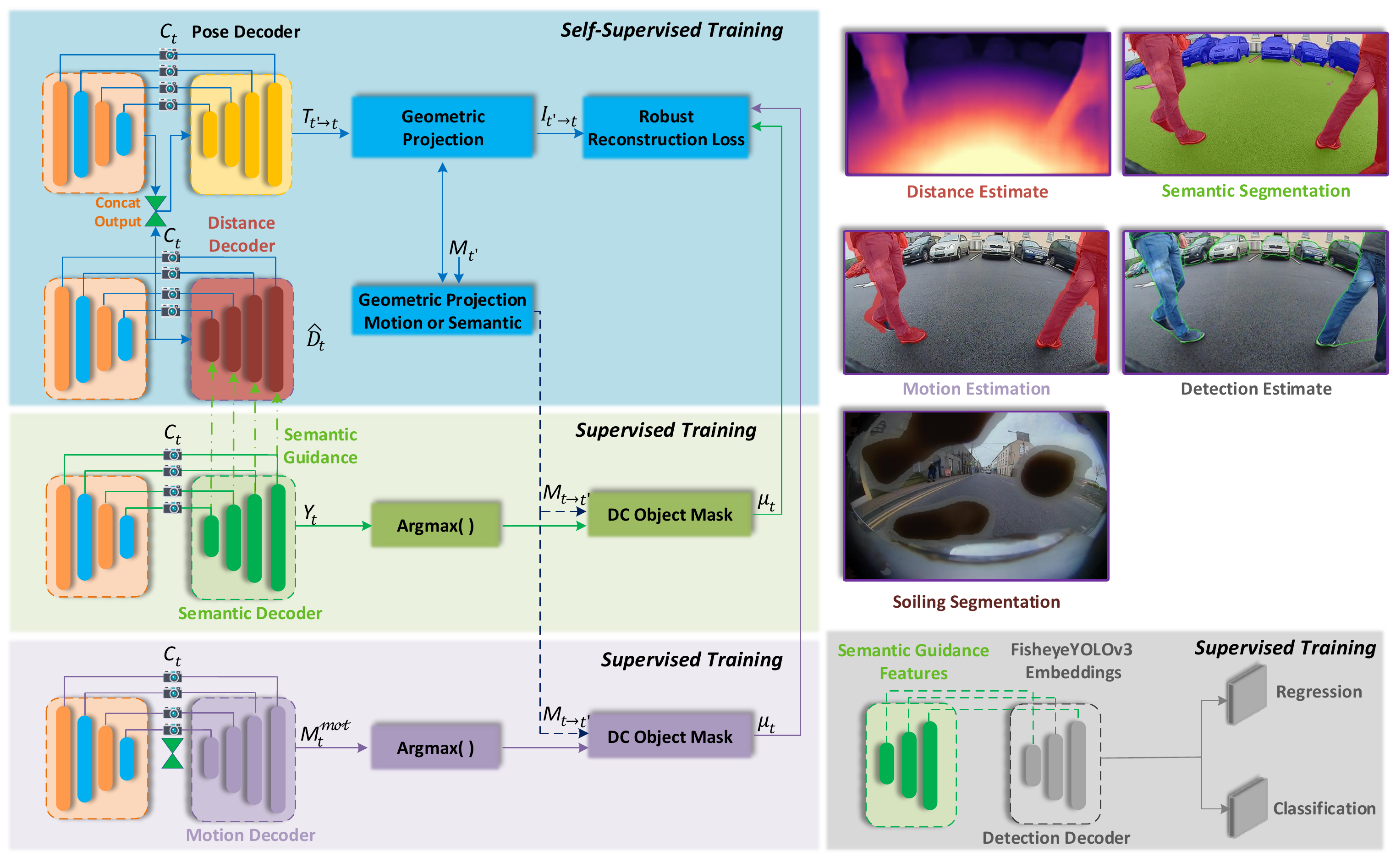}
    \caption{\bf Overview of the synergies established in OmniDet framework.}
    \label{fig:synergy-omnidet}
\end{figure*}

Most previous works employed hard parameter sharing techniques, \ie, a shared encoder that branches out into task-specific heads without any synergy. In addition, our goal is to induce synergies across the tasks in the form of loose coupling, maintaining the tasks to be independent. We achieve this by passing some decoder features from one task to another. The decoder features from the guiding task are combined with the intended task's decoder features using pixel-adaptive convolutions that contain an adaptive kernel to mix both feature types. For example, semantic segmentation produces holes in the road surface due to irregular textures, and depth maps corresponding to the road surface may help regularize the segmentation. There are multiple instances of synergy where semantic segmentation is guiding depth estimation, and object detection features and motion is guiding depth during the training phase. Figure~\ref{fig:synergy-omnidet} illustrates the synergies established in our OmniDet framework.\par
\subsubsection{Dealing With Dynamic Objects and Solving Infinite Depth Issue}
\label{sec:dynamic-object-mask-1}

\begin{figure}[!t]
  \centering
  \begin{minipage}[t]{0.495\textwidth}
    \centering
    \includegraphics[width=\textwidth]{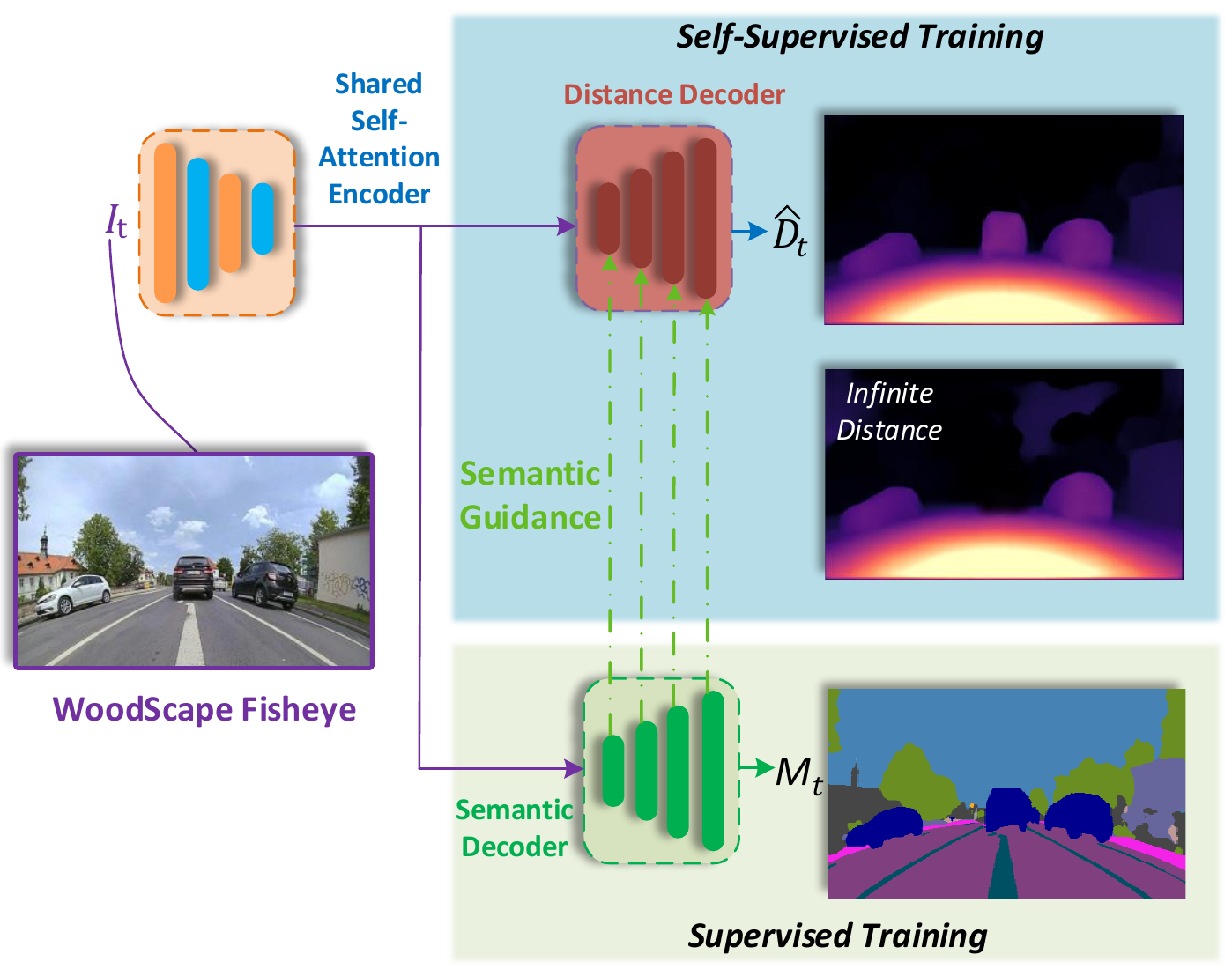}
    \caption{\bf Filtering of dynamic objects using semantic segmentation.}
    \label{fig:dynamic-semantic-seg}
  \end{minipage}%
  \hfill
  \begin{minipage}[t]{0.495\textwidth}
    \centering
    \includegraphics[width=\textwidth]{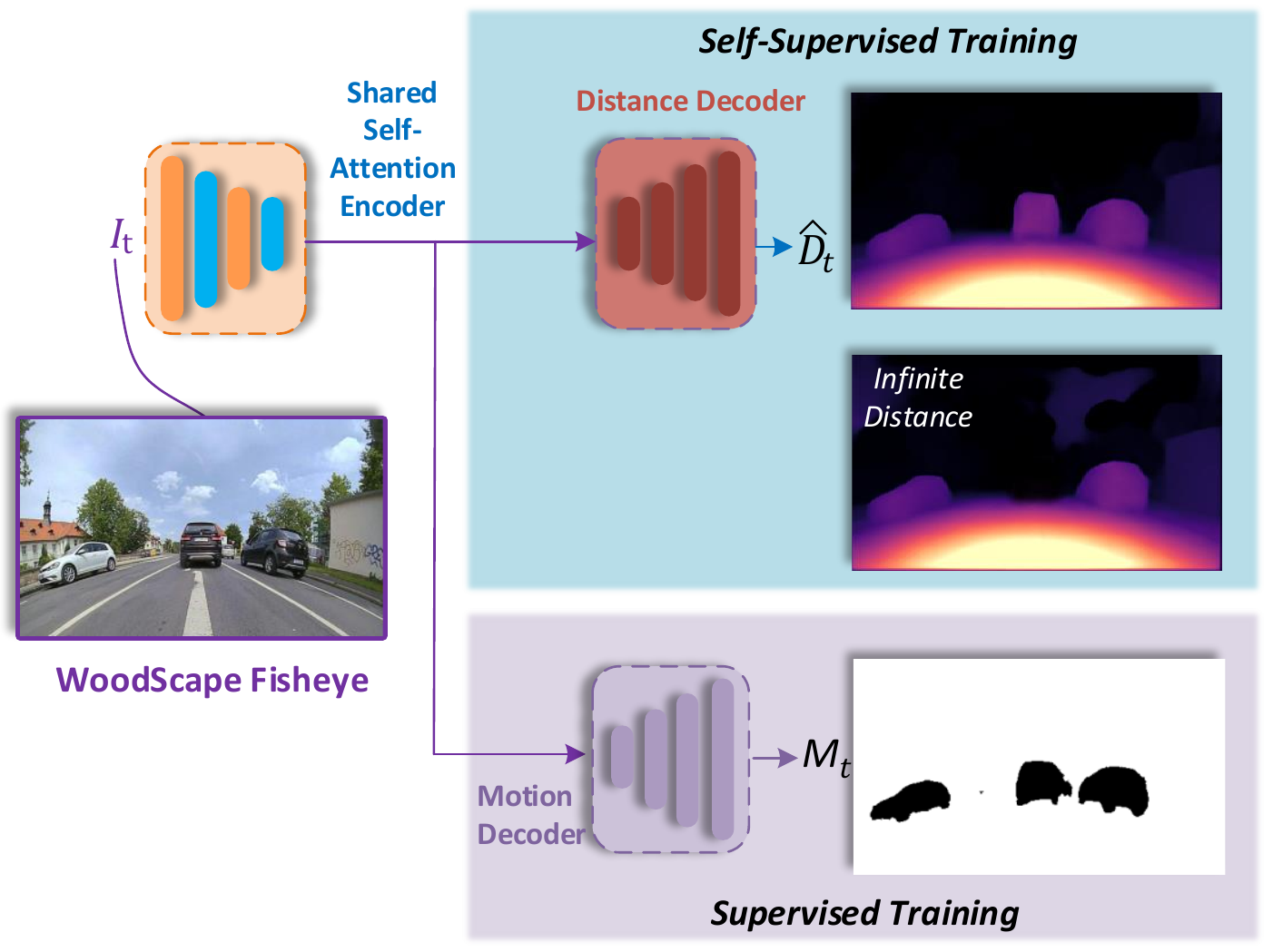}
    \caption{\bf Filtering of dynamic objects using motion segmentation.}
    \label{fig:dynamic-motion-seg}
  \end{minipage}
\end{figure}

This section discusses how we solved one of the concrete challenges using synergy, namely the dynamic object issue for depth estimation, which contaminates the photometric loss and causes infinite depth during inference. As dynamic objects violate the static world assumption, information about their depth/distance is essential in autonomous driving; else, we would encounter the infinite depth issue during the inference stage and not accurately reconstruct the scene and possibly even oversee other traffic participants. Compared to the previous Section~\ref{sec:dynamic-object-mask}, wherein we use the semantic segmentation task to obtain masks and filter dynamic objects as shown in Figure~\ref{fig:dynamic-semantic-seg}. The major drawback of using semantics is that it might not cover all the dynamic object classes in the semantic ground truth (\eg, cows). Therefore, we propose a robust alternative to enable the filtering of dynamic objects using the motion segmentation task as shown in Figure~\ref{fig:dynamic-motion-seg}, which yields either a static or a dynamic mask. However, the user can leverage either one of the tasks to solve this issue. We use the motion segmentation information to exclude potentially \textit{moving} dynamic objects, while the distance is learned from \textit{non-moving} dynamic objects. For this purpose, we define the pixel-wise mask $\mu_t$, which contains a $1$ if a pixel does not belong to a dynamic object from the current frame $I_t$ and also not to a wrongfully projected dynamic object from the reconstructed frames $\hat{I}_{t'\to t}$ and a $0$ otherwise. Accordingly, we predict a motion segmentation mask $M^{mot}_t$ belonging to the target frame $I_t$, as well as motion masks $M_{t'}$ for the source frames $I_{t'}$. Dynamic objects inside the source frame are canonically detected inside $M_t$. However, to obtain the wrongfully projected dynamic objects, we need to warp the motion masks by nearest-neighbor sampling to the target frame, yielding projected motion masks $M_{t' \to t}$. Dynamic objects can be masked through a pixel-wise multiplication of the mask with the reconstruction loss for images and features.\par
\subsubsection{Linking Self-Attention and Semantic features to Distance and Detection decoders}

To better incorporate the semantic knowledge extracted from the multi-task networks segmentation branch into the distance estimation, we incorporate it using pixel adaptive convolutions (PAC) described in Section~\ref{sec:Semantically-Guided-Distance-Decoder} to distill the knowledge from the semantic features into the distance decoder. This, in particular, breaks up the spatial invariance of the convolutions and allows the incorporation of location-specific semantic knowledge into the multi-level distance features. As shown in Figure~\ref{fig:mtl_pipeline} (\textcolor[HTML]{00b050}{green} block), the features are extracted at different levels of the segmentation decoder.
To leverage the multi-task learning setup, at first, we extract the SAN encoder features and feed it as an input signal to the Eq.~\ref{eq:pac} and bypass the spatial information from the SAN encoder to the semantic decoder and fuse these features (skip-connections). Finally, we fuse these features and the detection decoder embeddings by applying PAC and obtaining content-agnostic features. This novel fusion technique in the OmniDet framework significantly improves the detection decoder's accuracy, which can be seen in Table~\ref{table:features}.\par
\subsection{Joint Optimization}

We provided a brief overview on MTL optimization strategies in Section~\ref{sec:optimization-in-mtl}. Balancing the task losses is of significant importance in training a multi-task model. We evaluate various task weighting strategies for five tasks compared to the two task experiments in chapter~\ref{Chapter3} and chapter~\ref{Chapter4}. We evaluate the uncertainty loss from Kendall~\cite{Kendall2018}, the gradient magnitude normalization GradNorm~\cite{chen2018gradnorm}, the dynamic task prioritization DTP~\cite{guo2018dynamic}, the dynamic weight average DWA~\cite{liu2019end} and the geometric loss~\cite{chennupati2019multinet++}.\par

Secondly, we propose a novel method called \textbf{VarNorm} for variance normalization. It consists of normalizing each loss by its variance over the last $n$ epochs. The loss weight of task $i$ at epoch $t$ is formulated as below:
\begin{equation}
    w_i(t) = \frac{1}{\sigma_i(t-1)},
    \sigma_i(t) = \frac{1}{n-1} \sum_{k=0}^{n-1} (L_i(t-k) - \overline{L_i})^2
    \label{eq:varnorm}
\end{equation}
where $\overline{L_i}$ is the average of task loss $i$ over the last $n$ epochs. We chose $n=5$. This method is motivated by the simple idea that the task loss values can be seen as a distribution whose dispersion is its variance. Variance normalization re-scales the dispersion between the different task loss distributions based on the previous $n$ epochs. A large dispersion leads to a lower task weight, whereas a small dispersion to a higher one. Its final effect tends to homogenize the learning speed of tasks. As shown in Table~\ref{table:task-weighting}, equal weighting is the worst, and the multi-task network performs better than the single task networks by using any dynamic task weighting method presented above. We employ the proposed VarNorm method for all the further experiments as it achieved the best results.\par
\begin{table}[t]
\centering{
\begin{adjustbox}{width=0.8\columnwidth}
\small
\setlength{\tabcolsep}{0.15em}
\begin{tabular}{lccccccc}
\toprule
  \multicolumn{1}{c|}{\textit{
  \begin{tabular}[c]{@{}c@{}} Task \\ Weighting\end{tabular}}} &
  \multicolumn{2}{c|}{\cellcolor[HTML]{00b0f0} \textit{
  \begin{tabular}[c]{@{}c@{}} Distance \\ Estimation\end{tabular}}} &
  \multicolumn{2}{c|}{\cellcolor[HTML]{00b050} \textit{
  \begin{tabular}[c]{@{}c@{}} Semantic \\ Segmentation\end{tabular}}} &
  \multicolumn{2}{c|}{\textit{\cellcolor[HTML] {ab9ac0}
  \begin{tabular}[c]{@{}c@{}} Motion\\ Segmentation\end{tabular}}} &
  \multicolumn{1}{c|}{\cellcolor[HTML]{a5a5a5} \textit{
  \begin{tabular}[c]{@{}c@{}}Object \\ Detection\end{tabular}}} \\ 
\midrule
  & \multicolumn{1}{|c|}{\textbf{Sq. Rel} $\mathbf{\downarrow}$}
  & \multicolumn{1}{c|}{\textbf{Abs Rel} $\mathbf{\downarrow}$}
  & \multicolumn{1}{c|}{\textbf{mIoU} $\mathbf{\uparrow}$}
  & \multicolumn{1}{c|}{\textbf{PA} $\mathbf{\uparrow}$} 
  & \multicolumn{1}{c|}{\textbf{mIoU} $\mathbf{\uparrow}$}
  & \multicolumn{1}{c|}{\textbf{PA} $\mathbf{\uparrow}$}
  & \multicolumn{1}{c|}{\textbf{mAP} $\mathbf{\uparrow}$} \\
\midrule
Single Task                               & 0.060  & 0.304 & 72.5 & 94.8 & 68.1 & 94.1 & 63.5 \\
\midrule
Equal                                     & 0.058  & 0.302 & 70.3 & 92.7 & 67.3 & 93.3 & 64.6 \\
DTP~\cite{guo2018dynamic}                 & 0.047  & 0.281 & 75.8 & 95.6 & 75.3 & 95.3 & 67.9 \\
DWA~\cite{liu2019end}                     & 0.054  & 0.293 & 75.4 & 95.2 & 74.7 & 95.1 & 67.5 \\
Geometric~\cite{chennupati2019multinet++} & 0.061  & 0.297 & 74.2 & 94.1 & 73.2 & 94.3 & 66.7 \\
GradNorm~\cite{chen2018gradnorm}          & 0.050  & 0.283 & 75.9 & 95.7 & 74.9 & 96.0 & 67.7 \\
Uncertainity~\cite{Kendall2018}           & 0.044 & 0.279  & 76.1 & 96.2 & 75.1 & 95.8 & 68.0 \\
\textbf{VarNorm}                          & \textbf{0.046} & \textbf{0.276}  & \textbf{76.6} & \textbf{96.4} & \textbf{75.3} & \textbf{96.1} & \textbf{68.4} \\ 
\bottomrule
\end{tabular}
\end{adjustbox}
}
\caption[\bf Comparison of task-weighting methods on the WoodScape dataset.]
{\bf {Comparison of task-weighting methods on the WoodScape dataset.} PA denotes pixel accuracy.}
\label{table:task-weighting}
\end{table}
\section{Implementation Details}

We systematically train and test all single and multi-task models on the Woodscape and the pinhole camera datasets KITTI and Cityscapes described in the datasets Section~\ref{sec:benchmarks}. We use Pytorch~\cite{paszke2017automatic} and employ a single-stage learning process for the OmniDet framework to facilitate network optimization. We incorporate the recently proposed SAN in the encoder. Zhao~\etal~\cite{zhao2020exploring} proposed two convolution variants, namely \emph{pairwise} and \emph{patchwise}. We mainly use patchwise but perform an ablation study using pairwise self-attention convolutions. We employ the Ranger (RAdam~\cite{liu2019variance} + LookAhead~\cite{zhang2019lookahead}) optimizer to minimize the training objective function. We train the model for 20 epochs, with a batch size of 24 on a 24GB Titan RTX with an initial learning rate of ${4 \times {10}^{-4}}$ for the first 15 epochs, which is reduced to ${{10}^{-5}}$ for the last five epochs. The sigmoid output $\sigma$ from the distance decoder is converted to distance with $D = {m \cdot \sigma + n}$, where $m$ and $n$ are chosen to constrain $D$ between $0.1$ and $100$ units. Finally, we set $\beta$, $\gamma$, $\omega$ and $\mu$ to ${{10}^{-3}}$. Images are resized to $544 \times 288\,\mathrm{px}$ from the native $1\mathrm{MP}$ resolution for WoodScape. In the case of Cityscapes, images are resized to $640 \times 384\,\mathrm{px}$ for training and validation, and for KITTI, we resize it to input size of $640 \times 192\,\mathrm{px}$ for all the tasks. For the motion segmentation, we use the annotations provided by DeepMotion~\cite{vertens2017smsnet} for Cityscapes and KITTI MoSeg~\cite{siam2018modnet}, where labels are available only for the cars category.\par

All images from the surround-view cameras with multiple viewpoints are shuffled thoroughly and fed to the distance and pose networks along with their respective intrinsic to create the camera geometry tensor $C_t$, as shown in Figure~\ref{fig:mtl_pipeline}, and described in Section~\ref{sec:camera-geometry-tensor}. The soiling dataset is independently built, and thus it cannot be trained jointly in a traditional manner. Thus we freeze the shared encoder trained using five other tasks and train only the decoder for soiling. This demonstrates the addition of new tasks reusing the encoder features. We also trained soiling jointly using asynchronous backpropagation \cite{kokkinos2017ubernet}, but it achieved the same accuracy as using the frozen encoder. Compared to the previous work SoilingNet \cite{uvrivcavr2019soilingnet}, we moved from the tiled output to a pixel-level segmentation.\par
\section{Experiments}

\begin{table}[!t]
\centering
\begin{adjustbox}{width=0.85\columnwidth}
\small
\begin{tabular}{cccccccccccc}
\toprule
\multicolumn{1}{c}{\cellcolor[HTML]{00b0f0} \textit{
\begin{tabular}[c]{@{}c@{}} Dist. \& \\ Pose Est.\end{tabular}}}
& \multicolumn{1}{c}{\cellcolor[HTML]{00b050} \textit{
\begin{tabular}[c]{@{}c@{}} Sem.\\ Seg.\end{tabular}}} 
& \multicolumn{1}{c}{\cellcolor[HTML]{ab9ac0} \textit{
\begin{tabular}[c]{@{}c@{}} Mot.\\ Seg.\end{tabular}}} 
& \multicolumn{1}{c}{\cellcolor[HTML]{a5a5a5} \textit{
\begin{tabular}[c]{@{}c@{}} Obj.\\ Det.\end{tabular}}} 
& \multicolumn{1}{c}{\cellcolor[HTML]{e5b9b5} \textit{
\begin{tabular}[c]{@{}c@{}} Soil.\\ Seg.\end{tabular}}}
& \multicolumn{1}{c}{\cellcolor[HTML]{00b0f0} \textbf{\textit{RMSE}}} 
& \multicolumn{1}{c}{\cellcolor[HTML]{00b050} 
\begin{tabular}[c]{@{}c@{}} \textbf{\textit{mIoU}} \\ \textbf{\textit{Seg.}}\end{tabular}} & \multicolumn{1}{c}{\cellcolor[HTML]{ab9ac0} \begin{tabular}[c]{@{}c@{}}\textbf{\textit{mIoU}} \\ \textbf{\textit{Mot.}}\end{tabular}} & \multicolumn{1}{c}{\cellcolor[HTML]{a5a5a5}
\begin{tabular}[c]{@{}c@{}}\textbf{\textit{mAP}} \\ \textbf{\textit{Det.}}\end{tabular}}
& \multicolumn{1}{c}{\cellcolor[HTML]{e5b9b5}
\begin{tabular}[c]{@{}c@{}}\textbf{\textit{mIoU}} \\ \textbf{\textit{Soil.}}\end{tabular}} & \multicolumn{1}{c}{\textit{\cellcolor[HTML]{80CBC4}
\begin{tabular}[c]{@{}c@{}} Infer.\\ (fps)\end{tabular}}} \\
\midrule
\multicolumn{11}{c}{\cellcolor[HTML]{34FF34}\textit{WoodScape}} \\
\midrule
    \ch & \xm & \xm & \xm & \xm & 1.681 & \xmb & \xmb & \xmb & \xmb & 210 \\
    \xm & \ch & \xm & \xm & \xm & \xmb  & 72.5 & \xmb & \xmb & \xmb & 190 \\
    \xm & \xm & \ch & \xm & \xm & \xmb  & \xmb & 68.1 & \xmb & \xmb & 105 \\
    \xm & \xm & \xm & \ch & \xm & \xmb  & \xmb & \xmb & 63.5 & \xmb & 190 \\
    \xm & \xm & \xm & \xm & \ch & \xmb  & \xmb & \xmb & \xmb & 80.8 & 190 \\
    \ch & \ch & \xm & \xm & \xm & 1.442 & 74.8 & \xmb & \xmb & \xmb & 143 \\
    \xm & \ch & \xm & \ch & \xm & \xmb  & 77.1 & \xmb & 67.9 & \xmb & 143 \\
    \ch & \ch & \ch & \xm & \xm & 1.352 & 75.5 & 74.8 & \xmb & \xmb & 69  \\
    \ch & \ch & \ch & \ch & \xm & \textbf{1.332} & \textbf{76.6} & \textbf{75.3} & \textbf{68.4} & \xmb & \textbf{60}  \\
\midrule
\multicolumn{11}{c}{\cellcolor[HTML]{FD6864}\textit{KITTI}} \\
\midrule
    \ch & \xm & \xm & \xm & \nap & 4.126 & \xmb & \xmb & \xmb & \nap & 160 \\
    \xm & \ch & \xm & \xm & \nap & \xmb  & 67.7 & \xmb & \xmb & \nap & 148 \\
    \xm & \xm & \ch & \xm & \nap & \xmb  & \xmb & 68.3 & \xmb & \nap & 78 \\
    \xm & \xm & \xm & \ch & \nap & \xmb  & \xmb & \xmb & 80.1 & \nap & 182 \\
    \ch & \ch & \xm & \xm & \nap & 3.984 & 72.1 & \xmb & \xmb & \nap & 103 \\
    \ch & \ch & \ch & \xm & \nap & 3.892 & 71.9 & 71.7 & \xmb & \nap & 47 \\
    \ch & \ch & \ch & \ch & \nap & \textbf{3.859} & \textbf{72.4} & \textbf{72.2} & \textbf{82.3} & \nap & \textbf{43} \\
    \midrule
\multicolumn{11}{c}{\cellcolor[HTML]{448BE9}\textit{CityScapes}}  \\
\midrule
    \ch & \xm & \xm & \xm & \nap & 4.906 & \xmb & \xmb & \xmb & \nap & 156 \\
    \xm & \ch & \xm & \xm & \nap & \xmb  & 78.7 & \xmb & \xmb & \nap & 132 \\
    \xm & \xm & \ch & \xm & \nap & \xmb  & \xmb & 70.4 & \xmb & \nap & 64 \\
    \xm & \xm & \xm & \ch & \nap & \xmb  & \xmb & \xmb & 51.7 & \nap & 167 \\
    \ch & \ch & \xm & \xm & \nap & 4.741 & 79.4 & \xmb & \xmb & \nap & 91 \\
    \ch & \ch & \ch & \xm & \nap & 4.725 & 79.1 & 72.0 & \xmb & \nap & 36 \\
    \ch & \ch & \ch & \ch & \nap & \textbf{4.691} & \textbf{81.2} & \textbf{72.7} & \textbf{53.0} & \nap & \textbf{31} \\
    \bottomrule
\end{tabular}
\end{adjustbox}
\caption[\bf Comparative study of SAN10-Patch MTL model and the equivalent single task models on three datasets.]{\textbf{Comparative study of SAN10-Patch MTL model and the equivalent single-task models on three datasets.} The checkmark legends indicate \ch if the task is activated during training, \xm \, deactivated, \xmb \, no evaluation performed, and \nap \, task not being part of the MTL training.}
\label{table:mtl_ablation}
\end{table}
\subsection{Single-Task vs Multi-Task Learning} 
\label{sec:single_vs_mtl}

\begin{figure*}[!ht]
  \resizebox{\textwidth}{!}{\newcommand{\turnheightnew}{0.25\linewidth}
\centering

\begin{tabular}{@{\hskip 0.5mm}c@{\hskip 0.5mm}c@{\hskip 0.5mm}c@{\hskip 0.5mm}c@{\hskip 0.5mm}c@{}}

\includegraphics[height=\turnheightnew]{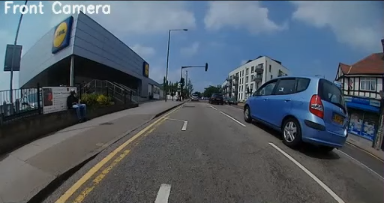} &
\includegraphics[height=\turnheightnew]{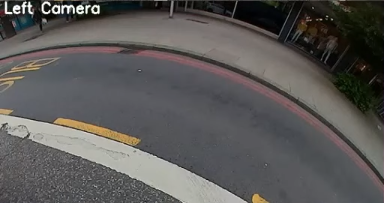} &
\includegraphics[height=\turnheightnew]{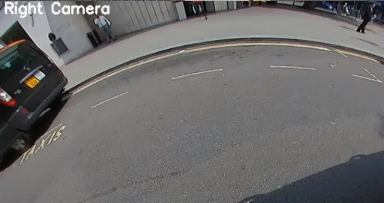} &
\includegraphics[height=\turnheightnew]{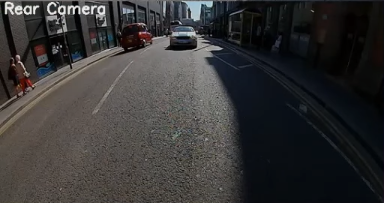} \\

\includegraphics[height=\turnheightnew]{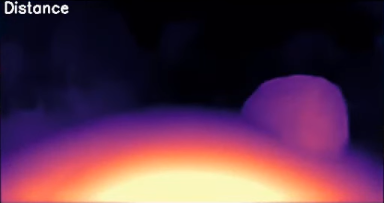} &
\includegraphics[height=\turnheightnew]{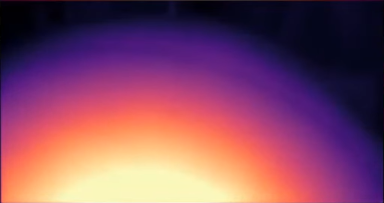} &
\includegraphics[height=\turnheightnew]{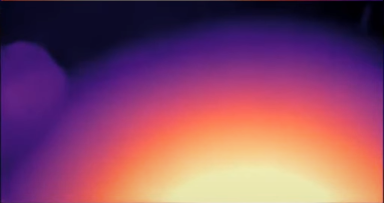} &
\includegraphics[height=\turnheightnew]{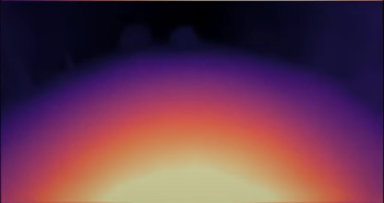} \\

\includegraphics[height=\turnheightnew]{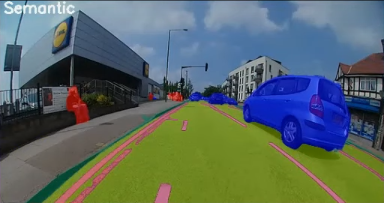} &
\includegraphics[height=\turnheightnew]{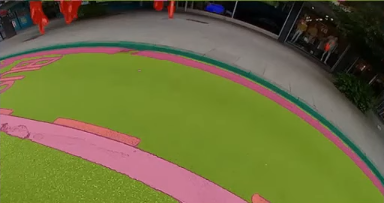} &
\includegraphics[height=\turnheightnew]{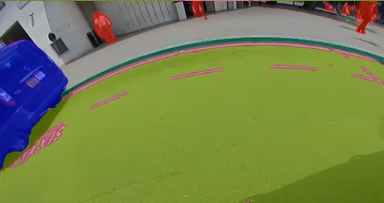} &
\includegraphics[height=\turnheightnew]{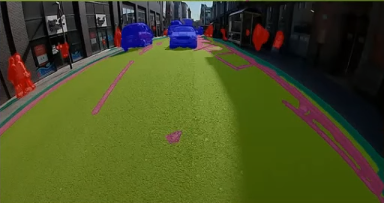} \\

\includegraphics[height=\turnheightnew]{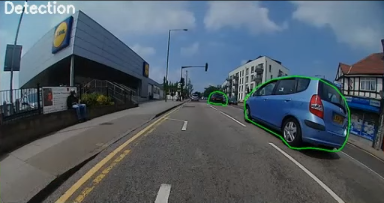} &
\includegraphics[height=\turnheightnew]{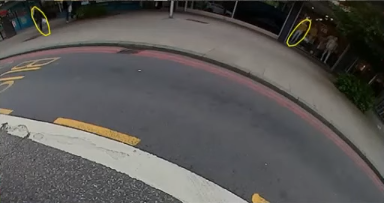} &
\includegraphics[height=\turnheightnew]{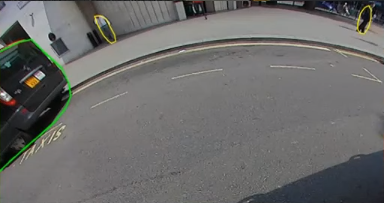} &
\includegraphics[height=\turnheightnew]{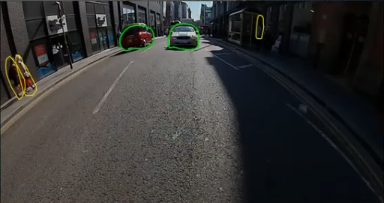} \\

\includegraphics[height=\turnheightnew]{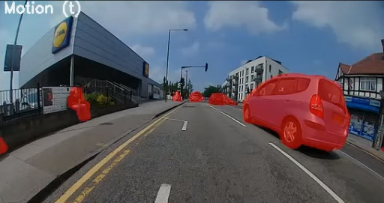} &
\includegraphics[height=\turnheightnew]{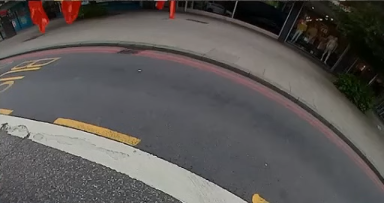} &
\includegraphics[height=\turnheightnew]{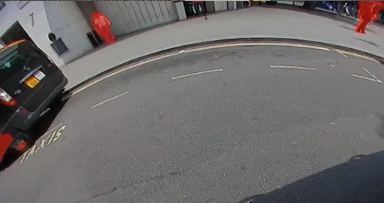} &
\includegraphics[height=\turnheightnew]{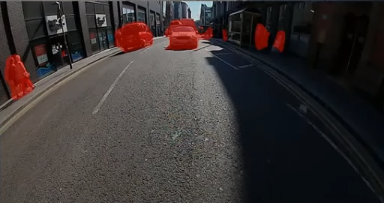} \\

\includegraphics[height=\turnheightnew]{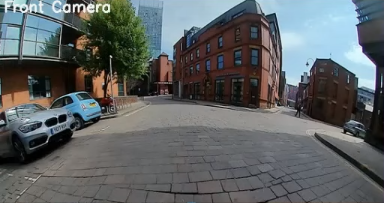} &
\includegraphics[height=\turnheightnew]{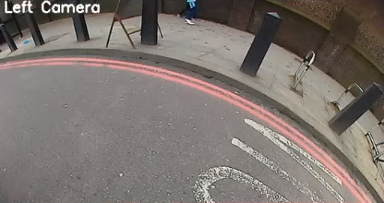} &
\includegraphics[height=\turnheightnew]{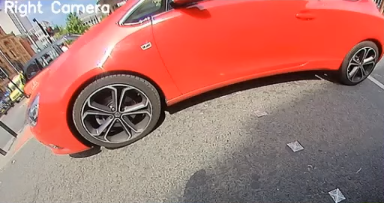} &
\includegraphics[height=\turnheightnew]{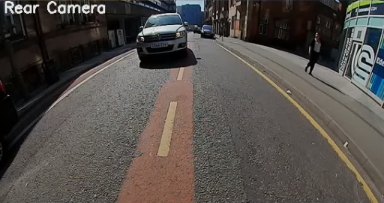} \\

\includegraphics[height=\turnheightnew]{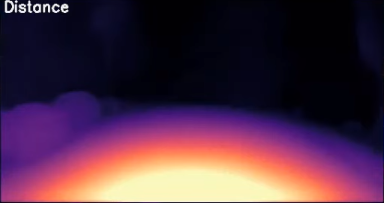} &
\includegraphics[height=\turnheightnew]{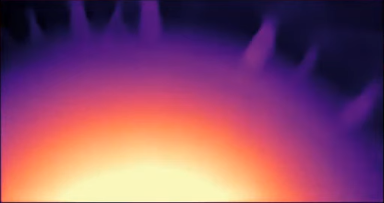} &
\includegraphics[height=\turnheightnew]{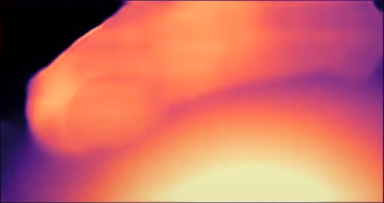} &
\includegraphics[height=\turnheightnew]{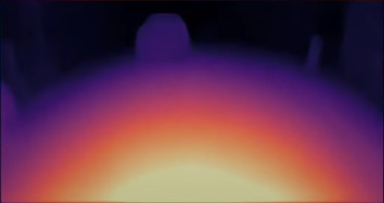} \\

\includegraphics[height=\turnheightnew]{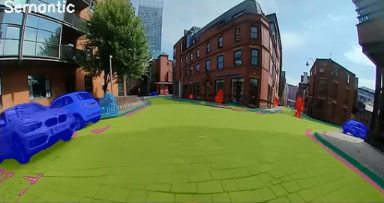} &
\includegraphics[height=\turnheightnew]{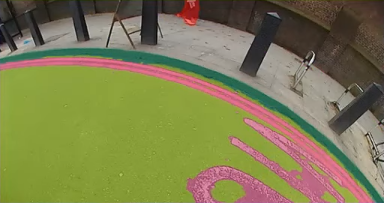} &
\includegraphics[height=\turnheightnew]{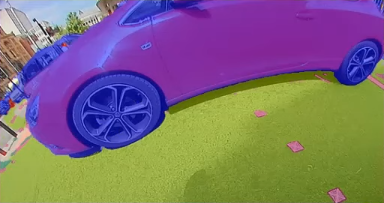} &
\includegraphics[height=\turnheightnew]{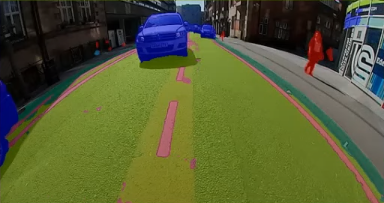} \\

\includegraphics[height=\turnheightnew]{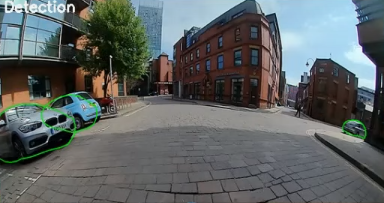} &
\includegraphics[height=\turnheightnew]{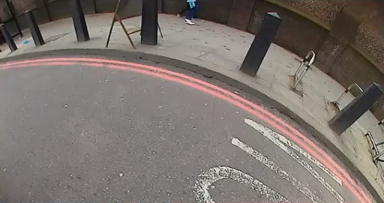} &
\includegraphics[height=\turnheightnew]{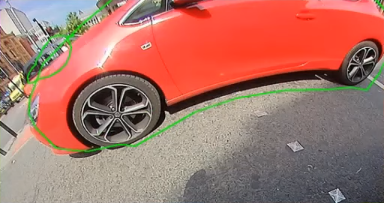} &
\includegraphics[height=\turnheightnew]{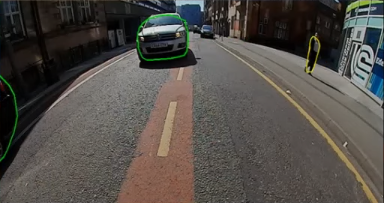} \\

\includegraphics[height=\turnheightnew]{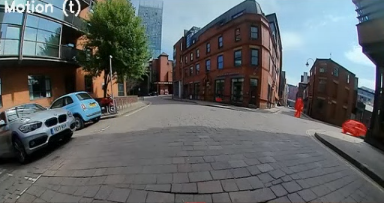} &
\includegraphics[height=\turnheightnew]{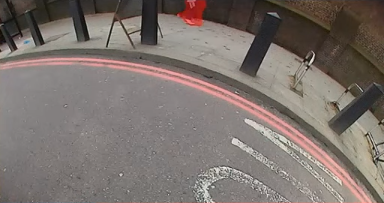} &
\includegraphics[height=\turnheightnew]{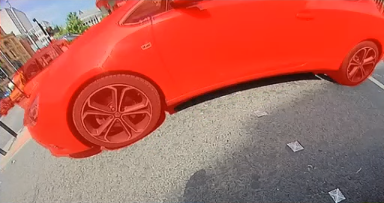} &
\includegraphics[height=\turnheightnew]{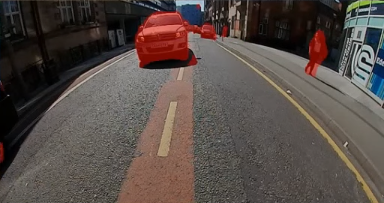} \\

\end{tabular}
}
  \caption[\bf Qualitative results of raw fisheye images from the OmniDet framework on the WoodScape dataset.]
          {{\bf Qualitative results of raw fisheye images from the OmniDet framework on the WoodScape dataset.} The 1\textsuperscript{st} and 6\textsuperscript{th} rows indicates the input images from Front, Left, Right and Rear cameras. 2\textsuperscript{nd} and 7\textsuperscript{th} rows indicate distance estimates, 3\textsuperscript{rd} and 8\textsuperscript{th} rows indicate semantic segmentation maps, 4\textsuperscript{th} and 9\textsuperscript{th} rows indicate generalized object detection predictions and finally 5\textsuperscript{th} and 10\textsuperscript{th} rows indicate the motion segmentation. For more qualitative results at a higher resolution, we refer to this video: \url{https://youtu.be/xbSjZ5OfPes}}
\label{fig:omnidet-qual_raw}
\end{figure*}

In Table~\ref{table:mtl_ablation}, we perform an extensive ablation of the proposed framework on all considered datasets. The soiling Segmentation task is indicated \nap \, (Not Applicable) as it is not included in the MTL training regime. Quantitative results from the experiments indicate that a multi-task network with six tasks, five diverse tasks performs better than the single task models along with the proposed synergies. The qualitative results on the raw fisheye streams from the surround-view camera system on the perception tasks are shown in Figure~\ref{fig:omnidet-qual_raw}. For KITTI and CityScapes, we employ the novel VarNorm task weighting technique. With this synergy of perception tasks, we obtain state-of-the-art depth and pose estimation results on the KITTI dataset, as shown in Table~\ref{tab:kitti-results} and Table~\ref{table:pose-ate} respectively. We infer the models using the TensorRT (FP16bit) on NVIDIA's Jetson AGX platform and report processed frames per second for all the tasks.\par
\subsection{Ablation Study of the Contributions}
\label{sec:features-ablation}

\begin{table}[t]
\centering
\small
\begin{adjustbox}{width=\columnwidth}
\setlength{\tabcolsep}{0.1em}
\begin{tabular}{lccccccccccccc}
\toprule
\textbf{Network} 
& \textit{\begin{tabular}[c]{@{}c@{}} Robust \\ loss\end{tabular}}
& \textit{\begin{tabular}[c]{@{}c@{}} Feature \\ loss\end{tabular}} 
& \textit{\begin{tabular}[c]{@{}c@{}} Semantic \\ Guide Dist.\end{tabular}}
& \textit{\begin{tabular}[c]{@{}c@{}} Semantic \\ Mask\end{tabular}}
& \textit{\begin{tabular}[c]{@{}c@{}} Motion \\ Mask\end{tabular}} 
& \textit{\begin{tabular}[c]{@{}c@{}} Semantic \\ Guide Det.\end{tabular}} 
& \textit{CGT} 
& \textit{\begin{tabular}[c]{@{}l@{}} Cyl \\ Rect. \end{tabular}} 
& \cellcolor[HTML]{00b0f0}\textbf{\textit{RMSE} $\mathbf{\downarrow}$}
& \cellcolor[HTML]{00b0f0} $\delta<1.25 \mathbf{\uparrow}$
& \multicolumn{1}{c}{\cellcolor[HTML]{00b050}
\begin{tabular}[c]{@{}c@{}}\textbf{\textit{mIoU}} \\ \textbf{\textit{Seg}} \end{tabular}}
& \multicolumn{1}{c}{\cellcolor[HTML]{ab9ac0}
\begin{tabular}[c]{@{}c@{}}\textbf{\textit{mIoU}} \\ \textbf{\textit{Mot.}}\end{tabular}} & \multicolumn{1}{c}{\cellcolor[HTML]{a5a5a5}
\begin{tabular}[c]{@{}c@{}}\textbf{\textit{mAP}} \\ \textbf{\textit{Det}}\end{tabular}} \\
\midrule
\multirow{10}{*}{\begin{tabular}[c]{@{}c@{}} OmniDet \\ (SAN10-patch) \end{tabular}}
& \ch & \xm & \xm & \xm & \xm & \xm & \xm & \xm & 2.153 & 0.875 & 73.2 & 71.8 & 63.3 \\  
& \ch & \ch & \xm & \xm & \xm & \xm & \xm & \xm & 1.764 & 0.897 & 73.6 & 72.3 & 63.5 \\  
& \ch & \ch & \xm & \xm & \xm & \xm & \ch & \xm & 1.681 & 0.902 & 74.2 & 73.5 & 63.8 \\  
& \ch & \ch & \ch & \ch & \xm & \xm & \xm & \xm & 1.512 & 0.905 & 74.5 & 73.4 & 63.6 \\  
& \ch & \ch & \ch & \ch & \xm & \xm & \ch & \xm & 1.442 & 0.908 & 74.8 & 74.0 & 64.1 \\   
& \ch & \ch & \ch & \xm & \ch & \xm & \xm & \xm & 1.397 & 0.915 & 75.2 & 74.3 & 64.0 \\
& \ch & \ch & \ch & \xm & \ch & \xm & \ch & \xm & 1.352 & 0.916 & 75.5 & 74.8 & 64.3 \\ 
& \ch & \ch & \ch & \xm & \ch & \ch & \xm & \xm & 1.348 & 0.915 & 75.9 & 74.9 & 67.8 \\   
& \ch & \ch & \ch & \xm & \ch & \ch & \ch & \xm & \textbf{1.332} & \textbf{0.918} & \textbf{76.6} & \textbf{75.3} & \textbf{68.4} \\
& \ch & \ch & \ch & \xm & \ch & \ch & \ch & \ch & 1.210 & 0.929 & 78.9 & 79.2 & 74.1 \\  
\midrule
\multirow{3}{*}{\begin{tabular}[c]{@{}c@{}} OmniDet \\ (SAN10-pair) \end{tabular}}
& \ch & \ch & \ch & \ch & \xm & \xm & \ch & \xm & 1.492 & 0.904 & 74.1 & 73.1 & 63.3 \\
& \ch & \ch & \ch & \xm & \ch & \ch & \ch & \xm & 1.321 & 0.911 & 75.4 & 74.6 & 67.6 \\   
& \ch & \ch & \ch & \xm & \ch & \ch & \ch & \ch & 1.272 & 0.919 & 77.1 & 77.4 & 72.6 \\
\midrule
\multirow{10}{*}{\begin{tabular}[c]{@{}c@{}} OmniDet \\ (SAN19-patch) \end{tabular}}
& \ch & \xm & \xm & \xm & \xm & \xm & \xm & \xm & 2.138 & 0.880 & 73.9 & 72.4 & 64.7 \\  
& \ch & \ch & \xm & \xm & \xm & \xm & \xm & \xm & 1.749 & 0.903 & 74.3 & 73.0 & 64.8 \\  
& \ch & \ch & \xm & \xm & \xm & \xm & \ch & \xm & 1.662 & 0.906 & 74.6 & 74.1 & 65.2 \\  
& \ch & \ch & \ch & \ch & \xm & \xm & \xm & \xm & 1.495 & 0.910 & 74.9 & 73.8 & 64.9 \\  
& \ch & \ch & \ch & \ch & \xm & \xm & \ch & \xm & 1.427 & 0.916 & 75.4 & 74.7 & 65.5 \\   
& \ch & \ch & \ch & \xm & \ch & \xm & \xm & \xm & 1.378 & 0.918 & 75.7 & 75.1 & 65.3 \\
& \ch & \ch & \ch & \xm & \ch & \xm & \ch & \xm & 1.331 & 0.922 & 76.2 & 75.6 & 65.9 \\ 
& \ch & \ch & \ch & \xm & \ch & \ch & \xm & \xm & 1.320 & 0.927 & 76.8 & 76.2 & 69.6 \\   
& \ch & \ch & \ch & \xm & \ch & \ch & \ch & \xm & \textbf{1.304} & \textbf{0.931} & \textbf{77.4} & \textbf{77.0} & \textbf{71.5} \\
& \ch & \ch & \ch & \xm & \ch & \ch & \ch & \ch & 1.177 & 0.938 & 80.2 & 80.5 & 76.3 \\
\bottomrule
\end{tabular}
\end{adjustbox}
\caption[\bf Ablation study on the effect of our contributions up to the final OmniDet model on the Woodscape.]
{\textbf{Ablation study on the effect of our contributions} up to the final OmniDet model on the Woodscape.}
\label{table:features}
\end{table}
For the ablation analysis of the main features shown in Table~\ref{table:features}, we consider two variants of the self-attention encoder, namely pairwise and patchwise, as described in Section~\ref{sec:vector-self-attention}. First, we replace the \lone loss with a generic parameterized loss function and test it using the self-attention encoder's patchwise variant. We cap the distance estimates to 40m. We achieve significant gains in this setting by attributing better-supervised signal provided by using discriminative features $\mathcal{L}_{dis}$ as described in Section~\ref{sec:discriminative-loss} where incorrect distance values are appropriately penalized with more considerable losses along with the combination of $\mathcal{L}_{cvt}$ in Section~\ref{sec:converget-loss} wherein a correct optimization direction is provided. These losses help the gradient descent approaches to transit smoothly towards optimal solutions. When adding the CGT to this setting, we observe a significant increase in accuracy since we train multiple cameras with different camera intrinsics and viewing angles. For the OmniDet framework to be operational in the first place, this an important feature. The aforementioned training strategy makes the network camera-independent and generalizes better to images taken from a different camera.\par
\begin{figure*}[!ht]
  \centering
  \newcommand{\turnheightnew}{0.33\columnwidth}
\begin{adjustbox}{width=\textwidth}
\begin{tabular}{c@{\hskip 0.4mm}c@{\hskip 0.4mm}c@{\hskip 0.4mm}}

\includegraphics[height=\turnheightnew]{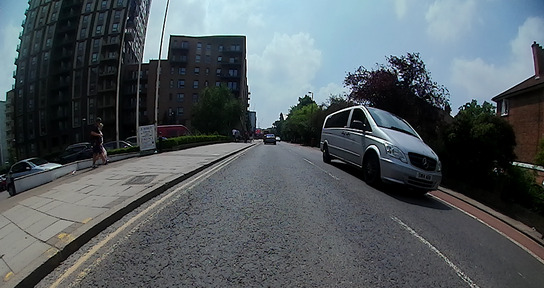} &
\includegraphics[height=\turnheightnew]{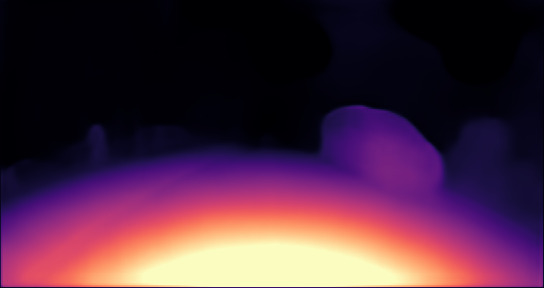} &
\includegraphics[height=\turnheightnew]{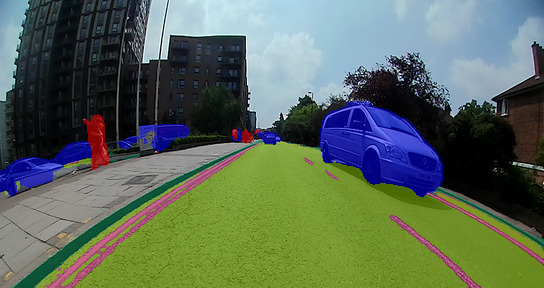} \\

\includegraphics[height=\turnheightnew]{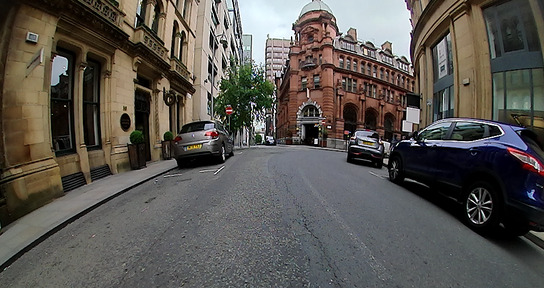} &
\includegraphics[height=\turnheightnew]{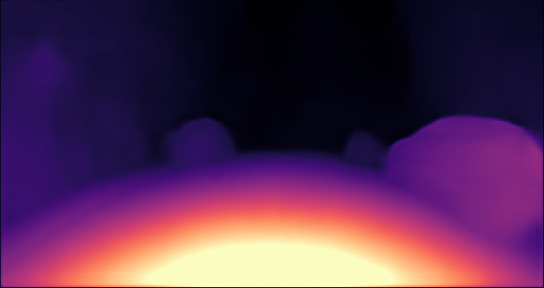} &
\includegraphics[height=\turnheightnew]{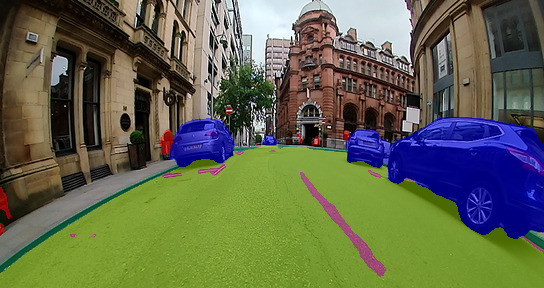} \\

\includegraphics[height=\turnheightnew]{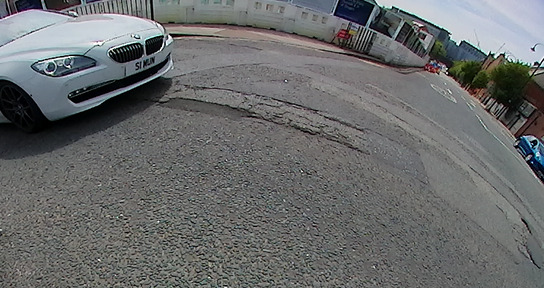} &
\includegraphics[height=\turnheightnew]{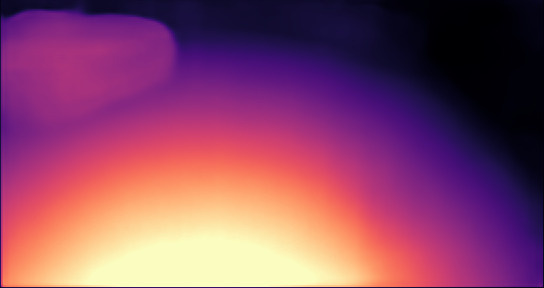} &
\includegraphics[height=\turnheightnew]{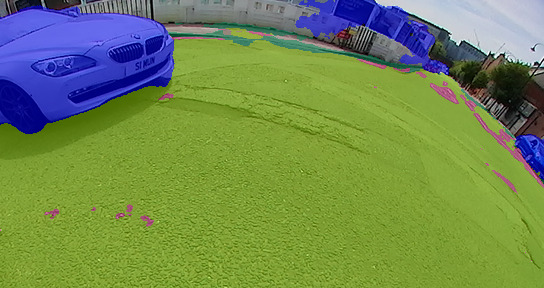} \\

\includegraphics[height=\turnheightnew]{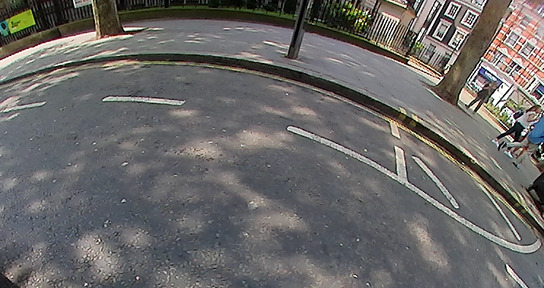} &
\includegraphics[height=\turnheightnew]{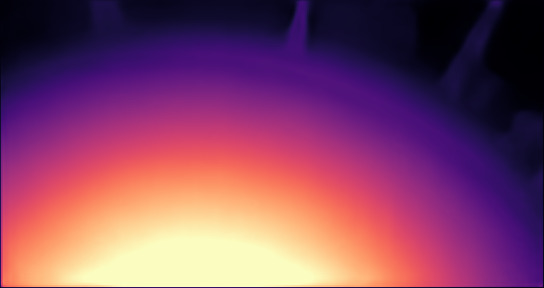} &
\includegraphics[height=\turnheightnew]{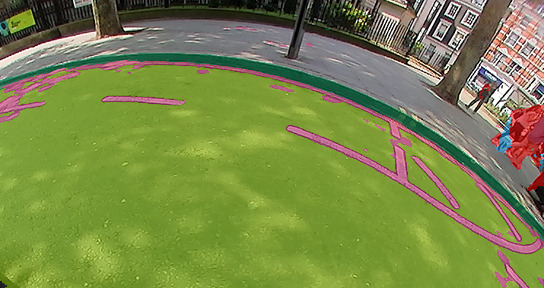} \\

\includegraphics[height=\turnheightnew]{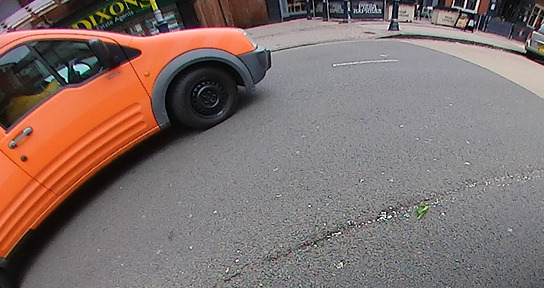} &
\includegraphics[height=\turnheightnew]{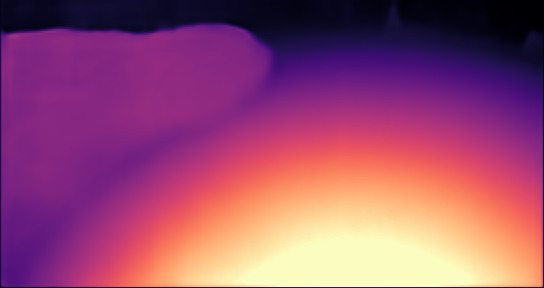} &
\includegraphics[height=\turnheightnew]{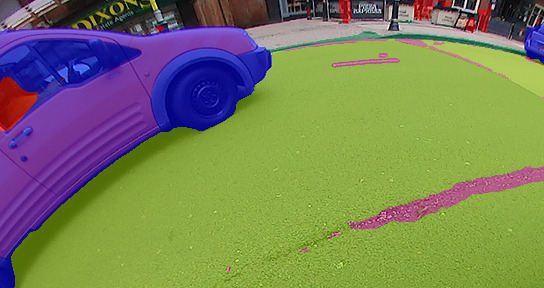} \\

\includegraphics[height=\turnheightnew]{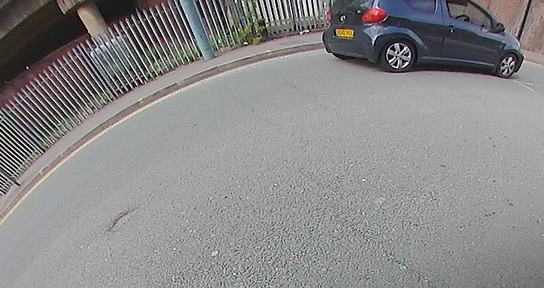} &
\includegraphics[height=\turnheightnew]{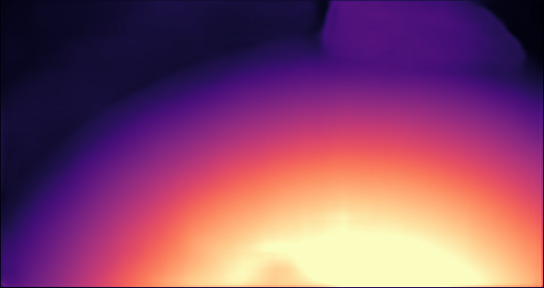} &
\includegraphics[height=\turnheightnew]{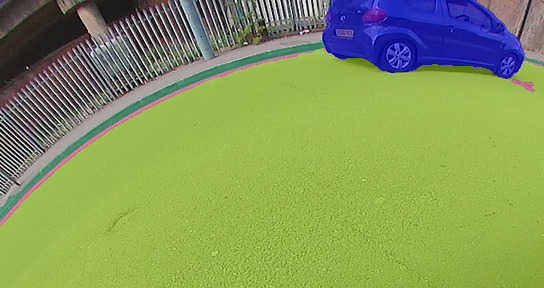} \\

\includegraphics[height=\turnheightnew]{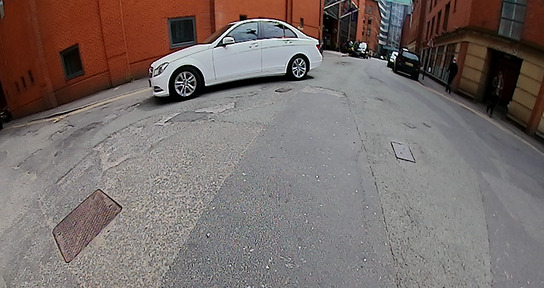} &
\includegraphics[height=\turnheightnew]{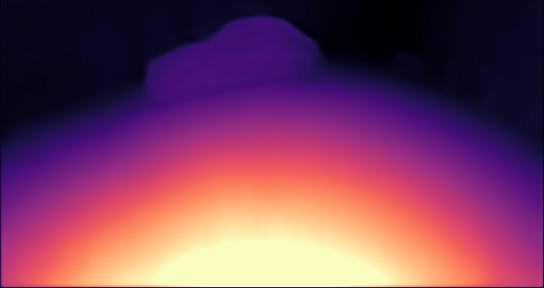} &
\includegraphics[height=\turnheightnew]{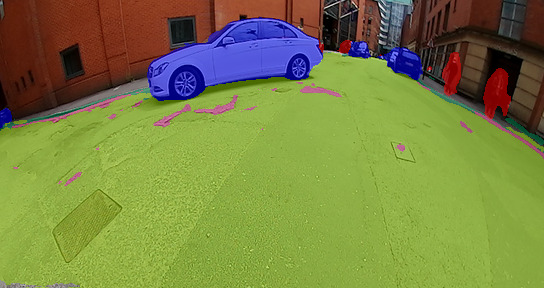} \\

\includegraphics[height=\turnheightnew]{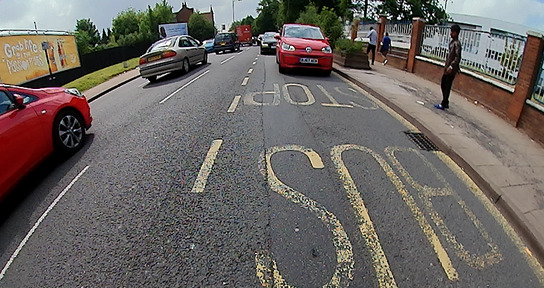} &
\includegraphics[height=\turnheightnew]{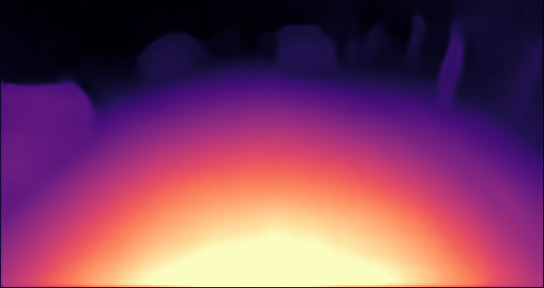}  &
\includegraphics[height=\turnheightnew]{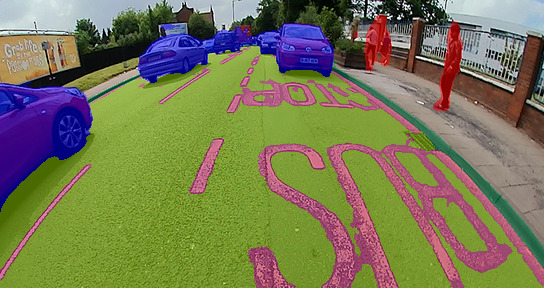} \\

\end{tabular}
\end{adjustbox}
  \caption[\bf Qualitative results of distance estimation and semantic segmentation on raw surround-view fisheye cameras on the WoodScape dataset.]
  {\bf Qualitative results of distance estimation and semantic segmentation on raw surround-view fisheye cameras on the WoodScape dataset. The 1\textsuperscript{st} column contains two rows of input images from each of the Front, Left, Right, and Rear cameras.}
  \label{fig:omnidet-norm-sem-raw}
\end{figure*}
\begin{figure*}[!ht]
  \resizebox{\textwidth}{!}{\newcommand{\turnheightnew}{0.25\linewidth}
\centering

\begin{tabular}{@{\hskip 0.5mm}c@{\hskip 0.5mm}c@{\hskip 0.5mm}c@{\hskip 0.5mm}c@{\hskip 0.5mm}c@{}}

\includegraphics[height=\turnheightnew]{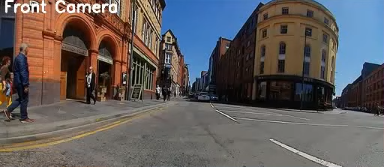} &
\includegraphics[height=\turnheightnew]{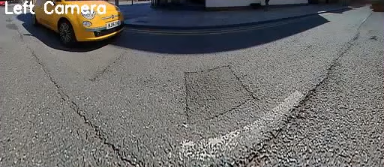} &
\includegraphics[height=\turnheightnew]{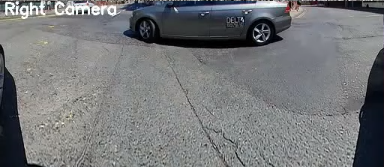} &
\includegraphics[height=\turnheightnew]{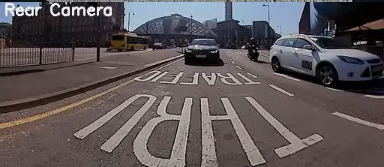} \\

\includegraphics[height=\turnheightnew]{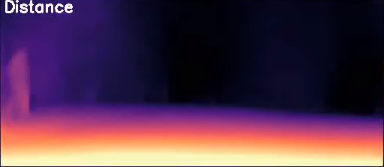} &
\includegraphics[height=\turnheightnew]{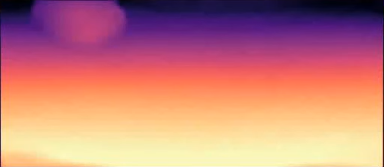} &
\includegraphics[height=\turnheightnew]{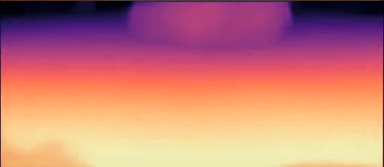} &
\includegraphics[height=\turnheightnew]{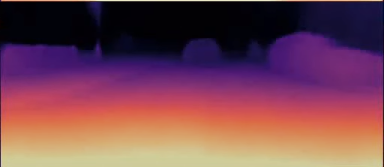} \\

\includegraphics[height=\turnheightnew]{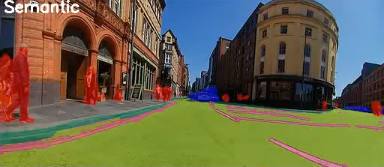} &
\includegraphics[height=\turnheightnew]{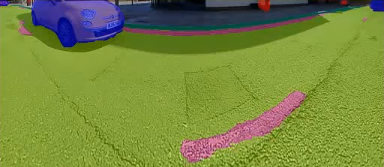} &
\includegraphics[height=\turnheightnew]{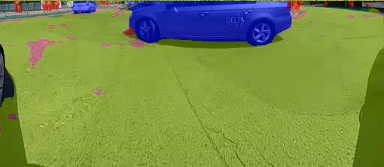} &
\includegraphics[height=\turnheightnew]{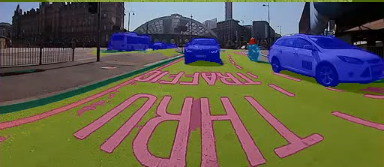} \\

\includegraphics[height=\turnheightnew]{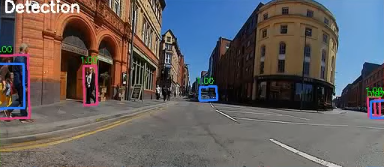} &
\includegraphics[height=\turnheightnew]{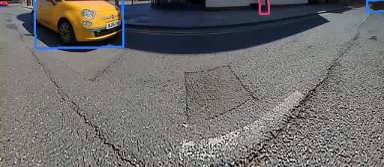} &
\includegraphics[height=\turnheightnew]{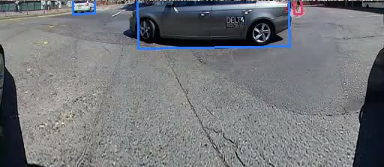} &
\includegraphics[height=\turnheightnew]{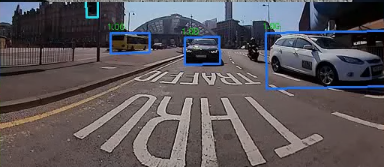} \\

\includegraphics[height=\turnheightnew]{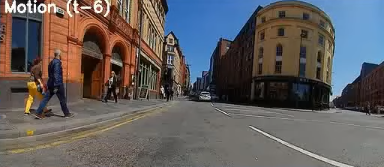} &
\includegraphics[height=\turnheightnew]{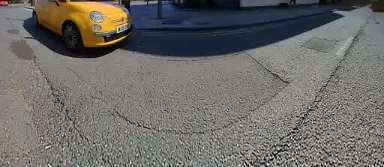} &
\includegraphics[height=\turnheightnew]{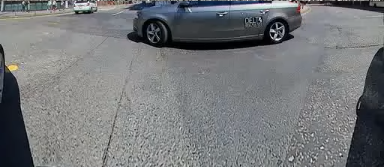} &
\includegraphics[height=\turnheightnew]{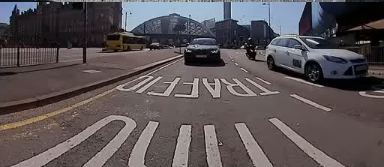} \\

\includegraphics[height=\turnheightnew]{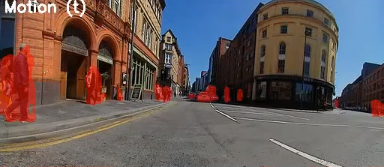} &
\includegraphics[height=\turnheightnew]{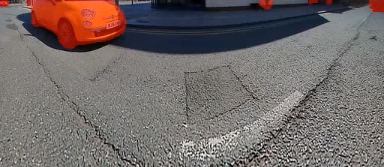} &
\includegraphics[height=\turnheightnew]{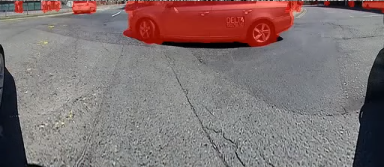} &
\includegraphics[height=\turnheightnew]{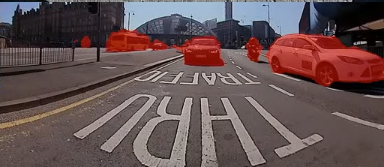} \\

\includegraphics[height=\turnheightnew]{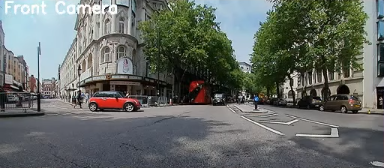} &
\includegraphics[height=\turnheightnew]{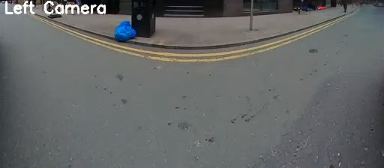} &
\includegraphics[height=\turnheightnew]{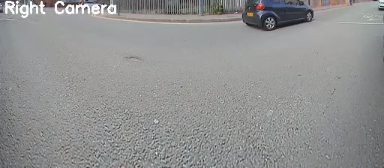} &
\includegraphics[height=\turnheightnew]{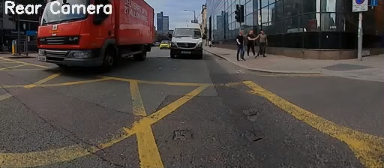} \\

\includegraphics[height=\turnheightnew]{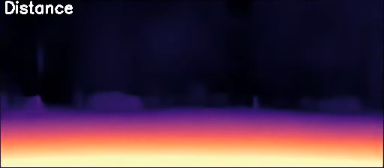} &
\includegraphics[height=\turnheightnew]{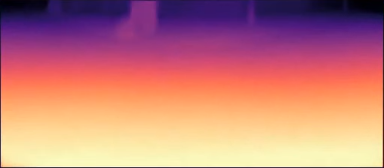} &
\includegraphics[height=\turnheightnew]{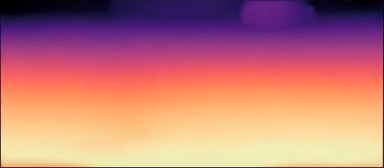} &
\includegraphics[height=\turnheightnew]{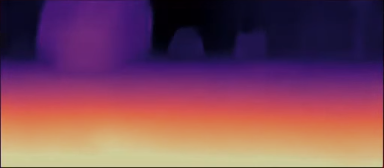} \\

\includegraphics[height=\turnheightnew]{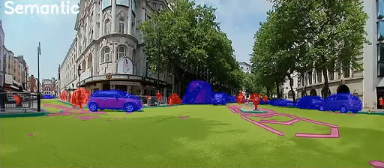} &
\includegraphics[height=\turnheightnew]{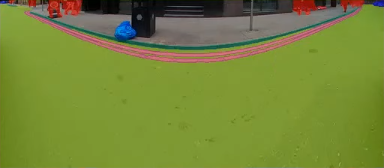} &
\includegraphics[height=\turnheightnew]{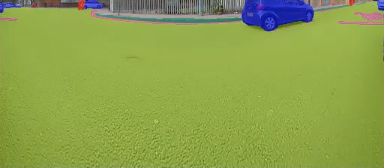} &
\includegraphics[height=\turnheightnew]{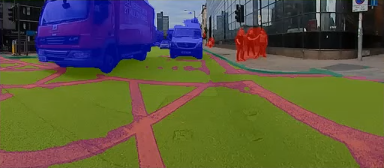} \\

\includegraphics[height=\turnheightnew]{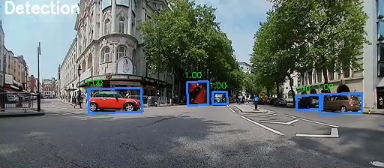} &
\includegraphics[height=\turnheightnew]{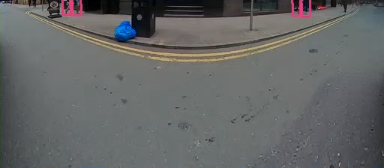} &
\includegraphics[height=\turnheightnew]{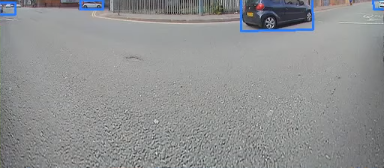} &
\includegraphics[height=\turnheightnew]{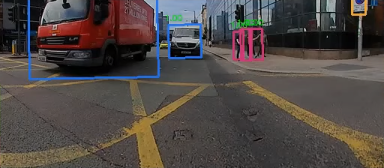} \\

\includegraphics[height=\turnheightnew]{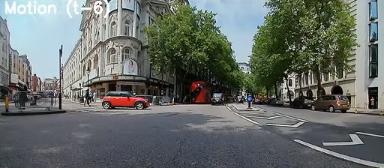} &
\includegraphics[height=\turnheightnew]{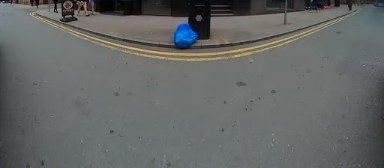} &
\includegraphics[height=\turnheightnew]{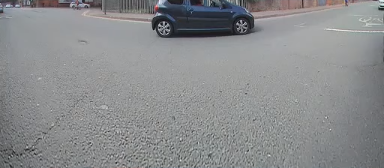} &
\includegraphics[height=\turnheightnew]{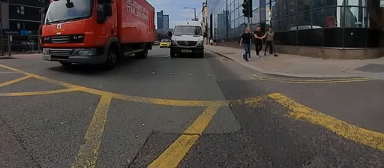} \\

\includegraphics[height=\turnheightnew]{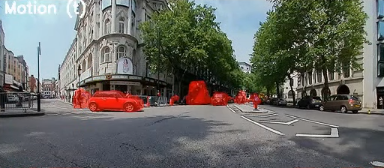} &
\includegraphics[height=\turnheightnew]{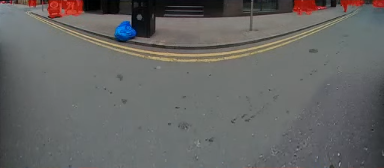} &
\includegraphics[height=\turnheightnew]{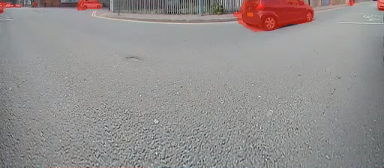} &
\includegraphics[height=\turnheightnew]{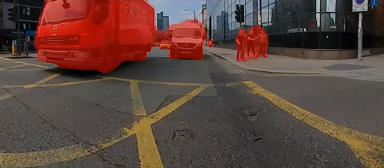} \\

\end{tabular}
}
  \caption[\bf Qualitative results of cylindrical rectified images from the OmniDet framework on the WoodScape dataset.]
          {{\bf Qualitative results of cylindrical rectified images from the OmniDet framework on the WoodScape dataset.} The 1\textsuperscript{st} and 7\textsuperscript{th} rows indicates the input images from Front, Left, Right and Rear cameras. 2\textsuperscript{nd} and 8\textsuperscript{th} rows indicate distance estimates, 3\textsuperscript{rd} and 9\textsuperscript{th} rows indicate semantic segmentation maps, 4\textsuperscript{th} and 10\textsuperscript{th} rows indicate object detection of standard boxes representations, and finally 5\textsuperscript{th} and 11\textsuperscript{th} rows show input frames at (t-6) for reference to compare the motion segmentation in 6\textsuperscript{th} and 12\textsuperscript{th} rows at time (t). For more qualitative results at a higher resolution, we refer to this video: \url{https://youtu.be/xbSjZ5OfPes}}
\label{fig:omnidet-qual_cyl}
\end{figure*}

\begin{table*}[!t]
\centering
\scalebox{0.85}{
\small
\setlength{\tabcolsep}{0.3em}
\begin{tabular}{ccccccc}
\toprule
  \textit{\begin{tabular}[c]{@{}c@{}}Semantic\\ Mask\end{tabular}} &
  \textit{\begin{tabular}[c]{@{}c@{}}Motion\\ Mask\end{tabular}} &
  \textit{CGT} &
  \cellcolor[HTML]{00b0f0}\textbf{\textit{RMSE} $\mathbf{\downarrow}$} &
  \multicolumn{1}{c}{\cellcolor[HTML]{00b050}\begin{tabular}[c]{@{}c@{}}\textbf{\textit{mIoU}}\\ \textbf{\textit{Seg}} \end{tabular}} &
  \multicolumn{1}{c}{\cellcolor[HTML]{ab9ac0}\begin{tabular}[c]{@{}c@{}}\textbf{\textit{mIoU}}\\ \textbf{\textit{Mot.}}\end{tabular}} &
  \multicolumn{1}{c}{\cellcolor[HTML]{a5a5a5}\begin{tabular}[c]{@{}c@{}}\textbf{\textit{mAP}}\\ \textbf{\textit{Det}}\end{tabular}} \\
\midrule
\ch & \xm & \xm & 1.512 & 74.5 & 73.4 & 63.6 \\  
\ch & \xm & \ch & 1.442 & 74.8 & 74.0 & 64.1 \\   
\xm & \ch & \xm & 1.397 & 75.2 & 74.3 & 64.0 \\
\xm & \ch & \ch & 1.352 & 75.5 & 74.8 & 64.3 \\
\bottomrule
\end{tabular}
}
\caption{\textbf{Ablation study of dynamic object filtering using semantic and motion segmentation masks.}}
\label{table:dynamic_sem_mot}
\end{table*}
To achieve synergy between geometry and semantic features, we add semantic guidance to the distance decoder. It helps to reason about geometry and content within the same shared features and to disambiguate photometric ambiguities. To establish a robust reconstruction loss free from the dynamic objects' contamination, we introduce semantic and motion masks as described in Section~\ref{sec:dynamic-object-mask-1}, to filter all the dynamic objects. Motion mask-based filtering yields superior gains along with CGT compared to using semantic masks, which is also shown in Table \ref{table:dynamic_sem_mot} as semantics might not contain all the dynamic objects in its set of classes as indicated in Eq.~\ref{eq:semantic_mask}. Figure~\ref{fig:omnidet-norm-sem-raw}. presents the qualitative results of the distance estimation and semantic segmentation tasks. Finally, to complete the synergy, we use semantically guided features to the detection decoder described in Section~\ref{sec:Semantically-Guided-Distance-Decoder}, which yields significant gains in mAP, and overall results for all the tasks are inherently improved with better-shared features. All the contributed features and the synergy between tasks help the OmniDet framework to achieve a good scene understanding with high accuracy in each task's predictions. To enable a single model to handle the different intrinsics, we re-projected all input images to the same central cylindrical projection in the first step. In vertical direction, it resembles a rectilinear projection $y_{\mathrm{I}}=f\cdot\tan(\theta')$, where $\theta'=\arctan\left(y/\sqrt{x^2+z^2}\right)$. In horizontal direction it resembles a equirectangular projection $x_{\mathrm{I}}=f\cdot\theta''$, where $\theta''=\arctan\left(x/z\right)$. Cylindrical rectification (Cyl Rect.) provides a good trade-off between loss of field-of-view and reducing distortion~\cite{yogamani2019woodscape}. The qualitative results of the Cylindrical rectified image predictions on the perception tasks are shown in Figure~\ref{fig:omnidet-qual_cyl}.
\subsection{State-of-the-Art Comparison on KITTI} 
\label{sec:kitti-depth-pose}

\begin{table}[!t]
\centering
\setlength{\tabcolsep}{0.25em}
\begin{adjustbox}{width=\columnwidth}
\small
\begin{tabular}{c|lcccccccc}
\toprule
& \multicolumn{1}{c}{\textbf{Method}} 
& \multicolumn{1}{c}{\textit{\cellcolor[HTML]{7d9ebf} Abs$_{rel}$}} 
& \textit{\cellcolor[HTML]{7d9ebf} Sq$_{rel}$} 
& \multicolumn{1}{c}{\cellcolor[HTML]{7d9ebf} \textit{RMSE}} 
& \multicolumn{1}{c}{\cellcolor[HTML]{7d9ebf} \textit{RMSE$_{log}$}}
& \multicolumn{1}{c}{\cellcolor[HTML]{e8715b} \textit{$\delta < 1.25$}}
& \multicolumn{1}{c}{\cellcolor[HTML]{e8715b} \textit{$\delta < 1.25^2$}}
& \multicolumn{1}{c}{\cellcolor[HTML]{e8715b} \textit{$\delta < 1.25^3$}} \\ 
\cmidrule(l){3-6} \cmidrule(lr){7-9} &
& \multicolumn{4}{c}{\cellcolor[HTML]{7d9ebf} lower is better}
& \multicolumn{3}{c}{\cellcolor[HTML]{e8715b} higher is better} \\
\midrule
\parbox[t]{2mm}{\multirow{10}{*}{\rotatebox[origin=c]{90}{Original~\cite{Eigen_14}}}}
& Monodepth2~\cite{godard2019digging}        & 0.115 & 0.903 & 4.863 & 0.193 & 0.877 & 0.959 & 0.981 \\
& PackNet-SfM~\cite{guizilini2019packnet}    & 0.111 & 0.829 & 4.788 & 0.199 & 0.864 & 0.954 & 0.980 \\
& FisheyeDistanceNet~\cite{kumar2020fisheyedistancenet} & 0.117 & 0.867 & 4.739 & 0.190 & 0.869 & 0.960 & 0.982 \\
& UnRectDepthNet~\cite{kumar2020unrectdepthnet} & 0.107 & 0.721 & 4.564 & 0.178 & 0.894 & 0.971 & \textbf{0.986} \\
& SynDistNet~\cite{kumar2020syndistnet}      & 0.109 & 0.718 & 4.516 & 0.180 & 0.896 & 0.973 & \textbf{0.986} \\
& Shu \etal~\cite{shu2020featdepth}          & 0.104 & 0.729 & 4.481 & 0.179 & 0.893 & 0.965 & 0.984 \\
& OmniDet                                    & \textbf{0.092} & \textbf{0.657} & \textbf{3.984} & \textbf{0.168} & \textbf{0.914} & \textbf{0.975} & \textbf{0.986} \\ 
\cmidrule{2-9}
& Struct2Depth${^*}$~\cite{casser2019depth}  & 0.109 & 0.825 & 4.750 & 0.187 &0.874 & 0.958 & 0.983 \\
& GLNet${^*}$~\cite{Chen2019b}               & 0.099 & 0.796 & 4.743 & 0.186 &0.884 & 0.955 & 0.979 \\
& Shu${^*}$ \etal~\cite{shu2020featdepth}    & 0.088 & 0.712 & 4.137 & 0.169 & 0.915 & 0.965 & 0.982 \\
& OmniDet${^*}$                              & \textbf{0.077} & \textbf{0.641} & \textbf{3.859} & \textbf{0.152} & \textbf{0.931} & \textbf{0.979} & \textbf{0.989} \\
\midrule
\parbox[t]{2mm}{\multirow{6}{*}{\rotatebox[origin=c]{90}{Improved~\cite{uhrig2017sparsity}}}}
& Monodepth2~\cite{godard2019digging}        & 0.090 & 0.545 & 3.942 & 0.137 & 0.914 & 0.983 & 0.995 \\
& PackNet-SfM~\cite{guizilini2019packnet}    & 0.078 & 0.420 & 3.485 & 0.121 & 0.931 & 0.986 & 0.996 \\
& UnRectDepthNet~\cite{kumar2020unrectdepthnet} & 0.081 & 0.414 & 3.412 & 0.117 & 0.926 & 0.987 & 0.996 \\
& SynDistNet~\cite{kumar2020syndistnet}      & 0.076 & 0.412 & 3.406 & 0.115 & 0.931 & 0.988 & 0.996 \\
& OmniDet                                    & \textbf{0.067} & \textbf{0.306} & \textbf{3.098} & \textbf{0.101} & \textbf{0.944} & \textbf{0.991} & \textbf{0.997} \\
& OmniDet${^*}$                              & \textbf{0.048} & \textbf{0.287} & \textbf{2.913} & \textbf{0.081} & \textbf{0.948} & \textbf{0.991} & \textbf{0.998} \\
\bottomrule
\end{tabular}
\end{adjustbox}
\caption[\bf Evaluation of depth estimation in OmniDet on the KITTI dataset.]
{\textbf{Evaluation of depth estimation} on the KITTI Eigen~\cite{Eigen_14} split.}
\label{tab:kitti-results}
\end{table}
To facilitate comparison to previous methods, we also train the distance estimation method in the classical depth estimation setting on the KITTI Eigen split~\cite{geiger2013vision} whose results are shown in Table~\ref{tab:kitti-results}. With the synergy between depth, semantic, motion, and detection tasks along with the features ablated in Table~\ref{table:features} and their importance explained in Section~\ref{sec:features-ablation}, \textit{we outperform all previous monocular methods.}
Following best practices, we cap depths at $80\,m$. We also evaluate using the \textit{Original}~\cite{Eigen_14} as well as \textit{Improved}~\cite{uhrig2017sparsity} ground truth depth maps. Method${^*}$ indicates the online refinement technique~\cite{casser2019depth}, where the model is also trained during inference. Using the online refinement method from~\cite{casser2019depth}, we obtain a significant improvement.\par
\begin{table}[t]
\centering
\begin{adjustbox}{width=0.75\columnwidth}
\begin{tabular}{lcccc}
\toprule
\multicolumn{1}{l}{\textbf{Method}} 
& \textit{\begin{tabular}[c]{@{}c@{}}No. of\\ Frames\end{tabular}} 
& \textit{GT} 
& \cellcolor[HTML]{7d9ebf}\textit{Sequence 09} 
& \cellcolor[HTML]{e8715b}\textit{Sequence 10} \\ 
\midrule
    ORB-SLAM~\cite{mur2015orb}                   & 5 & \ch & 0.014 $\pm$ 0.008 & 0.012 $\pm$ 0.011 \\
    DF-Net~\cite{zou2018df}                      & 5 & \ch & 0.017 $\pm$ 0.007 & 0.015 $\pm$ 0.009 \\
    SfMLearner~\cite{zhou2017unsupervised}       & 5 & \ch & 0.016 $\pm$ 0.009 & 0.013 $\pm$ 0.009 \\
    Klodt et al.~\cite{klodt2018supervising}     & 5 & \ch & 0.014 $\pm$ 0.007 & 0.013 $\pm$ 0.009 \\
    GeoNet~\cite{yin2018geonet}                  & 5 & \ch & 0.012 $\pm$ 0.007 & 0.012 $\pm$ 0.009 \\
    Struct2Depth~\cite{casser2019depth}          & 5 & \ch & 0.011 $\pm$ 0.006 & 0.011 $\pm$ 0.010 \\
    Ranjan~\cite{ranjan2019competitive}          & 5 & \ch & 0.011 $\pm$ 0.006 & 0.011 $\pm$ 0.010 \\ 
    PackNet-SfM~\cite{guizilini2019packnet}      & 5 & \ch & 0.010 $\pm$ 0.005 & 0.009 $\pm$ 0.008 \\
    PackNet-SfM~\cite{guizilini2019packnet}      & 5 & \xm & 0.014 $\pm$ 0.007 & 0.012 $\pm$ 0.008 \\
    OmniDet                                      & 5 & \ch & \textbf{0.009} $\pm$ \textbf{0.004} & 0.008 $\pm$ \textbf{0.005} \\
    OmniDet                                      & 5 & \xm & \textbf{0.010} $\pm$ \textbf{0.005} & \textbf{0.010} $\pm$ \textbf{0.008} \\
    \midrule
    DDVO~\cite{Wang_2018_CVPR}                   & 3 & \ch & 0.045 $\pm$ 0.108 & 0.033 $\pm$ 0.074 \\
    Vid2Depth~\cite{mahjourian2018unsupervised}  & 3 & \ch & 0.013 $\pm$ 0.010 & 0.012 $\pm$ 0.011 \\
    EPC++~\cite{luo2019every}                    & 3 & \ch & 0.013 $\pm$ 0.007 & 0.012 $\pm$ 0.008 \\
    OmniDet                                      & 3 & \ch & \textbf{0.011} $\pm$ \textbf{0.006} & \textbf{0.010} $\pm$ \textbf{0.007} \\
    OmniDet                                      & 3 & \xm & \textbf{0.012} $\pm$ \textbf{0.007} & \textbf{0.011} $\pm$\textbf{ 0.008} \\
    \midrule
    Monodepth2~\cite{godard2019digging}          & 2 & \ch & 0.017 $\pm$ 0.008 & 0.015 $\pm$ 0.010 \\
    OmniDet                                      & 2 & \ch & \textbf{0.015} $\pm$ \textbf{0.007} & \textbf{0.013} $\pm$ \textbf{0.007} \\
    OmniDet                                      & 2 & \xm & \textbf{0.016} $\pm$ \textbf{0.008} & \textbf{0.014} $\pm$ \textbf{0.009} \\
    \bottomrule
\end{tabular}
\end{adjustbox}
\caption[\bf Evaluation of the pose estimation in OmniDet on the KITTI Odometry benchmark.]
{\textbf{Evaluation of the pose estimation} on the KITTI Odometry Benchmark~\cite{geiger2013vision}.}
\label{table:pose-ate}
\end{table}
In Table~\ref{table:pose-ate}, we report the average trajectory error of the pose estimation network in meters by following the same protocols as Zhou~\cite{zhou2017unsupervised} on the official KITTI odometry split (containing $11$ sequences with ground-truth (GT) odometry acquired with the IMU/GPS measurements, which is used for evaluation purpose only), and use sequences 00-08 for training and 09-10 for testing. We outperform the previous methods listed in Table~\ref{table:pose-ate}, mainly by applying the bundle adjustment framework inflicted by the cross-sequence distance consistency loss~\cite{kumar2020fisheyedistancenet} which induces more constraints and simultaneously optimizes distances and camera poses for an implicitly extended training input sequence. This provides additional consistency constraints that are not induced by previous methods.\par
\begin{table}[!t]
\centering
\scalebox{0.7}{
\begin{adjustbox}{width=0.7\columnwidth}
\small
\begin{tabular}{lccc}
\toprule
\textit{\textbf{Representation}} 
& \multicolumn{1}{c}{\cellcolor[HTML]{7d9ebf}
\begin{tabular}[c]{@{}c@{}}\textit{mIoU} \\ \textit{GT}\end{tabular}}
& \textit{\cellcolor[HTML]{e8715b} mIOU} 
& \cellcolor[HTML]{00b050}
\begin{tabular}[c]{@{}l@{}}\textit{No. of} \\ \textit{params}\end{tabular} \\ 
\midrule
Standard Box     & 51.3 & 31.6 & 4 \\
Curved Box       & 52.5 & 32.3 & 6 \\
Oriented Box     & 53.9 & 33.6 & 5 \\
Ellipse          & 55.5 & 35.4 & 5 \\
\textbf{24-sided Polygon} & \textbf{86.6} & \textbf{44.6} & \textbf{48} \\
\bottomrule
\end{tabular}
\end{adjustbox}
}
\caption{\bf Evaluation of various object detection representations.}
\label{tab:fisheye-yolov3}
\vspace{-1em}
\end{table} 
For object detection on native fisheye images, in addition to the standard 2D box representation, we benchmark oriented boxes, ellipse, curved boxes, and 24-sided polygon representations in Table~\ref{tab:fisheye-yolov3}. Here mIoU GT represents the maximum performance we can achieve in terms of instance segmentation by using each representation. It is computed between the ground truth instance segmentation and the ground truth of the corresponding representation. Whereas mIoU represents the evaluation of the performance achieved on the network estimates. We also list the number of parameters involved in the model for each representation to provide a complexity comparison.\par
\subsection{Analysis on Adversarial Attacks}

We conduct the experiments across four visual perception tasks, excluding the pose estimation and soiling segmentation tasks, on a test set of 100 images, \ie, randomly sampled from the original test set of the target network. We generate adversarial examples for each image in the test set while attacking one task at a time. For white-box attacks, given the available gradients, we perform an iterative optimization process to add perturbation in the input image in a direction to harm the original predictions. For the black box attacks, we set up similar protocols as established for white box attacks; however, the gradients are not given but estimated. As a generic black-box optimization algorithm, we show that Evolution Strategies (ES) can be adopted as a black-box optimization method for generating adversarial examples. Precisely, the ES algorithm is used to update the adversarial example over the attacking steps. At each step, we take the adversarial example vector, \ie, the adversarial example is the perturbed image, and generate a population of 25 slightly different vectors by adding noise sampled from a normal distribution. Then, we evaluate each of the individuals by feeding it through the network. Finally, the newly updated vector is the weighted sum of the population vectors. Each weight is proportional to the task's desired performance, and the process continues till convergence defined by a stopping criterion.\par

In the untargeted case, the aim is to inflict maximum harm to the predictions without considering a certain target prediction: $f(x_{adv}) \neq y_{true}$, however in the targeted case, the aim is to harm the predictions in a desired specific way towards a certain target: $f(x_{adv}) = y_{target}$. The attack loss is based on the task. Mean square error (MSE) is used for the distance task, while cross-entropy loss is used for motion and semantic segmentation tasks. For the object detection task, only object confidence is attacked; hence the cross-entropy loss is adopted. Regarding untargeted attacks across all the tasks, the goal is to maximize the distance between the original output of the network and the adversarial example's output. Accordingly, we add the perturbations to achieve this simple goal where the output can be anything but the correct one. This can be formulated as $\theta = \theta + \alpha dJ/d\theta$ where $\theta$ is the image parameters \ie pixels, $J$ the loss functions and $\alpha$ is the learning rate. However, for the targeted attacks, the target output is defined. The aim is to minimize the distance between the original output and the target output according to $\theta = \theta - \alpha dJ/d\theta$.\par

For each perception task, the targets are as follows: The \textit{Targeted Depth} attack tries to convert the predicted near pixels to be predicted as far. The \textit{Targeted Segmentation} attack tries to convert the predicted vehicle pixels as void for randomly 50\% of the test set. For the other 50\%, the attack tries to convert the predicted road pixels to void. The \textit{Targeted Motion} attack tries to convert the predicted dynamic object pixels as static. Finally, similar to semantic segmentation, the \textit{Targeted Object Detection} attack tries to increase or decrease the predicted confidence randomly. In addition to attacks, we apply a simple blurring defense approach across all attacks. Similar to~\cite{dziugaite2016study}, the intuition is to try to remove the adversarial perturbations and restore the original output as much as possible. The hyperparameters of the attacks are empirically defined based on a minimal validation set of three samples. All white-box attacks are conducted with a learning rate of $\alpha = 0.00015$. In black-box attacks, hyperparameters are chosen to balance the attack effect and the severity of the perturbations. The learning rates range from $0.0001$ to $0.001$, and $\mu=0, \sigma = 0.05$ for ES population generation.\par
\subsubsection{Adversarial attacks results}

In this subsection, we present and discuss the results of w.r.t. adversarial attacks. As expected, white-box attacks, where the gradients are accessible, were easier to optimize than the black box case. White box attacks can generate adversarial examples with minimal and localized perturbations across all tasks. On the other hand, ES black-box attacks have more significant perturbations and require more hyperparameters to optimize.\par

\begin{figure*}[t]
    \centering

    \includegraphics[width=0.49\textwidth, trim={2.35cm 1.4cm 2.55cm 1.6cm},clip]{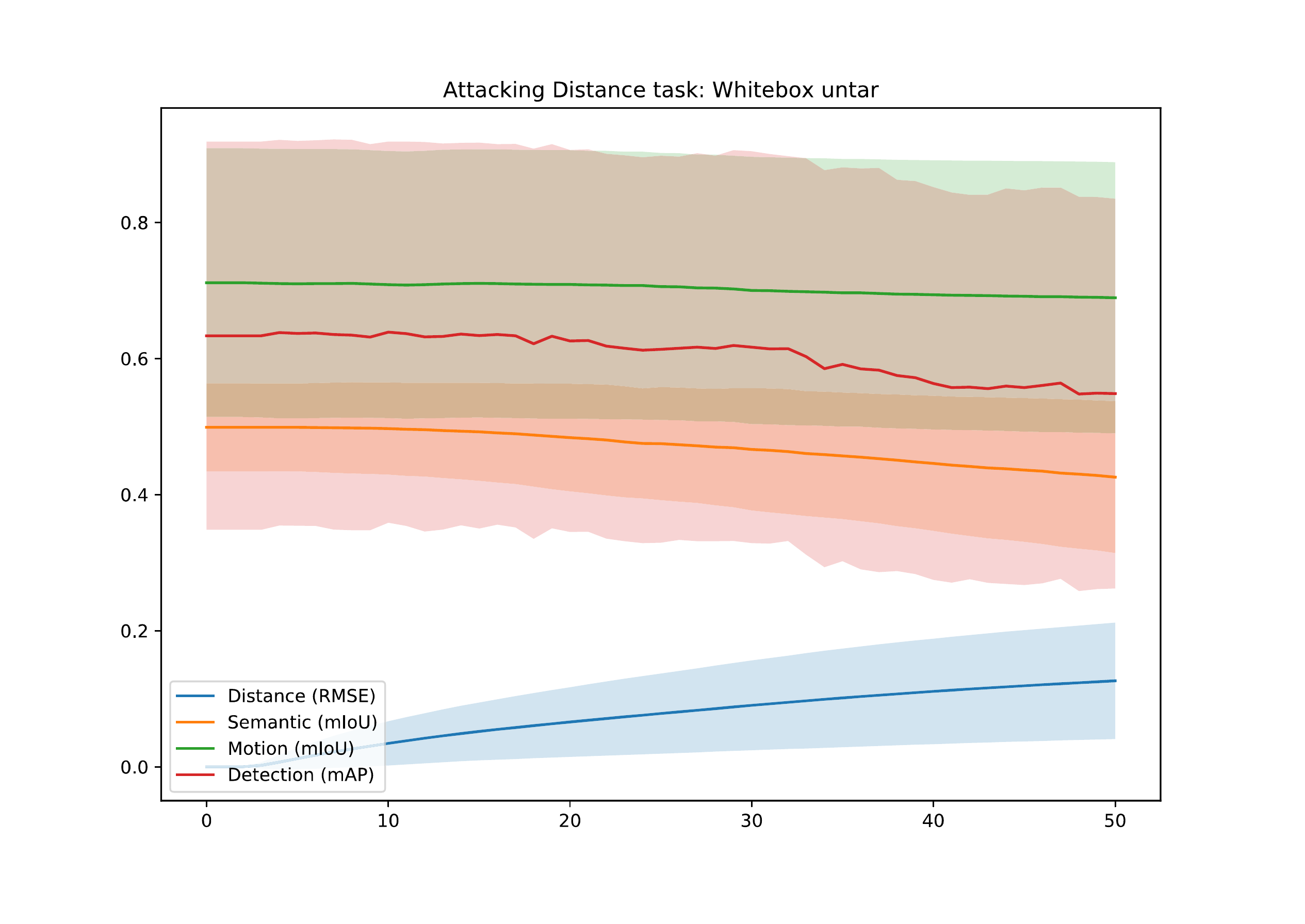}
    \includegraphics[width=0.49\textwidth, trim={2.35cm 1.4cm 2.55cm 1.6cm},clip]{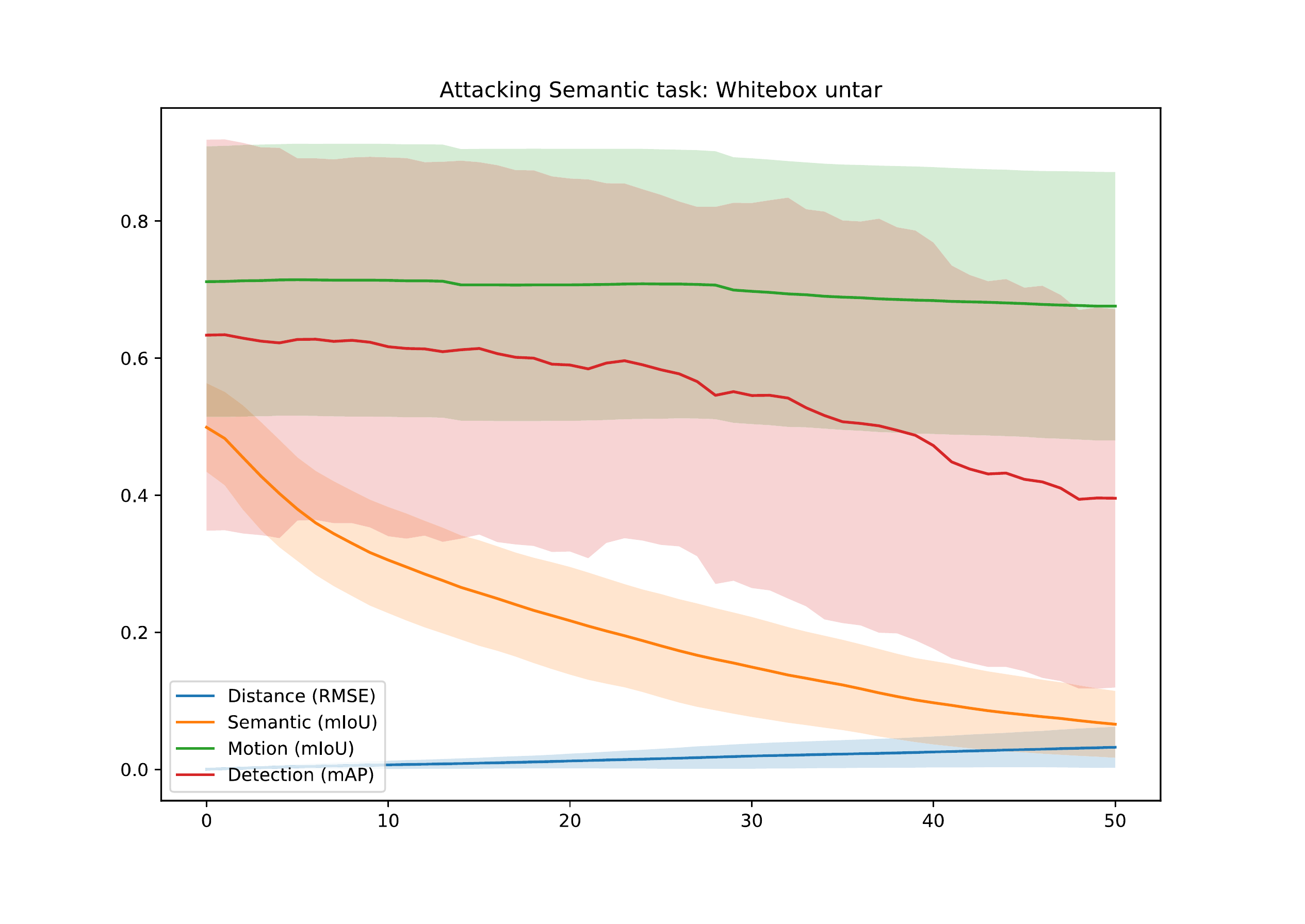}  
    \includegraphics[width=0.49\textwidth, trim={2.35cm 1.4cm 2.55cm 1.6cm},clip]{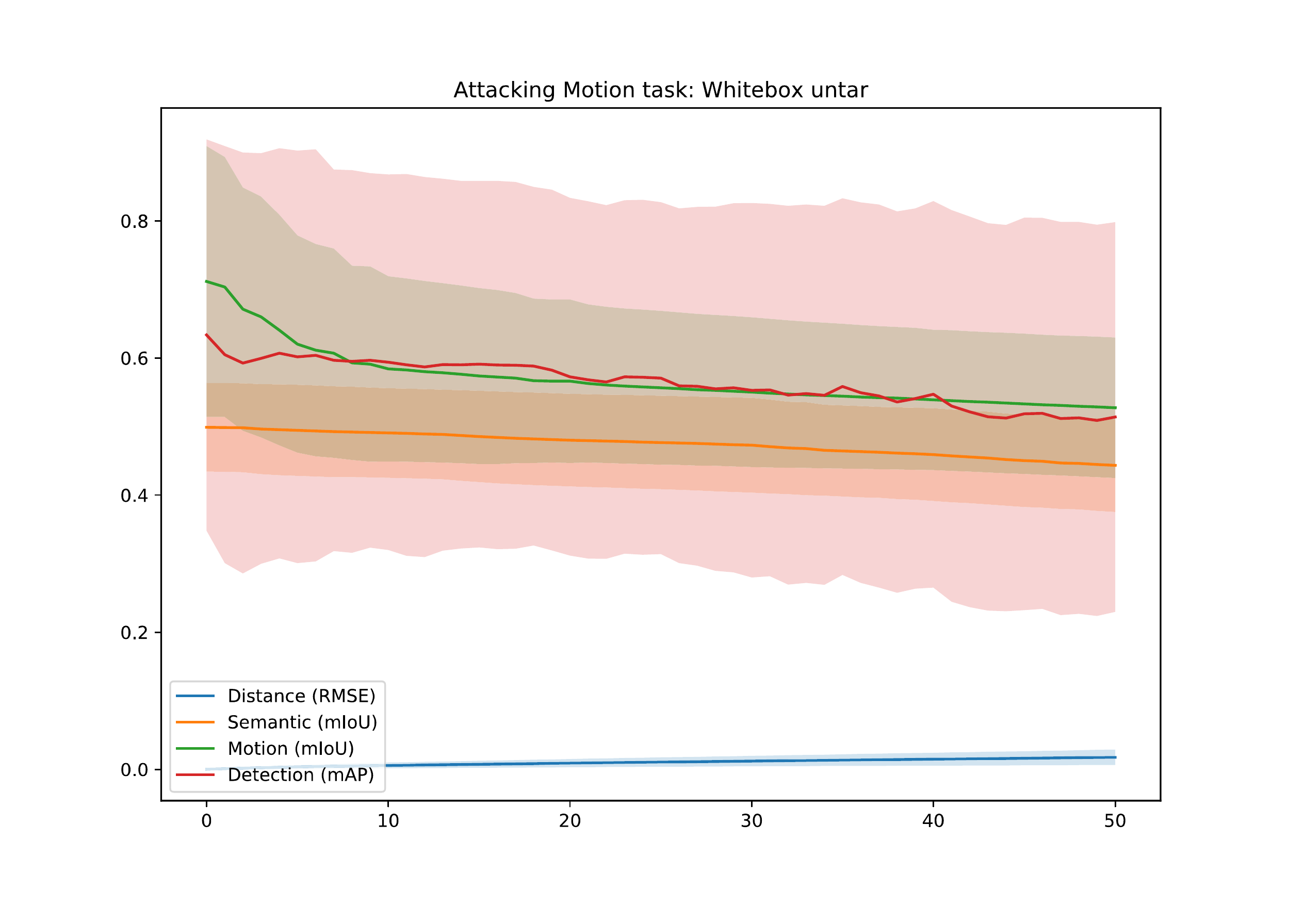}
    \includegraphics[width=0.49\textwidth, trim={2.35cm 1.4cm 2.55cm 1.6cm},clip]{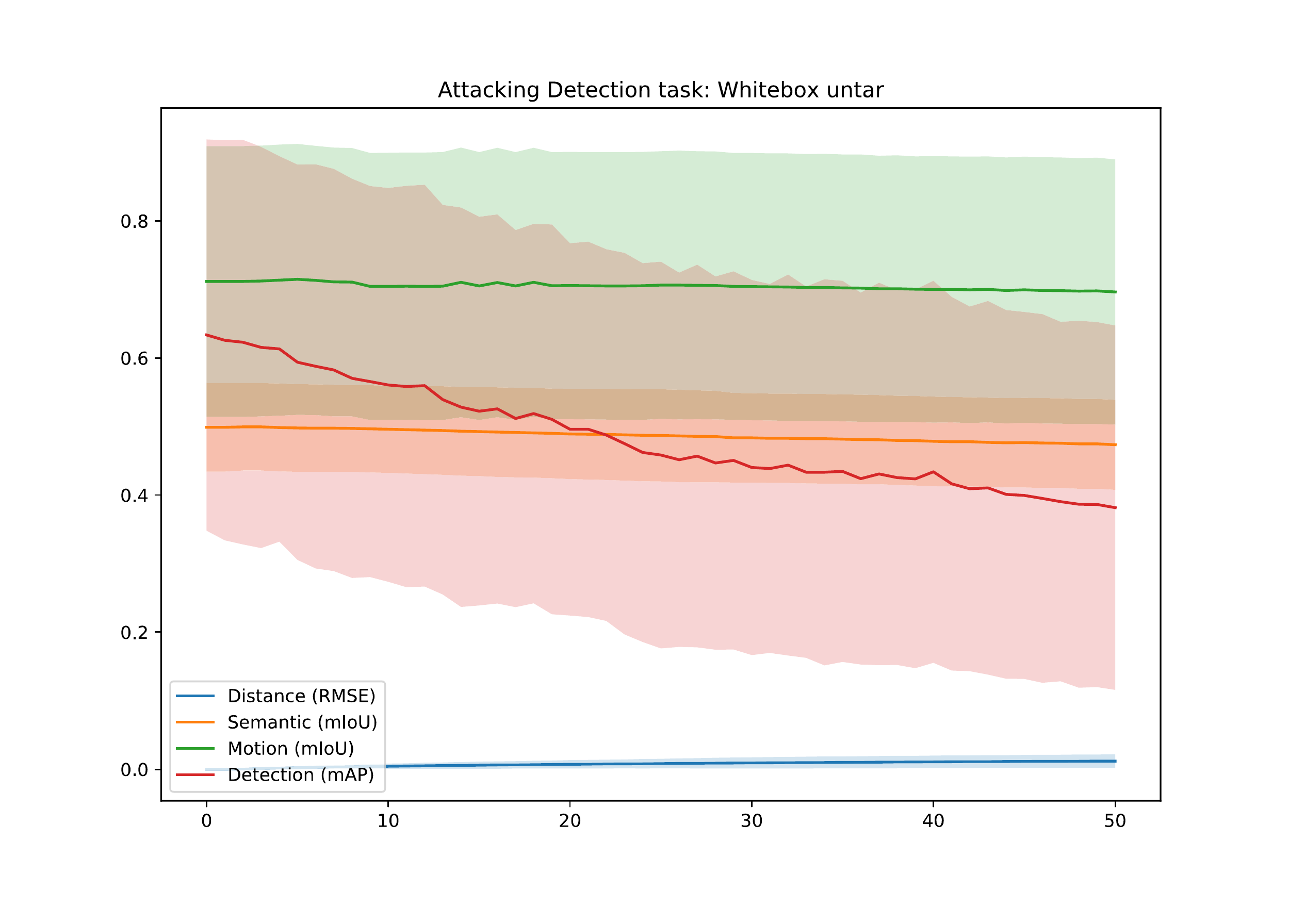}   
\\
\vspace{1mm}
    \includegraphics[width=0.49\textwidth, trim={2.35cm 1.4cm 2.55cm 1.6cm},clip]{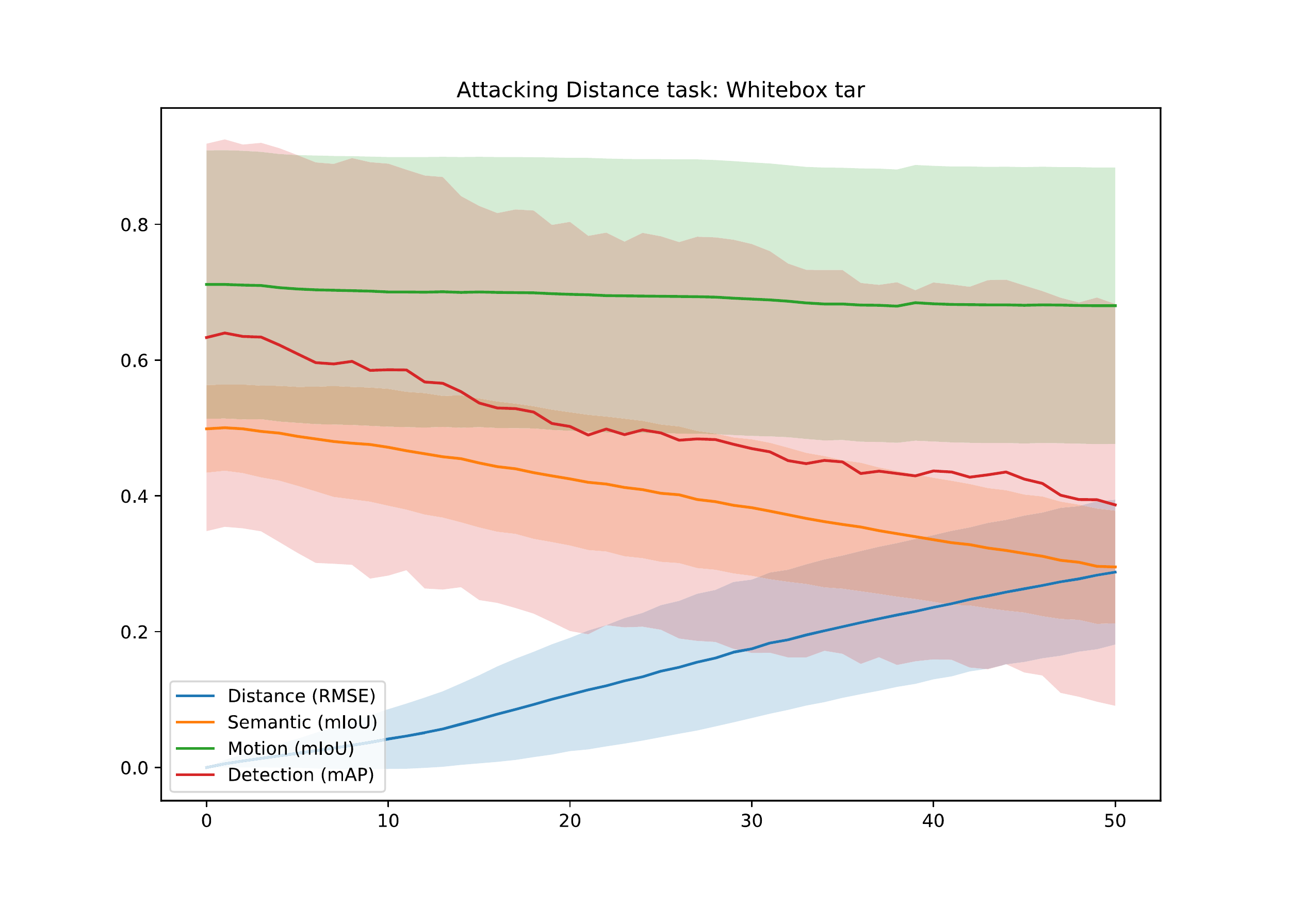}
    \includegraphics[width=0.49\textwidth, trim={2.35cm 1.4cm 2.55cm 1.6cm},clip]{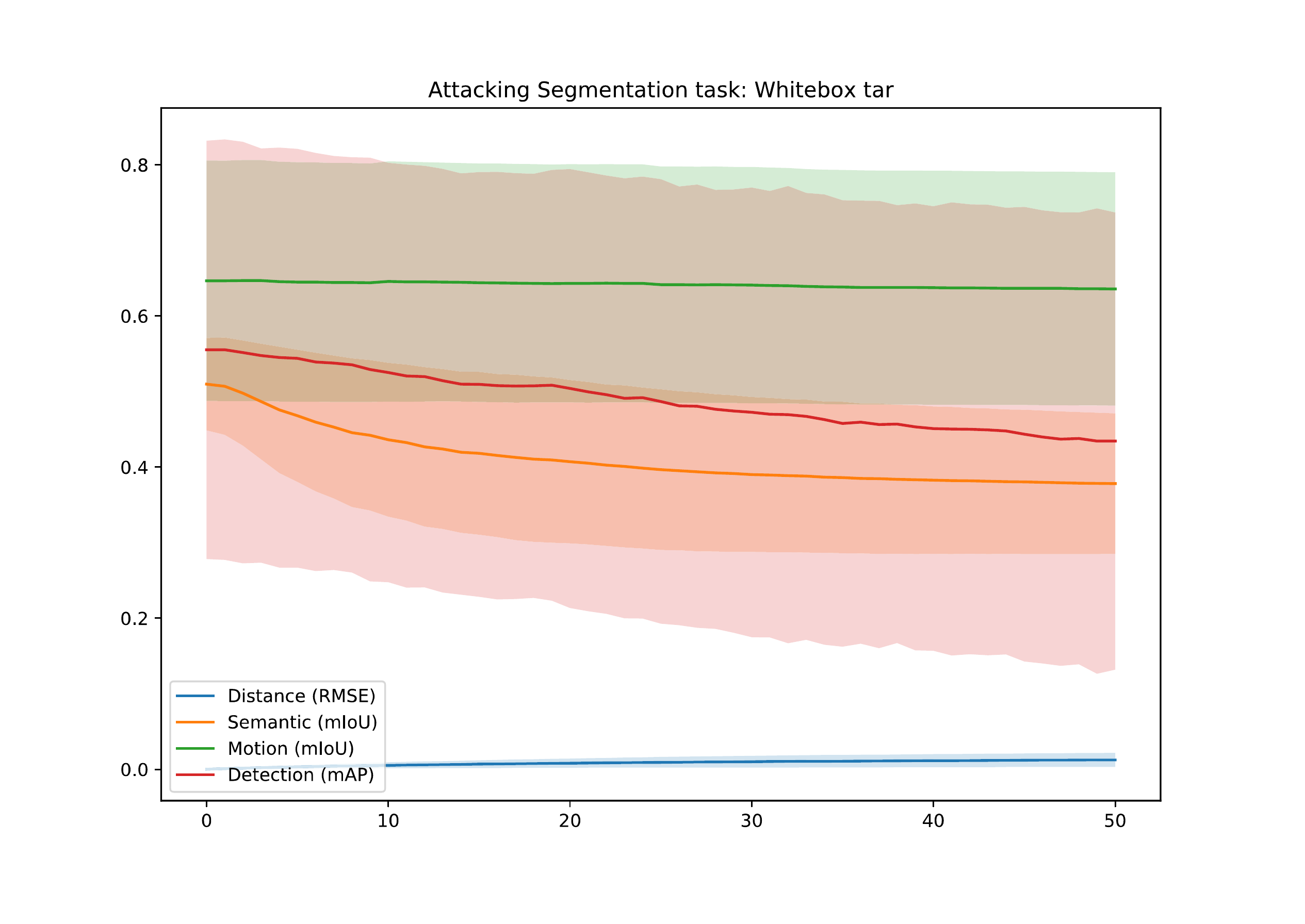}  
    \includegraphics[width=0.49\textwidth, trim={2.35cm 1.4cm 2.55cm 1.6cm},clip]{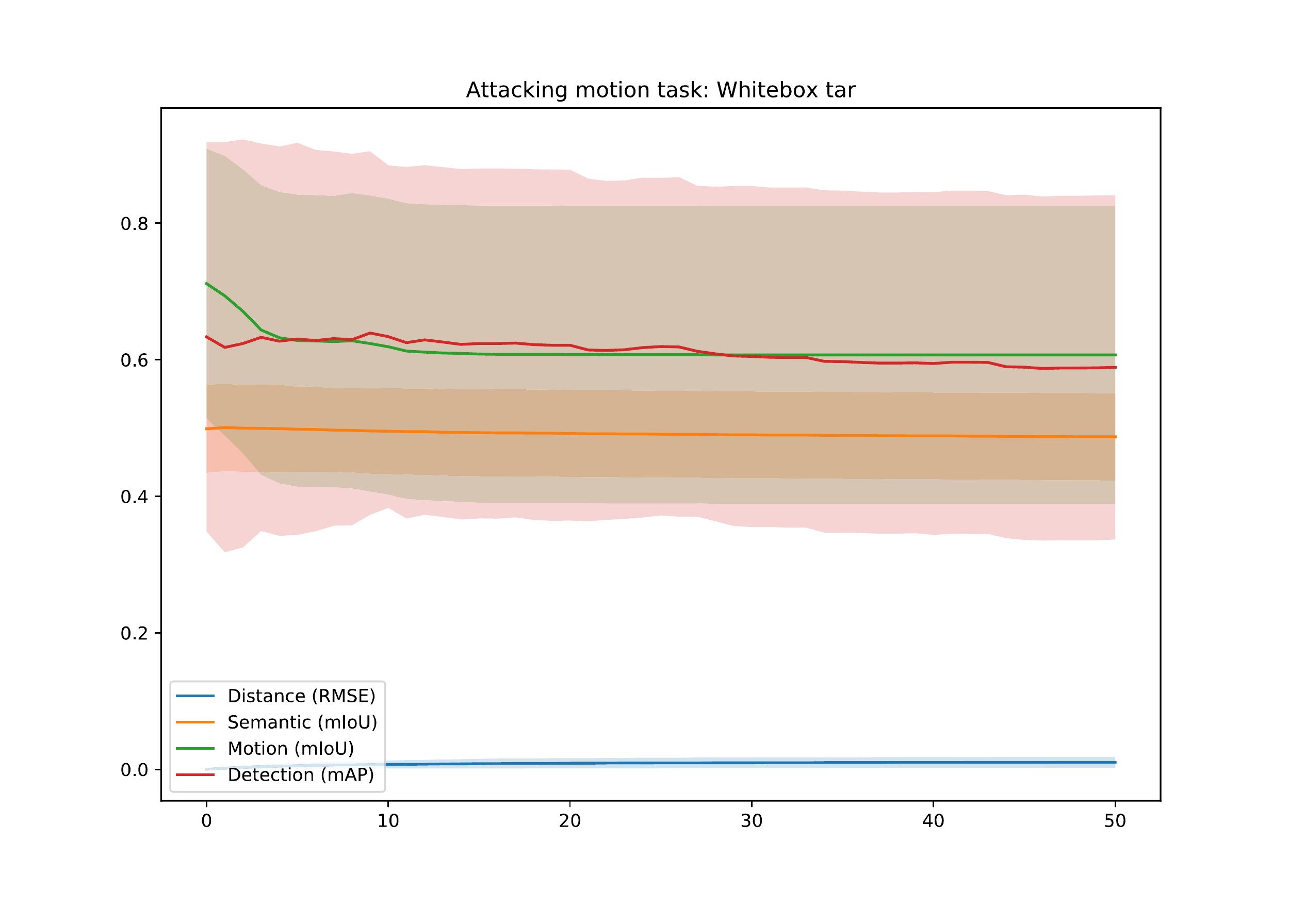}
    \includegraphics[width=0.49\textwidth, trim={2.35cm 1.4cm 2.55cm 1.6cm},clip]{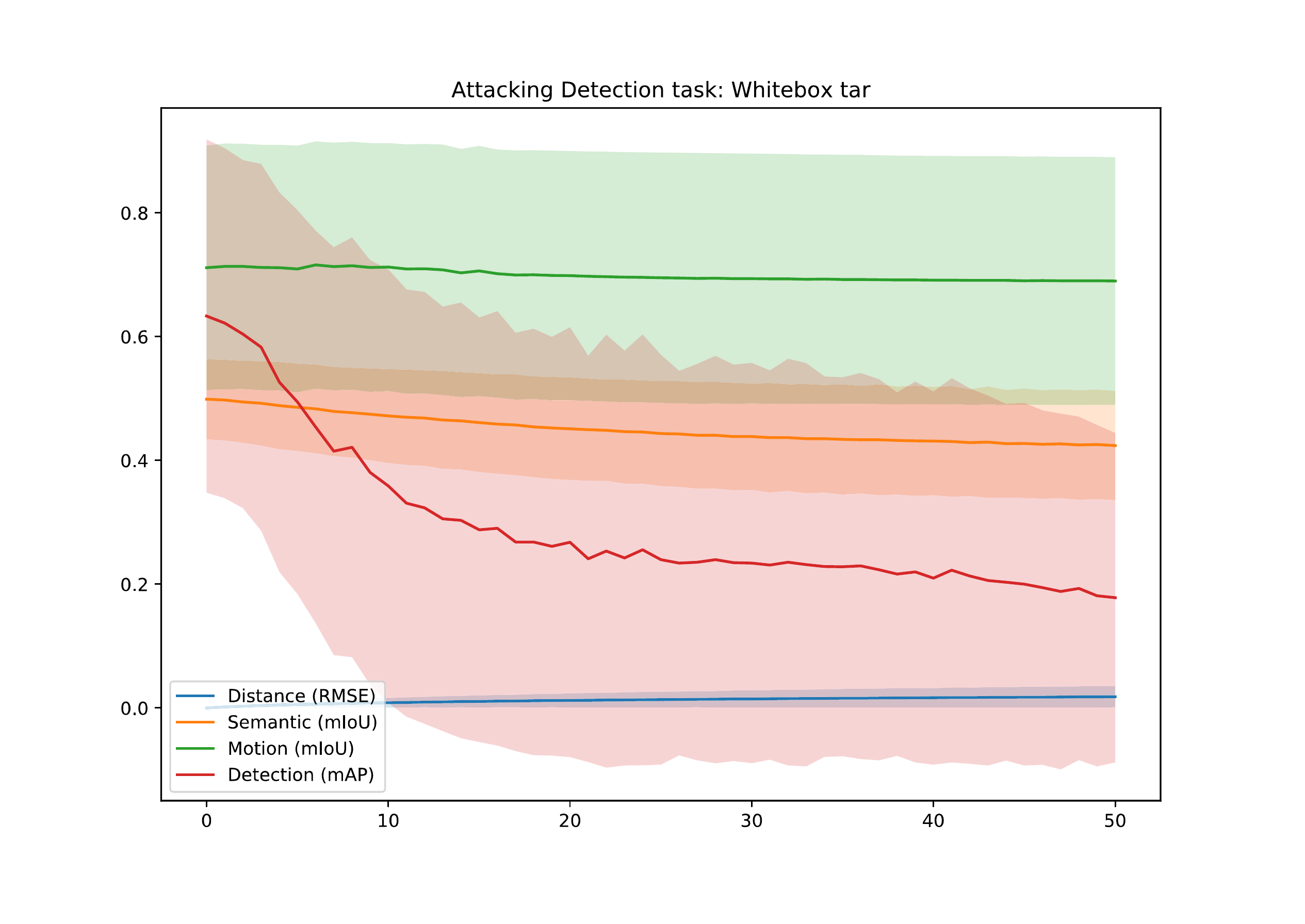}

    \caption[\bf Performance comparison of White-box attacks across different tasks]{Performance comparison of \textbf{White-box} attacks across different tasks. The 1\textsuperscript{st} and 2\textsuperscript{nd} rows show untargeted attacks, 3\textsuperscript{rd} and 4\textsuperscript{th} rows show targeted attacks, and columns represent the tasks.}
    \label{fig:whitebox}
\end{figure*}

\begin{figure*}[t]
    \centering
    \includegraphics[width=0.49\textwidth, trim={2.35cm 1.4cm 2.55cm 1.6cm},clip]{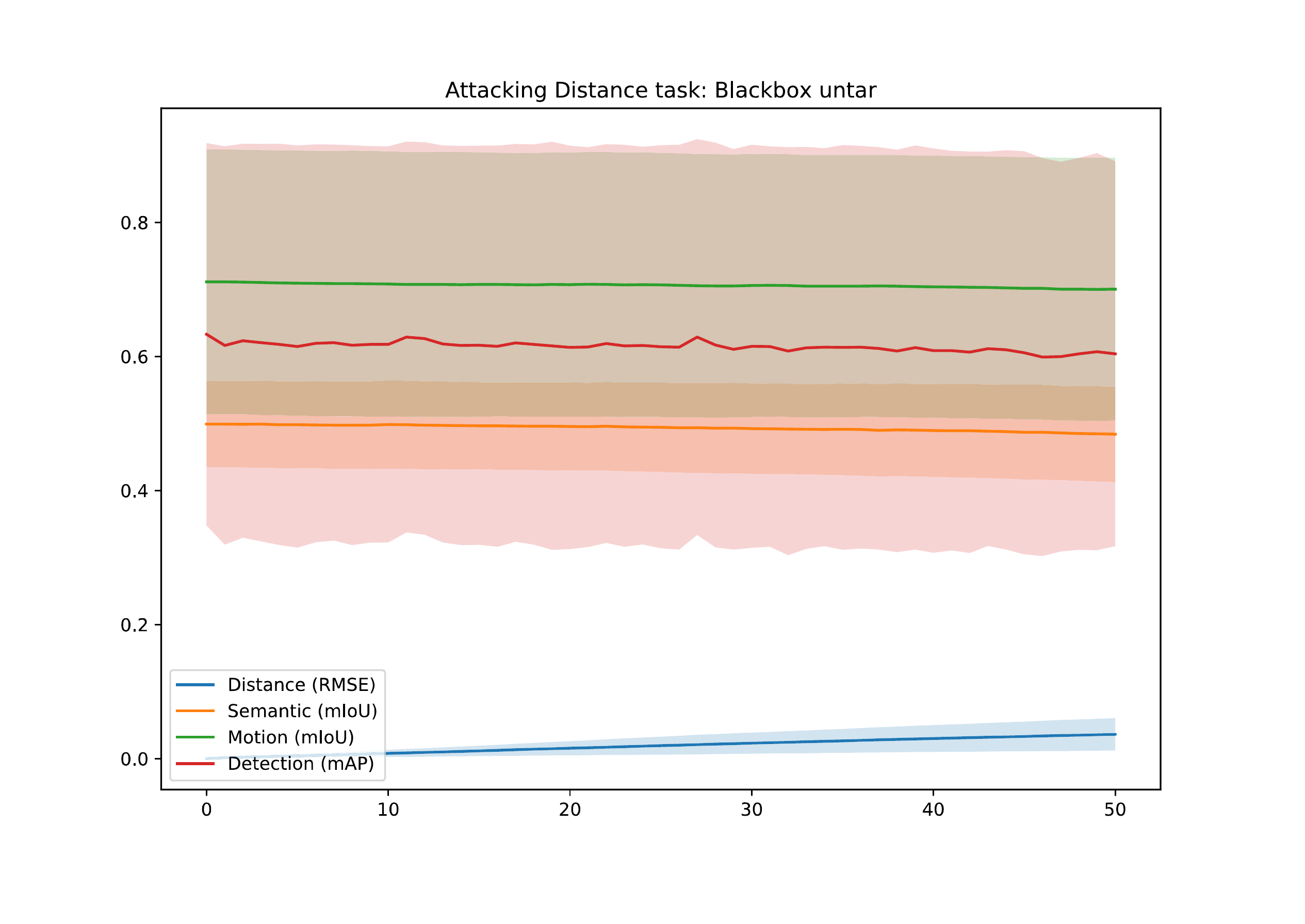}
    \includegraphics[width=0.49\textwidth, trim={2.35cm 1.4cm 2.55cm 1.6cm},clip]{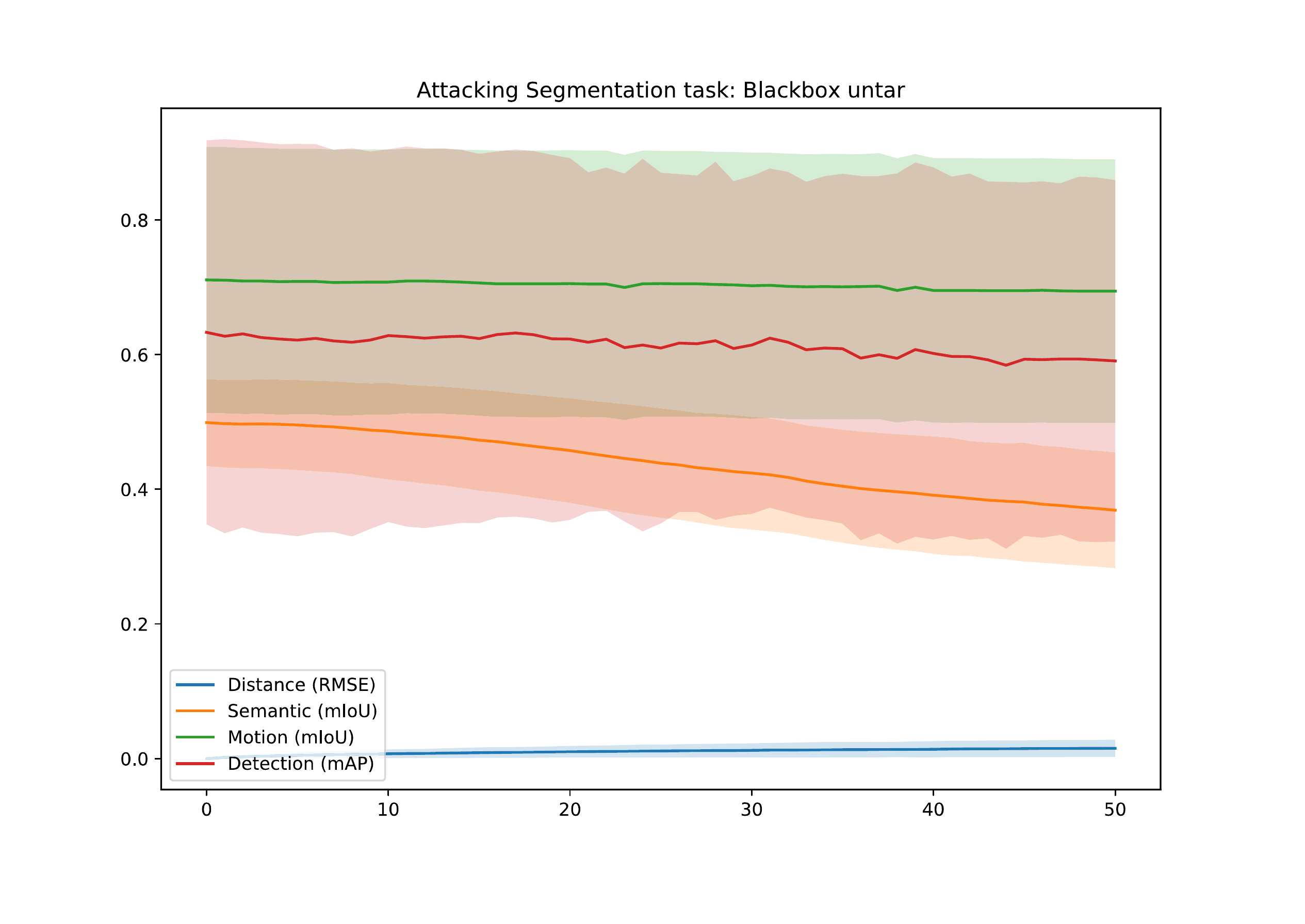}  
    \includegraphics[width=0.49\textwidth, trim={2.35cm 1.4cm 2.55cm 1.6cm},clip]{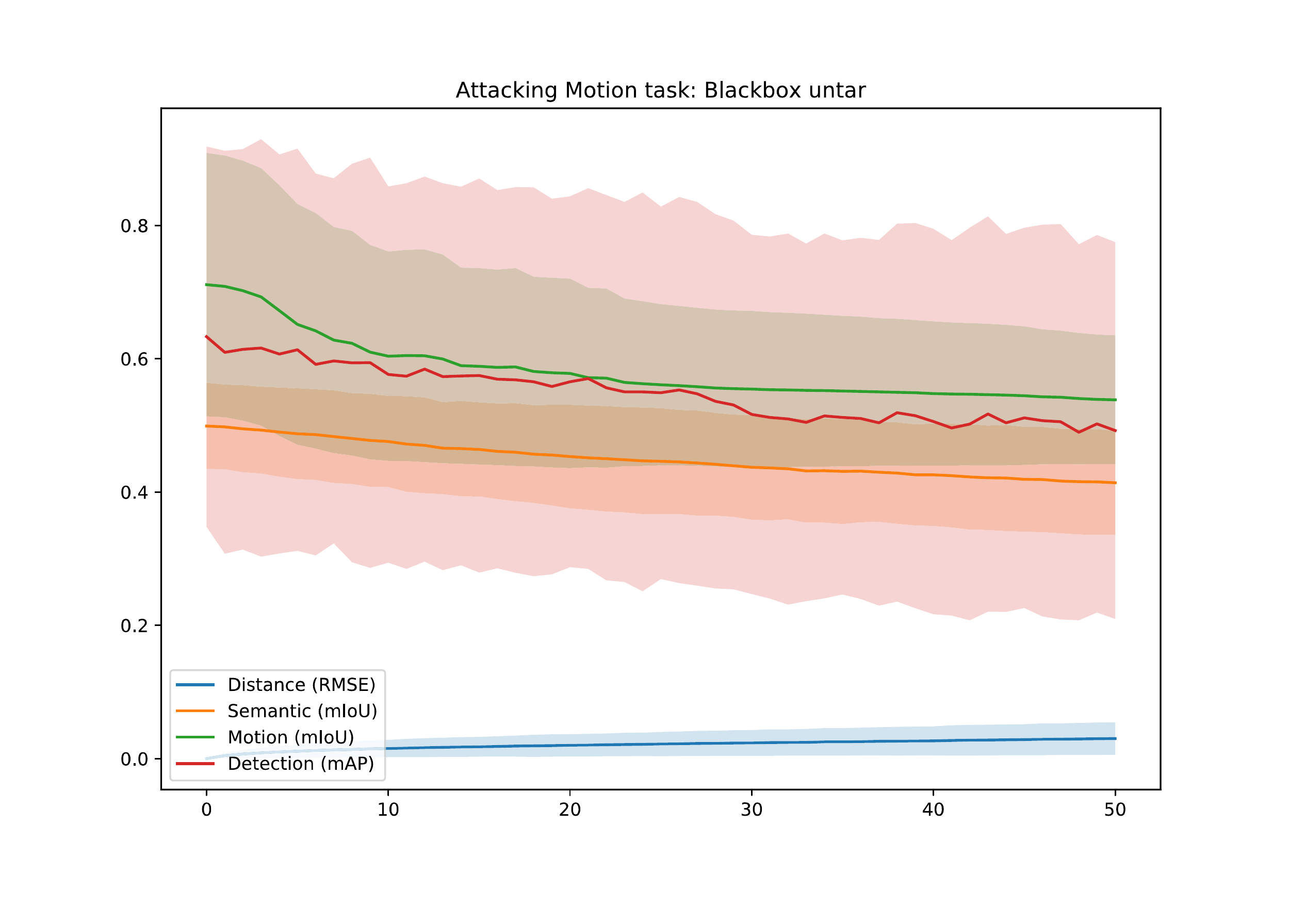}
    \includegraphics[width=0.49\textwidth, trim={2.35cm 1.4cm 2.55cm 1.6cm},clip]{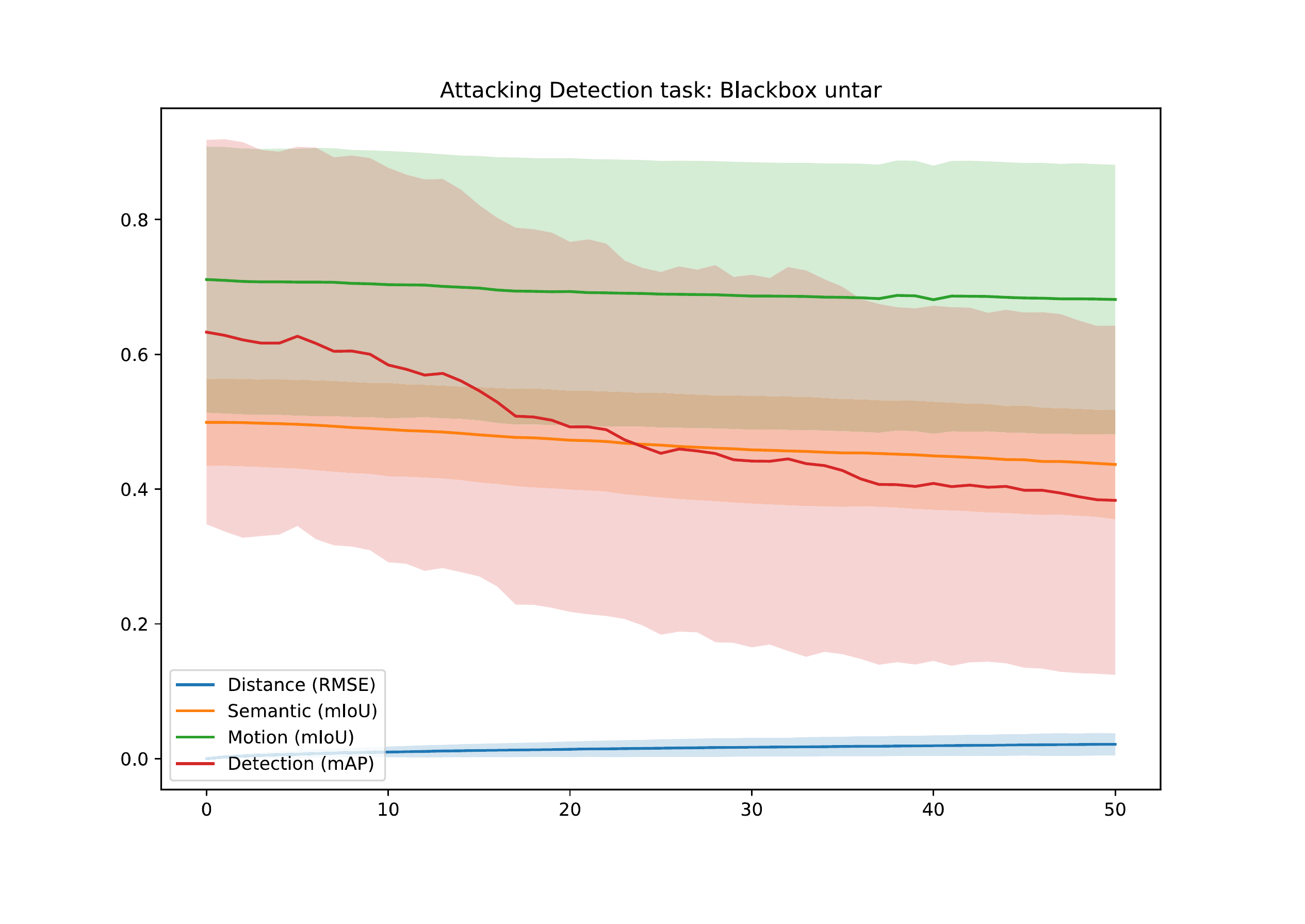}  
\\
\vspace{1mm}
    \includegraphics[width=0.49\textwidth, trim={2.35cm 1.4cm 2.55cm 1.6cm},clip]{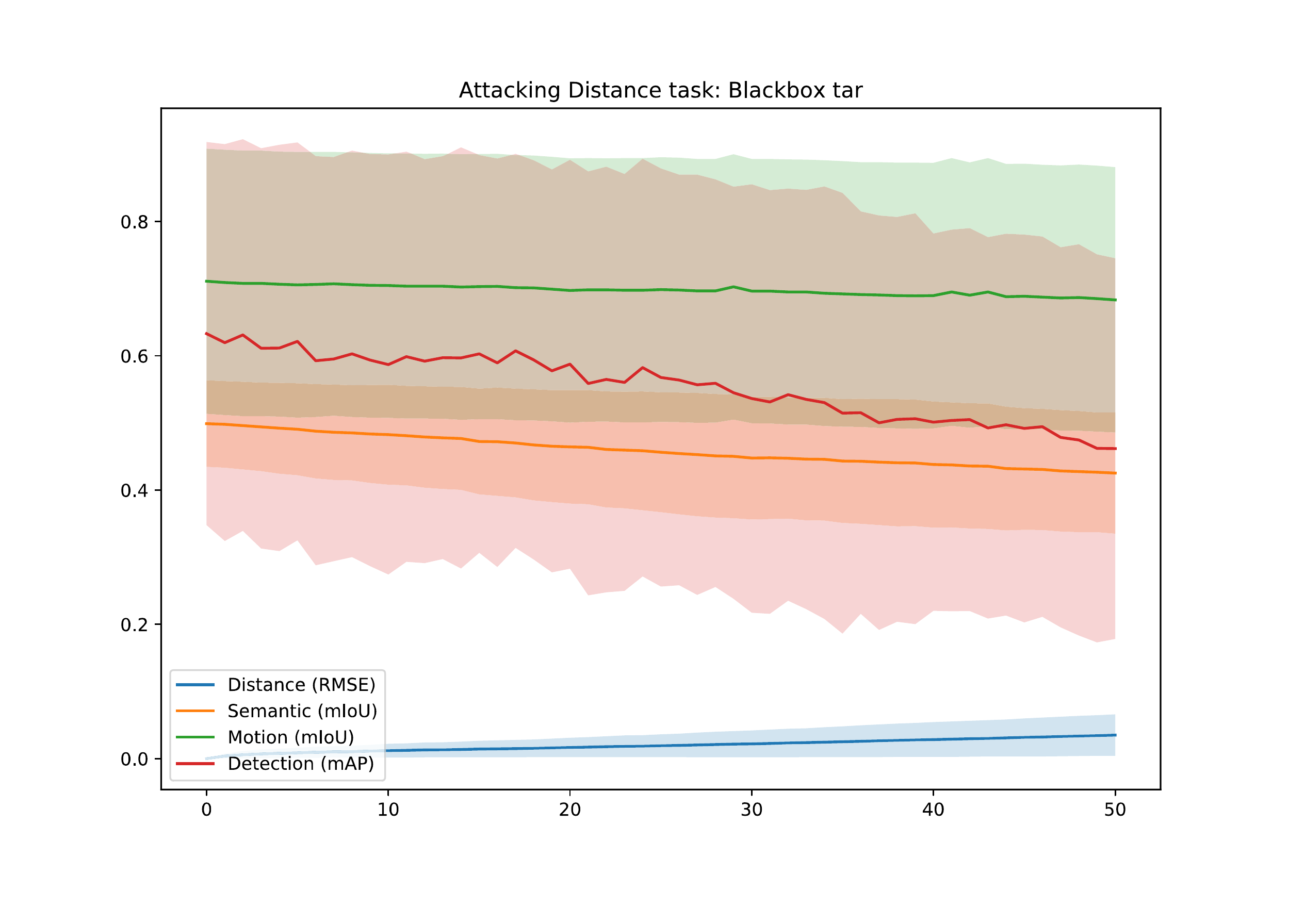}
    \includegraphics[width=0.49\textwidth, trim={2.35cm 1.4cm 2.55cm 1.6cm},clip]{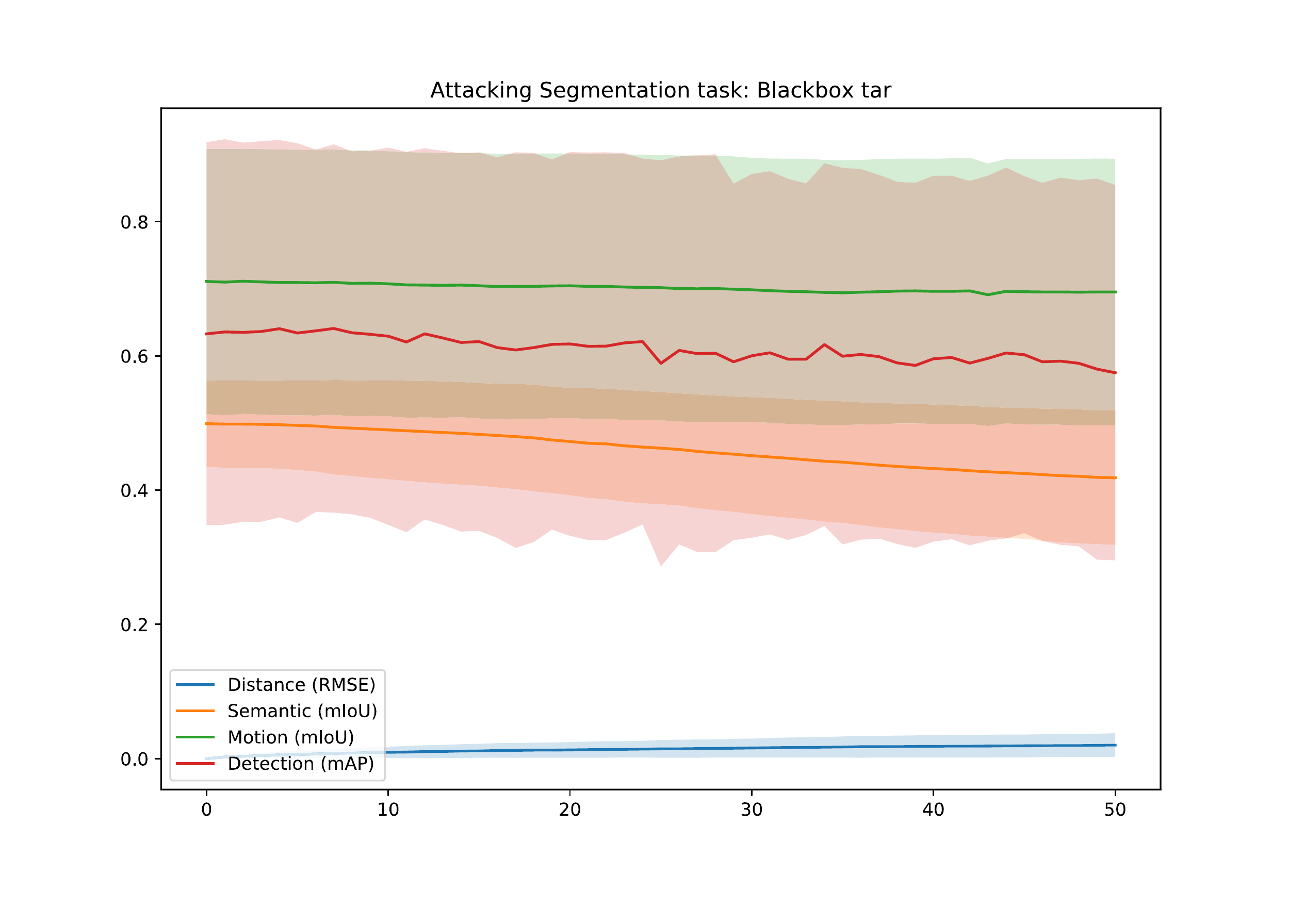}  
    \includegraphics[width=0.49\textwidth, trim={2.35cm 1.4cm 2.55cm 1.6cm},clip]{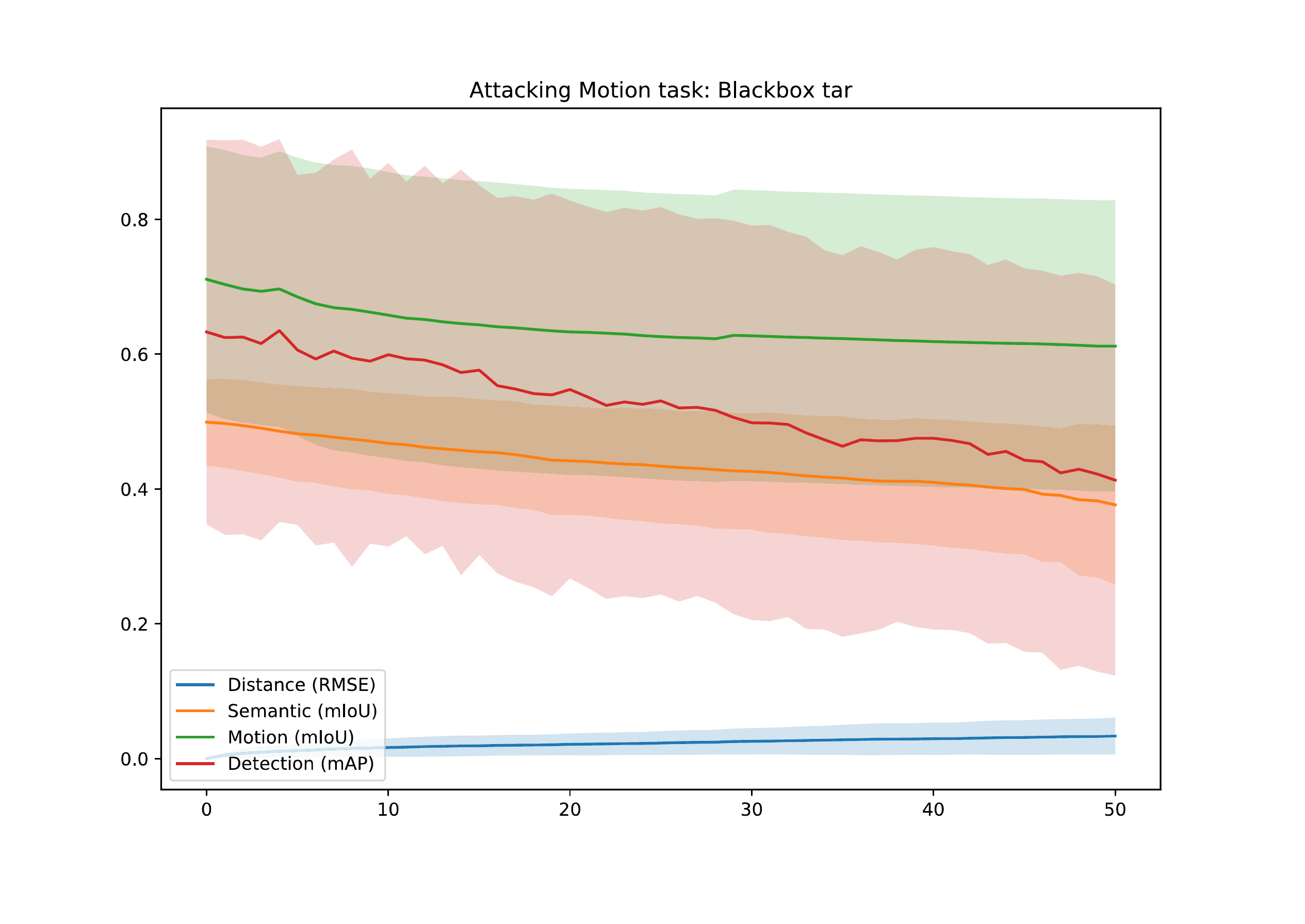}
    \includegraphics[width=0.49\textwidth, trim={2.35cm 1.4cm 2.55cm 1.6cm},clip]{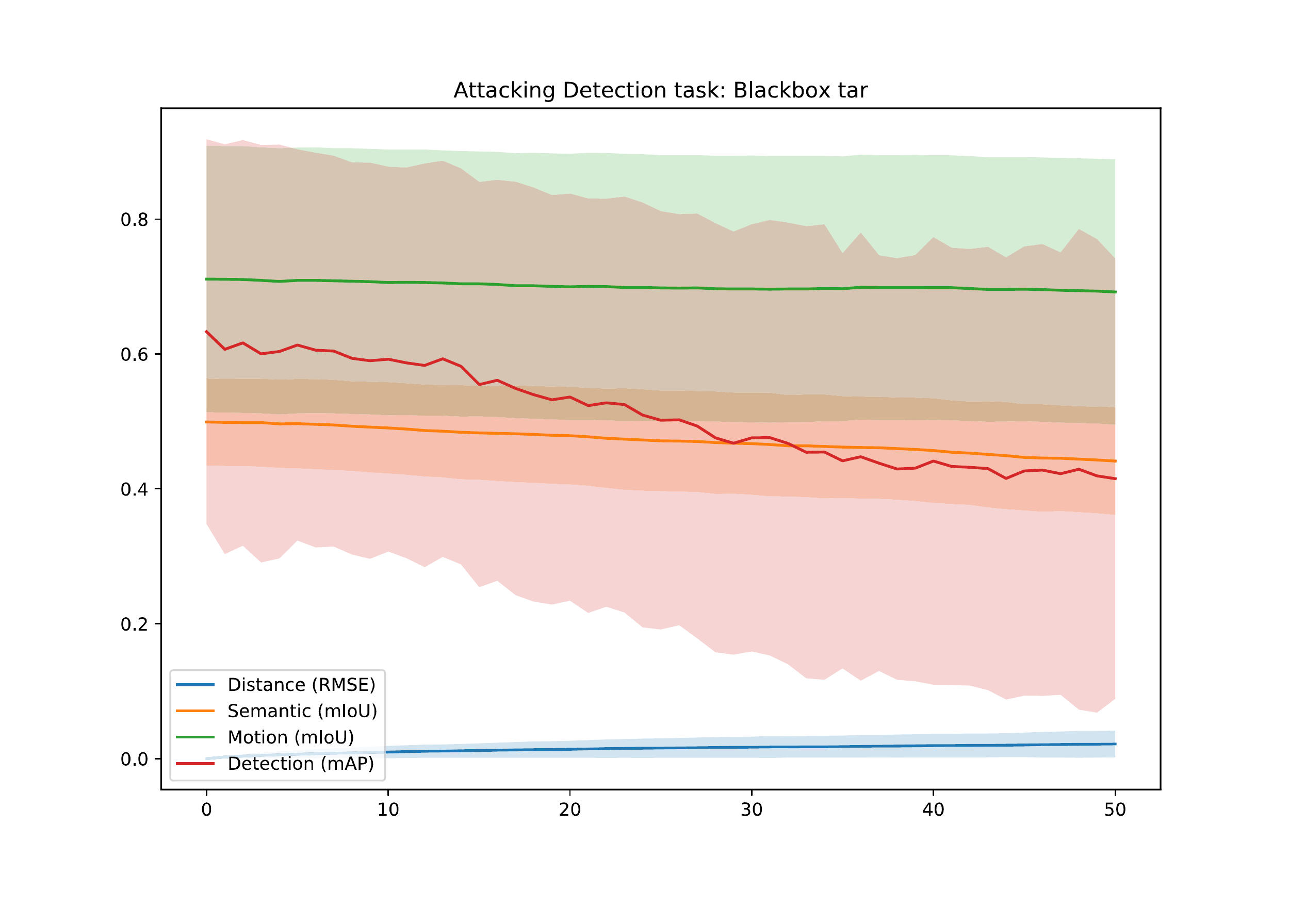}

    \caption[\bf Performance comparison of Black-box attacks across different tasks.]{Performance comparison of \textbf{Black-box} attacks across different tasks. The 1\textsuperscript{st} and 2\textsuperscript{nd} rows show untargeted attacks, 3\textsuperscript{rd} and 4\textsuperscript{th} rows show the targeted attacks, and columns represent the attacked tasks.}
    \label{fig:blackbox}
\end{figure*}

The attacking curves for white and black box attacks are shown in Figures~\ref{fig:whitebox} and~\ref{fig:blackbox} respectively. Each plot shows each perception task's performance over the 50 attacking steps where the first step at index $0$ represents the actual performance of the target network without applying any attack. Each curve shows the mean performance of a task over the test set, where the shaded area is the mean $\pm$ standard deviation. Generally, motion and detection tasks have a performance with a large standard deviation indicating the test set's diversity containing easy and hard examples. Across all curves, it is clear that the performance is decreasing along with the attacking steps.\par

Moreover, attacking one task by generating an adversarial example affects the other tasks' performance in different ways along the attacking curve. These curves enable the adversary to decide at which step the adversarial example is generated according to the required effect on the target task and the other tasks. As shown in Figures~\ref{fig:whitebox} and~\ref{fig:blackbox}, in most cases, attacking other tasks has a marginal negative effect on the motion task. The main reason is that the motion task takes two frames as input, and only one of them is attacked. Moreover, it is shown that attacking the distance task affects both segmentation and detection tasks. Attacking segmentation or detection showed to affect all other tasks. As mentioned, the attack effect depends on the parameters selected for the attack. Moreover, targeted attacks try to optimize the adversarial example to produce the required target prediction. In contrast, the untargeted attack continues to apply perturbations to produce as different as possible predictions.\par
\begin{figure*}[!ht]
  \resizebox{\textwidth}{!}{\newcommand{\turnheightnew}{0.25\linewidth}
\centering

\begin{tabular}{@{\hskip 0.2mm}c@{\hskip 0.2mm}c@{\hskip 0.2mm}c@{\hskip 0.2mm}c@{\hskip 0.2mm}}    

\includegraphics[height=\turnheightnew]{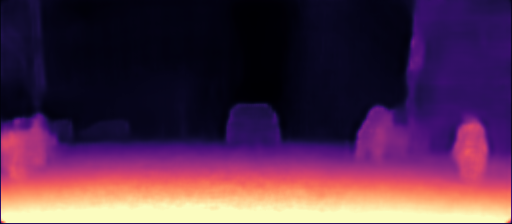}
\includegraphics[height=\turnheightnew]{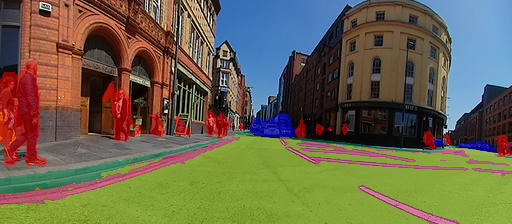}
\includegraphics[height=\turnheightnew]{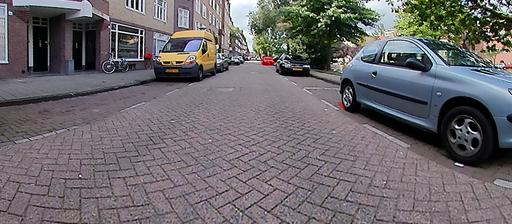}
\includegraphics[height=\turnheightnew]{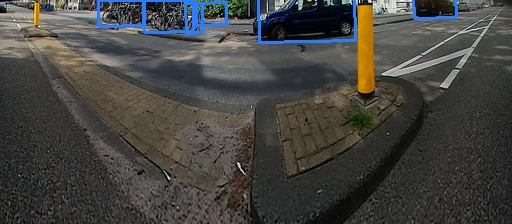} \\
    
\includegraphics[height=\turnheightnew]{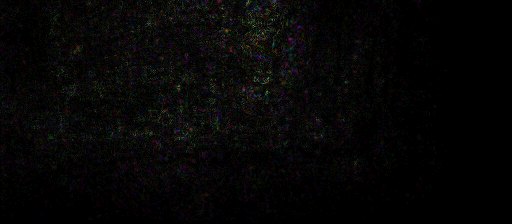}
\includegraphics[height=\turnheightnew]{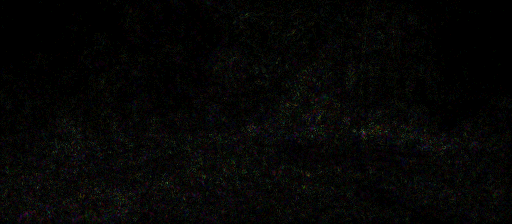}
\includegraphics[height=\turnheightnew]{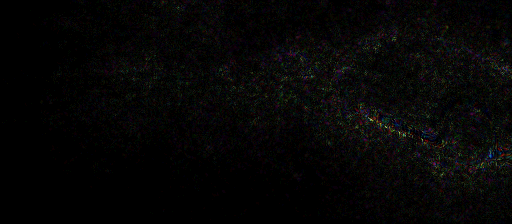}
\includegraphics[height=\turnheightnew]{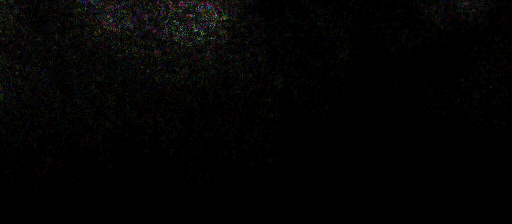} \\
    
\includegraphics[height=\turnheightnew]{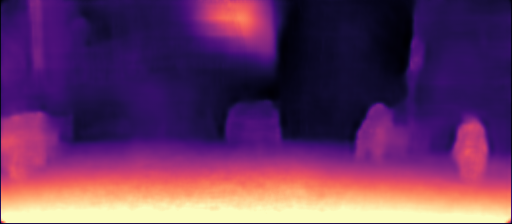}
\includegraphics[height=\turnheightnew]{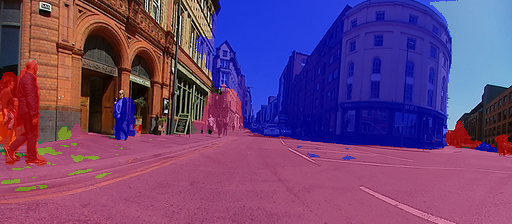}  
\includegraphics[height=\turnheightnew]{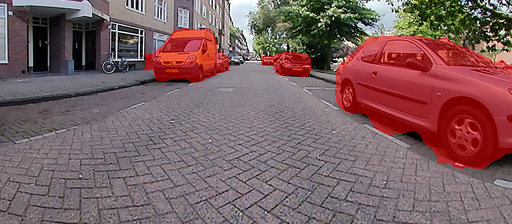}
\includegraphics[height=\turnheightnew, clip]{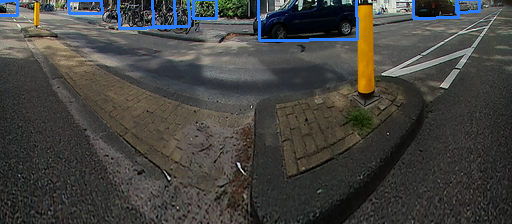} \\
    
\includegraphics[height=\turnheightnew]{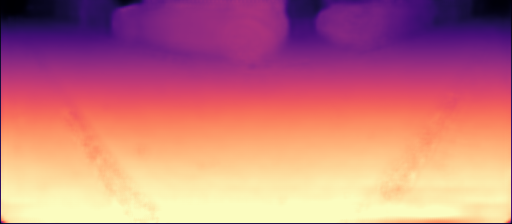}
\includegraphics[height=\turnheightnew]{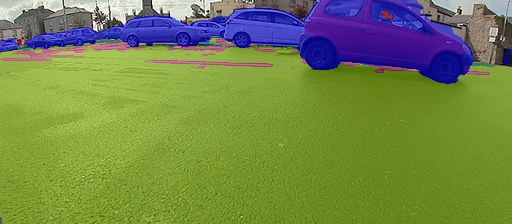}
\includegraphics[height=\turnheightnew]{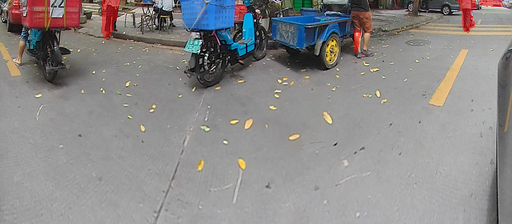}
\includegraphics[height=\turnheightnew]{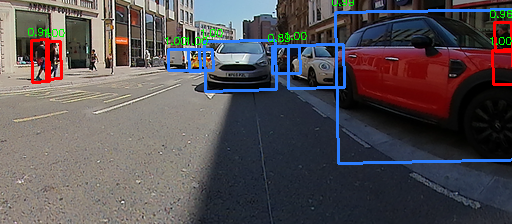} \\
    
\includegraphics[height=\turnheightnew]{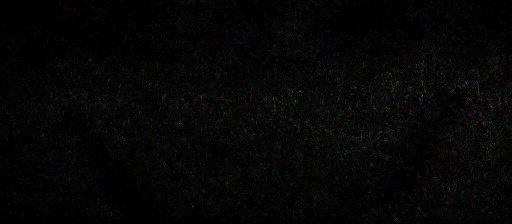}
\includegraphics[height=\turnheightnew]{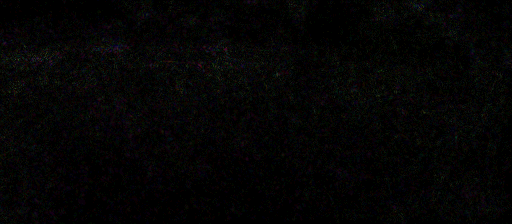}  
\includegraphics[height=\turnheightnew]{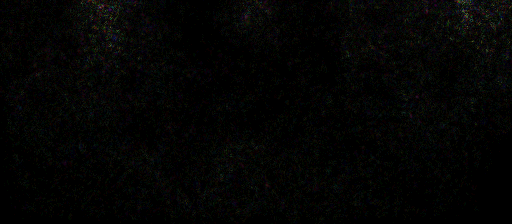}
\includegraphics[height=\turnheightnew]{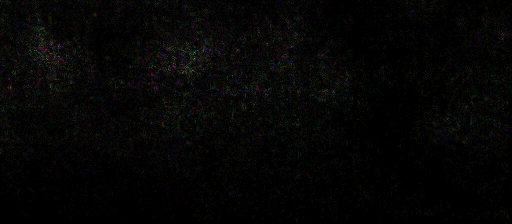} \\
    
\includegraphics[height=\turnheightnew]{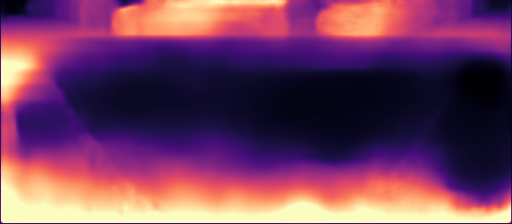}
\includegraphics[height=\turnheightnew,clip]{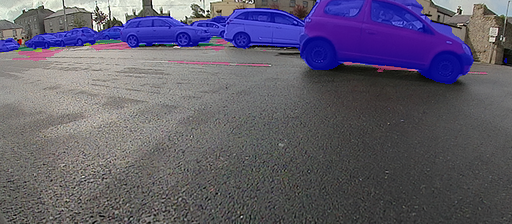}
\includegraphics[height=\turnheightnew]{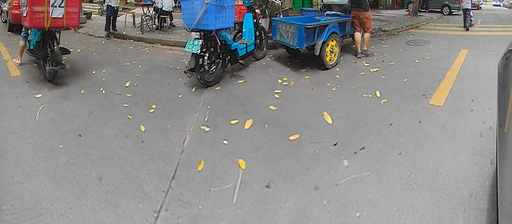}
\includegraphics[height=\turnheightnew]{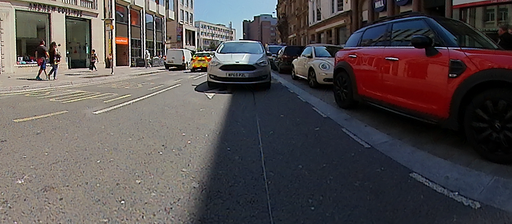} \\
    
\includegraphics[height=\turnheightnew]{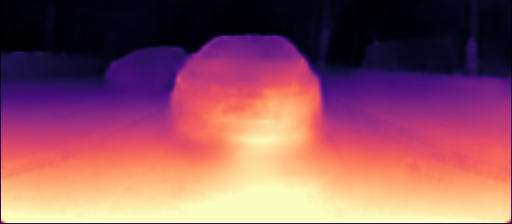}
\includegraphics[height=\turnheightnew]{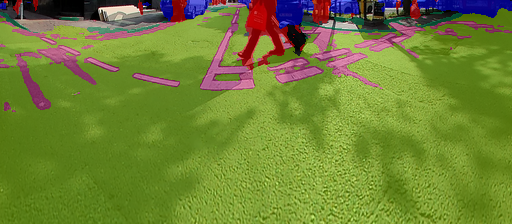}
\includegraphics[height=\turnheightnew]{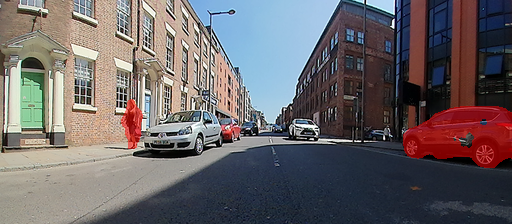}
\includegraphics[height=\turnheightnew]{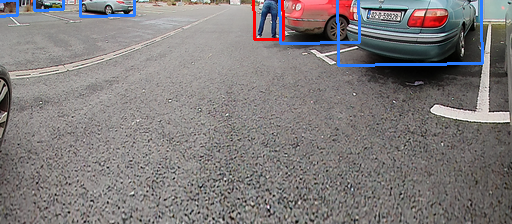} \\
    
\includegraphics[height=\turnheightnew]{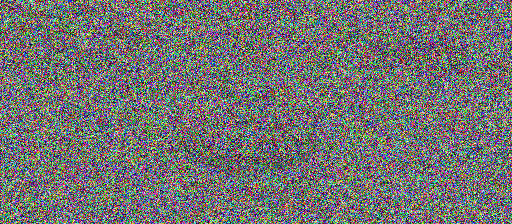}
\includegraphics[height=\turnheightnew]{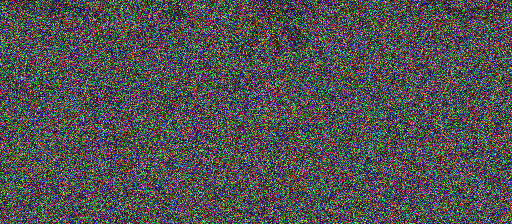}
\includegraphics[height=\turnheightnew]{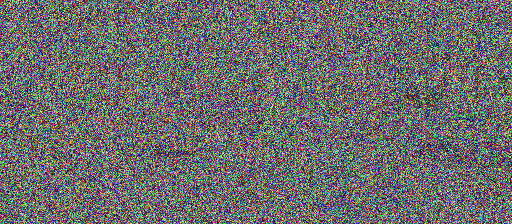}
\includegraphics[height=\turnheightnew]{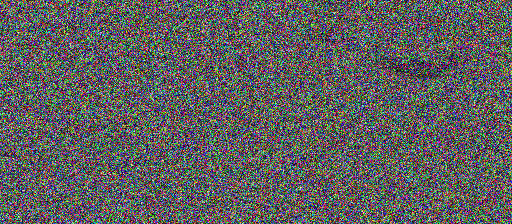} \\

\includegraphics[height=\turnheightnew]{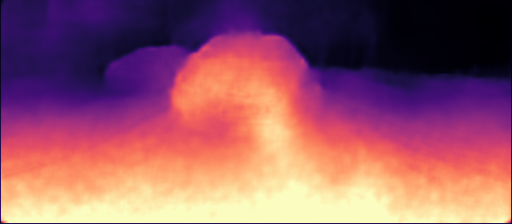}
\includegraphics[height=\turnheightnew,clip]{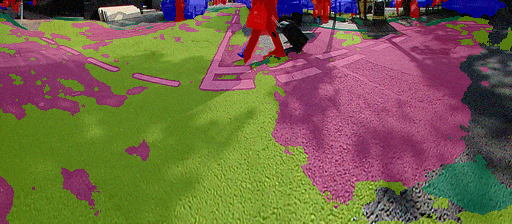}  
\includegraphics[height=\turnheightnew]{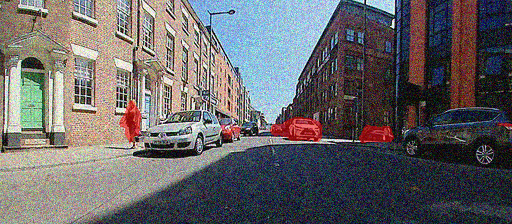}
\includegraphics[height=\turnheightnew]{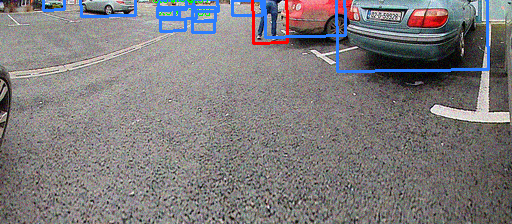} \\

\includegraphics[height=\turnheightnew]{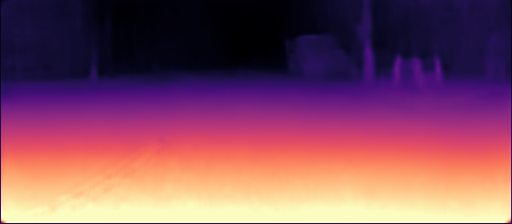}
\includegraphics[height=\turnheightnew]{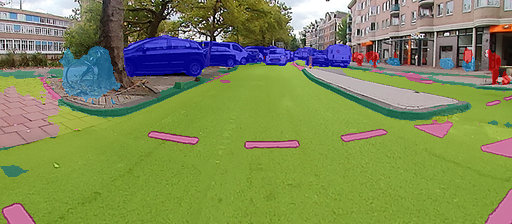}
\includegraphics[height=\turnheightnew]{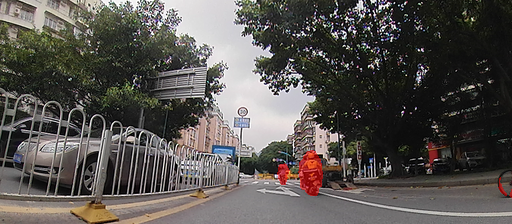}
\includegraphics[height=\turnheightnew]{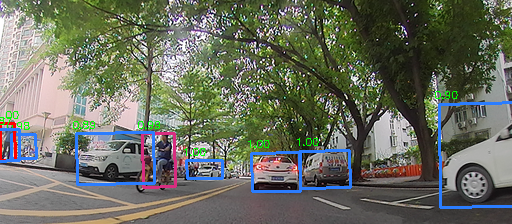} \\
    
\includegraphics[height=\turnheightnew]{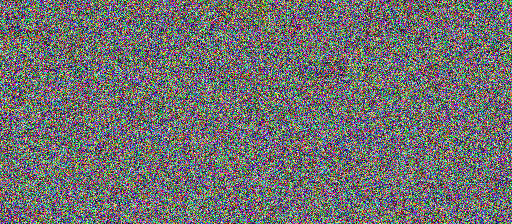}
\includegraphics[height=\turnheightnew]{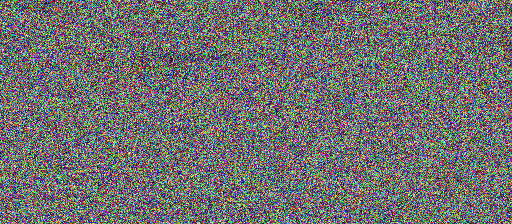}
\includegraphics[height=\turnheightnew]{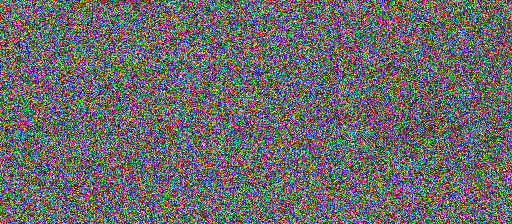}
\includegraphics[height=\turnheightnew]{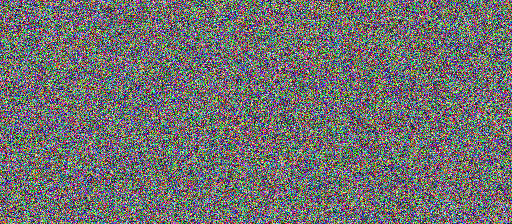} \\
    
\includegraphics[height=\turnheightnew]{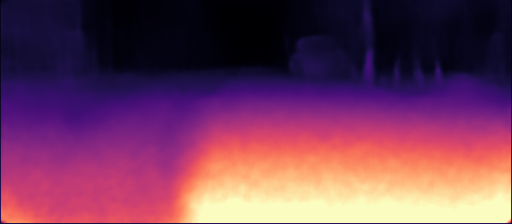}
\includegraphics[height=\turnheightnew]{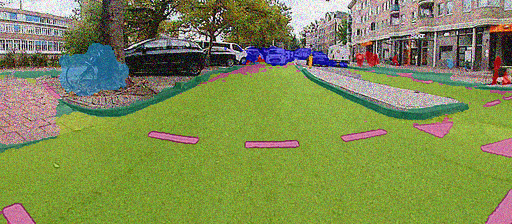}
\includegraphics[height=\turnheightnew]{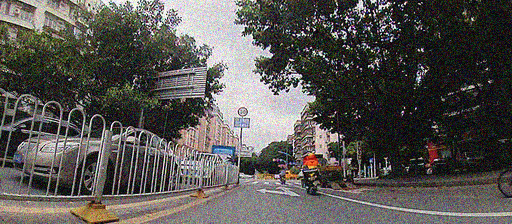}
\includegraphics[height=\turnheightnew]{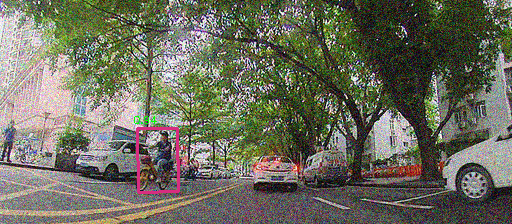}

\end{tabular}}
    \caption[\bf Qualitative illustration of White box Untargeted, White box Targeted, Black box Untargeted, and Black box Targeted Attacks.]{From Top to Bottom: \textbf{White box Untargeted, White box Targeted, Black box Untargeted, and Black box Targeted Attacks.} Within each group from top to bottom: Original results, adversarial perturbations, and the impacted results. For intuitive visualizations of the attacks, see {\url{https://youtu.be/R3JUV41aiPY}}.}
\label{fig:omnidet-qual_attacks}
\end{figure*}

\begin{table}[t]
\centering
\begin{adjustbox}{width=\columnwidth}
\small
\setlength{\tabcolsep}{0.3em}
\begin{tabular}{l|l|ll|ll|ll|ll}
\toprule
\textbf{Task} &  
& \multicolumn{2}{c}{\cellcolor[HTML]{00b0f0} \textit{
\begin{tabular}[c]{@{}c@{}} Distance\\ RMSE\end{tabular}}} 
& \multicolumn{2}{c}{\cellcolor[HTML]{00b050} \textit{
\begin{tabular}[c]{@{}c@{}} Segmentation\\ mIoU\end{tabular}}} 
& \multicolumn{2}{c}{\cellcolor[HTML]{ab9ac0} \textit{
\begin{tabular}[c]{@{}c@{}} Motion\\ mIoU\end{tabular}}}
& \multicolumn{2}{c}{\cellcolor[HTML]{a5a5a5} \textit{
\begin{tabular}[c]{@{}c@{}} Detection\\ mAP\end{tabular}}} \\ 
\midrule
& \multicolumn{1}{l|}{} 
& \multicolumn{1}{c|}{\cellcolor[HTML]{7d9ebf} A}
& \multicolumn{1}{c|}{\cellcolor[HTML]{e8715b} D}
& \multicolumn{1}{l|}{\cellcolor[HTML]{7d9ebf} A(\%)}
& \multicolumn{1}{l|}{\cellcolor[HTML]{e8715b} D(\%)}
& \multicolumn{1}{l|}{\cellcolor[HTML]{7d9ebf} A(\%)}
& \multicolumn{1}{l|}{\cellcolor[HTML]{e8715b} D(\%)} 
& \multicolumn{1}{l|}{\cellcolor[HTML]{7d9ebf} A(\%)}
& \multicolumn{1}{l|}{\cellcolor[HTML]{e8715b} D(\%)} \\ 
\cmidrule(l){3-10}
\multirow{4}{*}{\textit{Distance}} 
& \textit{wb\_untarget} & 0.126 & 0.047 & -14 & -7.3 & -3.1 & -3.0 & -13.4 & -25.5 \\
& \textit{wb\_target} & 0.288 & 0.031 & -40 & -7.5 & -4.4 & -2.7 & -38.9 & -33.2 \\
& \textit{bb\_untarget} & 0.036 & 0.033 & -3.0 & -6.5 & -1.5 & -3.7 & -4.6 & -25.8 \\
& \textit{bb\_target} & 0.035 & 0.036 & -14.8 & -13.4 & -3.9 & -2.9 & -27.1 & -37.6 \\
\midrule
\multirow{4}{*}{\textit{Segmentation}} 
& \textit{wb\_untarget} & 0.032 & 0.028 & -86.8 & -14.2 & -5.0 & -3.5 & -37.6 & -30.1 \\
& \textit{wb\_target} & 0.017 & 0.027 & -32.0 & -5.5 & -4.0 & -2.6 & -21.3 & -27.5 \\
& \textit{bb\_untarget} & 0.015 & 0.031 & -26.1 & -11.4 & -2.3 & -4.9 & -6.7 & -27.3 \\
& \textit{bb\_target} & 0.020 & 0.034 & -16.1 & -9.2 & -2.2 & -2.5 & 9.2 &  -28.8 \\
\midrule
\multirow{4}{*}{\textit{Motion}} 
& \textit{wb\_untarget} & 0.018 & 0.027 & -11.1 & -7.3 & -25.9 & -9.2 & -18.9 & -23.1 \\
& \textit{wb\_target}   & 0.010 & 0.027 & -2.4 & -6.0 & -14.7 & -9.2 & -7.1 & -30.0 \\
& \textit{bb\_untarget} & 0.030 & 0.039 & -17.1 & -15.8 & -24.3 & -17.6 & -22.2 & -38.6 \\
& \textit{bb\_target}   & 0.033 & 0.040 & -24.6 & -22.5 & -13.9 & -11.7 & -34.7 & -47.9 \\
\midrule
\multirow{4}{*}{\textit{Detection}} 
& \textit{wb\_untarget} & 0.012 & 0.027 & -5.1 & -5.9 & -2.1 & -2.5 & -39.8 & -31.4 \\
& \textit{wb\_target} & 0.018 & 0.027 & -15.0 & -6.2 & -3.0 & -4.0 & -71.9 & -30.6 \\
& \textit{bb\_untarget} & 0.021 & 0.033 & -12.5 & -10.4 & -4.2 & -5.8 & -39.4 & -35.3 \\
& \textit{bb\_target} & 0.022 & 0.034 & -11.6 & -10.6 & -2.7 &
-3.7 & -34.4 & -37.9 \\ 
\bottomrule
\end{tabular}
\end{adjustbox}
\caption[\bf Summary of attacking and defending results.]{\textbf{Summary of attacking and defending results} across the test data where \textit{A} and \textit{D} columns are for Attack and Defense respectively.}
\label{tab:table-attackdefense1}
\end{table}
\begin{table}[!ht]
\begin{center}
\begin{tabular}{@{}l|l|c|c|c@{}}
\toprule
\textbf{Task } 
& \textit{\cellcolor[HTML]{7d9ebf} Metric} 
& \multicolumn{1}{l|}{\textit{\cellcolor[HTML]{7d9ebf} Original}}
& \multicolumn{1}{l|}{\textit{\cellcolor[HTML]{e8715b} Blurred}}
& \multicolumn{1}{l}{\textit{\cellcolor[HTML]{e8715b} Effect (\%)}} \\ 
\midrule
\textit{Distance}     & RMSE & 0.0 & 0.026   & NA \\ 
\textit{Segmentation} & mIoU & 0.499 & 0.477 &  -4.4 \\ 
\textit{Motion}       & mIoU & 0.711 & 0.693 & -2.6 \\ 
\textit{Detection}    & mAP  & 0.633 & 0.416 & -34.3 \\ 
\bottomrule
\end{tabular}
\caption{\textbf{Input blurring effect on the tasks.}}
\label{tab:table-blur}
\end{center}
\end{table}
To understand the effect of applying a defense method on the attacks, Gaussian blurring with a radius = $1$ is applied to the final adversarial examples, which are then fed into the target network. As shown in Table~\ref{tab:table-attackdefense1}, this simple defense method has a positive effect for both segmentation and motion tasks in most cases compared to depth and detection tasks. Furthermore, the effect of blurring on the network's performance is inspected without applying any attacks, as shown in Table~\ref{tab:table-blur}. Both detection and distance tasks are affected the most. This explains why this defense method is more effective for segmentation and motion tasks. Figure~\ref{fig:omnidet-qual_attacks} shows different visual samples of the attacks organized into four groups. Each group has three images: the original output, the adversarial perturbations magnified to 10X, and the impacted results are overplayed on the adversarial examples. As expected, perturbations for the white box attacks are much more minor and more localized than the black box case. Moreover, the performance is harmed for the untargeted attacks without having a specific goal leading to arbitrary predictions. On the other hand, vehicles or roads are removed for the semantic segmentation task for targeted attacks. We add false objects or remove true objects for the detection task. Near pixels are converted as far for the distance task. Finally, we convert dynamic objects to static for the motion task.\par
\section{Conclusion}

This chapter successfully demonstrated a six-task network with a shared encoder and synergized decoders on fisheye surround-view images in this thesis. The majority of the automated driving community's research continues to focus on individual tasks, and there is much progress to be made in designing and training optimal multi-task models. We introduced several novel contributions, including the camera geometry tensor usage for encoding radial distortion, variance-based normalization task weighting, and generalized object detection representations. To enable a comparison, we evaluate the network on five tasks on KITTI and Cityscapes, achieving competitive results. There are still many practical challenges in scaling to a higher number of tasks: building a diverse and balanced dataset, corner case mining, stable training mechanisms, and designing an optimal map representation that combines all tasks, thereby reducing the post-processing. We hope that this thesis encourages further research in building a unified perception model for autonomous driving.\par

This chapter also applied various adversarial attacks to understand the vulnerabilities of the perception network. For each perception task, white and black box attacks are conducted for targeted and untargeted scenarios. Moreover, the attacking curves show the interactions between attacks on different tasks. It is shown how attacking a task affects not only that task but also the others. Moreover, by applying blurring on the adversarial examples as a defense method, it is found to positively affect segmentation and motion tasks in contrast to the object detection and distance estimation tasks for the considered network model.\par
\chapter{Discussion}
\label{Chapter8}
\minitoc

This chapter discusses the contributions and the limitations of the approaches presented in this thesis.
\section{Geometric Tasks}

Depth estimation models may be learned in a supervised fashion on LiDAR distance measurements, such as KITTI~\cite{geiger2013vision}. In previous work, we followed this approach and demonstrated the possibility to estimate high-quality distance maps using LiDAR ground truth on fisheye images~\cite{kumar2018monocular}. However, the limitation of this approach is that setting up the entire rig for such recordings is expensive and time-consuming, limiting the amount of data on which a model can be trained. Supervised learning remains a bottleneck for creating more intelligent generalized models which can do multiple tasks and acquire new skills without massive amounts of labeled data. To overcome this problem, we proposed {\em FisheyeDistanceNet} in \textbf{Chapter~\ref{Chapter4}}, the first end-to-end self-supervised monocular scale-aware training framework. \textit{FisheyeDistanceNet} uses CNNs on raw fisheye image sequences to regress a Euclidean distance map and provides a baseline for single frame Euclidean distance estimation.\par

Practically speaking, if we consider the supervised learning approach, it is improbable to label everything in the world. Suppose AI systems can gather a more profound, more nuanced knowledge of reality beyond what is specified in the training data set. In that case, they will be more valuable and eventually bring AI closer to human-level intelligence. We think that self-supervised learning is one of the most assuring means to develop such background knowledge and approximate common sense in AI systems. Self-supervised learning allows AI systems to learn from orders of magnitude more data, which is vital to identify and interpret patterns of more subtle, less common representations of the world.
\subsection{Contributions}
\begin{itemize}
\item A self-supervised training strategy that aims to infer a distance map from a sequence of distorted and unrectified raw fisheye images.
\item A solution to the scale factor uncertainty with the bolster from ego-motion velocity allows outputting metric distance maps. This facilitates the map's practical use for self-driving cars.
\item A novel combination of super-resolution networks and deformable convolution layers~\cite{zhu2019deformable} to output high-resolution distance maps with sharp boundaries from a low-resolution input. Inspired by the super-resolution approach~\cite{shi2016real}, we accurately resolve distances by replacing the deconvolution~\cite{odena2016deconvolution} and a naive nearest neighbor or bilinear upsampling. (see Figure~\ref{fig:overview}: the self-supervised model, \textit{FisheyeDistanceNet}, produces sharp, high quality distance and depth maps).
\item We depict the importance of using backward sequences for training and construct a loss for these sequences. Moreover, a combination of filtering static pixels and an ego mask is employed. The incorporated bundle-adjustment framework~\cite{zhou2018unsupervised} jointly optimizes distances and camera poses within a sequence by increasing the baseline and providing additional consistency constraints.
\item A novel generic end-to-end self-supervised training pipeline to estimate monocular depth maps on raw distorted images for various camera models.
\item Empirical evaluation of the approach on two diverse automotive datasets, namely KITTI and WoodScape.
\item First demonstration of depth estimation results directly on unrectified KITTI sequences (see Figure~\ref{fig:unrectdepthnet-overview}: the scale-aware model, \textit{UnRectDepthNet}, yields precise boundaries and fine-grained depth maps).
\item State-of-the-art results on KITTI depth estimation among self-supervised methods.
\end{itemize}
\subsection{Contextual Depth Limitations}

\begin{figure}[!t]
  \centering
  \begin{minipage}[t]{0.495\textwidth}
    \centering
    \includegraphics[width=\textwidth]{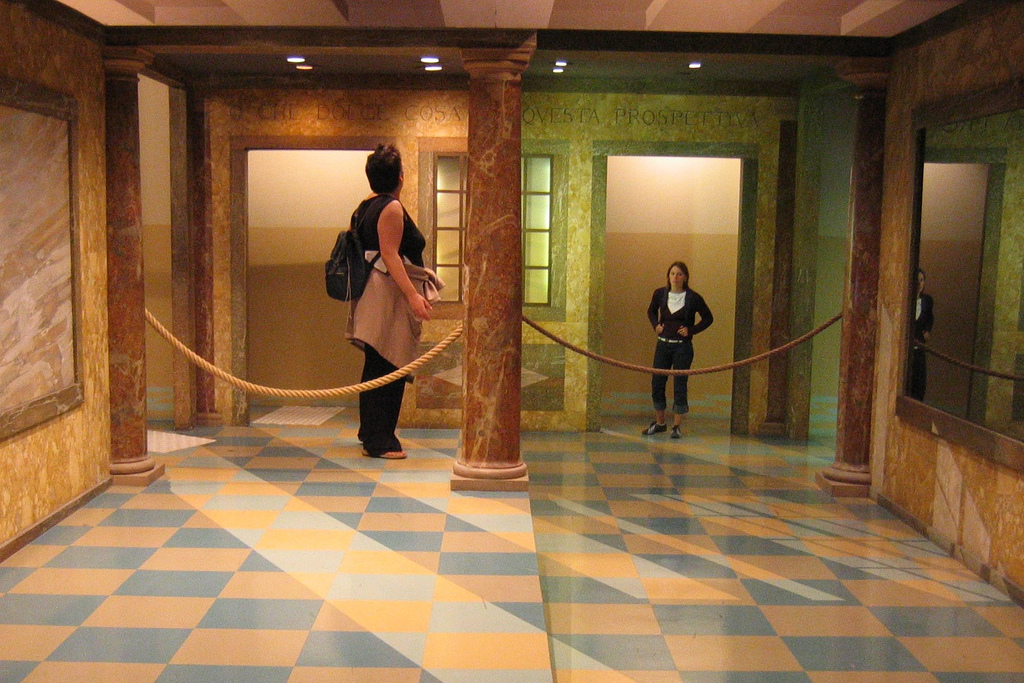}
    \caption[\bf Ames room, a famous example of forced perspective, taken at Cité des Sciences, Paris.]{\textbf{Ames room}, a famous example of forced perspective, taken at Cité des Sciences, Paris~\cite{ames_room}.}
    \label{fig:ames_room}
  \end{minipage}%
  \hfill
  \begin{minipage}[t]{0.495\textwidth}
    \centering
    \includegraphics[width=\textwidth]{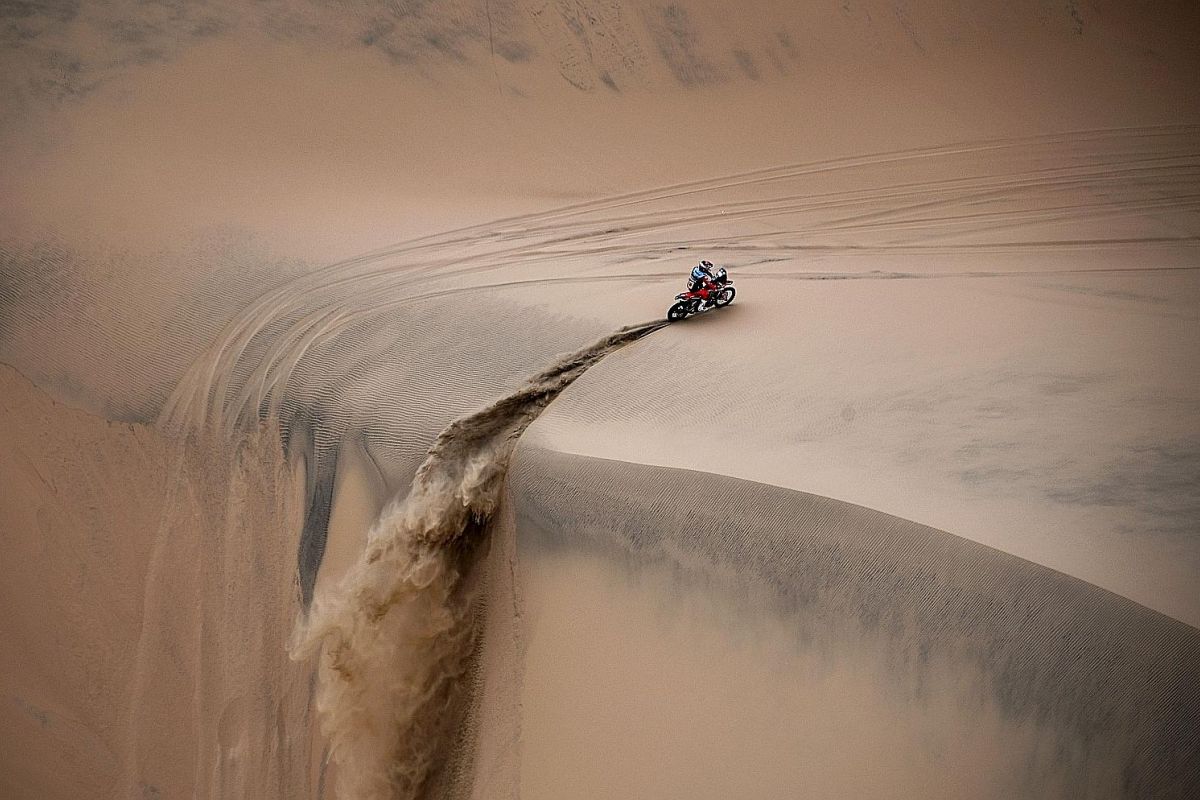}
    \caption[\bf Dakar 2019 photograph.]{\textbf{Dakar 2019} photograph by Frank Fife~\cite{dakar_2019}.}
    \label{fig:dakar_2019}
  \end{minipage}
\end{figure}

It has been demonstrated that depth inference networks can estimate depth solely based on perspective and context. They were able to obtain reasonable scale-invariant quality measures with only one image~\cite{Fu2018, saxena20083, Eigen_14, zhou2017unsupervised}. It has been deemed popular and convincing enough to prompt the development of a dedicated large-scale challenge~\cite{uhrig2017sparsity}. Without a way to link the estimation to the real world, scale-invariant quality is not particularly interesting in the context of autonomous driving. Using human perception as a reference, it was demonstrated by McManus~\etal~\cite{mcmanus2004light} that forcing perspective on humans was well-known and studied, especially in the art for dramatic effect. \emph{Ames room} is one of these methods. The apparent viewpoint is disputed by the person's height, as seen in Figure~\ref{fig:ames_room}; the depth is thus different on the left and right, although it appears to be the same. This demonstrates that the projection of a point P is independent of its distance from the camera and the human eye's lack of robustness in determining depth from context while viewing a single image. We can infer that a single frame depth network will face the same constraints~\cite{clement19}.\par
\begin{figure*}[t]
  \centering
    \includegraphics[width=\columnwidth]{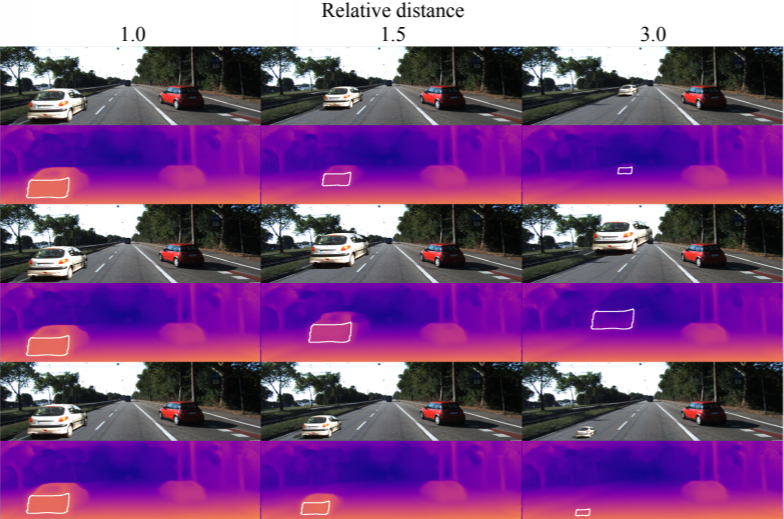}
    \caption[\bf Illustration of depth estimation's effect on change in position vs. apparent size on a test image from KITTI.]{\textbf{Illustration pf depth estimation's effect on change in position vs. apparent size on a test image from KITTI.} The white car on the left is inserted into the image at $1.0$ (left column), $1.5$ (middle column), and $3.0$ (right column), where $1.0$ corresponds to the same scale and position at which the car was cropped from its original image. The position and scale of the car vary with distance in the top row; in the middle row, only the position changes while the scale remains constant, and in the bottom row, the scale varies while the position remains constant. A white outline indicates the measurement region from which the estimated distance is obtained in the disparity maps. Figure and part of the caption reproduced from~\cite{dijk2019neural}.}
    \label{fig:depth-context-ablation}
\end{figure*}

One could argue that these forced perspective counter-examples are not really realistic: if they were discovered unintentionally, the human would have been resistant to it. Figure~\ref{fig:dakar_2019} depicts a Dakar 2019 competitor. At first glance, an illusory cliff can be seen before realizing the image is a high-angle shot (that the human eye is not familiar with). This demonstrates that even realistic imagery can lead to a perplexing perspective. CNN's can learn depth from single images by focusing their attention on specific features, similar to how humans see. Because this process is entirely data-driven (no geometry is involved in deployment), biases in the training data significantly impact the accuracy of a monocular network. Van Dijk and de Croon~\cite{dijk2019neural} investigate this by taking a set of cues (relative position, apparent size, and others) and examine how they affect the depth maps estimated by CNNs trained on KITTI~\cite{geiger2012we} as shown in Figure~\ref{fig:depth-context-ablation}. In summary, the position vs. apparent size is as follows:
\begin{itemize}[nosep]
    \item Varying position and size \textbf{affects} depth estimation (the higher and smaller, the farther) as seen in the 1\textsuperscript{st} row in Figure~\ref{fig:depth-context-ablation}.
    \item Varying position only \textbf{affects} depth as well (the higher, the farther) as seen in the 2\textsuperscript{nd} row.
    \item Varying size only \textbf{does not affect} estimated depth as seen in the 3\textsuperscript{rd} row.
\end{itemize}
These examples show that depth from context is not robust to new environments and that the depth error (even when scale-invariant) is not continuous with respect to appearance. It is also prone to changes in apparent camera mounting and object positions vs. apparent change in the size of the objects and if the image is cropped at a different aspect ratio for inference than the one used during training. Significant errors can even be found in the training set due to multi-stable perception~\cite{eagleman2001visual}. As a result, a training set must be comprehensive because a minor change in appearances, such as a change in lighting or orientation, can completely change the outcome.\par

Furthermore, a depth estimator trained on this hypothetically may be required to remember the most probable perspective layout in a given image, implying that heavy memory tasks may be required. We can assume that a neural network dedicated to this task will have to be large and thus complex to be embedded on a mobile system~\cite{clement19}.\par
\subsection{Shortcomings of Self-Supervised Distance Estimation}

\emph{SfM} pipelines have a varying degree of artifacts. Still, there exist several problems that need to be addressed. In this section, we will discuss some of the limitations.\par

In Chapter~\ref{Chapter4}, \textit{SfM} relies on the assumption of a static world. A real-world scene typically includes both \textit{static objects} such as walls, structures, and buildings, as well as \textit{moving objects} such as persons and vehicles. We are assuming a static world hypothesis employing \textit{SfM}, and it can only fulfill the static part.\par

What exactly does the static world assumption mean for the \emph{SfM} framework? \\
Camera motion and depth structure are solely responsible for the movement of static objects in a frame. The depth structure and camera motion will thoroughly evaluate the projected 2D image motion between frames. \\
On the other hand, dynamic objects possess the characteristics of large displacement (\eg, optical flow). They are caused by camera motion as well as actual object motion. This is not modeled in \emph{SfM} frameworks.\par
\subsection{Implications of the Used Loss Functions}

\begin{figure*}[t]
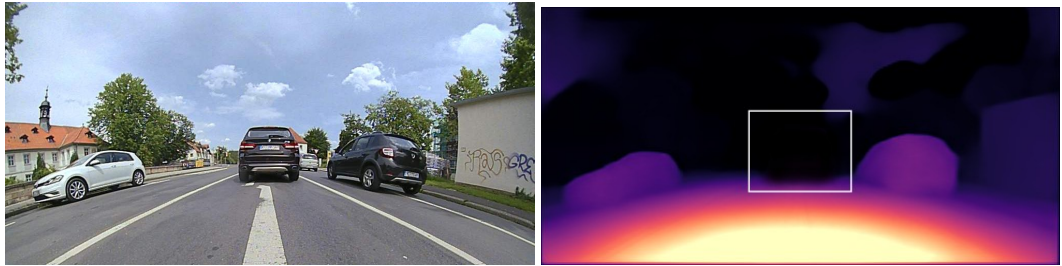

  \centering
    \includegraphics[width=0.49\columnwidth]{Figures/syndistnet/fisheye_fig/infinite_depth/20190614_143708_506_crop.jpg}
    \includegraphics[width=0.48\columnwidth]{Figures/syndistnet/fisheye_fig/infinite_depth/20190614_143708_506_norm.jpg} \\
    \caption[\bf Depiction of infinite distance due to dynamic objects on FisheyeDistanceNet.]{\textbf{Depiction of infinite distance due to dynamic objects} on FisheyeDistanceNet inducing holes during inference.}
    \label{fig:infinite_depth}
\end{figure*}

Aside from violating the static world assumption, the system introduced a slew of new considerations.
\begin{itemize}
    \item \textbf{Omission of object motion:} The object motion is ignored; when projecting the target pixels into the other camera view using the estimated pose, the reconstructed scene ignores motion in the source scene and is solely dependent on the estimated ego-motion. As a result, there is a mismatch between the target and predicted frames for pixels with motion. This means that even though the model correctly predicts the depth, it will still be penalized.
    \item \textbf{Depth efficiency is inextricably linked to the pose network performance:} Since the pose network is in charge of calculating the relative change in view, the performance of depth will suffer if the pose network underperforms.
    \item \textbf{The inevitable issue of predicting 'infinite depth' or holes:} It is mainly due to dynamic objects and a very particular case in which the object moves at the same speed as the camera (see Figure~\ref{fig:infinite_depth}). The object will appear to be stationary relative to the camera. It can also occur when things are infinitely far away. Any movement of the object is deemed unobservable and appears stationary. Since the object is not moving, it can cause the depth network to predict an infinite depth, manifesting as holes.\\
    To reduce the impact of stationary pixels on the loss, we incorporated an auto masking strategy that ignores these pixels when calculating the loss between the source and target frames~\cite{godard2019digging}. To further eliminate its impact on the loss, we tackled this issue in Chapter~\ref{Chapter5} with the help of semantic segmentation by employing the MTL approach.
\end{itemize}
\begin{figure}[!t]
  \centering
    \includegraphics[width=\columnwidth]{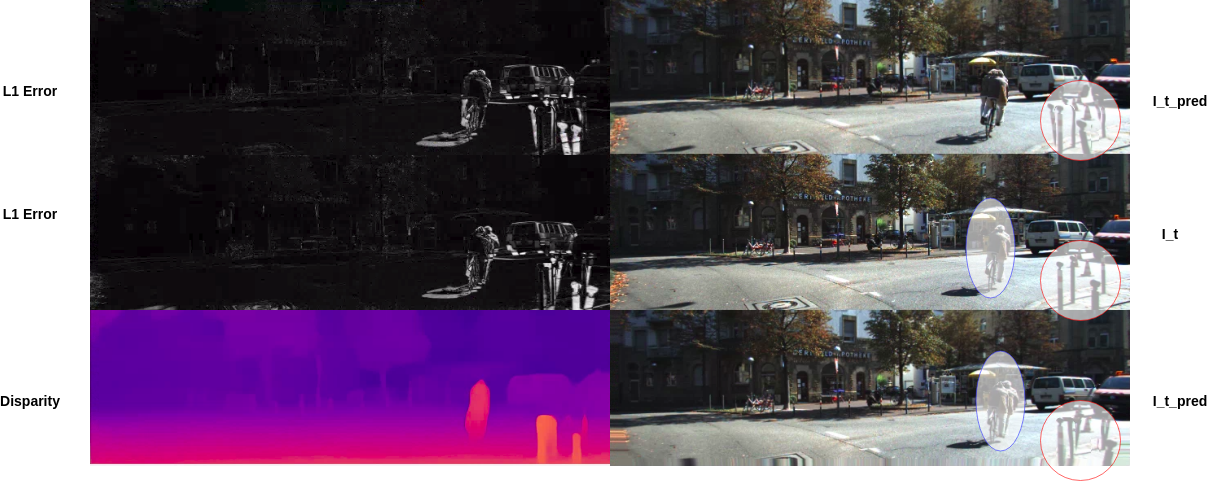}
    \caption[\bf Illustration of the \textit{SfM} framework's limitations.]{\textbf{Illustration of \textit{SfM} framework's limitations.} It is observed that the model struggles near boundaries and thin elongated structures, as illustrated by the pixel-wise absolute difference between prediction and target image. A higher intensity in the \lone error map indicates a higher absolute difference. Figure reproduced from~\cite{depth_discussion_2021}.}
    \label{fig:SfM_limitation}
\end{figure}
\subsection{Are there Better Choices than the Photometric Loss?}

Appearance-based loss has been a reliable alternative to depth loss. However, there are several drawbacks to using image intensity as a performance indicator (see Figure~\ref{fig:SfM_limitation}).\par
\begin{figure*}[!t]
  \centering
  \resizebox{\textwidth}{!}{\newcommand{\turnheightnew}{0.25\columnwidth}
\centering

\begin{tabular}{@{\hskip 0.5mm}c@{\hskip 0.5mm}c@{\hskip 0.5mm}c@{\hskip 0.5mm}c@{\hskip 0.5mm}c@{}}

{\rotatebox{90}{\hspace{0mm}}} &
\includegraphics[height=\turnheightnew]{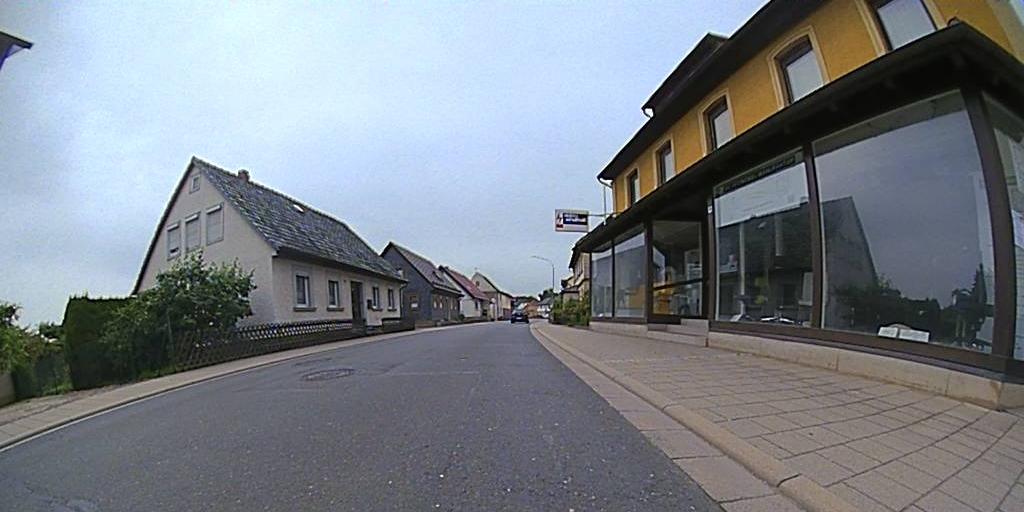} &
\includegraphics[height=\turnheightnew]{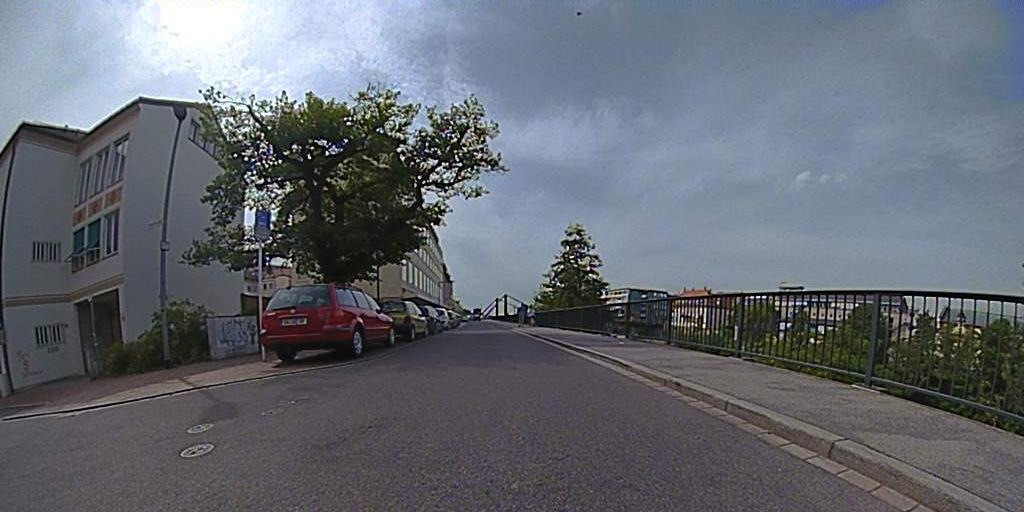} &
\includegraphics[height=\turnheightnew]{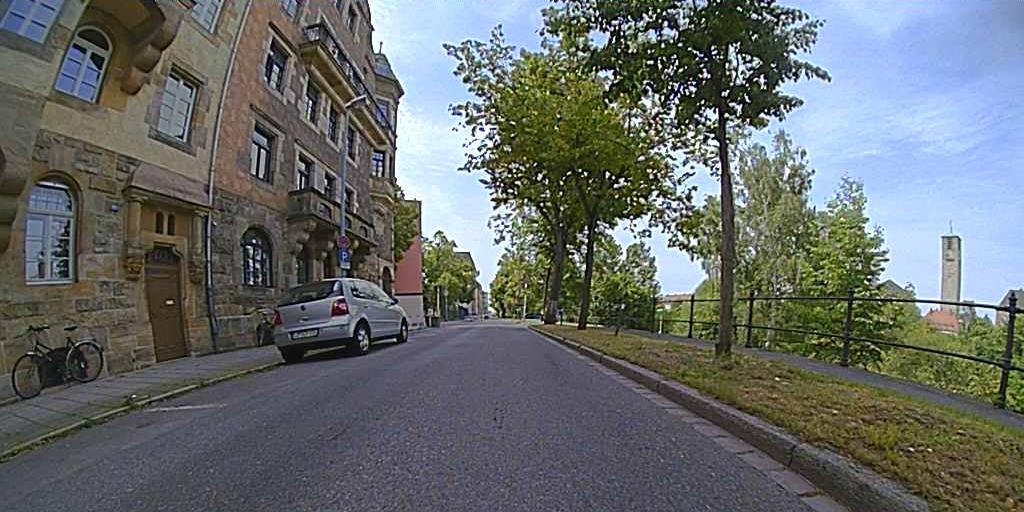} &
\includegraphics[height=\turnheightnew]{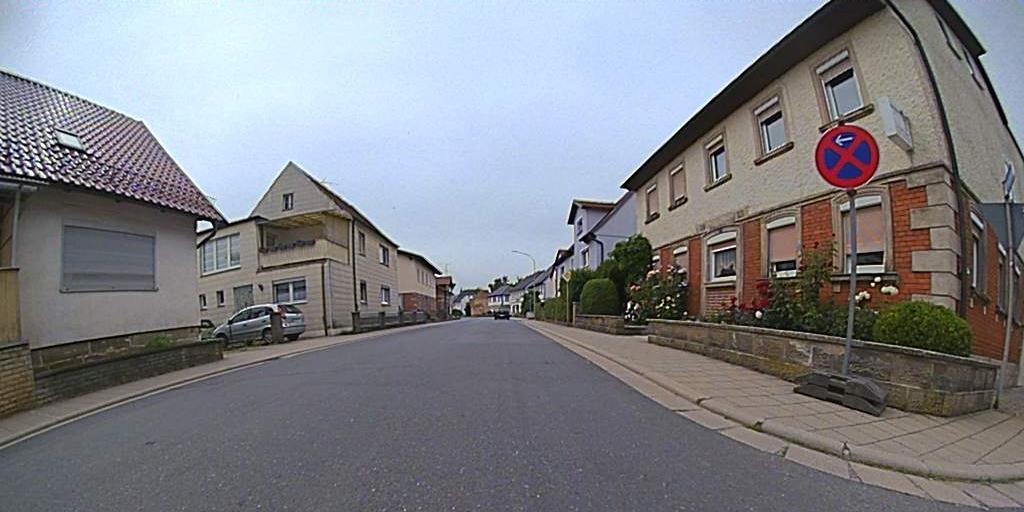} \\

{\rotatebox{90}{\hspace{0mm}\scriptsize}} &
\includegraphics[height=\turnheightnew]{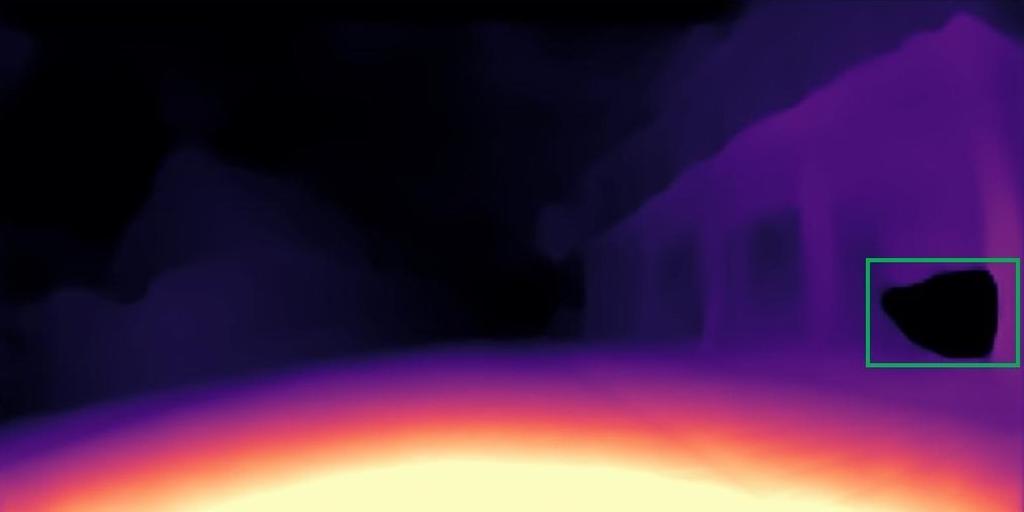} &
\includegraphics[height=\turnheightnew]{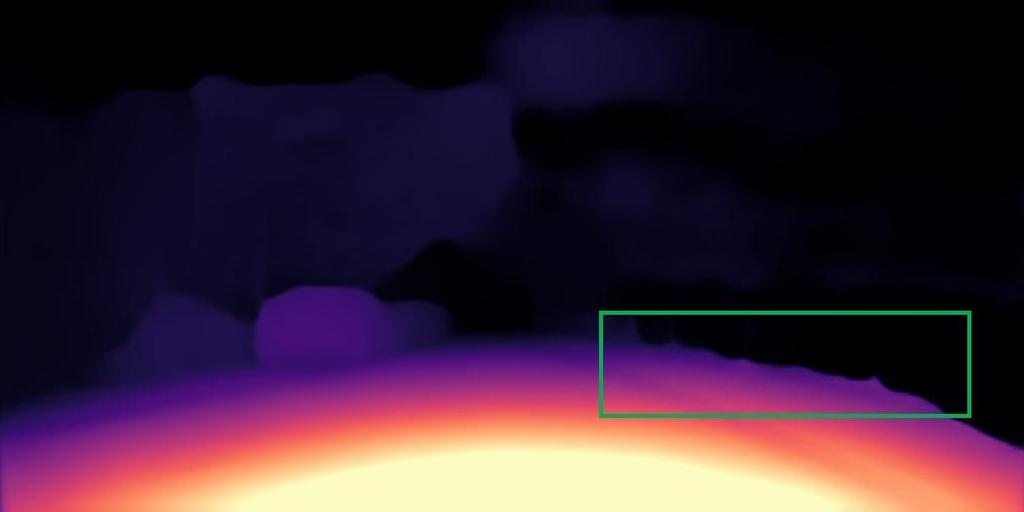} &
\includegraphics[height=\turnheightnew]{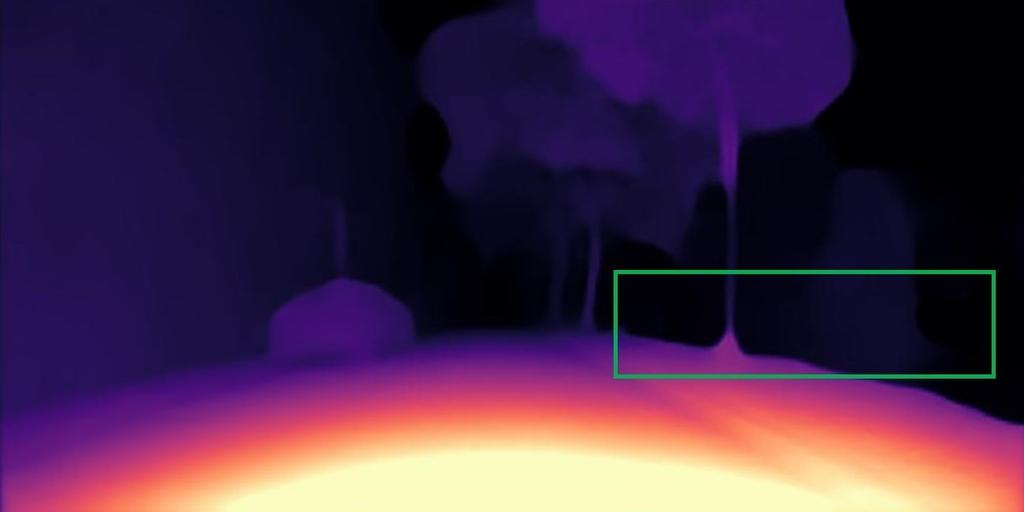} &
\includegraphics[height=\turnheightnew]{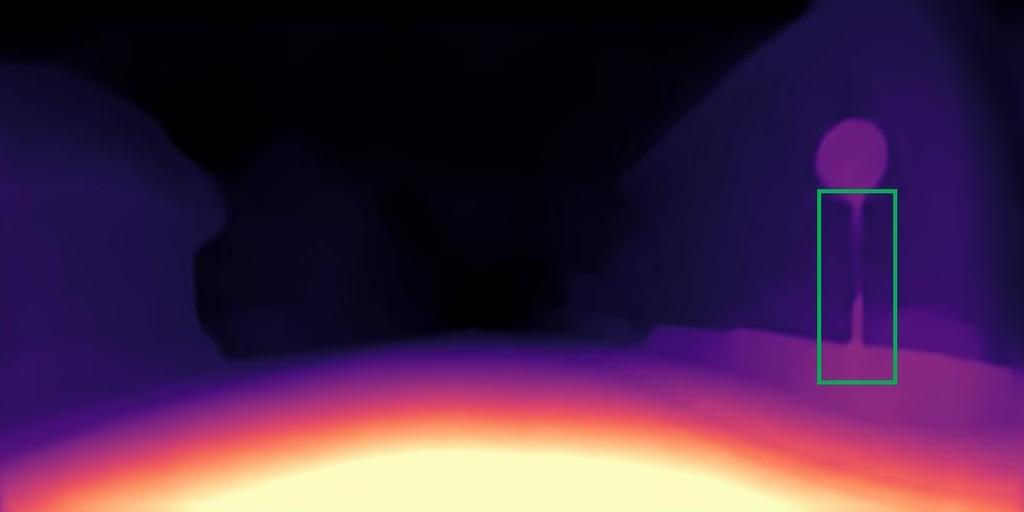} \\

\end{tabular}
}
   \caption[\bf Failure cases of FisheyeDistanceNet on the WoodScape.]
   {\textbf{Failure cases of FisheyeDistanceNet on the WoodScape.} The photometric loss fails to learn good distances for reflective regions which can be seen in the 1\textsuperscript{st} figure. In the following figures shown above, the model fails to accurately delineate objects where boundaries are ambiguous.}
  \label{fig:fisheye_failure}
\end{figure*}
\begin{figure}[!t]
  \centering
  \begin{minipage}[t]{0.495\textwidth}
    \centering
    \includegraphics[width=\textwidth]{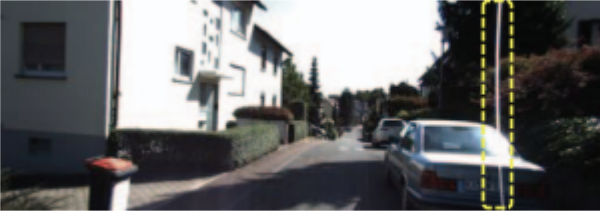}
    \caption[\bf An illustration of random lens error on the KITTI.]{\textbf{An illustration of random lens error on the KITTI.} Figure reproduced from~\cite{depth_discussion_2021}.}
    \label{fig:random_lens_error}
  \end{minipage}%
  \hfill
  \begin{minipage}[t]{0.495\textwidth}
    \centering
    \includegraphics[width=\textwidth]{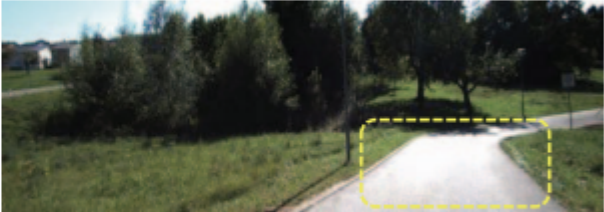}
    \caption[\bf An illustration of reflective surface on the KITTI.]{\textbf{An illustration of reflective surface on the KITTI.} Figure reproduced from~\cite{depth_discussion_2021}.}
    \label{fig:reflective_surface}
  \end{minipage}
\end{figure}

\begin{itemize}
    \item We cannot tell how big the world is from a 2D RGBD image. We lose scale information after the bilinear sampling stage since we compute the loss in 2D image space. In other words, several reconstructed images from the source view will result in the same correct target image. As a result, in 3D metric space, the learned depth will be scale uncertain but consistent in image space. To tackle this issue, we introduced scale awareness into the SfM framework~\ref{sec: scale-aware sfm} and computed direct depth/distance. In terms of KITTI Benchmark~\cite{geiger2012we} results, several works~\cite{zhou2017unsupervised, godard2019digging, mahjourian2018unsupervised, luo2019every} depend on determining the median depth based on the ground truth to overcome the scale ambiguity.
    \item Another risk of scale ambiguity is the probability of depth collapsing. Wang~\etal~\cite{Wang_2018_CVPR} demonstrated that as scale decreases, the same scene could be built with a smaller depth map until it degenerates to zero. Based on these findings, we introduced a normalization stage to prevent the mean depth from collapsing in Section~\ref{sec:edge-smoothness}.
    \item Even if the reconstructed image is not geometrically consistent, the loss can be satisfied. When depth is incorrectly predicted but pose is correct, or vice versa, the reconstructed image can still be well predicted in certain scenes. It is more common in texture-less regions such as the wall and the sky. The depth and pose networks are not penalized or compensated for incorrect predictions in this scenario.
    \item The value of the loss may be significant for differing views, such as when pixels are visible in the target frame but occluded or out of view in the source frames. There would be a mismatch in these regions when the points are re-projected. As a result, even though the depth is correctly estimated, the model will still be penalized.
    \item Reflective surfaces that violate the Lambertian assumption between views (see Figures~\ref{fig:fisheye_failure}, \ref{fig:random_lens_error} and~\ref{fig:reflective_surface}), and color-saturated images can cause mismatch and degrade the loss computation.
\end{itemize}
\begin{figure*}[t]
  \centering
    \includegraphics[width=0.48\columnwidth]{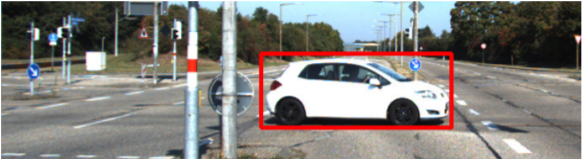}
    \includegraphics[width=0.48\columnwidth]{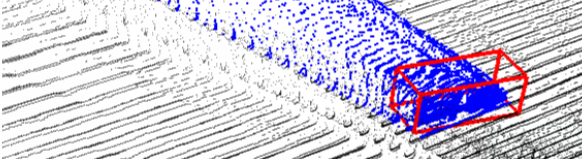} \\
    \caption[\bf Illustration of bleeding edge effect.]{\textbf{Illustration of bleeding edge effect. Figure reproduced from~\cite{weng2019monocular}.}}
    \label{fig:bleeding_edge_effect}
\end{figure*}
\subsection{Impact on Chosen Distance Estimation Network}

\begin{itemize}
    \item Instead of preserving edges, networks often predict smooth depths along occlusion boundaries and depth discontinuities. A probable cause could be the lack of edge awareness explicitly defined and smooth surfaces dominating the loss. As a result, the bleeding edge effect~\cite{weng2019monocular} occurs, and the back-projected points are poorly defined, as shown in Figure~\ref{fig:bleeding_edge_effect}.
    \item The distance map over consecutive frames, especially at the boundaries and thin structures in the video \url{https://youtu.be/Sgq1WzoOmXg?t=27}, causes a flickering effect since it is not geometrically consistent over time.
\end{itemize}
\section{Geometry Meets Semantics}

As heavily discussed in Chapter~\ref{Chapter4}, today's state-of-the-art regarding self-supervision for depth sensing mostly relies on heuristics, themselves based on validation results on the KITTI dataset. A clear understanding of the theoretical aspect of self-supervision is needed to perform a truly robust training workflow. To overcome the shortcomings of the SfM framework for distance estimation, we analyzed the issues of the photometric loss function and the impact caused by the dynamic objects on the distance predictions, as discussed in the previous section. In Chapter~\ref{Chapter5}, we dug deeper and took up the challenges posed by the SfM approach and the photometric losses.\par
\subsection{Contributions}
\begin{itemize}
    \item We introduce a novel architecture for the learning of self-supervised distance estimation synergized with semantic segmentation.
    \item We improve the self-supervised distance estimation by a general and robust loss function.
    \item We propose a solution for the dynamic object impact on self-supervised distance estimation by using semantic guidance. We show the approach's effectiveness on pinhole and fisheye camera datasets and present state-of-the-art results for both image types.
    \item We present a novel camera geometry adaptive multi-scale convolution to incorporate the camera parameters into the self-supervised distance estimation framework. We feed this camera geometry tensor (CGT) representation to the model as a generic way to adapt to new camera intrinsics.
    \item We create a training framework for self-supervised distance estimation, which jointly trains and infers images from multiple fisheye cameras and viewpoints.
    \item We demonstrate a single trained model for 23 fisheye cameras, which achieves the equivalent result as an individual specialized model that overfits a particular camera model.
    \item We present an improved version of the network architecture for multi-task learning of self-supervised distance estimation and semantic segmentation. We significantly improve upon the previous works explained in the previous chapters FisheyeDistanceNet~\cite{kumar2020fisheyedistancenet} and UnrectDepthNet~\cite{kumar2020unrectdepthnet}.
    \item We achieve state-of-the-art results on the WoodScape and KITTI datasets among monocular self-supervised depth estimation methods.
\end{itemize}
\subsection{Exploring Diverse Modalities}

\subsubsection{Rethinking the Heuristic Loss Function} 

The difference between images is determined as the absolute difference between RGB values. At first, we aimed to improve the photometric loss by replacing the \lone loss with a more robust loss function from~\cite{barron2019general} in the MTL framework of SynDistNet~\cite{kumar2020syndistnet}. We replaced the loss with various types of general loss functions and achieved improved efficiency. The absolute loss is the same as maximizing the probability of a Laplacian distribution on RGB pixel values with a fixed scale. We substitute the general distribution for the fixed Laplacian distribution, keeping our scale constant but allowing the shape parameter $\alpha$ to vary.\par

\textbf{Perspectives:} Finally, as indicated by the SSIM~\cite{wang2004image} loss, a photometric loss that takes neighboring pixels into account may be advantageous to obtain more information than just color to determine whether two points match or not. Brox \etal \cite{brox2010large} had already researched this and considered several features and descriptors to be matched in this work and concluded that including descriptor matching within the energy minimization process was beneficial, particularly for large displacement. The issue at the time was descriptors' lack of differentiability and sub-pixel accuracy. This problem could be easily adapted for photometric loss with convolution kernel descriptors, resulting in a feature matching loss as employed in Chapter~\ref{Chapter7}.\par

In general, self-supervision should rely more on techniques developed for optical flow using the variational approach because occlusion issues and efficient smoothing and illumination robustness are not novel~\cite{clement19}.\par
\subsubsection{Leveraging Semantic Segmentation} 

In terms of representing structure in images, surface normal, disparity, and optical flow is very close to depth. The direction of the depth gradient can be interpreted as the surface normal. Ego-motion and object motion can naturally relate optical flow to depth. Depth has an inverse relationship with disparity. Although semantics does not have any direct relation, many use segmentation maps to learn finer depth maps. There have been many studies that have explored these mutually beneficial properties~\cite{depth_discussion_2021}. Our MTL network employed pixel-adaptive convolutions~\cite{su2019kernel} to learn semantic-dependent representations that can better capture the aforementioned equivariance property compared to normal convolutions. We reduced the impact of dynamic objects on the photometric loss and inherently minimized the influence of 'infinite depth' or holes during inference. Our approach still lacked a complete solution \ie, if a dynamic object did not belong to the semantic class labels, this would still have a negative impact on the SfM framework during training.\par
\subsubsection{Model Advances} 

Recent work has shown that self-attention can assist as an essential building block for image recognition models. We adopted the self-attention modules from~\cite{zhao2020exploring} and assessed their effectiveness for estimating distance/depth. Initially, we used a scalar-based self-attention module~\cite{ramachandran2019stand}. Later, we employed two types of self-attention. The first is pairwise self-attention, which is essentially a set operator and generalizes standard dot-product attention. The other choice is patchwise self-attention, which is more efficient than the standard convolution. Pairwise self-attention networks match or outperform their convolutional counterparts, and patchwise models outperform the convolutional baselines significantly. We investigate the robustness of learned representations and conclude that when used in conjunction with our Camera Geometry Tensor, self-attention networks can provide significant benefits in terms of model robustness and depth generalization.\par
\begin{table}[!t]
\centering
\setlength{\tabcolsep}{2.5pt}
\begin{tabular}{l|c|c}
\toprule
\textbf{Operation} & 
\multicolumn{1}{c}{\textit{\begin{tabular}[c]{@{}c@{}}
\cellcolor[HTML]{7d9ebf} Content \\ \cellcolor[HTML]{7d9ebf} Adaptive\end{tabular}}} & \multicolumn{1}{c}{\textit{\begin{tabular}[c]{@{}c@{}}
\cellcolor[HTML]{e8715b} Channel \\ \cellcolor[HTML]{e8715b} Adaptive\end{tabular}}} \\
\midrule
Convolution~\cite{lecun1989backpropagation} & \xm & \ch \\
Scalar attention~\cite{vaswani2017attention, wang2018non, ramachandran2019stand, hu2019local} 
& \ch & \xm \\
Vector attention~\cite{zhao2020exploring} & \ch & \ch \\
\bottomrule
\end{tabular}
\caption[\bf Comparison of convolution vs self-attention.]{\textbf{Comparison of convolution vs self-attention.} The content of the image is not adapted by the convolution. Scalar attention yields scalar weights that are constant along the channel dimension. Zhao~\etal's~\cite{zhao2020exploring} self-attention modules compute attention weights that adapt across spatial dimensions and channels in an efficient manner.}
\label{tab:properties}
\end{table}
\subsubsection{Comparison between Convolution and Self-Attention} 

For the convolution operator, the fixed kernel weights are independent of the image's content. It does not adapt to the input content and can vary across channels, as shown in Table~\ref{tab:properties}. Scalar attention when compared to convolution, the aggregated weights can vary across different locations depending on the image's content. The main drawback in the formulation is that it does not adapt the attention weights at the different channels. We can alleviate this to a certain extent by introducing multiple heads~\cite{vaswani2017attention}; the number of heads is a small constant, and all channels within a head share scalar weights. Zhao~\etal's~\cite {zhao2020exploring} pairwise and patchwise modules can produce \emph{vector} output. The vector can be processed and mapped to the appropriate dimensionality, which can also accept input from position encoding channels. Convolution is generalized by the patchwise family of operators while preserving parameter and FLOP efficiency.\par
\section{Generalized Object Detection}

\subsection{Need for better 2D Object Representations} 

As discussed in Chapter~\ref{Chapter6}, surround-view coverage is critical for low-speed maneuvering autonomous driving applications such as automated parking~\cite{horgan2015vision, heimberger2017computer}. Four surround-view fisheye cameras are typically part of this sensor suite, enabling a dense $360\degree$ near field perception. The wide field of view of the fisheye image comes with the side effect of strong radial distortion. A common practice is to rectify distortions in the image using a 4\textsuperscript{th} order polynomial model or a unified camera model~\cite{barreto2006unifying}. However, undistortion comes with re-sampling distortion artifacts, especially at the periphery, a reduced FoV, and a non-rectangular image due to invalid pixels. Thus, we aimed to perform object detection on distorted fisheye images. Although semantic segmentation is an easier solution on fisheye images, object detection annotation costs are much lower~\cite{siam2017deep}.\par
\subsection{Contributions}

Chapter~\ref{Chapter6} aims to present a more detailed study of various techniques for fisheye object detection in autonomous driving scenes. The main contributions include:
\begin{itemize}
    \item Exploration of seven different object representations for object detection on fisheye images.
    \item Design of novel representations for fisheye images, including the curved box and adaptive step polygon.
    \item Release of a dataset of 10,000 images with annotations for all the object representations.
    \item Implementation and empirical study of FisheyeYOLO baseline, which can output different representations.
\end{itemize}
\subsection{Lack of Fisheye Object Detection Dataset} 

Our core objective was to present a more detailed study of various techniques for fisheye object detection in autonomous driving scenes. However, one of the main issues we faced in this research was the lack of a public dataset, particularly for autonomous driving scenarios. Henceforth, we compiled a 2D object detection dataset on fisheye images and released 10,000 images. We hope this encourages further research in this area leading to a mature object detection on undistorted fisheye images.\par
\subsection{Limitations of Generalized Object Detection} 

Polygon representation is a better way to represent objects, but it does not improve the CNN ability to detect objects on fisheye. Since we use the YOLOv3 decoder, an area of improvement could be its average precision for medium and large objects. Compared to YOLOv2, mean average precision increased, and localization errors decreased. When comparing YOLO to RetinaNet~\cite{lin2017focal}, the YOLOv3's average precision does show a trade-off between speed and accuracy. By having a larger dataset, the accuracy of detecting objects with YOLOv3 can be made equal to the accuracy of RetinaNet, making it an excellent choice for models that can be trained with large datasets. YOLOv3 may not be suitable for use with niche models where large datasets are difficult to access~\cite{yolov3_2021}.
\section{Holistic 360° Scene Understanding}

\subsection{Contributions}

This chapter demonstrates a multi-task perception model for the six essential perception tasks on unrectified fisheye images (shown in Figure~\ref{fig:abstract}). We discuss a full perception system building upon the tasks explained in the previous chapters, including depth estimation, pose estimation, semantic segmentation, and object detection. The contributions are as follows:
\begin{itemize}
    \item We demonstrate the first real-time six-task model for surround-view fisheye camera perception.
    \item We propose novel design techniques, including the VarNorm task weighting.
    \item We design synergized decoders where various tasks help each other in addition to a shared encoder.
    \item We showcase a 6-task model on WoodScape and a 5-task model on KITTI and Cityscapes performing better than the single task baselines.
    \item We obtain state-of-the-art results for depth and pose estimation tasks on KITTI among monocular methods.
\end{itemize}
In the following sections, we briefly discuss some significant aspects of MTL as to why and when it works, when it does not, and what are the vital things to note when training an MTL framework.\par

\textbf{Discussion on MTL:} Though MTL is becoming more common, the 20-year-old hard parameter sharing paradigm remains prevalent in CNN-based MTL. Recent advancements in learning what to share, on the other hand, are encouraging. At the same time, our understanding of tasks – their similarity, relationship, hierarchy, and benefit for MTL – is minimal. We need to learn more about them better to understand MTL's generalization capabilities in DNNs.\par

\subsection{Why Does MTL Work?}

Even though an inductive bias acquired through multi-task learning seems intuitively possible, in order to better understand MTL, we must look at the processes that shape it. For all examples, we will assume that we have two related tasks, $A$ and $B$, which rely on a common hidden layer representation $F$.
\begin{itemize}
    \item \textbf{Implicit data augmentation:} MTL effectively improves the sample size for which we are training our model. Since all tasks are at least somewhat noisy, our goal when training a model on some task $A$ is to learn a good representation for task $A$ that ignores the data-dependent noise and generalizes well. Since different tasks produce different noise patterns, a model that learns two tasks simultaneously may learn a more general representation. Learning only task $A$ risks overfitting to task $A$, while learning $A$ and $B$ together allows the model to achieve a better representation $F$ by averaging the noise patterns~\cite{mtl_basics_2021}.
    \item \textbf{Attention focusing:} It may be difficult for a model to distinguish between essential and irrelevant features when the task is very noisy or the data is limited and high-dimensional. MTL will assist the model in focusing its attention on the significant features, as other tasks will provide additional evidence for the importance or irrelevance of those features~\cite{mtl_basics_2021}.
    \item \textbf{Eavesdropping:} The certain features $G$ are simple to learn for one task $B$ but challenging to learn for another. This could be due to $A$ interacting with the features in a more nuanced way, or it could be due to other features impeding the model's ability to learn $G$. We can allow the model to eavesdrop through MTL, allowing it to learn $G$ via task $B$. The simplest method is to use hints~\cite{abu1990learning}, which involves explicitly training the model to predict the most relevant features~\cite{mtl_basics_2021}.
    \item \textbf{Regularization:} Finally, by adding an inductive bias, MTL serves as a regularizer. As a result, it reduces the chance of overfitting and the model's Rademacher complexity \ie, its ability to fit random noise~\cite{mtl_basics_2021}.
\end{itemize}
\subsection{When MTL Works -- And When It Does Not}

Multi-task learning aims to use data from related tasks to improve the generalization performance of all tasks simultaneously. Shared hidden layers can transfer knowledge between related tasks in a neural network, reducing overfitting and improving learned latent representations, especially when task-specific training data is scarce. On the other hand, MTL can be detrimental to performance when the tasks being considered are not sufficiently related. Although multi-task performance may improve on average across all tasks, multi-task performance for some specific tasks may be worse than a single-task model. 
This drop in performance is referred to as \textbf{negative transfer}. Negative transfer is especially problematic when a subset of tasks is of primary interest, and the others are only used to improve representation learning. Liu~\etal~\cite{liu2019loss} proposes two hypotheses for why negative transfer might occur. (1) Because all tasks are diverse and unrelated to one another, and there is no suitable common latent representation, multi-task learning produces poor representations. (2) The training process is dominated by a single group of related tasks. The performance of those tasks improves as more related tasks are added, but tasks outside the dominant group suffer~\cite{liu2019loss}.\par

\begin{figure}[!t]
  \centering
    \includegraphics[width=0.8\textwidth]{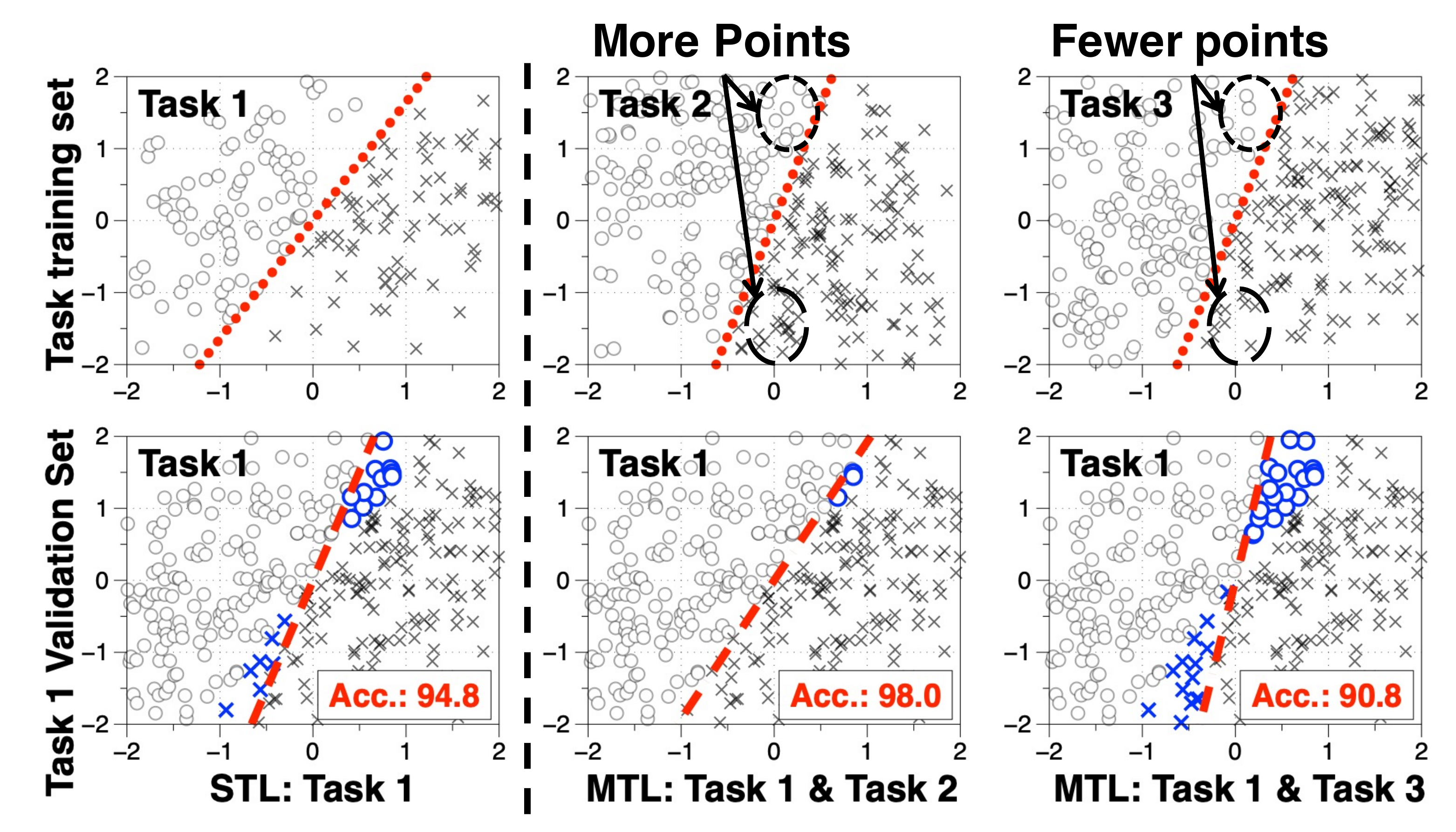}
    \caption[\bf Positive vs. Negative transfer is affected by the data.]{\textbf{Positive vs. Negative transfer is affected by the data -- not just the model}. See lower right-vs-mid. Task 2 and 3 have the same model (dotted lines) but different data distributions. Notice the difference of data in circled areas. Figure and caption reproduced from~\cite{wu2020understanding}.}
    \label{fig:mtl_model_impact}
\end{figure}

Wu~\etal~\cite{wu2020understanding} showed that MTL, when applied to heterogeneous task data, can frequently produce suboptimal models. They investigate an architecture with a shared module for all tasks (\ie, encoder) and a separate output module for each task (\ie, decoder) to determine whether two tasks interfere constructively or destructively. The motivating observation is that, in addition to model similarity, which impacts the type of interference, task data similarity has a second-order effect after controlling model similarity. They consider three tasks with the same number of data samples, where tasks 2 and 3 have the same decision boundary but different data distributions (see Figure~\ref{fig:mtl_model_impact} for an illustration). They discover that training task 1 with task 2 or task 3 can either improve or degrade task 1's performance, depending on the amount of data contributing along the decision boundary! This finding demonstrates that by analyzing task interference and attributing the cause more precisely, we can measure the similarities of task data and models separately.\par
\subsection{Key Components to Determine if MTL is better than STL}

The three significant factors that can result in negative transfer and help us to determine if MTL is better than Single Task Learning (STL) are:

\textbf{Model capacity:} The shared module's capacity, \ie, its output dimension, is critical because if the shared module is too large, there can be no interference (or transfer of knowledge) between tasks because each of them can be memorized in the shared module, resulting in zero training loss. There may be destructive interference if it is too small. As a general rule, according to~\cite{wu2020understanding} the shared module performs best when its capacity is less than the total capacity of the single-task models.\par

\textbf{Task covariance:} To determine the interference between different tasks, Wu~\etal~\cite{wu2020understanding} uses a fine-grained concept called task covariance to measure how similar two tasks are. Task covariance quantifies the alignment of two task input data points along their primary axes. According to intuition, if the principal directions of two task input data are not well-aligned, feeding them into the shared module can result in suboptimal models. Wu~\etal~\cite{wu2020understanding} proposed a covariance alignment algorithm that adds an alignment module between the task and the shared module to improve multi-task training to address this issue.\par

\textbf{Optimization scheme:} The order in which we optimize a multi-task learning neural network can also affect the interference between tasks. A common training strategy is to mix mini-batches of different task data at random. We can increase the task weight for an important task by duplicating its data. For further discussions on the optimization strategies employed in MTL refer Section~\ref{sec:optimization-in-mtl}.\par
\chapter{Conclusion}
\label{Chapter9}

In this thesis, we explored geometric and segmentation tasks for a holistic real-time scene understanding of the environment \textit{using cameras only}. The tremendous improvement in computer vision technology using deep learning-based models is the core reason why it is possible to drive on the road with camera-only perception. We build a near-field perception system that constitutes a \emph{Level $3$} autonomous stack. We develop a single real-time multi-task CNN capable of running on embedded NVIDIA's Jetson AGX platform, covering the necessary modules for near-field sensing use cases such as parking or traffic jam assistance. The model's scope is not just limited to near-field sensing use cases. We can deploy it for navigation on highways and urban scenarios. The final model encodes the input raw fisheye stream into a single, high-dimensional tensor representing geometry, semantics, motion, and object detection. This tensor can be used to make driving decisions for an autonomous car.\par

In \textbf{Chapter}~\ref{Chapter2} and \textbf{Chapter}~\ref{Chapter3} we provided an in-detail basic intuition on the fisheye camera models and geometry, perception tasks and related works. We reasoned why depth estimation is a challenging task compared to the other semantic tasks and provided the reader the fundamental background in vision. In \textbf{Chapter}~\ref{Chapter4}, we initially focused on solving one of the most challenging geometric problems \ie, distance estimation on raw fisheye cameras using image-based reconstruction techniques, which is a challenging task, as the mapping between 2D images to 3D surfaces is an under-constrained problem. Depth/distance estimation is also an ill-posed problem as there could exist many possible incorrect depths per pixel, which can also recreate the novel view. We also trained an additional model to predict the relative rigid transformation between the video frames to achieve the visual odometry required to place the detected objects in a temporally consistent map.\par

We took up a more challenging problem of obtaining distance on fisheye cameras which undergoes large distortion compared to pinhole cameras. Most of the previous works worked on rectified pinhole cameras and obtained inverse-depth estimates in the \emph{SfM} framework. The legacy continued to rely on the principle of rectification as the fundamental first step to get the view synthesis working in the first place. We broke this premise and showed that it is possible to obtain scale-aware direct distance estimates without rectifying fisheye and pinhole cameras on two diverse automotive datasets.\par

Following up on \textit{FisheyeDistanceNet} and \textit{UnRectDepthNet} we focused on improving the distance estimation further in \textbf{Chapter}~\ref{Chapter5} by using semantic segmentation in a multi-task learning set up. Our network in the \textit{Syndistnet} MTL framework learned semantic-aware geometric representations that could disambiguate photometric ambiguities in a self-supervised learning \emph{SfM} context. We integrated a generalized robust loss function, which improved performance significantly while removing the need for hyperparameter tuning with the reprojection loss. Finally, we reduced the artifacts caused by dynamic objects, violating static world assumptions, by using a semantic masking strategy. We significantly improved upon the RMSE of previous works on fisheye images in \textbf{Chapter}~\ref{Chapter4} by a 25\% reduction in RMSE. As there is little work on fisheye cameras, we also evaluated the proposed method on KITTI using a pinhole model. We achieved state-of-the-art performance among self-supervised methods without requiring an external scale estimation.\par

We later extended this MTL framework from \textit{Syndistnet} to multiple cameras and viewpoints. A $360^\circ$ perception of scene geometry is essential for automated driving, notably for parking and urban driving scenarios. Typically, it is achieved using surround-view fisheye cameras, focusing on the near-field area around the vehicle. Most of the current depth estimation approaches focused on employing just a single camera, which cannot be straightforwardly generalized to multiple cameras. Besides, the depth estimation model needs to be deployed across different-sized car lines with varying camera geometries. Even in a single-car line, there are variations in intrinsics due to manufacturing tolerances. Deep learning models are sensitive to these changes, and it is practically infeasible to train and test on each camera variant. Thus, we introduced novel camera-geometry adaptive multi-scale convolutions, which utilize the camera parameters as a conditional input, enabling the model to generalize to unseen fisheye cameras. We incorporated this into our previous work \textit{Syndistnet}.\par

Additionally, we improved the distance estimation by pairwise and patchwise vector-based self-attention encoder networks, further increasing the distance estimation's performance. We evaluated our approach on the Fisheye WoodScape surround-view dataset, significantly improving over previous approaches, \ie, \textit{FisheyeDistanceNet}, \textit{UnRectDepthNet}, and \textit{SynDistNet}. We also showed a generalization of our approach across different camera viewing angles and performed extensive experiments to support our contributions. To enable comparison with other approaches, we evaluated the front camera data on the KITTI dataset (pinhole camera images) and achieved state-of-the-art performance among self-supervised monocular methods. The following work series carried out on distance estimation was very influential from a product perspective to win next-generation projects and was influential in the academic community. It encouraged the researchers in the robotics community to deploy fisheye cameras to obtain point clouds from depth maps for SLAM by eliminating the need for rectification on the camera streams.\par

In \textbf{Chapter}~\ref{Chapter6}, we focused on the localization aspect of an autonomous car using 2D object detection. The standard bounding box fails in fisheye cameras due to the strong radial distortion, particularly in the image's periphery. We explored better representations such as oriented bounding box, ellipse, and generic polygon representations for object detection on fisheye images. One of the primary use cases is to find a free parking slot for the autonomous car with these representations on fisheye cameras.\par

In \textbf{Chapter}~\ref{Chapter7}, we achieved the goal of the thesis, \ie, a holistic real-time scene understanding for the near-field perception of the environment \textit{using cameras only} by creating \textit{OmniDet}: A surround-view camera-based multi-task visual perception network for autonomous driving. This thesis shows that a distinct approach can be considered to the immense challenge of autonomous driving: one that does not rely on infrastructures such as high-definition-maps or extremely costly sensor payloads yet can perform complex driving tasks using cameras. To achieve real-world driving efficiency, these learned vision representations are very crucial. With this framework's help, we can jointly understand and reason about geometry, semantics, motion, localization, and soiling from a single deep learning model at 60 frames/second on embedded systems. The accuracy, robustness, and performance of these models are much higher than before. These models were deployed in real-time on an autonomous vehicle using NVIDIA' Jetson AGX, and we were able to demonstrate these models performing robustly. We evaluated the network on five tasks on KITTI and Cityscapes, achieving competitive results. We set a benchmark of these tasks on the fisheye WoodScape dataset. We also applied various adversarial attacks to understand the vulnerabilities of the perception network. For each perception task, white and black box attacks were performed for targeted and un-targeted scenarios.\par

Finally, in Chapter~\ref{Chapter8}, we discussed the contributions and the shortcomings of the perception tasks solved in this thesis. We reasoned about the design choices and their impact employed in this work. For the task of depth estimation, we discussed the fundamental limitations of depth from context and structure. However, after conducting a thorough investigation into the potential drawbacks of depth from vision algorithms, we were able to identify several issues with depth networks that would need to be addressed before they could be used as a reliable source of information for parking scenarios for autonomous cars. We dug deep into the core of the \emph{SfM} approach and its implication on the loss. We reasoned about the apparent failures due to the heuristic design of the photometric loss function and its impact on the network. To overcome the shortcomings of the \emph{SfM} framework, we analyzed the issues put forth during our initial research and explored diverse modalities, including re-looking the heuristic loss function, leveraging semantic segmentation, model advances, and compared the standard convolution with self--attention modules. Coming to the task of 2D object detection, we discussed the need for better 2D representations on fisheye images and the limitations of our model. Finally, we discussed the critical aspects of an MTL approach to determine if there is any negative transfer.\par
\section{Future Work}

We have showcased state-of-the-art methods for distance and depth estimation tasks. There are many more avenues left unexplored, and we outline some of them here. The distance estimates are applied independently on each frame, indicating that adding temporal consistency~\cite{karsch2014depth} would likely improve results with the usage of recurrent~\cite{patil2020don} or a transformer model~\cite{li2020revisiting}. Additionally, assisting the model with depth hints~\cite{watson2019self} would improve the results and also improve the reconstruction loss function. It could also guide the network to learn better weights.
Hu \etal~\cite{hu2019visualization} claimed that only a portion of the image is relevant for a network to estimate depth, in analogy with human vision. Further research in this direction is required to understand how a CNN can infer depth from single images. For collision avoidance, it is intriguing to understand how confident the model is about a particular inference. Learning the uncertainty~\cite{poggi2020uncertainty} of the predicted distance maps would be of significant importance for practical applications such as autonomous driving. More generally, the distribution of possible values can be very beneficial in decision-making. Instance aware projection consistency~\cite{lee2021learning} can be introduced to the view synthesis process to improve the training.\par

Learning dense semantic representations in an unsupervised fashion is a significant problem in computer vision. As of now, most of the semantic and motion segmentation tasks are supervised. However, we can not rely on having a supervised label for every possible image type we encounter in the real world. One of the most prominent challenges for a real-world computer vision system is learning a representation from limited data that generalizes to novel situations. It is especially true for autonomous driving -- the variety of road scenes that exist is tremendously wide. There are many promising methods to learning robust representations. Self-supervision would be a better option as obtaining accurate, pixel-wise semantic and motion labels for every sample in a dataset is a labor-intensive process that costs significant amounts of money and time~\cite{bearman2016s}. Gansbeke \etal~\cite{van2021unsupervised} tries to perform unsupervised semantic segmentation by contrasting object mask proposals. Tosi \etal~\cite{tosi2020distilled} employs a self-supervised optical flow to segment the dynamic objects and obtain a motion mask.\par

For vision to become truly ubiquitous with intelligent robotic decision making~\cite{wayve_2021}, with all the perception algorithms using cameras, the key takeaway would be online learning. An intriguing viewpoint developed throughout this thesis is the concept of a robust and evolving geometric and semantic sensing system based on vision. In other words, such a system does not require any active sensors and can improve while in use. For example, Casser \etal \cite{casser2019depth} discussed this, where it was used to overfit each example of the test set (with a reset of the network between examples). This technique has the potential to be extremely useful in the context of stealth off-road autonomous vehicles. Because of their stealthiness, no active depth sensors can be used, and depth must be calculated using cameras.
Furthermore, off-road environments are typically more heterogeneous than on-road environments, and our robustness-oriented solution may be superior to a traditional \emph{SfM} analytical algorithm or an evolutive single frame depth solution. We find it engaging to examine whether and how it would be possible to self-adapt the \textit{OmniDet} framework online. A perception system would require determining a complex, plausibly high-dimensional state and needs to represent the geometry, semantics, motion, and localization. It would be intriguing to use this framework as a baseline and achieve end-to-end driving by combining the perception and decision-making algorithms.\par
To deploy the perception systems on an autonomous car would still involve several safety challenges. Development and validation of algorithms through mere brute force testing is not possible. We would need to quantify uncertainty, interpret saliency for decisions, understand intermediate representations, and reason with multiple sensors in real-time. Robustness is of higher priority than precision, \eg, we do not need to know other cars' position to the nearest millimeter to enable safe driving. We solely care if we can detect the car or not and its rough spatial layout within the scene. The same is true for the algorithms we build; we care about robustness rather than millimeter-level accurate geometry from a LiDAR laser scanner~\cite{wayve_2021}. It would also be of significant interest to improve the robustness by strengthening or equipping the MTL system to resist adversarial attacks~\cite{madry2017towards}. In the future, it would be necessary to conduct physical attacks on mature systems to assess the vulnerabilities of these networks and attack multiple tasks jointly. Adversarial attacks and defenses are still challenging tasks and an active area of research, especially for autonomous driving applications with multi-task deep networks.\par

\clearpage

\emph{To conclude, over the past few years, computer vision technology has evolved to a point where it operates robustly in the wild. We are backing that \textbf{camera-only perception} will drive the smart autonomous cars and robots of the future. We hope that this thesis encourages further research in building a unified perception model for autonomous driving.}\par
\bibliographystyle{Bibliography/IEEETran}
\bibliography{Bibliography/references}
\end{document}